\documentclass[11pt,a4paper,twoside,openright]{report}

\usepackage[utf8]{inputenc}
\usepackage{pdflscape}
\usepackage{tabulary}
\usepackage{longtable}
\usepackage{multirow}
\usepackage{mathtools}
\usepackage{array}
\usepackage{scrextend}
\usepackage[english]{babel}
\usepackage{verbatim}
\usepackage{todo}
\usepackage{afterpage}
\usepackage{lscape}
\usepackage{blindtext}
\usepackage{booktabs}
\usepackage{caption}
\usepackage{subcaption}
\usepackage{textcomp}
\usepackage{etaremune}
\usepackage{makecell}
\usepackage{pdfpages}

\usepackage[pdeec,onpaper]{feupteses}

\usepackage[sort&compress]{natbib}
\setcitestyle{numbers,open={[},close={]}}
\usepackage[mathscr]{euscript}
\usepackage{amsmath}
\usepackage{multirow}
\usepackage{tikz}
\usepackage{tcolorbox}
\usepackage{enumitem}
\usepackage{soul}
\usepackage{times}
\usepackage{latexsym}
\usepackage{mathtools}
\usepackage{bbm}
\usepackage{multicol}

\usepackage[nohyperlinks]{acronym}

\newcommand*{\doi}[1]{doi:~\href{https://doi.org/#1}{\nolinkurl{#1}}}

\usepackage{tocloft}
\tcbset{width=\textwidth, boxrule=0pt, colback=gray!10, arc=0pt, auto outer arc, left=5pt, right=5pt, boxsep=5pt}

\def\cmk{\tikz\fill[scale=0.3](0,.35) -- (.25,0) -- (1,.7) -- (.25,.15) -- cycle;} 
\usepackage{pifont}
\newcommand{\xmk}{\ding{53}}%
\graphicspath{{figures/}}

\newcommand*\diff{\mathop{}\!\mathrm{d}}
\newcommand{\me}{\mathrm{e}}

\renewcommand{\footnotesize}{\fontsize{9pt}{11pt}\selectfont}

\begin{document}


\school{}
\title{Seamless Multimodal Biometrics for Continuous Personalised Wellbeing Monitoring}
\author{João Tiago Ribeiro Pinto}

\thesisdate{December 21, 2022}

\copyrightnotice{João Tiago Ribeiro Pinto, 2022}

\supervisor{Supervisor}{Professor Jaime dos Santos Cardoso}
\supervisor{Co-Supervisor}{Professor Miguel Velhote Correia}


\committeetext{Approved in public examination by the Jury:}
\committeemember{President}{\hspace{0.11in}Professor Luís Miguel Pinho de Almeida}
\committeemember{Referee}{\hspace{0.2in}Professor Hugo Filipe Silveira Gamboa}
\committeemember{Referee}{\hspace{0.2in}Professor Armando José Formoso de Pinho\vskip 0.1\baselineskip}
\committeemember{Referee}{\hspace{0.2in}Professor Jaime dos Santos Cardoso\vskip 0.1\baselineskip}
\committeemember{Referee}{\hspace{0.2in}Professor Ana Maria Rodrigues de Sousa Faria de Mendonça}
\committeemember{Referee}{\hspace{0.2in}Professor Luís Filipe Pinto de Almeida Teixeira\vskip 3\baselineskip}

\logo{logo-feup-inesc-2.pdf}

\newcommand{\etal}{\textit{et~al}.}
\newcommand{\eg}{\textit{e.~g.}}
\newcommand{\ie}{\textit{i.~e.}}
\newcommand{\cmt}[2]{\textcolor{#1}{\# (#2) \#}}



\begin{Prolog}
  \chapter*{Resumo}

A perceção através da inteligência artificial está cada vez mais presente nas nossas vidas. Os veículos não são exceção, uma vez que sistemas avançados de assistência ao condutor auxiliam no cumprimento de limites de velocidade, na manutenção dentro das faixas e na prevenção de acidentes. Num futuro próximo, o reconhecimento de padrões terá um paper ainda mais preponderante nos veículos, uma vez que o automóvel autónomo necessitará de meios automáticos para compreender o que acontece ao seu redor (e no seu interior) para agir de forma adequada.

Em reconhecimento de padrões, a biometria oferece aplicações promissoras para veículos, do acesso \emph{keyless} à personalização automática de opções de condução com base no condutor reconhecido. De igual modo, as tecnologias de  reconhecimento de bem-estar têm atraído atenção pela possibilidade de reconhecer atividade, emoções, sonolência ou stress em condutores e passageiros. No entanto, estes dois tópicos são diametralmente opostos, uma vez que o reconhecimento de bem-estar usa a variabilidade intra-sujeito, enquanto a biometria se baseia na variabilidade inter-sujeito.

Apesar das diferenças, a biometria e o reconhecimento de bem-estar poderiam (e deveriam) co-existir. O reconhecimento contínuo de identidade em dados adquiridos de forma impercetível poderiam ser usados para personalizar modelos de reconhecimento de bem-estar e obter melhor desempenho. Estes modelos personalizados poderiam ser a chave para meios mais robustos de monitorizar sonolência e atenção em condutores e evitar acidentes. Num sentido mais amplo, estes poderiam ser aplicados a todos os ocupantes, abrindo o caminho em direção ao reconhecimento eficaz de atividade, emoções, conforto e até episódios de violência em veículos autónomos partilhados. 

Este doutoramento focou-se em avançar o tópico de perceção em veículos através do estudo de novas metodologias de visão computacional e reconhecimento de padrões para biometria e reconhecimento de bem-estar. O foco principal foi na biometria com electrocardiograma (ECG), um traço reconhecido pelo seu potencial em monitorização impercetível de condutores. Esforços foram dedicados à obtenção de desempenho melhorado em identificação e verificação de identidade em cenários \emph{off-the-person}, reconhecidos pelo elevado teor de ruído e variabilidade. Aqui, foram propostas soluções \emph{deep learning end-to-end} e analisados tópicos importantes como o desempenho \emph{cross-database} e a longo prazo, a importância relativa das ondas através da interpretabilidade, e a conversão entre canais.

A biometria com face, um complemento natural ao ECG em cenários impercetíveis, foi também estudada nesta tese. Os desafios em reconhecimento de faces com máscaras e na interpretabilidade em biometria foram abordados com o intuito de avançar para algoritmos mais transparentes, confiáveis e robustos a oclusões significativas. Dentro do tópico de reconhecimento de bem-estar, foram propostas soluções melhoradas para o reconhecimento multimodal de emoções em grupos de pessoas e de atividade/violência dentro de veículos partilhados. Por fim, foram propostos ainda uma forma inovadora de aprender segurança de \emph{templates} em modelos \emph{end-to-end}, evitando processos adicionais de encriptação, e um método auto-supervisionado adaptado a dados sequenciais, para garantir segurança de dados e desempenho otimizado.

Segundo os resultados deste trabalho, é possível concluir que o ideal de reconhecimento personalizado de bem-estar está ainda por atingir. No entanto, este trabalho construiu uma base sólida para suportar trabalho futuro em direção à integração da biometria com o reconhecimento de bem-estar de forma multimodal, impercetível, contínua e realista. Em geral, este doutoramento levou a múltiplas contribuições para os tópicos de biometria e reconhecimento de bem-estar, resultando diretamente em vinte e quatro publicações científicas em fóruns de renome em biometria e reconhecimento de padrões. A sua qualidade e impacto foram reconhecidas pela comunidade científica com mais de trezentas citações e múltiplos prémios, incluindo o prémio \emph{EAB Max Snijder} 2022.

\vspace{1cm}

\noindent\textbf{Palavras-chave:} Aprendizagem Computacional; Atividade; Áudio; Biometria; Electrocardiograma; Emoção; Face; Monitorização de Bem-Estar; Reconhecimento de Padrões; Processamento de Sinal; Veículos Autónomos; Vídeo; Visão Computacional.

\chapter*{Abstract}

Artificially intelligent perception is increasingly present in the lives of every one of us. Vehicles are no exception, as advanced driver assistance systems (ADAS) help us comply with speed limits, keep within the lanes, and avoid accidents. In the near future, pattern recognition will have an even stronger role in vehicles, as self-driving cars will require automated ways to understand what is happening around (and within) them and act accordingly.

Within pattern recognition, biometrics offer promising applications in vehicles, from keyless access control to the automatic personalisation of driving and environmental conditions based on the recognised driver. Similarly, wellbeing monitoring technologies have long attracted attention to the possibility of recognising activity, emotions, sleepiness, or stress from drivers and passengers. However, these two topics are starkly opposed, since wellbeing recognition relies on intrasubject variability while biometrics thrives on intersubject variability. 

Despite their differences, biometric recognition and wellbeing monitoring could (and should) coexist. Continuous identity recognition from seamlessly acquired data could be used to personalise wellbeing monitoring models and attain improved performance. These personalised models could be the key to more robust ways of monitoring drivers' drowsiness and attention and avoiding accidents. In a broader sense, they could be applied to all vehicle occupants, paving the way towards the accurate recognition of activity, emotions, comfort, and even violence episodes in shared autonomous vehicles.

This doctoral work focused on advancing in-vehicle sensing through the research of novel computer vision and pattern recognition methodologies for both biometrics and wellbeing monitoring. The main focus has been on electrocardiogram (ECG) biometrics, a trait well-known for its potential for seamless driver monitoring. Major efforts were devoted to achieving improved performance in identification and identity verification in \emph{off-the-person} scenarios, well-known for increased noise and variability. Here, end-to-end deep learning ECG biometric solutions were proposed and important topics were addressed such as cross-database and long-term performance, waveform relevance through explainability, and interlead conversion.

Face biometrics, a natural complement to the ECG in seamless unconstrained scenarios, was also studied in this work. The open challenges of masked face recognition and interpretability in biometrics were tackled in an effort to evolve towards algorithms that are more transparent, trustworthy, and robust to significant occlusions. Within the topic of wellbeing monitoring, improved solutions to multimodal emotion recognition in groups of people and activity/violence recognition in in-vehicle scenarios were proposed. At last, we also proposed a novel way to learn template security within end-to-end models, dismissing additional separate encryption processes, and a self-supervised learning approach tailored to sequential data, in order to ensure data security and optimal performance.

Following the results of this work, one can conclude that truly personalised wellbeing is yet to be achieved. However, this work has built a strong framework to support future work towards the goal of integrating biometric recognition and wellbeing monitoring in a multimodal, seamless, continuous, and realistic way. Overall, this doctoral work led to numerous contributions to biometrics and wellbeing monitoring in general, resulting directly in twenty-four scientific publications in major biometrics and pattern recognition venues. Its quality and impact have been recognised by the scientific community with over three hundred citations and multiple awards, including the \emph{EAB Max Snijder Award} 2022.

\vspace{1cm}

\noindent\textbf{Keywords:} Activity; Audio; Autonomous Vehicles; Biometrics; Computer Vision; Electrocardiogram; Emotion; Face;  Machine Learning; Pattern Recognition; Signal Processing; Video; Wellbeing Monitoring.
  \chapter*{Acknowledgements}


Doutor. Finalmente. Volvidos cinco anos desde o início do meu percurso em investigação, a escrita deste documento dá-me a oportunidade de reviver os sucessos, fracassos, ideias, desafios, pessoas e momentos que marcaram este meu doutoramento. Foi uma excelente aventura, mas não foi fácil. As vozes do \emph{impostor syndrome} fizeram-me duvidar se realmente teria a inteligência e a capacidade suficientes para completar um doutoramento. Foi inevitável, por vezes, tentar comparar com os doutoramentos de outros, o que frequentemente resultou em desilusão. Mas fui aprendendo a evitar os obstáculos no caminho e a compreender os avisos de quem já passara por eles. Diziam que um doutoramento não é um \emph{sprint}, mas sim uma maratona. Que não requer capacidades extraordinárias, mas sim resiliência, motivação e determinação. Duvidei, mas depois deste longo percurso reconheço a veracidade dessa afirmação. Podemos duvidar da nossa inteligência e capacidades, mas o importante é continuar, e insistir, e procurar até encontrar a meta.

Com tempo percebi que cada doutoramento é único, e cada um tem o seu caminho a traçar, com desafios e dificuldades específicos. A comparação com os doutoramentos dos outros, apesar de inevitável, será sempre incompleta e injusta. Aprendi ainda que um doutoramento completo deverá ir bem para além da investigação científica. Não devemos ser apenas ``máquinas de fazer artigos'', mas sim procurar explorar todas as outras vertentes de um investigador completo, como o ensino, a mentoria, as colaborações, e a organização de eventos científicos. É natural que rapidamente esgotemos as horas do dia (e a nossa energia) quando nos dividimos entre tantas atividades. Foram frequentes as longas noites de trabalho e os fins-de-semana que não o foram. Mas considero que encontrei o verdadeiro caminho para um bom doutoramento, e fico feliz por ter decidido segui-lo.

Foi perfeito? Não. Em retrospetiva, reconheço escolhas menos certas, certamente frutos da minha inexperiência, e imensos caminhos alternativos que teriam sido mais proveitosos. Relembro com pesar as ideias que faziam sentido mas não funcionaram. E imagino o potencial de tantas outras que não foram além de um rabisco esquecido numa folha qualquer. Mas não estou só. Ainda não conheci um único aluno de doutoramento que tenha chegado ao fim do seu trajeto plenamente contente e verdadeiramente confiante no resultado. Resta a ténue calma na ideia de que esta tese de doutoramento não é a minha ``obra-prima'' nem aquilo que irá definir toda a minha carreira. Mas sim, apenas, um livro de esboços, uma coleção de rascunhos de alguém que tropeçou, escorregou e errou inúmeras vezes durante quase cinco anos na tentativa de mapear um caminho inexplorado e, com sorte, tornar-se um investigador completo e autónomo. 

Mudaria alguma coisa? Também não. Suponho que teria conseguido publicar mais artigos se não tivesse dividido a minha atenção entre tantas atividades paralelas. Talvez tivesse um trabalho com mais impacto se não tivesse orientado tantas teses de mestrado e estágios. Deveria talvez ter dado prioridade à importante experiência de dar aulas. Mas tudo isto é incerto. Certos são alguns momentos marcantes que deram um brilho especial a estes cinco anos de esforço e dedicação.
Relembro o convívio e a partilha de conhecimento no \emph{BTAS}, no \emph{IbPRIA}, no \emph{BIOSIG}, nos vários \emph{RECPADs} e em tantas outras conferências.
O prazer de poder contribuir para a organização do \emph{IWBF 2020} e dos workshops \emph{xAI4Biometrics}.
O orgulho em ver o meu esforço reconhecido pela comunidade europeia de biometria através do prémio Max Snijder.
O entusiasmo ao ver (e poder demostrar a tantas pessoas) o meu trabalho a funcionar, ao vivo, no fim do projeto Easy Ride.
A felicidade ao assistir a cada uma das provas públicas dos meus alunos de mestrado, concluindo com sucesso as suas teses, depois do privilégio de os acompanhar ao longo de um ano de aprendizagem, dedicação, esforço e evolução.
A honra enorme de ser o escolhido da Inês, da Sofia e do Duarte para os cartolar na Imposição das Insígnias, um momento tão solene e significante a marcar o fim dos seus caminhos pela academia.
A satisfação dos últimos dias de cada \emph{VISUM}, onde depois de tanto trabalho nos damos conta do imenso valor da nossa escola de verão (e ainda a minha estranha proficiência na apresentação de \emph{quizzes}).
E ainda, a amizade e entreajuda que encontrei todos os dias no grupo VCMI.
Ao relembrar as pessoas encontradas e estes momentos vividos, chego à conclusão que não mudaria absolutamente nada, no receio de que perdesse sequer um deles. E afinal, talvez o meu doutoramento não esteja assim tão longe da perfeição.

Por tudo isto tenho a agradecer, em primeiro lugar, ao Professor Jaime Cardoso. O melhor professor que tive a sorte de conhecer durante o meu percurso académico. Ao aproximar-se o fim do meu mestrado, muitas foram as dúvidas relativamente à possibilidade de um doutoramento. Há muito que era algo que almejava fazer, mas a magnitude da tarefa impunha respeito. Recebi muitas e variadas opiniões sobre o doutoramento e como fazer um com qualidade. No meio de expectável discórdia, um ponto de consenso: ``o orientador é, de longe, o mais importante''. Mais vale o orientador certo numa universidade qualquer que o orientador errado na \emph{Ivy League}, diziam em uníssono. Estavam certos, todos eles, e sem dúvida estava certo eu também quando escolhi fazer o doutoramento consigo, Professor. Um verdadeiro exemplo de integridade, profunda dedicação, impressionante inteligência e contagiante amor pelo que faz. Apesar das diferenças em experiência e currículo, tão comumente enfatizadas na academia, sempre me deixou à vontade para expressar livremente todas as ideias, dúvidas e problemas como se fosse um simples colega. E tudo isto vale o mundo para um aluno de doutoramento a percorrer a sua maratona. Recordo-me de certos momentos, como quando chegou a Professor Catedrático, em que me senti tão alegre como se tivesse sido eu próprio a conseguir essa conquista. E sei que não fui o único. Deixar tal marca nos alunos é prova definitiva da sua capacidade, dedicação e entrega como Professor e orientador. Desejo apenas que todos os seus objetivos se continuem a cumprir, com sucesso e felicidade, e que eu possa continuar a colaborar consigo e a tê-lo como meu mentor.

Ao Professor Miguel Velhote Correia, por ter aceitado o desafio de fazer parte deste projeto. Apesar dos nossos planos ambiciosos para este doutoramento, acabei por não ter muitas oportunidades para colaborar consigo. No entanto, fico feliz por ter conseguido que deixasse a sua marca naquele que considero ser o meu melhor trabalho durante este doutoramento. Continuo a acreditar no valor da investigação em instrumentação para o futuro da biometria com electrocardiograma, e espero um dia voltar a poder contar com o seu conhecimento e a sua experiência neste e em outros tópicos.

Ao grupo VCMI e aos que dele fizeram parte e contribuíram para a sua história, plena de inúmeros sucessos. Tal como ninguém vive numa bolha, também ninguém faz um doutoramento sozinho (apesar de ser uma aventura bastante solitária). Recordo a afirmação atribuída a Isaac Newton - ``\emph{if I have seen further, it is by standing on the shoulders of giants}''. Os que por aqui passaram antes de mim abriram-me e mostraram-me o caminho com os seus sucessos, e aqueles com os quais tive o prazer de percorrer esta jornada ajudaram-me a encontrar novas oportunidades e a aprender com diversos desafios em biometria e tantos outros tópicos. Não teria certamente conseguido metade do que consegui neste doutoramento se não tivesse como alicerce este grupo de verdadeiros gigantes, repleto de genialidade, capacidade e dedicação. Entre eles, um agradecimento especial ao Professor Jaime pela criação e constante dedicação ao grupo e à Filipa Sequeira pelos excelentes esforços recentes para aumentar a colaboração do subgrupo de biometria com a comunidade internacional. Obrigado por fazerem do grupo VCMI um símbolo de qualidade e excelência em biometria além-fronteiras, e por encherem o meu doutoramento de desafios, oportunidades, sucessos e ainda muitos momentos de felicidade. Sem dúvida, não podia ter escolhido um melhor grupo para o meu doutoramento.

À \emph{VISUM}, indubitavelmente a melhor escola de verão por esse mundo fora. Obrigado pela oportunidade de organizar estes momentos de confluência entre culturas e aprendizagem, com tal magnitude e visibilidade internacional. Agradeço a todos os que deram um pouco de si para tornar a \emph{VISUM} possível (e excelente), em especial à Ana, à Sara, e ao Wilson, que sacrificaram bem mais que todos os restantes e puxaram a nossa escola de verão para uma 10ª edição que ficará para a história. Cada edição foi única, mas foram todas fantásticas, e fico feliz de ter tido a oportunidade de fazer parte de quatro delas.

A todos os que confiaram em mim para co-orientar as suas tese de mestrado. Ao Gabriel, à Carolina, ao Leonardo, ao Arthur, ao João, à Inês, à Sofia, à Telma, à Mariana, ao Duarte, ao Guilherme, ao Vítor e ao Erfan. Cada um aceitou um tema único, com objetivos e dificuldades diferentes, e cada um o abordou com perspetivas e ideias diversas. Mas todos estiveram em sintonia na vontade de aprender, na dedicação à procura do caminho certo, e no ânimo firme mesmo em momentos de maior dificuldade. Sei que fui excecionalmente sortudo em ter tantos e tão excelentes alunos durante estes anos de doutoramento, e adorei trabalhar com cada um de vocês. Espero que tenham também gostado de trabalhar comigo (ou que pelo menos não se arrependam da vossa escolha) e que eu tenha estado à altura das vossas expectativas. Espero ainda que tenham aprendido algo comigo, tal como eu aprendi com cada um de vocês. Obrigado pelas vossas excelentes contribuições para este projeto, que na verdade também é um pouco vosso, e lembrem-se que continuarei aqui para vos ajudar, sempre que precisarem.

A todos aqueles que colaboraram comigo durante este doutoramento, tanto em temas relacionados com biometria como naqueles que me permitiram alargar os meus horizontes e adquirir conhecimentos em tópicos diferentes, incluindo os projetos AUTOMOTIVE, Easy Ride e Aurora. A todos na CardioID Technologies, em especial ao André, ao Carlos, ao Roberto, ao Pedro e ao Lourenço, pela ajuda em biometria com ECG e no projeto AUTOMOTIVE. À Bosch Car Multimedia, em especial ao Joaquim, ao Filipe, à Carolina, ao Ricardo, à Margarida, ao Niklas e ao Jochen pela excelente colaboração em \emph{in-vehicle monitoring} no projeto Easy Ride. Aos da Fraunhofer IGD, em especial ao Fadi, ao Naser, ao Florian, ao Marco, ao Juan e à Meiling pela colaboração em \emph{masked face recognition} e ainda pela calorosa receção em Darmstadt. Entre todos, um agradecimento especial ao André Lourenço pela colaboração, apoio e amizade já desde a minha tese de mestrado. Espero um dia poder voltar a colaborar com todos vós.

E por fim, mas certamente não em último lugar, à minha família e aos meus amigos, por tudo o resto que fez de mim o que sou hoje e que me ajudou a chegar aqui.

\vspace{1cm}
Obrigado a todos,

\vspace{0.5cm}
\emph{João Tiago}

\newpage

\section*{Funding}

This work was financed by the Portuguese science and technology foundation, Fundação para a Ciência e a Tecnologia -- FCT -- and co-financed by the European Social Fund through the North Regional Operational Programme (NORTE 2020), under the grant ``SFRH/BD/137720/2018''.

\noindent\includegraphics[width=\linewidth]{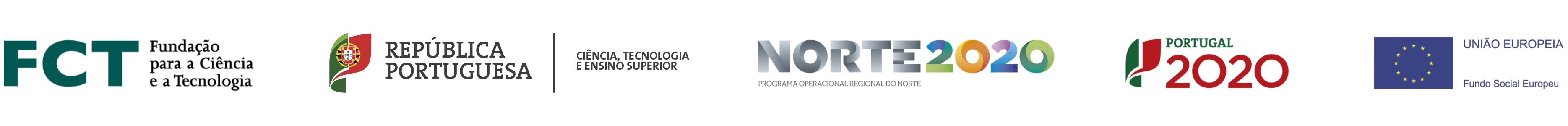}

\section*{Data}

The author wishes to thank the creators, contributors, and administrators of all databases and data collections used in the research work presented in this thesis. 

\vspace{0.2cm}

\noindent ECG signal databases and collections: the author acknowledges the creators of the PTB~\cite{Bousseljot1995} and PTB-XL~\cite{ptbxl1, ptbxl2} databases at the Physikalisch-Technische Bundesanstalt, Germany, the creators of the University of Toronto ECG Database (UofTDB)~\cite{Wahabi2014} at the University of Toronto, Canada, the creators of the Check Your Biosignals Here initiative~\cite{Silva2014} at Instituto de Telecomunicações, Portugal, the creators of the INCART database at the St. Petersburg Institute of Cardiological Technics, Russia, as well as the creators and administrators of the Physionet online data repository~\cite{Goldberger2000}. E-HOL-03-0202-003 (E-HOL) Data used for this research was provided by the Telemetric and Holter ECG Warehouse of the University of Rochester (THEW), NY.

\vspace{0.2cm}

\noindent Face biometric databases and collections: the author wishes to acknowledge the creators of the YouTube Faces~\cite{Wolf2011} database at Tel Aviv University, Israel, the creators of the ROSE Youtu~\cite{li2018unsupervised} database at the Nanyang Technological University, Singapore, the creators of the MFRC-21~\cite{Boutros2021MFR} database at Fraunhofer IGD, Germany, the creators of the VGGFace2~\cite{Cao2018} at the University of Oxford, UK, the creators of the MS1MV2~\cite{deng2019arcface} dataset at Imperial College London, UK, and the creators of the Labelled Faces in the Wild (LFW)~\cite{Huang2007} dataset at the University of Massachusetts, USA.

\vspace{0.2cm}

\noindent Other databases: the author wishes to acknowledge the EmotiW 2020 Grand Challenge~\cite{Dhall2020} organisers, and the authors of the Multi-Moments in Time (MMIT)~\cite{Monfort2019} dataset and pretrained models at the Massachusetts Institute of Technology, USA.

\section*{Credits}

All previously published copyrighted content reused, reprinted, or adapted in this thesis is appropriately referenced and acknowledged and has been licensed as detailed below:

\vspace{0.2cm}

\noindent Content in Chapter~\ref{ch:fundamentals}: Adapted from João Ribeiro Pinto, ``Continuous Biometric Identification on the Steering Wheel'', M.Sc. Thesis, University of Porto, Portugal, 2017. Figure~\ref{fig:face_variability} was reprinted from Cognition, vol.~121, Rob Jenkins, David White, Xandra Van Montfort, A. Mike Burton, ``Variability in photos of the same face'', pp. 313--323, Copyright 2011, with permission from Elsevier.

\vspace{0.2cm}

\noindent Content in Chapter~\ref{ch:ecgprior} and Appendix~\ref{ch:appendix_ecg}: © 2018 IEEE. Reprinted, with permission, from João Ribeiro Pinto, Jaime S. Cardoso, André Lourenço, ``Evolution, Current Challenges, and Future Possibilities in ECG Biometrics'', IEEE Access, June 2018.

\vspace{0.2cm}

\noindent Content in Chapter~\ref{ch:ecgiden} and Figure~\ref{fig:ecg_recog_structure}: Reproduced with permission of Taylor and Francis Group LLC (Books) US through PLSclear.

\vspace{0.2cm}

\noindent Content in Chapter~\ref{ch:ecgauth}: © 2019 IEEE. Reprinted, with permission, from João Ribeiro Pinto, Jaime S. Cardoso, ``An End-to-End Convolutional Neural Network for ECG-Based Biometric Authentication'', 2019 IEEE 10th International Conference on Biometrics Theory, Applications and Systems (BTAS), September 2019. 

\vspace{0.2cm}

\noindent Content in Chapter~\ref{ch:ecglongterm}: Reprinted/adapted by permission from Springer Nature: Springer Nature ``Don't You Forget About Me: A Study on Long-Term Performance in ECG Biometrics'' by Gabriel Lopes, João Ribeiro Pinto, Jaime S. Cardoso © 2019.

\vspace{0.2cm}





\noindent Content in Chapter~\ref{ch:faceprior}: Figures~\ref{fig:face_recog_history} and \ref{fig:face_recog_history_2} were reprinted from Neurocomputing, vol.~429, Mei Wang, Weihong Deng, ``Deep face recognition: A survey'', pp.~30, Copyright 2021, with permission from Elsevier.

\vspace{0.2cm}

\noindent Content in Chapter~\ref{ch:maskedFaceRecog}: © 2021 IEEE. Reprinted, with permission, from Pedro C. Neto, Fadi Boutros, João Ribeiro Pinto, Mohsen Saffari, Naser Damer, Ana F. Sequeira, Jaime S. Cardoso, ``My Eyes Are Up Here: Promoting Focus on Uncovered Regions in Masked Face Recognition'', 2021 International Conference of the Biometrics Special Interest Group (BIOSIG), September 2021.

\vspace{0.2cm}



\noindent Content in Chapter~\ref{ch:emotion}: © 2020 IEEE. Reprinted, with permission, from João Ribeiro Pinto, Tiago Gonçalves, Carolina Pinto, Luís Sanhudo, Joaquim Fonseca, Filipe Gonçalves, Pedro Carvalho, Jaime S. Cardoso, ``Audiovisual Classification of Group Emotion Valence Using Activity Recognition Networks'', 2020 IEEE 4th International Conference on Image Processing, Applications and Systems (IPAS), December 2020.

\vspace{0.2cm}



\noindent Content in Chapter~\ref{ch:secure}: © 2021 IEEE. Reprinted, with permission, from João Ribeiro Pinto, Miguel V. Correia, Jaime S. Cardoso, ``Secure Triplet Loss: Achieving Cancelability and Non-Linkability in End-to-End Deep Biometrics'', IEEE Transactions on Biometrics, Behavior, and Identity Science, April 2021.

\vspace{0.2cm}

\noindent Content in Chapter~\ref{ch:selfsupervised}: © 2020 IEEE. Reprinted, with permission, from João Ribeiro Pinto, Jaime S. Cardoso, ``Self-Learning with Stochastic Triplet Loss'', 2020 International Joint Conference on Neural Networks (IJCNN), July 2020.


  \cleardoublepage
\thispagestyle{plain}

\vspace*{8cm}

\begin{flushright}
   \textsl{``In theory, there is no difference between theory and practice,\\
   while in practice, there is.''} \\
\vspace*{1.5cm}
           Benjamin Brewster
\end{flushright}


  \cleardoublepage
  \pdfbookmark[0]{Table of Contents}{contents}
  \tableofcontents
  \cleardoublepage
  \pdfbookmark[0]{List of Figures}{figures}
  \listoffigures
  \cleardoublepage
  \pdfbookmark[0]{List of Tables}{tables}
  \listoftables
  \chapter*{Abbreviations}
\chaptermark{ABBREVIATIONS}


\begin{multicols}{2}
\raggedright
\begin{acronym}
\acro{1D}{Unidimensional}
\acro{2D}{Bidimensional}
\acro{2.5D}{Bidimensional with Depth Information}
\acro{3D}{Tridimensional}
\acro{AC}{Autocorrelation}
\acro{ADAS}{Advanced Driver Assistance Systems}
\acro{AE}{Autoencoder}
\acro{AHA}{American Heart Association (dataset)}
\acro{AMAE}{Average Mean Absolute Error}
\acro{ANN}{Artificial Neural Network}
\acro{APCER}{Attack Presentation Classification Error Rate}
\acro{AR}{Autoregressive (coefficients)}
\acro{AUC}{Area Under the Curve}
\acro{BF}{\emph{Bona Fide}}
\acro{BPCER}{\emph{Bona fide} Presentation Classification Error Rate}
\acro{BPF}{Bandpass Filter}
\acro{BIOSIG}{Biometrics Special Interest Group}
\acro{Bi-LSTM}{Bidirectional Long Short-Term Memory (network)}
\acro{CAM}{Class Activation Mapping}
\acro{CASIA}{Chinese Academy of Sciences}
\acro{CC}{(Pearson's) Correlation Coefficient}
\acro{CC0}{Creative Commons ``No Rights Reserved'' License}
\acro{CCC}{Concordance Correlation Coefficient}
\acro{CE}{Cross-Entropy}
\acro{CGFSI}{Cross-GAN Filter Similarity Index}
\acro{CIBB}{International Conference on Computational Intelligence Methods for Bioinformatics and Biostatistics}
\acro{CinC}{Computers in Cardiology (conference)}
\acro{CMC}{Cumulative Match Characteristic}
\acro{CNN}{Convolutional Neural Network}
\acro{CSIST}{Chung-Shan Institute of Science and Technology (datasets)}
\acro{CTM}{Centre for Telecommunications and Multimedia}
\acro{CYBHi}{Check Your Biosignals Here initiative}
\acro{CWT}{Continuous Wavelet Transform}
\acro{DBNN}{Decision-based Neural Network}
\acro{DCT}{Discrete Cosine Transform}
\acro{DET}{Detection Error Trade-off (curve)}
\acro{DNA}{Deoxyribonucleic Acid}
\acro{DT}{Direct Training (experimental scenario)}
\acro{DTW}{Dynamic Time Warping}
\acro{DWT}{Discrete Wavelet Transform}
\acro{EAB}{European Association for Biometrics}
\acro{ECCV}{European Conference on Computer Vision}
\acro{ECG}{Electrocardiogram}
\acro{EEG}{Electroencephalogram}
\acro{EEMD}{Ensemble Empirical Mode Decomposition}
\acro{EER}{Equal Error Rate}
\acro{EMD}{Empirical Mode Decomposition}
\acro{EMG}{Electromyogram}
\acro{ERDF}{European Regional Development Fund}
\acro{ETs}{Extremely Randomised Trees}
\acro{FAR}{False Acceptance Rate}
\acro{FC}{Fully-Connected (network layer)}
\acro{FCT}{Fundação para a Ciência e a Tecnologia}
\acro{FDDB}{Face Detection Database \& Benchmark}
\acro{FEC}{Forward Error Control}
\acro{FERET}{Facial Recognition Technology}
\acro{FG4COVID19}{Face and Gesture Analysis for COVID-19 (workshop)}
\acro{FLDA}{Fisher Linear Discriminant Analysis}
\acro{FLRP}{Focused Layer-wise Relevance Propagation}
\acro{FMR}{False Match Rate}
\acro{FNMR}{False Non-Match Rate}
\acro{FOGD}{First-Order Gaussian Derivative (filter)}
\acro{FPIR}{False Positive Identification Rate}
\acro{FRR}{False Rejection Rate}
\acro{FRVT}{Face Recognition Vendor Test}
\acro{FT}{Fine-Tuning}
\acro{FTDNN}{Focused Time-Delay Neural Network}
\acro{GAN}{Generative Adversarial Network}
\acro{GBFS}{Greedy Best-First Search}
\acro{GMM}{Gaussian Mixture Model}
\acro{GMean}{Genuine Mean (scores)}
\acro{GNMF}{Graph-regularised Non-negative Matrix Factorisation}
\acro{GPU}{Graphics Processing Unit}
\acro{GRU}{Gated Recurrent Unit}
\acro{Grad-CAM}{Gradient-weighted Class Activation Mapping}
\acro{HASLab}{High-Assurance Software Laboratory}
\acro{HE}{Homomorphic Encryption}
\acro{HLDA}{Heteroscedastic Linear Discriminant Analysis}
\acro{HPF}{Highpass Filter}
\acro{HRV}{Heart Rate Variability}
\acro{I3D}{Two-Stream Inflated Tridimensional Convolutional Network}
\acro{IARPA}{Intelligence Advanced Research Projects Activity}
\acro{ICA}{Independent Component Analysis}
\acro{IDR}{Identification Rate}
\acro{IJB-C}{IARPA Janus Benchmark-C}
\acro{IJCB}{International Joint Conference on Biometrics}
\acro{IJCNN}{International Joint Conference on Neural Networks}
\acro{IMDb}{Internet Movie Database}
\acro{IMean}{Impostor Mean (scores)}
\acro{IMF}{Intrinsic Mode Function}
\acro{IOMBA}{Interval Optimised Mapping Bit Allocation}
\acro{IoU}{Intersection over Union}
\acro{IPAS}{International Conference on Image Processing, Applications and Systems}
\acro{ISEL}{Instituto Superior de Engenharia de Lisboa}
\acro{ISM}{In-Vehicle Sensing Monitorisation (workshop)}
\acro{IT-CNN}{Identification Training Convolutional Neural Network }
\acro{IWBF}{International Workshop on Biometrics and Forensics}
\acro{KLD}{Kullback-Leibler Divergence}
\acro{kNN}{k-Nearest Neighbours}
\acro{KPCA}{Kernel Principal Component Analysis}
\acro{KSS}{Karolinska Sleepiness Scale}
\acro{LBP}{Local Binary Patterns}
\acro{LDA}{Linear Discriminant Analysis}
\acro{LDP}{Local Difference Patterns}
\acro{LFW}{Labeled Faces in the Wild (dataset)}
\acro{LLR}{Log-Likelihood Ratio}
\acro{LPF}{Lowpass Filter}
\acro{LRN}{Label Refinement Network}
\acro{LSTM}{Long Short-Term Memory (network)}
\acro{LWIR}{Long-Wave Infrared}
\acro{MB}{Megabyte}
\acro{MFCC}{Mel-Frequency Cepstral Coefficients}
\acro{MFR}{Masked Face Recognition}
\acro{MFRC-21}{2021 Masked Face Recognition Competition Dataset}
\acro{MIDR}{Misidentification Rate}
\acro{MIT-BIH}{Massachusetts Institute of Technology - Beth Israel Hospital}
\acro{MLP}{Multi-Layer Perceptron}
\acro{MMAE}{Maximum Mean Absolute Error}
\acro{MMIT}{Multi-Moments in Time (dataset)}
\acro{MRLBP}{Multi-Resolution Local Binary Patterns}
\acro{MSDF-1DMRLBP}{Multi-Scale Differential Features fusion of Unidimensional Multi-Resolution Local Binary Patterns}
\acro{MSE}{Mean Squared Error}
\acro{MSFS}{Multisession Feature Selection}
\acro{MTCNN}{Multi-Task Convolutional Neural Network}
\acro{MWIR}{Mid-Wave Infrared}
\acro{NCC}{Normalised Cross-Correlation}        
\acro{NCCC}{Normalised Cross-Correlation Clustering}  
\acro{NF}{Notch Filter}
\acro{NIR}{Near-Infrared}
\acro{NIS}{Networked Intelligent Systems (cluster)}
\acro{NIST}{National Institute of Standards and
Technology}
\acro{NN}{Neural Network}
\acro{NRC}{Normalised Relative Compression}
\acro{NSR}{Normal Sinus Rhythm}
\acro{OA}{One-Attack (evaluation scenario)}
\acro{OCFR}{Advanced Occluded Face Recognition (competition)}
\acro{PA}{Presentation Attack}
\acro{PAD}{Presentation Attack Detection}
\acro{PAI}{Presentation Attack Instrument}
\acro{PAISp}{Presentation Attack Instrument Species}
\acro{PaSC}{Point and Shoot Face Recognition Challenge}
\acro{PCA}{Principal Component Analysis}
\acro{PLI}{Powerline Interference}
\acro{PNN}{Probabilistic Neural Network}
\acro{PPG}{Photoplethysmogram}
\acro{PRD}{Percent Root-Mean-Square Difference}
\acro{PSD}{Power Spectral Density}
\acro{PTB}{Physicalisch-Technische Bundesanstalt (dataset)}
\acro{PTCD}{Probability of Time to Correct Decision (score)}
\acro{RBF}{Radial Basis Function}
\acro{RCNN}{Region-based Convolutional Neural Network}
\acro{ReLU}{Rectified Linear Unit}
\acro{RF}{Random Forest}
\acro{RGB}{Red-Green-Blue (colour model)}
\acro{RMFD}{Real Masked Face Data (evaluation scenario)}
\acro{RMSE}{Root Mean Squared Error}
\acro{ROC}{Receiver Operating Characteristic (curve)}
\acro{RR}{Reject Rate}
\acro{RSA}{Recurrent Scale Approximation}
\acro{S$^3$FD}{Single Shot Scale-Invariant Face Detector}
\acro{SAGR}{Sign Agreement Metric}
\acro{SAX}{Symbolic Aggregate Approximation}
\acro{SCG}{Seismocardiogram}
\acro{SecureTL}{Secure Triplet Loss}
\acro{SFA}{Simplified Fuzzy ARTMAP}
\acro{SGD}{Stochastic Gradient Descent}
\acro{SHAP}{Shapley Addictive Explanation}
\acro{SIMCA}{Soft Independent Modelling by Class Analogy}
\acro{SL}{Self-Learning}
\acro{SMFD}{Synthetic Masked Face Data (evaluation scenario)}
\acro{SSIM}{Structural Similarity Index Measure}
\acro{SVM}{Support Vector Machine}
\acro{StD}{Standard Deviation}
\acro{STFT}{Short-Time Fourier Transform}
\acro{SWIR}{Short-Wave Infrared}   
\acro{TCD}{Time to Correct Decision (score)}
\acro{TL-CNN}{Transfer-Learning Convolutional Neural Network }
\acro{TPIR}{True Positive Identification Rate}
\acro{TVCG}{Tridimensional Vectorcardiogram}
\acro{t-SNE}{t-Distributed Stochastic Neighbour Embedding}
\acro{UA}{Unseen-Attack (evaluation scenario)}
\acro{UBM}{Universal Background Model}
\acro{ULHT}{Universidade Lusófona de Humanidades e Tecnologias}
\acro{UMAP}{Face and Gesture Analysis for COVID-19}
\acro{UMD}{University of Maryland (dataset)}
\acro{USC}{Usability-Security Curve}
\acro{UofTDB}{University of Toronto ECG Database}
\acro{Var}{Variance}
\acro{VCMI}{Visual Computing and Machine Intelligence (research group)}
\acro{VGAF}{Video-level Group Affect (dataset)}
\acro{VGG}{Visual Geometry Group (dataset)}
\acro{VISAPP}{International Conference on Computer Vision Theory and Applications}
\acro{VGA}{Video Graphics Array}  
\acro{WACV}{Winter Conference on Applications of Computer Vision}
\acro{WDIST}{Wavelet Distance}
\acro{WWPRD}{Wavelet-Weighted Percent Root-Mean-Square Difference}
\acro{xaFCM}{Extended-alphabet Finite-Context Model}
\acro{xAI4Biometrics}{Explainable Artificial Intelligence for Biometrics (workshop)}
\end{acronym}
\end{multicols}

\end{Prolog}


\StartBody

\part{Prologue}\label{part:prologue}
\chapter{Introduction}\label{ch:intro}

\section{Context and Motivation}

The interactions between humans and machines are increasingly mediated by intelligent sensing technologies. Nowadays, the default way to unlock a smartphone is using face or fingerprint biometrics. Highly-populated countries like India and China are building unprecedentedly massive national identity networks relying entirely on biometric characteristics (and dealing with the societal intricacies of such colossal endeavours)~\cite{Jain2012, Singh2021, Liu2021}. Sophisticated algorithms monitor our attention levels while we drive~\cite{Hu2022, Fang2022}. Meanwhile, the gaming and entertainment industries are investing heavily in using affective computing to continuously personalise and enhance user experience~\cite{Bakkes2013Personalised, Lim2020, Melo2014}.

Among all of these applications, few have been so revolutionised (and so quickly) as vehicles, and there are plenty of great reasons for that. Whether in personal cars or public transport, people can typically spend up to multiple hours of their day inside vehicles, especially those with long daily commutes or those living/working in large urban centres. Moreover, a considerable fraction of preventable deaths and serious injuries occur in accidents involving vehicles, caused mainly by driver fatigue or the influence of psychotropic substances~\cite{Sahayadhas2012, Silveira2019}.

While we wait for fully autonomous vehicles, advanced driver assistance systems (ADAS) based on artificial intelligence help drivers comply with speed limits, stay in their lanes, beware of their blind spots, and avoid accidents~\cite{Cicchino2017, Cicchino2018, Souders2020}. One of the most interesting applications of this is drowsiness detection, where biometric data such as face video or physiological signals can be used to detect episodes of sleepiness and infer the fatigue levels of the driver~\cite{Esteves2021AUTOMOTIVE, Silveira2019, Oliveira2018}. When done continuously (or at least frequently) during vehicle usage, it can be used to warn the driver or even trigger the automatic safe interruption of the vehicle's operation.

Drowsiness, just like most other wellbeing parameters, reveals itself on biometric data as intrasubject variability: the way a person's data varies over time and across diverse conditions~\cite{Esteves2021AUTOMOTIVE}. This stands in stark opposition to biometric recognition applications, which rely on intersubject variability (the way a person's data differs from another's) to distinguish identities~\cite{Kaur2014, Pinto2018}. In biometric recognition, intrasubject variability is typically regarded as a nuisance, as it blurs the boundaries between individuals' data and hampers accurate and robust identity recognition~\cite{Jain1999}.

In spite of their differences, biometric recognition and wellbeing monitoring can coexist in a vehicle scenario. Applications of biometric recognition for vehicles are typically focused on access control or the automatic personalisation of driving and environmental conditions (such as seat position, mirror adjustments, or infotainment settings) based on the recognised driver and occupants~\cite{Pinto2017, Daimi2021}. However, in this thesis, we defend that this coexistence could (and should) be intensified, and eventually evolve to a level of symbiotic integration.

Despite the efforts devoted to wellbeing monitoring technologies, one challenge remains unvanquished: the fact that wellbeing patterns are deeply personal. For example, no two individuals experience drowsiness in the exact same way, and similar levels of fatigue can have dramatically different effects on each person. This is also true for the way wellbeing parameters reflect on biometric data. A drowsiness monitoring system can present acceptable accuracy levels for a given set of subjects and, simultaneously, be inadequate at recognising the specific drowsiness patterns of other people~\cite{Esteves2021AUTOMOTIVE, Silveira2019}.

Biometric recognition could be the key to unlocking the next generation of wellbeing monitoring technologies. Instead of using generalist models, identity predictions obtained through biometric recognition would enable the selection of specific models for each of the users. These models could also benefit from the influx of new data, continuously learning the subject-specific patterns of wellbeing. Thus, combining biometrics with wellbeing could enable the creation of more accurate and robust solutions for wellbeing monitoring~\cite{Esteves2021AUTOMOTIVE}.

In the future, fully autonomous driving may put an end to the need for drivers, but not to the need for automatic intelligent sensing technologies inside vehicles~\cite{Augusto2020, Murali2022, Pinto2022Streamlining}. The same biometric algorithms and wellbeing monitoring systems that once were focused on the driver may easily be adapted to target the occupants. In fact, self-driving vehicles unveil a new scenario for automatic passenger monitoring: since there is no driver, shared autonomous vehicles (such as autonomous taxis) lack an authority figure, responsible for the integrity of the vehicle and the comfort and security of the passengers. The driver could be replaced by pattern recognition solutions to monitor the shared vehicle interior and its passengers.

Overall, the advantages of introducing intelligent sensing technologies in vehicles are plenty. From a narrower subject-centric perspective, integrating the use of inter and intrasubject variability would enable the development of continuously personalisable models for more robust monitoring of wellbeing parameters. From a broader perspective, taking advantage of seamless multimodal data acquisitions for automatic monitoring of the interior and passengers is of utmost importance to ensure security and comfort inside autonomous shared vehicles.

\section{Objectives}

This doctoral work focused on advancing in-vehicle sensing technology through the conceptualisation and development of novel computer vision and pattern recognition methodologies. The goal has been to create automatic solutions for in-vehicle monitoring, robustly and efficiently. To achieve this, we targeted two specific scenarios, as follows:
\begin{itemize}
    \item \emph{Personalised wellbeing monitoring systems using biometrics}: Here, the objective was to advance biometric recognition technologies to be integrated with wellbeing monitoring methodologies, especially for driver assistance systems. We build upon the ECG biometrics research conducted in~\cite{Pinto2017b}, with additional work on face recognition and other important topics such as biometric security and learning from data streams. With this work, we aimed to pave the way towards a robust multimodal system to recognise the driver using ECG and face information. The automatic identity predictions enable the use of the drivers' data to continuously learn their personal patterns of wellbeing for more accurate monitoring. Within wellbeing, this research work focused on driver drowsiness and emotions, as part of the AUTOMOTIVE project, but it stands to reason that biometrics could be used for personalised monitoring of any wellbeing parameter;
    
    \item \emph{Occupant monitoring for autonomous shared vehicles}: Forecasting the advent of fully autonomous vehicles, this work aimed towards the use of data streams for monitoring occupants. Integrated within the Easy Ride project, we targeted the monitoring of emotions among passenger groups, as well as the recognition of activities and violence inside shared autonomous vehicles. Just like the first scenario, this aims towards contributions to more intelligent and robust wellbeing monitoring systems using multimodal data sources, albeit in a less personal subject-centric way, focused instead on passenger groups as a whole. Although activity recognition is a relatively mature research topic, the in-vehicle environment offers very specific challenges, mainly regarding perspective, lighting, and occlusions, which enabled the study of innovative solutions for robust passenger monitoring. Ultimately, the developed occupant wellbeing monitoring solutions could also be personalised, using the biometric recognition of individual passengers as additional information for improved accuracy.
\end{itemize}

Despite these two scenarios, used to contextualise and motivate the work conducted during this work, we aimed to build solutions that are applicable outside the target applications and beyond the fields of biometrics and wellbeing monitoring. This work also aimed to result in significant advances to the target fields, which are ready for real-life applications and capable to withstand the test of time. As such, throughout the entirety of this doctoral project, there was a constant concern to ensure the proposed methodologies were:
\begin{itemize}
    \item \emph{Multimodal}: The diversification of data sources is the key to truly robust models. On the topic of biometric recognition, considering the strengths and shortcomings of the ECG signal as a biometric trait, the face is the best complementary trait for better performance and robustness. Beyond biometric recognition, the fusion of ECG and face results in a larger availability of anatomical and physiological measurements, which enable the more comprehensive monitoring of wellbeing parameters;

    \item \emph{Seamless}: Regardless of the possible consequences to accuracy, user comfort should be of utmost importance. Hence, subjects should be as unaware of the acquisition process as possible, to avoid attention or behaviour changes that could impact their comfort or the realism of the collected data. As such, this work focused, as extensively as possible, on seamlessly acquired data, in nearly unconstrained settings. When drivers are the target, and thus the subject is in physical contact with the system nearly continuously (\eg, driving the car), ECG and other physiological signals can be acquired unobtrusively at the steering wheel. Face video can also be easily and inexpensively acquired with cameras. On the other hand, for shared vehicle occupant monitoring, physiological signals have been avoided in favour of less contact-intensive alternatives, namely video and audio;

    \item \emph{Continuous}: Continuous biometric methodologies offer unique advantages in usability and effectiveness, both for recognition and wellbeing monitoring. On the other hand, having a continuous stream of biometric data opens up new possibilities for improved accuracy, immersive systems, and error management. As such, beyond novelty and performance, the algorithms developed during this work aimed towards real-time operation, reflecting a constant preference for efficiency and simplicity whenever possible;

    \item \emph{Realistic}: Overall, performance results in ECG-based biometrics literature are unrealistic, mainly due to inadequate train-test splits and overly clean signal databases. The same happens in wellbeing monitoring when the problem of subject-independence is highlighted. This reveals the deeply flawed nature of typical evaluation frameworks, which should more realistically resemble actual application conditions and ensure reproducible results. To achieve this, this doctoral work included the reformulation of testing procedures, especially for ECG-based biometrics, through the definition of adequate test protocols, the benchmarking of literature methods, and the development of more robust recognition and monitoring algorithms.
\end{itemize}

\section{Contributions}

\begin{figure}[t]
    \centering
    \includegraphics[width=\linewidth]{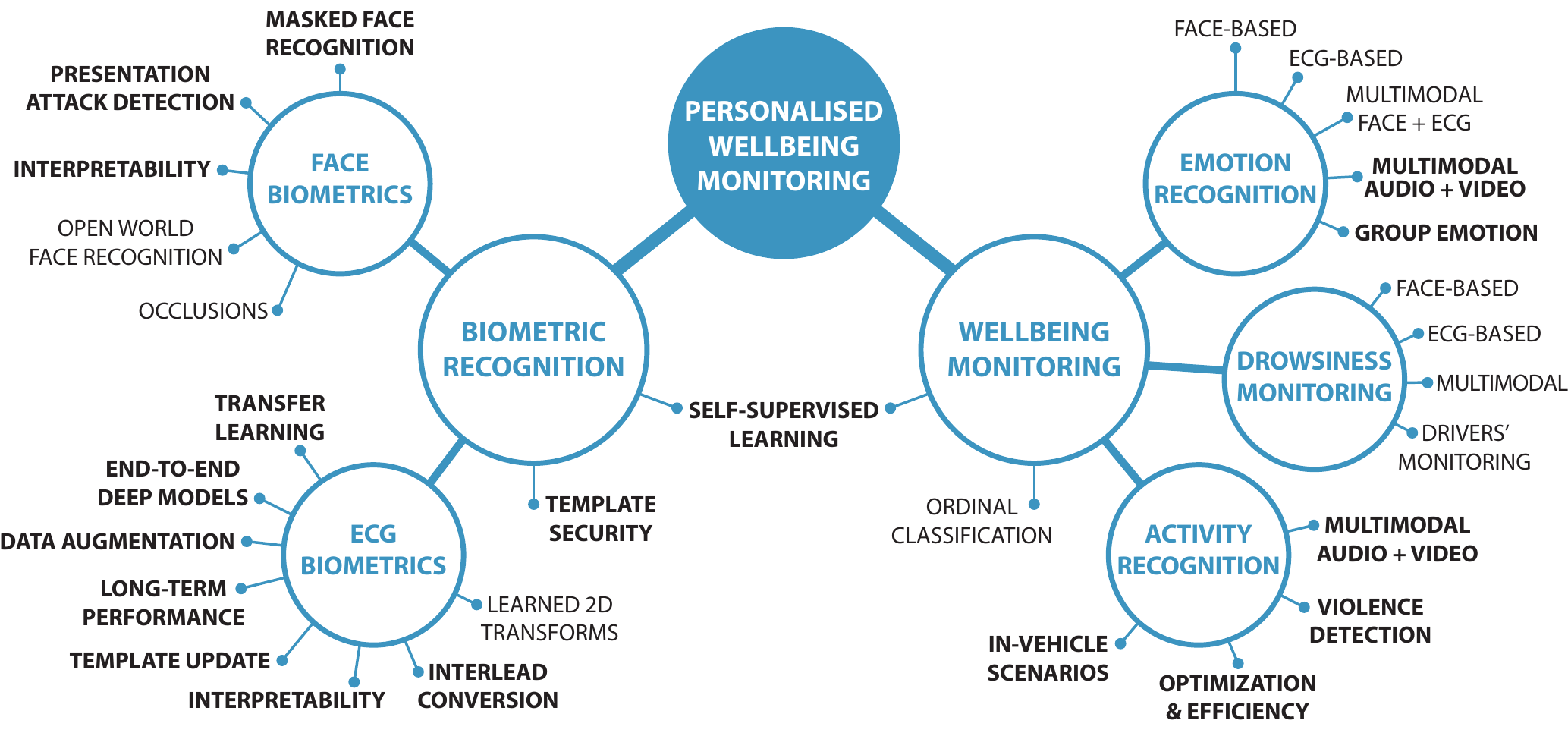}
    \caption[Schema of the topics covered during this doctoral project, their interconnections, and their link to the central topic of personalised wellbeing monitoring.]{Schema of the topics covered during this doctoral project, their interconnections, and their link to the central topic of personalised wellbeing monitoring (black topics in bold correspond to the strongest direct contributions, which are presented in this thesis).}
    \label{fig:intro_schema_topics}
\end{figure}

On the quest for personalised wellbeing monitoring, focused on the objectives detailed in the previous section, this doctoral project comprised several research topics related to both biometric recognition and wellbeing monitoring. Fig.~\ref{fig:intro_schema_topics} presents them and illustrates their interconnections. Most of these were the focus of research work during this doctoral project and resulted in innovative contributions towards the target of in-vehicle personalised wellbeing monitoring.

This thesis presents the innovative contributions of this work organised throughout four parts. The first two of these are focused on biometric recognition, aiming towards the development of personalised wellbeing monitoring solutions. The third part targets the scenario of passenger monitoring inside shared autonomous vehicles, specifically for emotion, activity, and violence recognition. The fourth presents broader contributions applicable to multiple scenarios in biometrics and pattern recognition. These contributions are concisely enumerated below.

In Part~\ref{part:ecgBiometrics}, \emph{Electrocardiogram Biometrics}:
\begin{itemize}
    \item a comprehensive survey of one hundred and twenty-five state-of-the-art methodologies for ECG-based biometric recognition, describing the evolution of the topic from 1999 to 2022 and open challenges and opportunities for future research (Chapter~\ref{ch:ecgprior});
    \item the first end-to-end methodology for ECG biometrics, alongside tailored data augmentation strategies for ECG signals and a study on the advantages of integrating typically separate processes inside a single deep architecture (Chapter~\ref{ch:ecgiden});
    \item an application of triplet loss and transfer learning for ECG-based identity verification, aiming towards higher robustness under realistic evaluation setups using off-the-person databases (Chapter~\ref{ch:ecgauth});
    \item a study on long-term performance with multiple state-of-the-art ECG biometric methodologies, including an assessment of the effect of diverse template and model update strategies (Chapter~\ref{ch:ecglongterm});
    \item a study on the relative importance of ECG waveforms for identification under diverse scenarios, less to more challenging, using explainability tools on our deep learning identification methodology (Chapter~\ref{ch:ecgexpl});
    \item a methodology for recovering missing ECG leads based on single-lead blindly-segmented input signals, paving the way towards more complete applications (even in clinical scenarios) with less obtrusive signal collection setups (Chapter~\ref{ch:ecginterlead}).
\end{itemize}

In Part~\ref{part:faceBiometrics}, \emph{Face Biometrics}:
\begin{itemize}
    \item a custom training methodology tailored for face recognition with masks, aiming to promote the use of unmasked parts of the face and close the performance gap relative to typical face recognition (Chapter~\ref{ch:maskedFaceRecog}); 
    \item a study using interpretability tools to understand the use of face image information in presentation attack detection (PAD), alongside a discussion on the need for explainability and transparency in biometric recognition (Chapter~\ref{ch:interpFacePAD}).
\end{itemize}

In Part~\ref{part:wellbeingMonitoring}, \emph{Wellbeing Monitoring}: 
\begin{itemize}
    \item an approach for classifying emotion valence in groups of people, based on the late fusion of parallel visual and audio data streams using deep neural networks (Chapter~\ref{ch:emotion});
    \item a cascade strategy for improved efficiency in continuous audiovisual activity recognition and violence detection inside vehicles (Chapter~\ref{ch:activity}).
\end{itemize}

In Part~\ref{part:broaderTopics}, \emph{Broader Topics on Biometrics and Pattern Recognition}:
\begin{itemize}
    \item the Secure Triplet Loss, a training methodology that promotes template cancelability and unlinkability, alongside identity discrimination, on end-to-end biometric algorithms without the need for separate encryption or hashing processes (Chapter~\ref{ch:secure});
    \item a methodology for self-supervised learning based on the triplet loss, taking advantage of the nature of balanced multiclass datasets, especially those composed of sequential data, for more adequate learning of the target tasks (Chapter~\ref{ch:selfsupervised}).
\end{itemize}

These research works may be categorised according to the specific contributions by the author of this thesis. The presented work in ECG-based biometric identification (Chapter~\ref{ch:ecgiden}), authentication (Chapter~\ref{ch:ecgauth}), and explainability (Chapter~\ref{ch:ecgexpl}), as well as emotion recognition (Chapter~\ref{ch:emotion}), activity recognition (Chapter~\ref{ch:activity}), biometric template security (Chapter~\ref{ch:secure}), and self-supervised learning (Chapter~\ref{ch:selfsupervised}) comprise the main contributions, resulting completely or mostly from the work performed by the author. The research on long-term performance in ECG biometrics (Chapter~\ref{ch:ecglongterm}), ECG interlead conversion (Chapter~\ref{ch:ecginterlead}), masked face recognition (Chapter~\ref{ch:maskedFaceRecog}), and interpretability for face PAD (Chapter~\ref{ch:interpFacePAD}) are secondary contributions that resulted from collaborations within the scope of this thesis and benefitted partially from the work of the author. Further details on the specific contributions can be found at the start of each chapter. Other topics have been addressed which are not presented in this thesis, due to the relevance of the respective topics or the relative weight of the author's contributions. They are, however, also mapped in Fig.~\ref{fig:intro_schema_topics} and listed in the sections below.

\section{Dissemination}

The contributions of the doctoral research to biometrics, wellbeing monitoring, and broader topics described in this thesis have been disseminated as part of twenty-four scientific publications. These are (clustered by type and in reverse chronological order):

\begin{itemize}
    \item Articles in journals:
        \begin{etaremune}
            \item S. Beco, \underline{J. R. Pinto}, and J. S. Cardoso, ``Electrocardiogram Lead Conversion from Single-Lead Blindly-Segmented Signals,'' \emph{BMC Medical Informatics and Decision Making}, 22: 314, 2022. \cite{Beco2022Electrocardiogram}
            
            \item T. Esteves, \underline{J. R. Pinto}, P. M. Ferreira, P. Costa, L. A. Rodrigues, I. Antunes, G. Lopes, P. Gamito, A. Abrantes, P. M. Jorge, A. Lourenço, A. F. Sequeira, J. S. Cardoso, and A. Rebelo, ``AUTOMOTIVE: A case study on AUTOmatic multiMOdal drowsiness detecTIon for smart VEhicles,'' \emph{IEEE Access}, 9: 153678--153700, 2021.~\cite{Esteves2021AUTOMOTIVE}
            
            \item A. F. Sequeira, T. Gonçalves, W. Silva, \underline{J. R. Pinto}, and J. S. Cardoso, ``An Exploratory Study of Interpretability for Face Presentation Attack Detection,'' \emph{IET Biometrics}, 10 (4): 441--455, 2021.~\cite{Sequeira2021Exploratory}
            
            \item \underline{J. R. Pinto}, M. V. Correia, and J. S. Cardoso, ``Secure Triplet Loss: Achieving Cancelability and Non-Linkability in End-to-End Deep Biometrics,'' \emph{IEEE Transactions on Biometrics, Behavior, and Identity Science}, 3 (2): 180--189, 2021.~\cite{Pinto2021Secure}
            
            \item \underline{J. R. Pinto}, J. S. Cardoso, and A. Lourenço, ``Evolution, Current Challenges, and Future Possibilities in ECG Biometrics,'' \emph{IEEE Access}, 6: 34746--34776, 2018.~\cite{Pinto2018}
        \end{etaremune}
        
    \item Articles in international conference proceedings:
        \begin{etaremune}
            \item \underline{J. R. Pinto}, P. Carvalho, C. Pinto, A. Sousa, L. G. Capozzi, and J. S. Cardoso, ``Streamlining Action Recognition in Autonomous Shared Vehicles with an Audiovisual Cascade Strategy,'' in \emph{17th International Conference on Computer Vision Theory and Applications (VISAPP)}, Feb.~2022.~\cite{Pinto2022Streamlining}
            
            \item P. C. Neto, F. Boutros, \underline{J. R. Pinto}, N. Damer, A. F. Sequeira, J. S. Cardoso, ``FocusFace: Multi-task Contrastive Learning for Masked Face Recognition,'' in \emph{Workshop on Face and Gesture Analysis for COVID-19 (FG4COVID19)}, Dec.~2021.~\cite{Neto2021Focus}
            
            \item P. C. Neto, F. Boutros, \underline{J. R. Pinto}, M. Saffari, N. Damer, A. F. Sequeira, and J. S. Cardoso, ``My Eyes Are Up Here: Promoting Focus on Uncovered Regions in Masked Face Recognition,'' in \emph{International Conference of the Biometrics Special Interest Group (BIOSIG 2021)}, Sep.~2021.~\cite{Neto2021Eyes}
            
            \item F. Boutros, N. Damer, J. Kolf, K. Raja, F. Kirchbuchner, R. Ramachandra, A. Kuijper, P.~Fang, C. Zhang, F. Wang, D. M. Martin, N. Aginako, B. Sierra, M. Nieto, M. E. Erakin, U. Demir, H. Ekenel, A. Kataoka, K. Ichikawa, S. Kubo, J. Zhang, M. He, D. Han, S. Shan, K. Grm, V. Struc, S. Seneviratne, N. Kasthuriarachchi, S. Rasnayaka, P. C. Neto, A. F. Sequeira, \underline{J. R. Pinto}, M. Saffari, and J. S. Cardoso, ``MFR 2021: Masked Face Recognition Competition,'' in \emph{International Joint Conference on Biometrics (IJCB 2021)}, Aug.~2021.~\cite{Boutros2021MFR}
            
            \item \underline{J. R. Pinto}, T. Gonçalves, C. Pinto, L. Sanhudo, J. Fonseca, F. Gonçalves, P. Carvalho, and J. S. Cardoso, ``Audiovisual Classification of Group Emotion Valence Using Activity Recognition Networks,'' in \emph{Fourth IEEE International Conference on Image Processing, Applications and Systems (IPAS 2020)}, Dec.~2020.~\cite{Pinto2020Audiovisual}
            
            \item \underline{J. R. Pinto} and J. S. Cardoso, ``Explaining ECG Biometrics: Is It All In The QRS?,'' in \emph{International Conference of the Biometrics Special Interest Group (BIOSIG 2020)}, Sep.~2020.~\cite{Pinto2020Explaining}
            
            \item \underline{J. R. Pinto} and J. S. Cardoso, ``Self-Learning with Stochastic Triplet Loss,'' in \emph{International Joint Conference on Neural Networks (IJCNN 2020)}, Jul.~2020.~\cite{Pinto2020SelfLearning}
            
            \item \underline{J. R. Pinto}, J. S. Cardoso, and M. V. Correia, ``Secure Triplet Loss for End-to-End Deep Biometrics,'' in \emph{8th International Workshop on Biometrics and Forensics (IWBF 2020)}, Apr.~2020.~\cite{Pinto2020iwbf}
            
            \item A. F. Sequeira, W. Silva, \underline{J. R. Pinto}, T. Gonçalves, and J. S. Cardoso, ``Interpretable Biometrics: Should We Rethink How Presentation Attack Detection is Evaluated?,'' in \emph{8th International Workshop on Biometrics and Forensics (IWBF 2020)}, Apr.~2020.~\cite{Sequeira2020Interpretable}
            
            \item \underline{J. R. Pinto} and J. S. Cardoso, ``An End-to-End Convolutional Neural Network for ECG-Based Biometric Authentication,'' in \emph{10th IEEE International Conference on
            Biometrics: Theory, Applications and Systems (BTAS 2019)}, Sep.~2019.~\cite{Pinto2019b}
            
            \item G. Lopes, \underline{J. R. Pinto}, and J. S. Cardoso, ``Don't You Forget About Me: A Study on Long-Term Performance in ECG Biometrics,'' in \emph{IbPRIA 2019: 9th Iberian Conference on Pattern Recognition and Image Analysis}, Jul.~2019.~\cite{Lopes2019}
        \end{etaremune}
        
    \item Short papers presented in international conferences:
        \begin{etaremune}
            \item S. Beco, \underline{J. R. Pinto}, and J. S. Cardoso, ``Interlead Conversion of Single-Lead Blindly-Segmented Electrocardiogram Signals,'' in \emph{17th International Conference on Computational Intelligence Methods for Bioinformatics and Biostatistics (CIBB 2021)}, Nov.~2021.~\cite{Beco2021Interlead}
        \end{etaremune}
    
    \item Chapters in books:
        \begin{etaremune}
            \item \underline{J. R. Pinto}, J. S. Cardoso, and A. Lourenço, ``Deep Neural Networks for Biometric Identification Based on Non-Intrusive ECG Acquisitions,'' in K. V. Arya and R.~S.~Bhadoria, Eds., \emph{The Biometric Computing: Recognition and Registration}, CRC Press, 2019.~\cite{Pinto2019Deep}
        \end{etaremune}
        
    \item Encyclopaedia entries:
        \begin{etaremune}
            \item \underline{J. R. Pinto} and J. S. Cardoso, ``ECG Biometrics,'' in S. Jajodia, P. Samarati, and M. Yung, Eds., \emph{Encyclopedia of Cryptography, Security and Privacy}, Springer, 2021.~\cite{Pinto2021Encyclopedia}
        \end{etaremune}
    
    \item Abstracts in national conference proceedings:
        \begin{etaremune}
            \item \underline{J. R. Pinto} and J. S. Cardoso, ``xECG: Using Interpretability to Understand Deep ECG Biometrics,'' in \emph{27th Portuguese Conference on Pattern Recognition (RECPAD 2021)}, Nov.~2021.
            
            \item \underline{J. R. Pinto}, M. V. Correia, and J. S. Cardoso, ``Achieving Cancellability in End-to-End Deep Biometrics with the Secure Triplet Loss,'' in \emph{26th Portuguese Conference on Pattern Recognition (RECPAD 2020)}, Oct.~2020.
            
            \item W. Silva, \underline{J. R. Pinto}, T. Gonçalves, A. F. Sequeira, and Jaime S. Cardoso, ``Explainable Artificial Intelligence for Face Presentation Attack Detection,'' in \emph{26th Portuguese Conference on Pattern Recognition (RECPAD 2020)}, Oct.~2020.
            
            \item G. Lopes, \underline{J. R. Pinto}, J. S. Cardoso, and A. Rebelo, ``Long-Term Performance of a Convolutional Neural Network for ECG-Based Biometrics,'' in \emph{25th Portuguese Conference on Pattern Recognition (RECPAD 2019)}, Oct.~2019.
            
            \item \underline{J. R. Pinto}, J. S. Cardoso, and A. Lourenço, ``Improving ECG-Based Biometric Identification Using End-to-End Convolutional Networks,'' in \emph{24th Portuguese Conference on Pattern Recognition (RECPAD 2018)}, Oct.~2018.
        \end{etaremune}
\end{itemize}

Beyond the aforementioned publications, the author has contributed to fourteen other scientific publications related to diverse pattern recognition and computer vision research topics not covered in this thesis. These are listed below:

\begin{itemize}
    \item Articles in journals:
        \begin{etaremune}
            \item P. C. Neto, T. Gonçalves, \underline{J. R. Pinto}, W. Silva, A. F. Sequeira, A. Ross, and J. S. Cardoso, ``Explainable Biometrics in the Age of Deep Learning,'' \emph{ACM Computing Surveys}, 2022.~\cite{Neto2022Explainable}~\emph{(submitted)} 
            
            \item L. G. Capozzi, V. Barbosa, C. Pinto, \underline{J. R. Pinto}, A. Pereira, P. M. Carvalho, and J. S. Cardoso, ``Toward Vehicle Occupant-Invariant Models for Activity Characterization,'' \emph{IEEE Access}, 10: 104215--104225, 2022.~\cite{Capozzi2022Actor}
            
            \item P. C. Neto, \underline{J. R. Pinto}, F. Boutros, N. Damer, A. F. Sequeira, and J. S. Cardoso, ``Beyond Masks: On the Generalization of Masked Face Recognition Models to Occluded Face Recognition,'' \emph{IEEE Access}, 10: 86222--86233, 2022.~\cite{Neto2022Beyond}
            
            \item S. P. Oliveira, \underline{J. R. Pinto}, T. Gonçalves, R. C. Marques, M. J. Cardoso, H. P. Oliveira, and J. S. Cardoso, ``Weakly-Supervised Classification of HER2 Expression in Breast Cancer Haematoxylin and Eosin Stained Slides,'' \emph{Applied Sciences}, 10 (14): 4728, 2020.~\cite{Oliveira2020Weakly}
        \end{etaremune}
        
    \item Articles in international conference proceedings:
        \begin{etaremune}
            \item P. C. Neto, F. Boutros, \underline{J. R. Pinto}, N. Damer, A. F. Sequeira, J. S. Cardoso, M. Bengherabi, A. Bousnat, S. Boucheta, N. Hebbadj, B. Yahya-Zoubir, M. E. Erakın, U. Demir, H. K. Ekenel, P. B. Q. Vidal, and D. Menotti, ``IJCB OCFR 2022: Competition on Occluded Face Recognition From Synthetically Generated Structure-Aware Occlusions,'' in \emph{International Joint Conference on Biometrics (IJCB 2022)}, Oct.~2022. \emph{(in press)}~\cite{Neto2022OCFR}
            
            \item L. G. Capozzi, P. Carvalho, A. Sousa, C. Pinto, \underline{J. R. Pinto}, and J. S. Cardoso, ``Impact of visual noise in activity recognition using deep neural networks - an experimental approach,'' in \emph{2nd International Conference on Pattern Recognition and Machine Learning (PRML 2021)}, Jul.~2021. \cite{Capozzi2021Impact}
            
            \item L. G. Capozzi, \underline{J. R. Pinto}, J. S. Cardoso, and A. Rebelo, ``End-to-End Deep Sketch-to-Photo Matching Enforcing Realistic Photo Generation,'' in \emph{25th Iberoamerican Congress on Pattern Recognition (CIARP'21)}, May~2021. \cite{Capozzi2021EndToEnd}
            
            \item L. G. Capozzi, \underline{J. R. Pinto}, J. S. Cardoso, and A. Rebelo, ``Optimizing Person Re-Identification using Generated Attention Masks,'' in \emph{25th Iberoamerican Congress on Pattern Recognition (CIARP'21)}, May~2021. \cite{Capozzi2021Optimizing}
            
            \item A. Matta, \underline{J. R. Pinto}, and J. S. Cardoso, ``Mixture-Based Open World Face Recognition,'' in \emph{9th World Conference on Information Systems and Technologies (WorldCIST'21)}, Apr.~2021. \cite{Matta2021}

            \item W. Silva, \underline{J. R. Pinto}, and J. S. Cardoso, ``A Uniform Performance Index for Ordinal Classification with Imbalanced Classes,'' in \emph{International Joint Conference on Neural Networks (IJCNN 2018)}, Jul.~2018.~\cite{Silva2018Uniform}
        \end{etaremune}
    
    \item Abstracts in national conference proceedings:
        \begin{etaremune}
            \item L. Capozzi, \underline{J. R. Pinto}, J. S. Cardoso, and A. Rebelo, ``Sketch-to-Photo Matching Enforcing Realistic Rendering Generation,'' in \emph{27th Portuguese Conference on Pattern Recognition (RECPAD 2021)}, Nov.~2021.
            
            \item S. P. Oliveira, \underline{J. R. Pinto}, T. Gonçalves, H. P. Oliveira, and Jaime S. Cardoso, ``IHC Classification in Breast Cancer H\&E Slides with a Weakly-Supervised Approach,'' in \emph{26th Portuguese Conference on Pattern Recognition (RECPAD 2020)}, Oct.~2020.
            
            \item P. Costa, P. Silva, \underline{J. R. Pinto}, A. F. Sequeira, and A. Rebelo, ``Face Anti Spoofing: Handcrafted and Learned Features for Face Liveness Detection,'' in \emph{25th Portuguese Conference on Pattern Recognition (RECPAD 2019)}, Oct.~2019.
            
            \item \underline{J. R. Pinto} and J. S. Cardoso, ``Fine Segmentation of Head and Torso Using Label Refinement Networks,'' in \emph{25th Portuguese Conference on Pattern Recognition (RECPAD 2019)}, Oct.~2019.
        \end{etaremune}
\end{itemize}

The research conducted during these doctoral studies has also been partially presented to the scientific community at the Doctoral Consortium of the \emph{2019 IEEE International Conference on Biometrics: Theory, Applications and Systems} (Tampa, FL, USA) and at the \emph{2020 International Summer School for Advanced Studies on Biometrics for Secure Authentication} (Alghero, Italy).

\section{Collaborations}

The doctoral work presented in this thesis included close collaborations with researchers from several institutions within the AUTOMOTIVE, Easy Ride, and Aurora projects. The author also collaborated frequently with VCMI group colleagues within their research work and master dissertations related to diverse biometrics, pattern recognition, and computer vision topics, as described below.

\subsection{Research projects}

The AUTOMOTIVE project\footnote{AUTOMOTIVE (``POCI-01-0145-FEDER-030707'') was financed by the European Regional Development Fund (ERDF) through the Operational Programme for Competitiveness and Internationalisation (COMPETE 2020 Programme), and by national funds through the Portuguese funding agency, Fundação para a Ciência e a Tecnologia (FCT).} was focused on ushering in the next generation of driver drowsiness monitoring technologies. Led by the VCMI research group at INESC TEC, this project featured the participation of CardioID Technologies, Instituto Superior de Engenharia de Lisboa (ISEL), and Universidade Lusófona de Humanidades e Tecnologias (ULHT). The author of this thesis has participated in the AUTOMOTIVE project from June 2019 to its conclusion in November 2021. He mainly contributed to the development of novel algorithms for ECG-based biometric recognition to enable personalised drowsiness models. The results of this project are nicely summed up in~\cite{Esteves2021AUTOMOTIVE}.

The Easy Ride project\footnote{Easy Ride (``POCI-01-0247-FEDER-039334'') was supported by European Structural and Investment Funds in the FEDER component, through the Operational Competitiveness and Internationalisation Programme (COMPETE 2020).}, under the motto ``Experience is Everything'', was a large effort led by Bosch Car Multimedia and Universidade do Minho aiming to improve passenger comfort and safety in autonomous shared vehicles. It featured INESC TEC's participation in the SP5 sub-project, in close cooperation with Bosch Car Multimedia, which focused on occupant emotional monitoring. The author of this thesis has participated in writing this project's proposal, and then in the research work from its start in February 2020 to its conclusion in December 2021. He has mainly contributed to the design, implementation, and evaluation of efficient multimodal deep learning algorithms using audio and RGB video for in-vehicle emotion, activity, and violence recognition. The work encompassed all stages of research and development from scratch to in-vehicle deployment and is presented in~\cite{Pinto2020Audiovisual, Pinto2022Streamlining}.

The Aurora project inherited Easy Ride's goals of bringing forth the future of shared autonomous vehicles. Specifically, it brought together Bosch Car Multimedia and INESC TEC's Centre for Telecommunications and Multimedia (CTM) and the High-Assurance Software Laboratory (HASLab) to continue SP5's mission of contributing towards accurate and efficient occupant emotional and activity monitoring. The author of this thesis participated in this project since its beginning in December 2021, having collaborated on the definition of the project's objectives and task planning.

\subsection{Synergies within the research group}

Within the VCMI research group, the author collaborated, in 2018, with Wilson Silva to develop a metric for ordinal classification which takes into account both accuracy and label ranking while retaining robustness to class imbalance~\cite{Silva2018Uniform}. In 2020, the author collaborated with Sara P. Oliveira, Tiago Gonçalves, and Hélder Oliveira, alongside Rita C. Marques and Maria João Cardoso at Champalimaud Foundation (Lisboa, Portugal), on a weakly-supervised methodology based on multiple instance learning to classify HER2 expression in breast cancer histology slides~\cite{Oliveira2020Weakly}.

Since 2020, the author has also been collaborating with Ana F. Sequeira, Wilson Silva, Tiago Gonçalves, and Pedro C. Neto, alongside Arun Ross from Michigan State University (East Lansing, MI, USA) to study biometrics from the perspective of interpretability, rethinking how model accuracy should be measured, and calling for more transparent biometric algorithms~\cite{Sequeira2020Interpretable, Sequeira2021Exploratory, Neto2022Explainable}.

In 2021, the author collaborated with Leonardo G. Capozzi and Ana Rebelo in their work related to person re-identification and scene geolocation for automatic missing person searching~\cite{Capozzi2021Optimizing, Capozzi2021EndToEnd}. Also in 2021 and 2022, the author has collaborated with Pedro C. Neto, Mohsen Saffari, and Ana F. Sequeira, alongside Fadi Boutros and Naser Damer at Fraunhofer IGD (Darmstadt, Germany), on novel strategies for masked face recognition to uphold state-of-the-art biometric performance amidst a global pandemic~\cite{Boutros2021MFR, Neto2021Eyes, Neto2021Focus}.

\subsection{Organisation of scientific events}

This doctoral project and the aforementioned collaborations and synergies motivated the contribution to the organisation of multiple scientific events. 

In regards to biometrics, the main example is that of the 2020 edition of the \emph{International Workshop on Biometrics and Forensics (IWBF)}, organised by INESC TEC and NTNU, where the author of this thesis collaborated as Demo Chair. He has also helped organise the \emph{Workshop on Explainable \& Interpretable Artificial Intelligence for Biometrics (xAI4Biometrics)}, hosted yearly at the \emph{IEEE/CVF Winter Conference on Applications of Computer Vision (WACV)}, as Publicity Chair in 2021 and 2022 and Programme Committee member in 2023. 

The collaboration with Pedro C. Neto and Ana F. Sequeira from INESC TEC and Fadi Boutros and Naser Damer from Fraunhofer IGD was enhanced by the co-organisation of the \emph{Advanced Occluded Face Recognition (OCFR)} competition at the \emph{2022 International Joint Conference on Biometrics (IJCB)}~\cite{Neto2022OCFR}.

Aligned with the topic of wellbeing monitoring, and as an extension to the Easy Ride and Aurora projects, the author has also co-organised the \emph{In-Vehicle Sensing and Monitorization Workshop (ISM)}. This first edition of the workshop was hosted at the \emph{European Conference on Computer Vision (ECCV)} in October 2022.

In a broader scope, the author of this thesis has also co-organised the special session on \emph{Machine Learning in Healthcare Informatics and Medical Biology} at the \emph{International Conference on Computational Intelligence Methods for Bioinformatics and Biostatistics (CIBB)}, in 2019 and 2021. He was also a member of the Technical and Programme committee of this conference for its 2021 edition.

Since 2019, the author has also helped organise the \emph{VISUM Summer School} on computer vision and machine intelligence. In the 2019 and 2020 editions, he was a member of the project team, while in 2021 and 2022 he was part of the main organisation team.

\subsection{Supervision of dissertations and internships}

The author of this thesis has collaborated, as co-supervisor or external supervisor, on the following master dissertations related to his doctoral studies (in reverse chronological order):
\begin{etaremune}
    \item Mariana Silva Xavier (2022), ``Inside Out: Fusing ECG and Face Information to Recognise Emotions'', Master in Bioengineering, Universidade do Porto - as co-supervisor, alongside Jaime S. Cardoso (supervisor)~\cite{Xavier2022};

    \item Guilherme Augusto Tiritan Romano Barbosa (2022), ``Going 2D: Exploring Learnable Bidimensional Approaches for ECG Biometrics'', Master in Bioengineering, Universidade do Porto - as co-supervisor, alongside Jaime S. Cardoso (supervisor)~\cite{Barbosa2022};

    \item Pedro Duarte da Cunha Nunes Lopes (2022), ``Deep Neural Networks for Face-based Emotion Recognition'', Master in Bioengineering, Universidade do Porto - as external institution supervisor, alongside Jaime S. Cardoso (supervisor) and Ana F. Sequeira (co-supervisor)~\cite{Lopes2022};
    
    \item Erfan Omidvar (2022), ``Single-Wrist Electrocardiogram Acquisition
    Application in Biometrics'', Master in Biomedical Engineering, Universidade do Porto - as second co-supervisor, alongside Miguel V. Correia (supervisor) and Duarte Dias (first co-supervisor)~\cite{Omidvar2022};
    
    \item Vítor Hugo Pereira Barbosa (2022), ``Robust occupant action classification in shared autonomous vehicles'', Master in Informatics and Computing Engineering, Universidade do Porto - as external institution supervisor, alongside Jaime S. Cardoso (supervisor) and Pedro Carvalho (co-supervisor)~\cite{Barbosa2022b};
    
    \item Telma Sofia Caldeira Esteves (2021), ``Sleepy Drivers: Drowsiness Monitoring Using ECG and Face Video'', Master in Biomedical Engineering, Universidade Nova de Lisboa - as external institution supervisor, alongside Ricardo Vigário (supervisor), André Lourenço (co-supervisor), and Ana Rebelo (external institution supervisor)~\cite{Esteves2021};
    
    \item Sofia Cardoso Beco (2021), ``Make My Heartbeat: Generation and Interlead Conversion of ECG Signals'', Master in Bioengineering, Universidade do Porto - as co-supervisor, alongside Jaime S. Cardoso (supervisor)~\cite{Beco2021};

    \item Inês Alexandra Teixeira Antunes de Magalhães (2021), ``Feel My Heart: Emotion Recognition Using the Electrocardiogram'', Master in Bioengineering, Universidade do Porto - as co-supervisor, alongside Jaime S. Cardoso (supervisor)~\cite{Antunes2021};

    \item Arthur Johas Matta (2020), ``Open-World Face Recognition'', Master in Informatics and Computing Engineering, Universidade do Porto - as co-supervisor, alongside Jaime S. Cardoso (supervisor)~\cite{Matta2020};

    \item Leonardo Gomes Capozzi (2020), ``Face Recognition For Forensic Applications: Methods for Matching Facial Sketches to Mugshot Pictures'', Master in Informatics and Computing Engineering, Universidade do Porto - as co-supervisor, alongside Ana Rebelo (supervisor)~\cite{Capozzi2020};

    \item João Manuel Guedes Ferreira (2020), ``Head Pose Estimation for Facial Biometric Recognition Systems'', Master in Informatics and Computing Engineering, Universidade do Porto - as co-supervisor, alongside Ana F. Sequeira (supervisor) and Jaime S. Cardoso (co-supervisor)~\cite{Ferreira2020};

    \item Carolina Martins Barbosa Rodrigues Afonso (2020), ``Changing Perspectives: Interlead Conversion in Electrocardiographic Signals'', Master in Network and Information Systems Engineering, Universidade do Porto - as co-supervisor, alongside Miguel Coimbra (supervisor)~\cite{Afonso2020};

    \item Gabriel Carneiro Lopes (2019), ``Don't You Forget About Me: Enhancing Long Term Performance in Electrocardiogram Biometrics'', Master in Bioengineering, Universidade do Porto - as co-supervisor, alongside Jaime S. Cardoso (supervisor)~\cite{Lopes2019thesis}.
\end{etaremune}

Beyond these aforementioned dissertations, the author of this thesis also collaborated in the supervision of more than twenty students on curricular, extra-curricular, and summer internships related to biometrics, pattern recognition, and computer vision topics.

\section{Awards and Distinctions}

Beyond the aforelisted peer-reviewed publications and presentations to the scientific community, the doctoral research work presented in this dissertation has also been the recipient of multiple awards and distinctions. These are listed below:

\begin{itemize}
    \item The author of this thesis was granted the \emph{EAB Max Snijder Award} at the \emph{European Biometrics Awards 2022} organised by the \emph{European Association for Biometrics (EAB)}. This award recognised the wider perspective and applicability of his work on ECG biometrics, which contributed towards a complete deep learning solution encompassing end-to-end models, learnable template security, and explainability;
    
    \item The initial work on the Secure Triplet Loss~\cite{Pinto2020iwbf}, focused on biometric template cancelability for end-to-end deep models, received the \emph{Computers Journal Best Paper Award} at the \emph{2020 International Workshop on Biometrics and Forensics (IWBF)};

    \item The work on audiovisual group emotion valence recognition~\cite{Pinto2020Audiovisual} received the \emph{Best Session Paper Award} at the \emph{2020 IEEE International Conference on Image Processing, Applications and Systems (IPAS)};
    
    \item The extended work on the Secure Triplet Loss~\cite{Pinto2021Secure}, presented at the \emph{2021 NIS Workshop} organised by INESC TEC's Networked Intelligent Systems cluster, received the \emph{Best Presentation Award} by the official jury.
\end{itemize}

\section{Document Structure}

This thesis is composed of six parts. Part~\ref{part:prologue} is the prologue, which includes this introduction, Chapter~\ref{ch:intro}, and offers an overview of the fundamental concepts related to biometric systems, biometric traits, and wellbeing monitoring in Chapter~\ref{ch:fundamentals}.

Part~\ref{part:ecgBiometrics} focuses on electrocardiogram biometrics, presenting the contributions to this topic produced during the doctoral work. It begins with an overview of the existing data, related literature methodologies, and a discussion of open challenges and opportunities, in Chapter~\ref{ch:ecgprior}. Chapter~\ref{ch:ecgiden} presents our study on end-to-end deep learning for ECG-based identification, including tailored unidimensional data augmentation strategies. Chapter~\ref{ch:ecgauth} showcases the work on triplet loss and transfer learning for ECG-based identity verification. A study on long-term performance and template update for identification is presented in Chapter~\ref{ch:ecglongterm}. Chapter~\ref{ch:ecgexpl} delves into the topic of explainability for ECG biometrics, aiming to better understand which parts of the signal are best for identification. Finally, Chapter~\ref{ch:ecginterlead} proposes a methodology for recovering missing ECG leads based on blindly-segmented single-lead segments.

Part~\ref{part:faceBiometrics} deals with face biometrics. Chapter~\ref{ch:faceprior} offers an overview of the data, existing methodologies, and open challenges in this topic. Two methodologies to close the performance gap in masked face recognition are presented in Chapter~\ref{ch:maskedFaceRecog}. Chapter~\ref{ch:interpFacePAD} describes a study on interpretability to understand the decisions of deep learning models in face presentation attack detection and motivate a more widespread usage of interpretability for more transparent biometrics.

Part~\ref{part:wellbeingMonitoring} is centred on wellbeing monitoring and covers the topics of emotion recognition, activity recognition, and violence detection.
Chapter~\ref{ch:emotion} describes an audiovisual approach developed to classify emotion valence in groups of people. Chapter~\ref{ch:activity} presents an adaptation of the aforementioned approach for activity recognition and violence detection, alongside a cascade strategy for increased efficiency in in-vehicle scenarios.

Part~\ref{part:broaderTopics} covers broader topics related to biometrics and pattern recognition which have been addressed during the doctoral work. Specifically, Chapter~\ref{ch:secure} introduces the Secure Triplet Loss, a novel approach to ensure biometric template security on end-to-end deep learning models. Lastly, a methodology for self-supervised learning formulated for minimal performance gaps when using sequential data is presented in Chapter~\ref{ch:selfsupervised}.

Part~\ref{part:epilogue} is the epilogue, which concludes this thesis. It includes an overview of the conducted work and the conclusions drawn from it, in Chapter~\ref{ch:summary}. Chapter~\ref{ch:future} offers a discussion on future work opportunities related to the results of this doctoral thesis.
\chapter{Fundamental Concepts}\label{ch:fundamentals}

Biometric systems are, in several ways, different from other pattern recognition applications. The need for storage of personal data from users and the different modes on which the systems can operate are only some of the special characteristics of biometric systems. When developing one, one should be aware of these specificities to ensure the best performance and robustness.

Hence, this chapter presents the fundamental concepts needed to build a biometric system, either for identity recognition or the monitoring of wellbeing parameters. It includes an overview of the general structure and operation of biometric systems, their security vulnerabilities, the different biometric traits (with a special focus on the electrocardiogram and face), and the metrics for thorough performance evaluations.

\section{Biometric Systems}

\subsection{General structure}

\subsubsection{Biometric recognition systems}

Biometric recognition systems are tools that use hardware and pattern recognition algorithms to compare the identity of a user with that of registered individuals based on their attributes (designated as \emph{biometric characteristics} or \emph{traits}). Like traditional identification systems based on keys, cards, or codes, biometric systems are mostly used for access control to restricted places, confidential information, or personal data and belongings~\cite{Dar2015b}.

\begin{figure}
\centering
\includegraphics[width=0.8\linewidth]{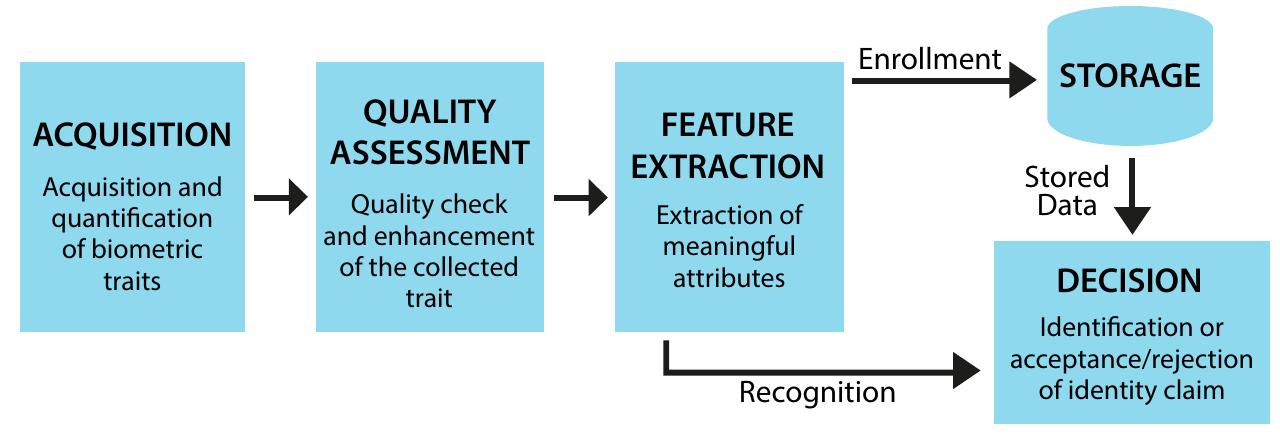}
\caption[General structure of a biometric recognition system.]{General structure of a biometric recognition system (from \cite{Pinto2018}, based on \cite{Prabhakar2003, Bolle2004, Jain2011}).}
\label{fig:modules}
\end{figure}

A biometric system is typically composed of an acquisition module, a storage module, and a biometric algorithm. The algorithm can, in turn, be divided into three modules: quality assessment, feature extraction, and decision (see Fig. \ref{fig:modules})~\citep{Bolle2004, Jain2011}. These modules are described below:
\begin{itemize}
\item \emph{Acquisition}: The acquisition module is the interface between the system and the subject and is responsible for the measurement of the biometric characteristic. The sensors used in this module should be carefully designed to fit the expected application settings and avoid, as much as possible, the noise and artefacts from environmental interference; 
\item \emph{Quality Assessment}: This module aims to evaluate the quality of the trait measurement and either accept it in its current form, enhance it to reduce noise and variability effects, or discard it if the quality is unacceptably low;
\item \emph{Feature Extraction}: The feature extraction module is focused on the processing of the acquired measurements, using pattern recognition tools, to extract the most meaningful attributes of the biometric trait and thus enable a robust decision. The feature extraction process should be designed to provide attributes that present high intersubject discrimination power and low intrasubject variability;
\item \emph{Decision}: This module uses the output from the feature extraction module and the stored information from registered users to identify the user, or validate or reject their identity claim. To achieve this, it compares the processed traits of the current user and the registered individuals;
\item \emph{Storage}: The storage module is typically composed of a database that stores \emph{biometric templates} (processed biometric trait measurements) from all individuals registered on the system. For security purposes, it can include template protection measures, such as hashing, to prevent leaks of sensitive personal information.
\end{itemize}

\subsubsection{Wellbeing monitoring systems}

Wellbeing monitoring systems are very diverse due to the variety of parameters these can monitor, which include emotions, stress, fatigue, and health conditions. However, most of these systems follow a general structure that is very similar to that of most biometric recognition systems.

Just like biometric recognition systems, an acquisition module performs the recording of data, which are checked and processed by the quality assessment module. After this, the feature extraction module extracts meaningful attributes from the trait measurements to enable accurate labelling by the decision module. 

Most wellbeing monitoring systems do not require a storage module, as biometric templates from the specific set of enrolled subjects will not be required, unlike in recognition systems. This is an advantage in terms of data security and privacy, which will be discussed later.

\subsection{Operation modes}

\begin{figure}[t!]
\begin{center}
\includegraphics[width=0.9\linewidth]{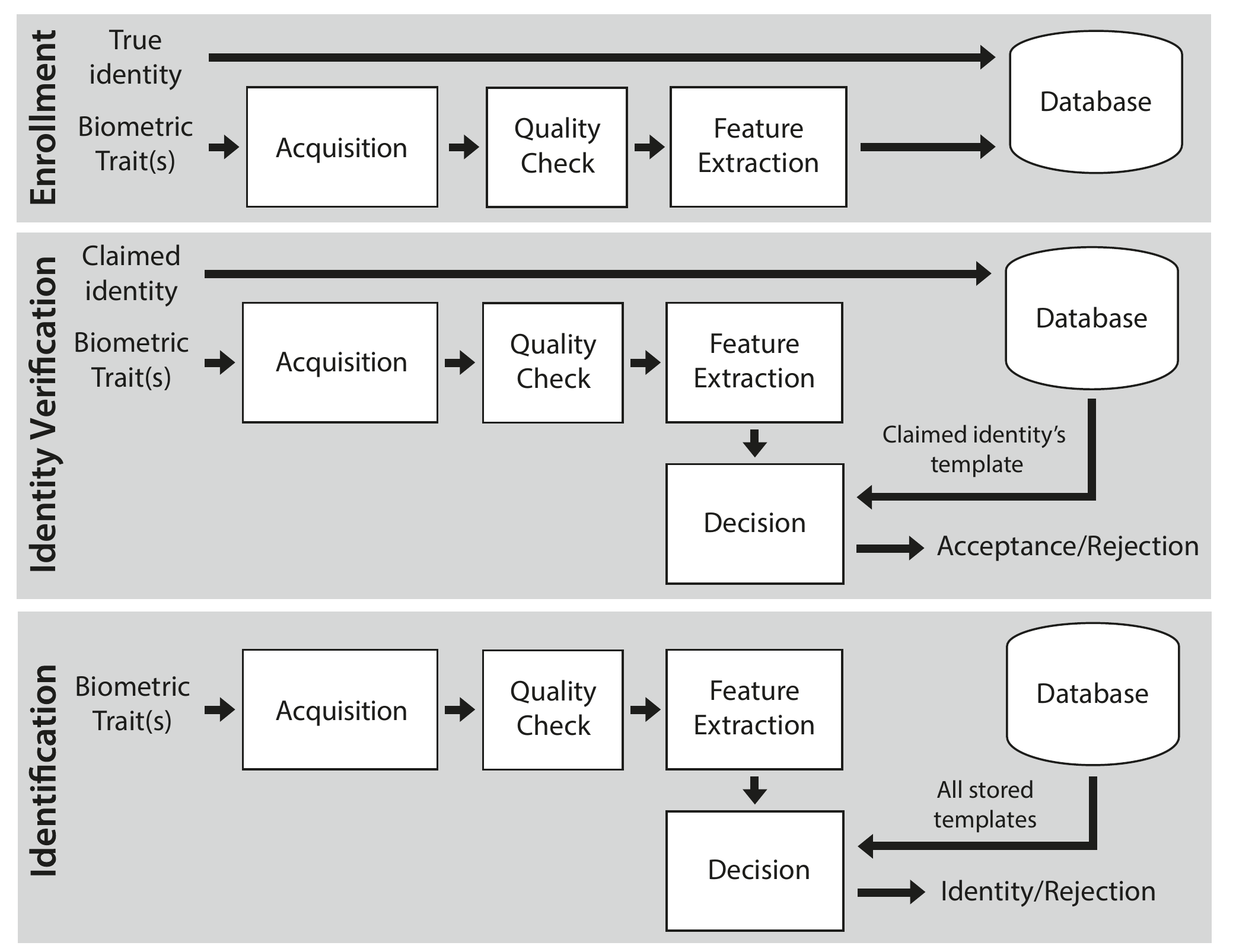}
\caption[Schematics of the operation of a biometric recognition system in identification and identity verification modes, and in the enrollment phase.]{Schematics of the operation of a biometric recognition system in identification and identity verification modes, and in the enrollment phase (adapted from~\citep{Pinto2017b}, based on \citep{Prabhakar2003,Sufi2010,Jain2011,Fratini2015}).}
\label{fig:modes}
\end{center}
\end{figure}

\subsubsection{Biometric recognition systems}

Depending on the application context and its requirements, a biometric system can either operate in \emph{identification} mode or \emph{identity verification} mode \citep{Prabhakar2003, Bolle2004, Agrafioti2011b, Jain2011, AboZahhad2014}.

In identity verification mode, also commonly called authentication, the biometric system will receive an identity claim along with the biometric measurement (the current user will claim to be a specific enrolled individual). Hence, the decision module will only compare the current measurement with the stored data from the claimed identity, performing a \emph{one-vs-one} comparison, and either accept or reject the claim.

In the identification mode, the biometric system will only receive the biometric measurement. Thus, the decision module performs a \emph{one-vs-all} comparison between the current biometric measurement and the data stored for each enrolled individual. Ultimately, the system will either assign one of the enrolled identities (corresponding to the strongest comparison) to the current user or reject to identify (if no comparison was strong enough).

Additionally, biometric systems include the \emph{enrollment} phase, which comprises the acquisition, processing, and storage of a biometric template of a subject for its registration on the system. After this, the system will be able to correctly perform identification or identity verification when used by the subject \citep{Prabhakar2003, Agrafioti2011b}.

\subsubsection{Wellbeing monitoring systems}

As aforementioned, wellbeing monitoring systems like emotion recognition devices rarely require the storage of personal information from the users. Hence, they dismiss enrollment phases and only operate in one mode: \emph{inference}. In this mode, a trait acquisition is performed by the system, which will output a corresponding label. The output is composed of discrete categories (\eg, sad, happy, or angry, in emotion recognition) or continuous scores (\eg, from very awake to very drowsy, in drowsiness recognition).  

\subsection{Security and privacy concerns}~\label{subsec:security}

\begin{figure}
    \centering
    \includegraphics[width=0.8\linewidth]{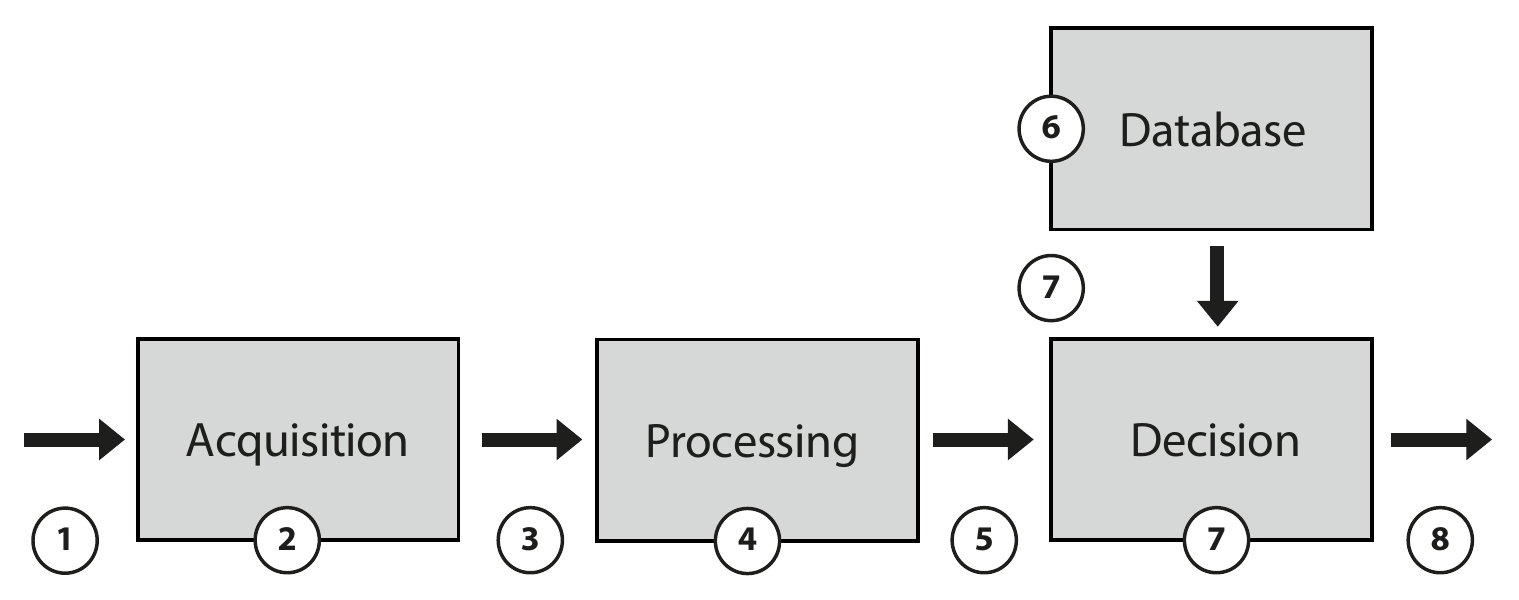}
    \caption[Attack points on a biometric system.]{Attack points on a biometric system (based on \citep{Galbally2007, Ratha2001}).}
    \label{fig:attacks}
\end{figure}

As key protectors of sensitive data, prized possessions, or restricted locations, biometric recognition systems are a prime target for attackers. The literature defines eight attack points (see Fig.~\ref{fig:attacks}) that sum up the different ways to unlawfully gain access to a biometric system~\cite{Mjaaland2011, Jain2015, Obied2006}. Considering a general structure (that groups the quality assessment and feature extraction into a single processing module) are described below:
\begin{itemize}
\item \emph{Type 1 -- At the acquisition module}: Such attacks are commonly called \emph{presentation attacks}, as they consist of physically forcing the biometric system to grant access to the attacker, either through the use of fake biometric traits (such as fake fingers, voice recordings, or prerecorded face videos) or even through the physical destruction of the system;
\item \emph{Type 2 -- Between the acquisition and the processing modules}: In these attacks, called \emph{replay attacks}, the attackers will target the communication link between the sensor and the processing module, to steal the trait acquisition. They can then bypass the sensor module by injecting the stolen trait measurements directly into the processing module;
\item \emph{Type 3 -- At the processing module}: The processing module can be attacked and overridden by another program, controlled by the attacker, that sends the desired features to the decision module upon request;
\item \emph{Type 4 -- Between the processing and decision modules}: These attacks are similar to replay attacks. The attacker will target the link between the processing module and the decision module and steal the features sent between them, to be later injected, bypassing the acquisition and processing modules;
\item \emph{Type 5 -- At the decision module}: Here, the attackers can replace the decision algorithms so they can generate high matching scores as requested, thus granting them access whenever desired, or to always output negative decisions, amounting to a denial-of-service attack;
\item \emph{Type 6 -- At the storage module}: This consists in exploiting database security flaws to add, modify, or delete templates, to ultimately grant access to unauthorised individuals or deny access to enrolled users;
\item \emph{Type 7 -- Between the storage and decision modules}: Here, the attackers intercept the communications between the database and the decision module, to steal biometric templates and replay them later;
\item \emph{Type 8 -- Between the decision module and the application}: These attacks consist in the manipulation of the data transmitted between the decision module and the application, \eg, to override a rejection decision.
\end{itemize}

In ECG-based biometrics, security vulnerabilities are still to be adequately addressed. Despite the pioneer studies of \citet{Eberz2017} and \citet{Karimian2017b}, no efforts have yet been devoted to better protect such systems. In general, biometric systems offer undeniable advantages when compared with traditional authentication systems, but this would be meaningless if the system introduced new vulnerabilities that paved the way to successful attacks. Hence, it is very important to remember the attack points presented above and their specificities throughout all stages of the development and deployment of a biometric system. 

Of all modules, storage is one of the most sensitive, as it stores personal data that could be used to unlawfully access private information and belongings. These intimate data are not specific to a single application and can be used by their legitimate owner as a single credential on several biometric systems (\eg, a user could use fingerprint-based access on two different computers). This is an obvious vulnerability as, just like using the same password for several online services, a single security failure can risk the privacy of the user on several applications \cite{Jain2008}.

Regardless of how sophisticated the database is, it can still be accessed or hacked by intruders who exert enough effort. Besides working towards more secure databases, it is paramount to prepare for possible successful attacks and ensure biometric templates cannot be retrieved in those cases \cite{Jain2008, Ignatenko2010}. Hence, regarding biometric template security, it is important to take into account the following factors:
\begin{itemize}
\item \emph{Non-Invertibility}: The processing module should be designed in a way that eases the creation of templates from biometric trait acquisitions. However, the retrieval of a close approximation of the original trait measurement or feature set, from a stored template, should be difficult and sufficiently time-consuming to render the process unfeasible or unattractive for attackers;
\item \emph{Revokability/Cancelability}: Keys can be changed when moving to a new house, and users can easily change e-mail passwords after they have become compromised. However, biometric systems rely on intrinsic personal characteristics, such as facial features or fingerprint minuti\ae, which are very difficult (or even impossible) to be changed. Hence, biometric systems should include measures that allow the easy invalidation of templates when these have become compromised. Thus, intruders will be denied access when using those credentials, but legitimate users will still be allowed to authenticate using their unchanged biometric trait;
\item \emph{Unlinkability}: For improved performance, a single biometric system can store, separately, more than one template for each user. The data protection scheme should ensure that the comparison between stored templates does not enable an attacker to cluster them by identity. This should also be difficult with templates from the same user in different biometric systems so that attackers are not able to attack multiple systems with a single stolen template~\cite{Ignatenko2010, Nandakumar2015}.
\end{itemize}

Besides these factors, the biometric system should also be able to deal with the characteristic variability of biometric traits and their measurements. Methods like hashing are commonly used for passwords, but such traditional credentials do not present variability. With biometric systems, small variations of the input should be considered normal and acceptable (\eg, with the face, different haircuts or beard styles), and should not influence the final decision. When designing a secure biometric system, it is necessary (although hard) to find an equilibrium between non-invertibility, unlinkability, and the controlled acceptance of natural intrasubject variability.

\section{Biometric Traits}

\subsection{General overview}

\begin{table}[!t]
\caption[Main benefits and drawbacks of different biometrics traits.]{Main benefits and drawbacks of different biometric traits (from~\citep{Pinto2018}, based on \citep{AboZahhad2014, Kaur2014}).}
\label{tab:traits_adavantages}
\centering
\renewcommand*{\arraystretch}{1.5}
\begin{tabular}{>{\raggedright}p{4cm} >{\raggedright}p{4.5cm} >{\raggedright}p{4.5cm}}
\textbf{Trait} & \textbf{Benefits} & \textbf{Drawbacks} \tabularnewline\hline
Electrocardiogram\\(ECG) & Universality\\Hidden nature\\Simple acquisition & Requires contact\\Variability over time \tabularnewline
Electroencephalogram\\(EEG) & Universality\\Hidden nature & Expensive equipment\\Vulnerability to noise\\Variability over time \tabularnewline
Face & Easily measurable\\Affordable equipment & Easy circumvention\\Depends on face visibility and lighting \tabularnewline
Fingerprint & High performance\\Permanent over time & Requires contact \tabularnewline
Gait & Easy to measure\\Affordable equipment & Low performance\\Variability over time\tabularnewline
Iris & High performance & Expensive equipment \tabularnewline
Palmprint & High measurability\\Permanent over time & Requires contact \tabularnewline
Photoplethysmogram (PPG) & Easy to acquire\\Hidden nature\\Affordable equipment & Low performance\\Variability over time \tabularnewline
Voice & Affordable equipment & Low performance\tabularnewline\hline
\end{tabular}
\end{table}

Biometric traits are human attributes that include enough personal information to reliably serve as the basis for the recognition and discrimination of individuals \citep{Agrafioti2011, Kaur2014}. According to the identity information they carry and the performance they can offer, biometric traits can either be considered \emph{hard traits}, strong enough to be standalone traits in a reliable biometric system, or \emph{soft traits}, which need further traits or information to offer acceptable recognition performance (see Table~\ref{tab:traits_adavantages}).

Traits can be categorised according to their nature, as \emph{anatomical}, \emph{physiological}, or \emph{be\-hav\-iour\-al} traits. Anatomical traits result from measurements of parts of the human body and include the fingerprints, the face, and the iris. Physiological traits are those that originate from physiological events in the body and include the heart rate, facial or hand thermography, the electrocardiogram (ECG), and the electroencephalogram (EEG). Behavioural traits originate from a person’s actions or behaviours, such as their gait (walking cadence), their signature, or their voice \citep{AboZahhad2014, Agrafioti2011}.

The quality of a biometric trait can be defined, as proposed by \citet{Jain1999}, through seven different aspects:
\begin{enumerate}
\item \emph{Universality}: the trait should be present in all subjects using the system;
\item \emph{Uniqueness}: the trait should include enough personal information to present differences between all subjects, and thus allow their identification;
\item \emph{Permanence}: despite the intersubject variability desired (uniqueness), the trait should be sufficiently stable over time (reduced intrasubject variability) to allow the identification through the comparison of measurements in different instances;
\item \emph{Measurability}: the trait should be easily and comfortably acquired and digitised, and its representation should allow easy processing and measurement;
\item \emph{Performance}: a system based on such a trait should meet or exceed the recognition accuracy requirements, set by the context in which it will be applied;
\item \emph{Acceptability}: there should be no foreseeable reservations that could make the subjects unwilling to allow the trait acquisition;
\item \emph{Circumvention}: the trait should be as hard as possible to mimic or counterfeit, in any way, to prevent spoofing of the biometric system.
\end{enumerate}

\citet{AboZahhad2014} compared sixteen biometric traits according to their compliance with each of the seven defined qualities. In that comparison, it is possible to verify that the traits with the lowest overall quality are the behavioural ones (gait, keystroke, signature, and voice) and phonocardiogram (heart sounds), with low performance, permanence, and distinctiveness. Low circumvention and universality are also downsides of behavioural traits. The traits with reported highest overall quality are the DNA, facial thermogram, fingerprint, iris, palm print, and ECG. Overall, the electrocardiogram excels in most factors, with just two `average’ scores (for collectability and acceptability).

Traits like fingerprint, face, signature, iris, and voice have been the most studied. However, these have long seen a quick growth and evolution of \emph{spoofing} methods (methods of counterfeiting a certain user’s trait to unlawfully gain access through the biometric system), which urges researchers to find more robust alternatives \citep{Kaur2014, Akhter2016}.

Throughout this doctoral work, the focus will be on the electrocardiogram (ECG), an emerging biometric trait that offers unique advantages regarding inherent liveness, anti-spoofing abilities, and wellbeing insight. Its performance and robustness drawbacks shall be mitigated through its fusion with face, a well-established and robust biometric trait. Hence, the next subsections consist of a presentation of both biometric traits, the ways to measure them, and the variability factors that provide them with identity and wellbeing information.

\subsection{Electrocardiogram}\label{subsec:ecg_fundamentals}

The electrocardiogram (ECG) is a physiological signal generated from the contraction and the recovery of the heart, that has been gaining traction as a biometric trait \citep{Pinto2018}. The heart has three main functions: generate blood pressure to keep blood circulating, route venous and arterial blood to the respective parts of the body, and regulate blood supply according to the metabolic demands \citep{Tate2009}. To do this, the heart needs to contract and relax its muscle, the myocardium, through the controlled generation and flow of depolarisation and repolarisation currents \citep{Scanlon2007,Tate2009}.

\begin{figure}[!t]
    \centering
    \includegraphics[width=0.6\linewidth]{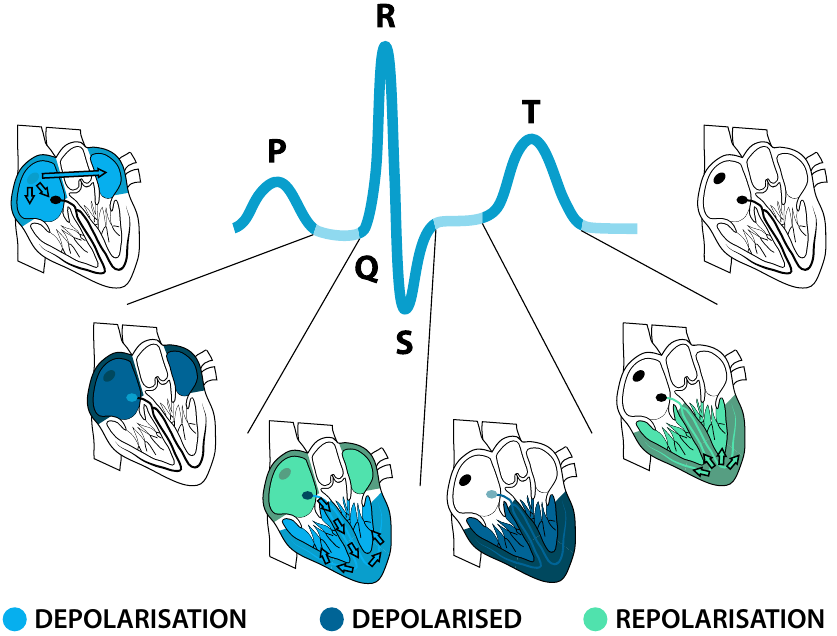}
    \caption[The sequence of depolarisation and depolarisation events in the heart, and their relationship with the different heartbeat waveforms in an ECG signal.]{The sequence of depolarisation and depolarisation events in the heart, and their relationship with the different heartbeat waveforms in an ECG signal (from \cite{Pinto2018}, based on \cite{Marieb2013}).}
    \label{fig:heart_conduction}
\end{figure}

The measurement of such currents using electrodes placed on the body is designated as electrocardiography and results in the electrocardiographic (ECG) signal. In normal conditions, the ECG is a cyclic repetition of five easily recognisable deflections: the P, Q, R, S, and T waves (see Fig. \ref{fig:heart_conduction}). A group of these deflections comprises a single heartbeat and each deflection can be traced back to the phase that originated it \cite{Scanlon2007,Tate2009,Marieb2013}.

\subsubsection{Acquisition}

The configurations used for the acquisition of ECG signals for biometric purposes have greatly evolved. From the first ECG-based biometric research works, considerable efforts have been devoted to more usable and comfortable acquisition technologies. This aims to place the ECG as a more attractive alternative to established biometric traits, mitigating the main disadvantage of the ECG, the obtrusive measurement techniques \citep{Pinto2018}.

\begin{figure}
    \centering
    \includegraphics[width=0.5\linewidth]{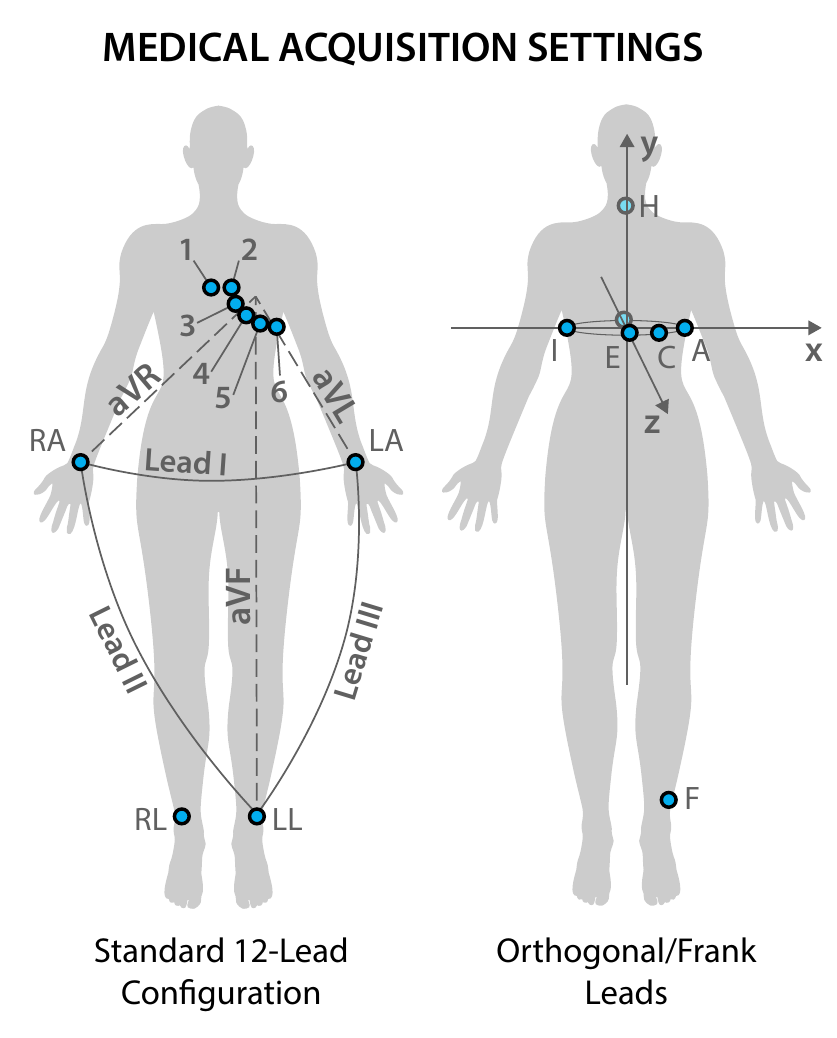}%
    \caption[Medical acquisition settings: electrode placement and leads on the standard 12-lead configuration and Frank leads.]{Medical acquisition settings: electrode placement and leads on the standard 12-lead configuration and Frank leads (from \cite{Pinto2018}, anterior electrodes depicted in blue, posterior electrodes depicted in lighter blue).}
    \label{fig:medical_acquisition}
\end{figure}

In early ECG-based biometrics research, recordings from the standard 12-lead or Frank leads were commonly used for the development and evaluation of algorithms \cite{Plataniotis2006,Wuebbeler2007,Ghofrani2010}. These are two defined and established configurations of electrodes for standardised and comparable ECG measurement (see Fig. \ref{fig:medical_acquisition}), widely used for the diagnosis of cardiac disorders. Authors frequently selected certain leads for their biometric algorithms, especially Lead I \citep{Palaniappan2004,Zhang2006,Molina2007} (because of its higher acceptability due to the electrode placement on the wrists), but also Lead II \citep{Kyoso2001,Kyoso2001b,Venkatesh2010,Pathoumvanh2014}, or chest leads \citep{Kyoso2000,Fang2009,Ye2010}.

\begin{figure}
    \centering
    \includegraphics[width=0.5\linewidth]{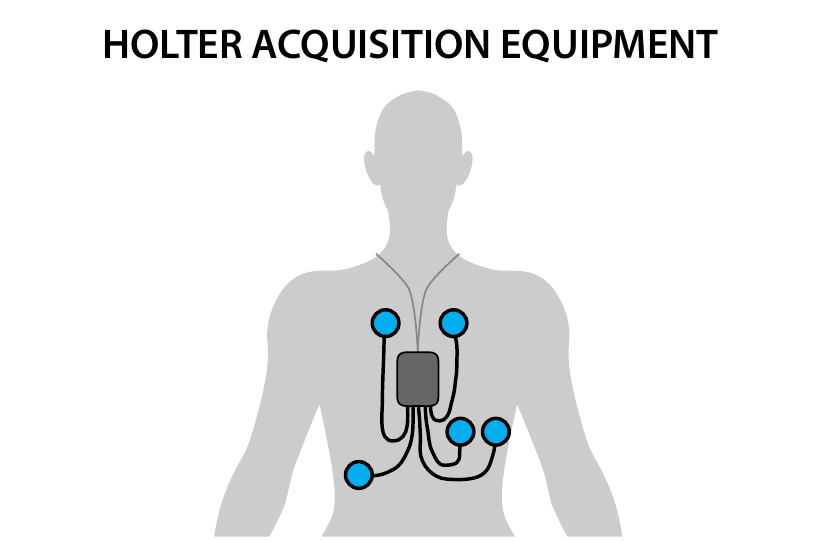}
    \caption[Acquisition settings with movement: example of a five-electrode Holter system for ambulatory recordings.]{Acquisition settings with movement: example of a five-electrode Holter system for ambulatory recordings (from \cite{Pinto2018}, electrodes depicted in blue).}
    \label{fig:holter_acquisition}
\end{figure}

Some researchers opted for acquisitions without movement restrictions and with fewer electrodes. Prominent choices in the literature include Holter systems (see Fig. \ref{fig:holter_acquisition}), which are prepared to acquire ECG signals for several hours while the subjects move and perform their daily activities. These were first used for ECG biometrics by \citet{Shen2002}, using ambulatory recordings from the MIT-BIH Normal Sinus Rhythm database, acquired for thirty minutes using Holter equipment. \citet{Labati2013,Labati2014} used 24-hour-long Holter acquisitions, from the E-HOL 24h signal collection, and seized the opportunity to study the effect of ECG variability over time on identification performance. Similarly, \citet{Zhou2014} used a mini-Holter system to continuously record ECG signals.

Medical and Holter systems are designated as \emph{on-the-person} acquisition settings. These present considerable drawbacks for biometric purposes, mainly concerning user comfort during acquisition due to the high number of electrodes and their placement on the chest and legs of the users. Although allowing for longer acquisitions with movement and activity, Holter acquisitions still require the placement of electrodes on the torso. This significantly reduces acquisition acceptability and comfort and damages the ECG strength as a biometric trait.

\begin{figure}
    \centering
    \includegraphics[width=0.5\linewidth]{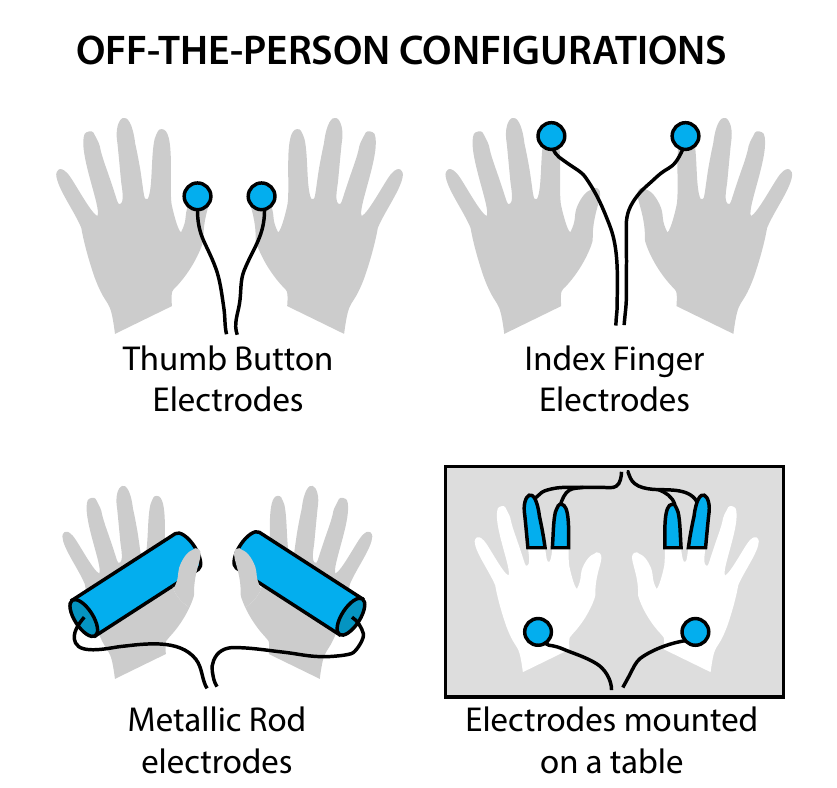}
    \caption[Examples of off-the-person ECG acquisition configurations, using thumb electrodes, index finger electrodes, metallic rods grabbed by the subjects, or electrodes mounted on a table.]{Examples of off-the-person ECG acquisition configurations, using thumb electrodes \cite{Chan2008}, index finger electrodes \cite{Matos2014}, metallic rods grabbed by the subjects \cite{Shen2011,Belgacem2012,Belgacem2013,Lin2014}, or electrodes mounted on a table \cite{Silva2014} (from \cite{Pinto2018}, electrodes depicted in blue).}
    \label{fig:offperson_acquisition}
\end{figure}

To improve acceptability and acquisition comfort, and get closer to biometric systems deployable in real settings, wet electrodes are being replaced by dry metallic electrodes, their number has been reduced to two or three, and their placement has been confined to the upper limbs, especially the on wrists, hands, or fingers (see Fig. \ref{fig:offperson_acquisition}). These acquisition configurations were designated as \emph{off-the-person} settings. The first research works in ECG biometrics to use such signals were, to the best of our knowledge, \citet{Molina2007}, who used commercial metallic electrodes strapped to the wrists of the subjects, and \citet{Chan2008}, who acquired ECG signals using dry button electrodes held by the subjects in contact with their thumbs.

Since then, ECG signals have been recorded using metallic rod electrodes~\cite{Shen2011,Belgacem2012,Belgacem2013,Lin2014}, and dry metallic electrodes mounted on plaques~\cite{Lourenco2011b} or attached to the users' fingers~\cite{Matos2014}, which offer increased comfort over on-the-person techniques. Nevertheless, off-the-person systems still require the user to hold the electrodes or deliberately place the fingers or palms over them. This prevents us from designating them as unconstrained systems, which puts the ECG at a disadvantage over other biometric traits. Besides this, the use of dry electrodes in farther placements makes the acquisition more vulnerable to interference, thus affecting the quality of the signal \cite{Belgacem2012,Silva2014}. 

\begin{figure}
    \centering
    \includegraphics[width=0.5\linewidth]{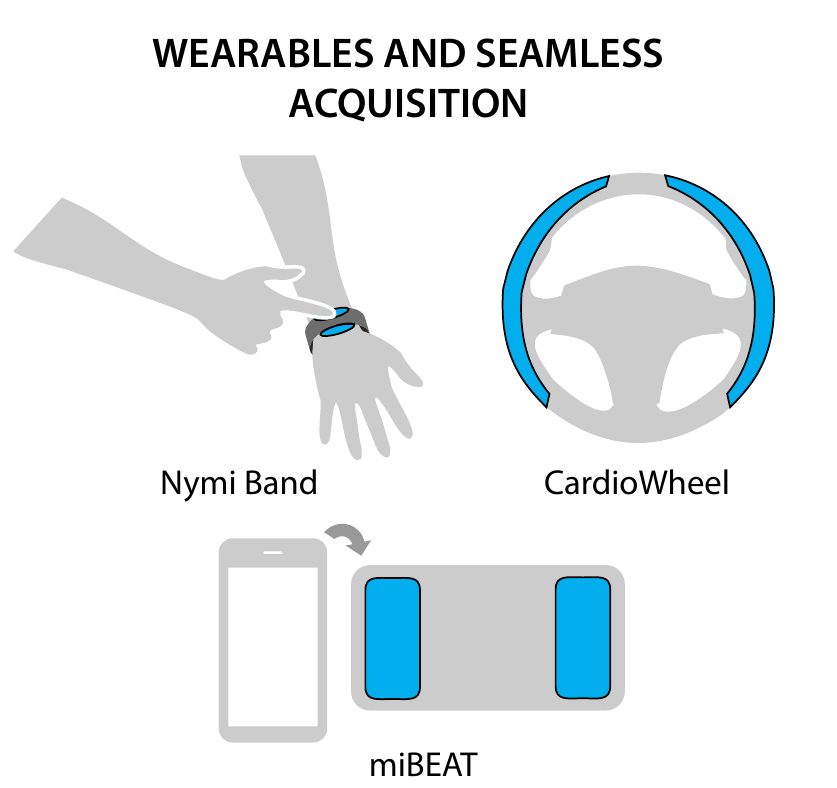}
    \caption[Wearable and seamless acquisition: examples of surveyed configurations.]{Wearable and seamless acquisition: examples of surveyed configurations (from \cite{Pinto2018}, electrodes depicted in blue).}
    \label{fig:seamless_acquisition}
\end{figure}

Recently, some researchers have tried to improve off-the-person configurations and approach unconstrained settings in ECG biometrics. They aimed to close the gap to real, commercial applications by developing wearable technologies for ECG acquisition or embedding the sensors into common objects (see Fig. \ref{fig:seamless_acquisition}). In research, the first example of this highly acceptable acquisition was a sensor pad to be used alongside a computer keyboard, to acquire ECG signals continuously during computer use \cite{Coutinho2010,Coutinho2011,Silva2013}. More recently, \citet{Zhang2017b} have shown it is possible to acquire ECG signals from a single arm and successfully used them for biometric recognition.

As for commercial applications, the Nymi Band \cite{Nymi2016}, a wearable wristband, acquires the ECG using two metallic electrodes on its inner and outer surfaces. Identity verification is performed when the band is put on and the session remains open until the band is taken off. While a session is open, the Nymi Band broadcasts an identity signal to authenticate the user in other nearby systems. The CardioWheel \cite{Lourenco2015} is a steering wheel cover that uses conductive leather for seamless and continuous biometric recognition and health monitoring of drivers, focused on automatic personalisation of driving settings and remote fleet supervision.

The miBEAT \cite{Yathav2017} is a versatile platform for the simultaneous acquisition of ECG and photoplethysmography (PPG) signals, which can be used for seamlessly integrated signal acquisition in smartphones or tablets. AliveCor provides a set of commercial solutions for easy ECG acquisition in the Kardia~\footnote{Kardia by AliveCor. Available on: \url{https://www.kardia.com/}.} lineup, including the KardiaMobile, to be used with typical smartphones, and the KardiaMobile Card, a credit card-sized slim single-lead acquisition device with integrated metallic electrodes.

These recent efforts have brought ECG biometrics closer to viable and unconstrained applicability. However, these newer technologies still require the users to wear certain products or perform specific actions and need contact with both limbs during acquisition. Besides this, the quality of the acquired signals is typically very low, because of the loose contact with the subject’s skin, suffering from wide impedance variations, sensor saturation, and contact loss artefacts.

Some researchers have already started to address these issues. The single-arm acquisition settings studied by \citet{Zhang2017b} and the contactless electrodes developed by \citet{Chi2010} raise new and inspiring possibilities for wearable ECG devices. For applications based solely on heart rate, techniques have been proposed to measure it at a distance, using microwave Doppler sensors \cite{Scalise2012, Obeid2012, Bounyong2017}.

These efforts pave the way for better ECG acquisition technologies. Such systems could consist of seamlessly integrated biometric systems that can acquire ECG signals at short distances from one hand of the user, without requiring contact and thus suffering from signal loss. For wearables, the future could reside in products that can continuously monitor the users’ ECG while only contacting with one of their wrists, or when inside their pockets separated from the body by clothes. Despite all efforts devoted so far to ECG biometrics, much work is still needed to reach true applicability in the form of real, comfortable, and easy-to-use ECG-based biometric systems.

\subsubsection{Variability}

\begin{figure}
    \centering
    \includegraphics[width=0.5\linewidth]{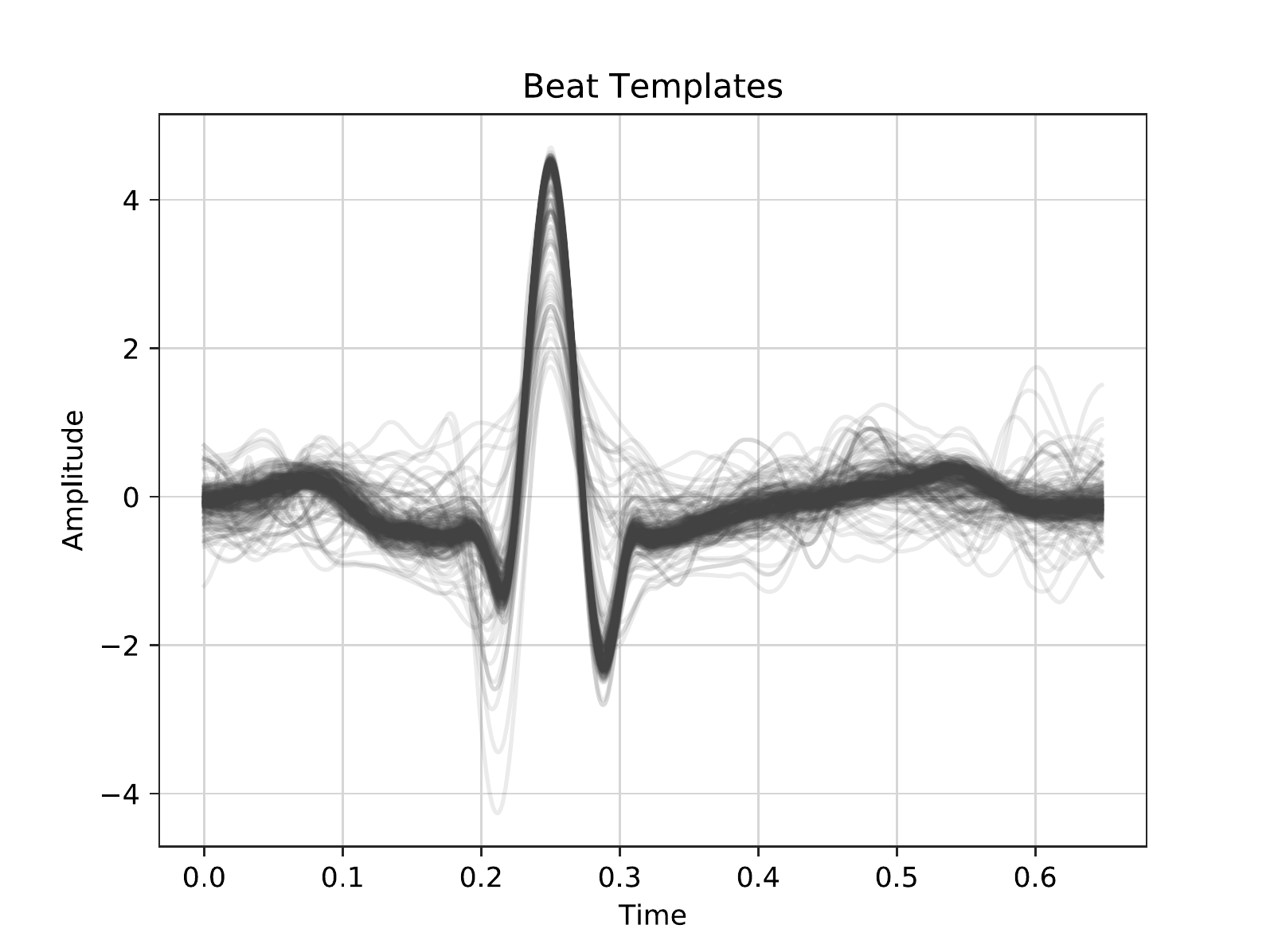}%
    \includegraphics[width=0.5\linewidth]{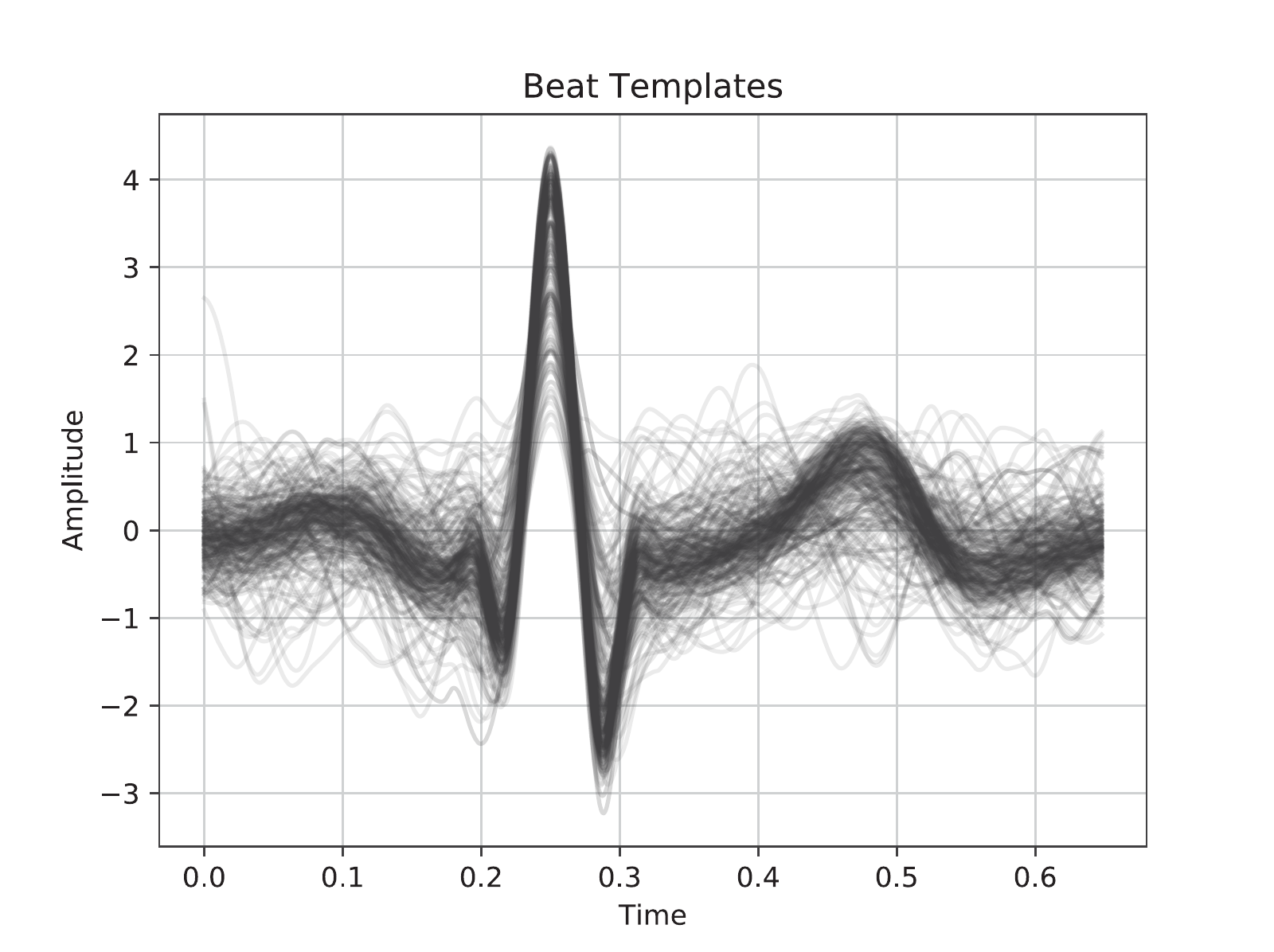}\\
    \includegraphics[width=0.5\linewidth]{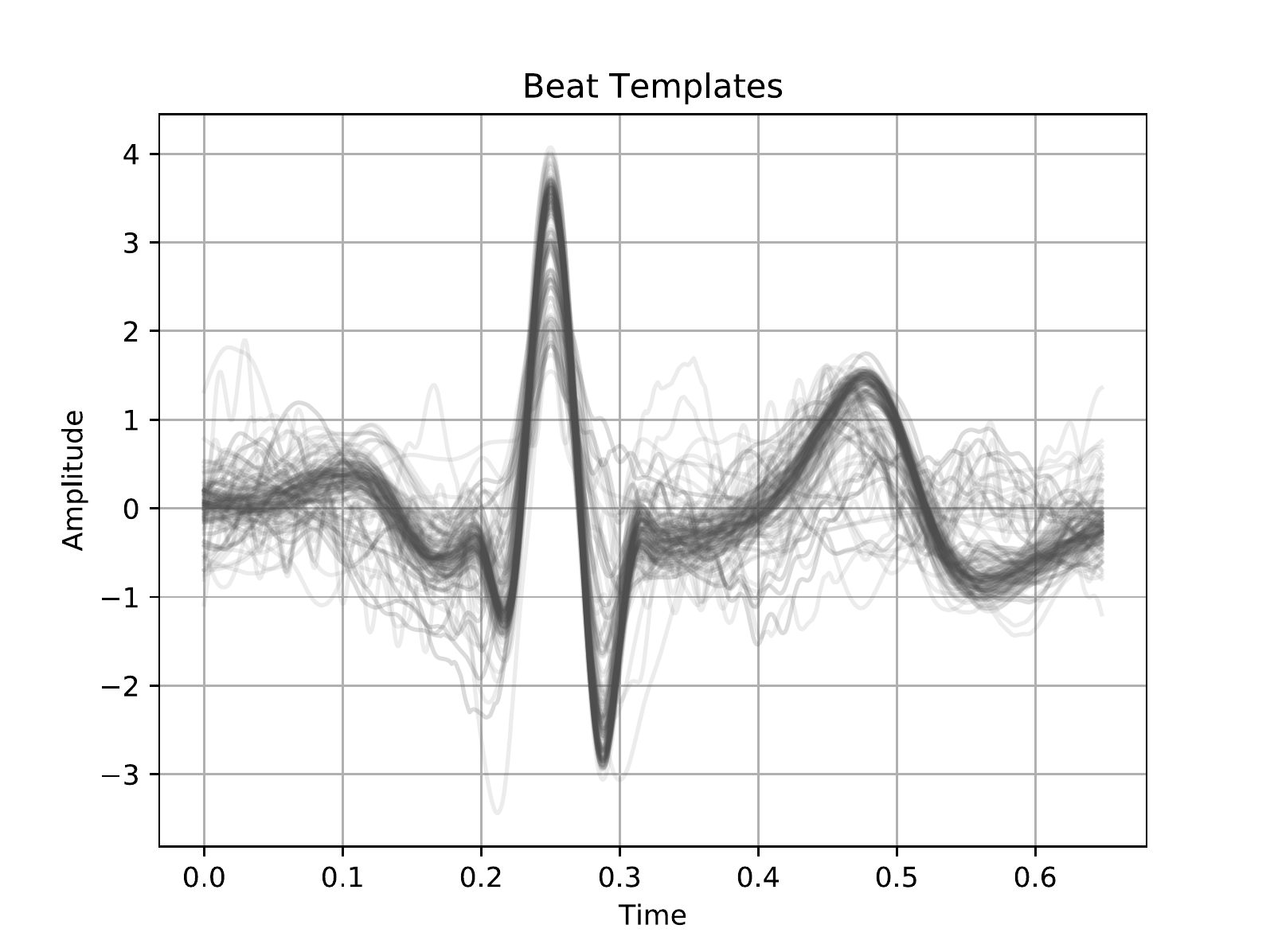}
    \caption[Variability in off-the-person ECG heartbeats from the same subjects.]{Variability in off-the-person ECG heartbeats from the same subjects (from~\cite{Pinto2017b}, individual heartbeats superposed after denoising, amplitude normalisation, and outlier removal).}
    \label{fig:ecg_variability}
\end{figure}

Although the ECG signals present, in normal conditions, the same deflections for all subjects at all times, these are characterised by a high degree of variability (see Fig.~\ref{fig:ecg_variability}). Variability in the ECG can be designated as \emph{intrasubject}, the variations between cycles (heartbeats) in the electrocardiogram of a single subject, or intersubject, the variations between heartbeats of different subjects. Intrasubject variations on the ECG signal are mainly explored for health monitoring and medical diagnosis \citep{Agrafioti2008,Sahayadhas2012,Rajpurkar2017}, while intersubject variations are especially useful to discriminate between subjects in biometric recognition. Both intrasubject and intersubject variability can originate from several factors, such as:
\begin{itemize}
\item \emph{Heart Geometry}: Heart size, cardiac muscle thickness, and the overall shape of the heart influence the trajectories of electrical currents throughout the heart, the number of muscle cells that will conduct those currents, and the time to do it across the whole heart. Athletes, because of intensive physical training, commonly have thicker myocardia, which affects the ECG with higher voltages in the QRS complex, and lower basal heart rates \citep{Hoekema1999,VanOosterom2000,Hoekema2001};

\item \emph{Individual Attributes}: Age, weight, and pregnancy are some individual attributes that can cause shifts in the heart position and orientation. These shifts will change the orientation of the electrical current conduction vectors across the heart, meaning the electrodes will detect the signal from a different perspective, thus altering the ECG waveforms \citep{Schijvenaars2000};

\item \emph{Physical Exercise or Meditation}: The duration of, and intervals between the different deflections of the heartbeats in an ECG signal, vary with the heart rate. These changes are especially visible on the interval between the QRS complex and the T wave in situations of tachycardia (higher heart rates) or bradycardia (lower heart rates). Changes in the heart rate caused by physical exercise or meditation are reflected on the electrocardiogram \citep{Agrafioti2012};

\item \emph{Cardiac Conditions}: Medical conditions of the heart can also interfere with the dynamics of electrical pulse conduction and generate variability. In biometrics, one of the most studied conditions is Arrhythmia, which causes wide variations in the heart rate across time and, as reported by several researchers, can consistently shrink the performance of ECG-based biometric systems \citep{Ye2010,Agrafioti2011,Safie2011}.

\item \emph{Posture}: Postures such as standing or laying down change the position and shape of internal organs. The heart is also affected by this, changing its position in the thorax, and thus its position in reference to the electrode placement, which will cause variations in the collected ECG signal \citep{Schijvenaars2000};

\item \emph{Emotions and Fatigue}: The sympathetic and parasympathetic systems of the autonomous nervous system work to, respectively, increase or reduce the heart rate. These systems are under the direct influence of psychological states and thus, under stress, fear and other strong emotions, fatigue or drowsiness, the heart rate and the ECG signal can be affected \citep{Agrafioti2012,Sahayadhas2012};

\item \emph{Electrode characteristics and placement}: The type, the size, and the number of electrodes, whether they are wet or dry, and the positioning on the chest or limbs, can influence the dominance of noise on the signal. The mispositioning of electrodes and reversal of leads are also sources of variability, as they change the perspective of detection of the electrocardiographic signal \citep{Hoekema1999,Schijvenaars2000}.
\end{itemize}

All the previously presented factors change the morphology of the electrocardiographic signals acquired from an individual. The first two factors contribute more to intersubject variability and the biometric potential of the ECG signals. The remaining factors are the main origins of intrasubject variability, which may undermine the process of biometric recognition, but offer the ECG information on several health and wellbeing parameters. When considering the acquisition of ECG for a specific application, whether for medical or biometric recognition, it is paramount to consider these factors, the way they can ease or difficult the task at hand, and how to mitigate their negative effects.

\subsection{Face}

The face may be considered the most intuitive of all biometric traits, since humans use facial features as the main clues for the identification of other people. Despite the challenges in its use for biometric recognition, the face offers the unique advantage of being the only trait that can be acquired using sensors at a considerable distance from the user, possibly without their knowledge \citep{Buciu2016, Hassaballah2014}. Below, the ways to measure the face trait and the factors that may affect the measurement are described and discussed.

\subsubsection{Acquisition}

\begin{figure}
    \centering
    \includegraphics[width=0.9\linewidth]{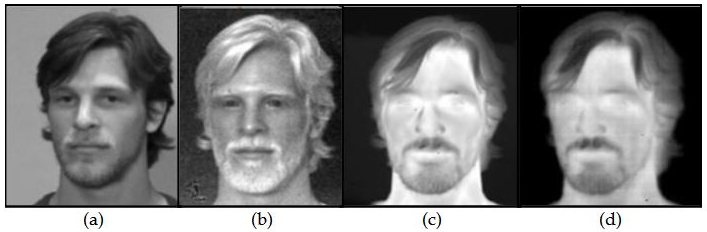}
    \caption[Comparison of face images acquired on the visible light, short-wave infrared, mid-wave infrared, and long-wave infrared spectra.]{Comparison of face images acquired on the (a) visible light, (b) short-wave infrared, (c) mid-wave infrared, and (d) long-wave infrared spectra (from~\cite{Bhowmik2011}).}
    \label{fig:rgb_jpg}
\end{figure}

Among widespread and inexpensive cameras and sophisticated thermal sensors, the face trait can be acquired in the following settings~\citep{Arya2015, Farokhi2016}:

\begin{itemize}
\item \emph{Visible Spectrum}: This is the most common setting in face biometrics. Visible light, with a wavelength in $[400,750]$ nm, generally from natural and unconstrained sources in the application environment, is reflected by the face of the user and captured by the sensor. Using the visible spectrum of light has some disadvantages, discussed in subsubsection~\ref{subsubsec:face_variability}, but it is commonly less expensive than the alternatives and, thus, the most fitting option for widespread applications.

\item \emph{Infrared Spectrum}: To overcome some limitations of the use of visible light, some sensors acquire the face trait using the infrared spectrum. These use thermal cameras that can obtain reliable face images in a much wider range of lighting conditions, capturing the heat from blood vessels and tissues on the user’s face. Based on the wavelength range, these can be designated as Near-Infrared (NIR, $[750, 1400]$~nm wavelength), Short-Wave Infrared (SWIR, $[1400,3000]$~nm), Mid-Wave Infrared (MWIR, $[3000, 8000]$~nm), or Long-Wave Infrared (LWIR, $[8000, 15000]$~nm). These are compared with visible light images in Fig.~\ref{fig:rgb_jpg}.
\end{itemize}

Depending on the frequency spectra used, the face trait can be categorised as either anatomical (visible light), or physiological (infrared spectrum). This is because the former mostly captures anatomical features of the face, while the latter is much more dependent on physiological factors that affect blood flow and face heat patterns.

\begin{figure}
    \centering
    \includegraphics[width=\linewidth]{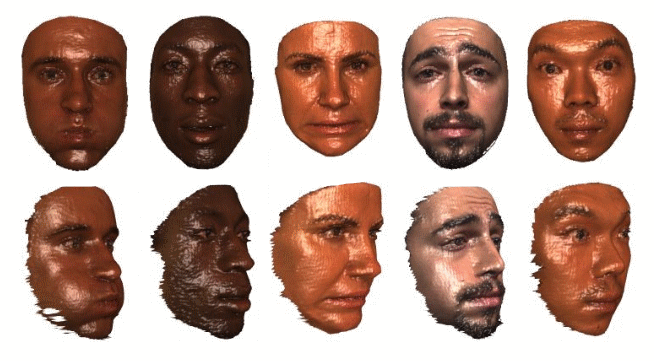}
    \caption[Examples of tridimensional face models.]{Examples of tridimensional face models (from the Bosphorus 3D Face Database~\citep{Savran2008}).}
    \label{fig:3d_faces}
\end{figure}

The face trait measurements can be acquired simultaneously with depth information, resulting in an upgrade of 2D images to 2.5D or 3D. While 2D data only includes colour or greyscale pixel information, 2.5D data combines that with depth information about the face of the subject, which is useful to increase recognition accuracy or avoid attack attempts. 3D data is a further step, where both sources of information are fused to build tridimensional models of the subject's face (see Fig.~\ref{fig:3d_faces})~\citep{Chong2019, Buciu2016}.

\subsubsection{Variability}\label{subsubsec:face_variability}

\begin{figure}
    \centering
    \includegraphics[width=0.9\linewidth]{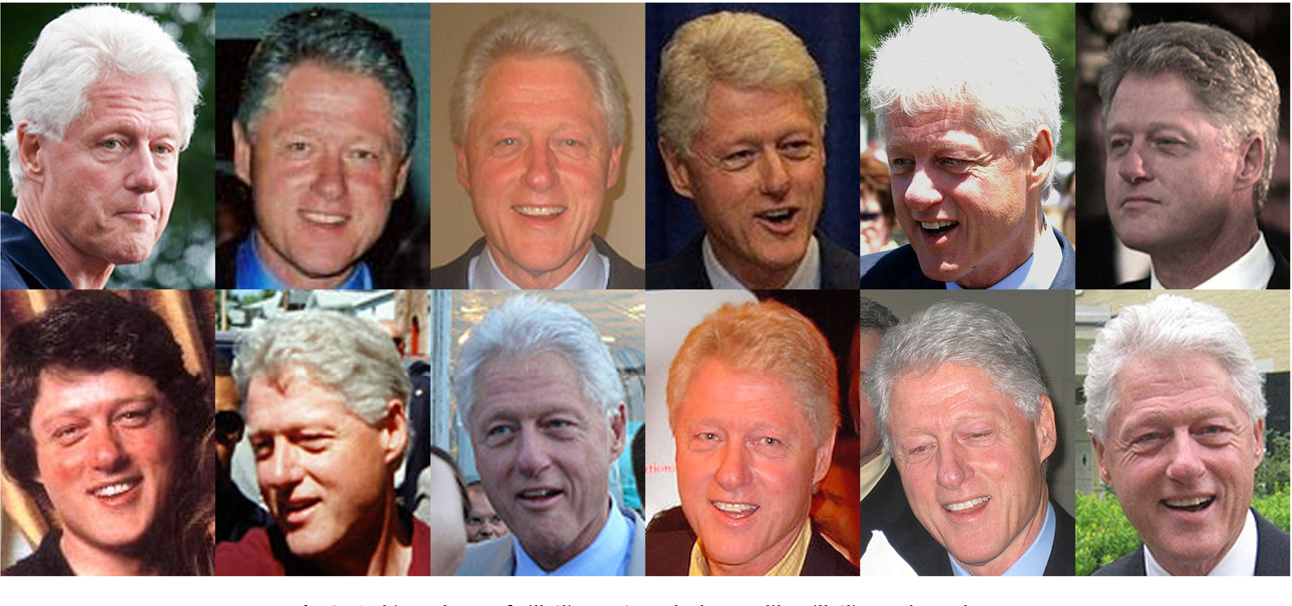}
    \caption[Variability in unconstrained face images of a subject.]{Variability in unconstrained face images of a subject (from~\citep{Jenkins2011}).}
    \label{fig:face_variability}
\end{figure}

The measurement of the face trait is affected by several intrinsic, environmental, or operational factors (see Fig.~\ref{fig:face_variability}). These can enhance intersubject or intrasubject variability, and thus make the recognition task easier or more difficult \citep{Admane2019, Arya2015, Buciu2016, Ding2015, Ding2017}. These factors are:
\begin{itemize}
\item \emph{Illumination and Heat Sources}: This factor is most relevant for visible spectrum acquisition settings, as the illumination with visible light will directly affect the light received by the sensor and change the face image. In infrared images, the influence of illumination can also be felt, due to the heat generated by light and heat sources, although generally in a lesser degree than in the visual spectrum; 
\item \emph{Pose Variations}: Variations in pitch, roll, and yaw of the subjects head, relative to the sensor, will result in the capture of facial features in different perspectives, which may encumber pattern recognition tasks. Moreover, the relative position of those facial features will also suffer from pose variations. Some authors argue infrared imaging is less susceptible to pose variations than visible spectrum sensors \citep{Friedrich2002}; 
\item \emph{Facial Expressions}: Like pose variations, facial expressions change the relative positions of the facial features, which may encumber the task at hand. Furthermore, it may create new features (\eg, wrinkles) or hide facial features that would be needed for the task. As with the previous factor, facial expressions were found to have less impact in infrared imaging \citep{Friedrich2002}; 
\item \emph{Stress, Emotions, and Illnesses}: Stress, emotions like happiness and fear, or illnesses like headaches or tooth infections can influence facial expressions. But more than that, they change the heat patterns on a subject’s head, which will influence the face images captured by infrared sensors; 
\item \emph{Age}: Growing old is generally accompanied by expression wrinkles, grey hair, among other changes that may affect someone’s appearance and make it more difficult for them to be recognised;
\item \emph{Cosmetics, Tattoos, Disguises, and Accessories}: Makeup, face tattoos, disguises, and accessories like rings or piercings can change, hide, or create new facial features. Recognition algorithms may not be prepared to deal with this new information and be induced into errors;
\item \emph{Occlusions and Surrounding Objects}: Like disguises and accessories, objects surrounding the person can sometimes occlude part of the face of the subject, and thus hide certain facial features that could be key for the recognition task at hand;
\item \emph{Sensor Quality and Stability}: The sensor is another important variability source. Using different cameras, with different characteristics, will result in different face images. Furthermore, the movement of the cameras (just like the movement of the subject) can cause blurred images that will disable the retrieval of finer facial features.
\end{itemize}

Like with the electrocardiogram, it is important to appropriately weigh all these factors when designing a biometric system. Some will contribute more to intersubject variability, and will be desirable for biometric recognition, while others will enhance intrasubject variability, and empower wellbeing monitoring. Considering the task at hand, it is key to design the acquisition and processing stages to focus on the factors that fit the application needs.

\section{Performance Evaluation}

\subsection{System design considerations}

When designing, developing, and deploying a biometric system, it is important to be aware of how it will behave in realistic conditions. Hence, designers and developers should consider some central aspects, as described by \citet{Bolle2004}:
\begin{itemize}
\item \emph{System accuracy}: This measures the frequency with which the biometric system makes correct or wrong decisions;
\item \emph{Computation time}: This corresponds to the time required by the system to output a decision, starting from the moment the user initiates contact with the acquisition module. This time depends on the processes of trait acquisition, quality assurance, feature extraction, and decision, and should be as low as possible, to improve usability. This aspect is especially important for continuous recognition systems, where decisions should be quickly output and frequently renewed;
\item \emph{Exception handling}: Because of universality issues and sensor flaws or unavailability, some users may find the system unable to acquire their biometric traits. Also, software errors may occur and render the system incapable of adequately performing its function. These possibilities should be addressed during system design and development, to ensure the application works even without the biometric system;
\item \emph{System cost}: The system cost includes all expenses related to acquisition and processing equipment needed, algorithm development and implementation, routine maintenance, and operational costs. It should be as low as possible, to make the biometric system more affordable;
\item \emph{Security}: Biometric systems decisions can serve as proof of the actions of a certain individual, and this, in some settings, can escalate to serious legal implications. Thus, it is important to minimise the possibility of decision flaws that allow impostors to act under the identity of an authorised person;
\item \emph{Privacy}: Biometric systems require the storage of templates, consisting in discriminating information about each of the enrolled individuals. That information, in the interest of anonymity and security, should be kept as safe as possible, resorting to encryption techniques that allow matching but minimise the possibility of reconstructing the original acquired trait data.
\end{itemize}

Some conflicts may exist between these aspects. Using more sophisticated approaches (such as deep learning) often leads to higher system accuracy, but at the expense of higher computational complexity. Using passwords or credentials as a fallback in case of an exception weakens system security. Tighter security measures generally lead to more frequent rejection of legitimate users, which translates into reduced accuracy.

Ultimately, trying to fully cover all aspects will increase costs. Hence, all these aspects should be carefully analysed and weighted, considering the expected application to get an affordable, efficient, accurate, secure, and usable system. In the next subsection, the most relevant metrics are presented, regarding the specific aspect of system accuracy.

\subsection{Recognition accuracy measurement}

\begin{table*}
\centering
\caption{Definition of the commonly used metrics for performance evaluation in identification and identity verification tasks.}
\label{tab:metrics}
\begingroup
\small
\setlength{\tabcolsep}{5pt}
\renewcommand*{\arraystretch}{2.0}
\begin{tabular}{clc}
 & \textbf{Metric} & \textbf{Definition}\tabularnewline\hline
\multirow{4}{*}{\rotatebox[origin=c]{90}{\textbf{Identity verification}}} & False Acceptance Rate & $FAR(T) = \frac{\text{Number of impostor trials with prediction score above } T}{\text{Total number of impostor trials}}$\tabularnewline
& False Rejection Rate & $FRR(T) = \frac{\text{Number of legitimate trials with prediction score below } T}{\text{Total number of legitimate trials}}$ \tabularnewline
& Equal Error Rate & $EER = FAR(T)$, for $T$ that gives $FAR(T)=FRR(T)$\\
& Area Under the Curve & $AUC = \int_0^1 1-FRR(FAR(T))~dT $\tabularnewline\hline
\multirow{3}{*}{\rotatebox[origin=c]{90}{\textbf{Identification}}} & True Positive Identification Rate & $TPIR(R) = \frac{\text{No. of trials where one of the strongest R predictions is correct}}{\text{Total number of trials}}$\tabularnewline
& Identification Rate & $IDR = TPIR(1) = \frac{\text{No. of trials where the strongest prediction is correct}}{\text{Total number of trials}}$ \tabularnewline
& Misidentification Rate & $MIDR = 1 - IDR = \frac{\text{No. of trials where the strongest prediction is incorrect}}{\text{Total number of trials}}$\tabularnewline\hline
\end{tabular}
\endgroup
\end{table*}

\subsubsection{Identity verification}

As described before, identity verification consists in accepting or rejecting an identity claim made by a user based on their biometric trait measurements. As such, there are four outcomes:
\begin{enumerate}
\item The claim is true and the system correctly accepts it;
\item The claim is false and the system correctly rejects it;
\item The claim is true, but the system incorrectly rejects it;
\item The claim is false, but the system incorrectly accepts it.
\end{enumerate}

In identity verification, the main goal is to minimise the frequency of outcomes three and four. Outcome three corresponds to a \emph{False Reject} or \emph{False Non-Match} error, and outcome four corresponds to a \emph{False Accept} or \emph{False Match} error. These errors are the foundation for most identity verification accuracy metrics (see Table~\ref{tab:metrics}), among which the two most commonly used are \citep{Bolle2004,Grother2010,Agrafioti2011b}:
\begin{itemize}
\item \emph{False Acceptance Rate} ($FAR$ or $FMR$, False Match Rate): measures the fraction of trials where the system accepted the identity claims, even though they were false. It represents how frequently the system erroneously grants access to impostors. The complement of $FAR$ is called \emph{Convenience};
\item \emph{False Rejection Rate} ($FRR$ or $FNMR$, False Non-Match Rate): measures the fraction of trials where the system rejected true identity claims. As such, it measures how frequently the system denies access to legitimate users. The complement of $FRR$ is designated as \emph{Security}.
\end{itemize}

\begin{figure}[t]
\begin{center}
\includegraphics[width=0.35\linewidth]{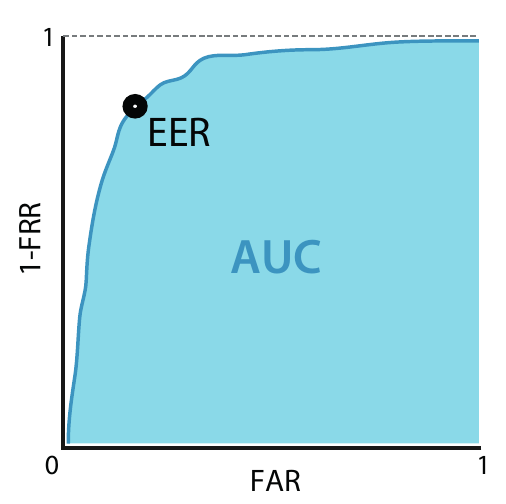}
\includegraphics[width=0.375\linewidth]{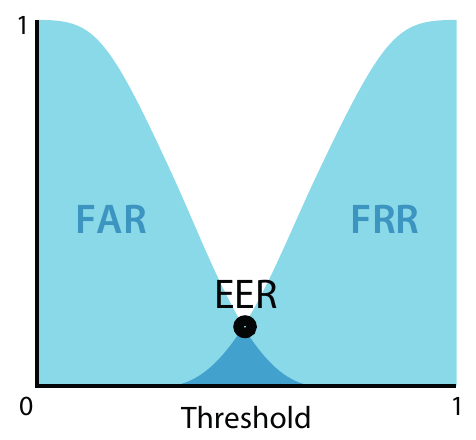}
\caption[Example of a Receiver Operating Characteristic (ROC) curve for an identity verification system and the evolution of False Acceptance and False Rejection rates with the threshold value.]{Example of a Receiver Operating Characteristic (ROC) curve for an identity verification system (left) and the evolution of False Acceptance and False Rejection rates with the threshold value (right) (from~\citep{Pinto2018}, adapted from~\citep{Syris2004}, example for a similarity-based matching method including the Equal Error Rate point and the Area Under the Curve).}
\label{fig:ROC}
\end{center}
\end{figure}

For a given system, these two metrics commonly depend on the chosen operation point. For example, in similarity or dissimilarity-based matching algorithms, the criterion for accepting or rejecting a claim is generally a threshold. Different threshold values will correspond to different $FAR$ and $FRR$ results. Hence, for a more complete performance evaluation, it is common to use performance characteristic curves such as the \emph{Receiver Operating Characteristic} (ROC), which plots $1-FRR$ versus $FAR$ for varying threshold values (see Fig.~\ref{fig:ROC}).

From the ROC curve, it is also common to extract two metrics that combine all results into a single performance value, easing performance comparison between algorithms. The \emph{Equal Error Rate} ($EER$) is the error that corresponds to the operation point where $FAR=FRR$ and represents an equilibrium between convenience and security (see Fig.~\ref{fig:ROC}). The \emph{Area Under the Curve} ($AUC$) measures the area under the ROC curve and serves as a measure of overall quality of a biometric system.

\subsubsection{Identification}

In identification, the system does not receive an identity claim. Hence, it will compare the current biometric measurements with the stored templates to assign one of the enrolled identities to the subject. Additionally, it can reject to identify when the strongest match is, still, not strong enough, which is generally asserted based on a threshold (as in identity verification mode).

Thus, in the identification mode, there are five possible outcomes:
\begin{enumerate}
\item The subject is enrolled and the system correctly identifies them, granting them access;
\item The subject is enrolled, but the system mistakes their identity for another enrolled subject, granting them access under a wrong identity;
\item The subject is enrolled, but the system fails to identify them with any enrolled subject, rejecting access;
\item The subject is not enrolled and the system correctly rejects access;
\item The subject is not enrolled, but the system erroneously identifies them as one of the enrolled subjects and grants them access.
\end{enumerate}

A good biometric system would maximise the frequency of outcomes 1 and 4 and minimise the frequency of the remaining outcomes. Most metrics for identification mode (see Table~\ref{tab:metrics}) are based on these frequencies \citep{Bolle2004,Grother2010,Agrafioti2011b}. Regarding the situations where the subject is enrolled (designated as legitimate trials), the most common metrics are:

\begin{itemize}
\item \emph{True-Positive Identification Rate} ($TPIR$ or Hit Rate): For a total number of legitimate trials, $TPIR$ corresponds to the fraction of those where one of the system’s $R$ strongest predictions corresponds to the true subject identity. As most other metrics, $TPIR$ depends on the defined threshold, the selected number of top ranks $R$, and the list of enrolled candidates (in identification, each enrolled subject is considered a candidate, and TPIR, as most metrics, varies not only with the size of $L$ but also with the variety and individual characteristics of each subject in $L$);

\item \emph{Identification Rate} ($IDR$ or Accuracy): $IDR$ corresponds to $TPIR$ when only the single highest ranking prediction is considered ($R=1$). It corresponds to the fraction of the legitimate trials where the true identity was the method’s strongest prediction above the threshold. In the literature, this is one of the most used metrics;

\item \emph{Reliability}: Reliability corresponds to $TPIR$ with $R=N$ (where $N$ is the number of enrolled subjects), we get the reliability metric. This measures how frequently the true identity satisfies the minimum threshold constraint, regardless of its ranking;

\item \emph{False-Negative Identification Rate} ($FNIR$ or Miss Rate): Represents the fraction of trials where the true identity does not correspond to one of the $R$ strongest predictions above the threshold $T$;

\item \emph{Reject Rate} ($RR$): This metric pertains to the very specific situations where all identity predictions stand below the defined threshold $T$, and the system has no choice but to reject to identify ($RR = 1 - TPIR - FNIR$);

\item \emph{Misidentification Rate} ($MIDR$ or, commonly, Misclassification Error): $MIDR$ is the complement of $IDR$ ($MIDR=1-IDR$), equivalent to $FNIR$ with $R=1$, and measures the fraction of legitimate trials where the true identity is not the system’s top ranking prediction above $T$.
\end{itemize}

Regarding situations where the subject is not enrolled in the biometric system (designated as impostor trials), the most common metrics are:
\begin{itemize}
\item \emph{False Positive Identification Rate} ($FPIR$): It is the fraction of impostor trials where at least one of the system’s predictions meets the threshold criterion, and the system thus grants access to the unenrolled subject;

\item \emph{Selectivity}: Similar to $FPIR$, selectivity counts the average number of predictions above the threshold $T$ across all impostor trials.
\end{itemize}

\begin{figure}
\begin{center}
\includegraphics[width=0.8\linewidth]{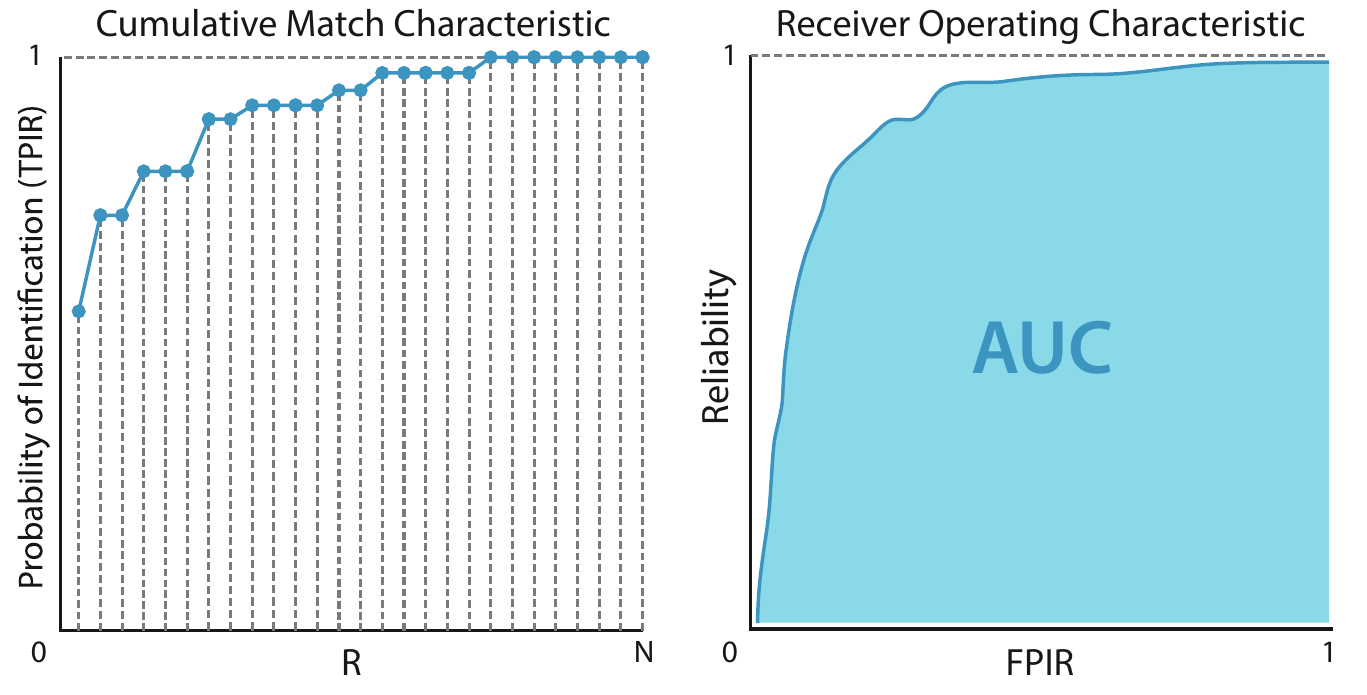}
\caption[Examples of a Cumulative Match Characteristic (CMC) curve, and a Receiver Operating Characteristic (ROC) curve for an identification system.]{Examples of a Cumulative Match Characteristic (CMC) curve, and a Receiver Operating Characteristic (ROC) curve for an identification system (from~\citep{Pinto2017b}, including a representation of the Area Under the Curve, AUC).}
\label{fig:CMCandROC}
\end{center}
\end{figure}

As in identity verification mode, some characteristic curves can be drawn based on these metrics and the defined thresholds, to help evaluate the algorithms more robustly. The first is the \emph{Cumulative Match Characteristic} (CMC), which plots $TPIR$ with threshold $T=0$ against $R$, varying $R\in[1, N]$. The second is the \emph{Receiver Operating Characteristic} (ROC) which, in the case of identification, plots Reliability against $FPIR$, for various threshold values (see Fig.~\ref{fig:CMCandROC}). From these plots, we can also extract the $AUC$ and $EER$ metrics.

\subsection{Time-based performance measurement}

There are biometric systems that perform recognition upon request, for example in computers or smartphones, keeping the session open until the user ends it or it reaches an idle time limit. This creates security issues, specifically when the user leaves the device unattended and forgets to close the session.

To solve this issue, there are \emph{continuous} (or \emph{online}) biometric systems, that aim to perform biometric recognition in real-time, acquire traits continuously, and renew decisions as frequently as possible based on the most recent acquisition. With such systems, if the user leaves and is replaced by an attacker, the system would be able to detect this and close the session before any harm could have been done.

\begin{figure}
\begin{center}
\includegraphics[width=0.4\linewidth]{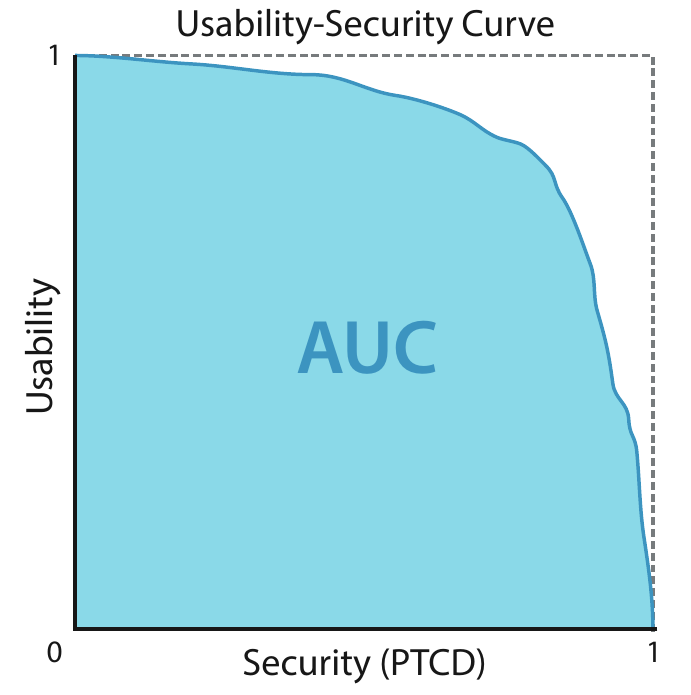}
\caption[Example of a Usability-Security characteristic curve.]{Example of a Usability-Security characteristic curve (from~\citep{Pinto2017b}, adapted from \citet{Sim2007}).}
\label{fig:USC}
\end{center}
\end{figure}

Besides accuracy, an important facet of performance evaluation for continuous biometric systems is timeliness. \citet{Sim2007} have addressed this issue and proposed some time-based metrics for the evaluation of continuous identity verification systems. These can be adapted for both identification and identity verification, as described below:
\begin{itemize}
\item \emph{Time to Correct Decision} ($TCD$): $TCD$ measures the time the system takes to detect an impostor, and take an appropriate decision, relative to the moment the impostor replaces a legitimate user. For an ideal system, this should be zero, but that is virtually impossible to achieve. Hence, \citet{Sim2007} state that this window should at least be always lower than $W$, called \emph{Window of Vulnerability} (the minimum access time required for the impostor or wrong individual to cause any kind of damage);

\item \emph{Probability of Time to Correct Decision} ($PTCD$): $PTCD$ measures the probability, for a given system, of $TCD$ being lower than $W$. The higher this value, the lesser the probability of an impostor having time to cause damage before the system acts;

\item \emph{Usability}: In a normal continuous biometric system, we should expect some decisions to be incorrect. Over $t$ seconds of usage, Usability measures the fraction of $t$ where the legitimate user is deprived of access due to wrong decisions of the system. For any biometric system, this should be as close to zero as possible.

\end{itemize}

From these metrics, the authors also define the \emph{Usability-Security curve} (USC), a new characteristic curve that plots Usability vs. $PTCD$ for a varying threshold $T$ (see Fig. \ref{fig:USC}). USC is similar to a ROC curve and, thus, $AUC$ can also be computed, being considered a good metric to evaluate the timeliness of a biometric system.

\subsection{Wellbeing monitoring performance measurement}

For wellbeing monitoring systems, several metrics have been used to measure their performance~\citep{Mollahosseini2019}. The choices depend on the nature of the ground-truths and system outputs.

For example, emotion recognition or affective computing algorithms generally focus either on a limited set of discrete emotion categories or on continuous ranges of emotion qualifiers. In the former case, they typically cluster all emotions onto six categories designated as the \emph{six basic emotions} by \citet{Ekman1992}: happiness, sadness, anger, fear, surprise, and disgust.

For systems based on categorical labels such as these, the most common metrics are:
\begin{itemize}
\item \emph{Accuracy}: The accuracy is the fraction of test samples that have been correctly classified by the algorithm. It is widely used in wellbeing monitoring applications;

\item \emph{F1-score}: The F 1-score is the harmonic mean of precision $p$ (the fraction of positive predictions that are truly positive) and recall $r$ (the fraction of positive samples that are classified as such), through the expression $F_1 = 2(pr)/(p+r)$. It is commonly used to evaluate performance in binary tasks;

\item \emph{AUC}: The Area Under the ROC Curve, which plots sensitivity (the fraction of positive samples correctly classified) \emph{versus} the complement of specificity ($1-$specificity, the fraction of negative samples incorrectly classified as positive) for several thresholds, is also commonly used for binary classification tasks.
\end{itemize}

\begin{figure}
    \centering
    \includegraphics[width=0.6\linewidth]{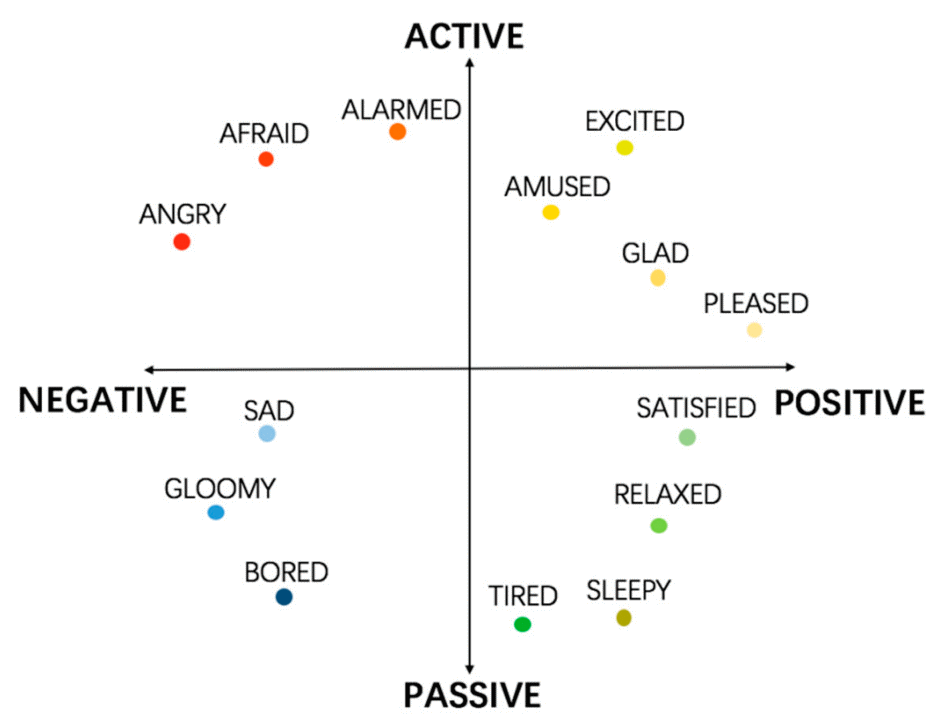}
    \caption[Illustration of the bidimensional valence-arousal space with example emotion categories.]{Illustration of the bidimensional valence-arousal space with example emotion categories (adapted from~\cite{Shu2018}, with valence on the horizontal axis and arousal on the vertical axis).}
    \label{fig:emotion_valence_arousal}
\end{figure}

In the case of continuous labels, emotion recognition systems typically consider a bidimensional space with two main emotion qualifier variables: valence and arousal~\cite{Lang1995}. Valence measures the pleasure of the emotion being felt, ranging from negative (unpleasant) to positive (pleasant). Arousal measures the level of activation or intensity of the emotion, ranging from low (passive) to high (active). Fig.~\ref{fig:emotion_valence_arousal} illustrates these concepts by offering examples of categorical emotions in the bidimensional valence-arousal space.

For systems focused on regression tasks such as these, outputting continuous scores, the most common metrics are:
\begin{itemize}
\item \emph{Root Mean Squared Error} ($RMSE$): $RMSE$ is the root square of the mean of all squared differences between corresponding predictions and ground-truths;

\item \emph{Pearson's Correlation Coefficient} ($CC$): To correct the limitations of $RMSE$, the Pearson's correlation coefficient uses the covariance between predictions $\hat{\theta}$ and ground-truths $\theta$. It follows the expression $CC=COV\{\hat{\theta},\theta\}/(\sigma_{\hat{\theta}}\sigma_{\theta})$;

\item \emph{Concordance Correlation Coefficient} ($CCC$): This metric is used for time-series predictions, based on the Pearson's correlation coefficient and the mean value of each time series, following the expression $\rho_c = \{2CC\sigma_{\hat{\theta}}\sigma_{\theta}\}/\{\sigma_{\hat{\theta}}^2 + \sigma_{\theta}^2 (\mu_{\hat{\theta}} - \mu_{\theta})^2 \}$. This is a very common metric for performance evaluation of emotion recognition over time;

\item \emph{Sign Agreement Metric} ($SAGR$): The sign agreement metric combines the magnitude of the prediction error with a penalisation for sign errors. It can be computed with $SAGR=\frac{1}{n}\sum_{i=1}^{n} \delta(sign(\hat{\theta_i}), sign(\theta_i))$, where $\delta$ is the Kronecker delta function. This metric has been mainly used for valence and arousal prediction, where the concordance of signs between the predictions and the ground-truths can be more important than the magnitude of the scoring errors.
\end{itemize}

Among these alternatives, it is important to consider the task at hand when selecting metrics for an adequate performance evaluation that also enables a simple and thorough comparison with the state-of-the-art.

\part{Electrocardiogram Biometrics}\label{part:ecgBiometrics}
\chapter{Prior Art in Electrocardiogram Biometrics}\label{ch:ecgprior}

\section{Data}

Numerous researchers, when working with ECG signals, for biometric recognition purposes or automatic diagnosis of medical cardiac conditions, opt for private acquisitions of data. However, as the needs grow for more complete datasets, with more subjects, including medical conditions, on more sessions, spread across wider time frames, and under different posture and activity conditions, researchers became more aware of the importance of public signal collections~\cite{Silva2014}.

Moreover, public ECG databases are needed to enable the comparison and benchmarking of algorithms in challenging conditions, across different publications, without requiring authors to implement algorithms and evaluate them again.  Below, we delve into the important aspects behind a well-structured ECG signal collection, we present the most relevant publicly available collections, and we discuss the current needs and future possibilities regarding data in ECG biometrics.

\subsection{Building a complete ECG data collection}

A well-structured ECG signal collection is key to appropriately guiding the development towards the exploitation of the best possibilities for the system, and accurately predicting its performance upon real-life application. To achieve such a complete collection, a few aspects have to be considered:

\begin{itemize}
\item \textit{Number of electrodes}: Fewer electrodes and leads have been shown to provide more challenging settings for biometrics \cite{Fang2009, Poree2016};

\item \textit{Electrode placement}: As shown by \cite{Zhang2006}, the use of chest leads is less challenging than limb leads, and the distance of the electrodes to the heart has a significant negative impact on the system's performance;

\item \textit{Sampling frequency}: Sampling causes the loss of fine details that influence the recognition process~\cite{Poree2016}. The lower the sampling frequency, the larger the amount of detail that can be lost, and the higher the risk of aliasing of high-frequency noise (such as electromyogram interference);

\item \textit{Subject posture, activity, and fatigue}: Several studies have shown that fatigue, exercise, or different postures have a negative effect on recognition performance if the systems have not been trained accordingly \cite{Poree2016,Wahabi2014,Pathoumvanh2014};

\item \textit{Subject health}: Some health issues, mainly arrhythmia, can generate intrasubject signal variability that encumbers the recognition process \cite{Safie2011,Dar2015,Dar2015b}. Thus, systems should be made robust against this, by including subjects with heart conditions in the datasets used during the development and validation of the methods;

\item \textit{Number of subjects}: The diversity of individuals and their own characteristics may ease or difficult the job of the biometric systems \cite{Doddington1998, Yager2010}, and successful state-of-the-art algorithms have been shown to be significantly worse when evaluated on larger datasets \cite{Odinaka2012}. The use of a collection with a large number of subjects ensures the presence of subject diversity, increasing the thoroughness of the performance assessment. As discussed in~\cite{Pinto2018}, the vast majority of literature in ECG biometrics reports the use of data from less than 100 subjects;

\item \textit{Acquisition sessions}: The ECG signal varies enough to cause recognition errors in most biometric systems, even over a short 24-hour period \cite{Labati2013,Labati2014}. Systems should be prepared with data from several sessions, weeks or months apart, to ensure their robustness \cite{Silva2013,Poree2016}.
\end{itemize}

All these factors can have an impact on the performance of an ECG-based biometric system. In order to correctly assess the capabilities of such systems, it is of the highest relevance to not only build a database that fits the system's expected application context, but also one that reflects all possibilities mentioned above, in order to study the use of the same biometric system in a wider set of contexts.

\subsection{Publicly available data}

\afterpage{
\begin{landscape}
\begin{table*}
\centering
\caption[Summary of the technical specificities of the most relevant publicly available ECG collections.]{Summary of the technical specificities of the most relevant publicly available ECG collections (from \citep{Pinto2018}, OP -- off-the-person; NS -- number of subjects; Fs -- sampling frequency (Hz); L / E -- number of leads/electrodes).}\label{tab:db_compar}
\begingroup
\small
\renewcommand*{\arraystretch}{1.25}
\begin{tabular}{>{\raggedright}p{3.0cm} >{\centering}p{0.6cm} >{\centering}p{0.7cm} >{\centering}p{0.8cm} >{\raggedright}p{2.3cm} >{\centering}p{1cm} >{\raggedright}p{2.5cm} >{\raggedright}p{4.0cm} >{\raggedright}p{4.2cm}}

\hline
\textbf{Collection} & \textbf{OP} & \textbf{NS} & \textbf{Fs} & \textbf{Electrode Placement} & \textbf{L / E} & \textbf{Health Conditions} & \textbf{Activity/Posture} & \textbf{Sessions}\tabularnewline\hline

AHA & No & 154 & 250 & Chest & 2 / - & Various & - & 3 h \tabularnewline

CEBSDB \cite{GarciaGonzalez2013} & No & 20 & 5000 & Chest & 2 / - & None & At rest, listening to music & 60 min. \tabularnewline

CYBHi \cite{Silva2014} & Yes & 128 & 1000 & Palms + Fingers & 2 / 4 & None & Reactions triggered by sound and video & Up to two 5 min. sessions, 3 months apart\tabularnewline

DriveDB \cite{Healey2005} & No & 9 & 456 & Chest & 1 / - & - & Rest, highway, and city driving & 50 min. to 1.5 h \tabularnewline

ECG-ID \cite{Lugovaya2005,Nemirko2005} & No & 90 & 500 & Wrists & 1 / - & - & Sitting, unrestrained movement & Various 20 s rec. per subject over 6 months \tabularnewline

E-HOL 24h & No & 203 & 200 & Chest & 3 / 4 & None & Ambulatory recordings & 24 h\tabularnewline

European ST-T \cite{Taddei1992} & No & 79 & 250 & Chest & 2 / - & Various & Ambulatory recordings & 2 h sessions \tabularnewline

FANTASIA \cite{Iyengar1996} & No & 40 & 250 & - & 1 / - & None & Supine, at rest, watching a movie & 120 min.\tabularnewline

LTST \cite{Jager2003} & No & 80 & 250 & Chest & 2-3 / - & Arrhythmia and ischaemia & Ambulatory recordings & 21-24 h  \tabularnewline

MIT-BIH Arrhythmia \cite{Mark1982,Moody1990} & No & 47 & 360 & Chest & 2 / - & None & Ambulatory recordings & 30 min.  \tabularnewline

MIT-BIH NSR    \cite{Mark1982,Moody1990} & No & 18 & 360 & Chest & 2 / - & None & Ambulatory recordings & 30 min. \tabularnewline

Physionet 2017 CinC Challenge & Yes & 8528 & 300 & Fingers & 1 / 2 & Various & At rest & 10-60s\tabularnewline

PTB    \cite{Bousseljot1995} & No & 290 & 1000 & Chest + Limbs & 15 / - & Various & At rest only & 1-5 per subject, 38.4-104.2 s  \tabularnewline

PTB-XL \cite{ptbxl1, ptbxl2} & No & 18~885 & 500 & Chest + Limbs & 12 / 10 & Various & At rest only & 1-2 per subject, 10 s \tabularnewline

QT \cite{Laguna1997} & No & 105 & 250 & Chest & - / - & Various & Rest and exercise & 15 min.\tabularnewline

UofTDB \cite{Wahabi2014} & Yes & 1019 & 200 & Fingers & 1 / 2 & None & Sit, stand, supine, exercise, and tripod & Up to six 2-5 min. recordings over 6 months  \tabularnewline\hline
\end{tabular}
\endgroup
\end{table*}
\end{landscape}
}

\begin{figure}
    \centering
    \includegraphics[width=\linewidth]{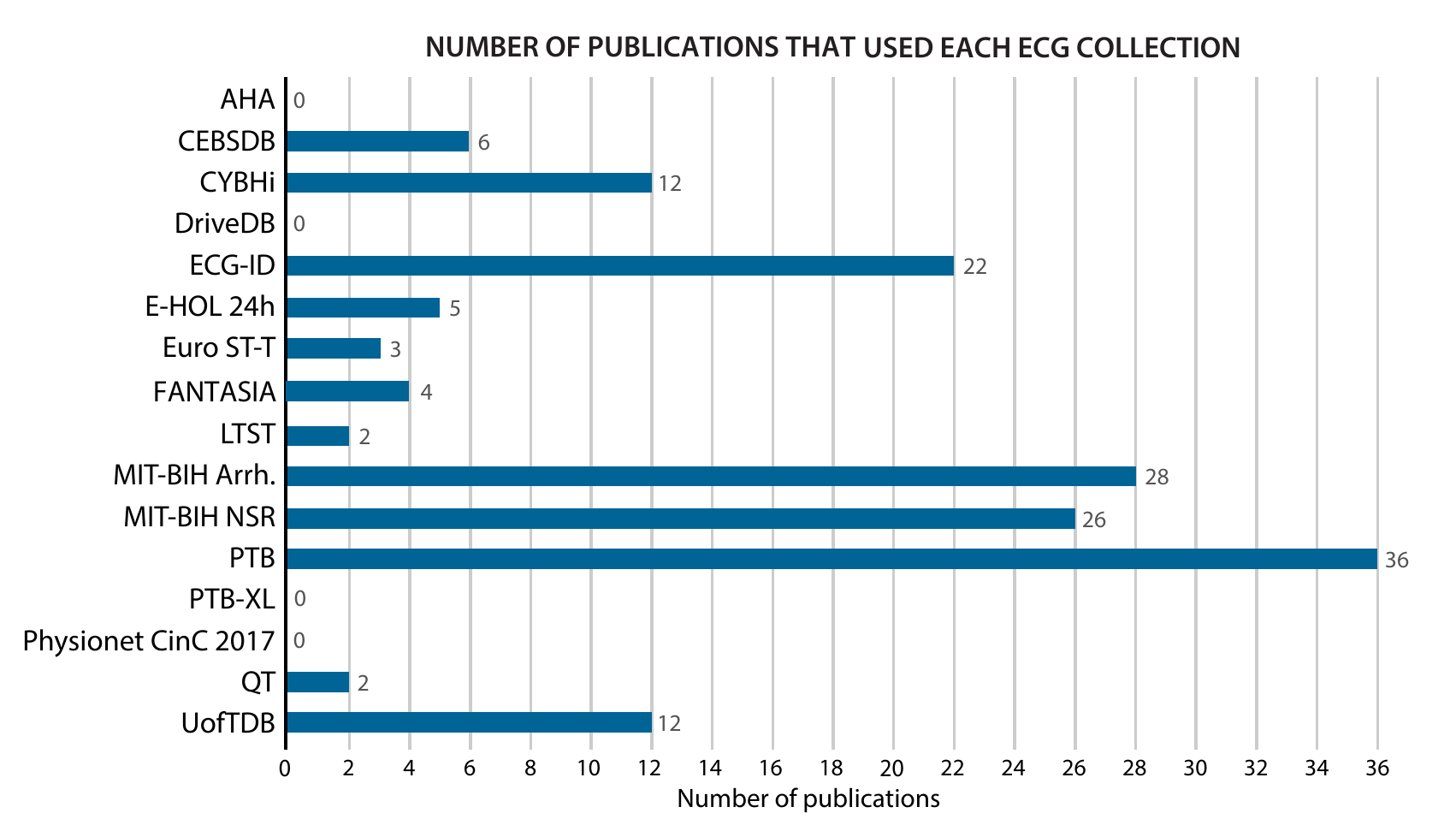}
    \caption[Currently available ECG collections and the number of surveyed publications that have used them.]{Currently available ECG collections and the number of surveyed publications that have used them (adapted from~\citep{Pinto2018}).}
    \label{fig:pubs_per_collection}
\end{figure}

Currently, there are several collections, publicly available for ECG biometrics research\footnote{Some of these databases may require prospective users to contact the respective administrators to request access to the data and/or sign agreements beforehand. Nevertheless, all presented databases are made available by the creators for research purposes.}, which try to cover the aforementioned factors to create a challenging environment for the development of robust biometric systems. Many are stored by Physionet\footnote{Physionet ECG databases. Available on: \url{https://www.physionet.org/physiobank/database/\#ecg}.}, while others are ceded by their owners upon request. Below, we present and characterise the most relevant of the currently available ECG collections (see Fig. \ref{fig:pubs_per_collection} for the number of publications that have used them), and Table \ref{tab:db_compar} summarises the characteristics of each.
\begin{itemize}
\item \emph{AHA}: The AHA ECG database\footnote{American Heart Association ECG database. Available on: \url{https://www.ecri.org/components/Pages/AHA_ECG_USB.aspx}.} was created by the American Heart Association to guide the training of health professionals on the diagnosis of arrhythmias. It includes 154 ECG recordings from real patients, donated by various institutions, each three hours long and composed of two lead signals. The last 30 minutes of each recording are annotated for seven types of arrhythmia;

\item \emph{CEBSDB}: The Combined measurement of ECG, Breathing and Seismocardiograms (CEBSDB) database~\cite{GarciaGonzalez2013} is a multimodal database available on Physionet. It includes two channels of ECG (standard leads I and II), thoracic respiratory signals, and seismocardiograms (SCG) from twenty healthy subjects at rest and in the supine position. Recordings include 50 minutes of classical music listening preceded and followed by five minutes at rest. ECG signals are sampled at 5 kHz and were acquired with foam tape and gel electrodes;

\item \emph{CYBHi}: The Check Your Biosignals Here initiative\footnote{CYBHi dataset for off-the-person ECG biometrics. Available on: \url{https://zenodo.org/record/2381823\#.YwDLmHbMKMp}.} \cite{Silva2014} is a collection of off-the-person ECG signals acquired with two dry electrodes at the palms, and two electrolycras at the middle and index fingers. It consists of a short-term dataset, with single-session recordings of 65 volunteers; and a long-term dataset, where 63 subjects were recorded in two sessions, three months apart. In each session, for 5 minutes, the subjects were exposed to videos designed to cause emotional reactions;

\item \emph{DriveDB}: Resulting from the Stress Recognition in Automobile Drivers initiative, this database was created with the purpose of monitoring stress in drivers \cite{Healey2005}. Various physiological parameters (electrocardiogram, electromyogram, and skin conductivity) were recorded from 9 subjects over a total of 18 driving sessions, including periods of rest (lower stress levels), highway driving, and city driving (higher stress levels);

\item \emph{ECG-ID}: The ECG-ID is a database entirely focused on biometrics \cite{Lugovaya2005, Nemirko2005}. 20-second ECG recordings were collected from 90 subjects, and are currently available on Physionet. For each subject, the database has between 2 and 20 recordings (a total of 310) collected over six months. The signals were acquired from Lead I using limb-clamp electrodes at the wrists;

\item \emph{E-HOL 24h Holter}: This is an ECG database, focused on biometrics, from the University of Rochester\footnote{University of Rochester Medical Center, Telemetric and Holter ECG Warehouse. Database E-HOL-03-0202-003. Available on: \url{http://thew-project.org/Database/E-HOL-03-0202-003.html}.\label{ref:uroch}}. A total of 203 healthy subjects were recorded using a Holter monitor for 24 hours, with four electrodes placed on the chest, from 3 leads following a pseudo-orthogonal configuration;

\item \emph{European ST-T}: The European ST-T database \cite{Taddei1992} was originally intended for the analysis of ST and T-wave changes. The database is composed of 90 two-hour excerpts of recordings from 79 subjects, from 2 leads, and includes abnormalities with origin in myocardial ischaemia, hypertension, ventricular dyskinesia, and effects of medication;

\item \emph{FANTASIA}: The FANTASIA database~\cite{Iyengar1996}, available on Physionet\footnote{FANTASIA database. Available on: \url{https://physionet.org/content/fantasia/1.0.0/}.}, is composed of 120 min. recordings of ECG, respiration, and blood pressure signals from forty people (twenty young and twenty old). All signals were acquired at 250~Hz as the subjects remained at rest, supine, watching Disney's Fantasia movie;

\item \emph{Long-Term ST}: The LTST database \cite{Jager2003}, available on Physionet, includes a variety of ST segment changes for the development of algorithms for the diagnosis of myocardial ischaemia. This database includes 86 records from 80 subjects, from ambulatory recordings between 21 and 24 hours, from two and three leads;

\item \emph{MIT-BIH Arrhythmia}: The MIT-BIH Arrhythmia database \cite{Mark1982, Moody1990}, one of the most used in ECG-based biometrics research, is available at the Physionet repository. The database is composed of a total of 48 signals, 30 minutes long excerpts from ambulatory two-lead recordings. The 47 subjects were selected to obtain a representation of a wide variety of arrhythmias;

\item \emph{MIT-BIH Normal Sinus Rhythm}: This database is composed of excerpts from 18 subjects, from the MIT-BIH Arrhythmia database \cite{Mark1982, Moody1990}, presented above, deemed to be free from arrhythmias or other abnormalities;

\item \emph{Physionet 2017 CinC Challenge}: This ECG database is available on the Physionet repository, and was used for the 2017 Computers in Cardiology (CinC) Challenge, consisting of arrhythmia detection in short single-lead ECG signals. It includes individual 10-60s recordings from a total of 8528 subjects, acquired at 300 Hz sampling frequency using the AliveCor KardiaMobile off-the-person acquisition device;

\item \emph{PTB}: The PTB Diagnostic ECG database \cite{Bousseljot1995, Kreiseler1995} includes 549 recordings from 290 healthy subjects and individuals with various cardiac conditions (such as myocardial infarction, dysrhythmia, hypertrophy, or heart failure). It has 1 to 5 recordings per subject, ranging between 38.4 and 104.2 seconds, from all 12 standard and 3 Frank leads;

\item \emph{PTB-XL}: From the creators of the PTB Diagnostic ECG database, described above, the PTB-XL database~\cite{ptbxl1, ptbxl2} gathers a very large number of ten-second ECG signals ($21~837$) from over eighteen thousand subjects in clinical scenarios. The signals are sampled at $500$ Hz (but also available at $100$ Hz) and include the twelve standard leads, annotated by up to two cardiologists considering diagnostic, form, and rhythm;

\item \emph{QT}: The QT database aims to aid the development of automatic methods of measurement of QT waveforms~\cite{Laguna1997}. This collection is a compilation of 105 15-minute relevant recording extracts from other public databases;

\item \emph{UofTDB}: The University of Toronto ECG Database \cite{Wahabi2014} was specifically created for biometrics and addresses several important criteria for a thorough evaluation of biometric performance. The off-the-person ECG signals were captured using dry electrodes at the thumbs of a total of 1019 subjects. For each subject, the database includes up to six recordings over a period of six months, in various postures: supine, tripod, exercise, sitting, and standing.
\end{itemize}

While many researchers opt to use private acquisitions of data for their studies on ECG biometrics, public datasets have been crucial in allowing the appropriate comparison of results across publications. Nevertheless, if our goal is to increase competitiveness between ECG-based biometrics and more developed traits, we should address some concerns regarding public collections.

Currently, countries like India, China, and the United States, are starting to invest in nationwide identification systems for their large populations~\cite{Jain2012}, which awakens the need for biometric systems that can robustly discriminate between several million enrolled subjects. To keep up with this trend, we need to work towards the creation of public ECG collections with a larger number of subjects. PTB-XL is impressive in this regard, as it includes signals from over eighteen thousand subjects. However, it is still quite limited considering the extent and diversity of data per subject, which are very important aspects for biometrics.

Moreover, researchers can currently choose from small on-the-person datasets that include health conditions and longer acquisition times (such as the AHA, European ST-T, and the MIT-BIH Arrhythmia databases), or the off-the-person UofTDB collection with short recordings from several healthy subjects. This calls for the creation of a public database with a number of subjects similar or superior to UofTDB, with several longer off-the-person recordings (ideally over one hour), taken over long periods (months to years), during different activities and postures. ECG-BG, used by \citet{Ingale2020}, is composed of off-the-person data from 1119 healthy and unhealthy subjects, but it is still not publicly available.

Finally, it would also be very beneficial to have publicly available collections of signals acquired using recent wearable and seamless technologies, such as the CardioWheel and the Nymi Band. The highly acceptable acquisition settings offered by such products places, undoubtedly, new challenges on signal noise and variability, that would be very useful for the development of robust biometric algorithms.

\section{Related Work}

Despite being more recent and less developed than the face or fingerprint, the electrocardiogram (ECG) is quickly growing as a biometric trait, especially due to its inherent liveness and anti-spoofing capabilities.

\begin{figure}
    \centering   \includegraphics{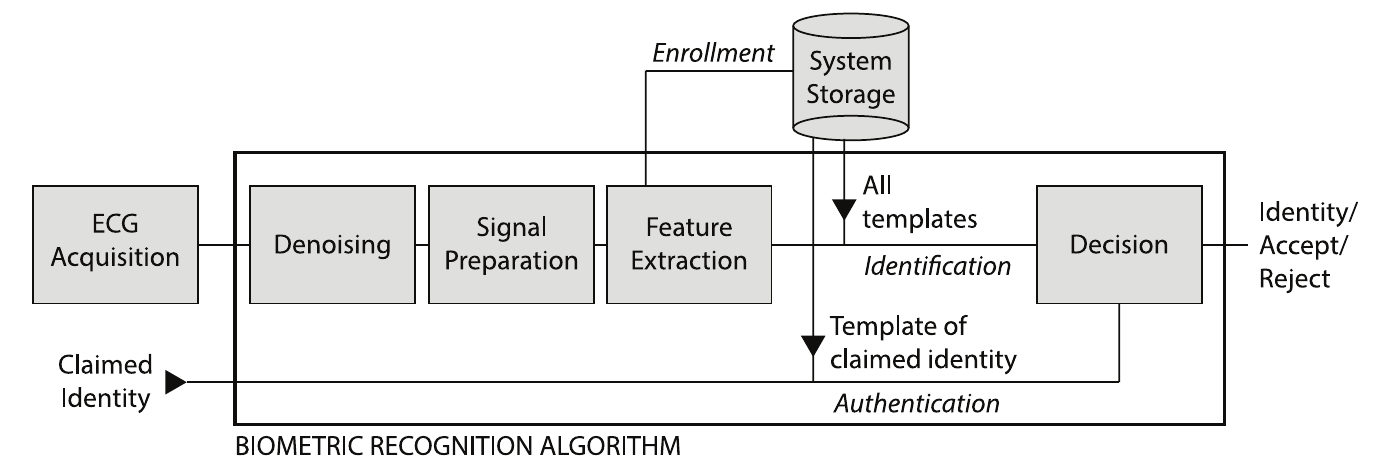}
    \caption[General structure of an ECG-based recognition system.]{General structure of an ECG-based recognition system (from \citep{Pinto2019Deep}).}
    \label{fig:ecg_recog_structure}
\end{figure}

A comprehensive review of prior art in ECG-based biometric recognition is available in~\citep{Pinto2018}, published in the scope of this doctoral work, and succinctly summarised in Table~\ref{tab:ecg_unimodal_works}. In this section, we present a summary of the survey, organised in the four common stages of an ECG-based recognition algorithm: signal denoising, signal preparation, feature extraction, and decision (see Fig.~\ref{fig:ecg_recog_structure}). We also offer a discussion on the approaches based on deep learning and the current challenges and possibilities in the topic.

\subsection{Signal denoising}\label{sec:denoising}

ECG signals are highly susceptible to interference during the acquisition stage~\cite{Rezgui2016}. The amplitude of their waveforms may vary depending on the electrode characteristics and placement but, under ideal conditions (using chest leads in medical settings), the QRS complex only reaches $2-3$ mV, the largest amplitude of the whole cyclic beat \cite{Fatemian2009}.

This means that when the electrodes are placed far from the heart, the signal is weaker and the noise is more dominant. This can result in different interference types, such as \emph{powerline interference} (PLI) from alternating current energy lines, \emph{baseline wander} from breathing movements, \emph{electrode movement} from motion, \emph{lead reversal} due to electrode mispositioning, or \emph{pacemaker interference} \cite{Singh2015,Fatemian2009}. The stage of signal denoising is, thus, of utmost importance for an ECG biometric system.

On the first initiatives in ECG biometrics, using on-the-person acquisitions, the signal-to-noise ratio was higher, and noise sources were mainly limited to powerline interference and baseline wander. Hence, filters, such as bandpass (BPF), lowpass (LPF), highpass (HPF), or notch (NF), were the first and have been the most frequent option, due to their simplicity and lower computational cost. Bandpass filters have been most common, with bands between $1-40$ Hz \cite{Agrafioti2008,Agrafioti2012,Belgacem2012,Louis2016}, $2-40$ Hz \cite{Israel2005, Safie2011, Rezgui2016}, or $2-30$ Hz \cite{Coutinho2010,Coutinho2011,Coutinho2013}, aiming to keep most useful individual information of the ECG while attenuating low and high-frequency noise.

Recently, \citet{Choudhary2015} proposed the use of the Discrete Cosine Transform (DCT) for simultaneous removal of baseline wander and powerline interference, which proved more successful than bandpass filters, when compared on simulated scenarios. The Discrete Wavelet Transform (DWT) has also been proposed for denoising of on-the-person signals \cite{Fatemian2009, Sasikala2010, Chun2016}, as it allows to decompose the signal into several levels, which may be separately processed to eliminate noise in certain frequency ranges.

When considering off-the-person approaches, wearables, or seamlessly integrated acquisition settings, it is reasonable to expect a considerable increase in the noise influence, with a lower signal-to-noise ratio. The ability to capture the ECG signal weakens, so the amplitude of the ECG components is smaller when compared with chest leads, and movement artefacts are much more frequent and dominant \cite{Lourenco2014, Matos2014, Silva2014}. 

For these, filters have also been widely applied \cite{Matos2014, Louis2016}, as well as DWT \cite{Hejazi2016}. However, the enhanced noise content motivated the proposal of new approaches based on line fitting algorithms, such as fitting of polynomial curves and the Savitzky-Golay algorithm \cite{Savitzky1964}. Their use or combination with moving average or median filters has been shown more successful than filters or transform denoising \cite{Molina2007, Pinto2017}, likely because noise is widely present across the ECG frequency range, and such methods avoid restricting their operation to narrow frequency ranges. 

Considering all this, it is possible to conclude that the trend in signal denoising has been the evolution towards methods that can adapt to increasingly unexpected and dominant noise. Considering the efforts devoted to more acceptable and comfortable acquisition settings, with an increasing focus on wearables and seamless settings, it is unreasonable to expect this trend would be reversed in the near future.

While filters appear to be a wise option if the noise is confined to known frequency ranges outside the ECG frequency range, for on-the-person signals, transforms (especially DCT) have shown to be good alternatives for denoising without causing distortions \cite{Choudhary2015}. However, when the noise is widespread and/or its frequency range is unpredictable (such as with off-the-person signals), line fitting algorithms such as the Savitzky-Golay filter may be a better option, as they smooth the signal without making strong assumptions about its noise content.

Nevertheless, research must continue to work towards increasingly robust and adaptable denoising methods. Researchers have recently started to use deep learning methodologies (as discussed further on in subsection \ref{subsubsec:deep}), that have shown remarkable robustness to noise and variability in several pattern recognition applications \cite{Hannun2014, Zhang2017b}. These, along with a deep study of data augmentation, may result in better alternatives to current and future methods devoted to signal denoising, and should certainly be explored in depth.

\subsection{Signal preparation}\label{subsubsec:signal_preparation}

ECG biometric algorithms frequently resort to the application of several processing operations over the acquired ECG signal, between denoising and feature extraction. These have the main goal to prepare the signal for the feature extraction phase, maximise the performance of the system (by reducing persistent noise and variability), segmenting specific useful parts of the acquired signal, and/or discarding undesirable or prejudicial parts \citep{Lourenco2014, Islam2017, Pinto2017}.

The noise and variability that may remain after the signal denoising stage, which this stage will aim to attenuate, are generally segment length and alignment inconsistencies, amplitude variations, heart rate variability, movement artefacts, and contact loss or impedance artefacts \citep{Irvine2008, Pinto2017, Islam2017}.

To fulfil its objective, this stage generally consists of reference point detection, signal segmentation, amplitude normalisation, time normalisation, and/or outlier detection processes. The most common in the literature methods are fiducial point detection, signal segmentation, and amplitude normalisation. Below, an overview of these processes is presented.
\begin{itemize}
    \item \emph{Fiducial Point Detection}: To aid posterior processes, such as signal segmentation, the preparation of the signal for recognition can include a step of detection of heartbeat reference points, designated as \emph{fiducials}. The majority of the surveyed research works have used this technique, varying in the methods used. On the other hand, some researchers have opted to make their algorithms completely non-fiducial, discarding the processes included in this stage \citep{Plataniotis2006, Agrafioti2008, Hejazi2016}.
    
    The Pan-Tompkins algorithm \citep{Pan1985}, based on moving-window integration and adaptive thresholding, has been the most frequent choice for fiducial detection \citep{Palaniappan2004, Venkatesh2010, Shen2011, Louis2016, Tuerxunwaili2016}. Alternatives include the Discrete Wavelet Transform (DWT) (used in \citep{Fatemian2009, Fatemian2010, Sasikala2010, Pal2018}), the Trahanias algorithm \cite{Trahanias1993} (based on morphological operations and adaptive thresholding, used in \citep{Molina2007, Pinto2017}), and the Engelse-Zeelenberg algorithm \cite{Engelse1979, Lourenco2012b} (based on differentiation, negative lobe detection, and adaptive thresholding, used in \cite{Lourenco2012, Louis2016, Pinto2017}). \citet{Pinto2017} have applied these methods and found Pan-Tompkins and Engelse-Zeelenberg gave better results for on-the-person signals, while Trahanias performed better in off-the-person settings. 
    
    \item \emph{Signal Segmentation}: Signal segmentation is used to limit the signal span for feature extraction, or to set a fixed size to ease template matching when the feature is the signal itself. In some cases, the segmentation uses the fiducial point locations and is used to crop QRS complexes and/or other waveforms \citep{Tawfiq2010, Sufi2011, Tuerxunwaili2016}. It can also be used to crop the whole heartbeat (or a majority of it), thus being performed at fixed distances before and after detected R-peaks or QRS complexes \cite{Zhou2014, Carreiras2016, Louis2016}. Some research works included signal segmentation using sliding windows, with or without overlap, regardless of the completeness of the heartbeat cycles inside it \cite{Odinaka2010, Matta2011, Ergin2014}.

    The alignment and averaging of various signal segments are closely related to the signal segmentation process. The alignment is generally performed using the R-peaks as a reference after these are located, or it is performed using cross-correlation \cite{Belgacem2012, Coutinho2013, Lourenco2014, Choudhary2015}. It usually serves as a way to ensure the template and the collected signal are not affected by variability, which distorts the personal information the signal contains, and could threaten the recognition task.

    \item \emph{Amplitude and Time Normalisation}:  As previously discussed, the electrocardiogram varies over time with several factors. These include differences in acquisition equipment or in the interaction of the subject that may cause differences in signal amplitude and DC offset \cite{Irvine2008}, or heart rate variability that causes significant changes in the duration of the heartbeats and their waveforms. To mitigate this, some methods include amplitude and time normalisation techniques.

    Amplitude normalisation techniques include the \emph{min-max} technique \cite{Irvine2008} (used in\cite{Fang2009,Li2010,Safie2011}), which normalises the signal to the range $[0,1]$, the $z-score$ method \cite{Odinaka2010}, which subtracts the signal and divides it by its standard deviation, or the \emph{max-div} method \citep{Tawfiq2010, Lourenco2011}, which divides the signal by the maximum amplitude value (generally, the R-peak amplitude).

    Time normalisation techniques aim to reduce the impact of heart rate variability on the electrocardiogram's heartbeats. Some methods perform normalisation by simply shrinking the segmented signal to a predefined length through resampling \citep{Saechia2005, Li2010, Lourenco2011}. As the heartbeat does not expand uniformly with lower heart rates, \citet{Tawfiq2010} normalised only the QT waveform, more prone to heart rate variations, using the Framingham study formula. This formula computes the linearly corrected QT duration ($QT_{LC}$), based on the time between the nearest R-peaks ($RR$) and the original duration of the waveform ($QT$), using $QT_{LC} = QT + 0.154 (1-RR)$. \citet{Fatemian2009} went further, segmenting each ECG heartbeat into its key waveforms (P, QRS, and T), and individually resampling them, before joining them back together, with regulated intervals between them. By reducing the effects of heart rate variability and avoiding the typical distortion of the individual waveforms, this is likely the best technique for time normalisation. However, it requires the detection of several waveforms' onset and offset fiducial points, making it potentially unreliable for off-the-person or seamlessly acquired signals.
    
    \item \emph{Outlier Detection}: Outlier detection is generally applied to discard false or deflected heartbeats, segmented from unacceptably noisy signal portions affected by movement or impedance artefacts or contact loss \cite{Pinto2017}. Such methods should be able to discriminate between normal deflections, noise interference, and health-related deflections. DMEAN was proposed by \citet{Lourenco2013} specifically to reject heartbeat outliers. It verifies the compliance of candidate heartbeats with four rules, regarding the distance to the average template, the minimum and maximum amplitudes, and the position of the maximum heartbeat amplitude (which must correspond to the R-peak location).
    
    \citet{Louis2016} opted to use Gaussian Mixture Models (GMM) as a supervised method for outlier detection, after being trained on a set of known clean and desirable heartbeats. \citet{Pinto2017} proposed a clustering algorithm, NCCC, based on normalised cross-correlation between candidate heartbeats. While GMM, due to the supervised data, can easily be biased towards certain patterns or subjects seen during training, clustering-based approaches like NCCC are not susceptible to this issue but can become unreliable for small sets of candidate heartbeats. Nevertheless, with the rise of off-the-person, wearable, and seamlessly integrated acquisition settings, it is expected that robust outlier detection methods will be increasingly necessary.
\end{itemize}

\subsection{Feature extraction}

The stage of feature extraction aims to translate the acquired signal into a representation that reduces the effects of remaining noise and intrasubject variability and emphasises differences between subjects. Several feature extraction methods have been proposed for ECG biometrics, which are generally grouped into three types -- \emph{fiducial}, \emph{non-fiducial}, or \emph{hybrid} approaches \cite{Matta2011,Matos2013}. 
\begin{itemize}
    \item \emph{Fiducial Approaches}: Fiducial approaches are those that exclusively use as features measurements of fiducial landmarks of the ECG signal in the time domain. These measurements vary widely throughout the state-of-the-art, including time intervals, amplitude, widths, and angles based on the heartbeat waveforms P, Q, R, S, and T, their onset and offset points \citep{Israel2005, Zhang2006, Shen2011, Rezgui2016, Tuerxunwaili2016}.
    
    Nevertheless, these approaches present the significant drawback of requiring the previous localisation of several fiducial points in the ECG heartbeats (see subsection \ref{subsubsec:signal_preparation}), which proves difficult to satisfy when using off-the-person or seamless signals. Hence, fiducial feature extraction approaches were considerably more frequent in early research works.
    
    \item \emph{Non-Fiducial Approaches}: Non-fiducial approaches are those that use the entirety of the signal (or segments of it), holistically, to extract features related to the waveform morphology \cite{Matta2011,Lourenco2011,Matos2013}. These approaches include the use of Fourier, Wavelet, or cosine transforms \citep{Saechia2005,Odinaka2010,Tawfiq2010,Ye2010,Belgacem2012, Matos2013, Matos2014, Pinto2017}, autocorrelation coefficients \citep{Plataniotis2006, Agrafioti2008,Agrafioti2012,Hejazi2016}, cardioid graphs \citep{Sufi2011, Iqbal2014}, generated tridimensional vectorcardiograms (TVCG) \citep{Dong2018}, multiresolution local binary patterns \cite{Louis2016}, and information-theoretical approaches based on Lloyd-Max quantisation~\cite{Coutinho2011,Coutinho2013} and Kolmogorov complexity~\cite{Bras2015}.

    Some methods do not perform feature extraction, alternatively using segmented heartbeats, average ensemble heartbeats, or segments between consecutive R peaks as features \cite{Lourenco2012, Labati2014, Carreiras2016, Molina2007, Zhou2014}. While more applicable to noisier signals, non-fiducial have still to reach the near-perfect performance reported by earlier works using fiducial features.

    \item \emph{Hybrid Approaches}: Hybrid approaches are those that use features from both fiducial and non-fiducial origins. These are rare among the surveyed literature works and include the approach from \citet{Palaniappan2004}, which combined common amplitude, interval and width fiducial features with a non-fiducial QRS complex form factor. \citet{Ergin2014} proposed the fusion of QRS fiducials, with several time-domain, Wavelet transform, and Power Spectral Density (PSD) features. Also, \citet{Dar2015b} opted for the extraction of a total of 46 features from Haar transform and heart-rate-variable R-R intervals.
\end{itemize}

Through the analysis of the surveyed research works, it is possible to conclude that fiducial approaches generally contribute more towards a high-performance biometric system, as the use of specific measurements reduces useless information to a minimum, and allows for feature sets with fewer dimensions. However, as noise increases, the relevance of robustness overcomes that of accuracy, and the former can only be offered by non-fiducial methods. The ideal feature extraction method would be one that combines the conditions for high performance offered by fiducial approaches with the robustness to noise and variability offered by non-fiducial approaches, perhaps using deep learning networks and their characteristic robustness to noise and versatile feature extraction capabilities.

Extracted features may, additionally, suffer dimensionality reduction to improve performance \cite{Wahabi2014}. Although frequently overlooked, dimensionality reduction has a very important goal in biometric systems, as the number of features extracted by biometric algorithms can easily become too high for a time-efficient and reliable recognition process \cite{Fratini2015}. Thus, dimensionality reduction aims to select or transform the extracted features, to reduce its number to a more computationally viable number, while keeping the maximum discriminant power to ensure the system's recognition performance \cite{Wahabi2014}.

Dimensionality reduction methods in the surveyed literature range from common methods such as Linear Discriminant Analysis (LDA) \citep{Agrafioti2008, Boumbarov2009,Matta2011,Agrafioti2012, Hejazi2016} and its Fisher (FLDA) \citep{Pathoumvanh2014} and Heteroscedastic variants (HLDA) \citep{Li2010}, Principal Component Analysis (PCA) \citep{Hejazi2016} and Kernel PCA (KPCA) \citep{Hejazi2016}, or Greedy Best-First Search (GBFS) \citep{Dar2015,Dar2015b}, to rarer methods such as Discrete Cosine Transform (DCT) \citep{Plataniotis2006}, Wilkes' lambda stepwise correlation \citep{Israel2005}, correlation matrices \citep{Biel1999, Biel2001}, or bin selection based on symmetric Kullback-Leibler divergence \citep{Matos2013,Matos2014}.

The work performed by \citet{Plataniotis2006}, \citet{Agrafioti2008}, and \citet{Hejazi2016} provides an adequate platform for the comparison of dimensionality reduction algorithms. According to their findings, LDA offers better performance than unsupervised techniques such as PCA and DCT coefficients, despite its supervised nature that requires knowledge of the subjects prior to the deployment of the biometric system \citep{Matta2011}. More recently, other supervised techniques such as the non-linear KPCA method \citep{Hejazi2016, Pal2018} rendered even better results. Hence, research should probably focus on more sophisticated dimensionality reduction methods and deep learning methodologies, which are tunable to provide optimised non-linear dimensionality reduction.

\subsection{Decision}

Based on the representation of the ECG acquisition, obtained through processes of feature extraction and dimensionality reduction, the decision stage aims to accurately attribute one of the enrolled identities to the user, in the case of identification tasks, or to accept or reject an identity claim, for authentication tasks~\cite{Bolle2004,Grother2010,Agrafioti2011}. In the case of identification, the decision stage usually consists of a classification process while, for authentication, the acceptance or rejection of the identity claim is generally based on a reference threshold $T$ that is applied to the prediction score. The decision stage can be based on:
\begin{itemize}
    \item \emph{Classifiers}: A classifier can be trained on enrollment templates from a set of subjects, and then be used to discriminate them, to output an accurate decision when needed. Classifiers are more commonly used for identification tasks. The most common classifiers in ECG-based biometrics are Support Vector Machines (SVM) \cite{Li2010,Ye2010, Lourenco2012,Lourenco2014, Silva2013,Lin2014, Hejazi2016,Rezgui2016}, mostly using Radial Basis Function (RBF) and Polynomial functions as kernels, Nearest Neighbour (kNN) classifiers \cite{Wuebbeler2007, Ghofrani2010, Agrafioti2012, Lourenco2012, Wang2013, Bras2015, Carreiras2016, Poree2016}, or Multilayer Perceptrons \citep{Palaniappan2004,Ghofrani2010,Iqbal2014,Waili2016}. 
    
    \item \emph{Metric-based Matching}: Some methods are based on the comparison between the currently acquired trait and the previously acquired templates, stored in the system database. The comparison is performed based on similarity or dissimilarity metrics. In ECG-Based biometrics, most metric-based matching methods have been based on distance metrics, among which the most popular was the Euclidean distance \citep{Plataniotis2006, Lourenco2011, Matta2011, Safie2011, Singh2012, Silva2013, Pathoumvanh2014, Chun2016}. However, since the Euclidean distance is regarded by some as unreliable in high dimensional spaces, some researchers have opted to use the cosine \citep{Silva2013} or the Mahalanobis distances \citep{Kyoso2000, Kyoso2001, Kyoso2001b, Guennoun2009}. Among similarity metrics, literature methods include the correlation coefficient \cite{Shen2002,Agrafioti2008,Chan2008,Fatemian2009,Fatemian2010,Sasikala2010,Shen2011}, normalised cross-correlation (NCC) \citep{Choudhary2015}, Gaussian log-likelihood \citep{Odinaka2010,Matos2013,Matos2014}, and Dynamic Time Warping (DTW) \citep{Molina2007,Venkatesh2010,Zhou2014}.
\end{itemize}

In the literature, it is hard to perform a thorough and fair comparison between the algorithms based on the results reported by the respective authors, as the data used to evaluate such algorithms is commonly not the same, or is used differently. However, it is important to compare algorithms to find the advantages and disadvantages of each and find opportunities for improvement. Hence, to help the comparison of state-of-the-art methods in terms of reported performance, the results of the surveyed publications that have used the six most common data collections (see Fig.~\ref{fig:pubs_per_collection}) -- PTB, ECG-ID, MIT-BIH NSR, MIT-BIH Arrhythmia, UofTDB, and CYBHi -- are presented in Tables~\ref{tab:results_ptb},~\ref{tab:results_ecg_id},~\ref{tab:results_mit_nsr},~\ref{tab:results_mit_arrh},~\ref{tab:results_uoftdb}, and~\ref{tab:results_cybhi}, respectively.

SVM and kNN have shown superior performance among traditional classifiers, even in situations with increased noise and variability. Hence, it is safe to assume that these would be wise options for new ECG biometric algorithms. However, there is the need for an equally accurate alternative that would not require re-training with every subject enrollment or update (as SVM does) or the memory-heavy storage of all subjects' templates (as kNN does). Recent studies indicate that Deep Learning models could solve these issues, but researchers will need to dedicate efforts to overcome the challenge of scarce supervised data.

\begin{table}
\caption[Results of surveyed approaches evaluated with PTB.]{Results of surveyed approaches evaluated with PTB (adapted from \citep{Pinto2018}, ordered by the number of subjects, NS; works that joined PTB with other databases are not included).}
\label{tab:results_ptb}
\centering
\begin{tabular}{>{\raggedright}p{6cm} >{\centering}p{1.25cm} >{\centering}p{1cm} >{\raggedright}p{2.5cm} >{\raggedright}p{1.25cm}}\hline
\textbf{Author} & \textbf{Year} & \textbf{NS} & \multicolumn{2}{l}{\textbf{Results}} \tabularnewline\hline
\citet{Ingale2020} & 2020 & 290 & IDR\\EER & 100\%\\2\%\tabularnewline
\citet{Ibtehaz2021} & 2021 & 290 & IDR\\EER & 99.7\%\\5.66\%\tabularnewline
\citet{Srivastva2021} & 2021 & 290 & IDR & 99.7\%\tabularnewline
\citet{Thentu2021} & 2021 & 290 & IDR & 99.5\%\tabularnewline
\citet{Chu2019} & 2019 & 290 & IDR\\EER & 99.3\%\\0.59\%\tabularnewline
\citet{Byeon2020} & 2020 & 290 & IDR & 99.1\%\tabularnewline
\citet{Wang2021} & 2021 & 290 & IDR & 98.9\%\tabularnewline
\citet{Hammad2019} & 2019 & 290 & IDR & 98.9\%\tabularnewline
\citet{Pinto2020Explaining} & 2020 & 290 & IDR & 97.7\%\tabularnewline
\citet{Jyotishi2020} & 2020 & 290 & IDR & 97.3\%\tabularnewline
\citet{Karimian2017} & 2017 & 290 & Reliability & 97.4\%\tabularnewline
\citet{Pinto2019b} & 2019 & 290 & EER & 11.0\%\tabularnewline
\citet{Wang2020} & 2020 & 248 & IDR\\EER & 98.2\%\\2.55\%\tabularnewline
\citet{Huang2022} & 2022 & 248 & IDR\\EER & 86.3\%\\2.26\%\tabularnewline
\citet{Zhang2019c} & 2019 & 234 & IDR & 99.5\%\tabularnewline
\citet{Zhang2021} & 2021 & 234 & IDR & 98.8\%\tabularnewline
\citet{Dong2018} & 2018 & 113\\99\\14 & IDR\\IDR\\IDR & 92.8\%\\93.3\%\\98.3\% \tabularnewline
\citet{Safie2011} & 2011 & 112 & EER & 19.2\% \tabularnewline
\citet{Alduwaile2020} & 2020 & 100 & IDR & 99.9\%\tabularnewline
\citet{Wang2013} & 2013 & 100 & IDR & 99.5\%\tabularnewline
\citet{Pal2018} & 2018 & 100 & IDR & 97.1\%\tabularnewline
\citet{Wuebbeler2007} & 2007 & 74 & IDR\\EER & 98.1\%\\2.8\% \tabularnewline
\citet{Li2022} & 2022 & 71 & IDR & 95.8\%\tabularnewline
\citet{Labati2018} & 2019 & 52 & IDR & 100\%\tabularnewline
\citet{Bras2015} & 2015 & 52 & IDR & 99.9\%\tabularnewline
\citet{Coutinho2013} & 2013 & 51 & IDR\\EER &  99.9\%\\0.01\% \tabularnewline
\citet{Plataniotis2006} & 2006 & 14 & IDR\\FAR & 100\%\\0.02\%\tabularnewline
\citet{Tuerxunwaili2016} & 2016 & 14 & IDR & 96\%\tabularnewline
\citet{Zhao2013} & 2013 & 12 & IDR & 96.0\%\tabularnewline
\citet{Ghofrani2010} & 2010 & 12 & IDR & 98.6\%\tabularnewline
\citet{Paiva2017} & 2017 & 10 & IDR\\FAR\\FRR & 97.5\%\\5.71\%\\3.44\% \tabularnewline
\hline
\end{tabular}
\end{table}

\begin{table}
\caption[Results of surveyed approaches evaluated with ECG-ID.]{Results of surveyed approaches evaluated with ECG-ID (adapted from \citep{Pinto2018}, ordered by the number of subjects, NS; works that joined ECG-ID with other databases are not included).}
\label{tab:results_ecg_id}
\centering
\begin{tabular}{>{\raggedright}p{6cm} >{\centering}p{1.25cm} >{\centering}p{1cm} >{\raggedright}p{2.5cm} >{\raggedright}p{1.25cm}}\hline
\textbf{Author} & \textbf{Year} & \textbf{NS} & \multicolumn{2}{l}{\textbf{Results}} \tabularnewline\hline
\citet{Wang2020} & 2020 & 90 & IDR\\EER & 100\%\\3.3\%\tabularnewline
\citet{Salloum2017} & 2017 & 90 & IDR & 100\% \tabularnewline
\citet{Li2020} & 2020 & 90 & IDR & 98.0\%\tabularnewline
\citet{Chu2019} & 2019 & 90 & IDR\\EER & 97.8\%\\2.00\%\tabularnewline
\citet{Wu2018} & 2018 & 90 & IDR\\EER & 97.5\%\\0.52\%\tabularnewline
\citet{Ranjan2019} & 2019 & 90 & EER & 2\%\tabularnewline
\citet{Bento2020} & 2020 & 90 & IDR & 96.9\%\tabularnewline
\citet{Ingale2020} & 2020 & 90 & IDR\\EER & 96.7\%\\2.3\%\tabularnewline
\citet{Zhang2019b} & 2019 & 90 & IDR\\EER & 96.3\%\\5.82\%\tabularnewline
\citet{Ibtehaz2021} & 2021 & 90 & IDR\\EER & 96.2\%\\1.29\%\tabularnewline
\citet{Benouis2021} & 2021 & 90 & IDR\\EER & 93.3\%\\3.05\%\tabularnewline
\citet{Jyotishi2020} & 2020 & 90 & IDR & 93.1\%\tabularnewline
\citet{Ivanciu2021} & 2021 & 90 & IDR\\FAR\\FRR\\Sensitivity & 86.5\%\\13.7\%\\12.7\%\\87.3\%\tabularnewline
\citet{Zaghouani2017} & 2017 & 90 & EER & 15\%\tabularnewline
\citet{Dar2015b} & 2015 & 90 & IDR\\FAR\\FRR & 83.88\%\\16.1\%\\0.3\%\tabularnewline
\citet{Dar2015} & 2015 & 90 & IDR & 82.3\%\tabularnewline
\citet{Tan2017} & 2017 & 89 & IDR & 100\%\tabularnewline
\citet{Li2022} & 2022 & 89 & IDR & 98.9\%\tabularnewline
\citet{Chun2016} & 2016 & 89 & EER & 5.2\%\tabularnewline
\citet{Kim2022} & 2022 & 83 & IDR & 95.7\%\tabularnewline
\hline
\end{tabular}
\end{table}

\begin{table}
\caption[Results of surveyed approaches evaluated with MIT-BIH NSR.]{Results of surveyed approaches evaluated with MIT-BIH NSR (adapted from \citep{Pinto2018}, ordered by the number of subjects, NS; works that joined MIT-BIH NSR with other databases are not included).}
\label{tab:results_mit_nsr}
\centering
\begin{tabular}{>{\raggedright}p{6cm} >{\centering}p{1.25cm} >{\centering}p{1cm} >{\raggedright}p{2.5cm} >{\raggedright}p{1.25cm}}\hline
\textbf{Author} & \textbf{Year} & \textbf{NS} & \multicolumn{2}{l}{\textbf{Results}} \tabularnewline\hline
\citet{Shen2002} & 2002 & 20 & IDR & 100\%\tabularnewline
\citet{Dar2015b} & 2015 & 18 & IDR\\EER & 100\%\\0\%\tabularnewline
\citet{Kim2020} & 2020 & 18 & IDR\\F-score & 100\%\\1.0\tabularnewline
\citet{Ibtehaz2021} & 2021 & 18 & IDR\\EER & 99.5\%\\5.17\%\tabularnewline
\citet{Dar2015} & 2015 & 18 & IDR & 99.4\%\tabularnewline
\citet{Ye2010} & 2010 & 18 & IDR\\FPIR & 99.3\%\\26.9\% \tabularnewline
\citet{Tan2017} & 2017 & 18 & IDR & 98.8\% \tabularnewline
\citet{Li2010} & 2010 & 18 & IDR\\EER & 98.3\%\\0.5\% \tabularnewline
\citet{Ergin2014} & 2014 & 18 & F-score & 0.97\%\tabularnewline
\citet{Zhang2019c} & 2019 & 18 & IDR & 95.3\%\tabularnewline
\citet{Zhang2017} & 2017 & 18 & IDR & 95.1\% \tabularnewline
\citet{Zhang2021} & 2021 & 18 & IDR & 93.6\%\tabularnewline
\citet{Li2020b} & 2020 & 18 & IDR & 91.4\%\tabularnewline
\citet{Venkatesh2010} & 2010 & 15 & IDR & 96\%\tabularnewline
\citet{Palaniappan2004} & 2004 & 10 & IDR & 96.2\% \tabularnewline
\citet{Camara2017} & 2017 & 10 & IDR & 94.8\% \tabularnewline\hline
\end{tabular}
\end{table}

\begin{table}
\caption[Results of surveyed approaches evaluated with MIT-BIH Arrhythmia.]{Results of surveyed approaches evaluated with MIT-BIH Arrhythmia (adapted from \citep{Pinto2018}, ordered by the number of subjects, NS; works that joined MIT-BIH Arrhythmia with other databases are not included).}
\label{tab:results_mit_arrh}
\centering
\begin{tabular}{>{\raggedright}p{6cm} >{\centering}p{1.25cm} >{\centering}p{1cm} >{\raggedright}p{2.5cm} >{\raggedright}p{1.25cm}}\hline
\textbf{Author} & \textbf{Year} & \textbf{NS} & \multicolumn{2}{l}{\textbf{Results}} \tabularnewline\hline
\citet{Salloum2017} & 2017 & 47 & IDR\\EER & 100\%\\3.4\%  \tabularnewline
\citet{Ingale2020} & 2020 & 47 & IDR\\EER & 100\%\\4\%\tabularnewline
\citet{Li2020} & 2020 & 47 & IDR & 100\%\tabularnewline
\citet{Tan2017} & 2017 & 47 & IDR & 100\% \tabularnewline
\citet{Kim2020} & 2020 & 47 & IDR\\F-score & 99.8\%\\0.99\tabularnewline
\citet{Wu2018} & 2018 & 47 & IDR\\EER & 99.7\%\\0.02\%\tabularnewline
\citet{Ye2010} & 2010 & 47 & IDR\\FPIR & 99.6\%\\12.3\% \tabularnewline
\citet{Ibtehaz2021} & 2021 & 47 & IDR\\EER & 98.2\%\\6.36\%\tabularnewline
\citet{Wang2021} & 2021 & 47 & IDR & 97.8\%\tabularnewline
\citet{Jyotishi2020} & 2020 & 47 & IDR & 96.8\%\tabularnewline
\citet{Lee2022} & 2022 & 47 & IDR & 96.0\%\tabularnewline
\citet{Dar2015b} & 2015 & 47 & IDR\\FAR\\FRR &  95.9\%\\4.1\%\\0.1\% \tabularnewline
\citet{Huang2022} & 2022 & 47 & IDR\\EER & 95.7\%\\0.36\%\tabularnewline
\citet{Chu2019} & 2019 & 47 & IDR\\EER & 94.9\%\\4.74\%\tabularnewline
\citet{Wang2020} & 2020 & 47 & IDR\\EER & 94.7\%\\2.73\%\tabularnewline
\citet{Dar2015} & 2015 & 47 & IDR & 93.1\%\tabularnewline
\citet{Zhang2017} & 2017 & 47 & IDR & 91.1\% \tabularnewline
\citet{Jahiruzzaman2015} & 2015 & 11 & IDR & 96.9\%\tabularnewline
\citet{Sasikala2010} & 2010 & 10 & IDR & 62.7\%\tabularnewline
\citet{Sufi2010} & 2010 & - & MIDR\\EER & 1\%\\0.5\% 
\tabularnewline\hline
\end{tabular}
\end{table}

\begin{table}
\caption[Results of surveyed approaches evaluated with UofTDB.]{Results of surveyed approaches evaluated with UofTDB (ordered by the number of subjects, NS; works that joined UofTDB with other databases are not included).}
\label{tab:results_uoftdb}
\centering
\begin{tabular}{>{\raggedright}p{6cm} >{\centering}p{1.25cm} >{\centering}p{1cm} >{\raggedright}p{2.5cm} >{\raggedright}p{1.25cm}}\hline
\textbf{Author} & \textbf{Year} & \textbf{NS} & \multicolumn{2}{l}{\textbf{Results}} \tabularnewline\hline
\citet{Pinto2019Deep} & 2019 & 1019 & IDR & 96.1\%\tabularnewline
\citet{Luz2018} & 2018 & 1019 & EER (raw)\\EER (spect.)\\EER (fusion) & 16.9\%\\19.4\%\\14.3\%\tabularnewline
\citet{Pinto2020Explaining} & 2020 & 1018 & IDR & 91.5\%\tabularnewline
\citet{Pinto2019b} & 2019 & 1018 & EER & 7.86\%\tabularnewline
\citet{Pinto2020iwbf} & 2020 & 1018 & EER & 12.6\%\tabularnewline
\citet{Pinto2021Secure} & 2021 & 1018 & EER & 12.6\%\tabularnewline
\citet{Louis2016} & 2016 & 1012 & EER & 7.89\%\tabularnewline
\citet{Komeili2017} & 2017 & 82 & EER (sess.)\\EER (post.) & 6.9\%\\3.7\%\tabularnewline
\citet{Ciocoiu2019, Ciocoiu2020} & 2019/20 & 52 & IDR\\EER & 95.6\%\\5.48\%\tabularnewline
\citet{Wang2020} & 2020 & 46 & IDR\\EER & 100\%\\2.17\%\tabularnewline
\citet{Huang2022} & 2022 & 46 & IDR\\EER & 87.4\%\\2.36\%\tabularnewline
\hline
\end{tabular}
\end{table}

\begin{table}
\caption[Results of surveyed approaches evaluated with CYBHi.]{Results of surveyed approaches evaluated with CYBHi (ordered by the number of subjects, NS; works that joined CYBHi with other databases are not included).}
\label{tab:results_cybhi}
\centering
\begin{tabular}{>{\raggedright}p{6cm} >{\centering}p{1.25cm} >{\centering}p{1cm} >{\raggedright}p{2.5cm} >{\raggedright}p{1.25cm}}\hline
\textbf{Author} & \textbf{Year} & \textbf{NS} & \multicolumn{2}{l}{\textbf{Results}} \tabularnewline\hline
\citet{Pinto2019b} & 2019 & 128 & EER & 16.3\%\tabularnewline
\citet{Ingale2020} & 2020 & 125 & IDR\\EER & 100\%\\0.5\%\tabularnewline
\citet{Hammad2019} & 2019 & 65 & IDR & 99.3\%\tabularnewline
\citet{Ciocoiu2019, Ciocoiu2020} & 2019/20 & 65 & IDR\\EER & 95\%\\8.6\%\tabularnewline
\citet{Belo2020} & 2020 & 63 & IDR\\EER & 100\%\\0.0\%\tabularnewline
\citet{Srivastva2021} & 2021 & 63 & IDR & 99.7\%\tabularnewline
\citet{Wang2021} & 2021 & 63 & IDR & 96.8\%\tabularnewline
\citet{Huang2022} & 2022 & 63 & IDR\\EER & 88.5\%\\1.48\%\tabularnewline
\citet{Jyotishi2020} & 2020 & 63 & IDR & 79.4\%\tabularnewline
\citet{Ibtehaz2021} & 2021 & 63 & IDR & 73.9\%\tabularnewline
\citet{Luz2018} & 2018 & 61 & EER (raw)\\EER (spect.)\\EER (fusion) & 14.1\%\\26.4\%\\12.8\%\tabularnewline
\hline
\end{tabular}
\end{table}

\subsection{Deep learning}\label{subsubsec:deep}

Deep learning methodologies are quickly revolutionising several fields in pattern recognition, galvanising the machine learning community with outstanding results and unforeseen robustness to input noise and variability in diverse tasks \cite{LeCun2015, Hannun2014, Rajpurkar2017, Um2017}. It achieved these milestones mainly due to the flexibility and robustness of convolutional layers for feature learning, the selective memory of recurrent layers connected to their previous instances, and the versatility of fully-connected layers \cite{LeCun2015, Zhang2017}. Their adaptability to scarce data through techniques such as data augmentation, fine-tuning, transfer learning, and weakly supervised learning, just add to their power for pattern recognition applications.

In the topic of ECG biometrics, the study of deep learning is still a pioneering affair. It has, however, been gathering steam throughout the past four years. Despite a few works which continued focusing on traditional feature extraction and decision models \cite{Li2020, Randazzo2020, TiradoMartin2020, Wang2020, Benouis2021}, the majority of literature methods proposed since 2020 already include some deep learning architecture responsible for the processes of feature extraction, decision, or both.

Initially, \citet{Zhang2017} proposed a multiscale CNN that receives, in parallel, selected autocorrelation coefficients of approximation and detail Wavelet transform coefficient sets of two-second ECG segments. \citet{Eduardo2017} replaced the feature extraction stage using an Autoencoder to learn lower-dimensional representations of segmented heartbeats, which were ultimately fed to a kNN classifier.

However, most researchers aim to integrate several stages into the deep learning model. \citet{Salloum2017}, after signal preprocessing and segmentation, replaced the stages of feature extraction and decision with an RNN with Long Short-Term Memory (LSTM) and Gated Recurrent Unit (GRU). \citet{Zhang2017b} replaced the stages of feature extraction and classification, by feeding 2D representations of single-arm ECG signals to a CNN. \citet{Luz2018} also integrated the feature extraction and decision stages, proposing the combined use of two separate CNN, one receiving segmented heartbeats as input and the other receiving the respective heartbeats' spectrograms, fused at score level.

\citet{Labati2018} detected, segmented, and selected QRS complexes from ECG signals, and concatenated them into a QRS vector that served as input to a unidimensional CNN that fulfilled the purposes of feature extraction and decision. With a softmax output, the method attained $100\%$ IDR on the PTB database and, with Hamming distance matching, achieved $2.75\%$ EER with long-term signals from the E-HOL 24h collection.

Several researchers have also tried to explore two-dimensional representations of ECG signals in order to use 2D deep learning models. \citet{Ciocoiu2019, Ciocoiu2020} have studied S-Transforms, Gramian Angular Fields, Phase-Space Trajectories, and Recurrence Plots of segmented heartbeats on a custom two-dimensional convolutional neural network. They found that S-Transforms offered the best performance, achieving around $95\%$ identification rates on the UofTDB and CYBHi off-the-person databases.

\citet{Bento2020} explored spectrograms as inputs to a custom 2D CNN architecture and a DenseNet. The latter achieved the best performance, with almost $97\%$ IDR on ECG-ID and $100\%$ on FANTASIA. \citet{Byeon2020} also studied DenseNets, alongside XCeption and ResNet architectures to build ensembles of models receiving spectrograms, log-spectrograms, melspectrograms, scalograms, and MFCCs, achieving approximately $99\%$ IDR on the PTB database.

More recently, \citet{Srivastva2021} used images of ECG heartbeat plots as inputs to an ensemble of ImageNet-pretrained DenseNet and ResNet models. With this methodology, they achieved $99.7\%$ IDR on PTB and CYBHi.  \citet{Thentu2021} explored continuous wavelet transform-based multi-scale representations of ECG heartbeats on various ImageNet pretrained two-dimensional architectures achieving over $99.5\%$ accuracy on both CEBSDB and PTB databases.

However, beyond the approaches proposed within this doctoral work, no truly end-to-end methodologies are present in the literature. Two-dimensional representations are promising, especially considering the possibility of using pretrained deep models, but the transformations themselves are still separately optimised processes that may lose important information and limit achievable performance.

Another promising category of approaches is temporal networks, such as LSTMs, which have attained interesting results and are a natural match with ECG signals (time series). However, the aforementioned problem is still valid: if separate processes of denoising, preparation, and/or feature extraction are added to the pipeline, the model may be limited in the information received and thus in the performance it can achieve.

\section{Open Challenges and Opportunities}

\begin{figure}
    \centering
    \includegraphics[width=\linewidth]{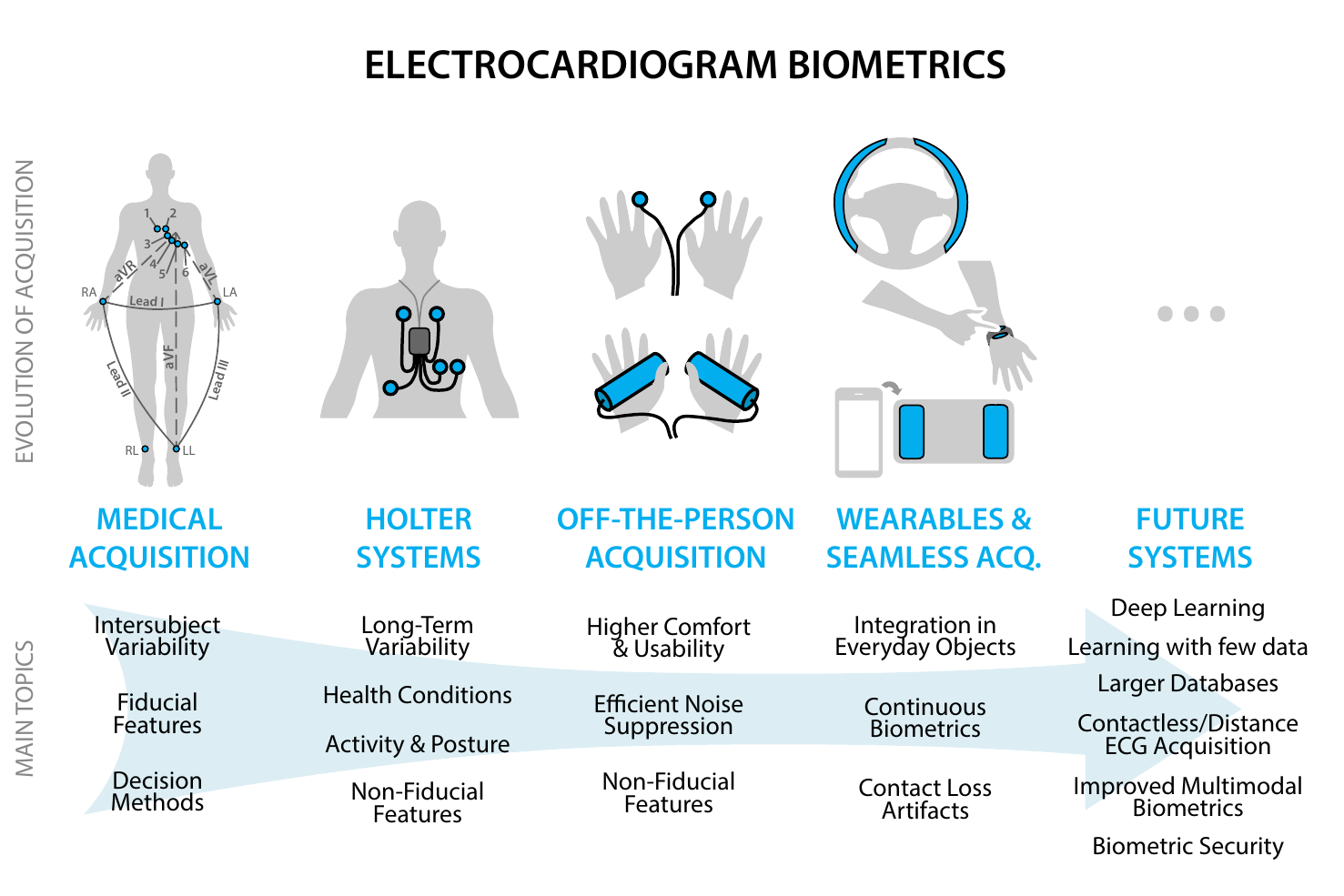}
    \caption[A panorama of ECG biometrics across time: the past, present, and future trends in ECG acquisition for biometrics and the corresponding research challenges.]{A panorama of ECG biometrics across time: the past, present, and future trends in ECG acquisition for biometrics and the corresponding research challenges (from~\cite{Pinto2018}).}
    \label{fig:ecg_history_panorama}
\end{figure}

Much of the great potential of deep learning for ECG biometrics is still to be explored. Considering the information presented and discussed throughout this chapter, one can align the main challenges in ECG biometrics with the corresponding trends in ECG acquisition, thus painting a panorama of the history and near future of ECG biometrics (see Fig.~\ref{fig:ecg_history_panorama}).

As illustrated in the aforementioned figure, the use of deep learning is a major research opportunity as we move into the future of ECG biometrics. Although deep learning typically brings significantly increased computational costs to biometric systems, these should be compensated by considerable boosts in performance and robustness due to the flexibility offered by deep models.

The integration of all processing stages in a single end-to-end deep model, alongside new techniques of data augmentation and regularisation, could enable the coordinated optimisation for individual recognition and lead us to new levels of robustness against noise and variability. Nevertheless, as we move towards ``black-box'' deep models, it is important to address the problem of trustworthiness and transparency through the study of model interpretability and explainability in ECG biometrics.

The growth of deep learning should further unveil a serious problem in ECG biometrics: the scarcity of data. This places significant hurdles on the development of accurate and robust biometric models, which should only worsen with data-hungry deep learning methodologies. As we delve deeper into deep learning methodologies for ECG biometrics, it is important to build larger and more complete off-the-person ECG collections. Additionally, researchers should devote further efforts to data augmentation, siamese architectures, triplet learning methodologies, unsupervised and self-supervised learning, and other strategies towards the mitigation of the effects of data scarcity.

Partially linked to data scarcity, another large problem currently plaguing ECG biometrics is the prevalence of unrealistic and mismatching evaluation settings. The diversity of databases and data subsets is noticeable throughout this literature review and makes it impossible to adequately compare different methodologies. The high frequency of random train/test subset splits makes results unrealistic. Moreover, the rarity of long multi-session data acquisitions makes long-term performance a hidden problem waiting to be truly unveiled and solved.

Hence, this thesis part focuses on five contributions to these open challenges and opportunities. Specifically:
\begin{itemize}
    \item In Chapter~\ref{ch:ecgiden}, we propose the first true end-to-end methodology for ECG-based identification, complete with a study on the progressive integration of pipeline stages within the deep model and various tailored data augmentation strategies;
    \item In Chapter~\ref{ch:ecgauth}, we adapt the previous model for identity verification and explore identification \emph{vs.} triplet loss training for template similarity matching. The proposed methodology is also benchmarked against state-of-the-art approaches on a restructured evaluation setup for more realistic results;
    \item In Chapter~\ref{ch:ecglongterm}, we study the effect of long-term variability on the performance of state-of-the-art approaches using a database of day-long Holter acquisitions, along with the application of template/model update strategies;
    \item In Chapter~\ref{ch:ecgexpl}, we study the relative importance of ECG waveforms, with a special focus on the QRS, throughout experimental setups with varying database size and noise/variability using interpretability tools; 
    \item In Chapter~\ref{ch:ecginterlead}, we propose an end-to-end methodology for the recovery of the entire set of twelve standard leads requiring as input just one single-lead blindly-segmented ECG signal.
\end{itemize}
\chapter[End-to-End Models and Augmentation Strategies for Identification]{End-to-End Models and Augmentation\\Strategies for Identification}\label{ch:ecgiden}

\begin{tcolorbox}\footnotesize
{\large\bf Foreword on Author Contributions}

The research work described in this chapter was conducted entirely by the author of this thesis, under the supervision of Jaime S. Cardoso and André Lourenço. The results of this work have been disseminated in the form of a chapter in a book and an abstract in national conference proceedings:
\begin{itemize}[noitemsep, leftmargin=1em, nosep]
    \item \underline{J. R. Pinto}, J. S. Cardoso, and A. Lourenço, ``Deep Neural Networks for Biometric Identification Based on Non-Intrusive ECG Acquisitions,'' in K. V. Arya and R. S. Bhadoria, Eds., \emph{The Biometric Computing: Recognition and Registration}, CRC Press, 2019.~\cite{Pinto2019Deep}
    \item \underline{J. R. Pinto}, J. S. Cardoso, and A. Lourenço, ``Improving ECG-Based Biometric Identification Using End-to-End Convolutional Networks,'' in \emph{24th Portuguese Conference on Pattern Recognition (RECPAD 2018)}, Oct.~2018.
\end{itemize}

\end{tcolorbox}

\section{Context and Motivation}

The state-of-the-art in ECG-based recognition mostly consists of pipeline algorithms, composed of separate stages of denoising, signal preparation, feature extraction, and decision, as discussed in Chapter~\ref{ch:ecgprior}. Even the most recent methods using deep learning techniques still rely on some of these separate processes. 

However, Convolutional Neural Networks (CNNs) possess the tools to integrate all phases of processing, from acquisition to decision, into a single model. This integration replaces separate, step-by-step tuning with a holistic optimisation process, synergically adapting the model to attain the best performance possible.

Furthermore, the flexibility of convolutional and fully-connected layers makes deep networks able to autonomously learn the most fitted features for the task at hand. Meanwhile, these keep the ability to generalise and be robust against high variability and noise dominance over the signals \citep{LeCun2015, Zhang2017}. Hence, it could be the key to improving the inferior performance results verified in off-the-person ECG biometrics. 

This work aimed to study the full extent of the capabilities of CNNs for biometric identification using non-intrusive ECG signal acquisitions. A CNN architecture is proposed for the complete integration of traditional pipeline stages in a single model, for higher accuracy and robustness in off-the-person settings. To obtain further improved performance, unidimensional data augmentation strategies are designed specifically for ECG-based biometrics.

\section{Methodology}

\begin{figure}%
\includegraphics[width=\columnwidth]{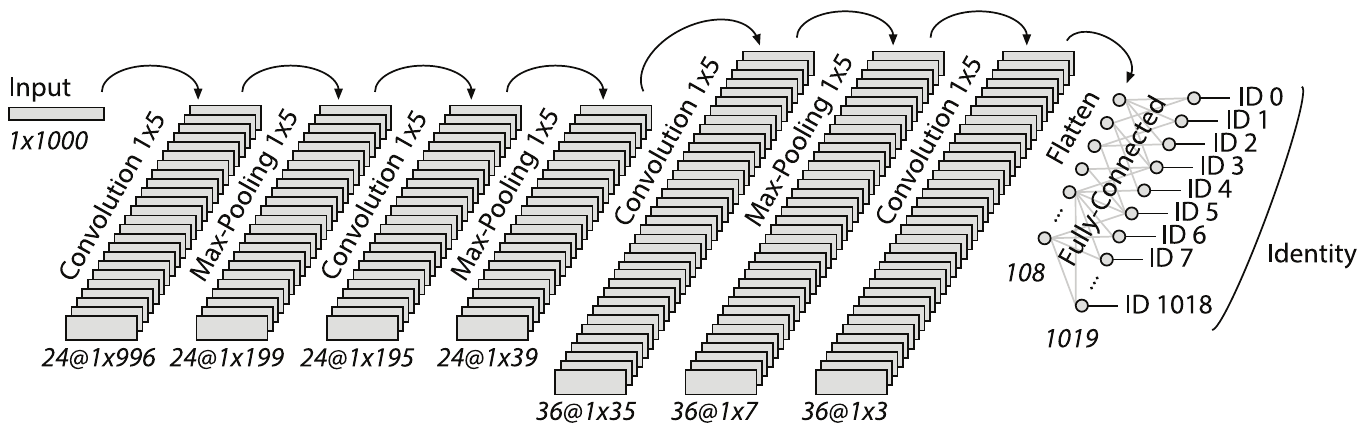}%
\caption[Architecture of the proposed CNN model for ECG-based identification.]{Architecture of the proposed CNN model for ECG-based identification (the number of neurons on the fully connected layer refers to the entire dataset with $1019$ possible identities).}%
\label{fig:endtoendiden_bc_archit}%
\end{figure}

\subsection{Model}
The proposed convolutional neural network (see Fig.~\ref{fig:endtoendiden_bc_archit}) integrates all common pipeline stages into a single, end-to-end model receiving raw five-second ECG segments and delivering the corresponding identity. It follows the typical structure of a convolutional neural network: the first part includes convolutional and max-pooling layers, and the second part includes one fully-connected layer.

Convolutional layers hold filter banks to learn the most advantageous representation of the input signal segment. Pooling quickly reduces the number of parameters, controlling the computational cost and training time, and improves robustness to small input variations. The proposed architecture uses Rectified Linear Unit (ReLU) activations, filters' size $5$, stride $1$, pooling size $5$, and pooling stride $1$.

The feature maps output by the last convolutional layer are concatenated into a single unidimensional vector of features. This serves as input to the fully-connected layer, which weighs and combines the received features at each neuron. The fully-connected layer is composed of $N$ neurons (where $N$ is the number of enrolled subjects), with softmax activations. The neuron that outputs the highest value will correspond to the predicted identity.

Based on a batch of train samples fed to the network, a measure of loss is computed by comparing the output of the network with the true labels of the batch. The weights/parameters that compose the neural network are adjusted to reduce that loss, using an optimiser function. In this work, the optimiser Adam \citep{Kingma2015} was used, with empirically adjusted initial learning rate in $[0.01, 0.001]$, and Sparse Categorical Cross-entropy loss.

To avoid overfitting, the network used dropout. Dropout will avoid learning overly specific patterns in the training data \citep{Krizhevsky2012, Srivastava2014}. They are placed between two layers and act upon the connections between them, setting the corresponding input to zero. In the proposed method, dropouts are used on the connections between the flattened vector of features and the fully-connected layer, effectively blocking the access of the classifier to a part of the features, and requiring it to become less specific to the training set, and more robust to unexpected variability and noise.

\subsection{Data augmentation strategies}
Data augmentation is used to obtain a more robust classifier. It consists of the application of small transformations or changes to the train samples while protecting the integrity of the underlying label of each sample, to simulate larger datasets and ensure the network is robust to such variability~\citep{Krizhevsky2012, Chatfield2014}. Like deep learning in general, data augmentation techniques are significantly more frequent in 2D networks (for images) than in 1D (signals).

\begin{figure}%
\includegraphics[width=\columnwidth]{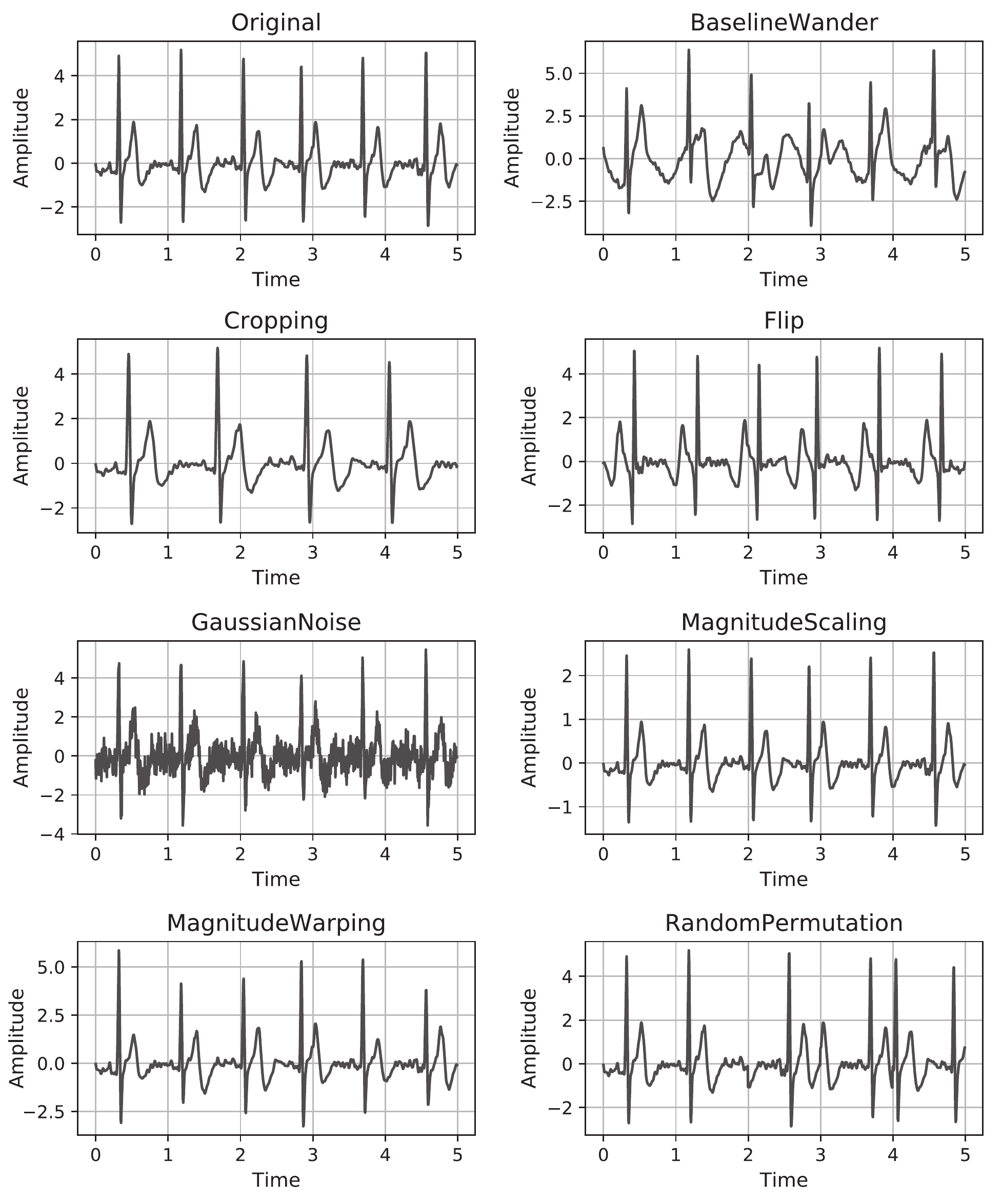}%
\caption[Illustration of the effects of the different data augmentation techniques on an example five-second ECG segment.]{Illustration of the effects of the different data augmentation techniques on an example five-second ECG segment (for easier visualisation, the original segment was denoised with a bandpass filter 1–30 Hz and had its amplitude z-score normalised).}%
\label{fig:endtoendiden_bc_da_examples}%
\end{figure}

Based on the recent work of \citep{Um2017}, and taking into account the unique characteristics of the ECG signals, seven different types of data augmentation are proposed and explored for 1D convolutional neural networks (see Fig.~\ref{fig:endtoendiden_bc_da_examples}). These types are:
\begin{itemize}
    \item \emph{Baseline Wander}: simulates a periodic undulation on the signal, by adding a sinusoidal wave with a frequency near $1$~Hz;
    \item \emph{Cropping}: takes a contiguous subsegment and resamples it to match the original length. In the case of ECG signals, this technique simulates slower cardiac frequencies;
    \item \emph{Flip}: inverts the signal along the time axis, which causes the inversion of the heartbeat waveforms and their relative locations;
    \item \emph{Gaussian Noise}: introduces Gaussian noise (with mean zero and standard deviation about ten times lower than the signal amplitude) to cause high-frequency distortions on the signal, similar to movement artefacts and powerline interference noise;
    \item \emph{Magnitude Scaling}: rescales the original train sample by multiplying it by a factor inferior or superior (but close) to $1$;
    \item \emph{Magnitude Warping}: similar to the previous technique, it rescales the signal in a non-uniform fashion, using a sinusoidal wave instead of a fixed factor, so that different parts of the signal will have their amplitude reduced or increased;
    \item \emph{Random Permutations}: divides the signal into $N$ contiguous subsegments with similar lengths, and their order is randomly changed. This may cause discontinuities in the heartbeats and their waveforms, simulating sensor faults or abrupt segment terminations.
\end{itemize}

\section{Experimental Setup}

The performance of the proposed convolutional neural network architecture, as previously described, was evaluated on off-the-person ECG recordings of the University of Toronto ECG Database (UofTDB)~\citep{Wahabi2014}. Besides the entire database of $1019$ subjects, two subsets were also used, with $25$ and $100$ subjects, to evaluate the performance in smaller datasets. The datasets were divided with $70\%$ of the data for training and $30\%$ for testing. 

\begin{figure}%
\includegraphics[width=\columnwidth]{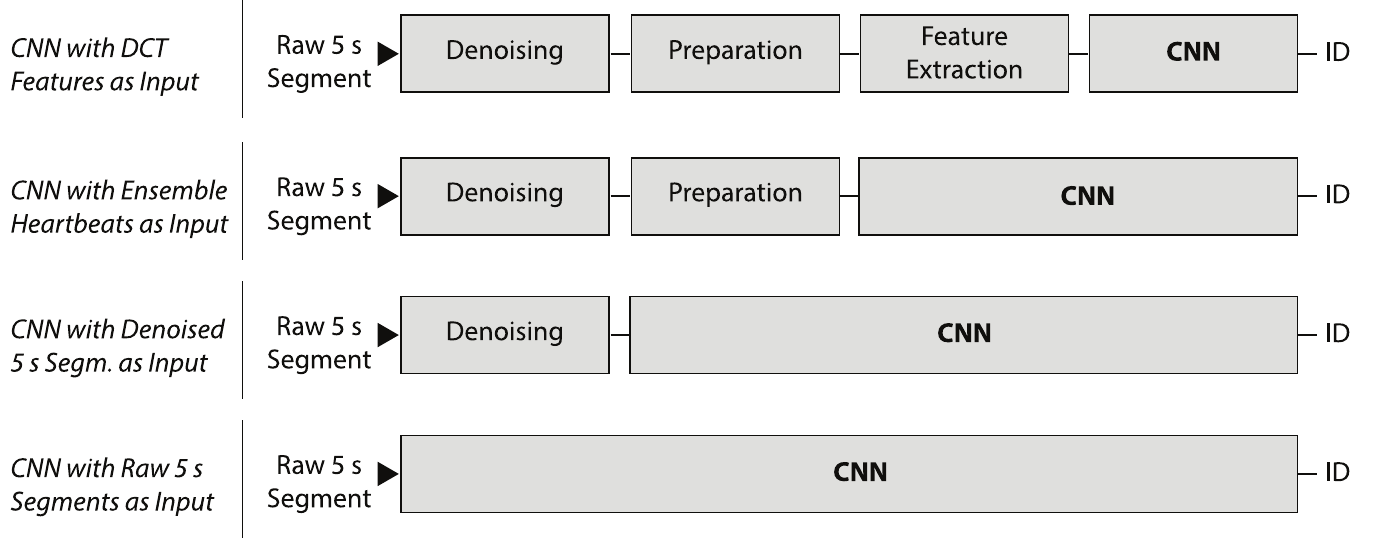}%
\caption{Illustration of the progressive phases of integration of the traditional pipeline stages into the CNN architecture.}%
\label{fig:endtoendiden_bc_integration}%
\end{figure}

The proposed method suffered slight adaptations to allow the study of the progressive integration of the traditional pipeline stages into the CNN model (see Fig.~\ref{fig:endtoendiden_bc_integration}). Thus, besides the proposed end-to-end version that receives raw five-second ECG segments, three other variants were evaluated. The first receives five-second ECG segments denoised using a $1-30$ Hz bandpass filter. The second receives the average of heartbeats detected using the Engelse-Zeelenberg algorithm and normalised to zero mean and unit variance. The third variant receives DCT features extracted from the average heartbeats. The pool size of max-pooling, which was set at $5$ for five-second segments as input, was changed to $3$ for ensemble heartbeats, or $2$ for DCT features.

The proposed method was compared with a baseline algorithm, adapted from the method proposed by \citet{Pinto2017} using SVM and kNN for decision, and the state-of-the-art algorithm based on autoencoders proposed by \citet{Eduardo2017}, and the algorithm based on AC/LDA features by \citet{Matta2011}, evaluated in the same conditions.

\begin{figure}%
\includegraphics[width=\columnwidth]{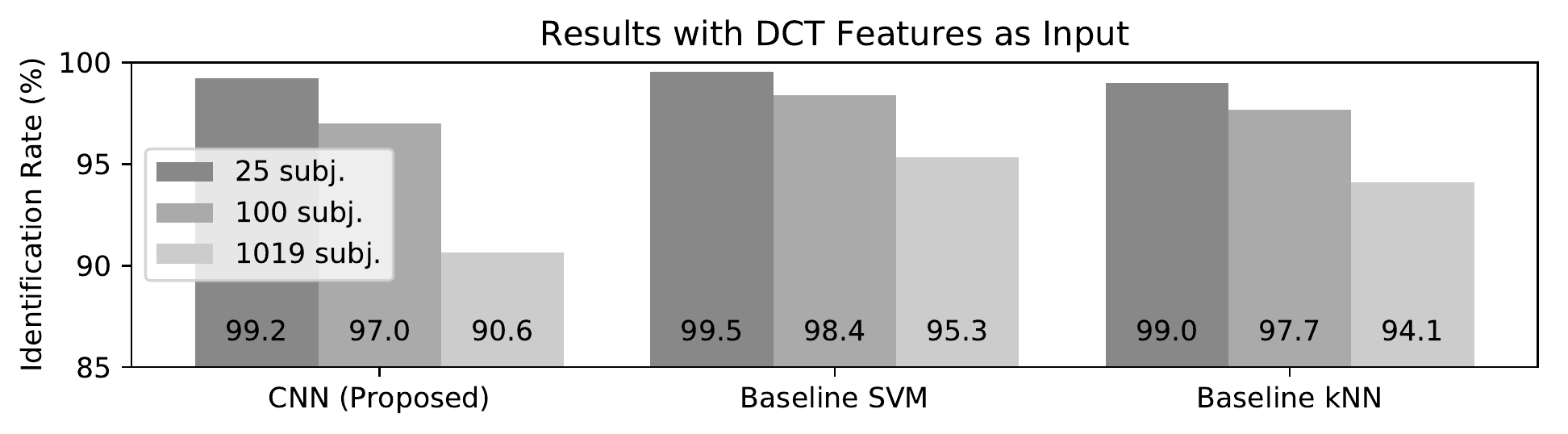}%
\caption{Results of the proposed and baseline algorithms, when using DCT features as input.}%
\label{fig:endtoendiden_bc_dct}%
\end{figure}

\section{Results and Discussion}

\begin{figure}%
\includegraphics[width=\columnwidth]{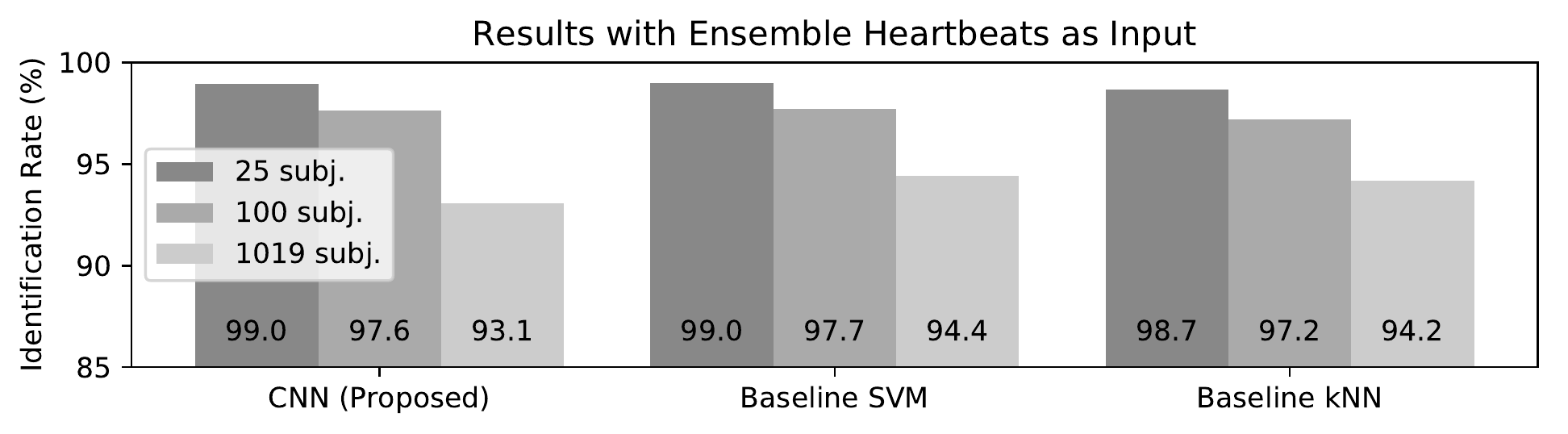}%
\caption{Results of the proposed and baseline algorithms, when using ensemble heartbeats as input.}%
\label{fig:endtoendiden_bc_ensemble}%
\end{figure}

With DCT features as input (see Fig.~\ref{fig:endtoendiden_bc_dct}), the performance of the proposed method is similar to that of the baseline algorithm for $25$ subjects. However, with the increase of subjects on the dataset (with $100$ and $1019$ subjects), the proposed algorithm falls behind. This may be caused by the very concise information that the input carries, fitted for typical pipeline algorithms as the baseline but not for deep learning. The results of the evaluation using ensemble heartbeats as input support this hypothesis (see Fig.~\ref{fig:endtoendiden_bc_ensemble}), as the performance increases and approaches that of the baseline methods, and even surpasses that of the kNN classifier on the two smaller datasets.

Integrating additional stages into the deep learning model allows us to simplify its structure, and use longer signal segments as inputs (in this case, five seconds). This means an increase in complexity of the input, which can harm the performance of the network, but also an increase in available information and variability, which can allow for a more robust model.

\begin{figure}%
\includegraphics[width=\linewidth]{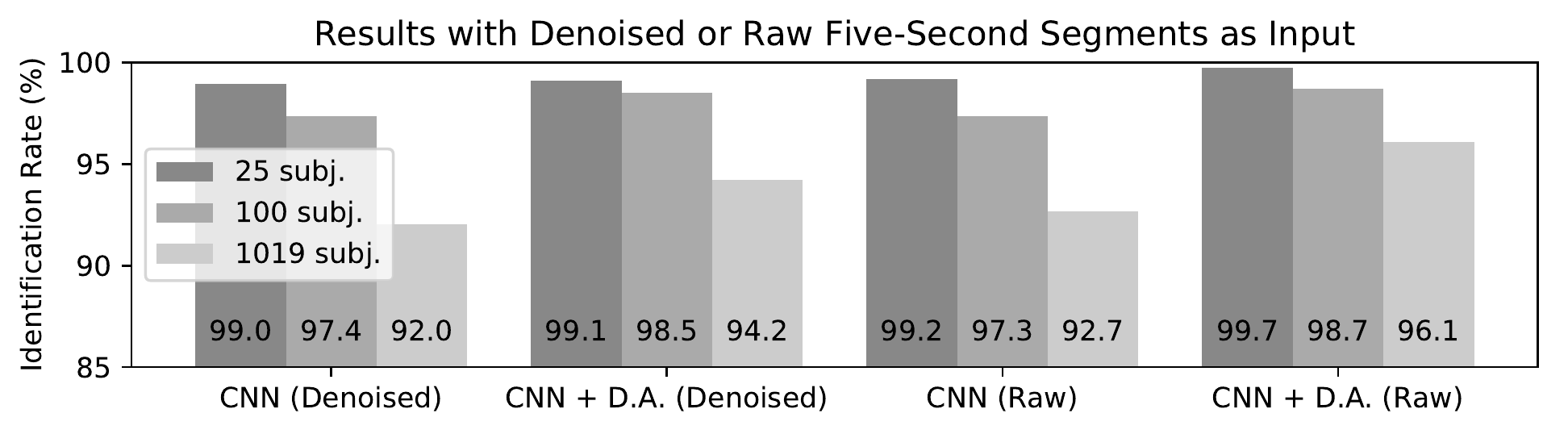}%
\caption{Results of the proposed and baseline algorithms, when using five-second ECG segments as input, raw or denoised.}%
\label{fig:endtoendiden_bc_raw_vs_denoised}%
\end{figure}

The results of the use of denoised and raw five-second segments (see Fig.~\ref{fig:endtoendiden_bc_raw_vs_denoised}) illustrate the trade-off between signal complexity and the increase of robustness due to extra information and variability, as the results were similar to those of the CNN receiving ensemble heartbeats. Moreover, in general, the results of the CNN with raw segments surpassed those of the CNN with denoised segments, which likely result from the benefit of increased variability during training.

\begin{figure}%
\includegraphics[width=\linewidth]{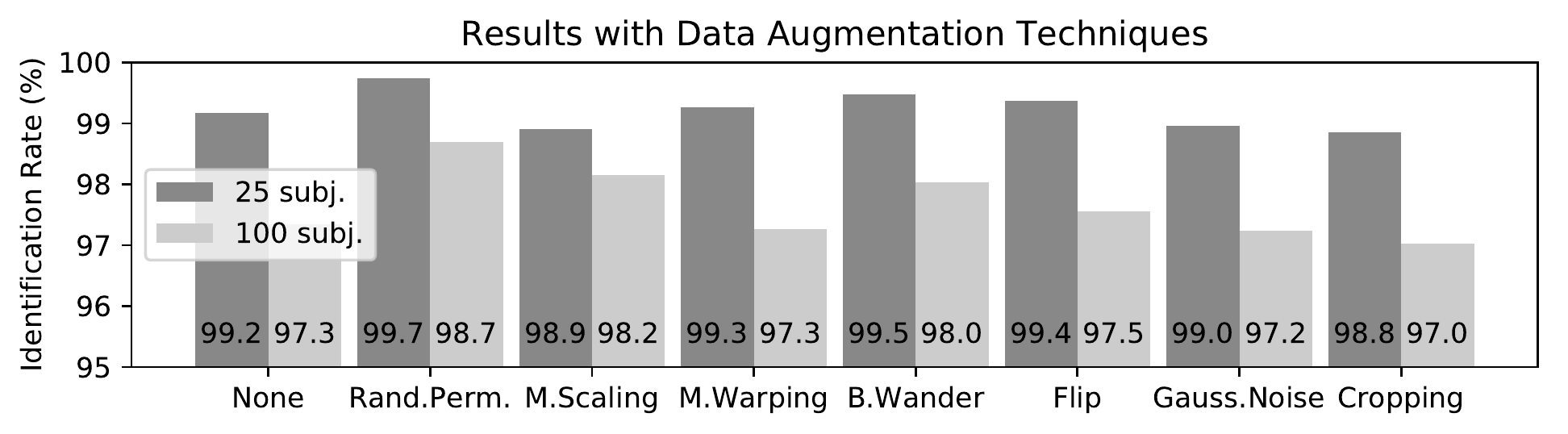}%
\caption{Results of the proposed algorithm receiving raw five-second segments, with each technique of data augmentation, on the datasets of $25$ and $100$ subjects.}%
\label{fig:endtoendiden_bc_da_1}%
\end{figure}

Increased variability is, in turn, the goal of data augmentation (D.A.). The aforementioned techniques were separately tested on the subsets of $25$ and $100$ subjects (see Fig.~\ref{fig:endtoendiden_bc_da_1}). Most of the techniques of data augmentation bring improvements to the algorithm's identification rates. The exceptions were magnitude scaling in the smallest dataset, cropping and magnitude warping in the $100$ subject dataset, and Gaussian noise in both. Likely, this performance decay is the result of a corruption of the underlying labels with these techniques.

\begin{figure}%
\includegraphics[width=\linewidth]{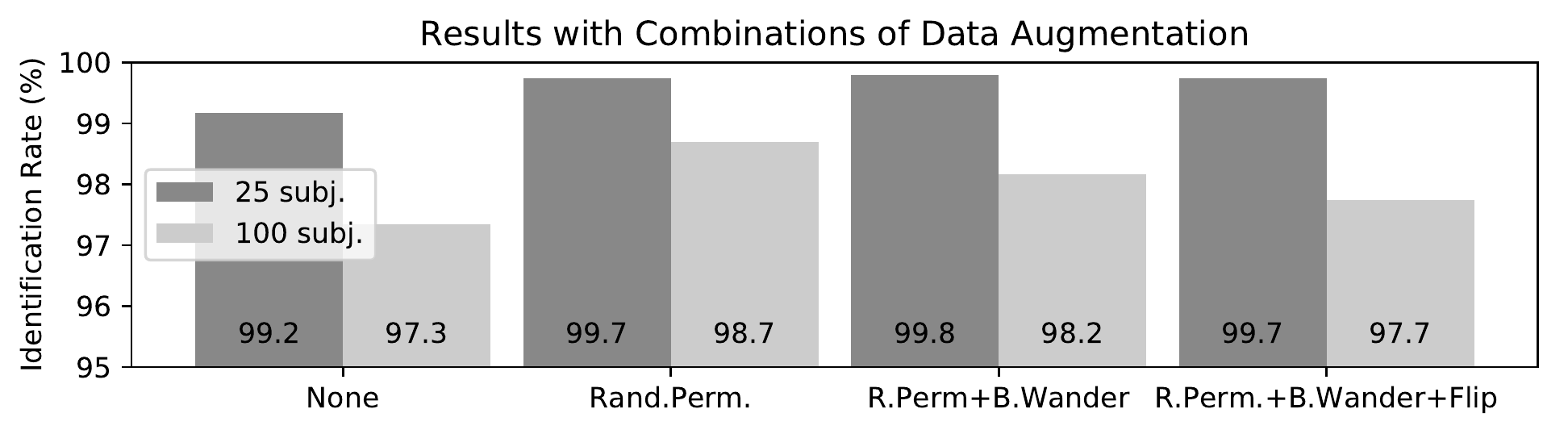}%
\caption{Results of the proposed algorithm, receiving raw five-second segments as input, with combinations of data augmentation techniques.}%
\label{fig:endtoendiden_bc_da_2}%
\end{figure}

The most promising data augmentation techniques were random permutations (that excelled in both datasets), baseline wander, and flip. These were evaluated in groups to assess if the combination of two or three techniques would offer performance improvements. The results (see Fig.~\ref{fig:endtoendiden_bc_da_2}) show that the sole use of random permutations is the best option, although the combinations also caused an improvement in identification performance.

\begin{figure}%
\includegraphics[width=\linewidth]{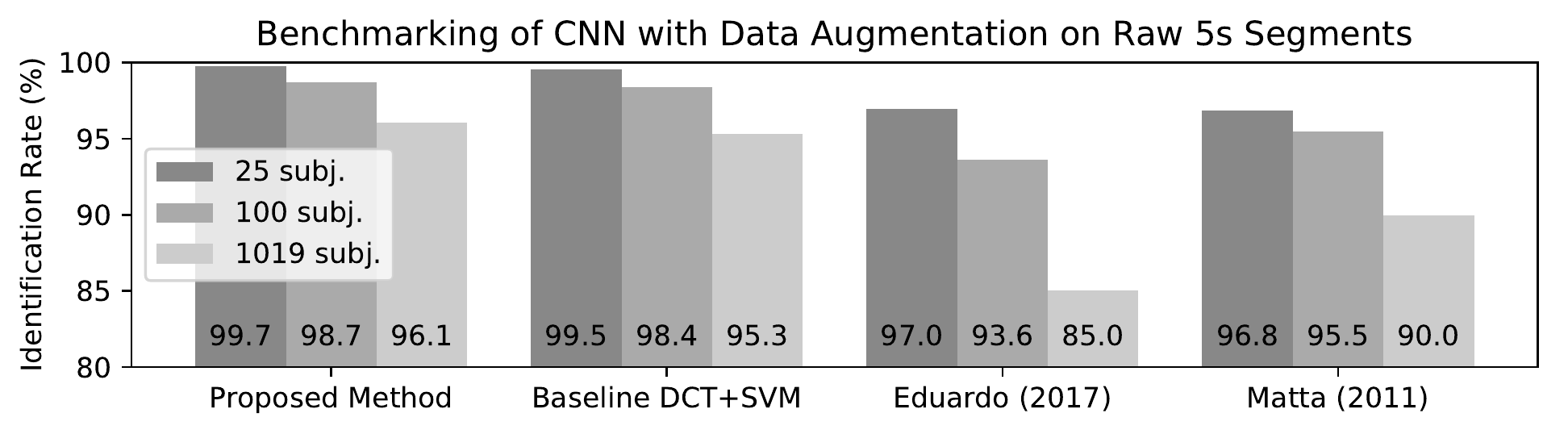}%
\caption{Direct benchmarking between the proposed architecture with the best baseline algorithm and the two implemented state-of-the-art algorithms.}%
\label{fig:endtoendiden_bc_baseline}%
\end{figure}

We compared the proposed and baseline algorithms with state-of-the-art algorithms. As they were implemented and tested in the same conditions, the algorithms of \citet{Eduardo2017} and \citet{Matta2011} can be used for a direct benchmarking (see Fig.~\ref{fig:endtoendiden_bc_baseline}). The proposed method presents better results than the alternatives and a slightly slower decay with the increase of the number of subjects, denoting better scalability to larger populations. The state-of-the-art algorithms likely suffer from using nearest neighbour classifiers, prone to overfit, as the results of the baseline algorithm with kNN were also consistently worse than with SVM. The method of \citet{Eduardo2017}, despite showing remarkably good results in the denoising of signals during our experiments (using the entire encoder-decoder), falls short in these conditions.

Finally, the results of the proposed and baseline algorithm can be compared with the results reported by the most recent prior artworks (see Table~\ref{tab:bc_benchmarking}). The IDR of the proposed and baseline algorithms may pale in comparison with some results reported in some of the considered prior works, but it is important to consider the evaluation settings. Only \citet{Wieclaw2017} used an off-the-person database, as opposed to the much cleaner signals of on-the-person databases still used by most researchers. Also, the UofTDB collection enabled the evaluation of the proposed algorithm with a much larger set of subjects than any other identification method.

\begin{table}
\caption{Comparison of the proposed and baseline algorithms with recent state-of-the-art methods.}\label{tab:bc_benchmarking}
\centering
\begin{tabular}{lllc}
\hline
\textbf{Authors}                                                    & \textbf{Brief Description}                                                            & \textbf{Dataset}                                                                    & \textbf{IDR} \\\hline
Proposed Method                                                     & \begin{tabular}[c]{@{}l@{}}Raw segments + CNN\\with data augment.\end{tabular}     & \begin{tabular}[c]{@{}l@{}}UofTDB\\   (off-the-person) - 1019 subj.\end{tabular} & 96.1\%       \\
Baseline                                                            & \begin{tabular}[c]{@{}l@{}}DCT features + SVM\end{tabular}                        & \begin{tabular}[c]{@{}l@{}}UofTDB\\   (off-the-person) - 1019 sub.\end{tabular}     & 95.3\%       \\
\citet{Salloum2017}   & LSTM-GRU RNN                                                                          & \begin{tabular}[c]{@{}l@{}}ECG-ID\\   (on-the-person) - 90 subj.\end{tabular}       & 100\%        \\
\citet{Zhang2017}     & Multiscale CNN                                                                        & \begin{tabular}[c]{@{}l@{}}Several\\   (on-the-person) - 18-47 subj.\end{tabular}   & 93.5\%       \\
\citet{Wieclaw2017}   & Heartbeats + MLP                                                                      & \begin{tabular}[c]{@{}l@{}}Private\\   (off-the-person) - 18 subj.\end{tabular}     & 89.0\%       \\
\citet{Tan2017}       & \begin{tabular}[c]{@{}l@{}}Fiducials + RF\\fused with DWT\\+ WDIST kNN\end{tabular} & \begin{tabular}[c]{@{}l@{}}Several\\   (on-the-person) - 184 subj.\end{tabular}     & 99.5\%       \\
\citet{Carreiras2016} & Heartbeats + kNN                                                                      & \begin{tabular}[c]{@{}l@{}}Private\\   (on-the-person) - 618 subj.\end{tabular}     & 84.4\%       \\
\citet{Bras2015}   & \begin{tabular}[c]{@{}l@{}}Kolmogorov-based\\compression\end{tabular}              & \begin{tabular}[c]{@{}l@{}}PTB\\   (on-the-person) - 52 subj.\end{tabular}          & 99.9\%       \\
\citet{Wang2013}      & \begin{tabular}[c]{@{}l@{}}Max-pooling of sparse\\coding coefficients\end{tabular} & \begin{tabular}[c]{@{}l@{}}PTB\\   (on-the-person) - 100 subj.\end{tabular}         & 99.5\%   \\\hline   
\end{tabular}
\end{table}


However, it is important to recall that deep learning both requires and benefits greatly from large datasets where each class is represented by a large number of samples. While, as visible in the results presented here, data augmentation attenuates the prejudicial effects of scarce data, it is difficult to acquire sufficient ECG signals from each subject to compensate for the increased noise and variability in off-the-person settings.

In the datasets used, each subject was represented, on average, by just $170$ five-second ECG segments, which is arguably too few to train a convolutional neural network to robustly discriminate between $1019$ individuals. Considering this, with future efforts devoted to adequately dealing with scarce data, deep learning methodologies could see their potential for ECG biometrics be better harnessed and place themselves as clearly better alternatives to traditional pipeline algorithms.

\section{Summary and Conclusions}

This work proposed a convolutional neural network for biometric identification based on non-intrusive electrocardiogram acquisitions. The proposed method was evaluated for incremental integration of traditional ECG biometric pipeline stages, including a complete substitution by the CNN architecture, that received raw five-second ECG segments and output a decision on the corresponding identity.

Besides this study, seven data augmentation techniques for unidimensional signals were explored and their individual and collective impact on the algorithm’s performance was assessed. The results on the UofTDB database were compared with those of a baseline algorithm and two promising state-of-the-art methods. 

The results show that the total integration of traditional pipeline processes in the CNN architecture was successful. The proposed CNN with data augmentation and receiving raw five-second segments surpassed, in all settings, the baseline and state-of-the-art algorithms in direct benchmarking. Among other recent state-of-the-art methods, considering the diverse dataset characteristics, the proposed method has also shown promise as an accurate and robust biometric identification algorithm.

\chapter[Triplet Loss and Transfer Learning for Identity Verification]{Triplet Loss and Transfer Learning\\for Identity Verification}\label{ch:ecgauth}

\begin{tcolorbox}\footnotesize
{\large\bf Foreword on Author Contributions}

The research work described in this chapter was conducted entirely by the author of this thesis, under the supervision of Jaime S. Cardoso. The results of this work have been disseminated in the form of an article in international conference proceedings:
\begin{itemize}[noitemsep, leftmargin=1em, nosep]
    \item \underline{J. R. Pinto} and J. S. Cardoso, ``An End-to-End Convolutional Neural Network for ECG-Based Biometric Authentication,'' in \emph{10th IEEE International Conference on Biometrics: Theory, Applications and Systems (BTAS 2019)}, Sep.~2019.~\cite{Pinto2019b}
\end{itemize}

\end{tcolorbox}

\section{Context and Motivation}



\begin{figure*}[t]
\centering
\footnotesize
\includegraphics[width=\textwidth]{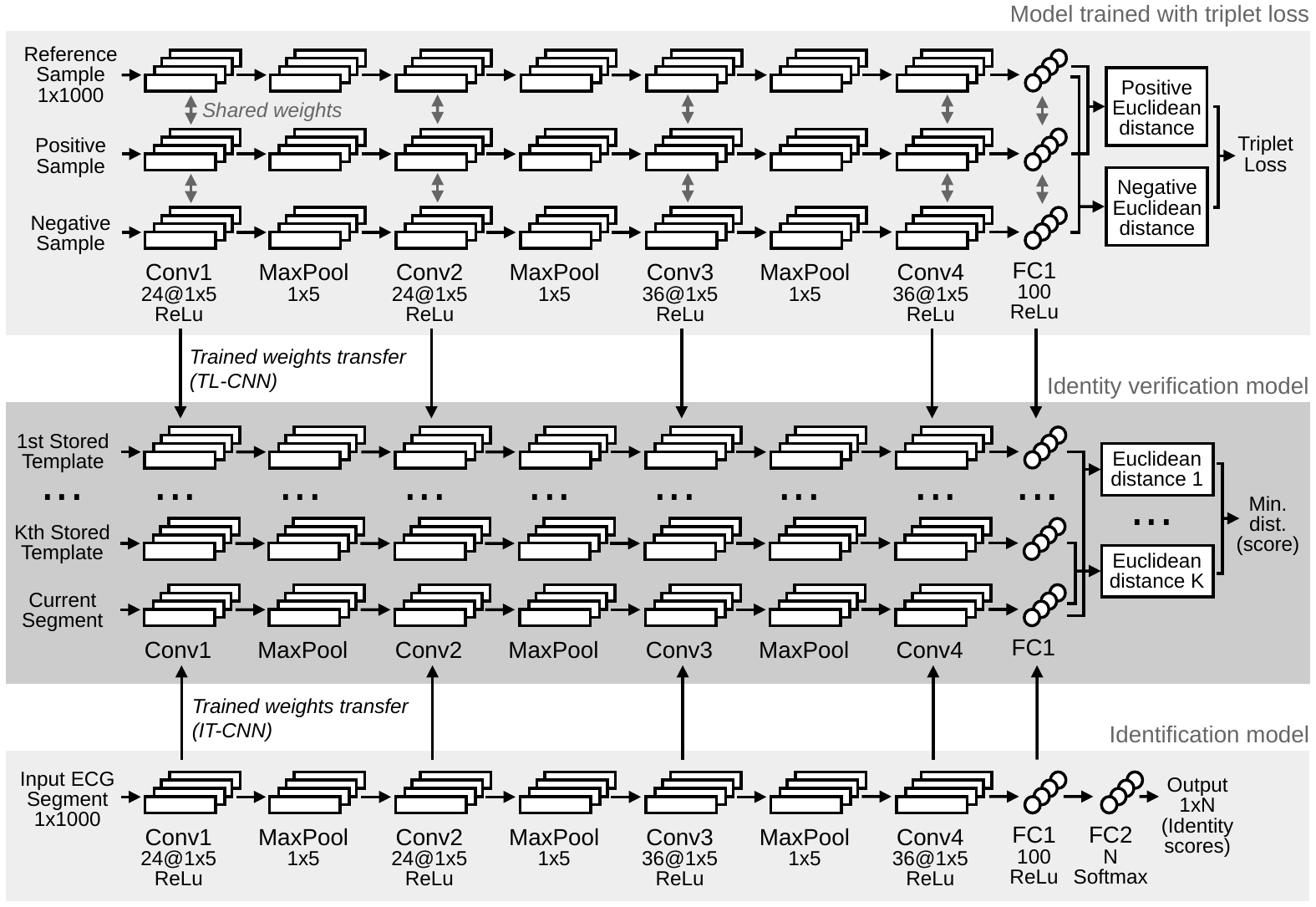}
\caption[Schemes of the proposed identity verification model, including the weight transfer between networks for both proposed training methodologies.]{Schemes of the proposed identity verification model, including the weight transfer between networks for both proposed training methodologies (the input shape $1\times1000$ refers to the five-second length of the segments used in this work, $1000$ samples at $200$ Hz sampling frequency).}
\label{fig:endtoendauth_method_schemas}
\end{figure*}



The field of ECG biometrics has been steadily evolving from on-the-person signals to off-the-person acquisition setups. Despite the enhanced usability and comfort, the increased dominance of noise and variability in off-the-person signals places serious hurdles to the real application of ECG biometric systems (more detailed information in Chapter~\ref{ch:ecgprior}).

Some researchers have resorted to deep learning in order to fight off noise and variability and achieve better performance and robustness~\cite{Eduardo2017,Luz2018,Pinto2019Deep,Zhang2017,Zhang2017b}. However, these still rely on separate predefined feature transforms and/or noise removal techniques, which are not optimised for the task at hand and therefore limit the achievable performance. In fact, the work presented in Chapter~\ref{ch:ecgiden} shows that end-to-end deep models offer considerable performance benefits in off-the-person ECG biometric identification, especially when using tailored augmentation techniques.

Building upon the work in Chapter~\ref{ch:ecgiden}, this work studied the use of end-to-end convolutional neural networks (CNN) for ECG identity verification. The main goal was to discover if dismissing all separate processes of denoising or preparation in favour of a single integrated model (granted complete control over the robustness to signal noise and variability) would also improve performance and robustness in the task of identity verification. Besides the use of metric learning through the triplet loss, this work introduces the technique of weight transfer from a similar model trained for identification. This aimed to assess whether parameters optimised for identification tasks would offer performance benefits in identity verification.

The proposed network and both training methodologies were extensively evaluated on three ECG collections, which include on-the-person and off-the-person signals with varying signal quality, multi-session recordings from several subjects, and the influence of emotions, posture, and exercise. This evaluation included the assessment of the trained model's applicability to other signal collections, through cross-database tests using transfer learning and fine-tuning.

\section{Methodology}

\subsection{Model architecture}

The proposed method for ECG biometric identity verification is based on a CNN (see Fig.~\ref{fig:endtoendauth_method_schemas}, darker grey). All enrolled users have one or more fixed-length ECG segments (templates) stored in the system, that have been blindly segmented (without requiring any process of reference point detection) from a recording obtained upon enrollment.

When a user claims to be an enrolled individual, the model receives and processes, simultaneously, the $K$ stored templates of the claimed identity and $1$ current segment of the user. The comparison between the processed current segment and each of the $K$ stored templates allows the model to output a dissimilarity score, which can be used to accept or reject the identity claim. 

After sample-wise normalisation to zero mean and unit variance, the processing of each input segment or template starts with a succession of convolutional and pooling layers. As visible in Fig.~\ref{fig:endtoendauth_method_schemas}, four unidimensional convolutional layers are alternated with three max-pooling layers. All have $1\times5$ filters, and the convolution is performed with unit stride and no padding. The first two convolutional layers hold $24$ feature maps, while the last two hold $36$.

The second part of the network is composed of a fully-connected layer. The outputs of this fully-connected layer for each stored template ($\textbf{a}$) and for the current segment ($\textbf{b}$) are compared using normalised Euclidean distance \cite{Wolfram} (see Eq.~\eqref{eq:ned}), using their variance ($\mathit{Var}$) so the output lies in $[0,1]$. Among the $K$ distances computed, the minimum is output as the final dissimilarity score for identity verification.
\begin{equation}\label{eq:ned}
    d(\textbf{a},\textbf{b}) =  \frac{\mathit{Var}(\textbf{a}-\textbf{b})}{2\left(\mathit{Var}(\textbf{a})+\mathit{Var}(\textbf{b})\right)}.
\end{equation}

\subsection{Model training}

The weights for the identity verification model layers are transferred either from a model trained for identification or from a model trained using triplet loss (see Fig.~\ref{fig:endtoendauth_method_schemas}). The training methodology of transferring weights from an identification model aimed to take advantage of the training process of identification deep neural networks and assess how it could benefit a neural network for identity verification. On the other hand, triplet loss has been recently and successfully used in biometrics, for identity verification and other similar tasks~\cite{Chen2017,Cheng2016,Ding2015}.

The training process requires specific structural changes to the model, which are illustrated in Fig.~\ref{fig:endtoendauth_method_schemas} and described below. In all cases, during training, the optimiser used was Adam \cite{Kingma2015} with an initial learning rate of $0.001$, $\beta_1 = 0.9$, $\beta_2 = 0.999$, and no decay. Dropout \cite{Srivastava2014} and data augmentation (random permutations, as in \cite{Pinto2019Deep}) were used to prevent overfitting. After training, the weights are transferred to the respective layers on the identity verification model.

\subsubsection{Transfer from identification network (IT-CNN)}

In the case of identification training (IT-CNN), the model is structured to receive $1$ input segment and contain one additional fully-connected layer (FC2), using softmax activation, that will output $N$ scores. It is trained for identification with data from $N$ identities (following the work of Pinto~\etal~\cite{Pinto2019Deep}).

After receiving a training segment, considering its true label and the network's output, the sparse categorical cross-entropy loss \cite{Tensorflow2015,Keras2015} is computed and used during training to ultimately prepare the model to adequately discriminate the subjects.

\subsubsection{Triplet loss training (TL-CNN)}

To be trained using triplet loss (TL-CNN), the identity verification model, which has $K+1$ inputs and $1$ output, is restructured to receive $3$ inputs and offer $2$ outputs. The three inputs are the reference template, a positive template (whose identity is the same as the reference), and a negative template (of a different identity). The network processes each input and computes the dissimilarities between the reference and the positive template ($p$) and between the reference and the negative template ($n$).

Using adequate triplets of signal segments, the goal is to minimize $p$ and maximize $n$. Hence, the model is trained using triplet loss \cite{Chen2017}, which can be computed for each triplet of inputs through the function:
\begin{equation}
    l(p,n) = \max(0, \alpha + p - n),
\end{equation}
where $\alpha$ controls the margin to be enforced between the scores of positive and negative pairs (in this work, $\alpha=0.5$). This margin eases the choice of an effective threshold for the purpose of identity verification.

\section{Experimental Setup}

In this work, one of the main concerns was ensuring the performance results were as realistic as possible. To achieve this, all databases were split between training subjects and testing subjects, to ensure the model can be trained and applied to data from two entirely disjoint sets of subjects. Furthermore, cross-database tests were performed to ensure the model can generalise to other population samples and acquisition settings. Subject enrollment was limited to realistic durations (5, 10, 15, or, at most, 30 seconds of the earliest data from each subject).

\subsection{Data and reference methods}

The three selected databases were UofTDB~\cite{Wahabi2014}, CYBHi~\cite{Silva2014}, and PTB~\cite{Bousseljot1995, Goldberger2000}. UofTDB (off-the-person, $1019$ subjects) was used for most experiments due to its intermediate but realistic signal quality. The PTB (on-the-person, $290$ subjects) and CYBHi (off-the-person, $128$ subjects) databases were used to assess performance in better and worse signal quality settings, respectively. To match UofTDB, CYBHi and PTB signals were resampled to $200$ Hz. For PTB, only Lead I signals were used.

Three literature methods were used as reference: the AC/LDA method, proposed by Agrafioti~\etal~\cite{Agrafioti2012}; the Autoencoder method, proposed by Eduardo~\etal~\cite{Eduardo2017}; and the DCT method, proposed by Pinto~\etal~\cite{Pinto2019Deep,Pinto2017} (adapted for identity verification, using cosine distance normalised to $[0,1]$ for matching).

\subsection{Evaluation scenarios}

The proposed and implemented methods were evaluated across four scenarios, as detailed below, using the Equal Error Rate (EER, see~\cite{Pinto2018} for more details). Here, each signal segment used as input for the proposed model was five seconds long ($1000$ samples at $200$ Hz sampling frequency).

In the \emph{single-database scenario}, the proposed model was evaluated on UofTDB data, and compared with the aforementioned reference state-of-the-art methods. The last $100$ subjects were reserved for training, while the data from the remaining $919$ subjects were used for testing. The number of enrollment templates varied between $1$, $2$, $3$, or $6$ five-second segments.

The \emph{varying identity set size scenario} aimed to study how the performance is affected by the number of subjects used to train the model. Instead of the original $100$ subjects, training was performed using the $20$, $50$, or $150$ last subjects of UofTDB, and the remaining $999$, $969$, or $869$ subjects, respectively, were used for testing.

The \emph{cross-database scenario} was designed to assess the proposed model's applicability to signals from other databases. The proposed model, previously trained on $100$ subjects from UofTDB, was directly tested on data from CYBHi and PTB, without fine-tuning.

At last, in the \emph{fine-tuning scenario}, the goal was to assess the performance benefits brought by fine-tuning. As in the cross-database scenario, the proposed model trained on UofTDB data (from $100$ subjects) was fine-tuned to CYBHi/PTB data (from $20$ subjects). This was compared to the model directly trained, from scratch, on data from CYBHi or PTB (from $20$ subjects, following the single-database scenario). With $20$ subjects reserved for training, the tests on this scenario were performed for $108$ (CYBHi) or $270$ (PTB) subjects.

\section{Results and Discussion}

\subsection{Single-database scenario}

The results obtained in the single-database scenario are presented in Table~\ref{tab:results_maintest}. In all cases, the IT-CNN model, which used weights trained for identification, attained better results than TL-CNN, which was trained using triplet loss. With $30$ seconds of user enrollment, IT-CNN achieved $7.86$\% EER, while TL-CNN offered $9.94$\% EER in the same circumstances.

When considering shorter enrollment recordings ($5$ s, $10$ s, and $15$ s), the performance of both proposed methods worsens, but always remained below $14$\% EER. It is noteworthy that IT-CNN presented a wider advantage over TL-CNN with more enrollment data, which may denote it takes better advantage of the availability of data.

\begin{table}[t!]
\centering
\caption[Single-database scenario: EER results (\%) when trained with data from $100$ UofTDB subjects and tested with $919$ UofTDB subjects.]{Single-database scenario: EER results (\%) when trained with data from $100$ UofTDB subjects and tested with $919$ UofTDB subjects (in italics: proposed methods; in bold: best results).}
\label{tab:results_maintest}
\begin{tabular}{lcccc}\hline
 & \multicolumn{4}{c}{\textbf{Enrolment duration}} \\
\textbf{Method} & \textit{\textbf{5 s}} & \textit{\textbf{10 s}} & \textit{\textbf{15 s}} & \textit{\textbf{30 s}} \\ \hline
\emph{IT-CNN} & \textbf{13.70} & \textbf{10.92} & \textbf{9.52} & \textbf{7.86} \\
\emph{TL-CNN} & 13.93 & 11.89 & 10.90 & 9.94 \\
AC/LDA \cite{Agrafioti2012} & 30.27 & 17.90 & 16.55 & 15.82 \\
Autoencoder \cite{Eduardo2017} & 21.82 & 19.68 & 18.84 & 17.09 \\
DCT \cite{Pinto2019Deep,Pinto2017} & 23.05 & 20.41 & 18.55 & 17.38 \\\hline
\end{tabular}
\end{table}

Among the reference methods, AC/LDA presented the best results in most settings. When compared with these results, both proposed methods offered consistently lower EER. Considering the best reference method for each enrollment duration, IT-CNN attained an EER reduction of around $7-8$\%, which can be regarded as a significant improvement over the state-of-the-art.

\begin{figure}[!t]
    \centering
    \includegraphics[width=0.7\linewidth]{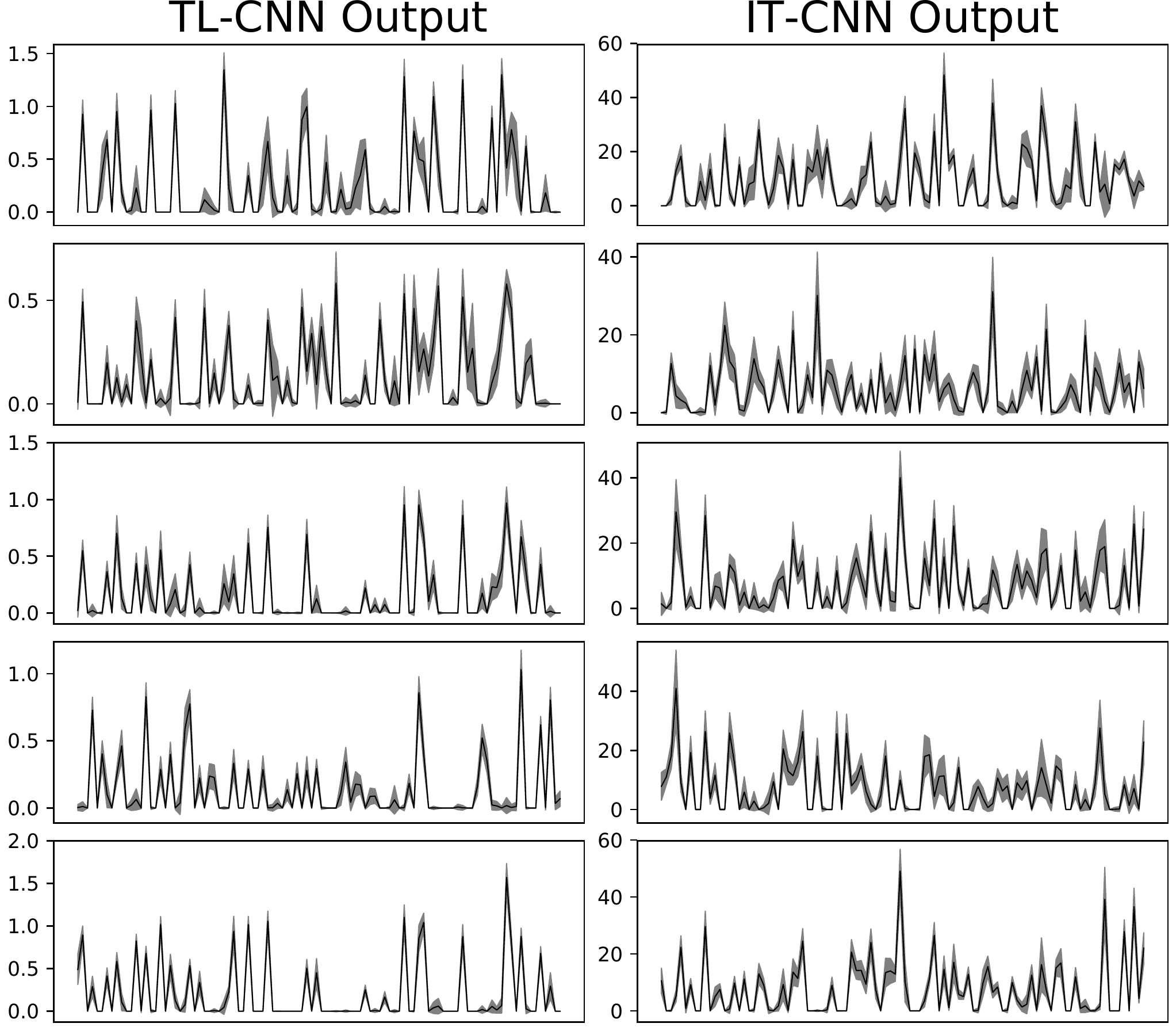}
    \caption[Network outputs for all training samples of five example subjects from the UofTDB collection.]{Network outputs for all training samples of five example subjects from the UofTDB collection (one subject for each row; the average output feature vector is presented as a black line, and the standard deviation as a grey area).}
    \label{fig:endtoendauth_output_visualisation}
\end{figure}

Among other state-of-the-art works, Luz~\etal~\cite{Luz2018}, under similar settings, reported $14.27$\% EER with UofTDB data. All IT-CNN and TL-CNN performance results are better, even when considering only $5$ seconds of enrollment (much less than what was used by Luz~\etal).

Moreover, Louis~\etal~\cite{Louis2016} reported $7.89$\% EER, but only using single session data from $1012$ UofTDB subjects. Using only data from subjects with more than one session ($82$ subjects), Louis~\etal~reported $10.10$\% EER, while Komeili~\etal~\cite{Komeili2017} reported $6.9$\% EER. Although the evaluation settings are different, the proposed method's results are aligned with these ($7.86$\% for IT-CNN with $30$ s enrollment).

\begin{table}[!t]
\centering
\caption[Single-database scenario: Mean and standard deviation of the EER results (\%) obtained on $100$ random data divisions.]{Single-database scenario: Mean and standard deviation of the EER results (\%) obtained on $100$ random data divisions (in italics: proposed methods; in bold: best results).}
\label{tab:results_statsig}
\begin{tabular}{lcccc}\hline
 & \multicolumn{4}{c}{\textbf{Enrolment duration}} \\
\textbf{Method} & \textit{\textbf{5 s}} & \textit{\textbf{10 s}} & \textit{\textbf{15 s}} & \textit{\textbf{30 s}} \\ \hline
\emph{IT-CNN}   & \textbf{11.3 $\pm$ 0.14} &  \textbf{9.4 $\pm$ 0.12} &  \textbf{8.4 $\pm$ 0.14} &  \textbf{7.0 $\pm$ 0.14} \\
\emph{TL-CNN}   & 11.6 $\pm$ 0.16 & 10.3 $\pm$ 0.11 &  9.7 $\pm$ 0.14 &  8.7 $\pm$ 0.11 \\
AC/LDA \cite{Agrafioti2012}  & 17.7 $\pm$ 0.18 & 15.6 $\pm$ 0.17 & 14.6 $\pm$ 0.17 & 13.3 $\pm$ 0.31 \\
Autoencoder \cite{Eduardo2017}  & 18.4 $\pm$ 0.17 & 16.3 $\pm$ 0.14 & 15.9 $\pm$ 0.16 & 13.8 $\pm$ 0.12 \\
DCT \cite{Pinto2019Deep,Pinto2017}  & 21.2 $\pm$ 0.16 & 18.6 $\pm$ 0.15 & 16.4 $\pm$ 0.14 & 15.5 $\pm$ 0.21 \\\hline
\end{tabular}
\end{table}

The statistical significance of the results was assessed, repeating the evaluation on one-hundred random subject data divisions between enrollment and testing (Table~\ref{tab:results_statsig}). Overall, the results were better, as this test is arguably less realistic than the remaining tests performed in this study (a real biometric system will always use the very first data of a subject for enrollment). Applying a paired two-sided \emph{t}-test to the EER estimates, the results of the proposed methods IT-CNN and TL-CNN were significantly different in all cases (the differences are statistically significant at the $1\%$ level), not only from each of the implemented state-of-the-art methods but also between themselves.

Additionally, the outputs of the network for five-second training segments from different subjects were visualised (see Fig.~\ref{fig:endtoendauth_output_visualisation}). These are, effectively, the feature vectors used for the identity verification decision. It is possible to observe that, despite the blind segmentation and the noise and variability carried by each five-second segment, the trained network was able to represent each input segment in a way that maximises similarity with other segments from the same subject. Although some variability is still present, it is reduced to a manageable level for the biometric identity verification task, and the differences between the subjects' output patterns are noticeable even through a simple visualisation of the plots.

\subsection{Varying identity set size scenario}

\begin{figure}[p]
    \centering
    \includegraphics[width=0.60\linewidth]{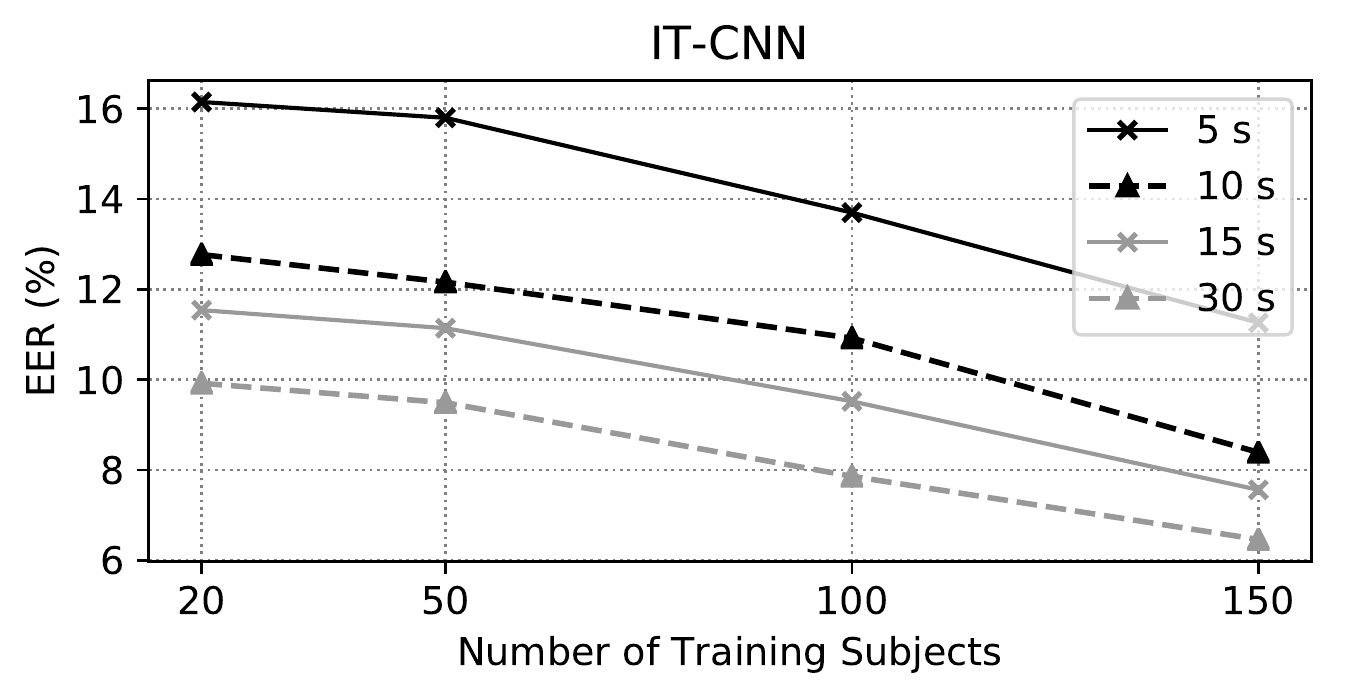}\\
    \includegraphics[width=0.60\linewidth]{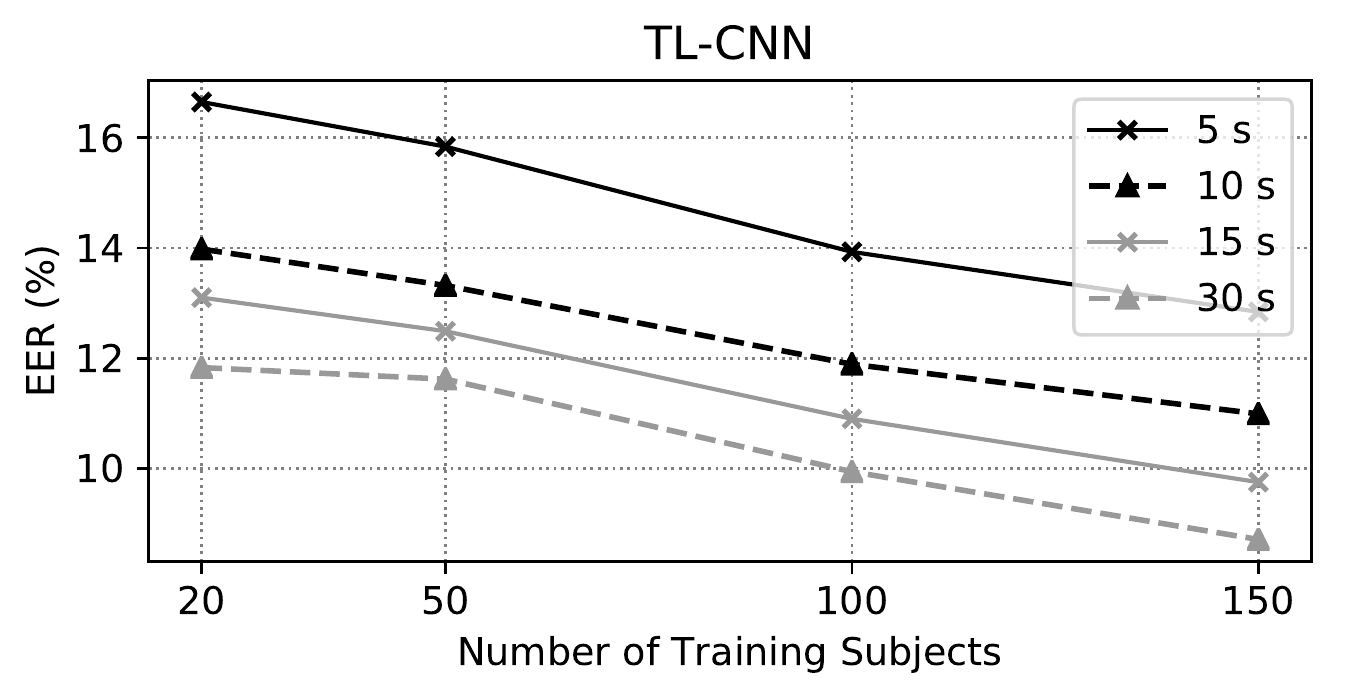}
    \caption{Varying identity set size scenario: EER evolution with number of subjects reserved for training, for diverse enrollment durations, for the proposed methods IT-CNN and TL-CNN.}
    \label{fig:endtoendauth_uoftdb_training_subjects_results}
\end{figure}

\begin{figure}[p]
    \centering
    \includegraphics[width=0.60\linewidth]{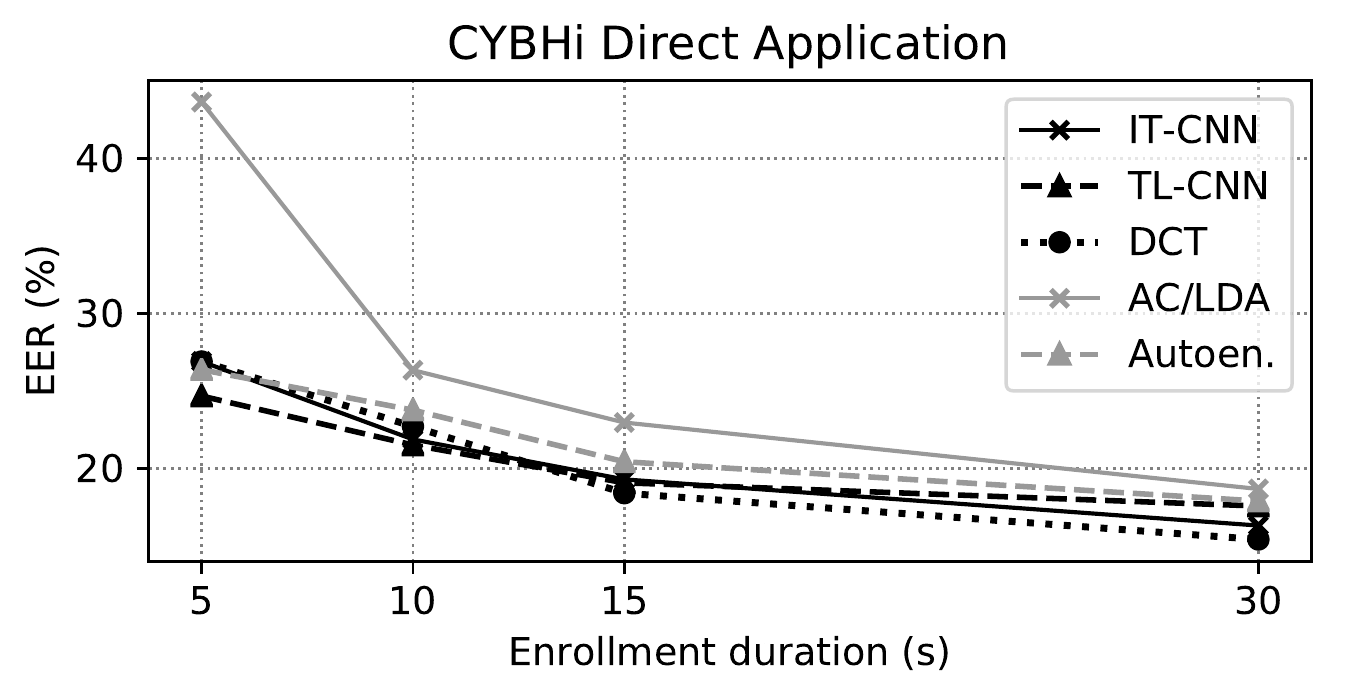}\\
    \includegraphics[width=0.60\linewidth]{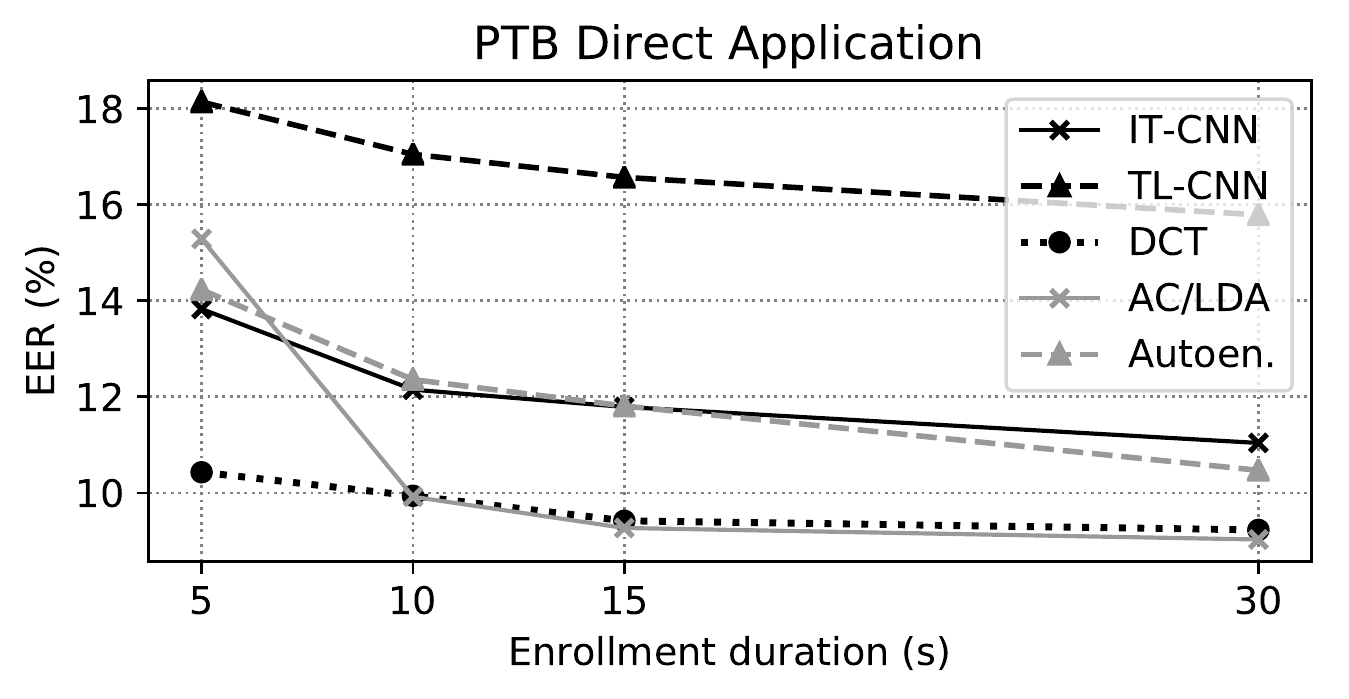}
    \caption{Cross-database scenario: EER for the proposed methods IT-CNN and TL-CNN when trained with UofTDB data and directly applied to CYBHi or PTB, and comparison with state-of-the-art methods.}
    \label{fig:endtoendauth_uoftdb_direct_application_results}
\end{figure}

\begin{figure}[p]
    \centering
    \includegraphics[width=0.60\linewidth]{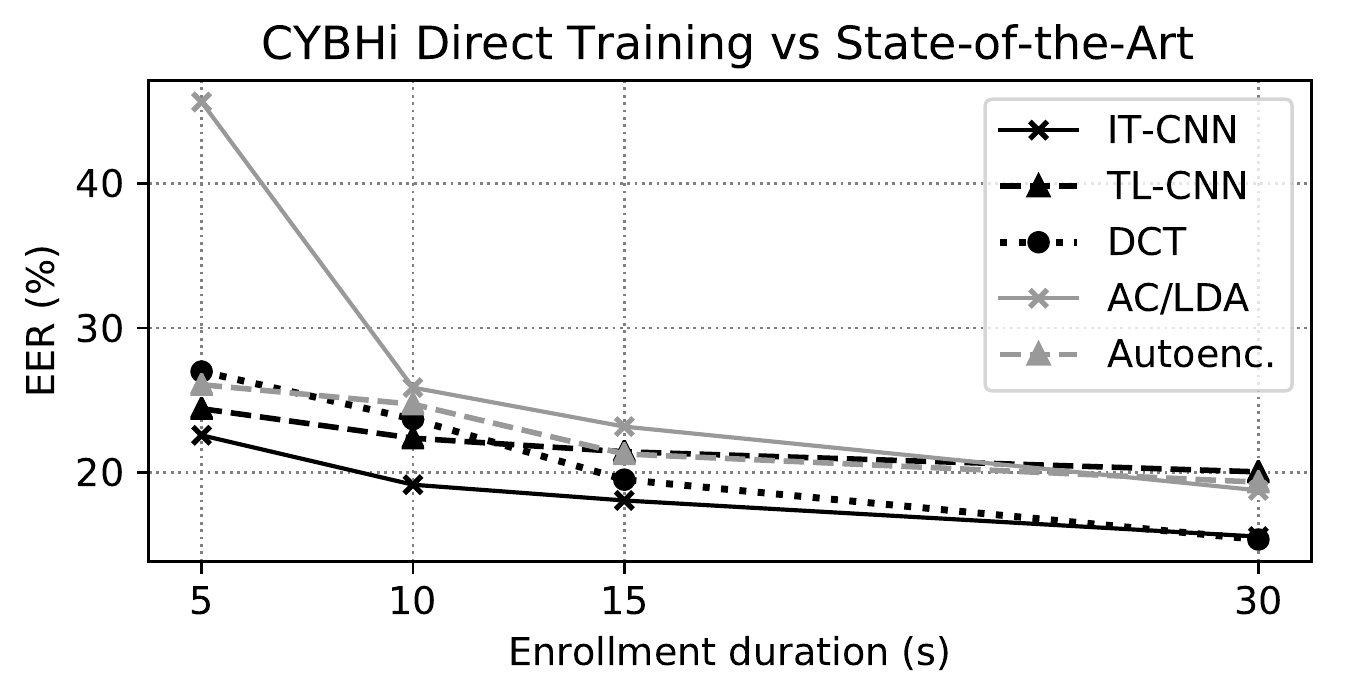}\\
    \includegraphics[width=0.60\linewidth]{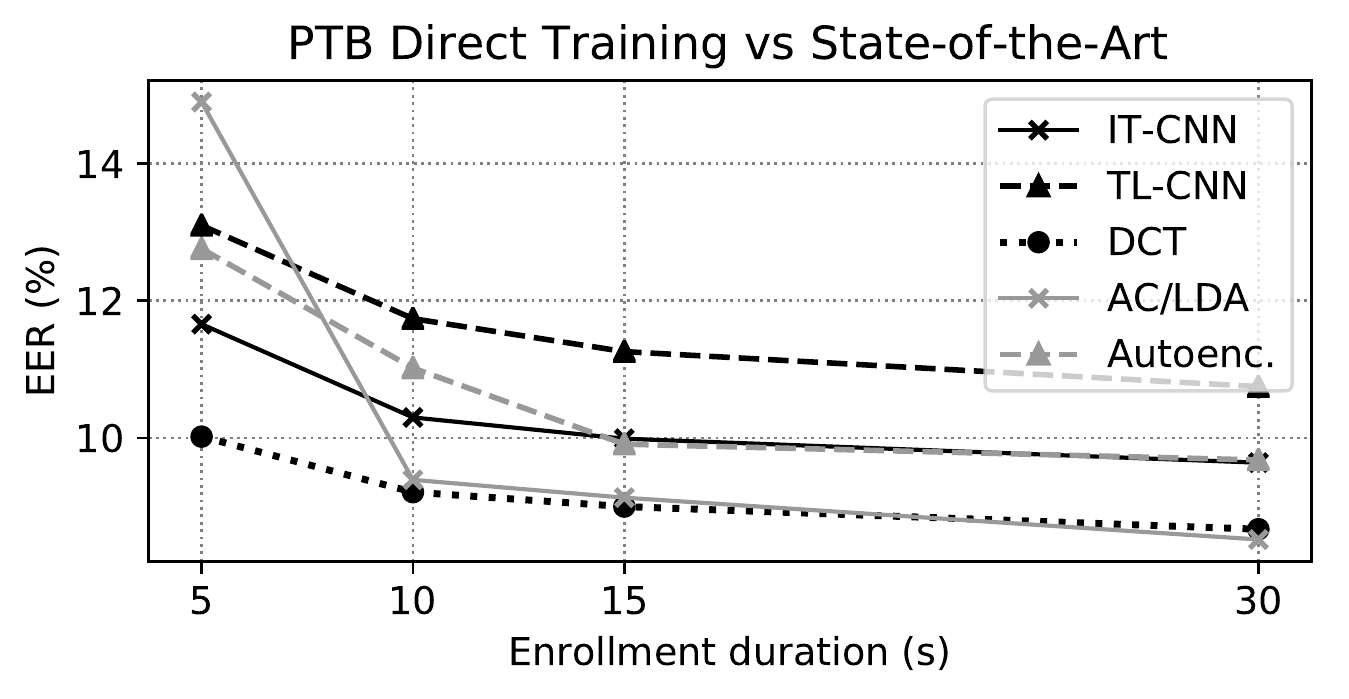}
    \caption{Fine-tuning scenario: EER results for the proposed methods IT-CNN and TL-CNN when directly trained with CYBHi or PTB data from $20$ subjects, and comparison with state-of-the-art methods.}
    \label{fig:endtoendauth_uoftdb_direct_training_results}
\end{figure}

\begin{figure}[p]
    \centering
    \includegraphics[width=0.60\linewidth]{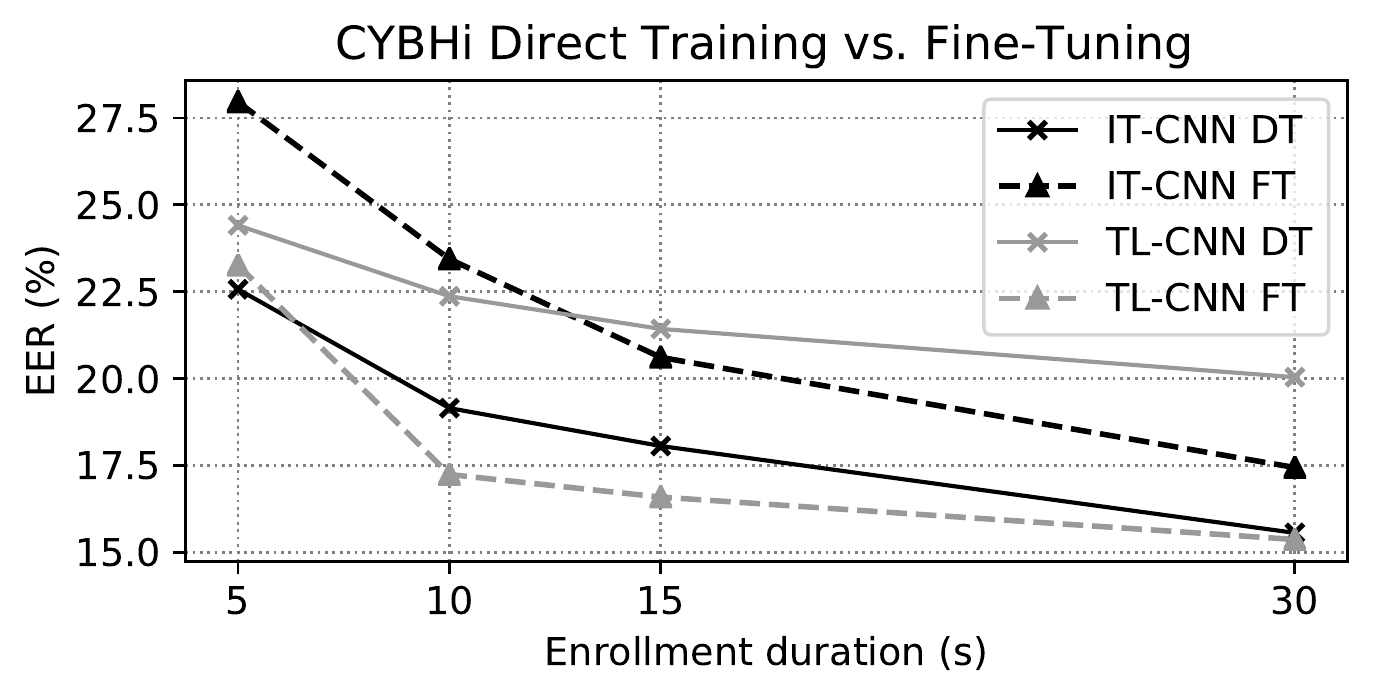}\\
    \includegraphics[width=0.60\linewidth]{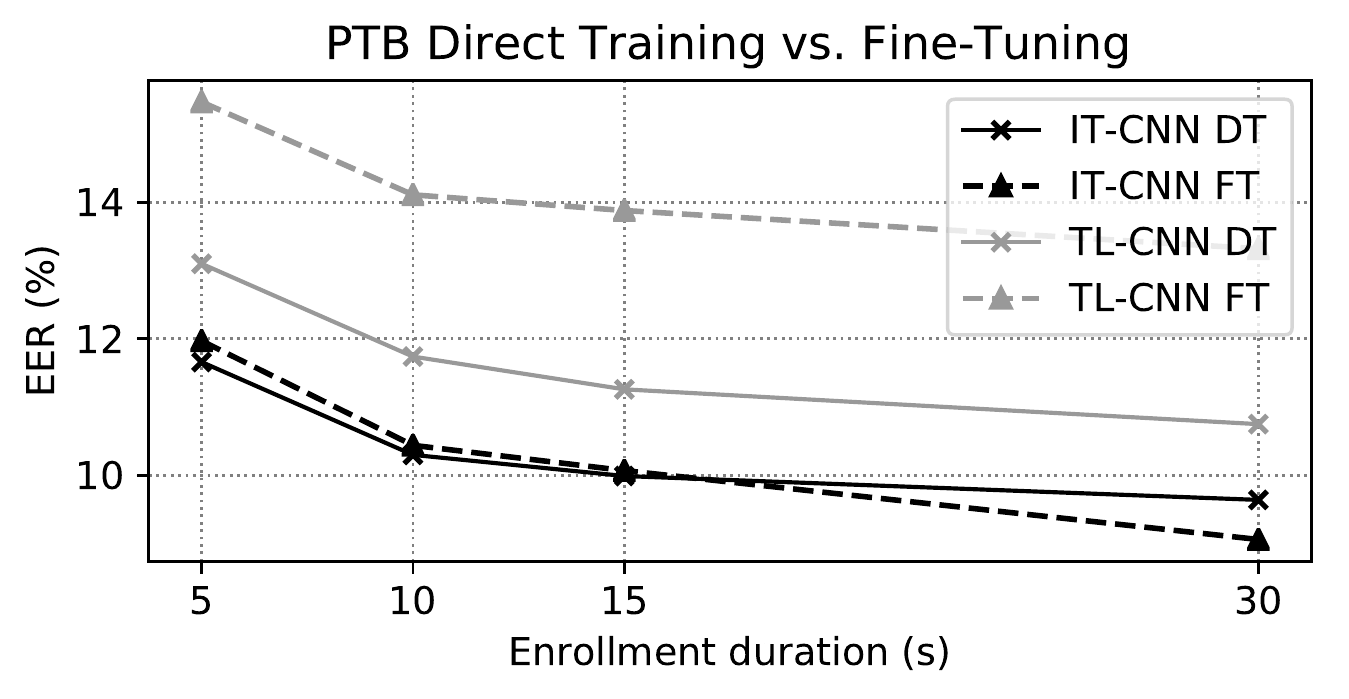}
    \caption{Fine-tuning scenario: EER results for the proposed methods when (DT) trained, from scratch, with data from CYBHi or PTB, or when (FT) trained with UofTDB data and fine-tuned to CYBHi/PTB.}
    \label{fig:endtoendauth_uoftdb_dt_vs_ft}
\end{figure}

In the varying identity set size scenario, multiple numbers of UofTDB subjects reserved for training were explored (Fig.~\ref{fig:endtoendauth_uoftdb_training_subjects_results}). In all cases, an increase in the number of training subjects resulted in performance improvements. The best results were obtained with $150$ training subjects and $30$ seconds enrollment, with $6.46$\% EER and $8.71$\% for IT-CNN and TL-CNN, respectively. Nevertheless, even with just $20$ training subjects, IT-CNN offered performance under $10$\% EER ($9.92$\%, with $30$ s enrollment).

As in the single-database scenario, it was noticeable that the performance advantage of IT-CNN over TL-CNN was greater when more data was available, either for model training or user enrollment. For example, the EER difference between IT-CNN and TL-CNN grew from $0.5$\% to $2.25$\% when increasing the number of training subjects from $20$ to $150$ and the enrollment duration from $5$ to $30$ s.

Despite this, one could expect the IT-CNN method to perform better than the state-of-the-art, even under scarce data conditions. Based on these results, when pretrained with only $20$ subjects with $10$ s enrollments, IT-CNN should offer an EER lower than $13$\% on a population of nearly one thousand individuals.

\subsection{Cross-database scenario}

In the cross-database scenario, the proposed methodologies were directly applied to CYBHi and PTB data, after training on data from $100$ UofTDB subjects (Fig.~\ref{fig:endtoendauth_uoftdb_direct_application_results}).

With CYBHi, IT-CNN offered better performance than TL-CNN when using $30$ s enrollment ($16.30$\% against $17.56$\% EER). However, with reduced enrollment duration ($5$ s), TL-CNN performed better ($24.66$\% against $26.89$\% EER). This reinforces the idea that TL-CNN is better in scarce data situations, while IT-CNN takes better advantage of the greater availability of data. With PTB, IT-CNN was, in all cases, the most successful proposed method ($13.83$\% EER with $5$ s enrollment).

Among the state-of-the-art methods, AC/LDA behaved as in the single-database scenario (see Table~\ref{tab:results_maintest}), offering the worst results when using $5$ s enrollment, but sharply improving with more enrollment data, offering the best result with PTB ($9.03$\% EER). DCT presented the best result with CYBHi ($15.40$\% EER), while IT-CNN offered the second-best result ($16.30$\% EER). Both proposed methods were, in general, worse than the state-of-the-art with the PTB database.

\subsection{Fine-tuning scenario}

In the fine-tuning scenario, the model was trained with CYBHi/PTB data and compared with the state-of-the-art (Fig.~\ref{fig:endtoendauth_uoftdb_direct_training_results}) and when trained with UofTDB data and fine-tuned to CYBHi/PTB (Fig.~\ref{fig:endtoendauth_uoftdb_dt_vs_ft}).

Directly trained on CYBHi data, TL-CNN attained $20.04$\% EER, but it offered $17.56$\% EER if trained with UofTDB data, and further improving to $15.37$\% EER if fine-tuning is performed. TL-CNN was able to attain better performance than IT-CNN in more difficult settings, once again indicating that this method may be better fitted for scarcer data or noisier signals.

On PTB, TL-CNN did not offer competitive results. For IT-CNN, fine-tuning ($9.06$\% EER with $30$ s enrollment) improved the results over the direct application, but it was not enough to significantly improve the results of direct training. Apparently, training with UofTDB data over-prepared the network for a degree of noise and variability that is not verified on PTB signals, which ultimately harmed its performance. A hybrid method where, before regular training, the neural network would be encouraged to mimic the behaviour of traditional methods, could be beneficial in cross-database settings.

Overall, the proposed methodologies presented more competitive results on CYBHi than on PTB, likely due to PTB signals' lesser noise and variability. Thus, while the proposed model has shown robustness to noise and variability in off-the-person settings, the state-of-the-art methods are more fitted to cleaner on-the-person signals.

\section{Summary and Conclusions}

In this work, an end-to-end model, based on a CNN, was proposed for biometric identity verification using ECG signals. It was designed to use a set of stored templates of a claimed identity and an ECG segment of the current user, and output a dissimilarity score used to accept or reject the identity claim. The model was trained using triplet loss or by transferring weights from a similar model trained for identification.

The proposed model was successful in improving the performance of state-of-the-art methods, especially in off-the-person signals, increasingly used in ECG-based biometrics. Using identification training has offered better performance than triplet loss when more training and enrollment data are available and could bring benefits for other tasks or biometric traits. Both methods have shown the ability to overcome increased noise and variability of off-the-person signals, focusing on subject-specific signal patterns for accurate identity verification. Nevertheless, further efforts should be devoted to improving performance and turning the ECG into a reliable biometric trait.

\chapter[Long-Term Performance and Template Update]{Long-Term Performance\\and Template Update}\label{ch:ecglongterm}

\begin{tcolorbox}\footnotesize
{\large\bf Foreword on Author Contributions}

The research work described in this chapter was conducted in collaboration with Gabriel C. Lopes, under the supervision of Jaime S. Cardoso. The author of this thesis contributed to this work on the formulation and implementation of the biometric recognition models, the conceptualisation of the template update methodologies, the preparation and conduction of experiments, the discussion of the results, and the writing of the scientific publications.

The results of this work have been disseminated as an article in international conference proceedings and an abstract in national conference proceedings:
\begin{itemize}[noitemsep, leftmargin=1em, nosep]
    \item G. Lopes, \underline{J. R. Pinto}, and J. S. Cardoso, ``Don't You Forget About Me: A Study on Long-Term Performance in ECG Biometrics,'' in \emph{IbPRIA 2019: 9th Iberian Conference on Pattern Recognition and Image Analysis}, Jul.~2019.~\cite{Lopes2019}
    \item G. Lopes, \underline{J. R. Pinto}, J. S. Cardoso, and A. Rebelo, ``Long-Term Performance of a Convolutional Neural Network for ECG-Based Biometrics,'' in \emph{25th Portuguese Conference on Pattern Recognition (RECPAD 2019)}, Oct.~2019.
\end{itemize}

\end{tcolorbox}

\section{Context and Motivation}

Modern ECG biometric techniques generally report relatively high identification rates and low verification error, while current off-the-person ECG acquisition techniques contribute towards increased simplicity, usability, and comfort~\cite{Pinto2017, Pinto2018}. Nevertheless, as with most alternatives, the performance decays over time, especially when considering long-term usage~\cite{komeili_liveness_2018}.

The natural variability of the input biometric data, the effects of ageing, and variations caused by the subject's interaction with the sensor contribute to intrasubject variability~\cite{Jain2011}. This causes stored individual templates to quickly lose representativity, resulting in poor recognition performance and placing serious challenges on long-term recognition. Thus, long-term biometrics benefits from the frequent update of stored templates to keep up with the variability and ageing of the users, thus maintaining acceptable performance over time~\cite{freni_template_2010}.

Specifically for ECG biometrics, long-term performance and template update remain open challenges. In the most thorough work yet on this topic, Labati~\etal~\cite{Labati2013} have studied the performance decay over time on their proposed algorithm for ECG-based authentication, finding that performance decays significantly even over relatively short periods of time. However, the focus of this study was limited to authentication and the algorithm proposed by the authors.

In this work, we build upon the study of Labati~\etal~\cite{Labati2013}. We
aim to more thoroughly explore the problem of long-term performance decay in ECG biometrics and how to correctly address it. Specifically, we extend it to (a) focus on the task of ECG biometric identification, (b) study diverse state-of-the-art biometric methods, and (c) evaluate how different update techniques may be able to improve long-term performance.

\section{Related Work}

In the literature, it is difficult to find a strong and widely accepted rule for template update. Most methods are based on heuristics and empirically determined thresholds, which are highly dependent on the data and the application scenario. For example, \citet{komeili_liveness_2018}, for authentication, have set the acceptance threshold equal to the point of zero false acceptance rate, thus ensuring updates with only genuine samples.

\begin{figure}[!t]
\centering
\includegraphics[width=0.85\linewidth]{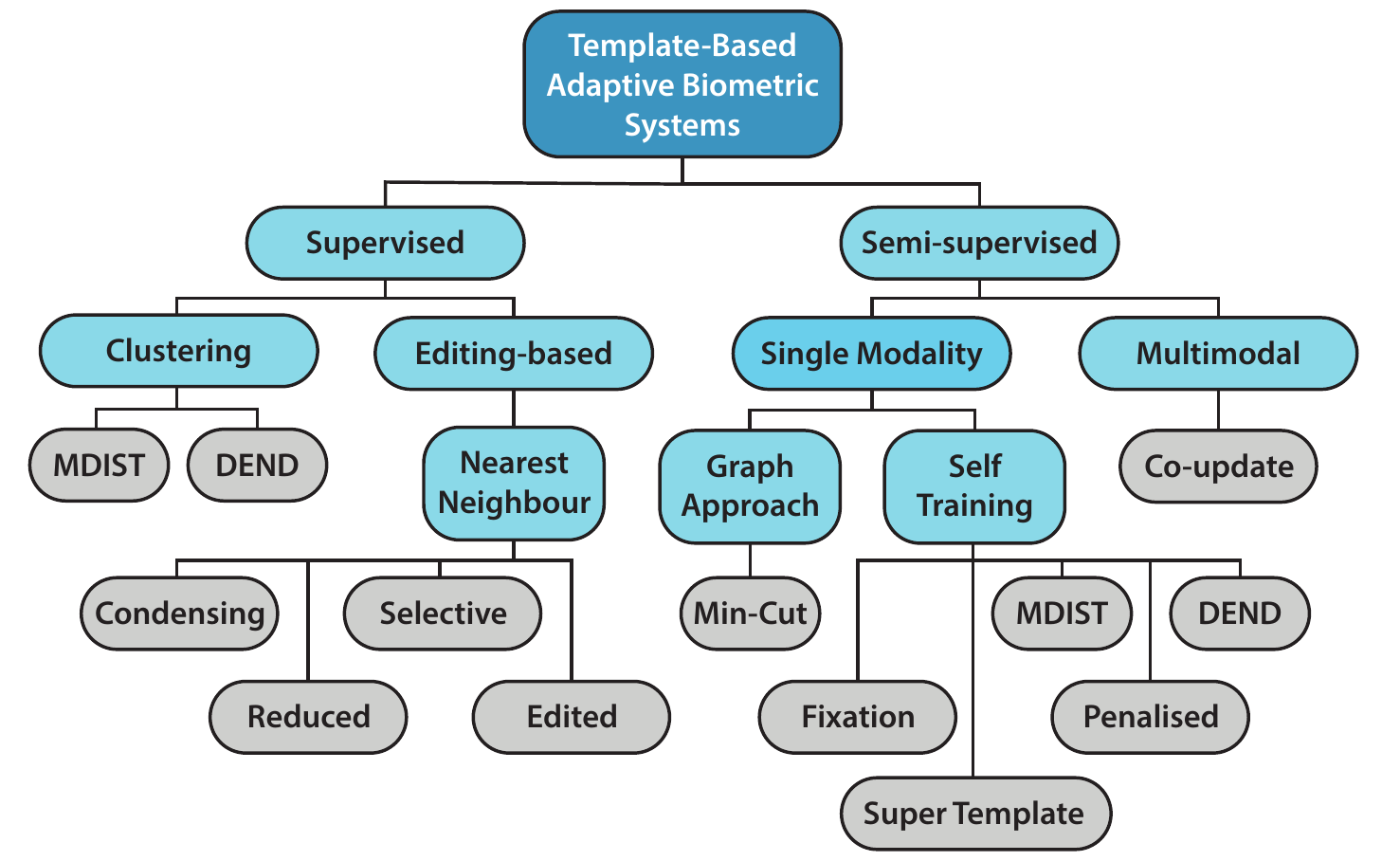}
\caption[Dendrogram representing the taxonomy of template update techniques.]{Dendrogram representing the taxonomy of template update techniques (based on~\cite{rattani_adaptive_2010}).}
\label{fig:longTerm_dendrograma_template}
\end{figure}

Nevertheless, it is possible to identify some common mechanisms that may vary depending on different factors: these include the choice of the update criterion (based on thresholds or graphs), the update periodicity (online or offline), the selection mechanism, and the template update working mode system (supervised or semi-supervised). The taxonomy of template update (see Fig.~\ref{fig:longTerm_dendrograma_template}) divides the existing techniques into two categories: supervised and semi-supervised. 

Supervised methods are offline methods in which label attribution is given by a supervisor. These contain the \emph{Clustering} subcategory, which includes the \emph{MDIST} that aims to search for the templates that minimise the distance among all the samples in the database (\emph{i.e.}, the most similar) and \emph{DEND} that aims to search for the templates that exhibit large intraclass variations resort to the dendrogram (\emph{i.e.}, the most different)~\cite{lumini_clustering_2006}.

The second subcategory comprises Editing-based methods, which are independent of the number of templates and give focus on the whole collected training set $T$. A subset $E \in T$ is generated, maintaining the classification performance offered by $T$. The best subsets were obtained by reviewing the structure of the data (which needs to be done for each subject)~\cite{freni_template_2010,hutchison_structural_2008}.  All the algorithms (based on k-Nearest Neighbours) must be representative of $T$ and can be roughly described as~\textit{incremental} when the $E$ starts empty and grows, or~\textit{decremental} when $E$ starts equal to $T$ and in each iteration some instances are deleted until some criterion is reached~\cite{freni_template_2010}.

Semi-supervised methods merge labelled 
(in biometrics, these correspond to the initial training samples) and unlabelled (corresponding to the samples available during system operation) data to improve the system's performance. This category comprises the \emph{Single Modality} (for unimodal biometric systems) and \emph{Multiple Modality} (for systems using more than one biometric trait) subcategories. The \emph{Single Modality} subcategory includes the~\emph{Self-Training} approaches such as \emph{FIFO} (first-in-first-out), \emph{Fixation}, \emph{Super Template} ($X$ composed by $N$ templates $x$) where new genuine date is always fused to a common single template~\cite{Labati2014} updated online during the execution of continuous verification,~\emph{Penalised template update} method based on the mean of the past ECG's and the actual ECG~\cite{Chun2016} and~\emph{clock method} where the current template is tested against all the others stored in the database~\cite{scheidat_automatic_2007}.

Generally, a new unknown trait measurement is used for template update if its score (returned by the biometric recognition system) is above a set threshold. Hence, the future performance of the system relies heavily on the chosen threshold value~\cite{rattani_adaptive_2010}.

The update threshold is commonly estimated using enrollment templates or training data. When training data are scarce or when using short enrollments, this may lead to some problems. First, important intrasubject variability information may be missed since only the patterns similar to the stored templates are used (and all others are discarded). Second, the effectiveness of the online methods depends on the order of the input data. Third, the methods are vulnerable to large intraclass variations. At last, since the algorithms normally look for the minimal cost (high scores), they may get stuck in local maxima and always only use high-confidence data for updating.

Semi-supervised methods also include \emph{Graph} approaches. These commonly define a graph where the nodes are either labelled (the identity is known) or unlabelled (unknown identity) data, and the edges (which can have different weights) are the similarity between those samples~\cite{rattani_adaptive_2010,zhu_semi-supervised_2008}. To be considered a graph-based semi-supervised method, it must estimate a function $f$, approximate the known $Y$ on the labelled nodes, and include two terms to smooth the graph: a loss function and a regulariser. These two terms are what define each approach (as can be seen in Table~\ref{tab:longtermecg_3})~\cite{zhu_semi-supervised_2008}, among which the most common in biometrics is min-cut graphs~\cite{rattani_adaptive_2010}.

Considering the topic of template update is still to be adequately addressed in ECG biometrics, this work studies the effect of ECG permanence and variability in long-term identification performance. Furthermore, it aimed to evaluate the effect of template update techniques, on the performance of several state-of-the-art methods.

\begin{table}[!t]
\centering
\caption[Graph-based template update methods and their respective loss and regulariser functions.]{Graph-based template update methods and their respective loss and regulariser functions (based on~\cite{zhu_semi-supervised_2008}).}\label{tab:longtermecg_3}
\begin{tabular}{lcll}
\hline
\textbf{Method} & \textbf{Source} & \textbf{Loss} & \textbf{Regulariser}
\\\hline
\begin{tabular}{l}Min Cut\end{tabular} & \cite{blum_learning_2001} & \(\displaystyle  \sum_{i \in L} (y_i-y_{i|L})^{2} \)  & \(\displaystyle\frac{1}{2} \sum_{ij} w_{ij} (y_i-y_j)^{2} \)\\

\begin{tabular}{l}Gaussian Random Fields\\and Harmonic Function\end{tabular}& \cite{zhu_semi-supervised_2003} & \(\displaystyle  \sum_{i\in L} (f_i-y_{i})^{2} \) &  \(\displaystyle f^{T}\Delta f\) \\

\begin{tabular}{l}Local and Global Consistency\end{tabular} & \cite{zha_graph-based_2009} & \(\displaystyle  \sum_{i=1}^{n} (f_i-y_{i})^{2} \) & \(\displaystyle  D^{-\frac{1}{2}}\Delta D^{\frac{1}{2}} \) \\

\begin{tabular}{l}Tikhonov Regularisation\end{tabular} & \cite{hutchison_regularization_2004} & \(\displaystyle\frac{1}{K} \sum_{i} (f_i-y_i)^{2} \) &  \(\displaystyle \gamma f^{T} S f\)\\

\begin{tabular}{l}Manifold Regularisation\end{tabular} & \cite{sindhwani_linear_2005} & \(\displaystyle\frac{1}{l} \sum_{i=1}^{l} V(x_i,Y_i,f) \) & \(\displaystyle\gamma_A\Vert f\Vert_{k}^{2} + \gamma_I\Vert f\Vert_I^{2}\) \\

\begin{tabular}{l}Graph Kernel from the\\Spectrum of Laplacian\end{tabular}& \cite{chapelle_semi-supervised_2006}  & \(\displaystyle \min \frac{1}{2} w^{T} W \) & \(\displaystyle \exp(-\frac{\sigma}{2} \lambda) \) \\

\begin{tabular}{l}Spectral Graph Transducer\end{tabular} & \cite{zhu_semi-supervised_2003}& \(\displaystyle \min c(f-\gamma)^{T} C (f-\gamma)\) & \(\displaystyle f^{T} L f\)\\

\begin{tabular}{l}Local Learning Regularisation\end{tabular} & \cite{kokkinos_local_2017} & \(\displaystyle \min \frac{1}{k} \sum_{i=1}^{k} (y_i-f_k(x_i))^{2}\) & \(\displaystyle \frac{\gamma}{k}\Vert f_k\Vert^{2}\) \\
\hline

\end{tabular}
\end{table}

\section{Methodology}

\subsection{Biometric identification methods}

To fully and objectively evaluate the effects of ECG variability on the performance of biometric algorithms, a study was conducted on four literature methods:

\begin{itemize}
    \item Plataniotis~\etal~\cite{Plataniotis2006} proposed an ECG biometric recognition method using a non-fiducial approach. Signals are preprocessed using a bandpass filter ($0.5-40$ Hz), followed by feature extraction with autocorrelation (AC) and dimensionality reduction using discrete cosine transform (DCT). The fifteen most relevant features were selected, and Euclidean distance was used for classification;
    
    \item Tawfik~\etal~\cite{Tawfiq2010} used a bandpass filter ($1-40$ Hz) in the preprocessing phase. QRS complexes (the most stable part of ECG) were cut from the signal using a $0.35$ second window. The average ensemble QRS was computed and features were extracted using the DCT technique (the thirty most relevant features were selected). A multilayer perceptron (MLP) is used for classification;
    
    \item Belgacem~\etal~\cite{Belgacem2012} also preprocessed signals with a bandpass filter ($1-40$ Hz). The QRS complexes were located and cut from the signal, and the average QRS was computed. The feature extraction resorted to Discrete Wavelet Transform (DWT). From all DWT decomposition levels, only the most relevant were selected, and a Random Forest is used for classification. This method was originally proposed for authentication and adapted here for identification;
    
    \item Eduardo~\etal~\cite{Eduardo2017} used a Finite Impulse Response ($5-20$ Hz) filter for preprocessing. Heartbeats were cut with a fixed length of $[-200, 400]$ ms around each R peak. Outliers were detected and removed using DMEAN ($\alpha = 0.5$ and $\beta = 1.5$, with Euclidean distance). For decision, the k-nearest neighbours (kNN) classifier was used with $k=3$ and cosine distance.
\end{itemize}

Beyond these literature methods, this work also explored the deep learning-based methodology presented in Chapter~\ref{ch:ecgiden} (proposed in Pinto~\etal~\cite{Pinto2019Deep}). This method uses an end-to-end unidimensional convolutional neural network, that receives five-second blindly-segmented z-score normalised ECG segments to perform biometric identification. The feature extraction part of the model is composed of four convolutional layers, interleaved with three max-pooling layers, with filter/pooling size $1\times5$ and ReLU activation units. For classification, the network uses a single fully-connected layer and softmax activation units.

\subsection{Template update methods}
\paragraph{FIFO:} First-In-First-Out is the most common strategy and, computationally, is very lightweight. Here, the database is updated using new samples whose score is above or below a threshold (whether the score represents similarity or dissimilarity, respectively), or between two threshold values (discarding previously stored  sample)~\cite{komeili_liveness_2018,Coutinho2011}. The score of a new sample can either be output by a classifier or be a measure of distance or similarity between that sample and the stored templates~\cite{Lourenco2011}.

\begin{figure}[!t]
\centering
\includegraphics[width=0.65\linewidth]{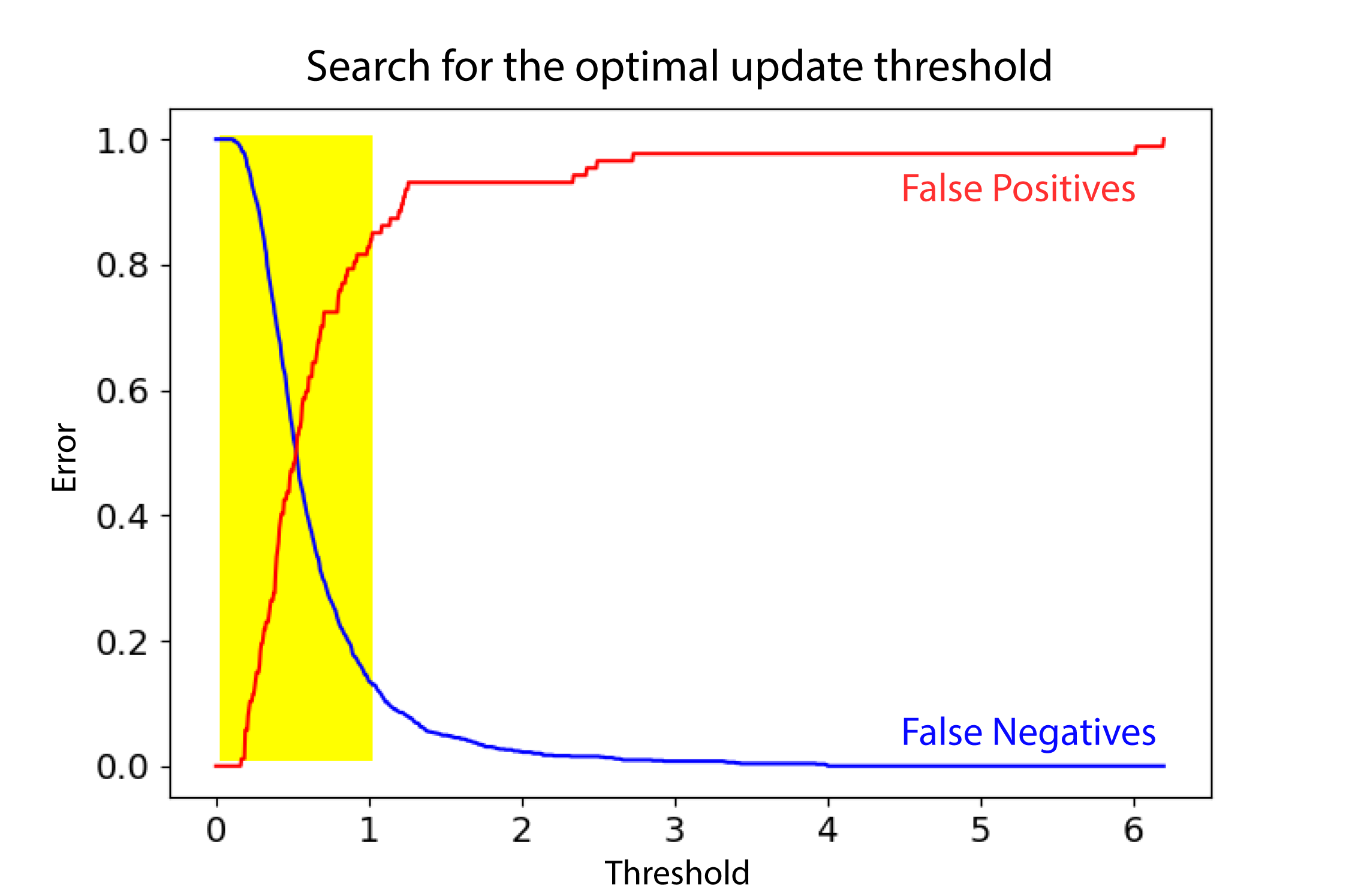}
\caption[Illustration of the search for the ideal threshold.]{Illustration of the search for the ideal threshold (the values were chosen near the intersection, inside the yellow zone).}
\label{fig:longTerm_threshold}
\end{figure}

In this work, the training data were used to search for threshold values. Among all training samples, $75$\% were used to train a model, which was used to obtain scores for the remaining data samples. Comparing the scores with several thresholds, the error at each threshold was analysed (Fig.~\ref{fig:longTerm_threshold}) to find one that simultaneously maximises true positives and minimises false positives. 

\paragraph{Fixation:} This method consists of fixing certain templates, allowing only the remaining stored samples to be updated~\cite{guerra-casanova_score_2011}. In this work, $25$, $50$, or $75$\% of the enrollment templates of the individual are fixed, while the rest of the samples are free to be updated. This ensures some initial, labelled information of the subjects remains on the system over time.

An adaptation of this technique was explored. Here, $n + j \times n$ samples were fixed, where $n\in[1,2,3]$ is the number of fixed initial templates, and $j$ increases over time. In this work, $j \in [0,6]$ increased by one at each testing moment ($j \in [0,6]$), which allowed the system to fix more and more samples over time, thus storing information on the subject's variability over time. In a real system with potentially endless use, the parameters $n$ and $j$ should be carefully chosen to avoid the eventual fixation of the entire template gallery.

\paragraph{Fine-Tuning:}  In this technique, the model is briefly optimised with the samples accepted for update, using the predicted labels. The model retains knowledge of the users' supervised training samples, as it was trained using their enrollment samples, but is slightly adapted to the new personal patterns carried by the new signals. This is explored exclusively for the CNN method~\cite{Pinto2019Deep}.

\section{Experimental Setup}

\subsection{Dataset}

For evaluation, the ECG signals used were from the E-HOL-03-0202-003 database\footnote{\label{note1}THEW. Available on: \url{http://thew-project.org/Database/E-HOL-03-0202-003.html}.} (most commonly designated as E-HOL 24h). This database consists of a study of 202 healthy subjects (only 201 were provided), recorded using three leads at $200$ Hz sampling frequency, after an initial resting supine period of 20 minutes.
From the available data of $201$ subjects, thirteen were discarded due to saturation or unacceptable noise (subjects $1043$, $9003$, $9005$, $9020$, $9021$, $9022$, $9025$, $9046$, $9061$, $9064$, $9071$, $9082$ and $9105$), similar to what was done by Labati~\etal~\cite{Labati2014}. From each of the remaining $188$ subjects, only the lead most closely resembling Lead I ECG was selected, to approximate off-the-person settings.

\subsection{Experiments}

Standard sample wise normalisation was performed following Eq.~\eqref{eq:25} for all methods except that of Eduardo~\etal~\cite{Eduardo2017}, which required $[-1,1]$ \emph{min-max} normalisation, Eq.~\eqref{eq:26}, where $x$ represents the input signal and $\tilde{x}$ the normalised signal.
 
\begin{equation}
  \tilde{x}[n] =\frac{x[n]-\overline{x}}{\sigma(x)}
  \label{eq:25}
\end{equation}

\begin{equation}
  \tilde{x}[n] =2\left(\frac{x[n]-\min(x)}{\max(x)-\min(x)}\right)-1
  \label{eq:26}
\end{equation}

In order to fit the used data, some changes were introduced to the original methods. On the method from Eduardo~\etal, the cutoff frequencies of the bandpass filter were changed to $1$ and $40$ Hz, to retrieve important information on higher frequencies. Outlier removal was reparametrized with $\alpha = 1.2$ and $\beta = 1.5$. The autoencoder had the topology $[120, 60, 40, 20, 40, 60, 120]$  and was trained using the Adam optimiser with a learning rate $0.01$. Classification used $k = 1$. For the method of Belgacem~\etal, DWT feature extraction was performed in four decomposition levels, due to lower data sampling rate, and $cd4$, $cd3$, $cd2$, and $cd1$ coefficients were used as features.

\begin{figure}[!t]
\includegraphics[width=\textwidth]{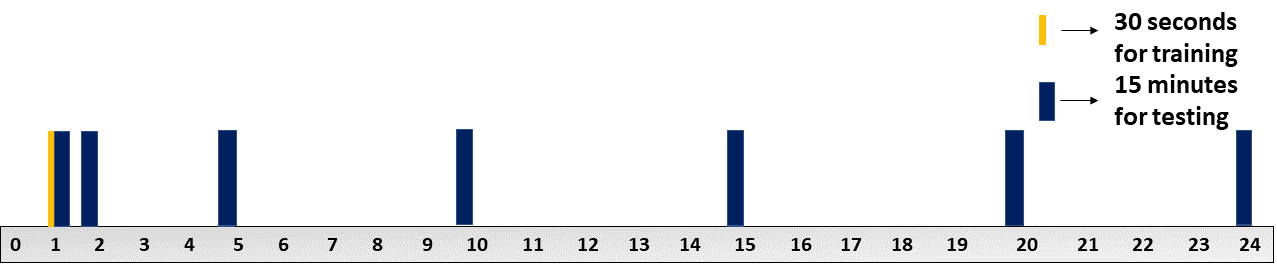}		
\centering
\caption[Schema illustrating the use of each E-HOL record for training and testing.]{Schema illustrating the use of each E-HOL record for training and testing (in orange - training segment; in blue - each test segment).}
\label{fig:longTerm_codigo_timeline}
\end{figure}

Data were divided into train and test sets. The training phase used the last $30$ seconds (mimicking short enrollments on real-life applications) of the first $60$ minutes (avoiding unrealistic calm after the initial resting period) of each subject. Five-second overlap was used to obtain $26$ samples from every $30$ seconds of training data. To study performance over time, testing was performed over seven time points (see Fig.~\ref{fig:longTerm_codigo_timeline}): one immediately after enrollment, another after one hour, and regularly until the end of the records. From each point, from 15 minutes of data, thirty $30$ s samples are extracted, and batches are built with one sample from each of the $188$ subjects.

\section{Results and Discussion}

\subsection{Handcrafted methodologies}

\begin{figure}[!t]
 	\begin{subfigure}[b]{0.5\textwidth}
  		\includegraphics[width=\textwidth]{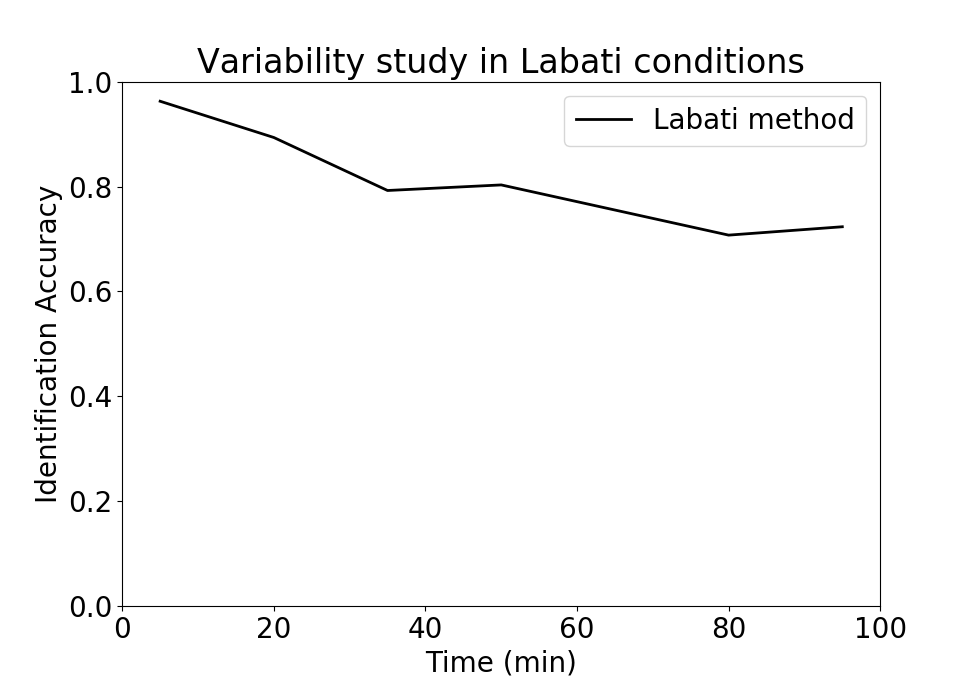}		
  		\caption{}	
  		\label{fig:longTerm_labati}
  	\end{subfigure}%
  	\begin{subfigure}[b]{0.5\textwidth}				
  		\includegraphics[width=\textwidth]{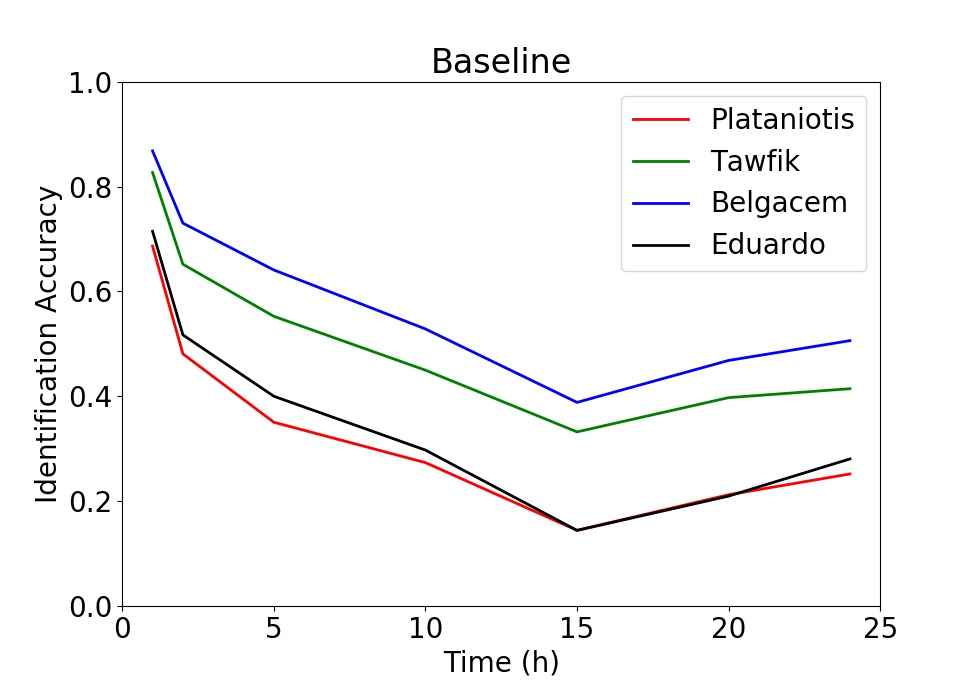}		
  		\caption{}
  		\label{fig:longTerm_variability_results}
  	\end{subfigure}
\centering
\caption{Identification performance over time corresponding to (a) the Labati~\etal~method, and (b) the implemented state-of-the-art methods.}
\label{variability}
\end{figure}

After implementing, for identification, the method proposed by Labati~\etal~\cite{Labati2013} (replicating their evaluation conditions), it was possible to conclude that the ECG signal is not fully permanent over 24 h. However, similarly to what was stated by Labati~\etal, the results are relatively good over the first two hours (see Fig~\ref{variability}\subref{fig:longTerm_labati}), although permanence was not verified.

The performance results at each test hour, obtained through the weighted average of the corresponding batches, for the state-of-the-art methods can be found in Fig.~\ref{variability}\subref{fig:longTerm_variability_results}. It was found that the performance is mostly acceptable in the first test point, but performance decays significantly over time and variability changes considerably over the day.

Moreover, a minimum around the 15\textsuperscript{th} hour occurs independently of the chosen method. Considering that most of the records start between 8-12 am, after 15 hours the subjects must be sleeping. In this perspective, it appears that the ECG is most different from normal when the subject is asleep.

\begin{figure}[!p]
 	\begin{subfigure}[b]{0.5\textwidth}
  		\includegraphics[width=\textwidth]{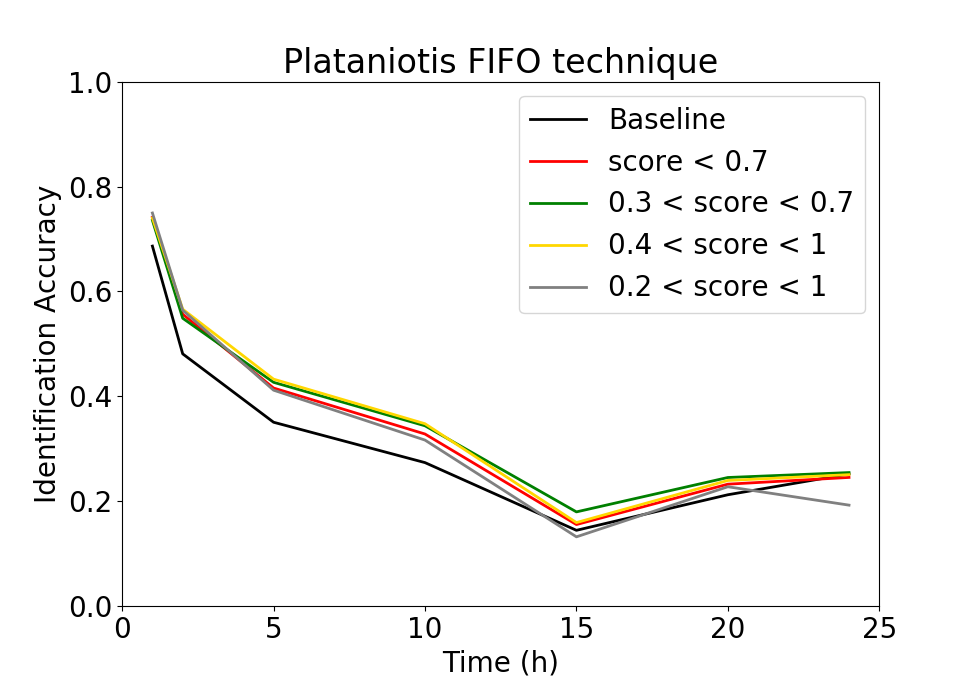}		
  		\label{fig:longTerm_fifo_plataniotis}
  	\end{subfigure}%
  	\begin{subfigure}[b]{0.5\textwidth}						
  		\includegraphics[width=\textwidth]{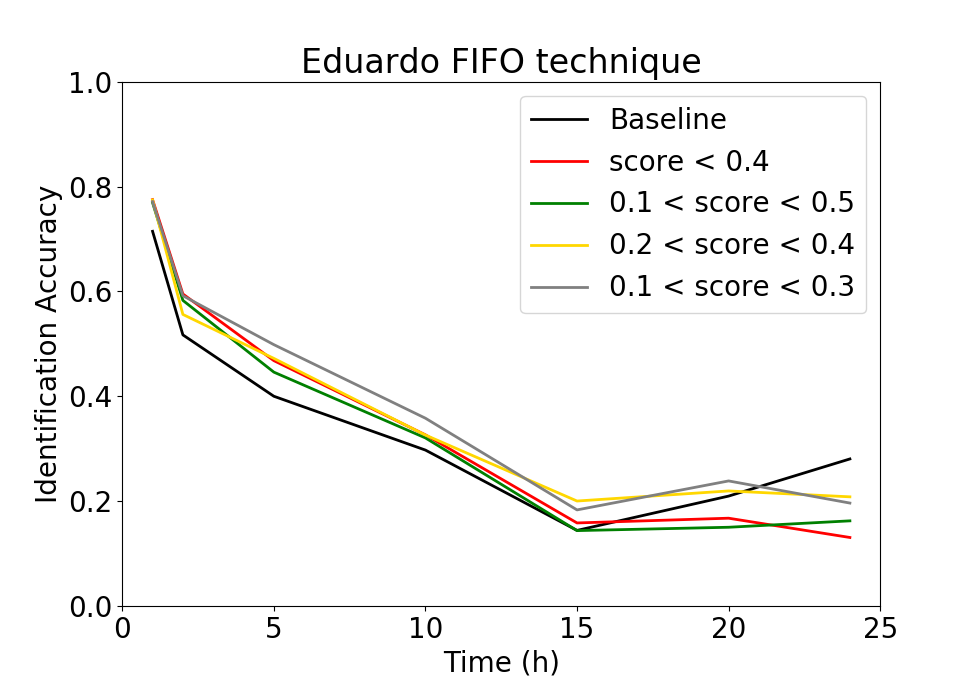}		
  		\label{fig:longTerm_fifo_autoencoder}
  	\end{subfigure}
  	\begin{subfigure}[b]{0.5\textwidth}						
  		\includegraphics[width=\textwidth]{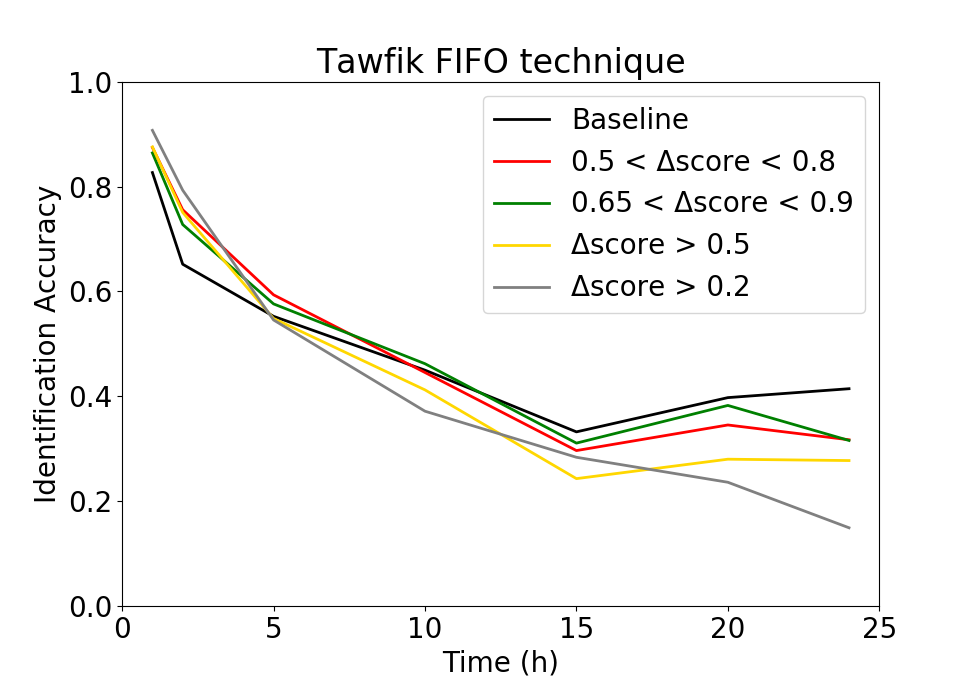}
  		\label{fig:longTerm_fifo_tawfik}
  	\end{subfigure}%
  	\begin{subfigure}[b]{0.5\textwidth}						
  		\includegraphics[width=\textwidth]{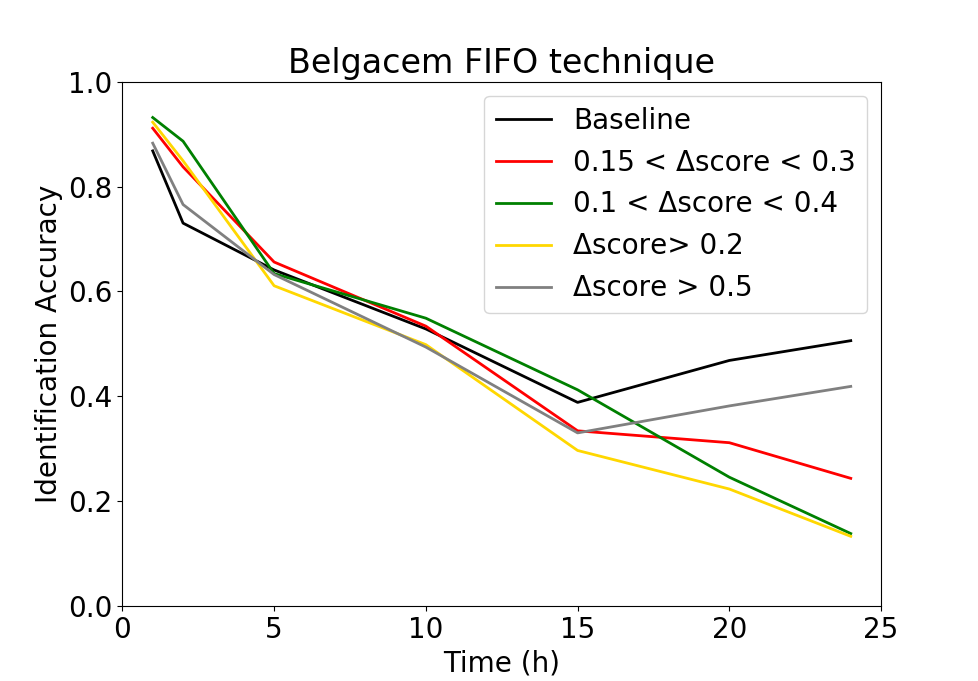}		
  		\label{fig:longTerm_fifo_belgacem}
  	\end{subfigure}
\centering
\caption{Comparison of the FIFO method applied with different thresholds to different identification methodologies.}
\label{fifo}
\end{figure}
\begin{figure}[!p]
 	\begin{subfigure}[b]{0.5\textwidth}
  		\includegraphics[width=\textwidth]{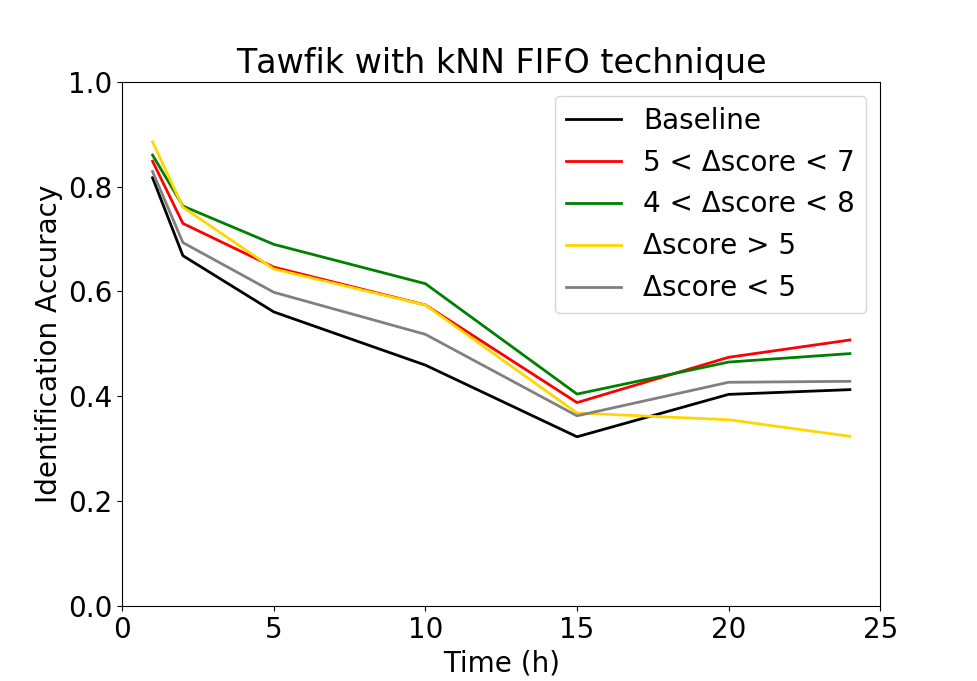}		
  		\label{fig:longTerm_fifo_knn_tawfik}
  	\end{subfigure}%
  	\begin{subfigure}[b]{0.5\textwidth}		
  		\includegraphics[width=\textwidth]{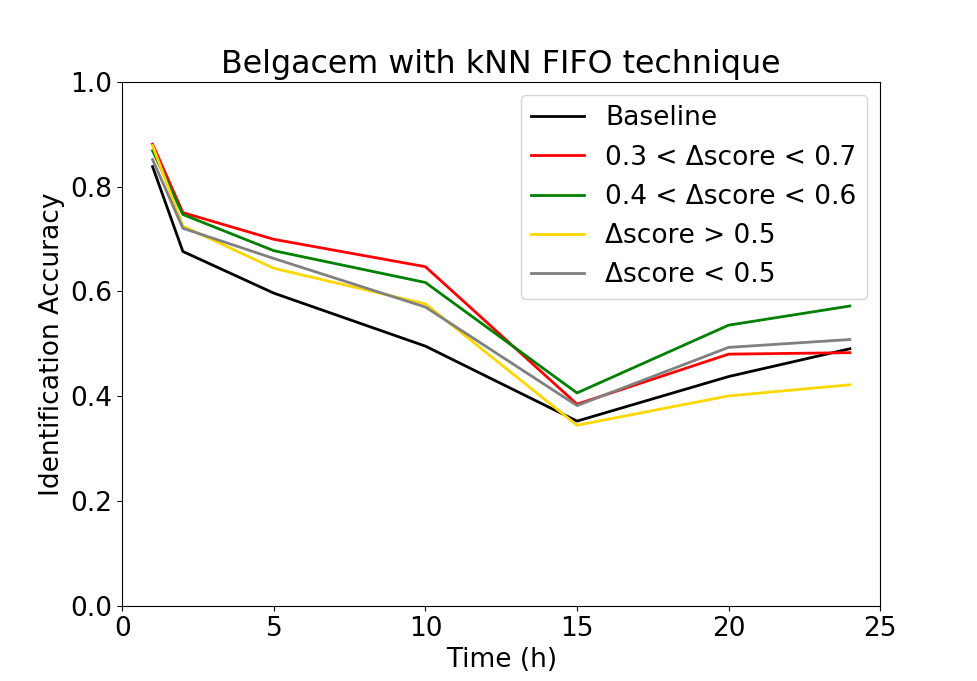}		
  		\label{fig:longTerm_fifo_knn_belgacem}
  	\end{subfigure}
\centering
\caption{Results using FIFO update with different thresholds.}
\label{change}
\end{figure}

\begin{figure}[!t]
 	\begin{subfigure}[b]{0.5\textwidth}
  		\includegraphics[width=\textwidth]{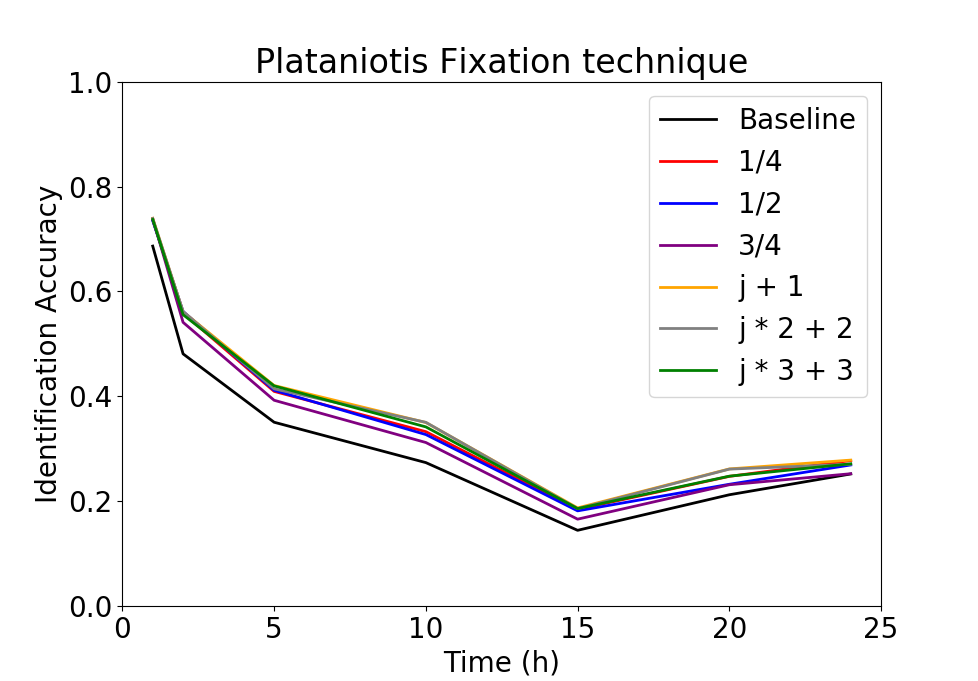}	
  		\label{fig:longTerm_fixation_plataniotis}
  	\end{subfigure}%
  	\begin{subfigure}[b]{0.5\textwidth}		
  		\includegraphics[width=\textwidth]{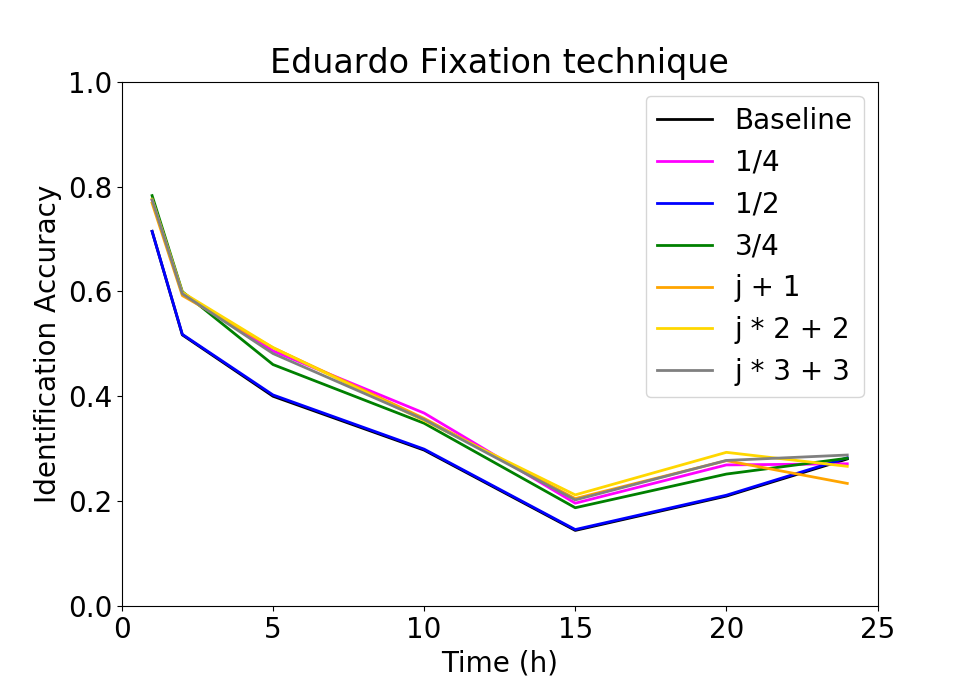}		
  		\label{fig:longTerm_fixation_autoencoder}
  	\end{subfigure}
  	\begin{subfigure}[b]{0.5\textwidth}		
  		\includegraphics[width=\textwidth]{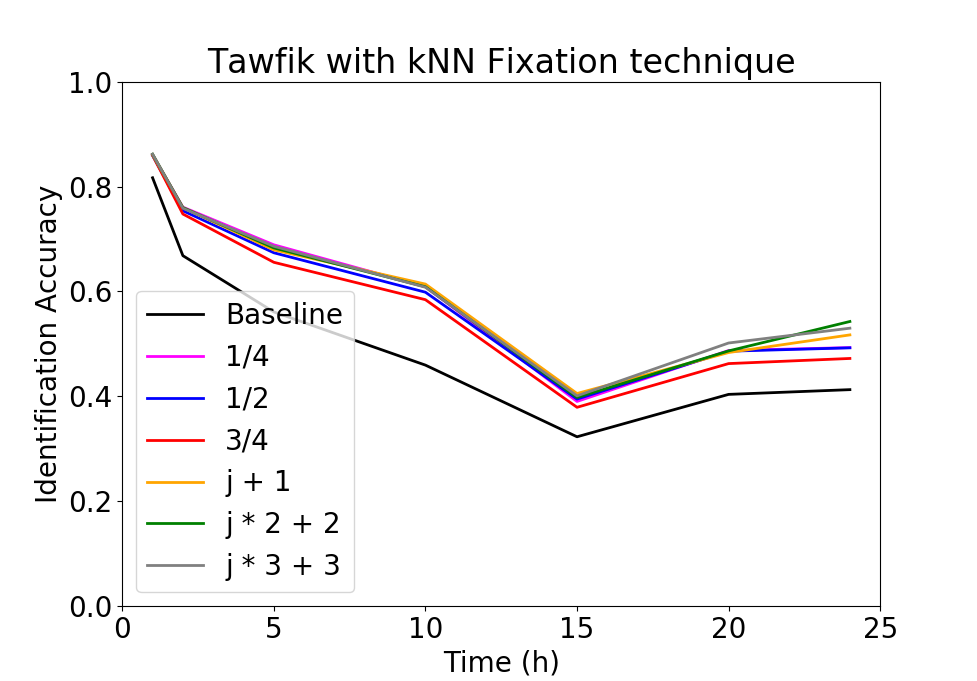}	
  		\label{fig:longTerm_fixation_tawfik_knn}
  	\end{subfigure}%
  	\begin{subfigure}[b]{0.5\textwidth}		
  		\includegraphics[width=\textwidth]{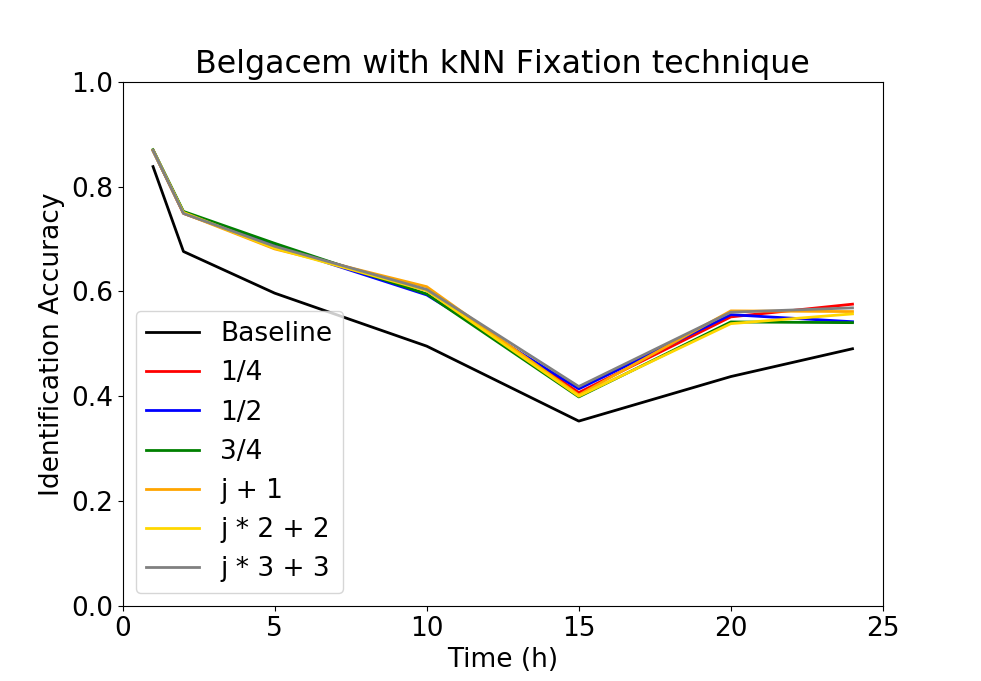}		
  		\label{fig:longTerm_fixation_belgacem_knn}
  	\end{subfigure}
\centering
\caption[Results using Fixation update.]{Results using Fixation update (the corresponding value represents the number of samples that were fixated per subject).}
\label{fixation}
\end{figure}

Considering the previous results, template update was applied to the methods, in an effort to avoid performance decay over time. Fig.~\ref{fifo} presents the results using the FIFO technique, with diverse thresholds.

For the methods of Plataniotis~\etal~and Eduardo~\etal, the best results were obtained using two thresholds, respectively, $\{0.3, 0.7\}$ ($4.7\%$ accuracy improvement) and $\{0.1, 0.3\}$ ($+$ $5.7\%$ accuracy), improving all performance results until the 15\textsuperscript{th} hour. However, for the method of Belgacem~\etal, the performance worsens with template update after the first two hours (best results were obtained when the difference between the highest and second highest scores $\Delta score \in [0.15, 0.3]$). The same was verified for the method of Tawfik~\etal~which, in the first two hours, offered the best results with $\Delta score > 0.2$. In general, using two thresholds instead of one offered the best results.

Considering this, it appeared that the Random Forest and MLP classifiers are not suitable for these kinds of template/model update. This was confirmed after a repetition of the evaluation of these methods, with kNN replacing the classifiers (see  Fig.~\ref{change}). With kNN, the template update was able to reduce the performance decay over time, improving accuracy, on average, by $7.9\%$ and $9.2\%$, respectively, for the methods of Belgacem~\etal~and Tawfik~\etal

As for the Fixation technique, the obtained results were more promising (see Fig.~\ref{fixation}). This template update technique brought performance improvements for all methods. The fixation technique that offered the best results was $j \times 3 + 3$, improving the baseline identification accuracy, on average, by $10.0\%$.

\subsection{Deep convolutional network}

\begin{figure}[!t]
    \centering
    \includegraphics[width=0.7\linewidth]{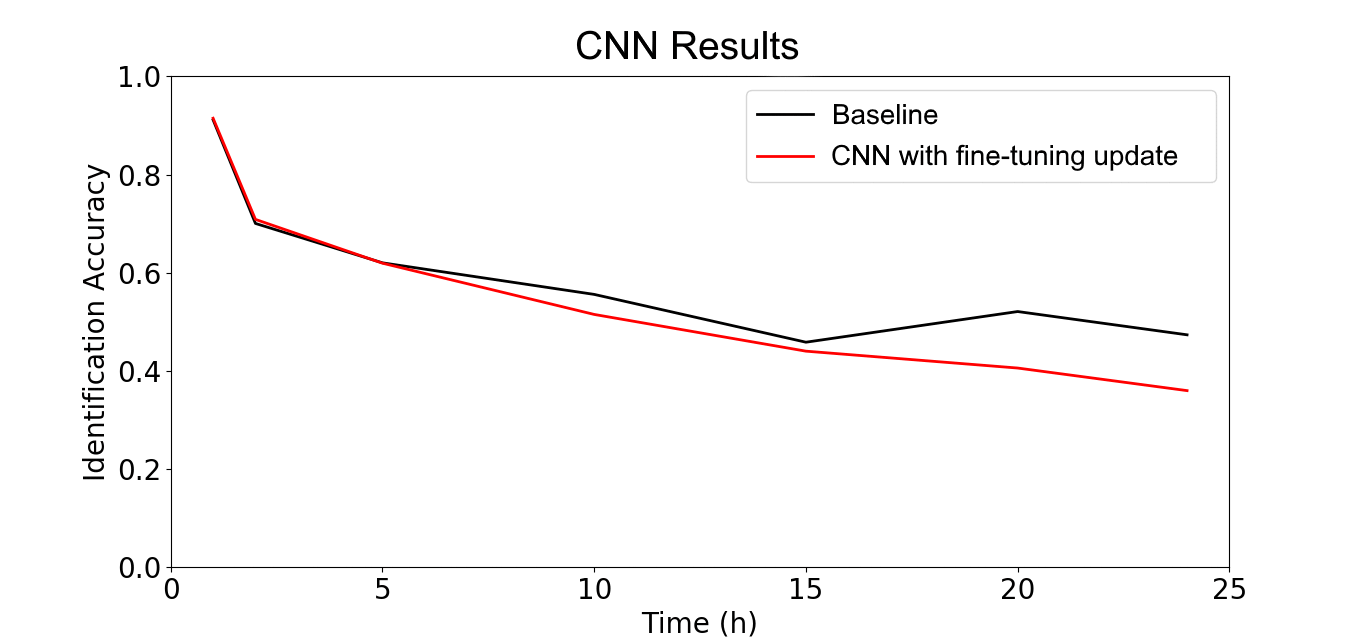}
    \caption{Performance results over time for the CNN model with fine-tuning-based model update.}
    \label{fig:longTerm_cnn_finetuning}
\end{figure}

The results for the implemented end-to-end convolutional neural network are presented in Fig.~\ref{fig:longTerm_cnn_finetuning}. Model update offered a small improvement in performance in the first test point ($91.48\%$ \emph{versus} $91.15$ without update). However, the model experiences sharp performance decay and, after the fifth hour, the model update is unable to improve identification accuracy. In fact, model update caused a decrease in identification rate, which is coherent with the findings regarding update with multilayer perceptron classifiers reported by Lopes~\etal~\cite{Lopes2019}.

\begin{figure}[!t]
    \centering
    \includegraphics[width=0.7\linewidth]{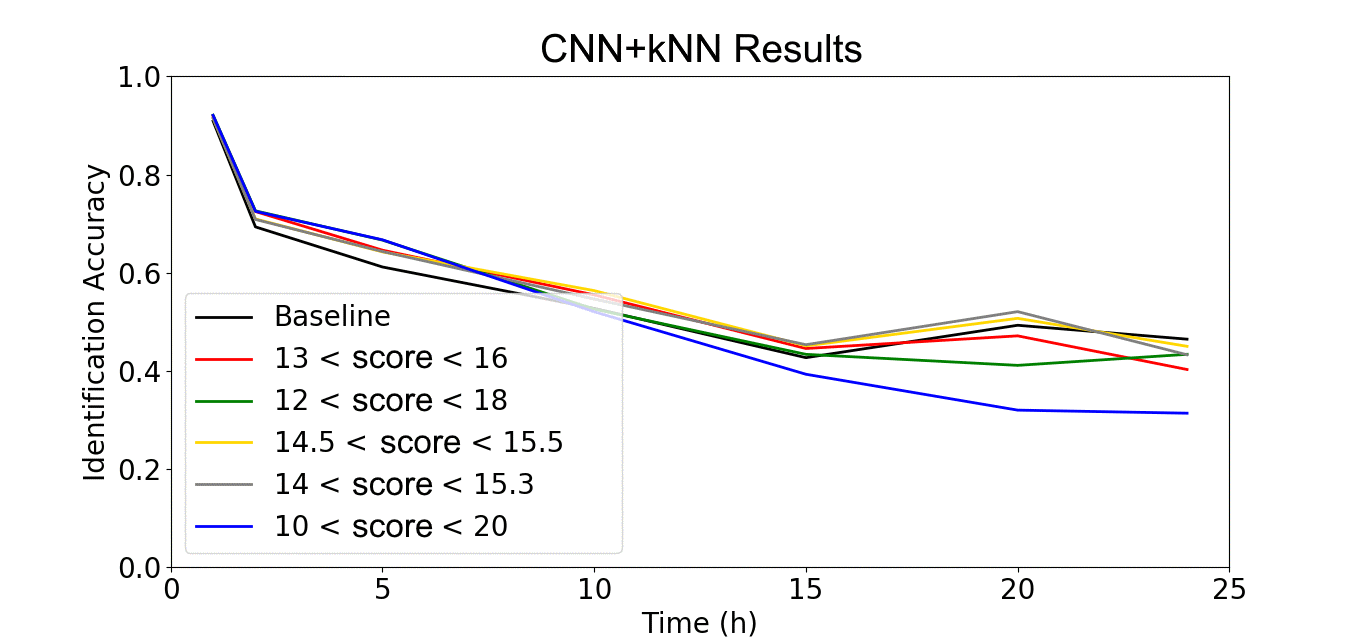}
    \caption[Performance results over time for the CNN model adapted with kNN decision and FIFO template update, for several threshold criteria.]{Performance results over time for the CNN model adapted with kNN decision and FIFO template update, for several threshold criteria (samples with scores between the presented values are accepted for update).}
    \label{fig:longTerm_cnn_fifo}
\end{figure}

Different results were obtained when the fully-connected layer of the network was replaced by a kNN classifier (see Fig.~\ref{fig:longTerm_cnn_fifo}). Although the results with kNN are slightly worse than those of the end-to-end network ($90.89\%$ \emph{vs.} $91.15\%$ for the first test point), the template update technique is more successful and is able to offer performance improvements for almost all test points.  

When compared with the results reported by Lopes~\etal, the CNN (either end-to-end or with kNN classification) offers the best performance in the first test points after enrollment. However, it gradually loses that advantage as time passes, even with template update. 

Likely, the network will require more data with more variability during the first training phase. Increasing the training data to thirty minutes or even a few hours per subject would enable the network to better learn the common variability patterns of the ECG. This should not only increase the initial performance, immediately after enrolment but also reduce the performance decay over time.

\section{Summary and Conclusions}

This work studied how ECG variability affects the performance of state-of-the-art biometric algorithms, and how template update could mitigate performance decay over time. The results have shown long-term identification performance in ECG biometrics is generally weak, despite the promising results often presented in the literature. 

Template update techniques proved successful in enhancing the long-term performance of handcrafted state-of-the-art methods, especially when using template fixation techniques. Additionally, with a deep learning algorithm, results are better than traditional methods immediately after enrollment, although it offers slightly worse performance as time progresses.

Generally, one can conclude that further efforts are needed for the study and development of more advanced techniques. The obtained results in these more realistic settings show that the performance levels commonly reported in the literature would likely not be verified upon real application. Special focus should be devoted to supervised update techniques, so that ECG-based biometric systems can offer reliable performances over long periods.
\chapter[Leveraging Explainability to Understand ECG Biometrics]{Leveraging Explainability\\to Understand ECG Biometrics}\label{ch:ecgexpl}

\begin{tcolorbox}\footnotesize
{\large\bf Foreword on Author Contributions}

The research work described in this chapter was conducted entirely by the author of this thesis, under the supervision of Jaime S. Cardoso. The results of this work have been disseminated in the form of an article in international conference proceedings and an abstract in national conference proceedings:
\begin{itemize}[noitemsep, leftmargin=1em, nosep]
    \item \underline{J. R. Pinto} and J. S. Cardoso, ``Explaining ECG Biometrics: Is It All In The QRS?,'' in \emph{International Conference of the Biometrics Special Interest Group (BIOSIG 2020)}, Sep.~2020.~\cite{Pinto2020Explaining}
    \item \underline{J. R. Pinto} and J. S. Cardoso, ``xECG: Using Interpretability to Understand Deep ECG Biometrics,'' in \emph{27th Portuguese Conference on Pattern Recognition (RECPAD 2021)}, Nov.~2021.
\end{itemize}

\end{tcolorbox}

\section{Context and Motivation}

Throughout the past twenty years, research on biometrics based on the electrocardiogram (ECG) has largely been a success story~\cite{Pinto2018}. After successful proofs-of-concept in cleaner medical signals (\emph{on-the-person}), the focus is quickly shifting to acquisitions in more realistic scenarios (\emph{off-the-person}). Deep learning approaches~\cite{Labati2018, Luz2018, Pinto2019Deep, Pinto2019b, Hammad2020resnet} have been essential in dealing with the increased noise and variability in off-the-person settings, despite the performance and robustness issues that still hinder application in real scenarios.

However, deep learning decisions are obscure: unlike traditional methods based on fiducial features, we don't know what information the model uses to distinguish people~\cite{samek2017explainable,doshi2017towards}. One can assume that the models look mainly to the QRS since it is the most stable part of the ECG in the face of noise and variability~\cite{Schijvenaars2000, Hoekema2001}. Several methods have thus focused on QRS complexes for ECG biometrics~\cite{Tuerxunwaili2016, Labati2018}, but this practice has become uncommon in recent works. This indicates the true role of this waveform complex in identity discrimination is still to be adequately recognised. 

Currently, pattern recognition researchers understand the importance of knowing what specific information is relevant for their models to reach decisions. Retreating to easily explainable traditional models (such as decision trees) is often unacceptable due to their performance limitations. Hence, various interpretability tools are being developed to peek into the inner workings of deep networks applied to diverse tasks~\cite{Carvalho2019, Silva2019, Sequeira2020Interpretable}.

This work uses, for the first time in the literature, such interpretability tools on a deep ECG biometric model, to understand what parts of the ECG are most useful for automatic human identification. The model is a competitive state-of-the-art method~\cite{Pinto2019Deep, Pinto2019b} applied for ECG-based identification in data subsets with diverse signal quality and number of identities. With this, we aim to assert the importance of the QRS and other waveforms for ECG biometrics and discuss future possibilities as this topic evolves towards more challenging and realistic scenarios. Additionally, we propose an intuitive way to visualise interpretations for unidimensional signals. The code and additional results are available online\footnote{xECG Github Repository. Available on: \url{https://github.com/jtrpinto/xECG}.}.

\section{The Electrocardiogram as a Biometric Trait}\label{sec:explainecg_ecg}

\begin{figure}[t]
    \centering
    \includegraphics[height=3.5cm]{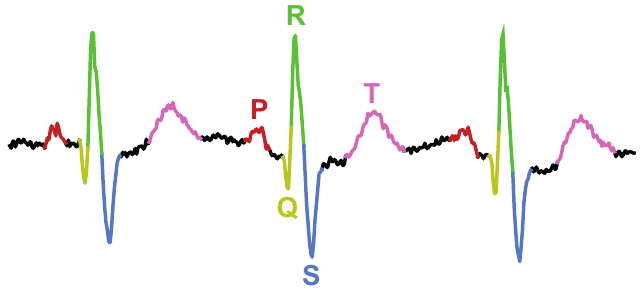}
    \caption{Illustration of the ECG waveforms on a sample PTB signal segment.}
    \label{fig:explainecg_ecg}
\end{figure}

As presented in Chapter~\ref{ch:fundamentals}, subsection~\ref{subsec:ecg_fundamentals}, the ECG is approximately a cyclical repetition of a set of waveforms (P, Q, R, S, and T) that corresponds to a heartbeat (see Fig.~\ref{fig:explainecg_ecg})~\cite{Marieb2013, Pinto2018}. Each of these waves corresponds to specific phenomena involved in the heart's contraction and relaxation.

As a measurement of the electrical currents spread across the heart, the ECG signals will reflect the geometry of this organ. For example, larger hearts, with more cells to depolarise and repolarise, will result in ECG waveforms with larger amplitudes. Higher or lower basal heart rates will also result in different signal morphologies. Since heart geometry and basal heart rates vary across individuals, this intersubject variability is what makes the ECG sufficiently unique to be used in biometric recognition~\cite{Hoekema2001, VanOosterom2000}.

However, the ECG signals are also susceptible to intrasubject variability factors. Noise sources during acquisition, the short-term and long-term effects of exercise, emotional states, stress, drowsiness, and fatigue are some of the factors that reflect mainly in the heart rate variability, changing the morphology of the P-R and S-T segments~\cite{Schijvenaars2000, Agrafioti2012}. These are the sources of uncertainty that hinder the use of the ECG as a biometric trait. While these are largely controlled in medical or on-the-person settings (where the subject is at rest, laying down, and signals are acquired using several high-quality gel electrodes), their effects are dominant for realistic off-the-person signals (acquired using fewer dry electrodes on the hands, during common daily activities)~\cite{Pinto2017, Pinto2018, Pinto2019b}. 

When compared with the P and T waves, the QRS corresponds to a larger polarisation event over a shorter period. In practice, this makes the QRS more dominant over noise and intrasubject variability than the other ECG waveforms~\cite{Pinto2017, Pinto2018}. Hence, the QRS is considered more stable over time and across variable conditions, which makes it better suited for biometric recognition.

Despite this, it is still unclear how much identity information is carried by the QRS complex compared to the other waveforms, and whether it is enough for an accurate and robust biometric recognition system. Studies on ECG-based biometric identification have shown it is possible to distinguish small sets of individuals in on-the-person settings using only the QRS complex or QRS fiducial amplitude and time measurements~\cite{Tuerxunwaili2016, Labati2018}. Nevertheless, this practice is becoming uncommon as research evolves towards realistic off-the-person signals and larger databases.

This denotes that the sole use of the QRS may not be adequate for off-the-person settings, or the individual information carried by the QRS may not be enough to distinguish individuals in large populations. This work aimed to address these doubts through a study on the role and relevance of the QRS and the other waveforms in ECG-based biometric identification. Interpretability tools are used to assess which parts of the ECG are more relevant to the decisions of an end-to-end identification model~\cite{Pinto2019Deep}, with on-the-person and off-the-person signals and data subsets with a varying number of identities.

\section{Methodology}\label{sec:explainecg_methodology}

\subsection{Biometric identification model}

\begin{figure}[t]
    \centering
    \includegraphics[height=3cm]{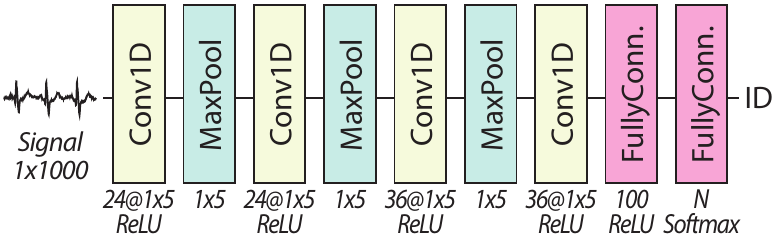}
    \caption{Architecture of the biometric identification model.}
    \label{fig:explainecg_model}
\end{figure}

The biometric model for identification followed the architecture proposed by Pinto~\emph{et al.}~\cite{Pinto2019Deep}, which has attained state-of-the-art results in off-the-person settings for both identification and, later, identity verification~\cite{Pinto2019b}. The model (see Fig.~\ref{fig:explainecg_model}) receives five-second blindly segmented ECG signals and outputs probabilities for each of the $N$ identities considered. Finding the highest probability score allows us to assign the respective identity to the input signal.

The model consists of an end-to-end 1D convolutional neural network (CNN) with four convolutional layers (with $1\times5$ filters, two layers with $24$ followed by two with $36$), followed by ReLU activation. Neighbouring convolutional layers are separated by $1\times5$ max-pooling layers. The last convolutional layer is followed by two fully-connected layers ($100$ neurons with ReLU and $N$ neurons with softmax activation).

\subsection{Interpretability tools}\label{subsec:explainecg_int_tools}

To capture the dynamics behind the decisions of the biometric model, four interpretability methods are applied to the trained model: Occlusion, Saliency, Gradient SHAP, and DeepLIFT. Occlusion and Saliency are two of the simplest interpretability methods, while Gradient SHAP and DeepLIFT are more sophisticated and powerful. These are implemented in the Captum toolbox~\cite{Captum2019github} for PyTorch and are described below.

\paragraph{Occlusion} The Occlusion method~\cite{Zeiler2014} consists in measuring the influence of hiding a portion of the input on the output of the model. When hidden, the more relevant input parts will cause larger changes in the output, and will thus be assigned greater relevance in the explanations offered by this method. This is the simplest method to interpret a model, although the size of occluded regions should be carefully defined to obtain meaningful explanations. 

\paragraph{Saliency} The Saliency method~\cite{Simonyan2014DeepInside} is based on the gradients of a model given a certain input. Through backpropagation, the gradient of target class scores w.r.t. the input is obtained. A saliency map is then generated by rearranging the class score derivatives, generating saliency maps that assign higher relevance to input regions that correspond to higher gradients. Requiring a single backpropagation pass, this method is a simple and fast way to obtain explanations of model predictions.

\paragraph{Gradient~SHAP} Gradient SHAP~\cite{Lundberg2017} is an approach based on game theory which considers the explanations of a model's predictions as models themselves. For sophisticated deep learning models, the explanation models are simplified and interpretable approximations of the respective models. SHapley Additive exPlanation (SHAP) values, inspired by game theory's Shapley values, are computed through the gradient of a random point between a baseline and the input with added random noise. The SHAP values denote how much a given part of the input raises the probability for the considered class, and are reportedly better aligned with human intuition and effective in discriminating among output classes.

\paragraph{DeepLIFT} DeepLIFT (Deep Learning Important FeaTures)~\cite{Shrikumar2017} performs backpropagation to track the contributions to the output to the responsible parts of the input. Throughout this process, it compares the difference in inputs and outputs considering a reference (or baseline) input, assigning contribution scores to each neuron of the model. It also allows for the study of negative contributions: how much a specific part of the input contributes to lower the probability for the considered class.

\subsection{Visualisation}\label{subsec:explainecg_vis}

Decision explanations obtained using interpretability tools are visualised using the multicoloured line plot feature of Matplotlib~\cite{Matplotlib}. ECG signals are plotted so that the colour of each signal component represents its relative relevance to the decision. In this case, lighter yellow colours represent less relevant time samples, whereas more relevant samples assume darker purple colours. This way, both the ECG morphology and the relevance of each of its components are easily and intuitively presented.

\section{Experimental Setup}\label{sec:explainecg_exp_settings}

The data used for model training and evaluation have been drawn from the Physikalisch-Technische Bundesanstalt ECG Database (PTB)~\cite{Bousseljot1995, Goldberger2000} and the University of Toronto ECG Database (UofTDB)~\cite{Wahabi2014}. The PTB database includes on-the-person (high-quality) 12-lead ECG signals acquired at 1 kHz from $290$ subjects at rest. The UofTDB includes single-lead off-the-person (more noisy and realistic) data acquired from $1019$ subjects. To match the UofTDB, PTB signals were downsampled to $200$ Hz and only Lead I was used.

Five-second segments were blindly extracted (without fiducial detection) from the recordings. Fifty per cent of those segments (\emph{per} identity) were used during training and the remaining were reserved for testing. This provided more challenging test settings than those commonly found in the literature, but also deliberately avoided the most realistic settings (see \cite{Pinto2019b}), for the sake of obtaining meaningful interpretations.

To simulate gradually increasing identification difficulty within each database, subsets of $N$ identities are considered, with $N\in\{2,5,10,20,50,100,200,500,1019\}$. The identities in each subset are the first $N$ in lexicographical order. Each subset includes all identities that compose smaller subsets, so subjects \#1 and \#2 are the main focus of analysis since these are present in all subsets. Throughout this paper, $T_N$ denotes the subset of UofTDB data from $N$ subjects and $P_N$ denotes the subset of PTB data from $N$ identities. As stated in Table~\ref{table:performance}, $P_{290}$ was used instead of $P_{200}$ to take advantage of the entire PTB dataset. Model training details can be found online at this project's repository.

Performance evaluation is based on the True Positive Identification Rate (or accuracy): the fraction of test samples that are correctly assigned to their true identity by the trained model. Interpretations are examined through the proposed visualisation method.

\section{Results and Discussion}\label{sec:explainecg_results}

\begin{table}[t]
\centering
\caption{\label{table:performance} True positive identification rate results (\%) on the test data.}
\begin{tabular}{l c c c c c c c c c}
\hline
\multirow{2}{*}{\textbf{Database}} & \multicolumn{9}{c}{\textbf{Number of Identities}} \\
& \textit{\textbf{2}} & \textit{\textbf{5}} & \textit{\textbf{10}} & \textit{\textbf{20}} & \textit{\textbf{50}} & \textit{\textbf{100}} & \textit{\textbf{200}}\textsuperscript{1} & \textit{\textbf{500}} & \textit{\textbf{1019}}\\
\hline
PTB & 100.0 & 100.0 & 99.63 & 99.50 & 98.92 & 98.76 & 97.73 & - & - \\
UofTDB & 100.0 & 97.26 & 98.30 & 95.46 & 93.86 & 91.16 & 89.70 & 91.20 & 91.45 \\
\hline
\end{tabular}\\
\textsuperscript{1}For PTB, this column corresponds to the entire set of 290 subjects.
\end{table}

The results of the performance evaluation are presented in Table~\ref{table:performance}. These results roughly follow the expected patterns considering the use of on-the-person \emph{versus} off-the-person ECG data. The model is able to attain high true positive identification rates in both databases when the population is small, but as the set of subjects grows, performance decreases and a wide gap distinguishes the more challenging off-the-person settings from the more controlled on-the-person settings.

Additionally, one can find some unusual patterns in the performance results. Considering $M > N$, one would expect identification performance with subset $T_N$ to be higher than with subset $T_M$. With UofTDB off-the-person data this is not always verified: \emph{e.g.}, from $T_5$ to $T_{10}$, performance increases from $97.26\%$ to $98.10\%$. In these cases, we need to consider that datasets with fewer identities have fewer data and, thus, more unstable results. Alternatively, the identities added to $T_N$ to create $T_{M}$ may be easier to discriminate (``sheep'', according to the concept of biometric menagerie~\cite{Doddington1998, Yager2010}) and thus contribute to improving accuracy. However, one should also regard the substantial regularisation needed to avoid overfitting and the instability during training as possible causes for these discrepancies. This is a very important insight into the increased difficulties of using off-the-person data and the need for improved and more robust biometric models. 

\begin{figure}[p]
    \centering
    \includegraphics[width=\linewidth]{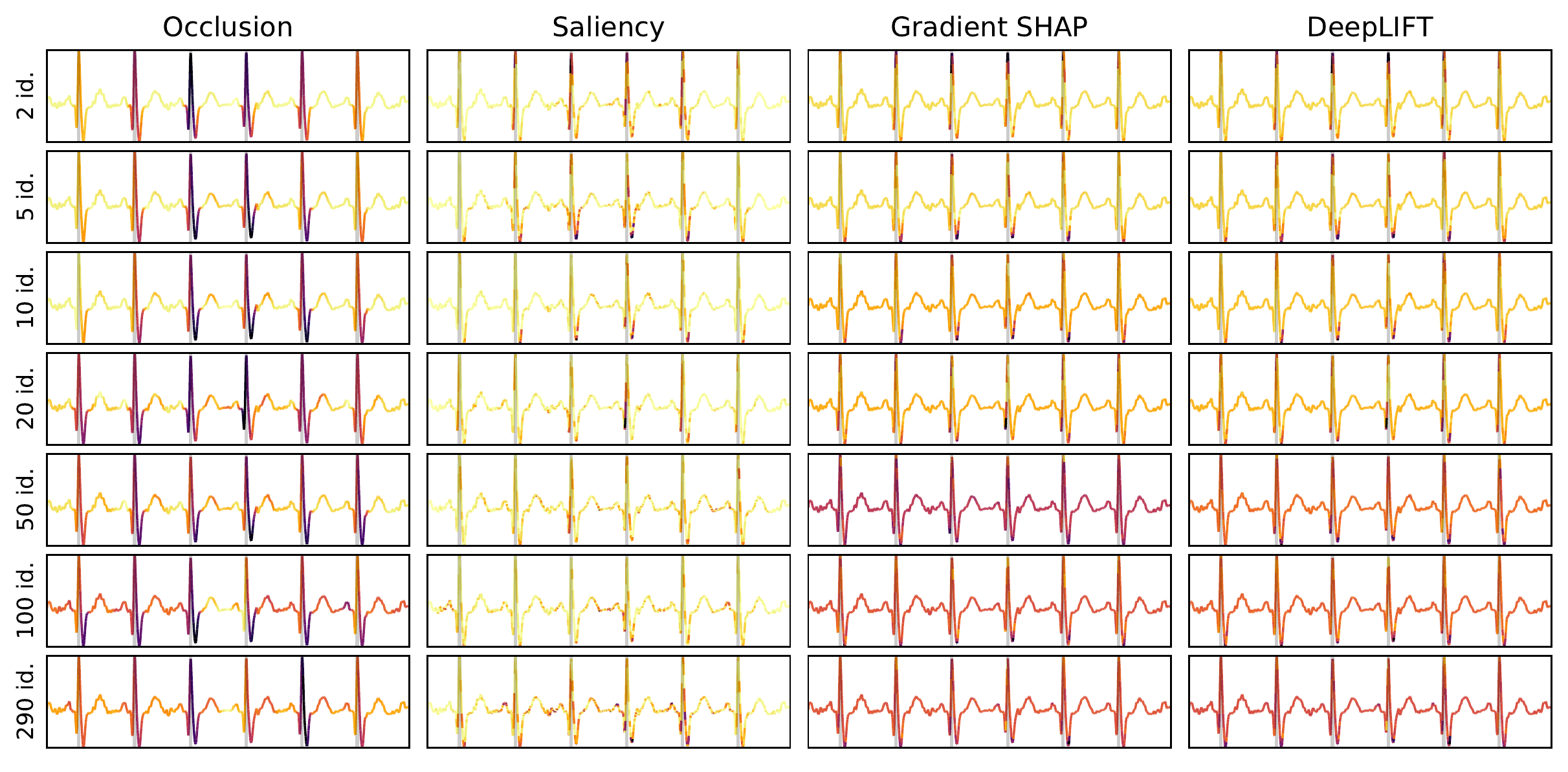}
    \caption[Explanations over an example five-second ECG segment from PTB.]{Explanations over an example five-second ECG segment from PTB (in each subplot, the yellow to dark purple colours correspond to increasing time sample relevance and vertical grey lines denote R-peak locations; signals were filtered for easier visualisation).}
    \label{fig:explainecg_ptb_segment}
\end{figure}

\begin{figure}[p]
    \centering
    \includegraphics[width=\linewidth]{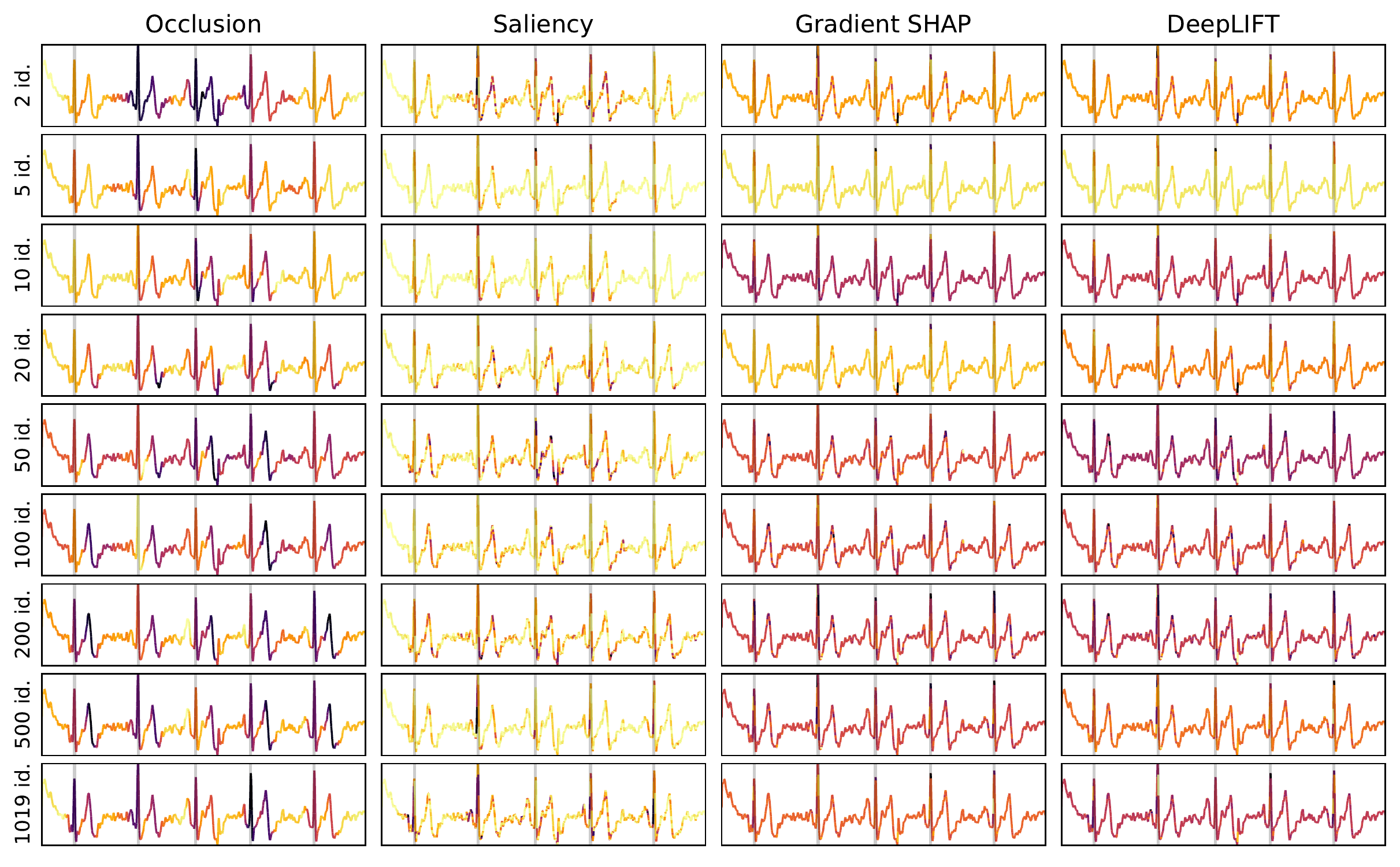}
    \caption[Explanations over an example five-second ECG segment from UofTDB.]{Explanations over an example five-second ECG segment from UofTDB (in each subplot, the yellow to dark purple colours correspond to increasing time sample relevance and vertical grey lines denote R-peak locations, signals were filtered for easier visualisation).}
    \label{fig:explainecg_uoftdb_segment}
\end{figure}

\begin{figure}[p]
    \centering
    \includegraphics[width=\linewidth]{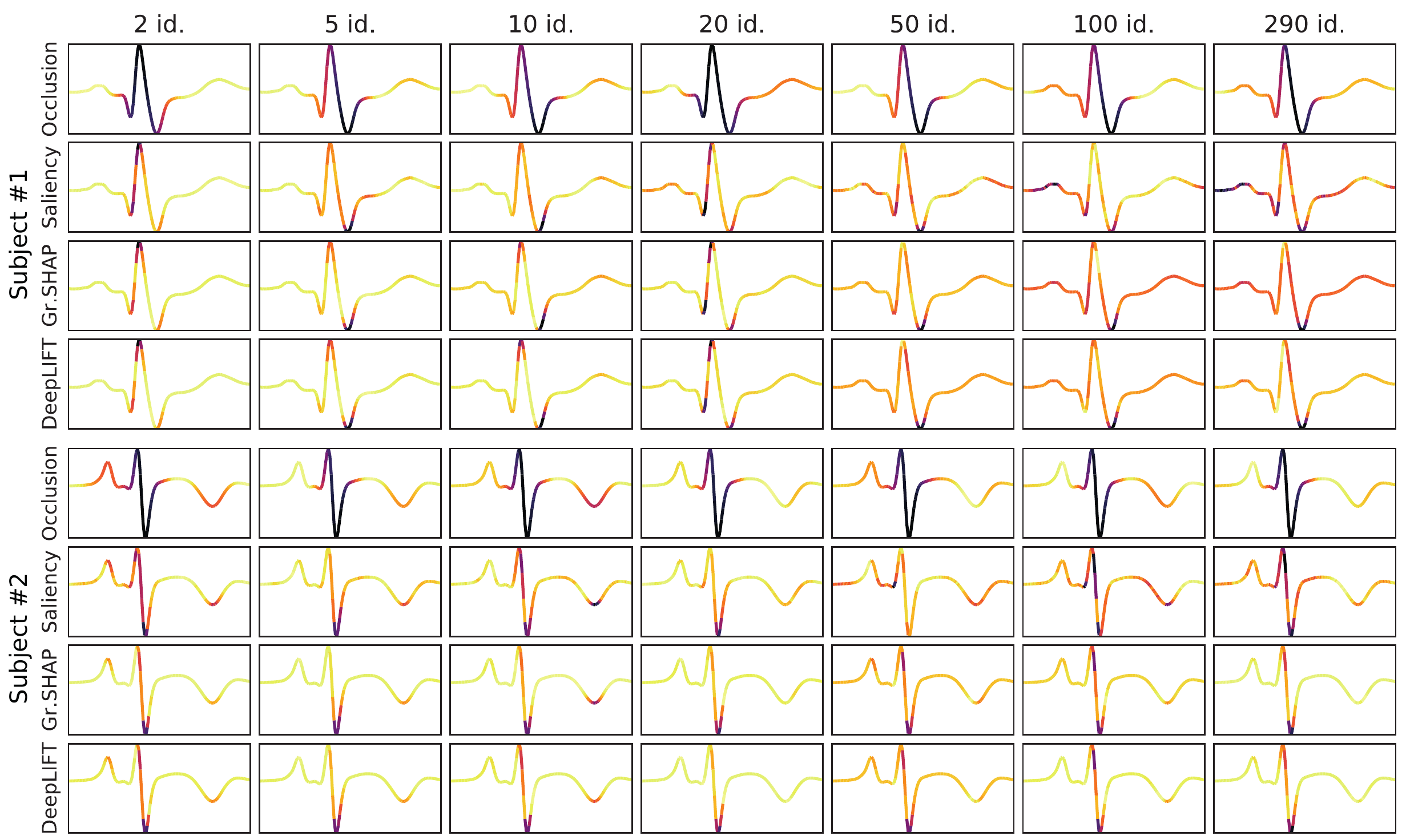}
    \caption[Average explanations over heartbeat waveforms of subjects \#1 and \#2 on the subsets of the PTB database.]{Average explanations over heartbeat waveforms of subjects \#1 and \#2 on the subsets of the PTB database (in each subplot, the yellow to dark purple colours correspond to increasing time sample relevance; signals were filtered for easier visualisation).}
    \label{fig:explainecg_ptb_heartbeats}
\end{figure}

\begin{figure}[p]
    \centering
    \includegraphics[width=\linewidth]{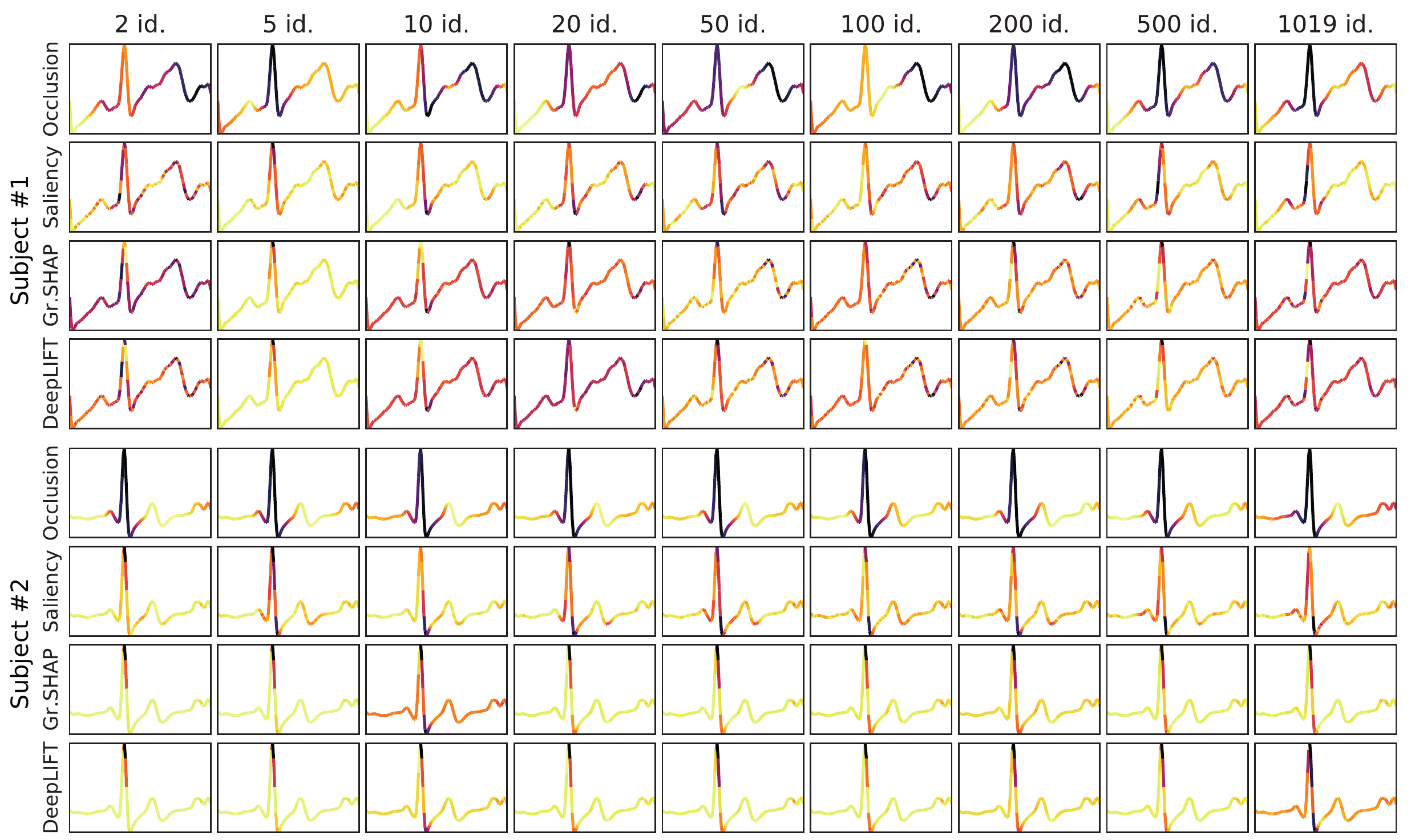}
    \caption[Average explanations over heartbeat waveforms of subjects \#1 and \#2 on the subsets of the UofTDB database.]{Average explanations over heartbeat waveforms of subjects \#1 and \#2 on the subsets of the UofTDB database (in each subplot, the yellow to dark purple colours correspond to increasing time sample relevance; signals were filtered for easier visualisation).}
    \label{fig:explainecg_uoftdb_heartbeats}
\end{figure}

Analysing the explanations obtained using the four interpretability tools (examples in Fig.~\ref{fig:explainecg_ptb_segment} and Fig.~\ref{fig:explainecg_uoftdb_segment}), a trend is verified from smaller to larger identity subsets, consisting on the deviation from focusing mainly on the QRS complex to the increasing relevance of other parts of the heartbeats. This is also confirmed when combining the explanations of all heartbeats of each person into a single average heartbeat (see Fig.~\ref{fig:explainecg_ptb_heartbeats} and Fig.~\ref{fig:explainecg_uoftdb_heartbeats}).

With the cleaner medical signals from PTB, the focus is mostly on the QRS complex, but information from other waveforms starts to become more and more relevant as more identities are added. It is noteworthy how, when discriminating PTB subjects \#1 and \#2 in a two-subject scenario (see Fig.~\ref{fig:explainecg_ptb_heartbeats}), the model still focuses mainly on the QRS, even though subject \#2 has a very specific characteristic, the inverted T-wave, that is arguably their most distinctive feature. This denotes how, in these cleaner signals, the QRS complex is so stable that the remaining waveforms, more susceptible to heart rate variability, are largely ignored by the model regardless of any visually obvious intersubject differences they may present.

With the more realistic off-the-person signals from UofTDB, the QRS retains high importance but the relevance is more evenly spread among the signal waveforms. In the specific case of subject \#2 (see Fig.~\ref{fig:explainecg_uoftdb_heartbeats}), it is evident that the QRS retains the highest importance for the decision, even in $T_{1019}$ (the largest subset). This may denote that, even in these more challenging settings, the identification models will still give preference to the QRS over other waveforms if it is sufficiently unique among the considered identities. Nevertheless, in such large sets of identities, the expected behaviour is that of subject \#1 (see Fig.~\ref{fig:explainecg_uoftdb_heartbeats}), since the limited identity information carried by the QRS will lead the model to also look to other parts of the signal.

One interesting aspect is the difference between the results with Occlusion \emph{versus} the other methods. Occlusion generally grants the QRS complex much more relevance, regardless of the settings. In the state-of-the-art approaches, the QRS complex is not only a source for identity features but also frequently used as an easily detectable reference landmark for the location of other ECG waveforms. This may also be the case in this end-to-end deep model. Although there are challenging contexts where the QRS may not be the main contributor to the decision, it may be essential to the deep model as a reference landmark to locate other waveforms in the signal. Hence, when occluded, it will be the signal component that most impacts the decision, causing the occlusion method to generally consider it the most relevant.

\section{Summary and Conclusions}\label{sec:explainecg_conclusion}

This work aimed to explain how deep models use ECG signals to distinguish people, using interpretability tools. Overall, the obtained results partially confirm the claim that the QRS is the key to ECG-based biometrics. With small populations in on-the-person settings, it can alone be used for reliable recognition. However, as we evolve towards larger populations and off-the-person settings, other components become relevant in discriminating people, as the models require more identity information to overcome the hurdles placed by enhanced intrasubject variability.

However, even though relevance is more evenly shared in off-the-person identification in large sets of identities, the QRS is shown as essential by the occlusion method. It appears that, just like several literature methods, the implemented end-to-end model learnt to use the QRS as a landmark for the location of other ECG components in the signal, resulting in large output changes when the QRS is occluded. Hence, despite the literature claims, one should avoid relying too heavily on any single part of the ECG, including the QRS complex, since all waveforms carry identity information that proves increasingly useful in more realistic settings and larger populations.

Beyond these insights, further efforts should be devoted to extending this study and offering a deeper, more thorough, and more objective analysis of the contribution of each ECG waveform to the model's decisions. Obtaining more systematic and complete explanations could create new opportunities for the use of interpretability tools during model training. Using explanations to regularise models and promote the focus on the most relevant signal components or the distributed use of the whole signal (instead of just the QRS) could lead to improved recognition accuracy and robustness.
\chapter{Interlead Conversion of Electrocardiographic Signals}\label{ch:ecginterlead}

\begin{tcolorbox}\footnotesize
{\large\bf Foreword on Author Contributions}

The research work described in this chapter was conducted in collaboration with Sofia C. Beco, under the supervision of Jaime S. Cardoso. The author of this thesis contributed to this work on the formulation, implementation, and improvement of the interlead conversion methodology, the preparation and conduction of the extended experiments, the discussion of the results, and the writing of the scientific publications.

The results of this work have been disseminated in the form of an extended journal article and a short paper presented at an international conference:
\begin{itemize}[noitemsep, leftmargin=1em, nosep]
    \item S. Beco, \underline{J. R. Pinto}, and J. S. Cardoso, ``Electrocardiogram Lead Conversion from Single-Lead Blindly-Segmented Signals,'' \emph{BMC Medical Informatics and Decision Making}, 22: 314, 2022.~\cite{Beco2022Electrocardiogram}
    \item S. Beco, \underline{J. R. Pinto}, and J. S. Cardoso, ``Interlead Conversion of Single-Lead Blindly-Segmented Electrocardiogram Signals,'' in \emph{17th International Conference on Computational Intelligence Methods for Bioinformatics and Biostatistics (CIBB 2021)}, Nov.~2021.~\cite{Beco2021Interlead}
\end{itemize}
\end{tcolorbox}

\section{Context and Motivation}

The electrocardiogram (ECG) is the measurement of electrical potentials that make the heart contract and relax as intended. The morphology of the ECG signal depends on the location of the electrodes used for acquisition: different electrode placement results in different perspectives over the heart~\cite{Pinto2018}. For medical purposes, the standard configuration acquires the ECG over twelve leads for more information, but it requires ten electrodes placed on the patient's arms, legs, and chest. Using fewer electrodes allows for more comfortable and inexpensive acquisitions, at the expense of certain leads that could be ideal for a more accurate diagnosis of certain conditions.

To get the best of both worlds, researchers have proposed methods for the automatic interlead conversion of ECG signals~\cite{Silva2020, Sohn2020, Lee2020, Matyschik2020, Smith2021}. These transform short ECG segments to mimic other perspectives, using acquired leads to reconstruct any leads that were not recorded. However, these methods still present limited applicability, since they typically require multiple leads as input. Even the most advanced methods \cite{Lee2020, Matyschik2020}, that only use one input lead, still require the inputs to be single heartbeat segments aligned in time, which makes them dependent on separate processes and, overall, less flexible and robust.

This chapter presents a study on the feasibility of ECG interlead conversion using short segments from just one limb lead without any kind of temporal alignment (blindly-segmented). With such input, the proposed methodology is trained to reconstruct other leads as faithfully as possible. This aims to open up new possibilities for more comfortable ECG acquisition in clinical scenarios or wearable devices without giving up the benefits of multi-lead recordings for medical diagnosis.

The proposed methodology, based on deep learning encoder-decoder structures, is explored for interlead conversion using either lead II or lead I (limb leads) signals as reference, and using a single shared encoder or an individual encoder for each target lead. Beyond the training and testing on the widely used PTB database, the conversion models are evaluated on cross-database scenarios with the INCART and PTB-XL databases. Additionally, the clinical annotations of the PTB-XL database are also used for a differential performance evaluation in the presence of medical conditions. The code is available online\footnote{Interlead ECG Conversion Github Repository. Available on: \url{https://github.com/jtrpinto/ecg-conversion}.}.

\section{Related Work}

At the onset of research on interlead conversion, methodologies commonly required several leads as reference for robust lead reconstruction. \citet{Zhu2004} performed a preliminary study on the conversion of ambulatory ECG recordings into standard 12-lead ECG signals using lead-field theory and the least-squares method. \citet{Nelwan2004} learned generic and patient-specific linear regression coefficient templates to reconstruct up to four missing leads with high correlation results.

Later, \citet{Yoshida2012} used 12 lead acquisitions to synthesise additional leads (right ventricular leads V3R, V4R, and V5R and posterior chest leads V7, V8, and V9) which provide important information for the diagnosis of acute myocardial infarction. Their algorithm was based on the transfer coefficient estimated from the learning data.

\citet{Silva2020} developed three methods for obtaining the Frank leads using the 12 standard leads as reference: the Kors Quasi-Orthogonal method, the Kors Linear Regression method, and the Dower Inverse Matrix. The conversion was successful for signals from healthy subjects but presented limitations on signals from subjects with pathologies. The recent work by \citet{Smith2021} was one of the first to use machine learning techniques for interlead conversion. They used a focused time-delay neural network (FTDNN), which is well suited for time series prediction. However, their methodology required seven input leads (all limb leads and V1).

\citet{Atoui2010} used ensembles of fully-connected neural networks to learn to synthesise V1, V3, V4, V5, and V6 heartbeats from three-lead inputs (I, II, and V2). \citet{Schreck2013} performed the first study on the synthesis of the entire set of 12 standard leads and scalar 3-lead derived vectorcardiogram from just three measured leads. Their proposed methodology used nonlinear optimisation to construct a universal patient transformation matrix. \citet{Hansen2015} applied linear generic and subject-specific transforms to convert recordings from adhesive patch-type ECG monitors to the standard 12-lead ECG signals. In~\cite{Trobec2011,Tomasic2013}, researchers also explored personalised statistically determined linear transforms and went on to achieve improved results.

\citet{Lee2017}~proposed methods based on linear regression and artificial neural networks to reconstruct the 12 standard leads from subsets of 35 channels acquired using one single large patch covering the subject's chest. Although accurate, the method is arguably incompatible with scenarios focused on ease of use and patient/user comfort. Similarly, \citet{GrandeFidalgo2021} used linear regression and fully-connected networks to reconstruct the entire set of twelve standard leads from a subset of just three input leads. \citet{Sohn2020} used long short-term memory (LSTM) networks to reconstruct the twelve ECG standard leads from a three-lead patch-type device. Their results show their method was able to correctly retain pathological abnormalities from medical conditions on the reconstructed signals. 

The work of \citet{Lee2020} was one of the few that studied the synthesis of standard leads using only one reference lead. In their study, chest leads (V1 to V6) were synthesised from lead II using a generative adversarial network (GAN). However, input segments had to be single heartbeats, aligned according to the R-peaks, which decreases the difficulty of the proposed method but also its applicability. \citet{Matyschik2020} developed patient-specific models to more accurately reconstruct eleven missing ECG signals from a single available lead of the standard 12-lead system. However, the reference lead was either V1, V2, or V3 which, being chest leads, do not enable the usage in less obtrusive setups which would preferentially use limb leads.

In this work, we explore the more challenging scenario of reconstructing the entire set of twelve standard leads using only one reference lead. Moreover, the reference signals are blindly-segmented (without any kind of temporal alignment) and pertain to one of the limb leads to allow for applications on the least obtrusive setups. Our main goal is to assess whether it is possible to reconstruct the electrocardiogram signal in such challenging scenarios and discuss the next steps towards the use of interlead conversion in less obtrusive clinical setups and wearable devices.

\section{Methodology}

\subsection{General overview}

The proposed methodology for interlead ECG conversion follows the encoder-decoder structure typically used for deep image segmentation. The encoder receives an input signal and processes it to create a compressed representation that retains relevant information for the task at hand. The decoder receives this representation and processes it so that the output matches the ground-truth as closely as possible. Here, the input to the encoder is a short ECG segment of one lead (X) and the ground-truth is the corresponding segment in a different lead (Y). Thus, the encoder is in charge of selecting the information from X that is needed for Y, and the decoder will use that information to reconstruct the corresponding lead Y signal.

\subsection{Model architectures}

The general encoder-decoder structure allows for diverse specific model architectures. This work focuses on the U-Net model, a fully convolutional architecture that has found many applications related to semantic segmentation and can also be adapted for the task of ECG lead conversion.

\subsubsection{U-Net}

\begin{figure}[t!]
    \centering
    \includegraphics[width=\linewidth]{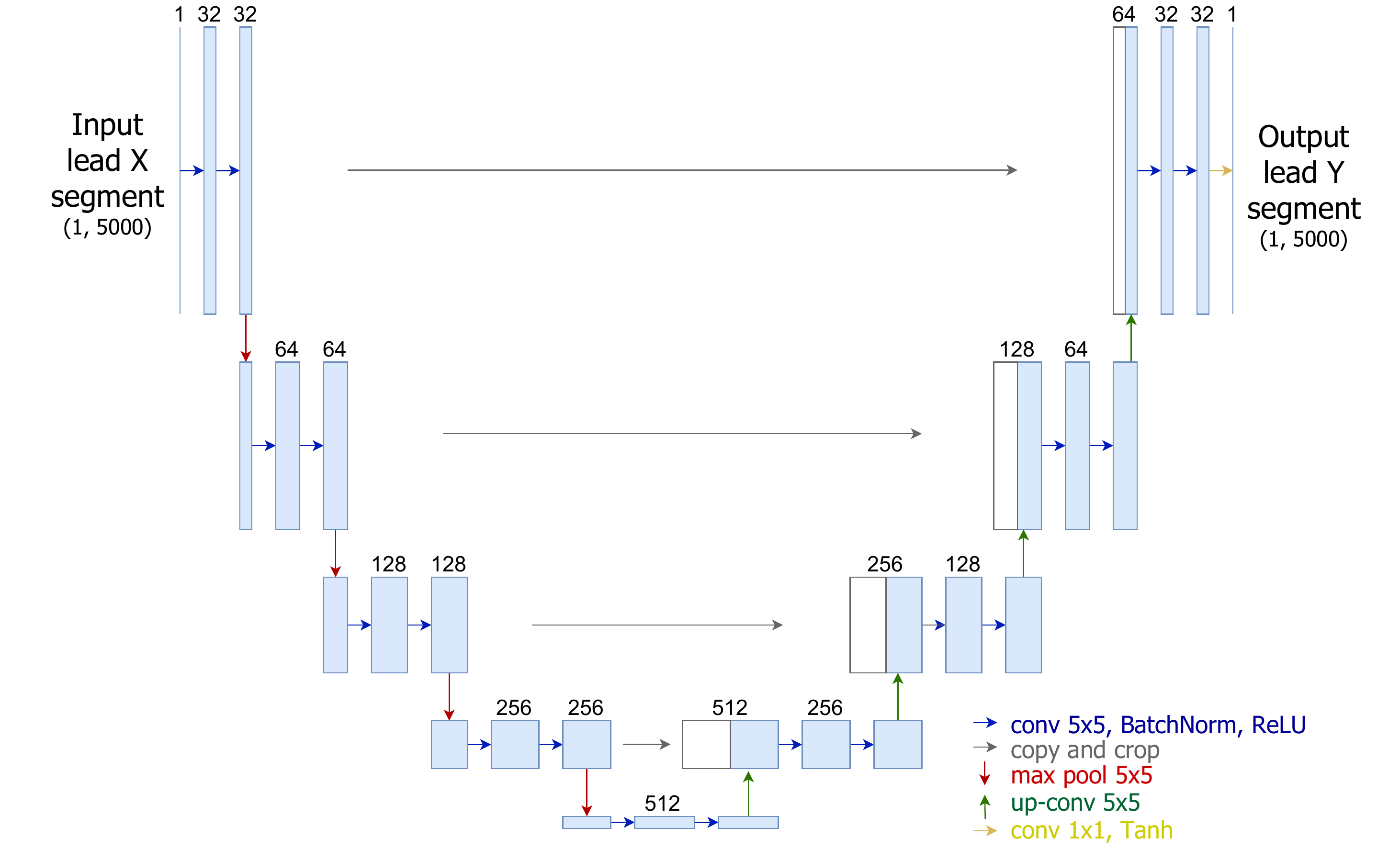}
    \caption{Schema of the U-Net architecture.}
    \label{fig:interleadconv_struc_unet}
\end{figure}

The U-Net was initially proposed by \citet{ronneberger2015unet} as a tool for biomedical image segmentation. In this work, the implemented architecture (see Fig.~\ref{fig:interleadconv_struc_unet}) receives an input segment of lead X, which initially goes through a chain of three sequential blocks, each with half the signal resolution of the previous block. Each block includes two convolutional layers (each followed by batch normalisation and ReLU activation) and ends with a max-pooling layer. 

Between the encoder and the decoder, two convolutional layers compose the latent space or bottleneck block, which corresponds to the maximum point of information compression. The decoder mirrors the encoder in its structure, with three similar blocks composed of an upsampling layer and two transposed convolutional layers. The last transposed convolutional layer outputs a single-channel signal whose size corresponds to the input segment. The activation function of this last layer is the hyperbolic tangent for an output signal with amplitudes in $[-1, 1]$.

One aspect of the U-Net which is often cited as the key to its widespread success is the skip-connection. U-Nets typically include skip-connections between corresponding blocks on the encoder and the decoder. This means the feature maps from the encoder blocks are directly routed to the corresponding decoder blocks, allowing the model to propagate context information from multiple resolutions between the encoder and the decoder for higher flexibility.

\subsubsection{Convolutional autoencoder (AE)}

\begin{figure}[t!]
    \centering
    \includegraphics[width=\linewidth]{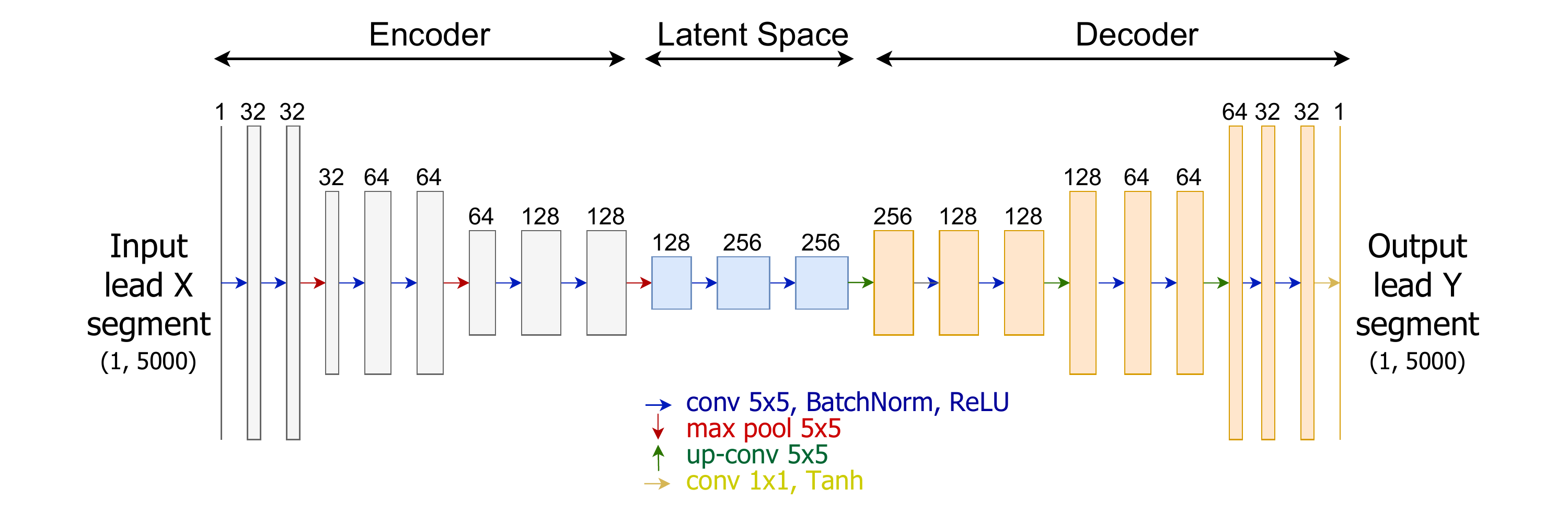}
    \caption{Schema of the convolutional autoencoder (AE) architecture.}
    \label{fig:interleadconv_struc_ae}
\end{figure}

Beyond the aforementioned U-Net architecture, adapted for unidimensional signal inputs, we also explore a convolutional autoencoder (AE, see Fig.~\ref{fig:interleadconv_struc_ae}). Its architecture is very similar to the U-Net, albeit without skip-connections. As a result, the structure is simplified, when compared to the U-Net, and the latent representation sent from the encoder to the decoder is smaller. Experiments with the AE architecture aim to assess if the skip-connections are essential for the task at hand or if the simplified structure could avoid overfitting and bring performance benefits.

\subsubsection{Label refinement network (LRN)}

\begin{figure}[t!]
    \centering
    \includegraphics[width=\linewidth]{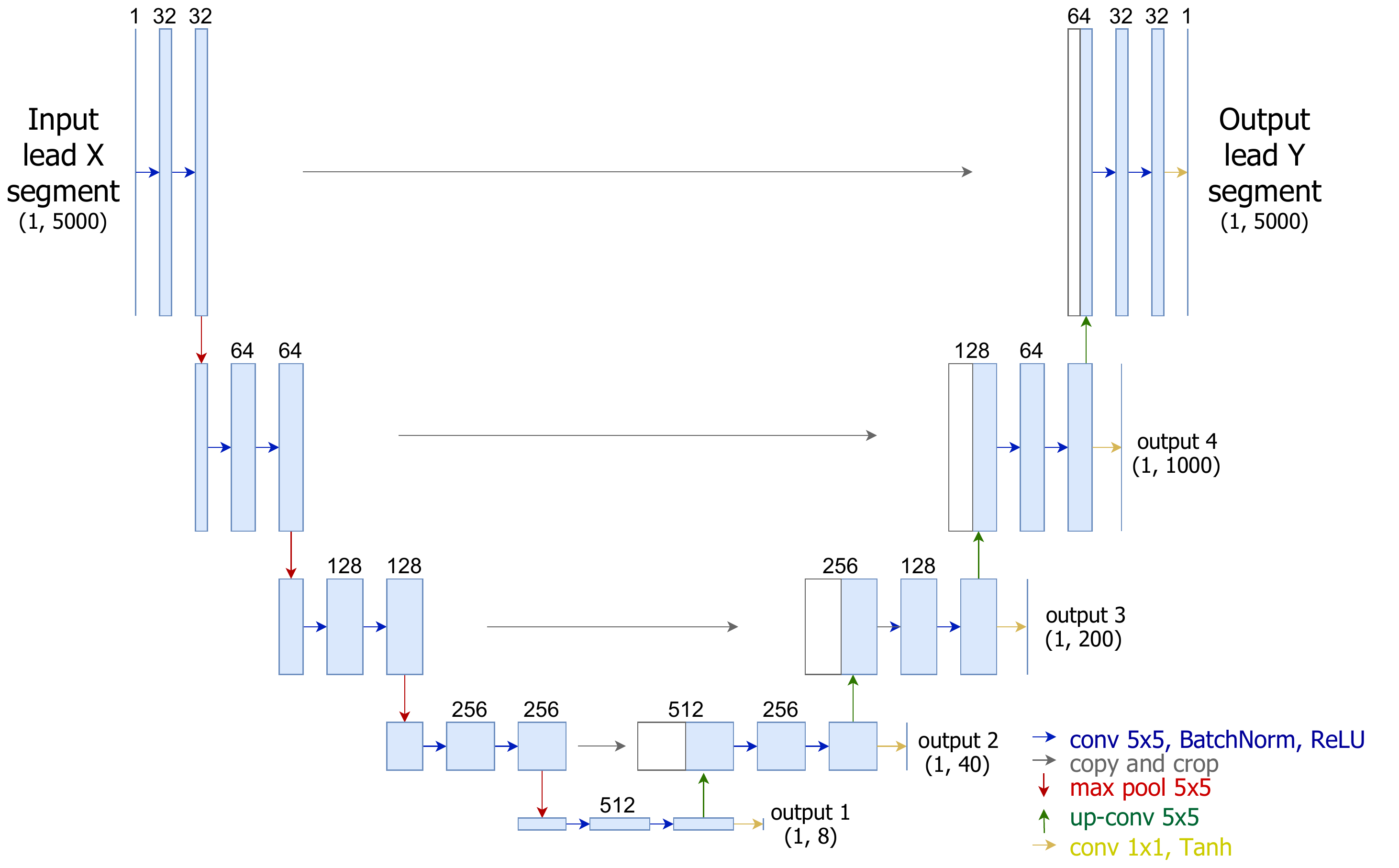}
    \caption{Schema of the architecture based on label refinement networks (LRN).}
    \label{fig:interleadconv_struc_lrn}
\end{figure}

The third architecture explored in this work was based on Label Refinement Network (LRN, see Fig.~\ref{fig:interleadconv_struc_lrn}) was originally proposed by \citet{Islam2017lrn} for semantic image segmentation. Its architecture is identical to the aforementioned U-Net. The singularity of the LRN lies in the supervision strategy: while the U-Net only uses the output of the last decoder block in the reconstruction loss, the LRN computes the loss at the outputs of every decoder block. This results in supervision at several resolution levels, leading the decoder to offer a coarse reconstruction right after the first block, which should be gradually refined by the subsequent blocks for improved results at higher resolutions. Experiments with the LRN architecture aim to assess if the multi-level resolution could bring improved performance to the task of signal lead conversion as they have for semantic segmentation.

\subsection{Shared \emph{vs.} individual encoders}

The conversion of one lead into multiple missing leads requires multiple decoders - each one will fulfil the task of reconstructing their respective lead based on the compressed latent representation. In the case of the encoder, however, it is possible to have a single one whose output will be shared by all decoders or have multiple encoders, each one dedicated to one individual decoder.

In this work, we explore both possibilities for 12-lead reconstruction - using one shared encoder connected to all 11 decoders, for all 11 output leads except the one corresponding to the input, or using one individual encoder for each of the 11 decoders. Using individual encoders grants more flexibility to each lead conversion process, as each encoder will be able to learn a unique way to obtain compressed representations and each encoder-decoder pair will work independently from all others. On the other hand, using one shared encoder results in a much lighter and faster algorithm and the added simplicity may contribute to avoiding overfitting.

\section{Experimental Setup}

\subsection{Data}

The experiments conducted in this work used mainly the data provided in the PTB Diagnostic ECG Database~\cite{Bousseljot1995}, available on Physionet~\cite{Goldberger2000}. The PTB database includes data from 16 channels, including all 12 standard leads, sampled at $1$ kHz. It contains a total of $549$ records from $290$ individuals, with one to five records per subject. Recordings were cropped into segments of $5$ s ($5000$ samples). A second-order Butterworth bandpass filter with cut-off frequencies $f_c = [1, 40]\ Hz $ was applied to each segment to remove noise while retaining the most useful ECG information. The amplitudes of the $n$ values of each signal $x$ were then min-max normalised to the interval $[-1, 1]$ following the equation:
\begin{equation}
    x_n = 2\times\frac{x_n-x_{min}}{x_{max} - x_{min}} - 1.
\end{equation}

The data from PTB was divided into train and test sets, with approximately $63\%$, $7\%$ and $30\%$ of the segments, respectively, for a total of $7086$, $787$, and $3509$ ECG segments for each set. For a more thorough and challenging evaluation, subjects are divided between the train/validation and test sets: the latter had recordings from subjects $1$ to $50$ while the former had recordings from subjects $51$ to $290$.

The INCART database (officially the St. Petersburg INCART 12-lead Arrhythmia Database), also available on Physionet, was used to test the performance of trained models on cross-database scenarios. This database contains $75$ Holter recordings from $32$ subjects undergoing tests for coronary artery diseases. Each record is $30$ minutes long and contains twelve standard leads sampled at $257$ Hz. Recordings from this database were resampled to $1$ kHz and processed as described above for PTB.

The PTB-XL database~\cite{ptbxl1, ptbxl2}, created by the same team as the PTB, includes $21837$ clinical ECG recordings from a total of $18885$ patients. Each recording is 10 seconds long, includes all twelve standard ECG leads, and is originally sampled at $500$ Hz. The waveforms were annotated by up to two cardiologists, who assigned annotations to each record. The 71 possible annotation statements have been clustered into five superclasses: NORM (normal ECG), MI (myocardial infarction), STTC (ST/T change), CD (conduction disturbance), and HYP (hypertrophy). This dataset was originally created for the training and evaluation of automatic ECG interpretation algorithms but also shows great promise for the development of lead conversion algorithms. In this work, we take advantage of expert clinical annotations to study the effect of medical conditions on the quality of the lead conversion results. From the total of $21837$ recordings, we selected the $16272$ that did not have conflicting superclass annotations. From each recording, the first $5$ seconds were cropped, resampled to $1$ kHz, and processed as described above for PTB.

\subsection{Model training and evaluation}

The models were trained using the $l1$-loss between the model outputs and the corresponding ground-truth signals as the objective function. The $l1$ was chosen empirically as it allowed the model to learn most adequately both the overall morphology of the signals and their finer details. The Adam optimiser was used with an initial learning rate of $1\times 10^{-3}$, over a maximum of $500$ epochs with batch size $32$ (shared encoder) or $16$ (individual encoder) and early stopping patience of $50$ epochs.

To compare lead conversions with the corresponding measured ground-truth signals, this work used the following metrics: the average and median Pearson correlation coefficient ($r$, used in the majority of the related literature), the average root mean square error (RMSE), and the average Structural Similarity Index Measure (SSIM).

\section{Results and Discussion}

\subsection{Architecture comparison}

To compare the selected architectures, the first experiment entailed the one-to-one lead conversion from II to I, two of the most used ECG leads for medical purposes (see Table~\ref{tab:interleadconv_conv_NN}). According to the results, the U-Net performs better than both alternatives AE and LRN. Although the AE achieves the same median $r$ as the U-Net, the average $r$ is lower, meaning that the least successful results are generally worse with the AE than with the U-Net.

\begin{table}[!t]
\caption{Comparison of encoder-decoder architectures on one-to-one lead conversion.}\label{tab:interleadconv_conv_NN}
\centering
\begin{tabular}{lcccccc}
\hline
\textbf{Model} & \textbf{$r$ (avg.)} & \textbf{$r$ (med.)}\\\hline 
U-Net       & 0.69 & 0.78 \\
Autoencoder & 0.67 & 0.78 \\
LRN         & 0.65 & 0.75 \\\hline
\end{tabular}
\end{table}

\begin{table}[!t]
\caption{Average correlation between lead II signals and the remaining leads on the PTB, INCART, and PTB-XL databases.}
\centering
\begin{tabular}{cccccccccccc}
    \hline
    & \multicolumn{11}{c}{\textbf{Average Correlation to Lead II}}\\
    \textbf{} & \textbf{\textit{I}} & \textbf{\textit{III}} & \textbf{\textit{aVR}} & \textbf{\textit{aVL}} & \textbf{\textit{aVF}} & \textbf{\textit{V1}} & \textbf{\textit{V2}} & \textbf{\textit{V3}} & \textbf{\textit{V4}} & \textbf{\textit{V5}} & \textbf{\textit{V6}} \\ \hline
    PTB & 0.45 & 0.36 & -0.71 & 0.01 & 0.77 & -0.34 & -0.20 & 0.00 & 0.28 & 0.72 & 0.81 \\
    INCART & 0.46 & 0.80 & -0.86 & -0.49 & 0.95 & -0.45 & -0.19 & 0.25 & 0.65 & 0.82 & 0.77 \\
    PTB-XL & 0.70 & 0.31 & 0.25 & -0.82 & 0.83 & -0.44 & -0.04 & 0.37 & 0.68 & 0.81 & 0.84 \\\hline
    \end{tabular}\label{tab:interleadconv_corr_between_leads_ii}
\end{table}

\begin{table}[!t]
\caption{Test results of the U-Net used for multi-lead conversion from lead II, with shared or individual encoders.}
\centering
\begin{tabular}{ccccccccccc}
    \hline
    && \multicolumn{4}{c}{\textbf{Shared Encoder}} && \multicolumn{4}{c}{\textbf{Individual Encoders}} \\
    \textbf{Lead} && \textbf{$r$ (avg.)} & \textbf{$r$ (med.)} & \textbf{\textit{RMSE}} & \textbf{\textit{SSIM}} && \textbf{$r$ (avg.)} & \textbf{$r$ (med.)} & \textbf{\textit{RMSE}} & \textbf{\textit{SSIM}} \\ \hline
    I    && 0.67    & 0.73    & 0.28 & 0.28 && 0.66    & 0.71    & 0.29 & 0.26 \\
    III  && 0.56    & 0.65    & 0.29 & 0.63 && 0.56    & 0.70    & 0.29 & 0.64 \\
    aVR  && 0.89    & 0.95    & 0.12 & 0.92 && 0.90    & 0.95    & 0.12 & 0.92 \\
    aVL  && 0.47    & 0.58    & 0.36 & 0.15 && 0.47    & 0.61    & 0.36 & 0.16 \\
    aVF  && 0.81    & 0.88    & 0.20 & 0.64 && 0.83    & 0.90    & 0.19 & 0.64 \\
    V1   && 0.77    & 0.84    & 0.20 & 0.87 && 0.80    & 0.86    & 0.18 & 0.88 \\
    V2   && 0.66    & 0.72    & 0.26 & 0.80 && 0.67    & 0.75    & 0.24 & 0.81 \\
    V3   && 0.56    & 0.62    & 0.33 & 0.65 && 0.59    & 0.66    & 0.31 & 0.66 \\
    V4   && 0.50    & 0.57    & 0.36 & 0.43 && 0.50    & 0.58    & 0.36 & 0.44 \\
    V5   && 0.70    & 0.77    & 0.27 & 0.36 && 0.74    & 0.80    & 0.26 & 0.40 \\
    V6   && 0.79    & 0.87    & 0.21 & 0.49 && 0.80    & 0.87    & 0.21 & 0.49 \\\hline
\end{tabular}\label{tab:interleadconv_one2many_ii}
\end{table}

The skip-connections give it the capability to send more information (and at more resolution levels) from the encoder to the decoders, granting it more flexibility and ultimately better performance than the AE. The multi-resolution supervision of the LRN, expected to improve overall performance, appears to excessively draw the model's attention away from the details, which results in worse performance. Following the results of this comparison, subsequent experiments focus solely on the U-Net architecture.

\subsection{One-to-all leads conversion}

Not all leads can be converted equally: the correlation between leads depends on their perspectives of the heart. Table~\ref{tab:interleadconv_corr_between_leads_ii} presents an overview of the average correlation between lead II and the remaining eleven standard leads, computed using the PTB, INCART, and PTB-XL test segments. Specifically for the PTB data, one can observe that some leads such as aVF or aVR are highly (positively or negatively) correlated with lead II. On the other hand, aVL is almost orthogonal. Hence, one should expect aVL to be much harder to accurately convert from lead II than aVF or aVR, since the former shares much less information with lead II than the latter.

\begin{table}[!t]
\caption{Average correlation between lead I signals and the remaining leads on the PTB, INCART, and PTB-XL databases.}
\centering
\begin{tabular}{cccccccccccc}
    \hline
    & \multicolumn{11}{c}{\textbf{Average Correlation to Lead I}}\\
    \textbf{} & \textbf{\textit{II}} & \textbf{\textit{III}} & \textbf{\textit{aVR}} & \textbf{\textit{aVL}} & \textbf{\textit{aVF}} & \textbf{\textit{V1}} & \textbf{\textit{V2}} & \textbf{\textit{V3}} & \textbf{\textit{V4}} & \textbf{\textit{V5}} & \textbf{\textit{V6}} \\ \hline
    PTB & 0.45 & -0.49 & -0.82 & 0.82 & -0.05 & -0.47 & -0.21 & 0.03 & 0.30 & 0.64 & 0.68 \\
    INCART & 0.46 & 0.02 & -0.62 & 0.32 & 0.26 & -0.36 & 0.01 & 0.11 & 0.35 & 0.51 & 0.44 \\
    PTB-XL & 0.70 & -0.24 & 0.77 & -0.86 & 0.32 & -0.54 & -0.06 & 0.33 &  0.63 & 0.80 & 0.83 \\\hline
    \end{tabular}\label{tab:interleadconv_corr_between_leads_i}
\end{table}

\begin{table}[!t]
\caption{Test results of the U-Net used for multi-lead conversion from lead I, with shared or individual encoders.}
\centering
\begin{tabular}{ccccccccccc}
    \hline
    && \multicolumn{4}{c}{\textbf{Shared Encoder}} && \multicolumn{4}{c}{\textbf{Individual Encoders}} \\
    \textbf{Lead} && \textbf{$r$ (avg.)} & \textbf{$r$ (med.)} & \textbf{\textit{RMSE}} & \textbf{\textit{SSIM}} && \textbf{$r$ (avg.)} & \textbf{$r$ (med.)} & \textbf{\textit{RMSE}} & \textbf{\textit{SSIM}} \\ \hline
    II   && 0.49    & 0.54    & 0.37 & 0.17 && 0.50    & 0.55    & 0.36 & 0.19 \\
    III  && 0.44    & 0.49    & 0.35 & 0.53 && 0.46    & 0.52    & 0.35 & 0.55 \\
    aVR  && 0.89    & 0.92    & 0.14 & 0.92 && 0.90    & 0.93    & 0.13 & 0.93 \\
    aVL  && 0.76    & 0.84    & 0.25 & 0.45 && 0.77    & 0.85    & 0.26 & 0.46 \\
    aVF  && 0.26    & 0.29    & 0.43 & 0.28 && 0.28    & 0.32    & 0.42 & 0.27 \\
    V1   && 0.81    & 0.88    & 0.18 & 0.88 && 0.79    & 0.88    & 0.19 & 0.88 \\
    V2   && 0.73    & 0.80    & 0.23 & 0.81 && 0.70    & 0.77    & 0.25 & 0.80 \\
    V3   && 0.67    & 0.73    & 0.28 & 0.70 && 0.67    & 0.74    & 0.29 & 0.68 \\
    V4   && 0.59    & 0.65    & 0.33 & 0.48 && 0.62    & 0.71    & 0.32 & 0.48 \\
    V5   && 0.62    & 0.73    & 0.30 & 0.31 && 0.64    & 0.73    & 0.30 & 0.28 \\
    V6   && 0.66    & 0.75    & 0.26 & 0.39 && 0.67    & 0.77    & 0.26 & 0.39 \\\hline
\end{tabular}\label{tab:interleadconv_one2many_i}
\end{table}

\begin{figure}[p!]
    \centering
    \includegraphics[width=0.91\linewidth]{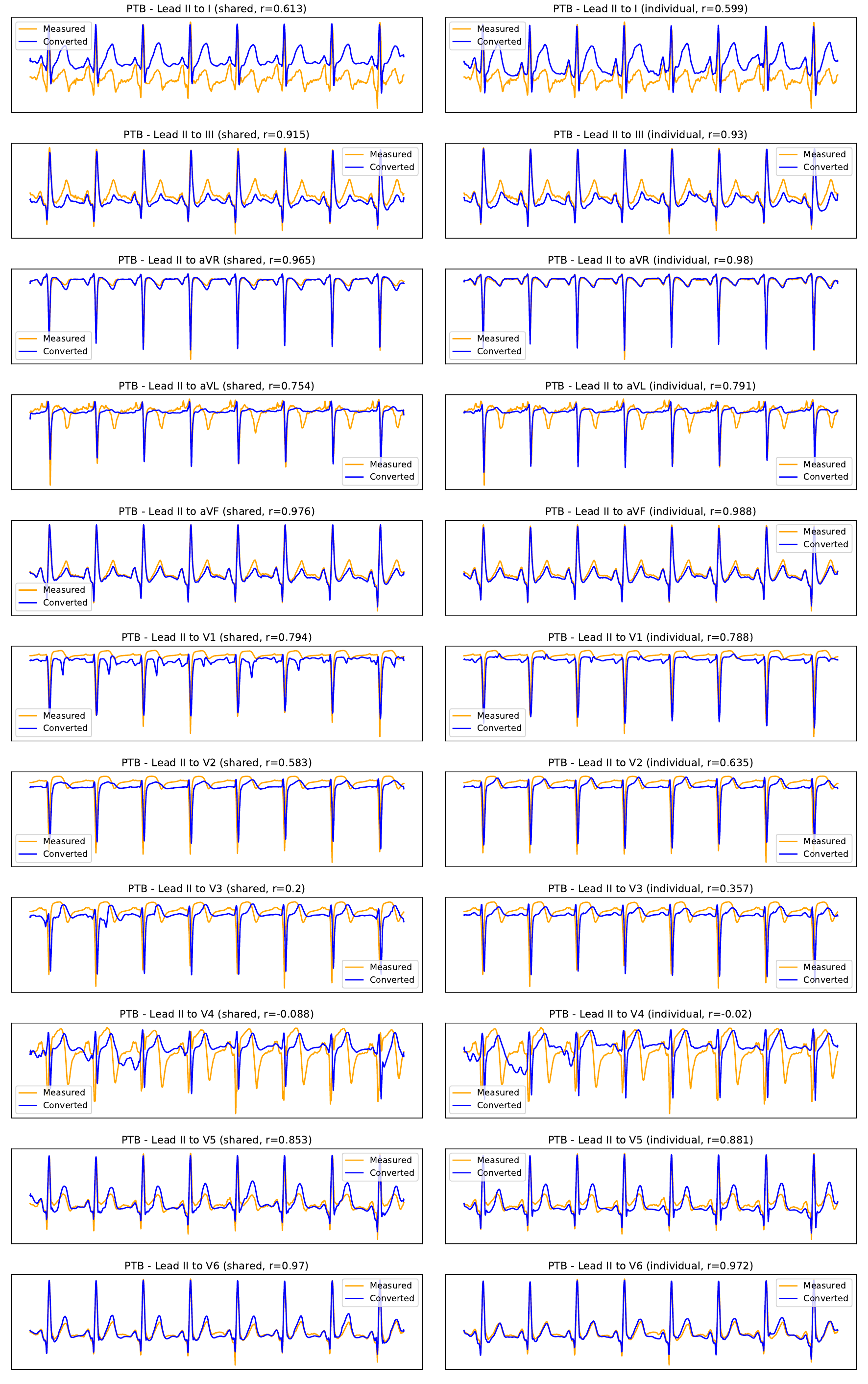}
    \caption[Example result of lead II to all conversion on the PTB test dataset.]{Example result of lead II to all conversion on the PTB test dataset (each row depicts one converted lead; shared encoder on the left; individual encoders on the right).}
    \label{fig:interleadconv_results_ii_to_all_ptb}
\end{figure}

This is verified in the results for multi-lead conversion on the PTB database (see Table~\ref{tab:interleadconv_one2many_ii}). Conversion from lead II to aVF, aVR, and V6 consistently offer good results, while the conversions to aVL, lead I, or V4 were overall the least successful. This behaviour is also visible in the example of Fig.~\ref{fig:interleadconv_results_ii_to_all_ptb}\footnote{Examples were selected among all test samples to correspond to the median overall $r$ result for each scenario. Hence, they represent a median result and the methodology should offer better results in half of the occasions.} where the model is unable to capture the finer details of the signals in lead aVL and leads V1-V4. The opposite happens in lead III, aVF, V6, and especially aVR, where the model was consistently able to capture the morphological details of the signals.

\begin{figure}[p!]
\centering
    \includegraphics[width=0.91\linewidth]{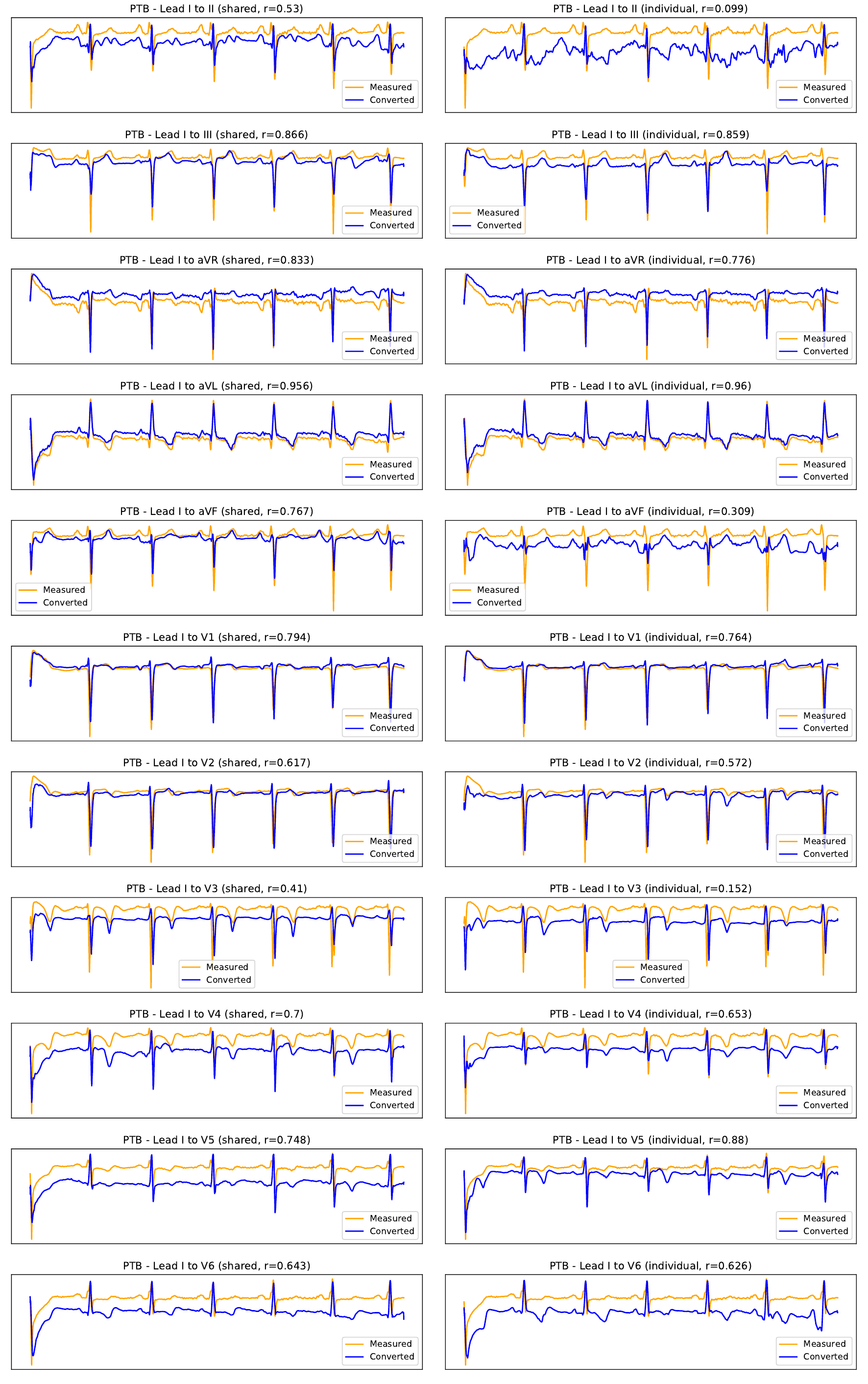}
    \caption[Example result of lead I to all conversion on the PTB test dataset.]{Example result of lead I to all conversion on the PTB test dataset (each row depicts one converted lead; shared encoder on the left; individual encoders on the right).}
    \label{fig:interleadconv_results_i_to_all_ptb}
\end{figure}

While lead II ECG signals are generally better for medical diagnosis in clinical scenarios, lead I is becoming increasingly important. The widespread implementation of ECG acquisition equipment in smartwatches, fitness bands, and other gadgets for daily use allows for the collection of lead I signals. Combining these growing applications with robust conversion algorithms would enable the recovery of missing leads on wearables and empower the next generation of robust continuous health monitoring.

Table~\ref{tab:interleadconv_corr_between_leads_i} presents the average correlation between lead I and the remaining eleven standard leads on the PTB, INCART, and PTB-XL test segments. Like lead II, lead I is more correlated (positively or negatively) with certain leads, such as aVR, aVL, or V6, while it is almost orthogonal with aVF or V3. As such, one can observe, in Table~\ref{tab:interleadconv_one2many_i}, that the proposed methodology obtains better performance with aVR and aVL while struggling to convert from lead I to lead aVF. The same can be observed in Fig.~\ref{fig:interleadconv_results_i_to_all_ptb}: for aVR and aVL, the model is able to correctly capture the target morphology, while the reconstructions of aVF and V3-V6 are largely unsuccessful.

From the example result in Fig.~\ref{fig:interleadconv_results_i_to_all_ptb}, one can also identify a shortcoming of the proposed methodology: the occasional offsets between the baseline of the measured and converted signals. We suspect this is due to the min-max normalisation of the signals, drawing them into the $[-1, 1]$ amplitude range. Alternatives to this normalisation, such as standard normalisation, should be further investigated.

Considering the overall results, no lead is perfect for converting all twelve standard leads. Hence, lead II should be chosen as reference input when aVF or V5-V6 are the most important leads for the application at hand. Lead I serves better as a reference when aVR, aVL, or V1-V2 are more important. Otherwise, other leads (such as lead III) should probably be explored. Nevertheless, the results show it is possible to nicely reconstruct several leads using only one input lead without temporal alignment.

Using either lead as a reference, there is apparently no considerable or consistent difference between using one single shared encoder or using an individual encoder for each target lead. It appears as if the additional flexibility of having multiple encoders is only beneficial up to a point, and the higher complexity ends up opening the door to overfitting and loss of robustness. As such, for this application, one should expect a shared encoder to be the best option, considering its higher simplicity and similar performance.

\subsection{Comparison with the state-of-the-art}

For a comparison with the state-of-the-art, we implemented the method recently proposed by Grande-Fidalgo~\etal~\cite{GrandeFidalgo2021} as a baseline. This method is based on a simple fully-connected model that receives each signal point's amplitude in three reference leads as inputs and returns the same point's amplitude in all twelve leads. Here, we adapt the methodology so it receives signal point amplitudes from one single lead (leads I or II), to exactly match the evaluation conditions of the proposed method.

Unlike what has been reported in~\cite{GrandeFidalgo2021}, the baseline was not successful in learning to retrieve the entire set of leads from just one reference lead. In fact, across all leads, the average test $r$ of this method ranged from $-0.005$ to $0.002$, considerably worse than the proposed methodology. One can assume that, although such a simplistic model presents advantages in terms of lightweight operation and robustness to overfitting, single-lead information is not enough for it to achieve reliable interlead conversion.

The fact the baseline method reconstructs signals point-by-point, unable to analyse broader local context information, makes it hard to reconstruct the signal without already having data from more than one channel. On the other hand, using convolutional layers allows the proposed method to use broader local information as context to adequately learn to reconstruct signals using only one lead as reference.

\begin{table}[!t]
\caption{Cross-database test results for INCART conversion from lead II.}
\centering
\begin{tabular}{ccccccccccc}
    \hline
    && \multicolumn{4}{c}{\textbf{Shared Encoder}} && \multicolumn{4}{c}{\textbf{Individual Encoders}} \\
    \textbf{Lead} && \textbf{$r$ (avg.)} & \textbf{$r$ (med.)} & \textbf{\textit{RMSE}} & \textbf{\textit{SSIM}} && \textbf{$r$ (avg.)} & \textbf{$r$ (med.)} & \textbf{\textit{RMSE}} & \textbf{\textit{SSIM}} \\ \hline
    I    && 0.46    & 0.51    & 0.34 & 0.18 && 0.44    & 0.50    & 0.35 & 0.16 \\
    III  && 0.57    & 0.63    & 0.28 & 0.49 && 0.57    & 0.63    & 0.29 & 0.45 \\
    aVR  && 0.91    & 0.95    & 0.11 & 0.92 && 0.92    & 0.95    & 0.11 & 0.92 \\
    aVL  && 0.13    & 0.11    & 0.41 & 0.33 && 0.10    & 0.09    & 0.44 & 0.27 \\
    aVF  && 0.88    & 0.93    & 0.15 & 0.68 && 0.92    & 0.95    & 0.13 & 0.71 \\
    V1   && 0.63    & 0.79    & 0.23 & 0.82 && 0.65    & 0.82    & 0.23 & 0.83 \\
    V2   && 0.53    & 0.64    & 0.26 & 0.73 && 0.55    & 0.69    & 0.27 & 0.74 \\
    V3   && 0.42    & 0.51    & 0.33 & 0.56 && 0.42    & 0.53    & 0.34 & 0.55 \\
    V4   && 0.52    & 0.59    & 0.35 & 0.28 && 0.52    & 0.60    & 0.35 & 0.30 \\
    V5   && 0.73    & 0.80    & 0.25 & 0.31 && 0.74    & 0.80    & 0.25 & 0.29 \\
    V6   && 0.73    & 0.83    & 0.23 & 0.42 && 0.72    & 0.81    & 0.24 & 0.39 \\\hline
\end{tabular}\label{tab:interleadconv_incart_ii}
\end{table}

\begin{table}[!t]
\caption{Cross-database test results for INCART conversion from lead I.}
\centering
\begin{tabular}{ccccccccccc}
    \hline
    && \multicolumn{4}{c}{\textbf{Shared Encoder}} && \multicolumn{4}{c}{\textbf{Individual Encoders}} \\
    \textbf{Lead} && \textbf{$r$ (avg.)} & \textbf{$r$ (med.)} & \textbf{\textit{RMSE}} & \textbf{\textit{SSIM}} && \textbf{$r$ (avg.)} & \textbf{$r$ (med.)} & \textbf{\textit{RMSE}} & \textbf{\textit{SSIM}} \\ \hline
    II   && 0.35    & 0.37    & 0.37 & 0.18 && 0.36    & 0.38    & 0.37 & 0.19 \\
    III  && 0.17    & 0.19    & 0.41 & 0.27 && 0.17    & 0.19    & 0.43 & 0.28 \\
    aVR  && 0.65    & 0.74    & 0.23 & 0.81 && 0.67    & 0.78    & 0.22 & 0.83 \\
    aVL  && 0.40    & 0.49    & 0.32 & 0.52 && 0.36    & 0.46    & 0.35 & 0.47 \\
    aVF  && 0.17    & 0.15    & 0.41 & 0.23 && 0.17    & 0.14    & 0.41 & 0.22 \\
    V1   && 0.55    & 0.62    & 0.25 & 0.78 && 0.57    & 0.64    & 0.24 & 0.79 \\
    V2   && 0.50    & 0.57    & 0.27 & 0.73 && 0.50    & 0.56    & 0.28 & 0.73 \\
    V3   && 0.35    & 0.35    & 0.37 & 0.46 && 0.36    & 0.37    & 0.37 & 0.44 \\
    V4   && 0.27    & 0.26    & 0.43 & 0.16 && 0.34    & 0.36    & 0.41 & 0.19 \\
    V5   && 0.46    & 0.51    & 0.36 & 0.11 && 0.45    & 0.53    & 0.35 & 0.11 \\
    V6   && 0.44    & 0.49    & 0.34 & 0.22 && 0.45    & 0.52    & 0.34 & 0.21 \\\hline
\end{tabular}\label{tab:interleadconv_incart_i}
\end{table}

\subsection{Cross-database evaluation}

The cross-database tests aimed to assess the behaviour of the proposed methodology in more diverse scenarios. Here, the models used were the same as in the previous experiments (trained with PTB data), and the evaluation was conducted using data from the INCART and PTB-XL databases.

\begin{table}[!t]
\caption{Cross-database test results for PTB-XL conversion from lead II.}
\centering
\begin{tabular}{ccccccccccc}
    \hline
    & \multicolumn{4}{c}{\textbf{Shared Encoder}} & \multicolumn{4}{c}{\textbf{Individual Encoders}} \\
    \textbf{Lead} && \textbf{$r$ (avg.)} & \textbf{$r$ (med.)} & \textbf{\textit{RMSE}} & \textbf{\textit{SSIM}} && \textbf{$r$ (avg.)} & \textbf{$r$ (med.)} & \textbf{\textit{RMSE}} & \textbf{\textit{SSIM}} \\ \hline
    I    && 0.74    & 0.80    & 0.25 & 0.31 && 0.72    & 0.79    & 0.26 & 0.29 \\
    III  && 0.44   & 0.50     & 0.32 & 0.50 && 0.45     & 0.52    & 0.32 & 0.50 \\
    aVR  && -0.38    & -0.53     & 0.72 & 0.09 && -0.39     & -0.55    & 0.72 & 0.09 \\
    aVL  && -0.33    & -0.44     & 0.64 & 0.23 && -0.33     & -0.45    & 0.67 & 0.17 \\
    aVF  && 0.83    & 0.90     & 0.19 & 0.61 && 0.84      & 0.92    & 0.19 & 0.62 \\
    V1   && 0.79    & 0.87     & 0.17 & 0.91 && 0.81     & 0.89    & 0.16 & 0.91 \\
    V2   && 0.71    & 0.79     & 0.22 & 0.84 && 0.72     & 0.82    & 0.21 & 0.85 \\
    V3   && 0.61    & 0.68     & 0.30 & 0.66 && 0.62     & 0.70    & 0.30 & 0.66 \\
    V4   && 0.64    & 0.71     & 0.31 & 0.31 && 0.66     & 0.74    & 0.32 & 0.32 \\
    V5   && 0.79    & 0.86      & 0.22 & 0.37 && 0.80     & 0.87    & 0.23 & 0.39 \\
    V6   && 0.85    & 0.91     & 0.18 & 0.58 && 0.85     & 0.91    & 0.18 & 0.58 \\\hline
\end{tabular}\label{tab:interleadconv_ptbxl_ii}
\end{table}

\begin{table}[!t]
\caption{Cross-database test results for PTB-XL conversion from lead I.}
\centering
\begin{tabular}{ccccccccccc}
    \hline
    && \multicolumn{4}{c}{\textbf{Shared Encoder}} && \multicolumn{4}{c}{\textbf{Individual Encoders}} \\
    \textbf{Lead} && \textbf{$r$ (avg.)} & \textbf{$r$ (med.)} & \textbf{\textit{RMSE}} & \textbf{\textit{SSIM}} && \textbf{$r$ (avg.)} & \textbf{$r$ (med.)} & \textbf{\textit{RMSE}} & \textbf{\textit{SSIM}} \\ \hline
    II   &&  0.60   &  0.66  & 0.33 & 0.23 && 0.62    & 0.70 & 0.31 & 0.26  \\
    III  &&  0.31   &  0.34  & 0.38 & 0.43 && 0.33    & 0.38 & 0.37 & 0.45 \\
    aVR  &&  -0.61   & -0.76 & 0.74 & 0.09 && -0.62    & -0.78 & 0.74 & 0.09 \\
    aVL  &&  -0.63   & -0.75 & 0.75 & 0.11 && -0.66    & -0.79    & 0.76 & 0.08 \\
    aVF  &&  0.29   & 0.31   & 0.41 & 0.23 && 0.32    & 0.36    & 0.40 & 0.24 \\
    V1   &&  0.79   & 0.86   & 0.16 & 0.91 && 0.81    & 0.88    & 0.15 & 0.91 \\
    V2   &&  0.71   & 0.78   & 0.22 & 0.84 && 0.70    & 0.77    & 0.23 & 0.83 \\
    V3   &&  0.65   & 0.72   & 0.29 & 0.64 && 0.67    & 0.76    & 0.28 & 0.66 \\
    V4   &&  0.62   & 0.72   & 0.32 & 0.30 && 0.69    & 0.80    & 0.30 & 0.32 \\
    V5   &&  0.76   & 0.85   & 0.24 & 0.35 && 0.76    & 0.86    & 0.25 & 0.32 \\
    V6   &&  0.80   & 0.87   & 0.20 & 0.53 && 0.81    & 0.89    & 0.21 & 0.51 \\\hline
\end{tabular}\label{tab:interleadconv_ptbxl_i}
\end{table}

For both INCART and PTB-XL, some differences in interlead correlations can be observed when compared to PTB (see Table~\ref{tab:interleadconv_corr_between_leads_ii} and Table~\ref{tab:interleadconv_corr_between_leads_i}). This can be explained due to the different acquisition setups, especially the positioning of the electrodes, which potentially causes each lead to offer a different perspective.

\begin{figure}[p!]
\centering
    \includegraphics[width=0.91\linewidth]{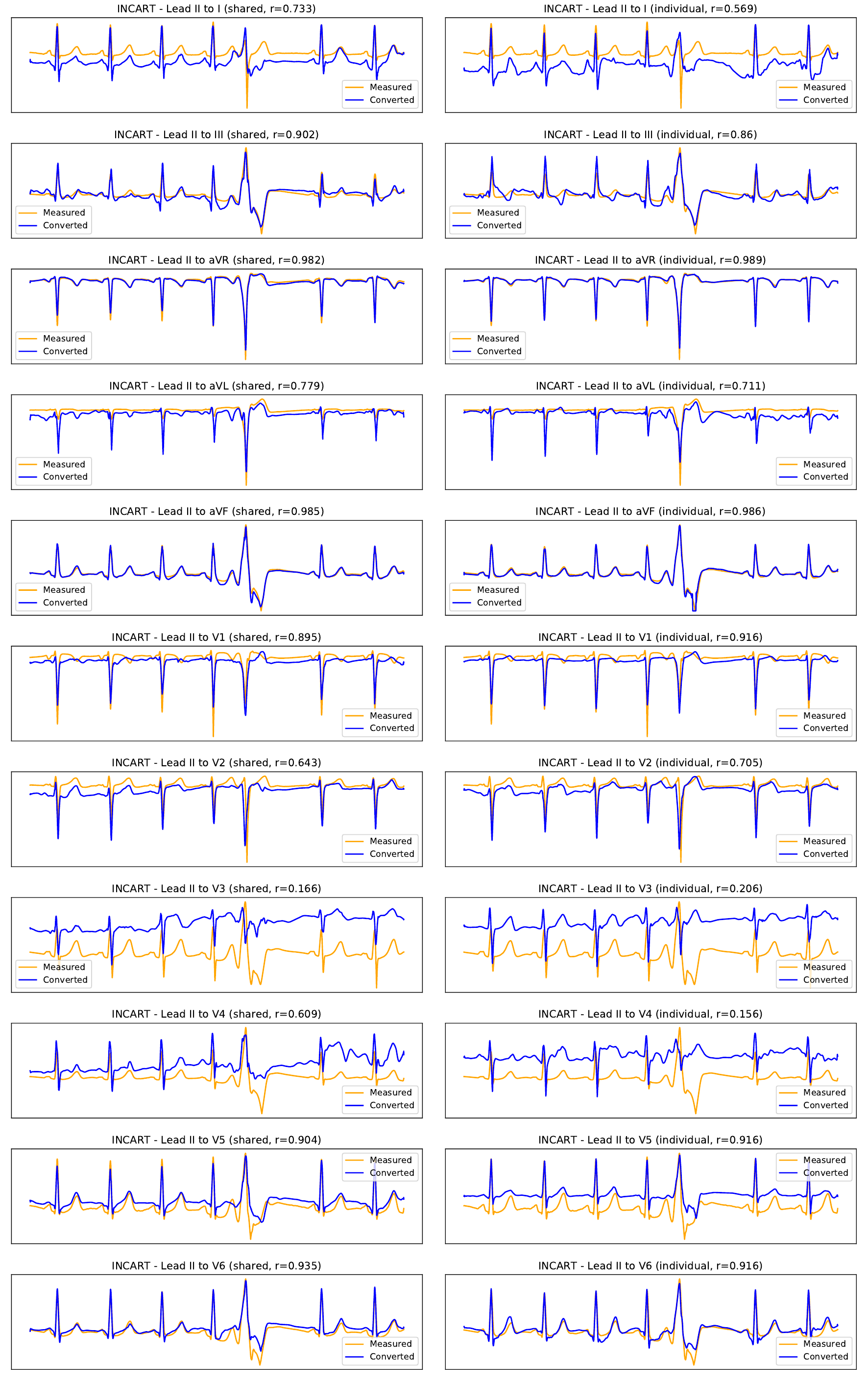}
    \caption[Example cross-database result of lead II to all conversion on INCART.]{Example cross-database result of lead II to all conversion on INCART (each row depicts one converted lead; shared encoder on the left; individual encoders on the right).}
    \label{fig:interleadconv_results_ii_to_all_incart}
\end{figure}

\begin{figure}[p!]
\centering
    \includegraphics[width=0.91\linewidth]{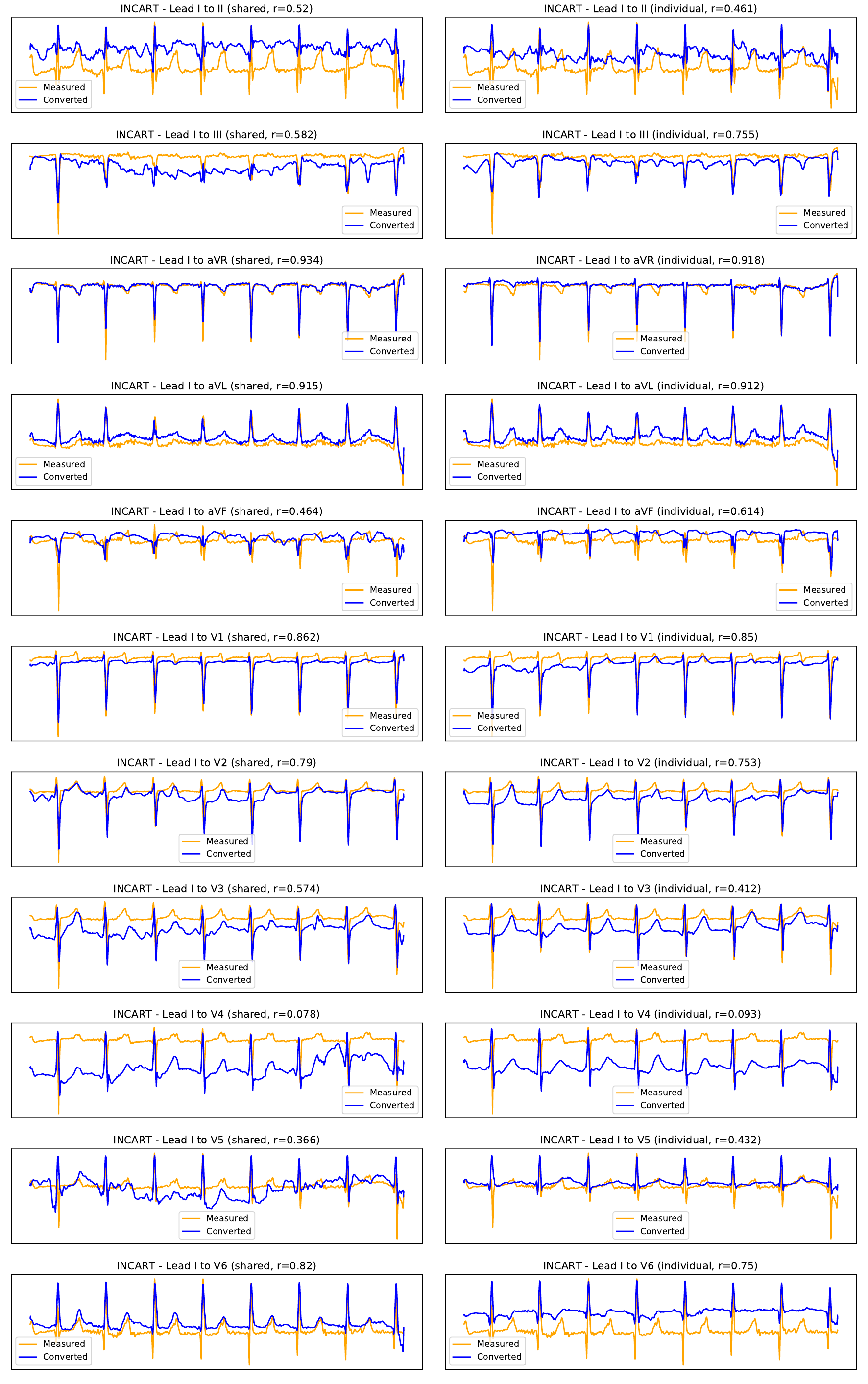}
    \caption[Example cross-database result of lead I to all conversion on INCART.]{Example cross-database result of lead I to all conversion on INCART (each row depicts one converted lead; shared encoder on the left; individual encoders on the right).}
    \label{fig:interleadconv_results_i_to_all_incart}
\end{figure}

For INCART (see Table~\ref{tab:interleadconv_incart_ii} and Table~\ref{tab:interleadconv_incart_i}), the overall quality of the results is inferior to those with PTB. This is as expected since PTB data was seen by the models during training and the INCART database is arguably more challenging regarding noise and variability. Despite these metrics, it is noticeable in Figure~\ref{fig:interleadconv_results_ii_to_all_incart} and Figure~\ref{fig:interleadconv_results_i_to_all_incart} that both reference leads can offer good conversion results in some leads, especially with lead II. Using this lead as reference, the proposed methodology is relatively good at converting most leads except I, V2, and V3.

\begin{figure}[p!]
\centering
    \includegraphics[width=0.91\linewidth]{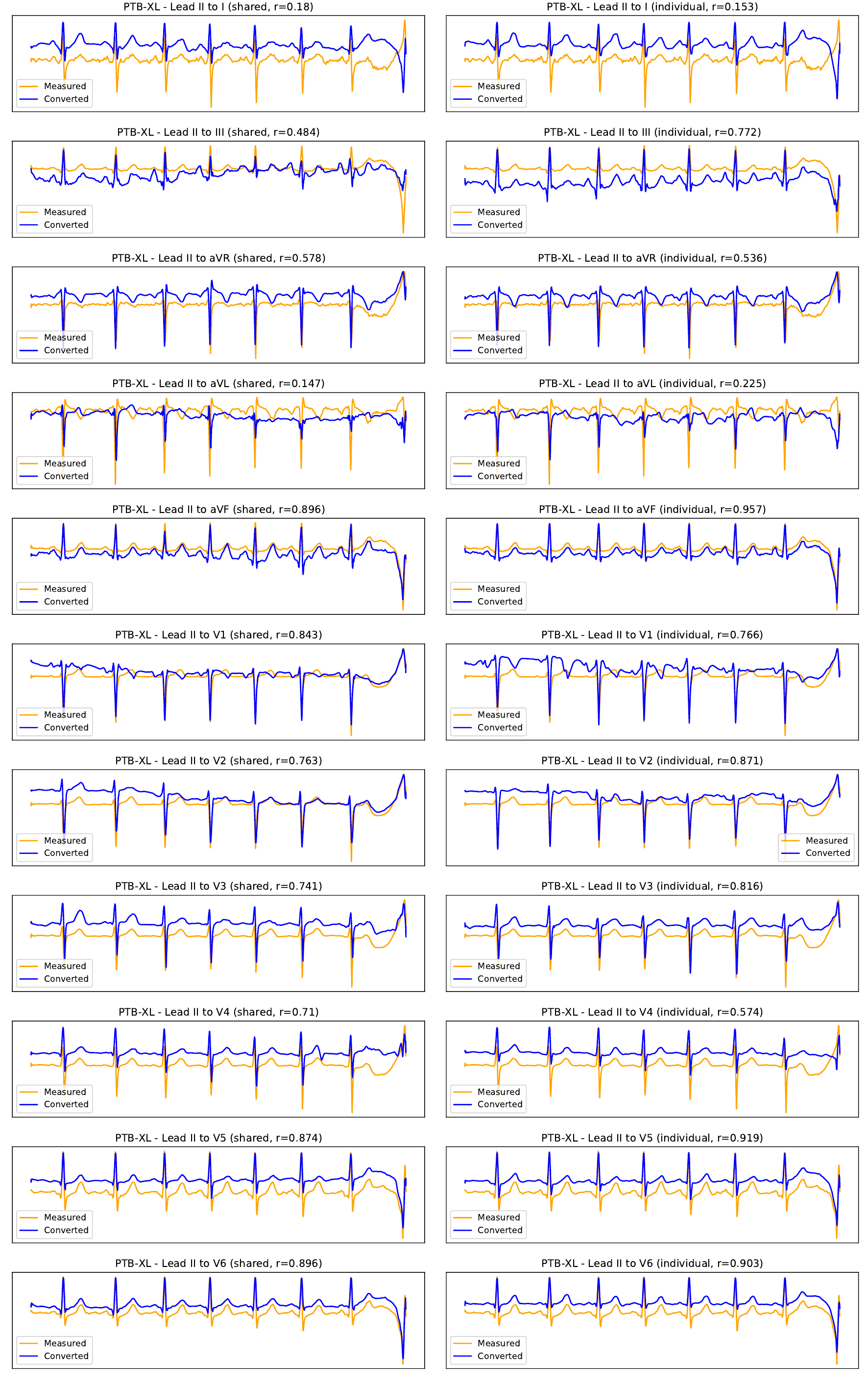}
    \caption[Example cross-database result of lead II to all conversion on PTB-XL.]{Example cross-database result of lead II to all conversion on PTB-XL (each row depicts one converted lead; shared encoder on the left; individual encoders on the right).}
    \label{fig:interleadconv_results_ii_to_all_ptbxl}
\end{figure}

\begin{figure}[p!]
\centering
    \includegraphics[width=0.91\linewidth]{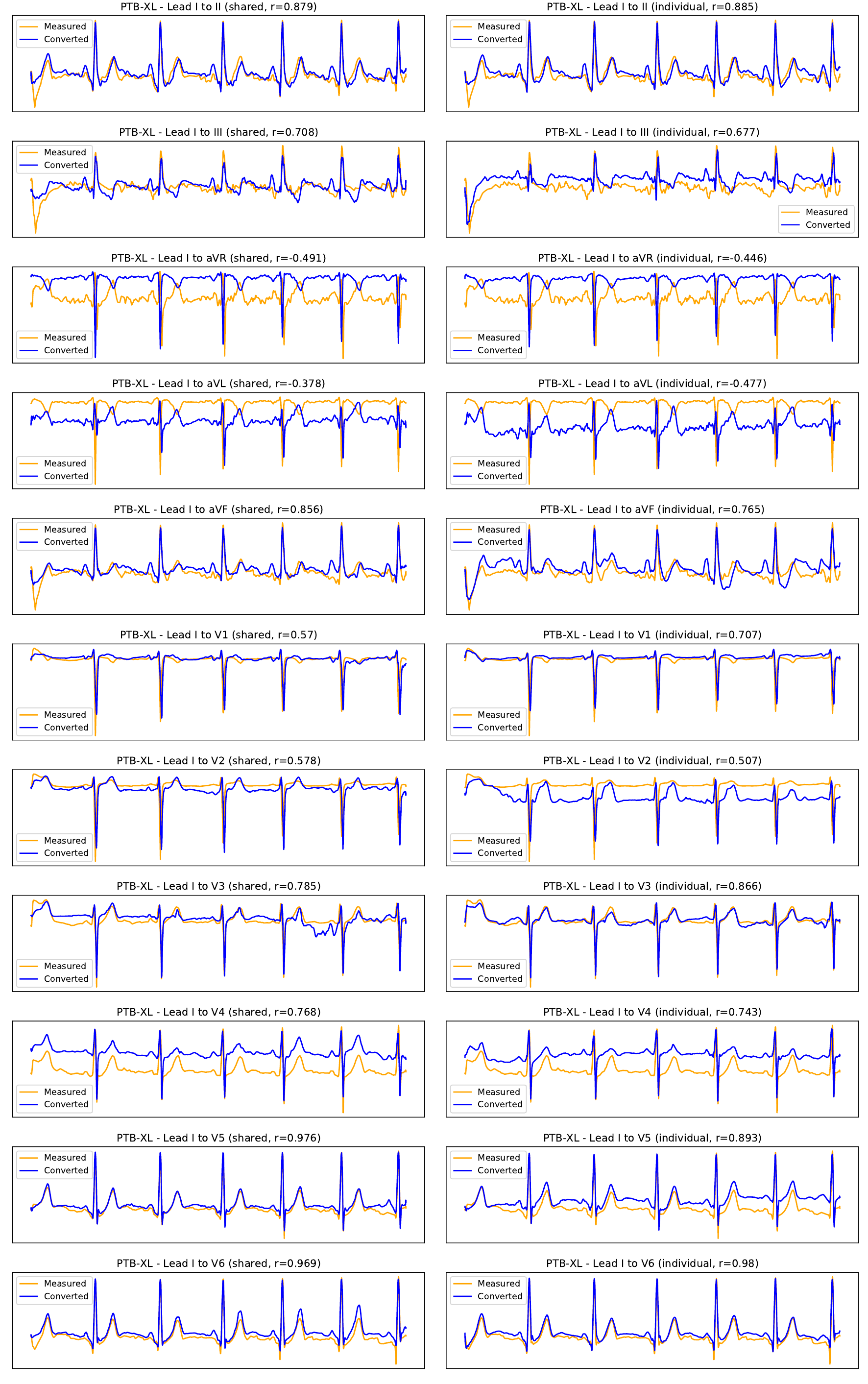}
    \caption[Example cross-database result of lead I to all conversion on PTB-XL.]{Example cross-database result of lead I to all conversion on PTB-XL (each row depicts one converted lead; shared encoder on the left; individual encoders on the right).}
    \label{fig:interleadconv_results_i_to_all_ptbxl}
\end{figure}

For PTB-XL (see Table~\ref{tab:interleadconv_ptbxl_ii} and Table~\ref{tab:interleadconv_ptbxl_i}), results are, overall, the worst, although some leads (namely V4, V5, and V6), due to higher correlation with the reference leads, are better reconstructed than with the PTB database. In Fig.~\ref{fig:interleadconv_results_ii_to_all_ptbxl} and Fig.~\ref{fig:interleadconv_results_i_to_all_ptbxl}, it is possible to observe that, despite occasional baseline offset and prevalent noise, both reference leads enable the approximate reconstruction of most of the set of twelve standard leads.

For either database, differences in acquisition settings and electrode placement result in inferior performance. The ideal solution is to always make sure the acquisition details of training and inference data match, to ensure optimal performance upon deployment. Nevertheless, the robustness in cross-database scenarios is a relevant issue that merits further research.

\subsection{Influence of medical conditions}

As aforementioned, medical conditions may affect differently the various leads of an ECG signal. While this is the main motivation behind the quest to reconstruct missing leads it may also be one of the main hurdles. If the medical condition is somehow not evident in the input lead, the algorithm could be led to reconstruct the remaining leads incorrectly without the proper information on the respective medical condition.

As such, we conducted a differential performance evaluation according to the existence and type of diagnosed medical conditions on the signals. To do this, we use the expert clinical annotations on the PTB-XL database and separate the results by the superclass labelling of each test sample. The average $r$ results for each converted lead and each superclass are presented in Table~\ref{tab:interleadconv_ptbxl_medical_ii} (using lead II as reference) and Table~\ref{tab:interleadconv_ptbxl_medical_i} (using lead I as reference).

Overall, no dominant difference could be observed between the results with normal signals and the results with signals with medical conditions. Similarly, no specific medical condition superclass presents considerably different performance results. This is likely due to the presence of medical conditions on the PTB signals originally used for training the model. Thus, although the behaviour of the proposed methodology should be expected to vary slightly in the presence of medical conditions, it should not have a considerable impact on its baseline performance.

\section{Summary and Conclusions}

This work implemented and compared the performance of three deep learning architectures for interlead conversion of ECG signals. Unlike the literature, this work focused on the more challenging scenario of single-lead blindly-segmented inputs from limb leads. The proposed model was explored on 12-lead acquisitions from three different databases. Ablation studies were conducted on the architectures used for conversion and on the use of a shared encoder vs. individual encoders. Moreover, the model was evaluated in both single-database and cross-database scenarios, including an experiment on the effect of medical conditions on signal reconstruction and the study of diagnosis performance with original \emph{vs}. converted signals.

Despite the considerably more challenging scenario, the proposed methodology based on a U-Net was capable of obtaining relatively good results. Each reference lead enabled the high-quality reconstruction of several of the twelve standard ECG leads, in some cases reaching state-of-the-art level performance. Both lead I and II appear to be especially suitable for certain sets of leads, and could be used on specific target applications that focus on those.

\begin{table}[!t]
\caption[Average correlation results for PTB-XL conversion from lead II according to medical condition class.]{Average correlation results for PTB-XL conversion from lead II according to medical condition class (using the U-Net with a shared encoder).}
\centering
\begin{tabular}{cccccccccccc}
    \hline
    & \multicolumn{11}{c}{\textbf{Converted leads}} \\
    \textbf{Class} & \textbf{\textit{I}} & \textbf{\textit{III}} & \textbf{\textit{aVR}} & \textbf{\textit{aVL}} & \textbf{\textit{aVF}} & \textbf{\textit{V1}} & \textbf{\textit{V2}} & \textbf{\textit{V3}} & \textbf{\textit{V4}} & \textbf{\textit{V5}} & \textbf{\textit{V6}} \\ \hline
    NORM & 0.79 & 0.42 & -0.39 & -0.32 & 0.86 & 0.83 & 0.76 & 0.64 & 0.70 & 0.85 & 0.90 \\
    MI & 0.65 & 0.47 & -0.37 & -0.38 & 0.76 & 0.75 & 0.65 & 0.56 & 0.54 & 0.68 & 0.77 \\
    STTC & 0.71 & 0.41 & -0.41 & -0.37 & 0.80 & 0.79 & 0.69 & 0.58 & 0.59 & 0.78 & 0.84 \\
    CD & 0.65 & 0.60 & -0.31 & -0.28 & 0.84 & 0.59 & 0.54 & 0.58 & 0.58 & 0.63 & 0.71 \\
    HYP & 0.75 & 0.42 & -0.40 & -0.24 & 0.80 & 0.85 & 0.70 & 0.56 & 0.61 & 0.81 & 0.87 \\\hline
\end{tabular}\label{tab:interleadconv_ptbxl_medical_ii}
\end{table}

\begin{table}[!t]
\caption[Average correlation results for PTB-XL conversion from lead I according to medical condition class.]{Average correlation results for PTB-XL conversion from lead I according to medical condition class (using the U-Net with a shared encoder).}
\centering
\begin{tabular}{cccccccccccc}
    \hline
    & \multicolumn{11}{c}{\textbf{Converted leads}} \\
    \textbf{Class} & \textbf{\textit{II}} & \textbf{\textit{III}} & \textbf{\textit{aVR}} & \textbf{\textit{aVL}} & \textbf{\textit{aVF}} & \textbf{\textit{V1}} & \textbf{\textit{V2}} & \textbf{\textit{V3}} & \textbf{\textit{V4}} & \textbf{\textit{V5}} & \textbf{\textit{V6}} \\ \hline
    NORM &  0.73 & -0.23 & -0.35 & -0.66 &  0.23 & 0.79 & 0.73 & 0.68 & 0.69 & 0.84 & 0.85 \\
    MI & 0.50 & -0.15 & -0.26 & -0.57 & 0.036 & 0.67 & 0.58 & 0.47 & 0.38 & 0.56 & 0.61 \\
    STTC & 0.57 & -0.20 & -0.36 & -0.64 & 0.00 & 0.74 & 0.64 & 0.55 & 0.48 & 0.71 & 0.76 \\
    CD   & 0.40 & -0.31 & -0.11 & -0.48 & -0.06 & 0.35 & 0.38 & 0.42 & 0.39 & 0.52 & 0.54 \\
    HYP  & 0.65 & -0.29 & -0.37 & -0.60 & 0.07 & 0.78 & 0.65 & 0.58 & 0.59 & 0.78 & 0.82 \\\hline
\end{tabular}\label{tab:interleadconv_ptbxl_medical_i}
\end{table}

In the cross-database scenario, despite the acquisition setup differences, results were promising especially with the INCART database. Finally, the analysis of the influence of medical conditions has shown no considerable effect of pathologies on the performance of the proposed methodology. However, a state-of-the-art methodology for automatic diagnosis revealed lower accuracy when using reconstructed signals, a problem that should be addressed in future research.

Although the results are promising, further efforts should be devoted towards the improvement of the methodologies for interlead conversion using single-lead blindly-segmented inputs. Namely, the pre-processing and normalisation of the signals, as well as the robustness to diverse acquisition setups, should be the target of further research. Additionally, task-oriented objective functions should be explored to ensure useful signal information is kept and avoid performance losses in subsequent diagnoses.

With some consolidation, the proposed methodology could be the key to better cardiac health monitoring in wearable devices and less obtrusive clinical scenarios. Taking the example of emergency rooms, if we can retrieve all twelve leads (or the most important among these) from Lead I signals, then patients will only need two electrodes placed on the wrists to have their ECG collected, instead of the full set of 10 electrodes on wrists, ankles, and chest. This is a meaningful step towards higher comfort and usability for patients in clinical settings or users in other scenarios involving the monitoring of ECG signals. Additionally, albeit outside the scope of this work, this methodology could also be applicable to other multi-channel signals where the different channels correspond to different perspectives over the same physiological phenomenon.

\part{Face Biometrics}\label{part:faceBiometrics}
\chapter{Prior Art in Face Biometrics}\label{ch:faceprior}

\section{Data}

There are several publicly available databases for research purposes, to develop and benchmark face biometric recognition algorithms. Considering the face is one of the most developed biometric traits, the databases available are some of the largest and most complete, thoughtfully structured for deep and adequate evaluation of recognition algorithms. Table~\ref{tab:priorartface_face_databases} compiles some relevant information about the most important databases currently available, which are also described below:

\begin{table}[t!]
\caption[Details on the main face recognition databases that are currently available.]{Details on the main face recognition databases that are currently available.}\label{tab:priorartface_face_databases}
\centering
\begingroup
\small
\renewcommand*{\arraystretch}{1.5}
\begin{tabular}{lcccccc}\hline
\textbf{Database}  & \textbf{Spectrum} & \textbf{Subjects} & \textbf{Images} & \textbf{Videos} & \multicolumn{1}{c}{\textbf{Resolution}} & \textbf{Unconstr.} \\ \hline
CASIA NIR-VIS \citep{Li2013}         & Vis. + NIR & 725      & 17~580      &  None     &   640x480         &  \xmk    \\
CASIA WebFace                        & Visible    & 10~575   & 494~414     &  None     &   250x250         &  \cmk    \\
CelebA \citep{Liu2015}               & Visible    & 10~177   & 202~599     &  None     &   -               &  \cmk    \\
COX Face \citep{Huang2015}           & Visible    & 1000     & 1000        &  3000     &   -               &  \xmk    \\
CSIST Lab1 \citep{Xu2011}            & Vis. + NIR & 50       & 1000        &  None     &   100x80          &  \xmk    \\
CSIST Lab2 \citep{Xu2011}            & Vis. + NIR & 50       & 2000        &  None     &   200x200         &  \xmk    \\
FERET                                & Visible    & 1199     & 14~126      &  None     &   512x768         &  \cmk    \\
IJB-C                                & Visible    & 3531     & 138~836     &  11~779   &   -               &  \cmk    \\
IMDb-Face  \citep{Wang2018}          & Visible    & 59~000   & 1~700~000   &  None     &   -               &  \cmk    \\
LFW \citep{Huang2007, Huang2014}     & Visible    & 5749     & 13~233      &  None     &   250x250         &  \cmk    \\
MegaFace \citep{Nech2017}            & Visible    & 672~057  & 4~753~520   &  None     &   -               &  \cmk    \\
MS-Celeb                             & Visible    & 99~892   & 8~200~000   &  None     &   -               &  \cmk    \\
PaSC \citep{Beveridge2013}           & Visible    & 293      & 9376        &  2802     &   -               &  \cmk    \\
PolyU \citep{Zhang2010}              & NIR        & 335      & 34~000      &  None     &   768x576         &  \xmk    \\
UMD Faces \citep{Bansal2017}         & Visible    & 8277     & 367~888     &  None     &   -               &  \cmk    \\
UMD Videos \citep{Bansal2017}        & Visible    & 3107     & None        &  22~075   &   -               &  \cmk    \\
VGGFace \citep{Parkhi2015}           & Visible    & 2622     & 2~600~000   &  None     &   Diverse         &  \cmk    \\
VGGFace2 \citep{Cao2018}             & Visible    & 9131     & 3~310~000   &  None     &   Diverse         &  \cmk    \\
Yale Face                            & Visible    & 15       & 165         &  None     &   320x243         &  \xmk    \\
YouTube Faces \citep{Wolf2011}       & Visible    & 1595     & None        &  3495     &   -               &  \cmk    \\\hline
\end{tabular}
\endgroup
\end{table}

\begin{itemize}
    \item \emph{CASIA NIR-VIS}: Also known as CBSR NIR, the CASIA NIR-VIS 2.0 database was created by the Institute of Automation of the Chinese Academy of Sciences (CASIA). It includes pairs of mugshots and corresponding NIR images from 725 people, acquired over four recording sessions, from 2007 to 2010 \citep{Li2013};
        
    \item \emph{CASIA WebFace}: This database was created and made available by the Institute of Automation of the Chinese Academy of Sciences (CASIA). It includes almost five hundred thousand images from more than ten thousand identities collected to support the research in unconstrained face recognition;
    
    \item \emph{CelebA}: This database results from a previous one, CelebFaces+, which has been enriched with fiducial and attribute annotations. It includes over two hundred thousand pictures from over ten thousand celebrities, with five fiducial locations, and forty binary attributes per image \citep{Liu2015};
    
    \item \emph{COX Face}: The COX Face database was designed to study recognition across still images and videos. Thus, it includes still images from one thousand subjects in a controlled environment with high quality, and surveillance videos from the subjects in unconstrained and low-quality settings \citep{Huang2015};
    
    \item \emph{CSIST Lab1}: The CSIST database was developed by the Chung-Shan Institute of Science and Technology, with images from volunteers at the Harbin Institute of Technology of Shenzhen. The Lab1 dataset contains ten visible light and ten NIR images from each of fifty subjects \citep{Xu2011};
    
    \item \emph{CSIST Lab2}: The CSIST Lab2 dataset is part of the CSIST database, and includes twenty visible light and twenty NIR images from fifty volunteers, with natural lighting and artificial lighting \citep{Xu2011};
    
    \item \emph{FERET}: The Facial Recognition Technology (FERET) Program was sponsored by the US Department of Defence, and the FERET database is distributed by the National Institute of Standards and Technology (NIST). The database, with more than fourteen thousand face images collected between 1993 and 1996, has the goal to support the development of new techniques for automatic recognition of faces; 
    
    \item \emph{IJB-C}: The IARPA Janus Benchmark-C (IJB-C) is a database resulting from a group of challenges created by NIST addressing verification, identification, detection, clustering, and processing of videos. It includes over one hundred and thirty-eight thousand images and eleven thousand face videos from over three thousand identities;
    
    \item \emph{IMDb-Face}: This dataset includes approximately 1.7 million images with faces from fifty-nine thousand celebrities on the IMDb movie database. According to the authors, efforts have been devoted to ensuring this database is cleaner than most other large available databases, making it better for training robust algorithms \citep{Wang2018};
    
    \item \emph{LFW}: The Labelled Faces in the Wild dataset was one of the first databases of unconstrained face images, and includes over thirteen thousand face images of almost six thousand identities \citep{Huang2007, Huang2014}; 
    
    \item \emph{MegaFace}: The MegaFace collection includes an average of 7 unconstrained face photos (between 2 and 2469) for each of almost seven hundred thousand identities. It is the surveyed database with the most identities, and the second with most total images \citep{Nech2017};
    
    \item \emph{MS-Celeb}: The Microsoft Celeb is a dataset of over eight million images obtained online. It includes faces from nearly one hundred thousand celebrities in unconstrained settings, aiming to accelerate research in face recognition with large target sets;
    
    \item \emph{PaSC}: The Point and Shoot Face Recognition Challenge (PaSC) dataset was created by NIST to encourage the development of face recognition algorithms that are more robust to very unconstrained settings and inexpensive camera technology. It includes over nine thousand images and almost three thousand videos with faces from almost three hundred identities with different distances to the camera, perspective, and locations \citep{Beveridge2013}; 
    
    \item \emph{PolyU}: This database was created by the Biometric Research Centre (UGC/CRC) at The Hong Kong Polytechnic University, using their own real-time NIR face capture device. It includes approximately thirty-four thousand NIR face images from over three hundred subjects \citep{Zhang2010};
    
    \item \emph{UMD Faces}: This dataset was created at the University of Maryland (UMD) and contains almost four hundred thousand face images from over eight thousand subjects. Each image includes human-curated face bounding boxes and annotations on pose, gender, and twenty-one keypoints \citep{Bansal2017}; 
    
    \item \emph{UMD Videos}: This dataset is similar to UMD Faces, only it includes over three million video frames from twenty-two thousand videos with over three thousand identities. The frames are also annotated with pose, keypoints, and gender information \citep{Bansal2017};
        
    \item \emph{VGGFace}: This database was created by the Visual Geometry Group (VGG) of the University of Oxford, UK. The group developed a method for minimally-supervised online image collection, which enabled the creation of this large database, with over two million faces in unconstrained conditions \citep{Parkhi2015};
    
    \item \emph{VGGFace2}: The VGGFace database was later extended to create the VGGFace2 database, which includes more than three million unconstrained face images from almost ten thousand identities \citep{Cao2018}; 
    
    \item \emph{Yale Face}: The Yale Face database includes eleven images from each of fifteen subjects. Although not an unconstrained database, it includes annotations on certain expressions and configurations simulated by the subjects, which can be useful in training models for other tasks such as emotion recognition;
    
    \item \emph{YouTube Faces}: This dataset is composed exclusively of faces on YouTube videos. It includes over three thousand videos with over one thousand identities. The videos range from 48 to 6070 frames (average of 181 frames per video) \citep{Wolf2011}.
\end{itemize}

These databases already cover most bases and offer a good starting point for the study and development of strong biometric algorithms. Nevertheless, it is important to acknowledge the growth of heterogeneous approaches in facial recognition, and the subsequent need for databases of face images acquired in different modalities (\eg, visual spectrum vs. NIR). These databases are still too small and too controlled for the development of robust algorithms. Moreover, it would be useful to have larger databases focused on more specific applications (\eg, face images and videos of car drivers).

\section{Related Work}

\begin{figure}%
\centering
\includegraphics[width=\linewidth]{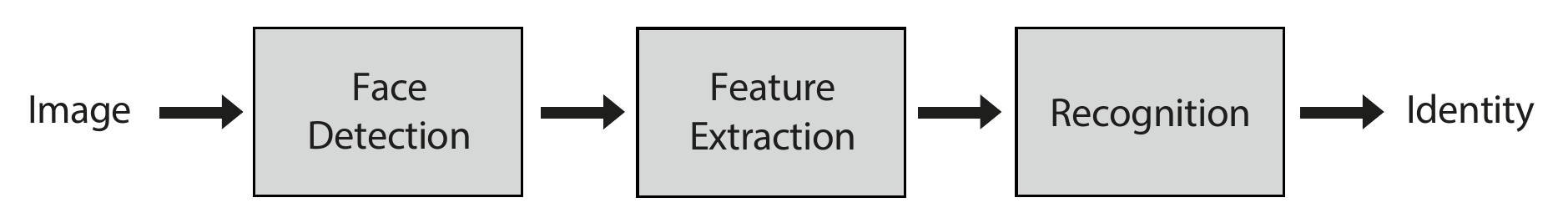}%
\caption[Stages of a biometric recognition algorithm based on face images.]{Stages of a biometric recognition algorithm based on face images (based on~\citep{Barnouti2016}).}%
\label{fig:priorartface_face_stages}%
\end{figure}

According to \citet{Barnouti2016}, face recognition can be decomposed into three processes: face detection, feature extraction, and face recognition (see Fig.~\ref{fig:priorartface_face_stages}). Below, we delve into the state-of-the-art in each of these processes. Due to the currently common practice of joining feature extraction and face recognition into a single model using deep learning, these two processes are jointly discussed.

\subsection{Face detection}

Given an image or video stream, the process of face detection has the goal of locating and extracting all human faces visible in the received input. It is an extremely important process not only for face-based biometric recognition, but also for face tracking and person re-identification across surveillance cameras, recognition of expressions and emotions, and analysis of soft-biometrics such as gender or age \citep{Zafeiriou2015, Kumar2019}.

\begin{figure}%
\centering
\includegraphics[height=0.23\linewidth]{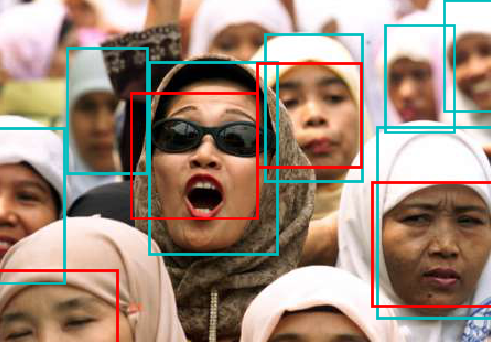}
\includegraphics[height=0.23\linewidth]{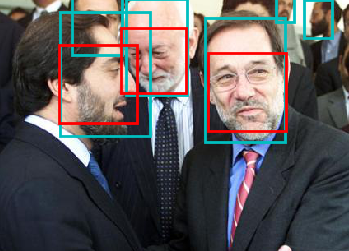}
\includegraphics[height=0.23\linewidth]{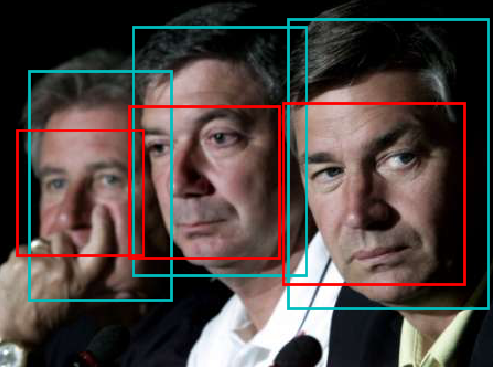}
\caption[Examples of face detection in unconstrained settings.]{Examples of face detection in unconstrained settings (images from the FDDB database~\citep{Jain2010}, ground-truths in green and predictions in red).}%
\label{fig:priorartface_face_detection}%
\end{figure}

Some earlier, simpler methods relied on skin colour for face detection on images \citep{Crowley1997, Kjeldsen1996}. These presented the advantage of being orientation-invariant, as it would serve to detect a face even if it did not present a frontal pose. However, they fail to consider the great variety of skin colours, both due to natural differences between individuals, and due to diverse illumination conditions.

More sophisticated algorithms have been proposed, including the method by \citet{Sirovich1987} based on eigenvectors from large face image datasets, the method by \citet{Viola2001} which uses cascades of Haar transform filters selected using AdaBoost, or the method by \citet{Dalal2005} which uses histograms of intensity gradients from image regions and their orientation. These commonly offer very fast detection, adequate for real-time systems, but often fail on non-frontal face detection and faces of very diverse scales.

Like most pattern recognition tasks, traditional methods from earlier literature have been recently replaced with deep learning algorithms. These offer more robust and accurate face detections, especially for non-frontal face detection, making better use of very large datasets currently available.

Some of these datasets currently offer a public benchmark for fair and direct comparison with state-of-the-art methods. These include the WIDER face dataset~\citep{Yang2016} and the Face Detection Database (FDDB)~\citep{Jain2010}. These benchmarks are currently largely dominated by deep learning approaches.

The FDDB benchmark is currently dominated by the S$^3$FD, the DeepIR, and the RSA algorithms. S$^3$FD~\cite{Zhang2017c} is based on a single deep network specifically fitted to better detect small faces. The DeepIR method~\citep{Sun2018} uses a Faster Region-based CNN (RCNN) adapted with feature concatenation, multiscale training, and hard negative mining for more robust detections. The RSA algorithm~\citep{Liu2017} is based on a convolutional network with a recurrent strategy for detection at different scales.

The best scores on the WIDER Face benchmark belong to the AInnoface and the RetinaFace algorithms. The RetinaFace algorithm~\citep{Deng2019} is a single-stage pixelwise face detector that takes advantage of extra-supervised and self-supervised multitask learning. AInnoface~\citep{Zhang2019} is based on RetinaFace with two-step classification and regression, an IoU loss function, improved data augmentation, and several other structural network changes. 

These sophisticated algorithms perform accurate and robust detection for faces in different poses and at different scales. But they are still computationally heavy and require GPUs for real-time operation in images with VGA resolution. Having overcome most challenges in unconstrained face detection, research should now focus on making algorithms faster and more efficient.

\subsection{Feature extraction and recognition}

Having extracted the detected faces from the input, face-based recognition systems need to extract appropriate features from those faces to accurately decide on their identities. \citet{Wang2018b}, in their survey of deep learning face recognition, have pointed out how the field of face biometrics has moved from traditional machine learning approaches to deep learning (see Fig.~\ref{fig:face_recog_history}). However, the best results were only obtained when the development of sophisticated tailored objective functions began (see Fig.~\ref{fig:face_recog_history_2}).

\afterpage{
\begin{landscape}
\begin{figure}
    \centering
    \includegraphics[width=0.95\linewidth]{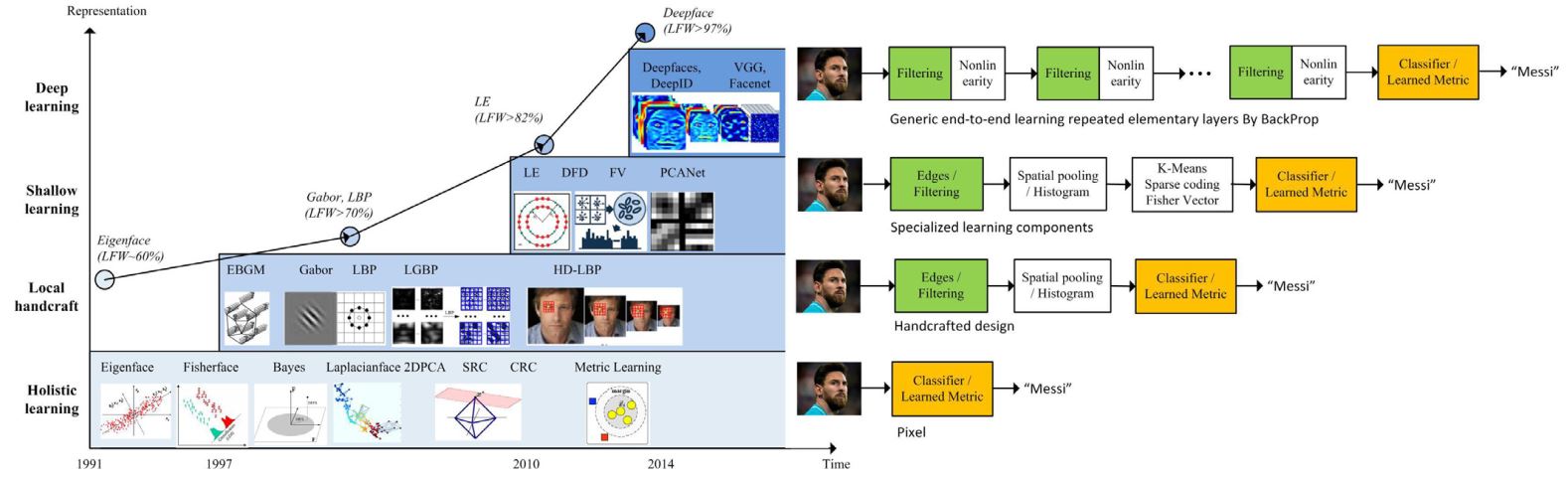}
    \caption[Evolution of face recognition approaches, from holistic to deep learning.]{Evolution of face recognition approaches, from holistic to deep learning (from~\cite{Wang2018b}).}
    \label{fig:face_recog_history}
\end{figure}

\begin{figure}
    \centering
    \includegraphics[width=0.85\linewidth]{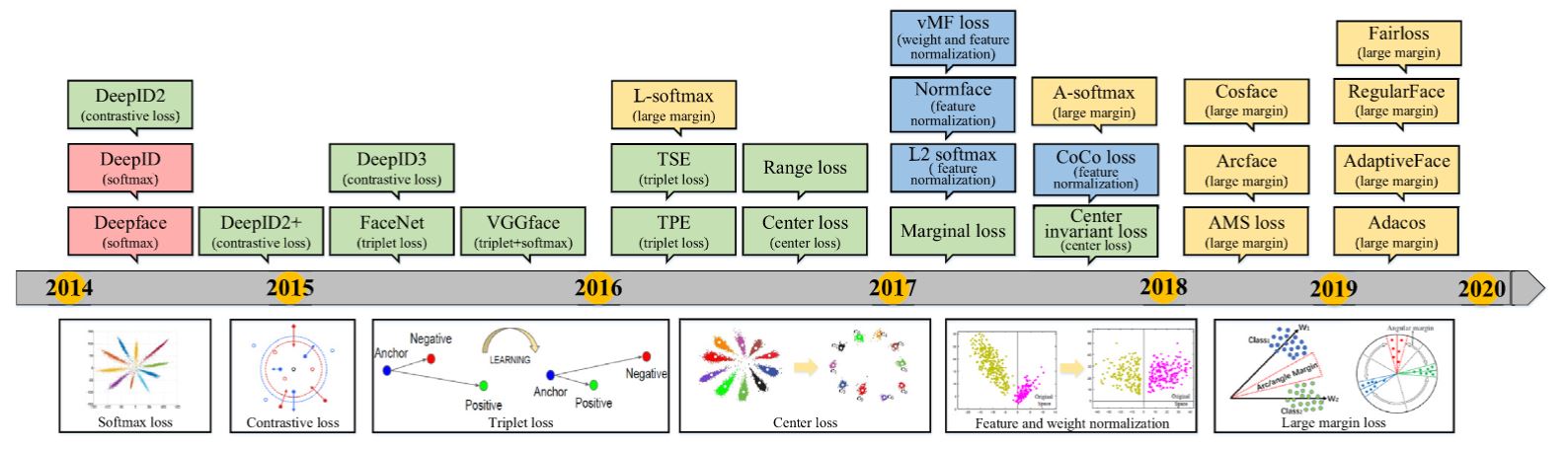}
    \caption[Recent history of face recognition, from deep learning to tailored objective functions.]{Recent history of face recognition, from deep learning to tailored objective functions (from~\cite{Wang2018b}).}
    \label{fig:face_recog_history_2}
\end{figure}
\end{landscape}
}

As such, \citet{Wang2018b} divide approaches into four categories: holistic learning, local handcrafted, shallow learning, and deep learning methods. Below, the most relevant examples of each category are presented, along with their advantages and shortcomings.

Holistic learning approaches are those that use the whole face image to obtain representations that ease the process of face recognition. The Eigenface method~\citep{Sirovich1987}, described for face detection, is one of these methods, along with the Fisherface method~\citep{Belhumeur1997}, which is similar to Eigenface, but uses the Fisher Linear Discriminant Analysis (FLDA, instead of PCA) for dimensionality reduction. Such methods are simple and fast, but lack robustness to several variability factors. 

Eventually, researchers started to explore methods that extracted features from regions of the face image. These methods mostly used Gabor filters and Local Binary Patterns for feature extraction based on intensity gradients and image edges~\citep{Ahonen2006, Liu2002}. Methods like these and the Elastic Bunch Graph Matching~\citep{Wiskott1997} were able to improve recognition accuracy, but not to make it high enough for real use.

To improve accuracy and robustness to pose variations, researchers proposed learning-based methods, that used available data to learn the best features. The first method to use shallow learning was proposed by \citet{Cao2010}, using gradient filtering after facial landmark alignment and clustering methods to learn encodings for better recognition.

But truly high accuracy and robustness in face recognition were only attained with the rise of deep learning. The first models were convolutional neural networks with conventional architectures, such as DeepFace~\citep{Taigman2014}, VGG-Face~\citep{Parkhi2015}, or VGG-Face2~\citep{Cao2018}. Over time, researchers started to focus on adapting the networks for specific details of face recognition, such as custom loss functions that force increased intersubject separability. This resulted in improved methods such as DeepID~\citep{Sun2014}, L2-Softmax~\citep{Ranjan2017}, and ArcFace~\citep{deng2019arcface}.

Both Facenet and DeepID are among the top five non-commercial methods in the LFW benchmark, with $99.63\%$ and $97.45\%$ accuracy, respectively. However, as discussed by~\citet{Wang2018b}, the best results yet haven't been achieved with traditional deep learning losses such as softmax or even triplet loss, but with tailored objective functions specifically designed to make the most of available data for the task of face recognition. According to results reported by \citet{Wang2018b}, both L2-softmax and ArcFace offer even better performance in the LFW benchmark, with $99.78\%$ and $99.83\%$ accuracy, respectively. In fact, ArcFace, using the tailored ArcLoss objective function, is still widely recognised as the state-of-the-art approach for face recognition.

These methods and results show the high potential offered by adapted deep learning networks for face recognition. However, these present the same problem as deep learning approaches for face detection: high computational cost. These models are generally very complex, and researchers should devote efforts to making them more efficient. Furthermore, video-based benchmarks should be used more to evaluate methods also based on their timeliness. Finally, as stated by \citet{Arya2015}, research in visible or infrared spectra may be reaching its limits, and the future may be based on multispectral imaging. 

Overall, face biometric recognition is already a thoroughly developed topic, unlike other biometric characteristics (such as the electrocardiogram). Joint efforts dedicated by several international research groups throughout multiple decades have brought this topic to a stage of high maturity that enables real applications. Proof of that is the currently endless variety of applications of face biometrics in our day-to-day routine, from unlocking our phones to border control, including opening bank accounts remotely.

\subsection{Presentation attack detection}

Just like any other biometric solution or access control system, face biometric systems are prone to attacks. These systems commonly guard goods or information whose value entices attackers to try to fool it into granting them access. In face biometrics, one of the main ways to do this is through presentation attacks: here, an attacker presents to the sensor fake or altered samples of the biometric trait that contains identity information from an authorised person~\cite{Sequeira2021Exploratory}.

Presentation attack detection (PAD) algorithms aim to automatically recognise when captured biometric samples have been faked or altered in such a way, preventing a biometric system from granting access to an attacker. They consist of binary classifiers that distinguish between presentation attacks or \emph{bona fide} samples. Presentation attacks involve presentation attack instruments (PAI) which can be printed photographs, digital screens, paper masks, or even tridimensional silicone masks: each of these types of PAI is called a PAI species (PAISp)~\cite{ISOPAD2017}. 

Earlier PAD approaches focused on one single PAISp (\emph{i.e.}, trained and tested only with one type of PAI), a scenario that can be designated as \emph{one-attack}. This can naturally lead to overly optimistic results that may not be verified in real-life applications, since attackers are perpetually working on new and improved PAI species and face biometric systems will expectedly be faced with more than one during application.

Another (more challenging) scenario is the one where multiple PAISp are used for training and a set of unknown PAISp are used to evaluate performance. This is called \emph{unseen-attack} and typically enables performance results much more closely related to what would be verified in real applications. A PAD algorithm that is able to generalise well from the seen PAISp to the unseen PAISp and offer good performance in the unseen-attack scenario is much more likely to perform well when faced with novel PAISp once deployed.

As with several other pattern recognition tasks, deep learning architectures have been widely applied in PAD, especially in recent years~\cite{Perez-Cabo_2019_CVPR_Workshops,Bhattacharjee2019}. These have enabled reaching improved results that allude to the possibility of real implementation. Nevertheless, the evaluation scenarios typically resemble the one-attack scenario and, thus, raise serious doubts about the realism of the reported results. In fact, several works on PAD have doubled down on this issue and called for the widespread adoption of unseen-attack scenarios for more realistic evaluation~\cite{rattani2015openset,sequeira2016realistic,ferreira2019adversarial}.

The issue of performance degradation in the presence of unseen attacks is currently the biggest challenge in face PAD. As such, some have addressed it using one-class classification or anomaly detection strategies~\cite{george2020learning,kittler2017faceanomaly,nikisins2018effectiveness, xiong2018unknown, perera2019learning,fatemifar2019combining,Perez-Cabo_2019_CVPR_Workshops} achieving meaningful breakthroughs and fostering the robustness of face PAD to unknown attacks. Others defend that the true solution lies in the use of domain adaptation and interpretability to better control model behaviour and lead PAD systems to achieve true generalising capabilities~\cite{Sequeira2020Interpretable, Sequeira2021Exploratory, phillips2020four}.

\subsection{Robustness and trustworthiness}

Faced with the current state of face recognition, some would say face recognition is a solved topic. Advances in deep learning architectures, tailored loss functions, and massive online-sourced databases have enabled the topic of face biometrics to achieve near-perfect performance metrics. This is true even for edge scenarios, on challenging datasets with significant pose, illumination, and environment variability.

However, diverse challenges remain to be solved or have surged over the recent years due to developments in face recognition or society in general. Here, we focus on two of the most pressing problems: trustworthiness and robustness. The first relates to the significant opacity of deep learning-based state-of-the-art approaches. The second is linked to the difficulty of current methodologies to recognise faces under occlusions, especially masks. Below, we delve deeper into each of these challenges.

\subsubsection{Face recognition in a masked society}

\begin{figure}[t!]
\centering
\includegraphics[width = 0.49\linewidth]{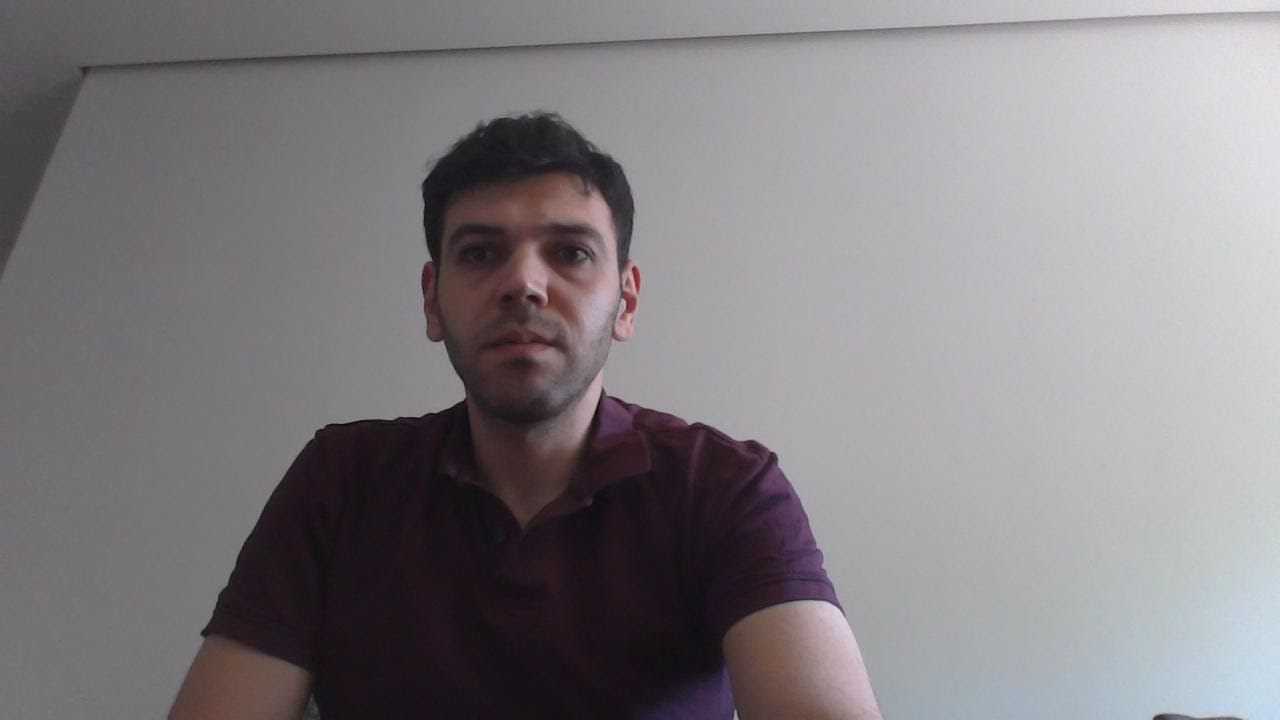}
\includegraphics[width = 0.49\linewidth]{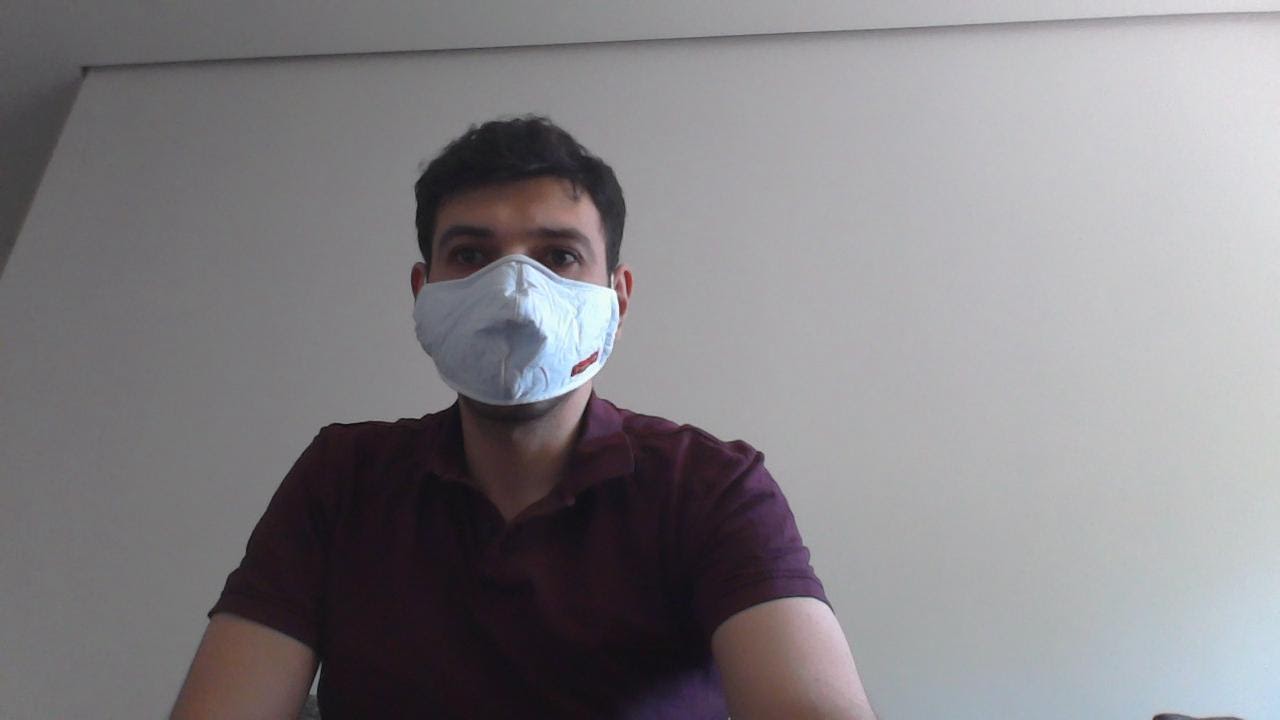} 
\caption[Example of how a mask can significantly occlude a face and limit the information that can be used by a face recognition algorithm.]{Example of how a mask can significantly occlude a face and limit the information that can be used by a face recognition algorithm (from~\cite{DamerBiosig2020}).}
\label{fig:facebiometricsprior_impact_face_masks}
\end{figure}

The ongoing Covid-19 pandemic has had a meaningful negative impact on face recognition systems~\cite{MartaCovid}. The widespread (and generally mandated) use of face masks covering the nose, mouth, and chin regions of the face has been reported to significantly degrade the accuracy of existing face recognition solutions~\cite{DamerBiosig2020,ngan2020ongoinga,ngan2020ongoingb,JEEVAN2022108308}.

Before the pandemic, research on the robustness to occlusions in face recognition was fairly common. However, it was also rather limited to small occlusions like sunglasses or scarves~\cite{opitz2016grid,song2019occlusion}, which do not typically hide as much information as a face mask (see Fig.~\ref{fig:facebiometricsprior_impact_face_masks}). One should easily understand how unprepared the existing face recognition solutions were for this new global paradigm.

Since the dawn of these challenging circumstances, multiple authors have studied in detail the effects of wearing masks on face recognition~\cite{MartaCovid}. Two of the most relevant studies were conducted by the National Institute of Standards and Technology (NIST), focusing on pre-Covid-19~\cite{ngan2020ongoinga} and post-Covid-19 algorithms~\cite{ngan2020ongoingb}. These studies were part of the ongoing 
Face Recognition Vendor Test (FRVT), an independent and thorough benchmark of face recognition solutions, and the results indicate that competitive algorithms, which fail to authenticate less than $1\%$ of probes in typical circumstances, failed up to $50\%$ more frequently in the presence of synthetic masks.

Beyond these studies, some have studied the effect of real masks in academic and commercial solutions. \citet{DamerBiosig2020} has found that the FMR100 score of the state-of-the-art method SphereFace can increase from $0.065\%$ to $27.35\%$ when evaluated with real masked face images from twenty-four participants. Wearing masks has also been reported to significantly impact verification performance of human operators~\cite{DBLP:journals/corr/abs-2103-01924}, face image quality estimation~\cite{FuMaskedQuality} and presentation attack detection performance~\cite{FANG2021108398}.

Early works on this topic included the automatic detection of face masks in images~\cite{loey2021hybrid,qin2020identifying} which, however, do not effectively solve the problem. Later, \citet{li2021cropping} proposed an attention-based method to train a model on the periocular face region, avoiding the mask areas and achieving promising results. 

Most other solutions have focused on the adaptation of existing pretrained face recognition networks, likely due to the limited amount and diversity of data currently available for masked face recognition. \citet{anwar2020masked} studied the benefits of synthesising masks on the training data. \cite{geng2020masked} augmented existing datasets with realistic masked images using a generative adversarial network (GAN) specifically tailored to retain identity information.

More recently, \citet{boutros2022self} proposed an efficient solution to be integrated on top of existing face recognition models, which attempts to unmask the embedding produced by the backbone. The unmasking module is based on a neural network trained with a self-restraining triplet loss which prioritises more affected genuine pairs.

\citet{huber2022mask_invariant} used a template-level knowledge distillation approach to approximate embeddings produced from masked and unmasked images. The teacher model is trained with ElasticFace loss, which is also used to ensure the resulting embeddings retain identity information. Similarly, \citet{li2020look} used knowledge distillation combined with image-level face completion.

Despite all these recent efforts and meaningful strides, there are still plenty of hurdles to overcome. The rise of masks in our society uncovered the feeble nature of face recognition systems in the face of extensive occlusions, and researchers are organising to face the challenge (\emph{e.g.}, through large competitions in masked face recognition~\cite{Boutros2021MFR,deng2021masked}). Nevertheless, a definitive solution should only be achievable with strong concerted efforts on improving available databases, adapting existing algorithms, and designing novel training strategies.

\subsubsection{Interpretability in face biometrics}

The growing use of increasingly sophisticated and elusive deep learning models has been sparking the need for strategies to better understand their inner workings. This is also true for several biometric applications. Namely, as face recognition systems permeate further into more critical applications such as border control or law enforcement, trustworthiness and transparency are paramount~\cite{Almeida2016,Prema2019,Abomhara2020,Furberg2017}.

In recent years, multiple studies have found that biometric systems (especially those based on face) generally present significant demographic biases which have been able to survive largely unnoticed. For example, lower false match scores among men (when compared to women) have been linked to a significantly larger variety of facial hair styles. Similarly, the variety of hairstyles, face morphology, and makeup styles have also been cited as possible causes behind gendered differences in face recognition accuracy~\cite{Albiero2020,Qiu2021,Albiero2022}.

\begin{figure}[t]
\centering
\includegraphics[width=0.9\linewidth]{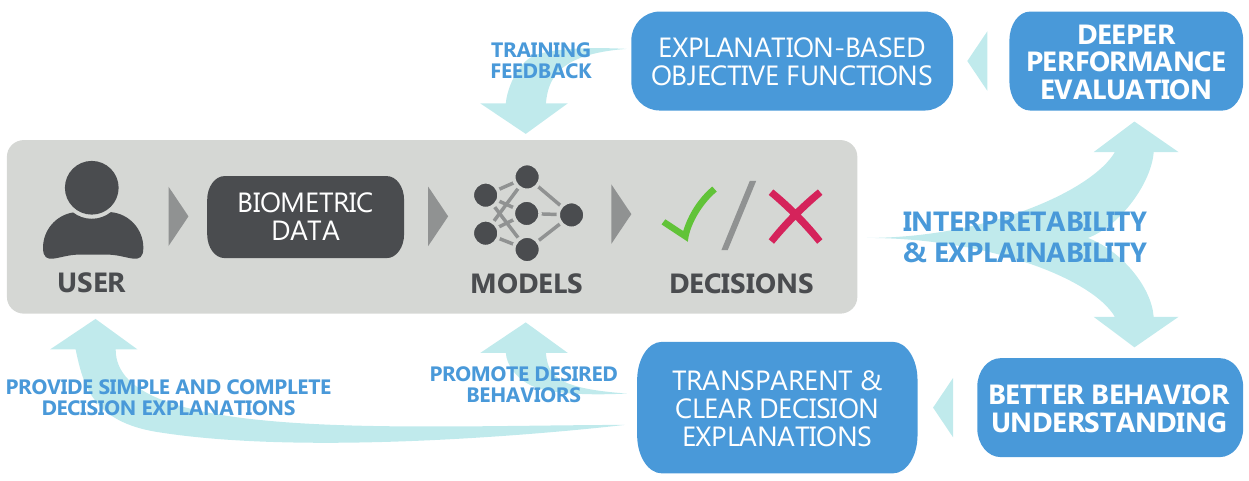}
\caption[Illustration of how interpretability/explainability can be used to understand and improve a biometric model.]{Illustration of how interpretability/explainability can be used to understand and improve a biometric model (from~\cite{Neto2022Explainable}).}
\label{fig:facebiometrics_priorart_evaluation_scheme}
\end{figure}

If a biometric model suffers from bias, traditional metrics or visualisation methods will not be able to detect it or explain that these features played a prominent role in the decision. Having more thorough mechanisms of understanding the behaviour of biometric models, for instance through interpretability, could be the key to unveiling and avoiding such biases and ensuring fair treatment at all times (see Fig.~\ref{fig:facebiometrics_priorart_evaluation_scheme}).

The topic of interpretability in biometrics is still in its early stages, although some researchers have already delved into the study of transparency for face biometrics. Some of these focused on attention mechanisms~\cite{jiang2021explainable,shao2020regularized,aghdaie2021attention}, which allow models to focus on relevant areas that can be adjusted to avoid certain undesirable behaviours.

One of the first approaches for interpretable face recognition was proposed by \citet{yin2019towards}, using feature and spatial activation diversity losses. These were used to promote, respectively, filter robustness against occlusions and the inclusion of semantic information. \citet{williford2020explainable} explored a new way to obtain explanations by combining triplets and an inpainting game. Using subtree excitation backpropagation and density-based input sampling for the explanation, model interpretability is promoted and saliency maps can be built to support explanations. \citet{liu2021heterogeneous} used adversarial training for heterogeneous face recognition, explicitly promoting the introduction of semantic and interpretable information in the model's latent space.

As for face PAD, most works so far are limited to the simple application of explainability tools (typically GradCAM), or the use of t-distributed stochastic neighbour embedding (t-SNE) as a way to interpret the way features lead to certain decisions~\cite{jourabloo2018face,pinto2020leveraging,wang2020deep,yu2020searching,fang2022learnable,wang2022disentangled}. Beyond these works, some authors proposed the application of auxiliary supervision techniques to promote interpretability \cite{liu2018learning,liu2020disentangling,yang2019face}. 

Beyond face recognition and presentation attack detection, Seibold~\etal~\cite{seibold2021focused} studied face morphing attack detection using focused layer-wise relevance propagation (FLRP) as a way to explain decisions to humans. Similarly, Xu~\etal~\cite{xu2022supervised} proposed the use of FLRP to explain decisions of their deepfake detection method. Alongside a method to simulate face aging, \citet{Genovese2019} proposed the cross-GAN filter similarity index (CGFSI) that can be used to explain the behaviour of face GANs.

Despite these first works on interpretability for face biometrics, there is still a long way to go. The effect of causality on the production of explanations and the possibility for multiple simultaneous explanations are still open topics. Likewise, the production of semantic and textual explanations (beyond simply visual ones) is yet to be addressed. All of these issues remark the relevance of calling for transparency in face biometrics and adopting the new technologies in interpretability for this topic.

\section{Open Challenges and Opportunities}

As discussed throughout this chapter, face recognition is an intensively and thoroughly researched topic with praised results and a significant impact in the real world. Through the study of increasingly sophisticated deep learning methodologies and the design of tailored objective functions, face recognition was able to conquer a prominent place in our society with robust and reliable commercial solutions.

Nevertheless, it is also clear that some problems remain largely unsolved. Nowadays, those are mainly related to robustness in masked face recognition scenarios, the trustworthiness of sophisticated deep learning-based solutions, and the detection of presentation attacks using unseen species. These are the topics which currently pose the greatest threats to face recognition applications and, thus, the most promising research opportunities.

Hence, this thesis part focuses on two contributions to these open challenges and opportunities. Specifically:
\begin{itemize}
    \item In Chapter~\ref{ch:maskedFaceRecog}, we propose two methodologies based on triplet and contrastive learning strategies, combined with ArcFace and mean squared error losses to promote similarity between masked and unmasked image embeddings and close the performance gap on masked face recognition;
    \item In Chapter~\ref{ch:interpFacePAD}, we study the use of interpretability to better understand the decisions of deep learning models in face PAD, in order to motivate the broader application of interpretability and explainability for more transparent and trustworthy biometrics.
\end{itemize}
\chapter{Masked Face Recognition}~\label{ch:maskedFaceRecog}

\begin{tcolorbox}\footnotesize
{\large\bf Foreword on Author Contributions}

The research work described in this chapter was conducted in collaboration with Pedro C. Neto, Fadi Boutros, Mohsen Saffari, and Naser Damer, under the supervision of Jaime S. Cardoso and Ana F. Sequeira. The author of this thesis contributed to this work on the conceptualisation of the training strategies, the discussion and comparison of the results, and the preparation of the scientific publications.

The results of this work have been disseminated in three articles in international conference proceedings:
\begin{itemize}[noitemsep, leftmargin=1em, nosep]
    \item P. C. Neto, F. Boutros, \underline{J. R. Pinto}, N. Damer, A. F. Sequeira, J. S. Cardoso, ``FocusFace: Multi-task Contrastive Learning for Masked Face Recognition,'' in \emph{Workshop on Face and Gesture Analysis for COVID-19 (FG4COVID19)}, Dec.~2021.~\cite{Neto2021Focus}
    
    \item P. C. Neto, F. Boutros, \underline{J. R. Pinto}, M. Saffari, N. Damer, A. F. Sequeira, and J. S. Cardoso, ``My Eyes Are Up Here: Promoting Focus on Uncovered Regions in Masked Face Recognition,'' in \emph{International Conference of the Biometrics Special Interest Group (BIOSIG 2021)}, Sep.~2021.~\cite{Neto2021Eyes}
    
    \item F. Boutros, N. Damer, J. Kolf, K. Raja, F. Kirchbuchner, R. Ramachandra, A. Kuijper, P. Fang, C. Zhang, F. Wang, D. M. Martin, N. Aginako, B. Sierra, M. Nieto, M. E. Erakin, U. Demir, H. Ekenel, A. Kataoka, K. Ichikawa, S. Kubo, J. Zhang, M. He, D. Han, S. Shan, K. Grm, V. Struc, S. Seneviratne, N. Kasthuriarachchi, S. Rasnayaka, P. C. Neto, A. F. Sequeira, \underline{J. R. Pinto}, M. Saffari, and J. S. Cardoso, ``MFR 2021: Masked Face Recognition Competition,'' in \emph{International Joint Conference on Biometrics (IJCB 2021)}, Aug.~2021.~\cite{Boutros2021MFR}
\end{itemize}

\end{tcolorbox}

\section{Context and Motivation}

Face recognition is one of the most advanced biometric traits. The adoption of sophisticated deep learning architectures and tailored loss functions led to the achievement of very high accuracies. This enabled the widespread adoption of face recognition solutions, from automated border control to personal biometric applications~\cite{gorodnichy2014automated,lovisotto2017mobile}.

However, as detailed in Chapter~\ref{ch:faceprior}, the Covid-19 pandemic and the global mask mandates created yet another hurdle to face recognition solutions. The occlusion of information by the use of a mask has been reported to result in meaningful performance degradation in state-of-the-art face recognition solutions~\cite{DamerBiosig2020,ngan2020ongoinga,ngan2020ongoingb,JEEVAN2022108308}.

Recent works have focused mainly on training existing architectures or fine-tuning pretrained models with relatively small datasets of synthetic masked images~\cite{anwar2020masked,geng2020masked}. Some have proposed more sophisticated approaches for avoiding information that could be occluded by a face mask~\cite{li2020look,huber2022mask_invariant,boutros2022self}. However, it is clear that further efforts are needed in the improvement of available databases, the adaptation of existing algorithms, and the design of novel training strategies in order to close the performance gap.

This chapter presents a work partially framed within the Masked Face Recognition (MFR 2021) challenge hosted by the \emph{2021 International Joint Conference on Biometrics (IJCB)}. Here, we propose two learning strategies based on the combination of triplet and contrastive losses with mean squared error loss to promote both identity information capture and similarity between embeddings of masked and unmasked images. With this, we aimed to adapt face recognition state-of-the-art deep learning methodologies to close the performance gap between face recognition in generic unconstrained scenarios and in the presence of face masks.

\section{Methodology}

Two approaches were explored to close the performance gap between unmasked and masked face recognition. The first one is an adaptation of the triplet loss training strategy~\cite{Neto2021Eyes}. The second one is a multi-task contrastive learning combination of cross-entropy (CE), ArcFace, and mean squared error (MSE) objective functions~\cite{Neto2021Focus}. Both aim to influence the behaviour of the models through the respective loss functions by leading them to steer clear of information that can be occluded by the use of face masks.

\subsection{Adapted triplet loss}

\begin{figure}[t]
   \centering
    \includegraphics[width=0.8\linewidth]{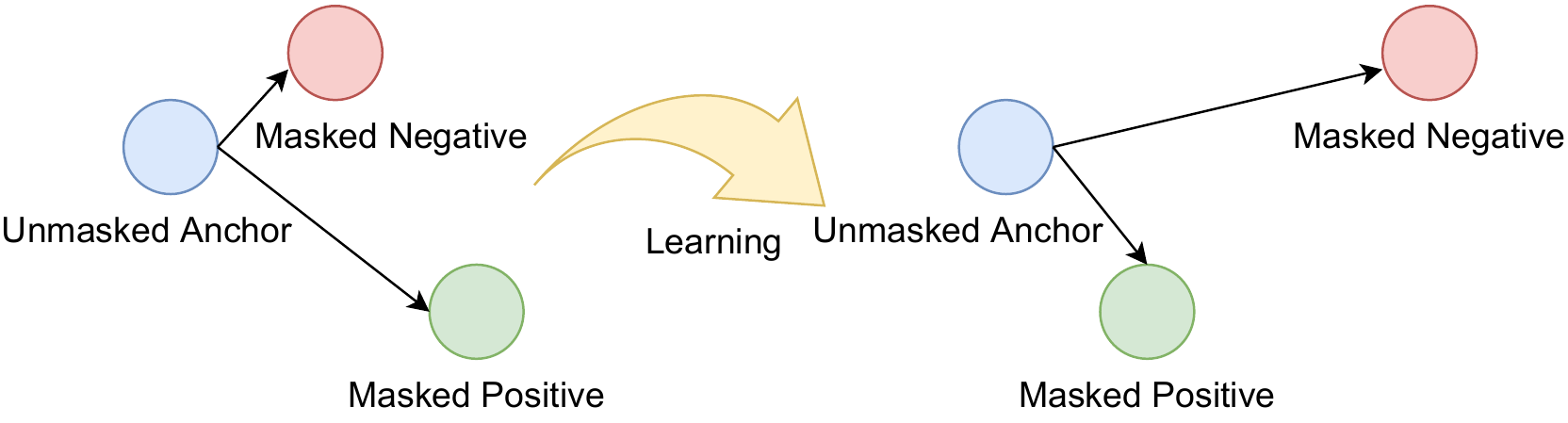}
  \caption{Expected effect of the original triplet loss on the output embedding space.}
  \label{fig:maskedfacerecog_triplet_objective} 
\end{figure}

\begin{figure}[t]
\centering
   \includegraphics[width=0.85\linewidth]{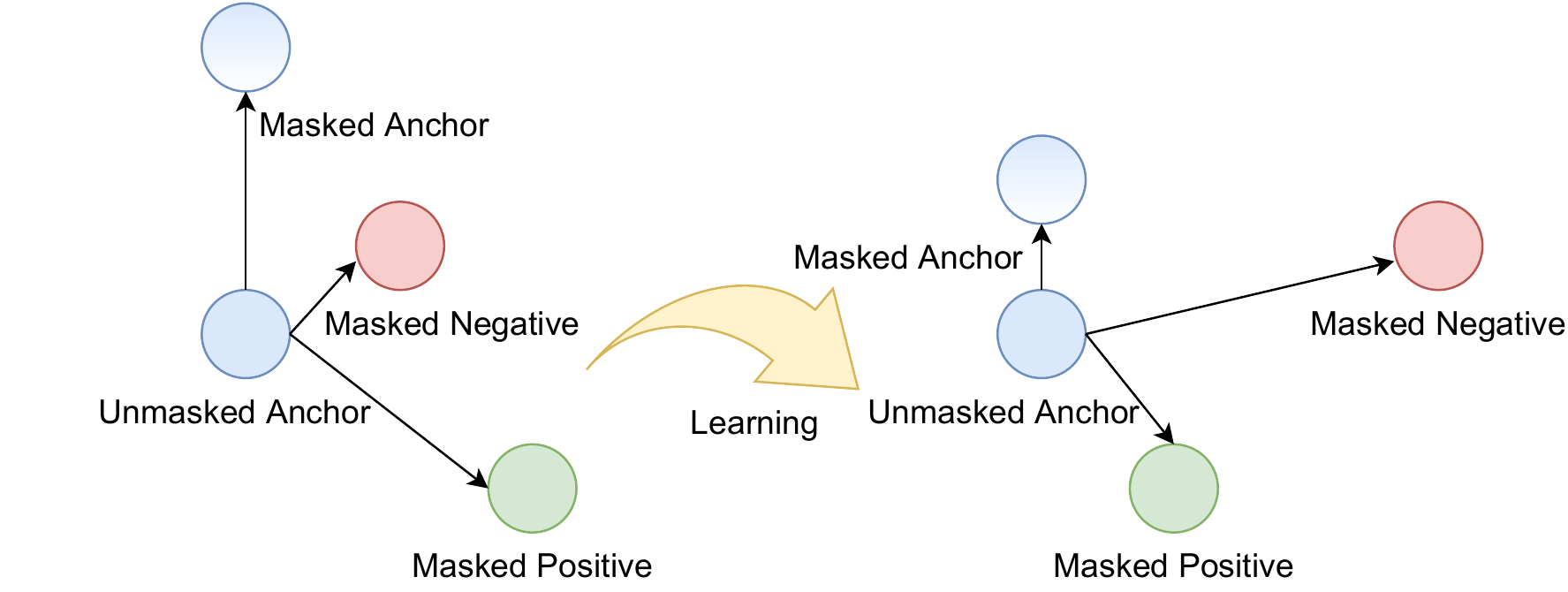}
  \caption{Expected effect of the proposed adapted triplet loss on the output embedding space.}
  \label{fig:maskedfacerecog_triplet_objective_new} 
\end{figure}

The triplet loss~\cite{Chechik2010} has been frequently used in general pattern recognition tasks, including the particular case of face recognition models. Instead of instance-based learning, this methodology organises training data into triplets composed of an anchor ($x_A$, which serves as a reference), a positive ($x_P$, which shares the same identity as the anchor), and a negative sample ($x_N$, of a different identity). Each input corresponds to a model embedding representation in the learned output space: $y_A$, $y_P$, and $y_N$, respectively. With these, the triplet loss follows the equation:
\begin{equation}
    l = \max(0, \alpha - d(y_A, y_N) + d(y_A, y_P)),
\end{equation}
where $d(a, b)$ denotes the Euclidean distance between two embeddings $a$ and $b$, and $\alpha$ is a tunable margin parameter. By optimising towards the minimisation of this loss, the model is effectively led to bring closer the representations $y$ which share the same identities and draw apart those that do not, in the learned output space (see Fig.~\ref{fig:maskedfacerecog_triplet_objective}).

The triplet loss will lead to embeddings of the same identity being clustered together in the output space. However, they are still allowed some variability as long as it verifies the $\alpha$ margin to the negative samples. In the scenario of masked face recognition, adding a mask to an image can be enough to result in a large difference in the embedding within the margin allowance. This is, however, undesirable. A method that is truly robust to face masks should be able to completely ignore masks and, ideally, offer the exact same outputs for a face image and the exact same image but with a face mask. Only thus do we have complete certainty that the model is robust to such occlusions.

As such, we adapt the triplet loss learning strategy to promote such behaviour. Alongside the typical members of a triplet ($y_A$, $y_P$, and $y_N$), the model also receives a version of the anchor with a synthetically added face mask ($y_{A_m}$) covering the mouth, chin, and nose according to health and safety guidelines. The mean squared error (MSE) between $y_A$ and $y_{A_m}$ is added to the aforementioned triplet loss formulation to further promote the similarity between the anchor and the masked anchor:
\begin{equation}
    l_{adapted} = \max(0, \alpha - d(y_A, y_N) + d(y_A, y_P)) + MSE(y_{A_m}, y_A).
\end{equation}
With this, we aim to lead the trained model to avoid the regions of the face that are commonly occluded by masks and thus retain performance levels when performing inference on masked face images (see Fig.~\ref{fig:maskedfacerecog_triplet_objective_new}).

\subsection{Multi-task contrastive learning}

\begin{figure}[!t]
\centering
\includegraphics[width=0.75\textwidth]{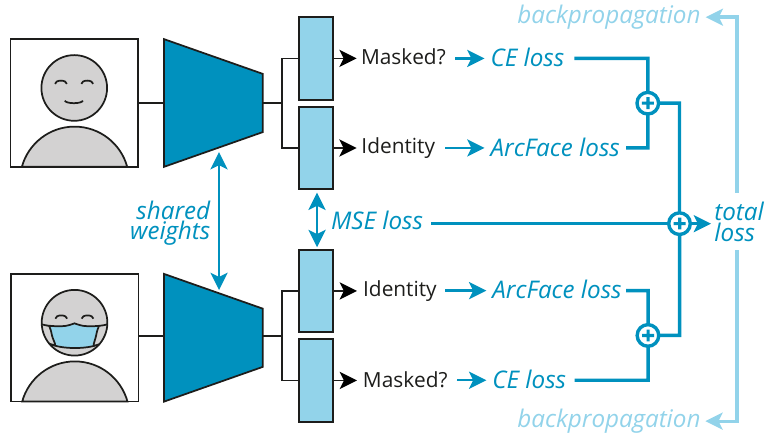}
\caption{Schema of the proposed multi-task contrastive learning approach.}
\label{fig:maskedfacerecog_architecture_scheme}
\end{figure}

Building upon the idea of the adapted triplet loss, we propose a multi-task approach for masked face recognition combining multiple objective functions. The proposed methodology, illustrated in Fig.~\ref{fig:maskedfacerecog_architecture_scheme}, receives pairs of masked and unmasked images and is composed of two symbiotic parts: one makes the network aware if a face mask exists in the input image, and the other uses mask awareness for higher stability in identity learning among masked and unmasked images.

The first part consists of the detection of masks in the input image. Embedding features from a common backbone (which can be a pretrained face recognition network) are delivered to a fully-connected module which performs binary classification (masked or unmasked). This module is optimised by minimising the cross-entropy objective function:
\begin{equation}
    L_{CE} = -\frac{1}{N}\sum_{i\in N}{\log\frac{\mathrm{e}^{y_{it}}}{\sum_{k\in n}\mathrm{e}^{y_{ik}}}},
\label{eq:ce}
\end{equation}
where $N$ represents the number of samples in the mini-batch, $n$ is the class number, and $y_{it}$ represents the output of the module for the sample $i$ and target class $t$. This strategy leads the model to become aware of the presence of a mask in the input image and to include this information in the output embedding.

The second part focuses on the identity recognition task. For improved results, we take the ArcFace approach~\cite{deng2019arcface}, widely considered the state-of-the-art in face recognition. ArcFace learns identity by minimising the following loss function:
\begin{equation}
    L_{arc} = -\frac{1}{N}\sum^{N}_{i=1}\log\frac{\mathrm{e}^{s\left(\cos\left(\theta_{it} + m\right)\right)}}{\mathrm{e}^{s\left(\cos\left(\theta_{it} + m\right)\right)} + \sum^{n}_{j=1, j \neq t}\mathrm{e}^{s\cos\theta_{ij}}},
\label{eq:arc_face}
\end{equation}
where $m$ represents the embedding distance margin, $s$ denotes the scale, and $\theta_{it}$ the angle between features ($x_i$) and weights ($W_t$) for sample $i$ and target class $t$. Both $m$ and $s$ are tunable hyperparameters of the loss. ArcFace loss explicitly promotes higher intraclass similarity and diversity among inter-class samples. Also, thanks to the $l_2$ normalisation of both the weights and the feature vector, the loss becomes equal to the geodesic distance margin penalty in a normalised hypersphere.

The global loss is the combination of ArcFace and CE losses computed for both masked and unmasked inputs as well as the MSE between the embeddings of each input. As with the aforedescribed adapted triplet loss approach, the MSE is intended to reinforce the similarity between the latent representations of masked and unmasked inputs throughout the training process, thus further promoting robustness to this kind of face occlusions.

\section{Experiments and Results}

\subsection{Adapted triplet loss}

\subsubsection{Experimental setup}

The development of the adapted triplet loss methodology was partly motivated by the Masked Face Recognition (MFR) competition hosted at the 2021 International Joint Conference on Biometrics (IJCB). As such, the developed methodology is mainly evaluated using the official MFR dataset (MFRC-21)~\cite{Boutros2021MFR}. This dataset is composed of face images of 47 subjects acquired over three different days. The first day is considered the reference, while the second and third days are considered probe sessions. Each session encompassed the acquisition of three videos, two with face masks and one without, using a webcam. In total, 470 non-masked images and 940 masked images are available as references, and 940 non-masked images and 1880 masked images are available as probes.

The training methodology is applied to a ResNet-50 backbone architecture~\cite{He2016} that is used to extract features from masked/unmasked face images. This backbone was trained using cross-entropy loss and stochastic gradient descent for 150 thousand epochs with a batch size of $400$. The initial learning rate was set as $0.1$, decreased by a factor of 10 whenever validation accuracy began to decay. After convergence, the backbone was fine-tuned either with the original triplet loss or the adapted triplet loss, with $\alpha$ empirically set to $0.2$. Triplets were randomly generated during training, without using any mining strategy, across a total of 65 thousand epochs.

Training used synthetic masked face images. The VGGFace2 dataset~\cite{Cao2018}, composed of over 3.3 million face images from more than nine thousand identities, was adapted to include masked faces. The NIST Dlib C++ toolkit~\cite{ngan2020ongoingb} is used to obtain sixty-eight face landmarks which are then used to generate masks appropriately fitted to the mouth/nose region, as detailed in~\cite{ngan2020ongoingb,king2009dlib}, allowing for some controlled variability regarding mask shape and colour. No face alignment was performed, and input images to the model were shaped $224\times224\times3$.

The trained models were evaluated in a biometric verification task in two scenarios: U-M, where the reference is an unmasked image and the probe is a masked image, and M-M, where both the reference and probe are masked face images. The performance is reported through the false non-match rates (FNMR), the FMR100 and FMR10, which are the lowest FNMR for a false match rate (FMR) $< 1.0\%$ and $< 10.0\%$, respectively. Additionally, the equal error rate (EER) and the area under the receiver operating characteristic curve (AUC) results are also reported. The genuine mean (GMean) and impostors mean (IMean) scores, which represent the mean distances between the mated and non-mated embedding pairs, were also computed.

\subsubsection{Results and discussion}

The proposed adapted triplet loss methodology was evaluated on two distinct datasets: one with synthetic masks (SMFD) and the other with real masked face images (RMFD). A detailed step-wise ablation study is used to understand the behaviour of the models and the impact of the proposed triplet loss adaptation in masked face recognition.

\begin{table}
\caption{Results with the adapted triplet loss on synthetic masked face data (SMFD).}
\label{tab:maskedfacerecog_res-smfd1}
\centering
\begin{tabular}{lcccccc}
\toprule
\textbf{Method}     & \textbf{GMean} & \textbf{IMean} & \textbf{AUC} & \textbf{EER} & \textbf{FMR100} & \textbf{FMR10} \\ 
\midrule
 VGG Face~\cite{Cao2018,Schroff2015}  & 0.505 & 0.325 & 0.951 & 11.8\% & 38.2\% & 13.5\% \\
 CE Loss & 0.528 & 0.426 & 0.941 & 13.2\% & 38.5\% & 21.5\%\\
 CE + TL & 0.601 & 0.320 & 0.977 & 7.8\% & 28.9\% & 11.9\%\\
 CE + Adapted TL   & 0.596 & 0.319 & \textbf{0.985} & \textbf{6.2\%} & \textbf{18.5\%} & \textbf{4.1\%} \\
\bottomrule
\end{tabular}
\end{table}

Table~\ref{tab:maskedfacerecog_res-smfd1} presents the results on SMFD data. The proposed methodology, combining cross-entropy pretraining, triplet loss, and MSE loss offers the best results across all performance metrics. As visible in the referenced table, the triplet loss training enables considerable performance gains over the CE baseline, outperforming the literature VGG Face model~\cite{Cao2018, Schroff2015} trained on unmasked face images. However, it is the MSE loss between masked and unmasked face images that allows the model to achieve promising results, especially in FMR100 and FMR10.

\begin{table}
 \caption{Results with the adapted triplet loss on real masked face data (RMFD).}
\label{tab:maskedfacerecog_res-rmfd1}
\centering
\begin{tabular}{lccccccc}
\toprule
\textbf{Method}   & \textbf{Mode}  & \textbf{GMean} & \textbf{IMean} & \textbf{AUC} & \textbf{EER} & \textbf{FMR100} & \textbf{FMR10} \\ 
\midrule
 VGG Face~\cite{Cao2018,Schroff2015} & \makecell{U-M\\M-M}& \makecell{0.523\\0.616} & \makecell{0.426\\0.461}  &  \makecell{0.769\\0.847}  & \makecell{29.419\%\\23.552\%}  & \makecell{90.587\%\\68.979\%} & \makecell{58.959\%\\38.159\%}   \\
 \cline{1-8}
 CE Loss & \makecell{U-M\\M-M} & \makecell{0.610\\0.702} & \makecell{0.475\\0.503} & \makecell{0.931\\0.936}& \makecell{11.687\%\\\textbf{9.002\%}} & \makecell{32.041\%\\\textbf{16.628\%}} & \makecell{12.852\%\\\textbf{8.791\%}}\\
 \cline{1-8}
 CE + TL& \makecell{U-M\\M-M} & \makecell{0.647\\0.699} & \makecell{0.396\\0.414}& \makecell{0.943\%\\0.945\%}& \makecell{11.213\\10.806}  & \makecell{34.744\%\\26.457\%} & \makecell{11.874\%\\11.249\%}\\
 \cline{1-8}
 CE + Adapted TL & \makecell{U-M\\M-M}  & \makecell{0.649\\0.699} & \makecell{0.383\\0.390}& \makecell{\textbf{0.957}\\\textbf{0.959}}& \makecell{\textbf{9.799\%}\\9.292\%} & \makecell{\textbf{28.252\%}\\23,507\%} & \makecell{\textbf{9.678\%}\\9.035\%}  \\
\bottomrule
\end{tabular}
\end{table}

\begin{figure}[t]
    \centering
    \includegraphics[width=0.16\linewidth, height=0.16\linewidth]{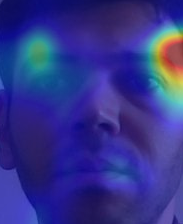}
    \includegraphics[width=0.16\linewidth, height=0.16\linewidth]{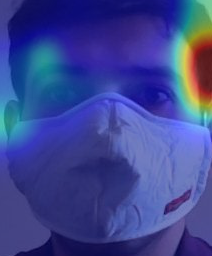}
    \includegraphics[width=0.16\linewidth, height=0.16\linewidth]{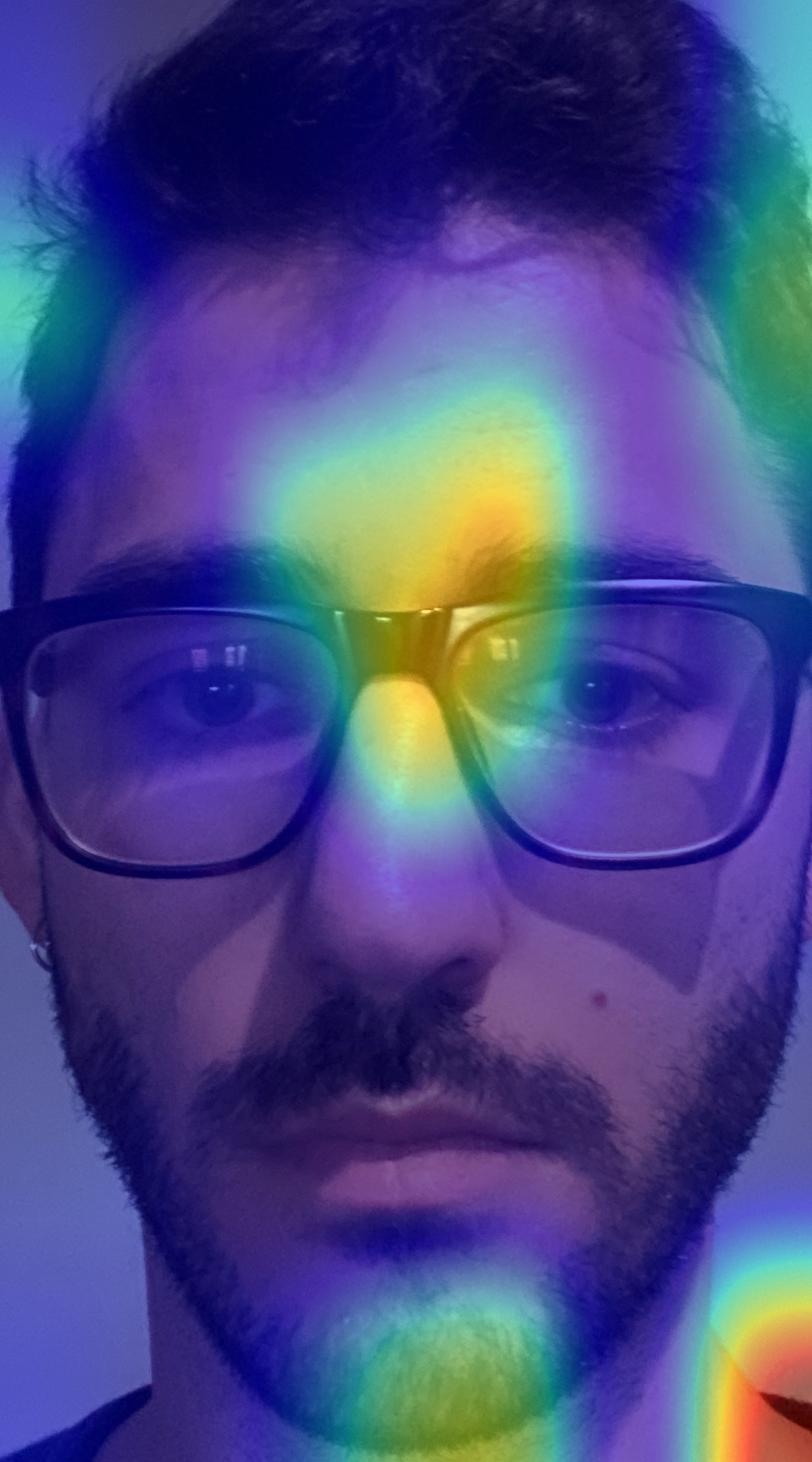}
    \includegraphics[width=0.16\linewidth, height=0.16\linewidth]{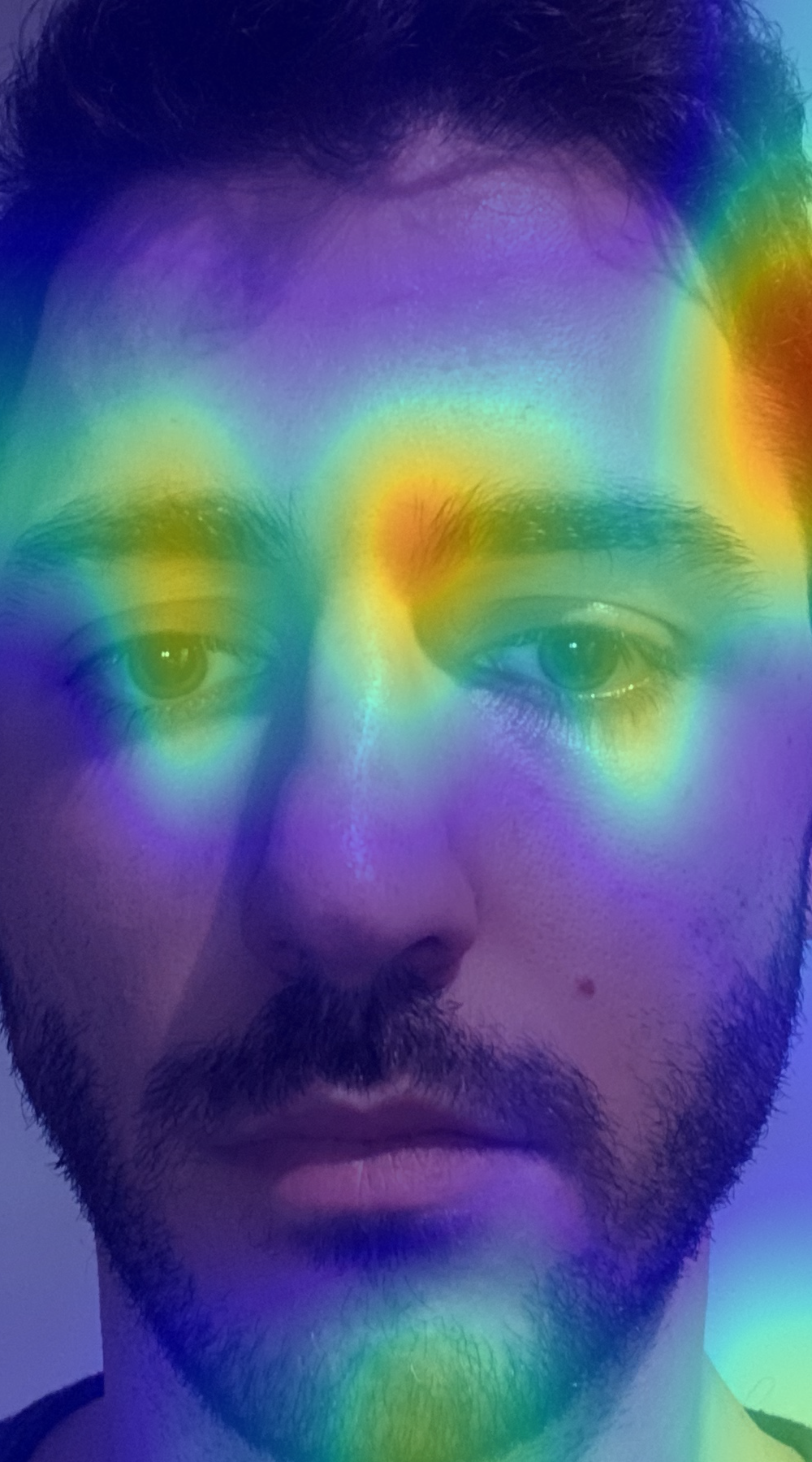}
    \includegraphics[width=0.16\linewidth, height=0.16\linewidth]{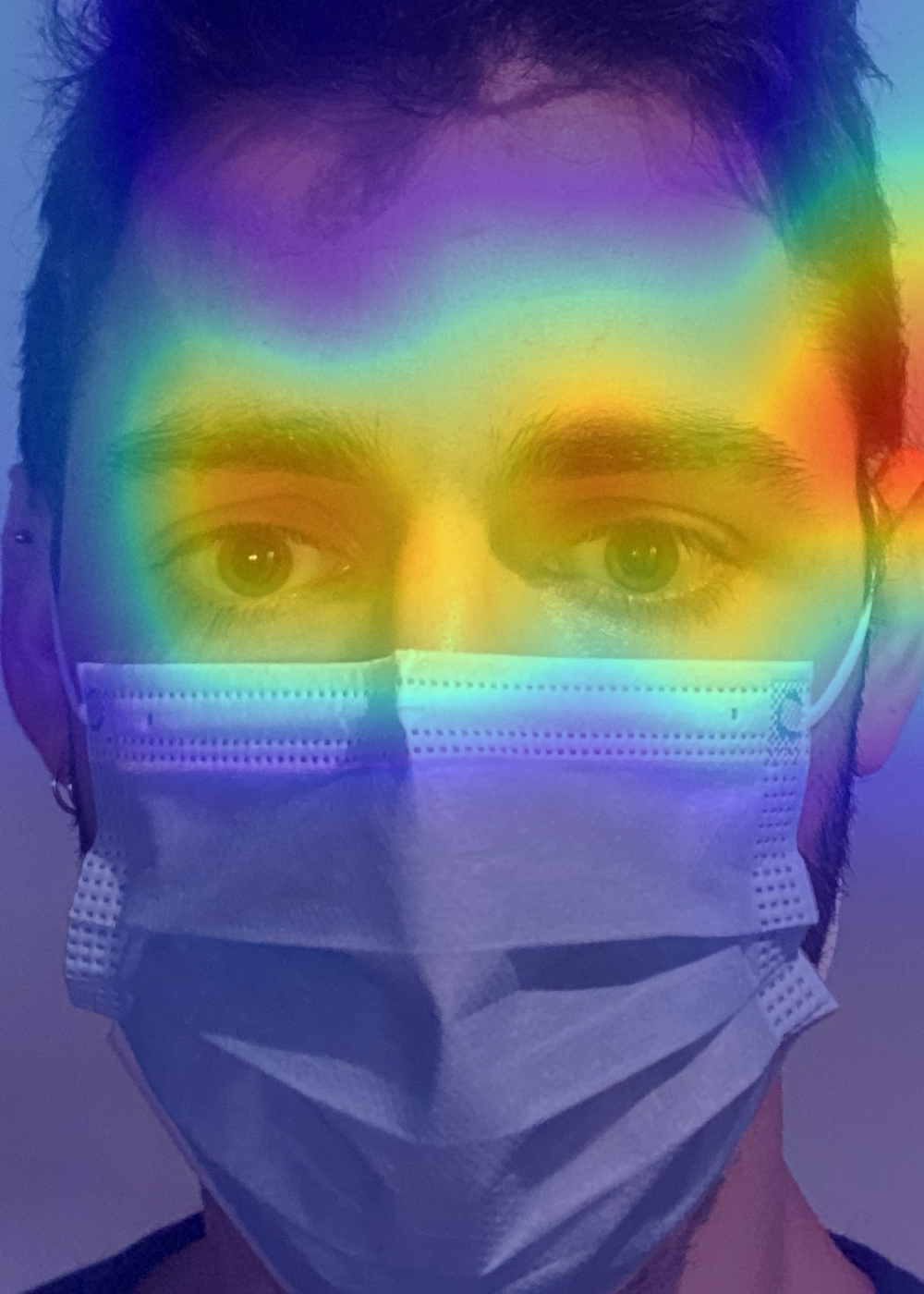}
    \includegraphics[width=0.16\linewidth, height=0.16\linewidth]{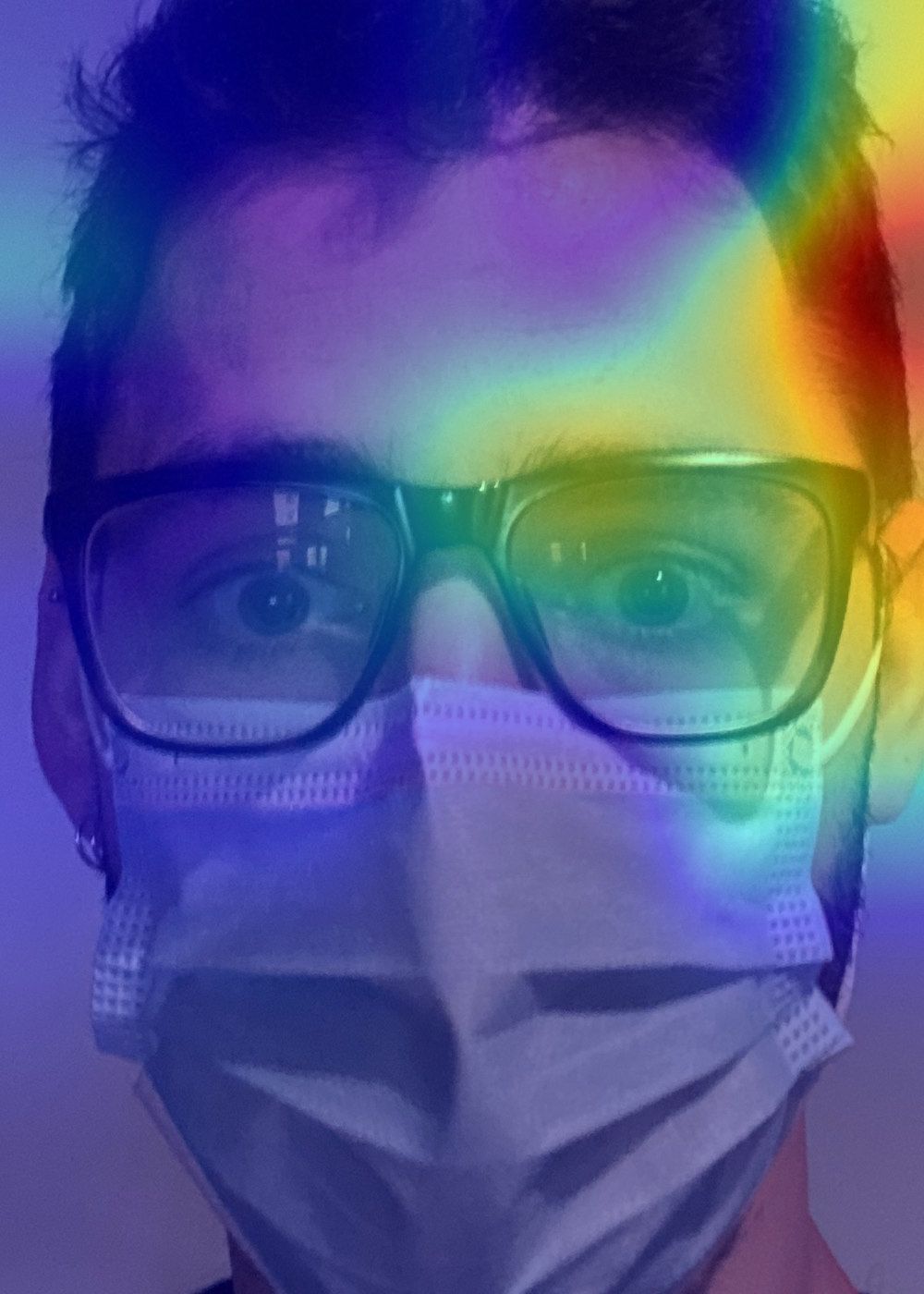}\\
  
    \includegraphics[width=0.16\linewidth, height=0.16\linewidth]{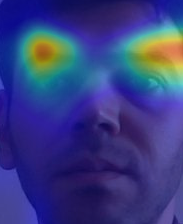}
    \includegraphics[width=0.16\linewidth, height=0.16\linewidth]{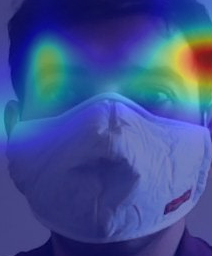}
    \includegraphics[width=0.16\linewidth, height=0.16\linewidth]{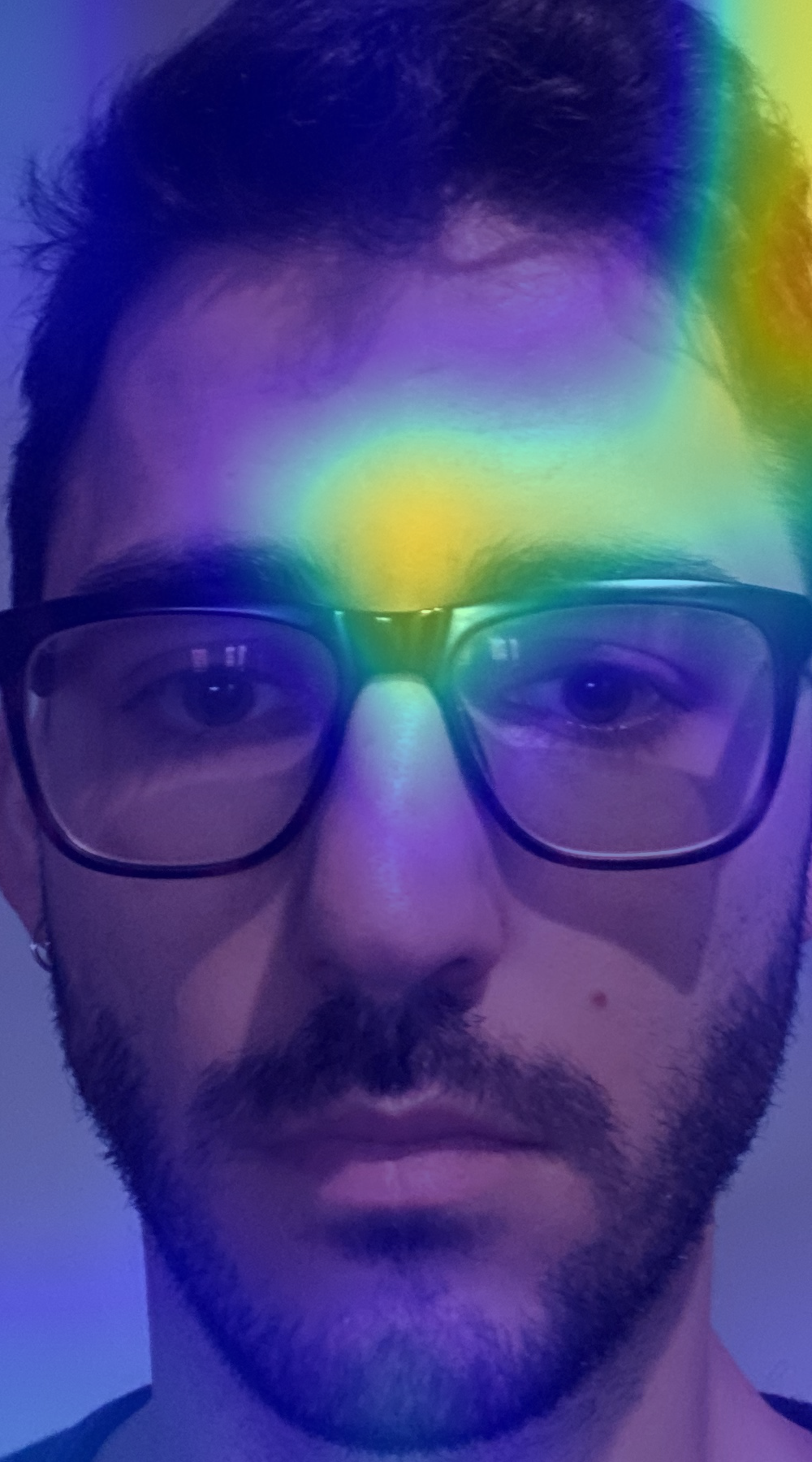}
    \includegraphics[width=0.16\linewidth, height=0.16\linewidth]{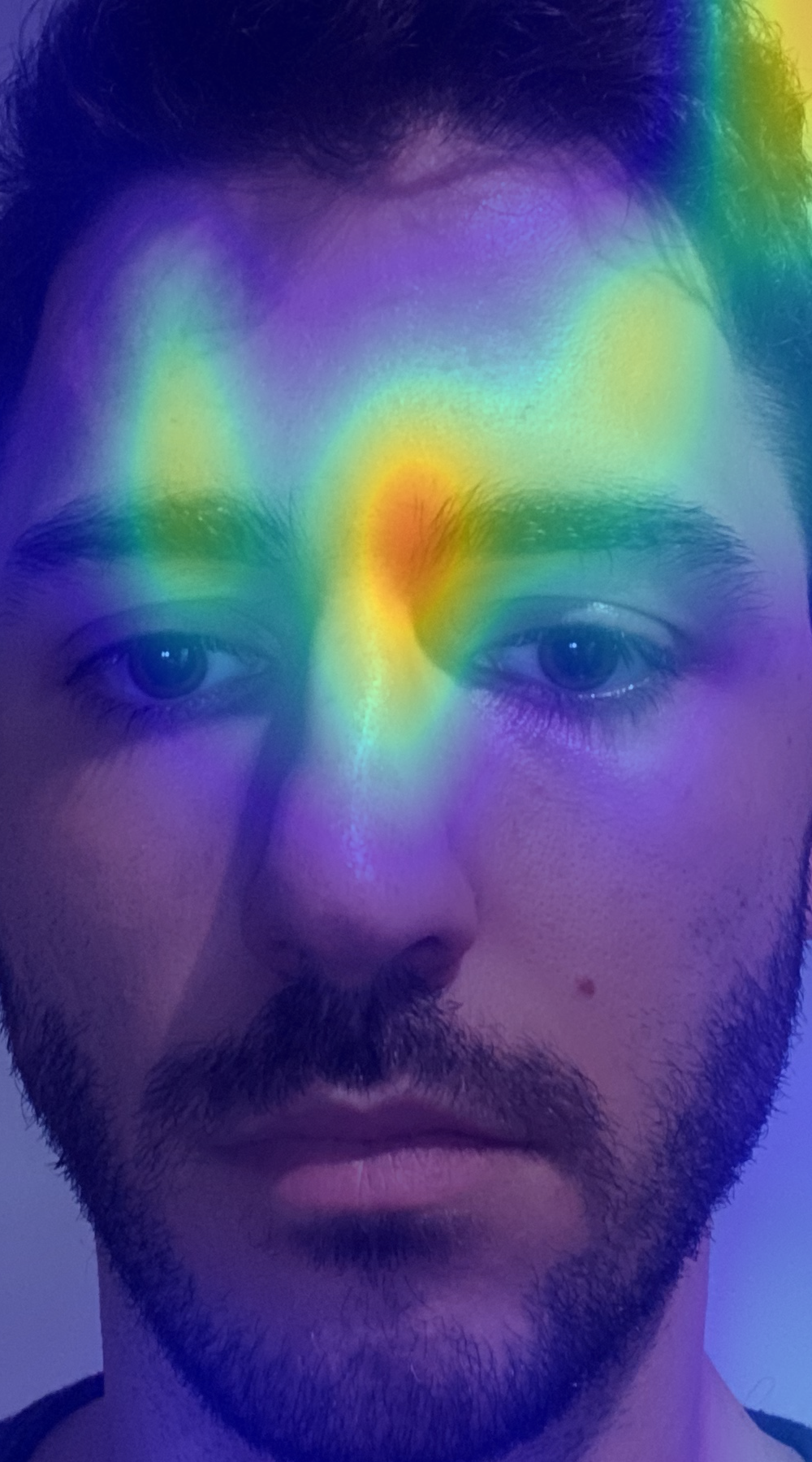}
    \includegraphics[width=0.16\linewidth, height=0.16\linewidth]{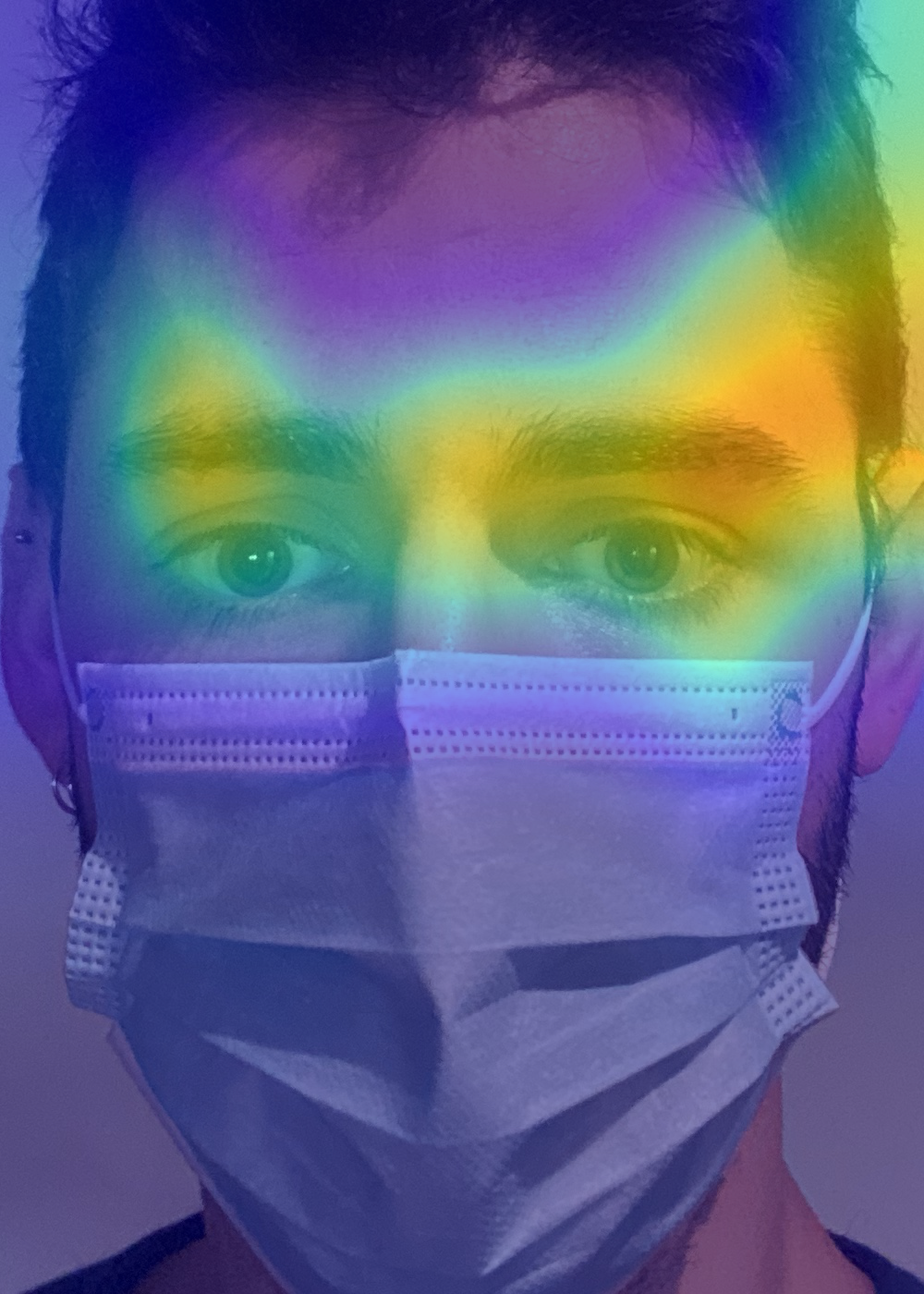}
    \includegraphics[width=0.16\linewidth, height=0.16\linewidth]{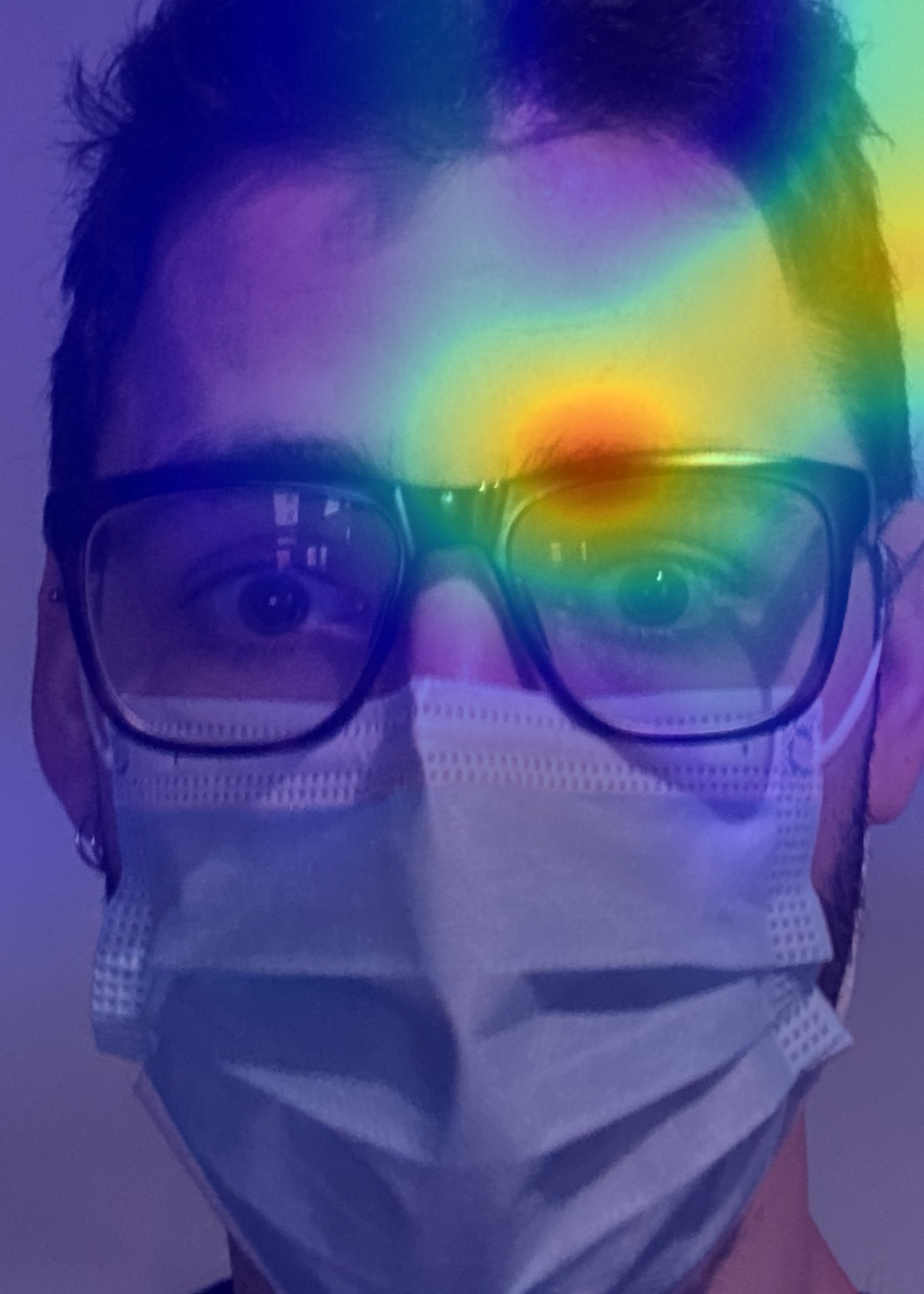}\\
    
    \includegraphics[width=0.16\linewidth, height=0.16\linewidth]{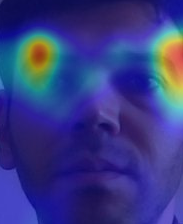}
    \includegraphics[width=0.16\linewidth, height=0.16\linewidth]{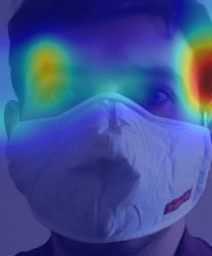}
    \includegraphics[width=0.16\linewidth, height=0.16\linewidth]{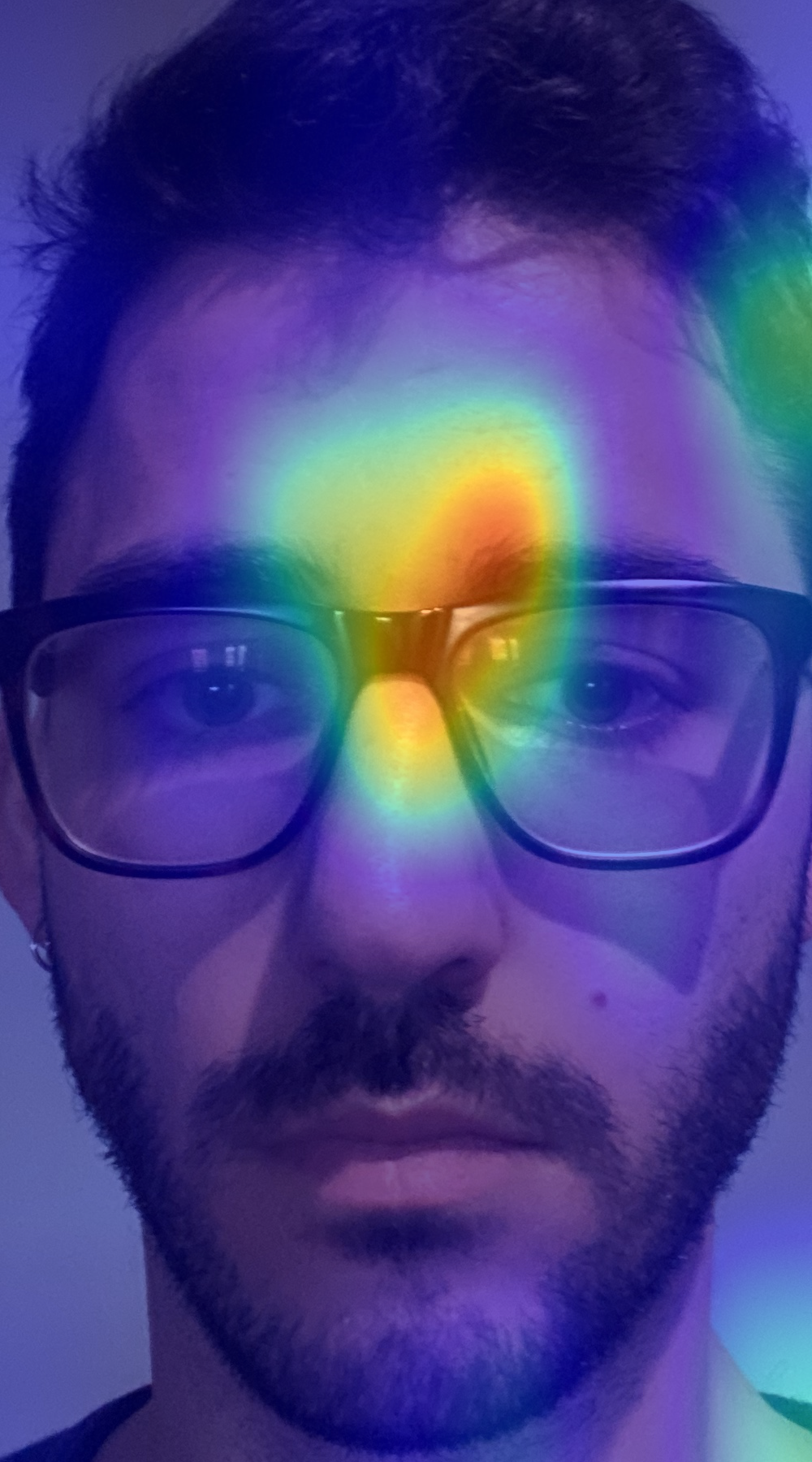}
    \includegraphics[width=0.16\linewidth, height=0.16\linewidth]{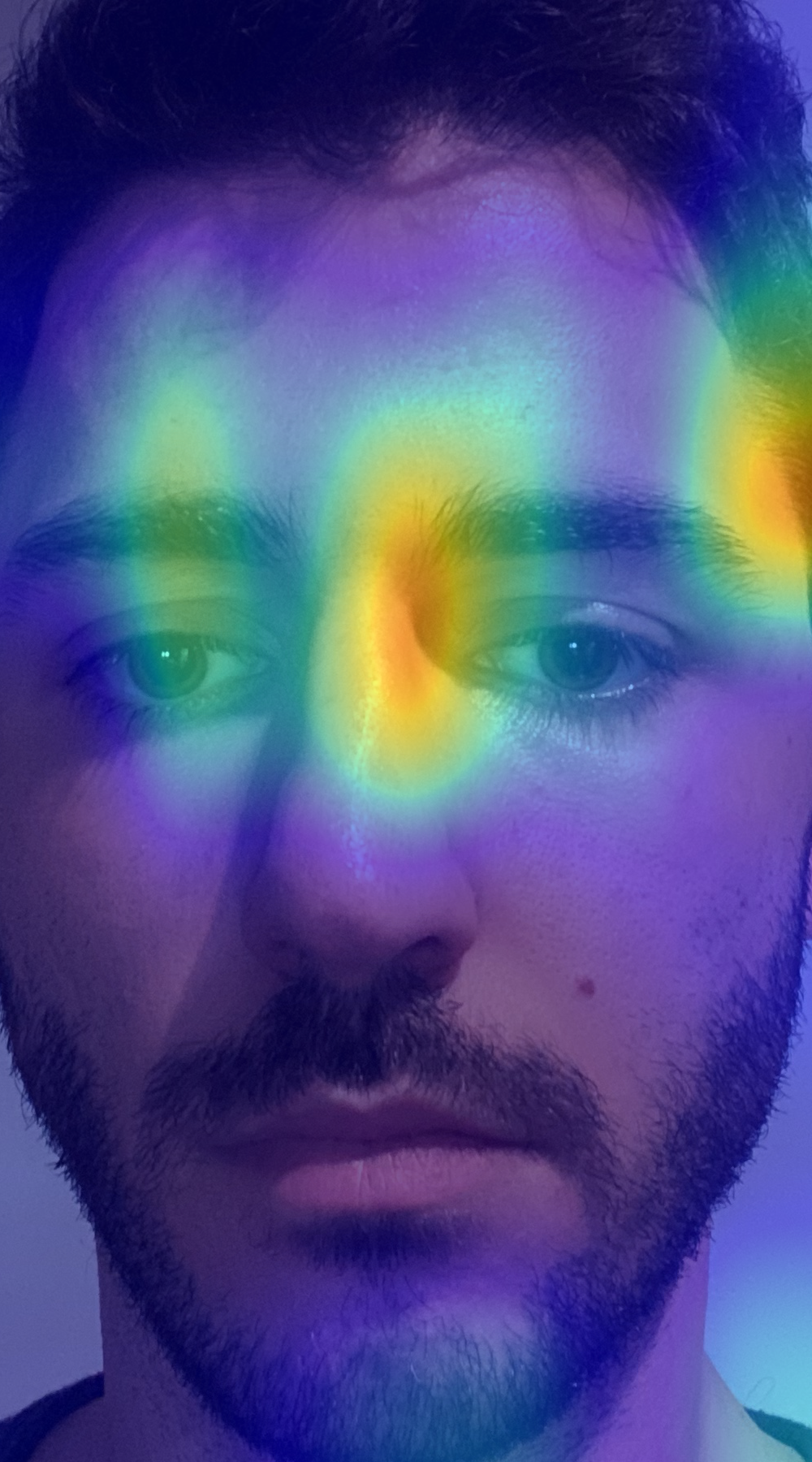}
    \includegraphics[width=0.16\linewidth, height=0.16\linewidth]{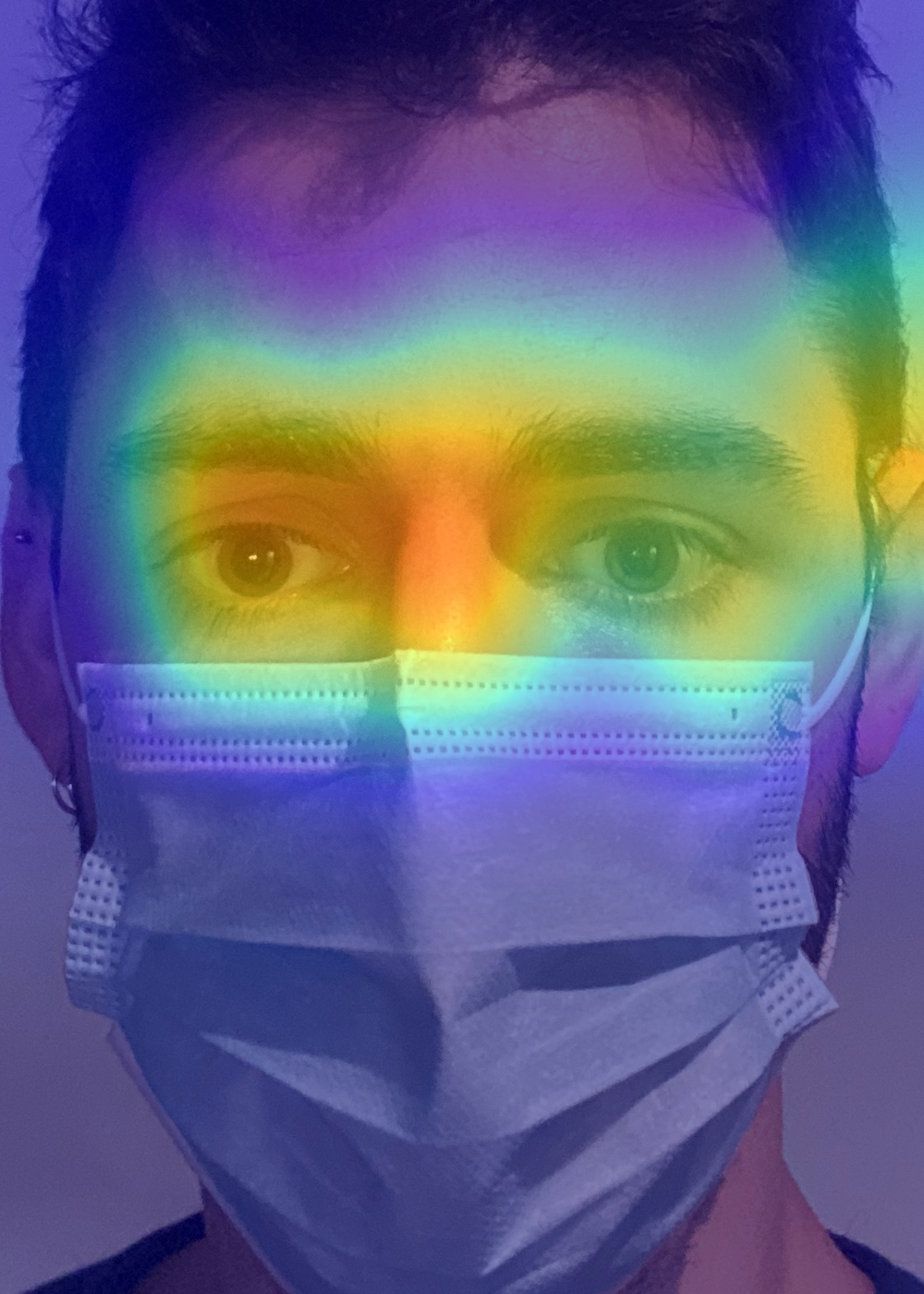}
    \includegraphics[width=0.16\linewidth, height=0.16\linewidth]{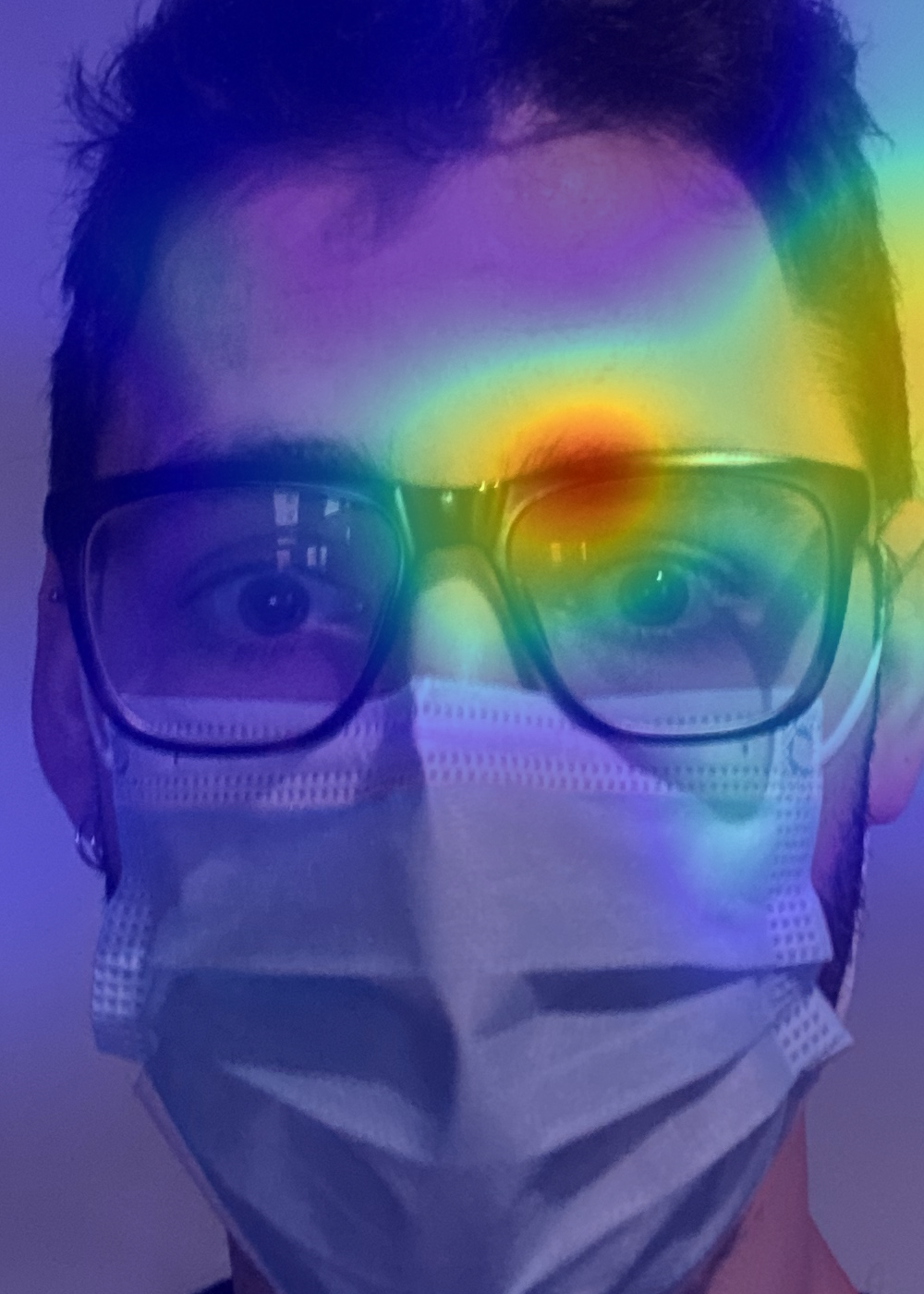}\\
       
  \caption[Explanations obtained for each trained model with the Smooth Grad-CAM++ explainability tool.]{Explanations obtained for each trained model with the Smooth Grad-CAM++ explainability tool (top row images were computed from the cross-entropy model - CE; middle row images were computed from the triplet loss model - CE + TL; bottom row images were computed from the adapted triplet loss model - CE + Adapted TL).}
    \label{fig:maskedfacerecog_grad_cam} 
\end{figure}

The results on real masked data are presented in Table~\ref{tab:maskedfacerecog_res-rmfd1}, including experiments on the U-M and M-M scenarios. It can be observed that, in general, the performance results are considerably inferior to those on SMFD, likely due to the models being trained with synthetic masks and tested with real masks. However, the proposed adapted triplet loss attained superior performance across all metrics.

It is noteworthy that M-M results, in general, do not differ considerably or consistently from U-M ones. This could be a result of the optimisation process of the adapted triplet loss, which leads the model to minimise the distance between masked and unmasked face embedding pairs. However, it is expected that some applications, such as border control, could consist of the comparison of masked probes (face images captured \emph{in loco}) with unmasked references (passport photographs).

Alongside the quantitative results presented above, the behaviour of the models trained with the proposed and alternative approaches was also evaluated through explainability. Fig.~\ref{fig:maskedfacerecog_grad_cam} presents six example images for which the Smooth Grad-CAM++ method~\cite{Omeiza2019} was used to assess the relevance of the input pixels for the output embedding features. Relevance maps were computed for each embedding feature and then averaged.

It can be observed that the first method, trained only with cross-entropy loss, was already largely able to ignore mask regions in the figures, even when masks are not present. However, it also commonly uses the region of the chin to construct the output embeddings, which does not happen with the remaining two methods. When comparing the explanations of the triplet loss and the adapted triplet loss, it can be observed that the latter typically considers more information from wider regions of the face, thus capturing more information that could be useful for more robust decisions.

\subsection{Multi-task contrastive learning}

\subsubsection{Experimental setup}
After the proposal of the adapted triplet loss, the multi-task contrastive learning methodology aimed to be a more thorough study of masked face recognition. The multi-task contrastive learning methodology was explored for two backbone architectures, ResNet-100 and ResNet-50, both widely used in the face recognition literature~\cite{He2016}. Following the example of \citet{boutros2021elasticface}, the parameter $s$ is set to 64 for the ResNet-100 and 30 for the ResNet-50, with $m=0.5$. Training used the Stochastic Gradient Descent (SGD) optimiser with an initial learning rate of $0.1$, a momentum of $0.9$, and weight decay of $5\times10^{-4}$.

Following the example of several literature works~\cite{deng2019arcface,an2021partial,huang2020curricularface,meng2021magface}, this research used the MS1MV2 dataset~\cite{deng2019arcface}. The MS1MV2 dataset is based on the MS-Celeb-1M dataset~\cite{guo2016ms} and is composed of 5.8 million images from 85 thousand identities. In this work, this dataset has been augmented, offline, with face masks by generating one image with a face mask for each image in the original dataset.

Face masks were synthesised using the MaskTheFace open source tool~\cite{anwar2020masked}, which includes five mask options (N95, KN95, surgical, cloth, and gas masks). For each image in the MS1MV2 dataset, one of the first four mask types was randomly selected for the process of synthesising masked face images, and the MaskTheFace tool took care of correctly reshaping and rotating the mask templates to match the detected face landmarks in each image.

\begin{figure}[t!]
    \centering
       \includegraphics[width=0.24\linewidth]{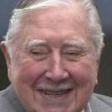}
       \includegraphics[width=0.24\linewidth]{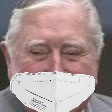}
       \includegraphics[width=0.24\linewidth]{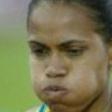}
       \includegraphics[width=0.24\linewidth]{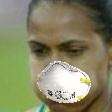}
  \caption[Example of masked face images generated to validate the multi-task contrastive learning approach.]{{Example of masked face images generated to validate the multi-task contrastive learning approach: images of the two individuals before and after mask generation. The first example is correctly masked, while the second illustrates a scenario of incorrect masking.}}
  \label{fig:maskedfacerecog_example_validset} 
\end{figure}

The described process was similarly followed for the images in the Labeled Faces in the Wild (LFW) dataset~\cite{Huang2007}, used for model validation during training (see Fig.~\ref{fig:maskedfacerecog_example_validset} for some examples). Although trained and validated on synthetic data, the method was evaluated on the MFR dataset, composed of real masked face images, just as the adapted triplet loss methodology. Also, like with the adapted triplet loss, we explore the evaluation scenarios U-M (with unmasked references and masked probes) and M-M (with masked references and probes).

\subsubsection{Results and discussion}

This section presents the results achieved by the multi-task contrastive loss approach. Comparisons focus mainly on the adapted triplet loss methodology, alternative approaches submitted to the MFR competition, and multiple ablation studies. 

\begin{table}
 \caption[Ablation results with the multi-task contrastive learning approach on the test dataset.]{Ablation results with the multi-task contrastive learning approach on the test dataset (original selection stands for the use of the masked version of the unmasked image on the second run of the network, whereas random selection denotes the process of choosing a random masked image from the same subject).}
\label{tab:maskedfacerecog_res-rmfd}
\centering
\begin{tabular}{lcccccccc}
\toprule
\textbf{Model} & \textbf{Selection} & \textbf{Mode} & \textbf{GMean} & \textbf{IMean} & \textbf{AUC} & \textbf{EER} & \textbf{FMR100} & \textbf{FMR10}\\
\midrule
 ResNet-50 & Original  & \makecell{U-M\\M-M} &\makecell{0.561\\0.625} & \makecell{0.333\\0.341}& \makecell{0.988\\0.983} & \makecell{5.377\%\\5.346\%} & \makecell{7.285\%\\6.546\%} &\makecell{4.519\%\\4.642\%} \\
 \midrule
 \makecell{ResNet-100\\(Pretrained)} & Original & \makecell{U-M\\M-M} &\makecell{0.571\\0.627} & \makecell{0.357\\0.365}& \makecell{0.987\\0.983} & \makecell{4.917\%\\5.005\%} & \makecell{5.986\%\\6.109\%} &\makecell{3.922\%\\3.882\%} \\
 \midrule
 \makecell{ResNet-100\\(Pretrained)} & Random & \makecell{U-M\\M-M} &\makecell{0.567\\0.625} & \makecell{0.354\\0.361}& \makecell{0.984\\0.984} & \makecell{5.371\%\\5.581\%} & \makecell{5.986\%\\6.184\%} &\makecell{4.991\%\\5.056\%} \\
 \midrule
 ResNet-100 & Original & \makecell{U-M\\M-M} &\makecell{0.624\\0.675} & \makecell{0.373\\0.383}& \makecell{\textbf{0.992}\\\textbf{0.992}} & \makecell{\textbf{4.594\%}\\\textbf{4.329\%}} & \makecell{\textbf{5.750\%}\\\textbf{5.509\%}} &\makecell{\textbf{2.582\%}\\\textbf{2.836\%}} \\
 \midrule
 ResNet-100  & Random & \makecell{U-M\\M-M} &\makecell{0.621\\0.671} & \makecell{0.371\\0.382}& \makecell{0.988\\0.991} & \makecell{5.164\%\\4.757\%} & \makecell{5.917\%\\5.695\%} &\makecell{4.197\%\\3.668\%} \\
 
\bottomrule
\end{tabular}
\end{table}

The conducted ablation studies explored smaller backbone models, the use of pretrained networks, and diverse image selection approaches for the contrastive learning module. According to the results presented in Table~\ref{tab:maskedfacerecog_res-rmfd}, four of the explored variants were capable of outperforming the official MFR competition baseline~\cite{Boutros2021MFR} under the U-M scenario. Two of them were capable of outperforming the baseline under the M-M scenario as well. 

The non-pretrained ResNet-100 backbone model offered the best overall performance, especially when using the original pair selection process where the masked image is the masked version of the unmasked image. Nevertheless, the results with pretrained models reveal that the proposed method can effectively make existing models able to recognise masked faces with reasonable accuracy and reduced training effort.

In fact, even though accuracy is important, time is also a key factor with such processing-heavy algorithms. Most of the MFR competition submissions using a ResNet-100 backbone required the training of more than 65 million parameters in the backbone model and up to 44 million parameters in the ArcFace layer. Using pretrained weights for the entire backbone (except for the last layer), the number of trainable parameters is reduced by approximately 47.6\% while still outperforming the baseline algorithm (see Table~\ref{tab:maskedfacerecog_module-parameters}).

\begin{table}[!t]
 \caption[Comparison of the modules of the proposed approach by the total number of parameters.]{Comparison of the modules of the proposed approach by the total number of parameters (includes information regarding the use of each module for inference).}
\label{tab:maskedfacerecog_module-parameters}
\centering
\begin{tabular}{lcc}
\toprule
\textbf{Module}  & \textbf{No. Parameters} & \textbf{Inference}  \\ 
\midrule
 ResNet-100 w/o Embedding Layer & 52~309~568 & Yes\\
 Face Recognition Embedding Layer &  12~846~080 & Yes \\
 Mask Detection Embedding Layer & 802~880 & No \\
 ArcFace Layer & 43~899~904 & No\\
 Mask Detection Fully-Connected Layer & 66 & No \\
 \midrule
 Total (Training w/o pretraining) & 109~858~498 & - \\
 Total (Training w/ pretraining) & 57~548~930 & - \\
 \midrule
 Total (Inference) & 65~155~648 & - \\
\bottomrule
\end{tabular}
\end{table}

In general, one can observe that better results were obtained when the masked image corresponds to the unmasked image (original pair selection), likely as this better allowed to reinforce in the model the behaviour of avoiding information that could be occluded by a mask. As for the U-M and M-M scenarios, performance differences were not relevant nor consistent for any of the models. As such, the proposed method is able to outperform the challenge baseline in both scenarios.

A comparison with other submissions to the MFR 2021 competition is presented in Table~\ref{tab:maskedfacerecog_res-sota}, for the U-M and M-M scenarios. The compared methods (A1\_Simple, VIPLFACE-M, and MaskedArcFace) were selected based on the competition results, the chosen loss functions (ArcFace loss), the input and feature vector sizes ($112\times112\times3$ and $512$, respectively), and the dataset used (MS1MV2). Additionally, the proposed method is also compared to the challenge baseline and the adapted triplet loss method~\cite{Neto2021Eyes}.

\begin{table}[t]
\caption[Comparison of the FMR100 results of the methods presented in the MFR competition, the official baseline, the adapted triplet loss, and the multi-task contrastive learning approach.]{Comparison of the FMR100 results of the methods presented in the MFR competition~\cite{Boutros2021MFR}, the official baseline, the adapted triplet loss, and the multi-task contrastive learning approach.}
\label{tab:maskedfacerecog_res-sota}
\centering
\begin{tabular}{lcc}
\toprule
 & \multicolumn{2}{c}{\textbf{FMR100}} \\
\textbf{Method} & \textbf{\textit{U-M}} & \textbf{\textit{M-M}}  \\ 
\midrule
 Baseline~\cite{Boutros2021MFR} &  6.009\%  & 5.925 \%   \\
 A1\_Simple~\cite{Boutros2021MFR} & \textbf{5.538\%} &  5.771\% \\
 VIPLFACE-M~\cite{Boutros2021MFR} & 5.681\% & 5.759\% \\
 MaskedArcFace~\cite{Boutros2021MFR} & 5.687\% & 5.825\%  \\
 Adapted Triplet Loss~\cite{Neto2021Eyes} & 28.252\% & 23.507\%\\
 \textbf{Multi-Task Contrastive Learning}& 5.750\% & \textbf{5.509\%}  \\
\bottomrule
\end{tabular}
\end{table}

As presented in Table~\ref{tab:maskedfacerecog_res-sota}, it is possible to see that, despite presenting very similar performance between the U-M and M-M scenarios, the proposed multi-task contrastive learning methodology is only able to outperform the selected competition submissions on the M-M scenario. However, the difference between the methods' performances is relatively small and the contrastive approach was able to consistently outperform both the official baseline and the adapted triplet loss methodology in both scenarios.

In the U-M scenario, a possible explanation for the advantage of A1\_Simple is that it uses a larger backbone model than ours, with roughly 34\% more parameters. As for VIPLFACE-M, the proponents used a different (improved) mask synthesis technique, which could be responsible for the performance benefit. MaskedArcFace appears to also apply synthetic masks to the test set. As for the M-M scenario, the proposed contrastive solution was able to outperform all alternative methods by a margin of at least $0.250\%$ FMR100.

\begin{figure}[t!]
\centering
\includegraphics[width=0.75\textwidth]{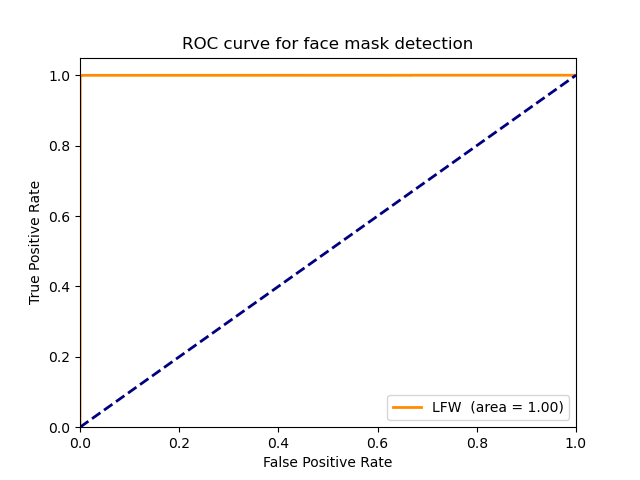}
\caption{Receiver operating characteristic (ROC) curve for mask detection on the LFW dataset with simulated masks.}
\label{fig:maskedfacerecog_roc_mask_detection}
\end{figure}

Thanks to the multi-task nature of the proposed methodology, it is possible to leverage the second embedding to perform simultaneous mask detection. Fig.~\ref{fig:maskedfacerecog_roc_mask_detection} shows the ROC curve results of the proposed model on the mask detection task, evaluated on the augmented LFW dataset. The model achieved a perfect face detection score, which could be exaggerated by the relative simplicity of the validation data, but nevertheless show the effectiveness of the proposed multi-task methodology.

Overall, these results showcase the capabilities of the proposed methodology, not only \emph{versus} the official MFR 2021 competition baseline and the submitted algorithms, but also against the proposed adapted triplet loss. Moreover, the ability to achieve competitive results with largely frozen pretrained backbones enables obtaining multi-task architectures with up to $80\%$ fewer trainable parameters.

\section{Summary and Conclusions}

The work presented in this chapter addressed the challenge of masked face recognition, made relevant by the worldwide face mask mandates born out of the recent Covid-19 pandemic. Despite the extensive work on the robustness to occlusions in face biometrics, the presence of a mask hiding the nose, chin, and mouth has resulted in significant performance decay.

Here, two approaches were presented to mitigate this issue. The first is an adaptation of the triplet loss by combining it with the mean squared error (MSE) loss, which will minimise the distance between embeddings from masked and unmasked versions of a face image. The second is a multi-task contrastive learning approach that aims to make the model aware of the presence of a mask and uses it alongside the ArcFace and MSE losses to achieve a more robust recognition model. Both approaches aim to lead a biometric model into offering the same templates for a face image, whether or not it contains a mask, effectively leading the model to avoid information from the face regions that could be occluded by a mask.

Results show that the combination of different losses and the promotion of embedding similarity between masked and unmasked versions of a face image are successful at improving masked face recognition performance. Experiments with an explainability tool also show the models are indeed led to avoid regions that could be occluded by masks and to better use information from visible face regions.

However, the second proposed approach, based on a multi-task contrastive learning strategy, is the one which offered the best results. By combining the ArcFace loss (widely praised in face recognition literature) with the MSE loss (which showed promise in the adapted triplet loss results) and a strategy to make the network aware of the presence of masks, it was able to outperform the official baseline and the best algorithm submissions in the MFR 2021 challenge.

Although the results are promising, further efforts should be devoted to the improvement of the contrastive learning approach, the preprocessing stage of adding synthetic masks to test data, and the study of larger and more accurate backbone architectures. The topic of masked face recognition would also benefit heavily from larger and improved databases, with real masks of diverse types worn by larger sets of individuals.

\chapter{Interpretability for Face Biometrics}\label{ch:interpFacePAD}

\begin{tcolorbox}\footnotesize
{\large\bf Foreword on Author Contributions}

The research work described in this chapter was conducted in collaboration with Ana F. Sequeira, Wilson Silva, and Tiago Gonçalves, under the supervision of Jaime S. Cardoso. The author of this thesis contributed to this work on the formulation and implementation of the face presentation attack detection framework, the preparation and conduction of experiments, the discussion of the results, and the writing of the scientific publications.

The results of this work have been disseminated as a journal article, an article in international conference proceedings, and an abstract in national conference proceedings:
\begin{itemize}[noitemsep, leftmargin=1em, nosep]
    \item A. F. Sequeira, T. Gonçalves, W. Silva, \underline{J. R. Pinto}, and J. S. Cardoso, ``An Exploratory Study of Interpretability for Face Presentation Attack Detection,'' \emph{IET Biometrics}, 10 (4), 2021.~\cite{Sequeira2021Exploratory}
    \item A. F. Sequeira, W. Silva, \underline{J. R. Pinto}, T. Gonçalves, and J. S. Cardoso, ``Interpretable Biometrics: Should We Rethink How Presentation Attack Detection is Evaluated?,'' in \emph{8th International Workshop on Biometrics and Forensics (IWBF 2020)}, Apr.~2020.~\cite{Sequeira2020Interpretable}
    \item W. Silva, \underline{J. R. Pinto}, T. Gonçalves, A. F. Sequeira, and Jaime S. Cardoso, ``Explainable Artificial Intelligence for Face Presentation Attack Detection,'' in \emph{26th Portuguese Conference on Pattern Recognition (RECPAD 2020)}, Oct.~2020.
\end{itemize}

\end{tcolorbox}

\section{Context and Motivation}

Like plenty of other artificial intelligence fields, biometrics has been witnessing a steadily increasing dominance of deep learning-based approaches. This is largely due to the availability of unprecedentedly large datasets and major computational gains offered by powerful graphics processing units (GPU)~\cite{LeCun2015, karpathy2014large}. As discussed in Chapter~\ref{ch:faceprior}, the largest downside of such a trend is the lack of transparency of the resulting algorithms, which has been increasingly criticised by the research community~\cite{samek2017explainable,holzinger2019causability,doshi2017towards}.

Beyond simple decision accuracy-based metrics, there are several other aspects of a model that may be useful (and even  to understand during its development and evaluation to avoid undesirable future consequences. One great example of this was described by \citet{lapuschkin2016analyzing}, who observed that for the detection of the class ``horse'' by a deep neural network, the model assigned relevance to the bottom left corner of the images. A careful inspection revealed the presence of a copyright tag in that location, meaning the model was relying on this tag for the decision instead of using meaningful image information.

For the example of a face presentation attack detection (PAD) model, it may be important for a model to verify certain properties that may not be immediately obvious. First, the model should look for the same information in a given sample whether or not that sample has been seen during training. Second, a presentation attack sample should be processed similarly by a model whether or not it was trained to detect that specific attack. Third, the behaviour of the model should be coherent (similar) for different samples with the same predicted label. At last, the model's choice of information within a sample should be meaningful (as in, a human would likely look to the same regions to provide the same decision).

These ideal behaviours may not be entirely objective and consensual. However, they are aspects of deep learning models that cannot be measured by traditional metrics. The ability to peek deeper into the inner workings of biometric models is the true advantage behind integrating interpretability in their development and evaluation.

The exploratory study presented in this chapter focused on the face PAD task to illustrate how interpretability could be integrated with biometrics to reach the aforementioned goal. As such, this work did not aim to push forward the state-of-the-art in face PAD, but to push forward the almost nonexistent field of the explainability analysis in biometrics. An end-to-end CNN was implemented to perform face PAD, and Grad-CAM was used to explain its decisions, delving deeper into the behaviour of the model and assessing whether or not it verifies the desirable properties discussed above.

\section{Methodology}

\subsection{Implemented face PAD network}

A PAD method receives a biometric trait measurement as input and returns a prediction of whether that measurement belongs to a live individual (referred to as a \textit{bona fide} sample) or a spoof attempt to intrude the system (in this case referred to as a \textit{presentation attack}).

\begin{figure}[!t]
    \centering
    \includegraphics[width=0.95\linewidth]{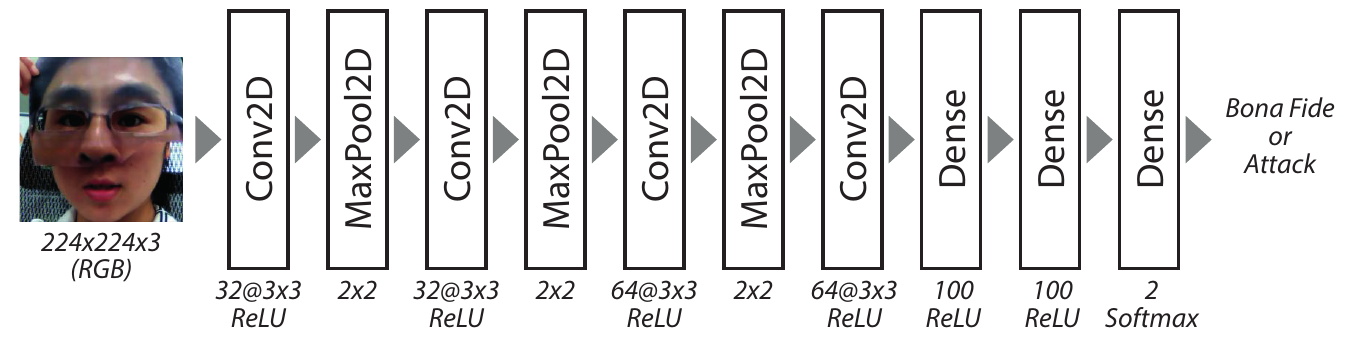}
    \caption{Architecture of the PAD end-to-end deep model used in this work.}
    \label{fig:interpfacebiom_arch_model}
\end{figure}

In this work, the model used for PAD is an end-to-end convolutional neural network (CNN). As an end-to-end CNN, the model was granted the flexibility to freely learn the most appropriate features for the task at hand. This provides the most interesting context for interpretability studies focused on gaining insight into the inner workings of a classifier. A relatively simple architecture was chosen (see Fig.~\ref{fig:interpfacebiom_arch_model}), as the emphasis of this work was to study the interpretability of the face PAD model and not necessarily to surpass the PAD state-of-the-art. 

The input to the network is a $224\times224$ RGB image and the output is a bidimensional softmax layer providing two probability scores. The network is composed of four convolutional layers, with three max-pooling layers interposed between them, and three fully-connected layers. The four convolutional layers are composed of $32$, $32$, $64$, and $64$ filters, respectively, with size $3\times3$, unit stride, and padding. The max-pooling is performed in $2\times2$ regions with stride $2$. The dense layers are composed of $100$, $100$, and $2$ neurons, respectively. All convolutional and fully-connected layers are followed by rectified linear unit (ReLU) activations, except for the last dense layer, which is followed by softmax activation.

\subsection{Interpretability method}

The Gradient-weighted Class Activation Mapping (Grad-CAM)~\cite{selvaraju2017grad} method is inspired by the Class Activation Mapping (CAM)~\cite{zhou2016learning}. CAM was introduced for the identification of discriminative regions in CNNs without fully-connected layers and restricted to the architectures that perform global average pooling over convolutional maps immediately before prediction. Grad-CAM is a generalisation of CAM in the sense that it was designed to be used with any type of CNN architecture.

Grad-CAM consists of the combination of feature maps using the gradient signal. The gradient information flows to the last convolutional layer of the model, thus assigning different importance values to each neuron according to a particular decision of interest. This interpretability tool enables the generation of explanations for any layer of the network. Additionally, it is possible to obtain explanations per class, allowing the analysis of the model predictions at a class-level.

\section{Experimental Setup}

\subsection{Data}

This work used \emph{bona fide} and presentation attack images extracted from the ROSE-Youtu Face Liveness Detection Dataset~\cite{li2018unsupervised}. This dataset is composed of $3497$ videos of twenty subjects, including attack videos of seven different PAI species (see Table~\ref{tab:interpfacepad_attack_types}). From each video, frames were extracted every five seconds and faces were detected on each frame using an MTCNN~\cite{Zhang2016}. Face regions were cropped, resized to $224\times224$, and normalised to $[0,1]$. The samples from subjects $\{2,3,4,5,6\}$ were reserved for testing, while the data from the remaining fifteen subjects were used for training and validation.

\begin{table}
    \centering
	\caption{PAI species in the ROSE Youtu DB and the respective number of extracted samples.}
    \label{tab:interpfacepad_attack_types}
	\begin{tabular}{clc}
	\hline
	\textbf{}  & \textbf{PAI Species}  & \textbf{No. Frames} \\ \hline 
	-       & Genuine (\emph{bona fide})                        & 2794 \\ 
	$\#1$   & Still printed paper                              & 1136 \\ 
	$\#2$   & Quivering printed paper                          & 1188 \\ 
	$\#3$   & Video of a Lenovo LCD display                    & 923 \\ 
	$\#4$   & Video of a Mac LCD display                       & 1113 \\ 
	$\#5$   	& Paper mask with two eyes and mouth cropped out   & 608 \\ 
	$\#6$   & Paper mask without cropping                      & 1194 \\ 
    $\#7$   & Paper mask with the upper part cut in the middle & 1162 \\\hline
	\end{tabular}
\end{table}

\subsection{Implementation details}

The PAD end-to-end model was trained on the extracted ROSE Youtu data using the Adam optimiser, with an initial learning rate of $10^{-4}$ for a maximum of $150$ epochs with batch size $8$. Early stopping was used, monitoring the validation loss, with patience of $20$ epochs. For regularisation, dropout ($0.5$) was used between each pair of consecutive dense layers. Horizontal flips, rotations with a range of $20$ degrees, and width and height shifts with a range of $0.2$ were used for data augmentation.

Explainability experiments were performed using the Grad-CAM~\cite{selvaraju2017grad} implementation provided by the Keras Visualisation Toolkit~\cite{raghakotkerasvis} for Python, which is a library that enables visualising and debugging a trained Keras model. It supports the visualisation of class activation maps, saliency maps, and activation maximisation. In the obtained maps, each pixel is assigned a relevance value that corresponds to a specific colour from blue (less relevant to the decision) to yellow (more relevant).

\subsection{Experimental scenarios and evaluation}

The explored scenarios follow those discussed in Chapter~\ref{ch:faceprior}:
\begin{itemize}
    \item In \emph{one-attack}, the PAD model is trained and tested with \emph{bona fide} samples and only one type of attack/PAD species (PAISp). Therefore, the only type of attack shown to the network during the test phase was already seen in the training step. The expression \emph{One-Attack$\#i$}, used throughout the remainder of this chapter, denotes the respective model was trained and tested with \textit{bona fide} samples and presentation attack samples of type $i$;

    \item In \emph{unseen-attack}, the PAD model is trained with all but one PAISp and tested with the remaining PAISp, besides the \emph{bona fide} samples in the train and test steps. Therefore, during the test phase, the network is evaluated only with the type of PAD attack that was not present in the training step (the unseen attack). This scenario enables a more thorough evaluation of the generalisation capabilities of the PAD model. The expression \emph{Unseen-Attack$\#i$}, used throughout the remainder of this chapter, denotes the respective model was tested with \textit{bona fide} samples and presentation attack samples of type $i$ and trained with \textit{bona fide} samples and the remaining types of attacks (\emph{i.e.}, trained with $j\in \{1,...,7\}\setminus{i}$).
\end{itemize}

Evaluation results are reported using standardised metrics commonly used by the literature in PAD. The \textit{Bona fide Presentation Classification Error Rate (BPCER)} measures the proportion of \textit{bona fide} samples erroneously classified as attacks and the \textit{Attack Presentation Classification Error Rate (APCER)} indicates the proportion of presentation attacks wrongly classified as \textit{bona fide}~\cite{ISOPAD2017}. The \textit{Equal Error Rate (EER)} is the error at the operation point where the APCER and BPCER present the same value.

\section{Conducted Studies on PAD Interpretability}

\subsection{Representation of a model's explanations}

Beyond qualitative visualisation of explanations, this work entails a quantitative analysis consisting of the comparison between the explanations obtained with \textit{bona fide} or presentation attack samples and taking into account the different evaluation scenarios (one-attack or unseen-attack). \textit{Bona fide} samples are present in both training and testing in any scenario, hence these can be tested and explained in any framework One-Attack$\#i$ or Unseen-Attack$\#i$.

The presentation attack samples belong to a specific type of attack. Considering the one-attack evaluation scenario, an attack sample of type $\#i$ can only be tested for the respective One-Attack$\#i$ scenario. It is not meaningful to test this sample with the models of One-Attack$\#j$ (with $j \neq i$) because then this would be an Unseen-Attack$\#i$ scenario.

As such, let $I=\{I_1$,...,$I_n\}$ and $E^{xi}=\{E_1^{xi}$,...,$E_n^{xi}\}$ be a set of images and the respective set of explanations. For each image $I_k$ (for $k=1,...,n$) there is a corresponding explanation $E_k^{xi}$. Note that $x=o$ or $x=u$ whether the explanation refers to a classification result within the one-attack or unseen-attack scenario, respectively, and $i=1,...,7$ is the type of attack that defines the model used for testing. Thus, each explanation is obtained with a specific model determined by the evaluation framework and the attack used in testing.

\subsection{Semantic representation of explanations}

\begin{figure}[!t]
\centering
\includegraphics[width=\linewidth]{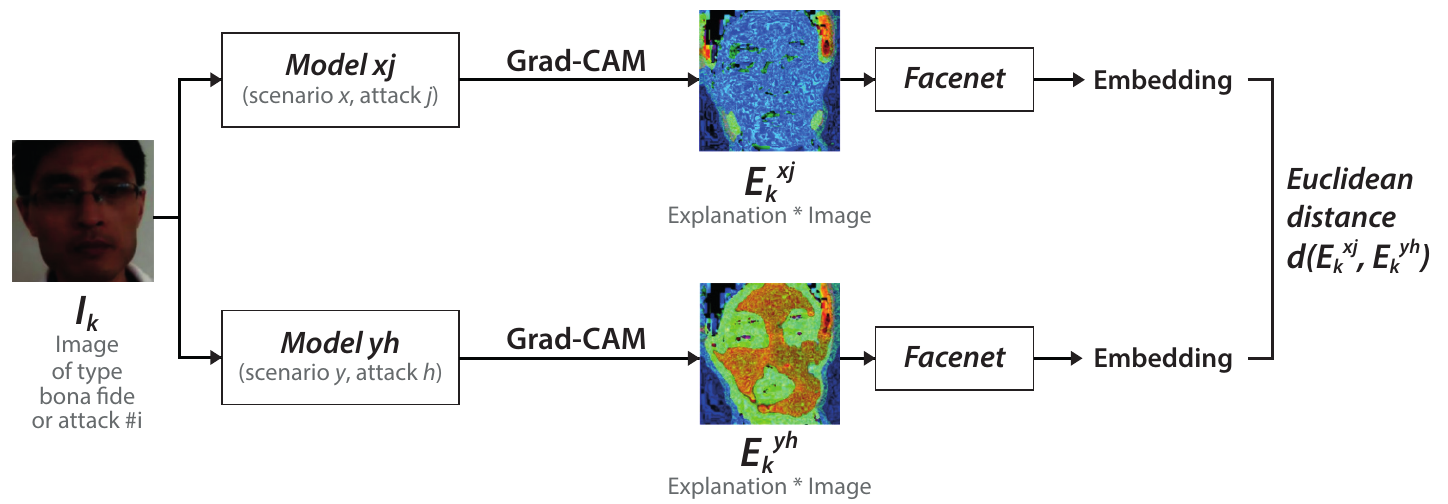}
\caption{Example of the approach used to quantify the difference between two explanations.}
\label{fig:interpfacebiom_representations}
\end{figure}

One of the objectives of this work was to measure the variability of the explanations for both \textit{bona fide} and presentation attack samples in the two different evaluation scenarios. To do this, one needs a suitable representation of the produced explanations, so that it becomes possible to quantify how much two explanations differ from each other. In this work, this comparison is performed in a semantic context. An illustrative example of how the approach used to perform a quantitative comparison between explanations is depicted in Fig.\ref{fig:interpfacebiom_representations}.

As stated before, the explanations are generated using Grad-CAM, which highlights the image regions that maximise the predicted class. Since Grad-CAM typically produces blobby and coarse explanations that fail to preserve finer details, it was decided to multiply the saliency maps by the respective images. However, this space is still not ideal for image comparison, as it would be highly impacted by the spatial location of important features.

To overcome this issue, and inspired by what is being done in image retrieval~\cite{hofmanninger2015mapping,silva2020interpretability} and concept-based interpretability~\cite{ghorbani2019towards} to find similar images, the learned features computed by a pre-trained CNN were used as the space to measure the distance between two explanations. This follows the finding of \citet{zhang2018unreasonable} that the Euclidean distance in the activation space of final layers is an effective similarity metric.

Since this work focused on face images, FaceNet~\cite{Schroff2015}, a face-specific network pretrained on the VGGFace2 dataset~\cite{Cao2018}, was used for the extraction of the deep features. This CNN was trained using a triplet loss, aiming to optimise the embedding space and ensure that the FaceNet could learn a function that correctly maps the face images to a compact Euclidean space where distances directly correspond to a measure of face similarity.

To take advantage of these FaceNet properties and achieve meaningful mappings of the Grad-CAM explanations, we start by multiplying the original images by their respective Grad-CAM explanations. We then input the resulting image into FaceNet, and extract the features generated in the penultimate layer. All the Euclidean distances reported in this paper are computed in this semantic space.

\subsection{Comparison of explanations across different scenarios}

One could hypothesise that, for a robust PAD model, explanations for the same sample should be similar whether or not the model is trained to detect that specific attack. Verifying this property means we are in the presence of a PAD algorithm with thorough generalisation capabilities. In this section, we describe a study that addresses the assessment of this property.

\begin{figure}[!t]
\centering
\includegraphics[width=\linewidth]{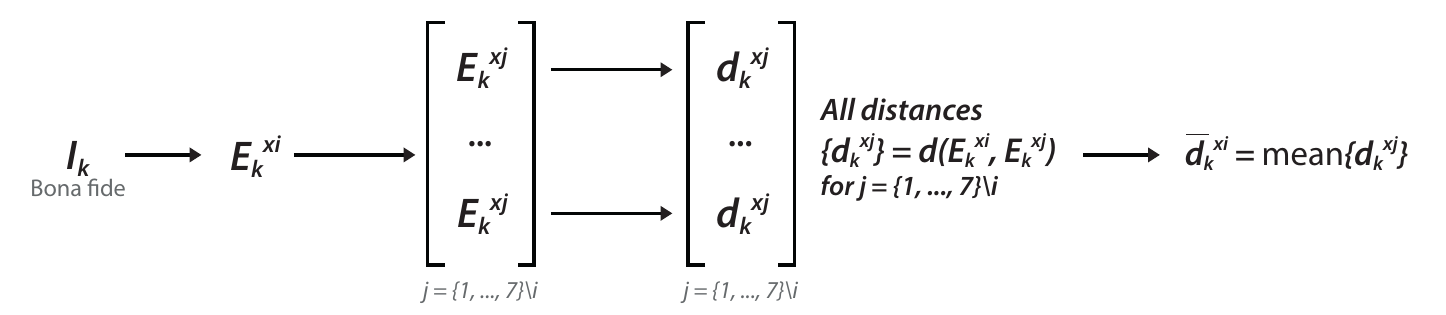} 
\caption{Comparison of explanations for a \textit{bona fide} sample $I_k$, on the evaluation scenario $x$, and fixing Attack $\#i$.}
\label{fig:interpfacebiom_calc_bf1}
\end{figure}

For \emph{bona fide} images, Fig.~\ref{fig:interpfacebiom_calc_bf1} illustrates the process to compare the explanations, having an image $I_k$, the evaluation framework $x$ (either one-attack or unseen-Attack), and fixing as reference the explanation obtained by the model for the PAISp $\#i$ in both frameworks. 

\begin{figure}[!t]
\centering
\includegraphics[width=\linewidth]{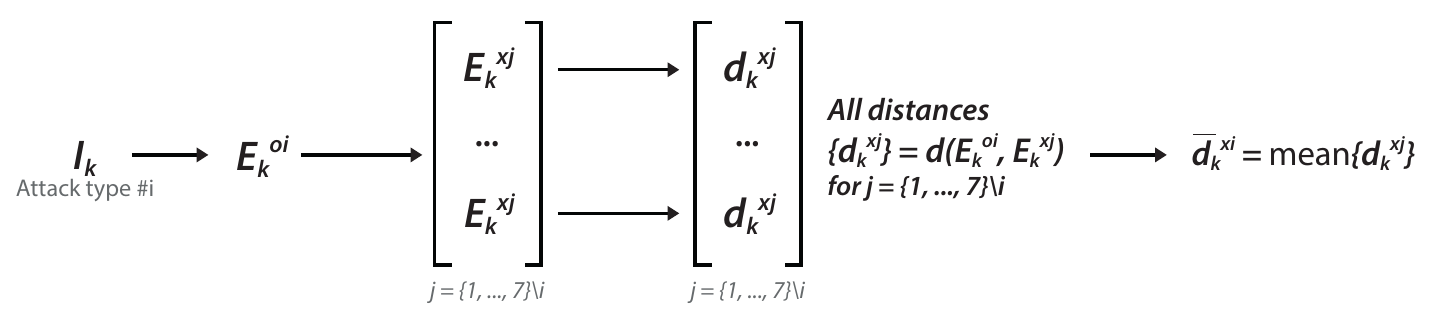} 
\caption{Comparison of explanations for a presentation attack sample $I_k$, on the evaluation scenario $x$, and fixing Attack $\#i$.}
\label{fig:interpfacebiom_calc_pa1}
\end{figure}

As for presentation attack images, Fig.~\ref{fig:interpfacebiom_calc_pa1} illustrates the process to compare explanations, having a presentation attack image $I_k$ of type Attack $\#i$. In this case, the comparison is made by fixing as reference the explanation of the result of the classification in the one-attack framework obtained by the model for the Attack $\#i$. Thus, the evaluation framework is $x=o$. As mentioned before, this is done in order to have a more stable benchmark for comparison, since in the one-attack scenario the model is trained and tested with the same type of attack.

So, for each image $I_k$ (either \textit{bona fide} or presentation attack), evaluation framework $x$, and using the model regarding attack $\#i$, a set of six total values $\{d_k^{xj}: j\in \{1,...,7\}\setminus{i}\}$ is obtained. It results of the comparison between the explanation $E_k^{xi}$ (always $E_k^{oi}$ in the presentation attack case) and the explanation $E_k^{xj}$ (for $j\in \{1,...,7\}\setminus{i}$). 

The distance measurements between explanations $\{d_k^{xj}\}$ provide a quantitative measure of the variability of the explanations produced by the different models when processing the same image. Averaging these values will provide the $\bar{d}_k^{xi}$ values for comparison, given by:
\begin{equation}
\label{eq:eq1}
\bar{d}_k^{xi} = \frac{1}{6}\sum_{j}{d_k^{xj}}.
\end{equation}

For sake of clarity, Table~\ref{tab:interpfacebiom_average_values} presents these distance values and their correspondence to the images and scenarios. The values $\bar{d}_k^{xi}$ obtained for each image $I_k$ are unique for a presentation attack sample and multiple (one for each $i\in\{1,...,7\}$) for a \textit{bona fide} sample. They can be used for other quantitative interpretability endeavours, including to obtain image average and attack average distances:

\begin{itemize}
    \item The Image Average ($I\mu$) provides a quantitative measure of the variability, across all models, of the explanations produced by each model under the evaluation framework defined by $xi$ (for $x=o$ or $x=u$ and $i=1,...,7$) regarding one image $I_k$. The $I\mu$ for the \emph{bona fide} is given by Eq.~\eqref{eq:IAbf} and for the presentation attacks is given by Eq.~\eqref{eq:IApa}.    
    \begin{align}
        \label{eq:IAbf}
        \text{for a \emph{bona fide} sample: } \hspace{0.5cm} & {I\mu}_k^{x} = \frac{1}{7}\sum_{i=1}^{7}\bar{d}_k^{xi}
        \\
        \label{eq:IApa}
        \text{for an attack sample (type $\#i$): } \hspace{0.5cm} & {I\mu}_k^{x} = \bar{d}_k^{xi}
    \end{align}

    \item The Attack Average ($A\mu$) provides a quantitative measure of the variability, across all samples, of the explanations produced by the model under one evaluation scenario defined by $xi$ (for $x=o$ or $x=u$ and $i=1,...,7$). Consider the values $\bar{d}_k^{xi}$ as defined in Table~\ref{tab:interpfacebiom_average_values} and being $n$ ad $m$ the number of \textit{bona fide} and \textit{attack} samples, respectively. The $A\mu$ for the \textit{bona fide} is given by Eq.~\eqref{eq:AAbf} and for the presentation attacks is given by Eq.~\eqref{eq:AApa}.
    \begin{align}
        \label{eq:AAbf}
        \text{for a \emph{bona fide} sample: } \hspace{0.5cm} & {A\mu}^{xi} = \frac{1}{n}\sum_{i=1}^{n} \bar{d}_k^{xi}, \text{ for $i=1,...,7$} \\
        \label{eq:AApa}
        \text{for an attack sample (type $\#i$): } \hspace{0.5cm} &  {A\mu}^{xi} = \frac{1}{m}\sum_{i=1}^{m} \bar{d}_k^{xi},\text{ for $i=1,...,7$}
    \end{align}

\end{itemize}

\begin{table}[!t]
\caption{Overview of the strategy to compare explanations for a sample across different evaluation scenarios.}
\label{tab:interpfacebiom_average_values}
\centering
\begin{tabular}{ccc}
\hline
\textbf{Sample Class}     & \textbf{Scenario}  & \textbf{Comparisons} \\ \hline
\multirow{2}{*}{\emph{Bona Fide}}& One-Attack      & \multirow{2}{*}{$\{\bar{d}_k^{xi}:\bar{d}_k^{xi} \text{ for } i=1,...,7\}$}  \\ 
& Unseen-Attack   &  \\ \hline 
Presentation attack & One-Attack      & \multirow{2}{*}{$\bar{d}_k^{xi}$}\\
(type $\#i$)  & Unseen-Attack   &      \\ \hline
\end{tabular}
\end{table}

\subsection{Interclass comparison in the unseen-attack scenario}

This study investigates the interclass comparison between explanations obtained using the models in the unseen-attack framework. In other words, it investigated the variability of the explanations between \textit{bona fide} and presentation attack samples. To achieve the desired goal, the explanations obtained from the classification of each image, with the different models trained in the Unseen-Attack scenario, are compared in a pairwise manner. This comparison is performed for all images within each class.

By using the unseen-attack models it is possible to test the robustness of the models to the variability in the attacks present in the training and testing steps. Recall that a model resulting from unseen-attack$\#i$ is trained with attacks $j$ for $j\in \{1,...,7\}\setminus{i}$.

\begin{figure}[!t]
\centering
\includegraphics[width=\linewidth]{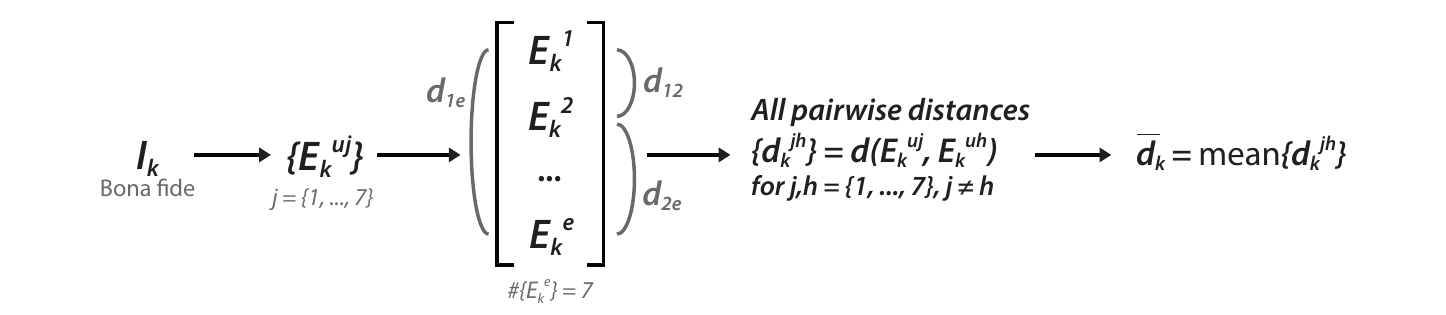} 
\caption{Pairwise comparison of explanations produced by the models in unseen-attack scenarios, for a \textit{bona fide} sample $I_k$.}
\label{fig:interpfacebiom_pwd_bf2}
\end{figure}

For a \emph{bona fide} sample $I_k$, the process to obtain the comparison of all explanations is illustrated in Fig.~\ref{fig:interpfacebiom_pwd_bf2}. It shows one example of how to obtain the pairwise distances $D_k$ given by: $D_k=\{d_k^{jh}: d_k^{jh}=d(E_k^{uj},E_k^{uh}),\text{ with }  j,h \in \{1,...,7\} \text{ and } j \neq h\}$. Then the values in $D_k$ are averaged and $\bar{d}_k$ is obtained for image $I_k$. A global value is obtained averaging all these values, $\bar{d}_{BF}$, as given by Eq.~\eqref{eq:dbf}:
\begin{equation}
   \label{eq:dbf}
     \bar{d}_{BF} = \frac{1}{n}\sum_{k} \bar{d}_k,
\end{equation}
for each \textit{bona fide} image $I_k$, $k=1,...,n$.

\begin{figure}[!t]
\centering
\includegraphics[width=\linewidth]{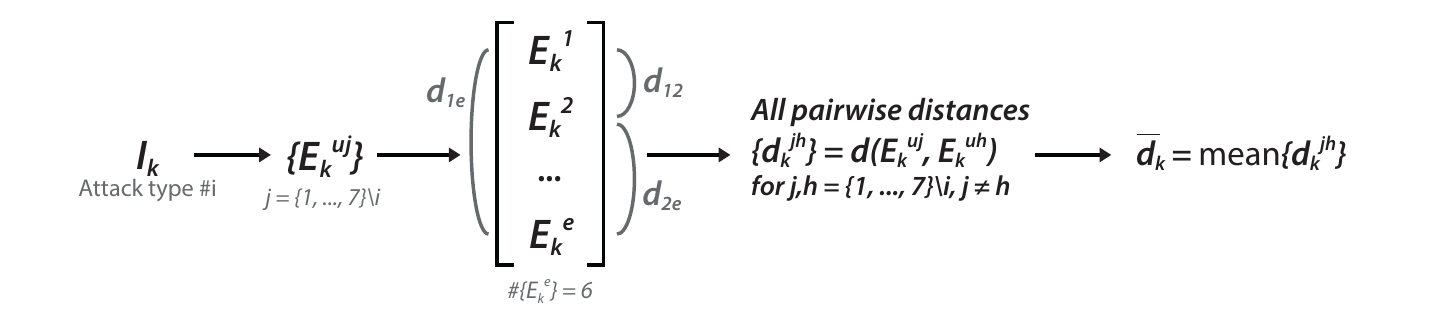} 
\caption{Pairwise comparison of explanations produced by the models in unseen-attack scenarios, for the presentation attack sample $I_k$ of type $\#i$.}
\label{fig:interpfacebiom_pwd_pa2}
\end{figure}

As for a presentation attack image $I_k$ (of type $\#i$), Fig.~\ref{fig:interpfacebiom_pwd_pa2} shows the process to obtain the comparison between the explanations obtained for the models Unseen-Attack$\#j$ for $j\in \{1,...,7\}\setminus{i}$. The figure shows one example, regarding image $I_k$ of type $Attack\#i$, on how to obtain the pairwise distances $D_k=\{d_k^{jh}: d_k^{jh}=d(E_k^{uj},E_k^{uh}),\text{ with } j,h\in \{1,...,7\}\setminus{i}\} \text{ and } j \neq h\}$. The values in $D_k$ are averaged and $\bar{d}_k$ is obtained for image $I_k$. A global value is obtained averaging all these values, $\bar{d}_{PA}$, as given by Eq.~\eqref{eq:dpa}:
\begin{equation}
  \label{eq:dpa}
     \bar{d}_{PA} = \frac{1}{m}\sum_{k} \bar{d}_k,
\end{equation}
for each presentation attack image $I_k$, $k=1,...,m$, of type $\#i$.

\subsection{Intraclass comparison across different samples}

One other property that a robust PAD solution should verify is that explanations should be similar for different samples with the same label. In other words, the explanations should reveal that the model looks for the same regions of the face even when the images are very different, since the ground-truth label is the same. This means the algorithm is coherent in its decisions and has effectively pinpointed the features that allow for accurate and stable detection of presentation attacks.

It is important to investigate the coherence of explanations for the \textit{bona fide} samples, in order to understand if the considered PAD solution knows indeed what is a ``real face''. However, it is also important to understand how much the explanations vary for presentation attack samples, so we can assess if our algorithm is indeed capable of generalising well to the multiple possible attack species.

This analysis can be performed by measuring how much the explanations for the models' decisions are affected by variations in the types of presentation attacks known in the learning phase. In this work, this comparison is done by comparing features extracted from the explanations (using the FaceNet model described above) and not in a pixel-to-pixel manner, making it possible to compare the explanations obtained for different samples.

\section{Results and Discussion}

\subsection{Performance of the face PAD algorithm}

The main focus of this work was interpreting the decisions of a face PAD model to illustrate and motivate the application of interpretability to biometrics. Achieving improved face PAD results \emph{versus} the state-of-the-art was not a priority. Nevertheless, one should not aim to interpret a model that lacks PAD abilities by design, as this will impair and bias the drawn conclusions. As such, the performance results of the implemented model in the one-attack and unseen-attack scenarios are presented in Table~\ref{tab:interpfacebiom_tabela_PAD_perf}. 

The results were generally inferior in the unseen-attack scenario when compared to the one-attack scenario, as expected given the considerably higher difficulty of the former scenario and the insight found in face PAD literature~\cite{sequeira2016realistic}. One reason for such results may be the large variability between the different PAI species found in the ROSE Youtu dataset. Hence, to obtain a truly robust PAD algorithm, the test set should always be composed of PAI species which have not been seen by the algorithm during training~\cite{ferreira2019adversarial}.

Attack $\#7$, an upper face mask with eyes cropped out, is a good example of the implemented model's difficulty to generalise. Although the results are acceptable in the one-attack scenario, the error rates are considerably higher in the unseen-attack scenario. Some exceptions can also be found, such as Attack $\#1$ (full face printed photo), for which both EER and APCER are actually lower in the unseen-attack scenario \emph{vs}. one-attack. For this specific PAI species, the model is probably learning most of the needed features from various other paper-based attack types in the unseen-attack scenario, taking advantage of the greater availability of data despite the absence of Attack $\#1$ samples in the training.

\begin{table}[!t]
\centering
\caption[Performance of the PAD models in the one-attack and unseen-attack evaluation scenarios.]{Performance of the PAD models in the one-attack and unseen-attack evaluation scenarios (EER, APCER, and BPCER in \%; APCER and BPCER calculated for a threshold of $0.5$).}
\label{tab:interpfacebiom_tabela_PAD_perf}
\begin{tabular}{ccccccccc}
\hline
\multirow{2}{*}{\textbf{Attack}} && \multicolumn{3}{c}{\textbf{One-Attack}}   && \multicolumn{3}{c}{\textbf{Unseen-Attack}}    \\  
                                 && \textbf{\textit{EER}} & \textbf{\textit{APCER}} & \textbf{\textit{BPCER}} &&
                                 \textbf{\textit{EER}} & \textbf{\textit{APCER}} & \textbf{\textit{BPCER}} \\ \hline
1 && 7.29          & 12.15         & 3.06          && 5.90          & 6.94          & 4.90          \\
2 && 3.62          & 6.67          & 1.35          && 5.55          & \textbf{3.00} & 10.65         \\
3 && 2.79          & 8.37          & 0.12          && 10.38         & 26.29         & 4.28          \\
4 && 12.66         & 30.38         & 1.84          && 25.34         & 45.73         & \textbf{3.92} \\
5 && 1.61          & \textbf{1.61} & 1.59          && \textbf{4.84} & 3.55          & 7.10          \\
6 && 4.46          & 5.10          & 1.10          && 10.19         & 12.74         & 7.71          \\
7 && \textbf{0.73} & 5.23          & \textbf{0.00} && 15.49         & 34.31         & 7.71          \\ \hline
\end{tabular}
\end{table}

\subsection{Comparison of explanations across different scenarios}

\subsubsection{Image average (${I\mu}$)}

One of the objectives of this work was to study the behaviour of the PAD models under varying data diversity in the training dataset. For an ideally robust PAD model, the same explanation should be obtained for a given \emph{bona fide} image, regardless of the PAI species seen during the training stage. However, in reality, models are often sensitive to variations in the training data, both for \emph{bona fide} and presentation attack samples.

\begin{figure}[!t]
\centering
\includegraphics[width=0.75\linewidth]{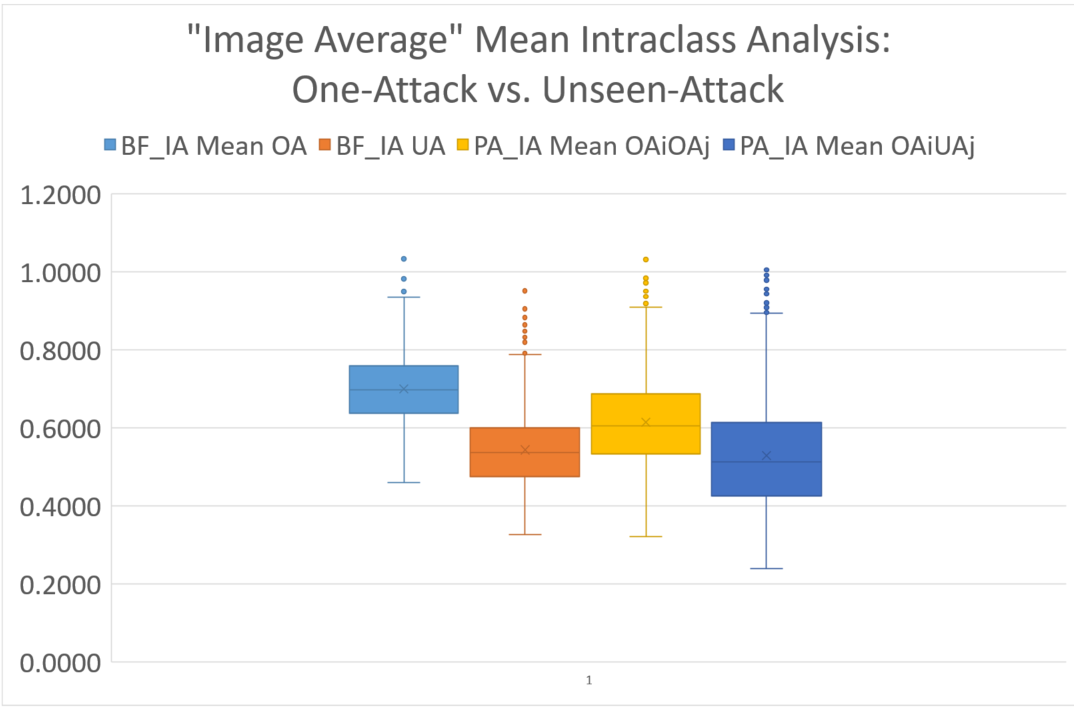}
\includegraphics[width=0.75\linewidth]{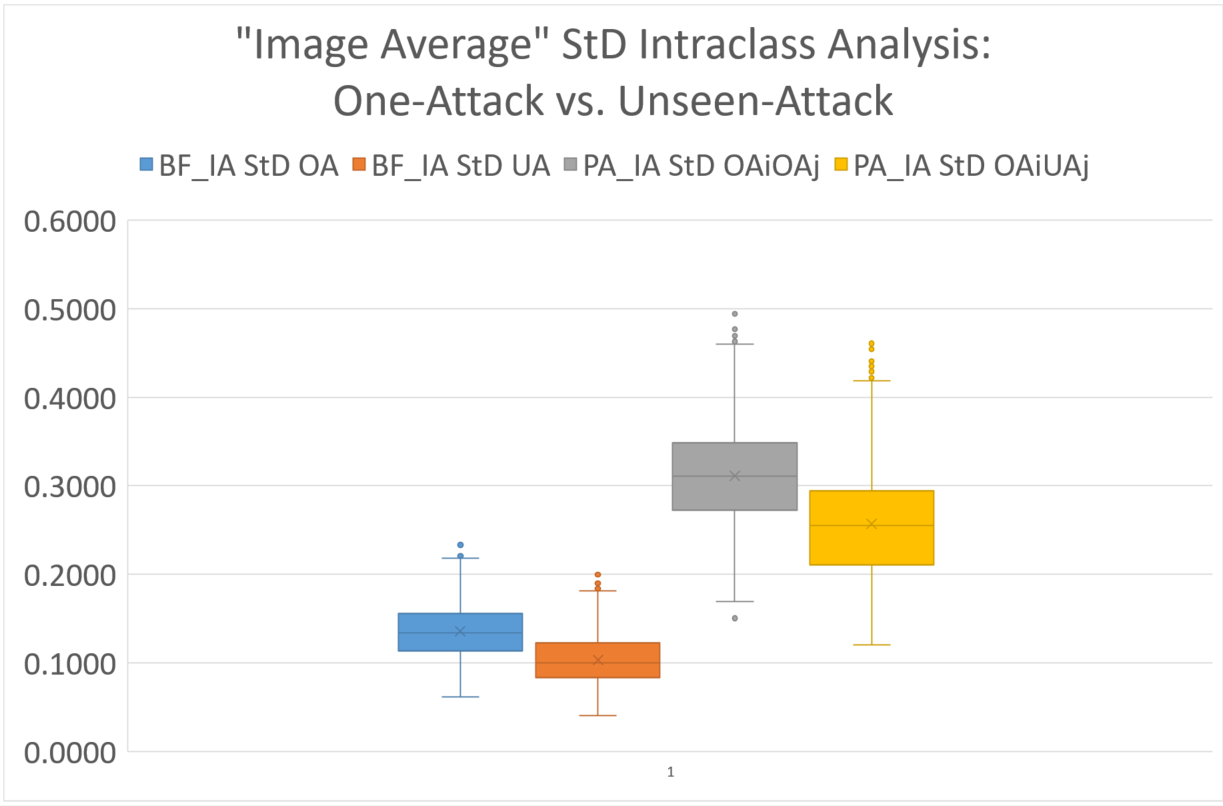}  
\caption{Image Average mean and standard deviation (StD) results for \emph{bona fide} (BF) and presentation attack (PA) samples in the comparison across the one-attack (OA) and unseen-attack (UA) scenarios.}
\label{fig:interpfacebiom_BF_IA_OA_UA_D2_v2}
\end{figure}

Fig.~\ref{fig:interpfacebiom_BF_IA_OA_UA_D2_v2} presents the mean results ($\mu({I\mu})$) and respective standard deviation ($\sigma({I\mu})$) of the Image average ($I\mu$) experiments for \emph{bona fide} (BF) and presentation attack (PA) samples across the one-attack (OA) and unseen-attack (UA) scenarios. These results show that intraclass variability is higher in the one-attack scenario, as indicated by higher $\mu({I\mu})$ values.

The $I\mu$ values also show higher variability in one-attack, according to the $\sigma(I\mu)$ results. This suggests that the models in the unseen-attack scenario are better able to generalise in the recognition of \emph{bona fide} samples, since they see a wider variety of attacks during training. For the PA samples, the variability results also suggest the models are more robust when a wider diversity of attacks is available during training.

\begin{figure}[!t]
\centering
\begin{tabular}{cccc}
\textbf{Attack $\#5$} & One-Attack 1 & One-Attack 2 & One-Attack 3
\\
\includegraphics[width=0.2\linewidth]{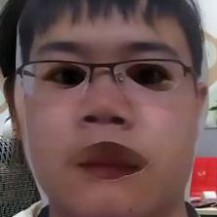}
& 
\includegraphics[width=0.2\linewidth]{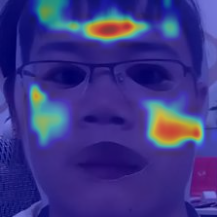}
&
\includegraphics[width=0.2\linewidth]{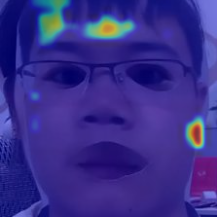}
&
\includegraphics[width=0.2\linewidth]{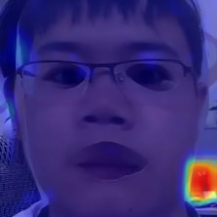}
\\
\textbf{One-Attack} 5 & One-Attack 4 & One-Attack 6 & One-Attack 7 
\\
\includegraphics[width=0.2\linewidth]{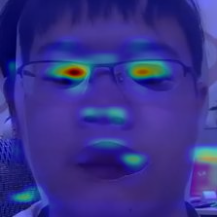}
&
\includegraphics[width=0.2\linewidth]{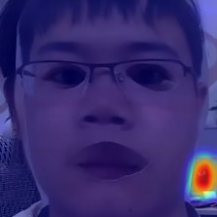}
&
\includegraphics[width=0.2\linewidth]{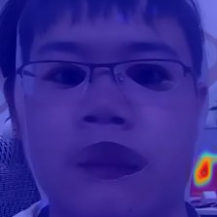}
&
\includegraphics[width=0.2\linewidth]{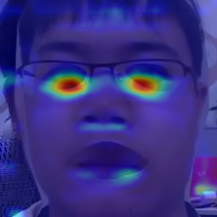}
\end{tabular}
\caption[Comparison of explanations in intraclass one-attack for an example PA sample of type $5$ presenting high $I\mu$ value.]{Comparison of explanations in intraclass one-attack for an example PA sample of type $5$ presenting high $I\mu$ value (obtained when the One-Attack$\#5$ is compared against the One-Attacks$\{\#1,...,\#7\}\setminus{\#5}$).}
\label{fig:interpfacebiom_3pa_oa_ua_HL}
\end{figure}

\begin{figure}[!t]
\centering
\begin{tabular}{cccc}
\textbf{Attack $\#5$} & Unseen-Attack 1 & Unseen-Attack 2 & Unseen-Attack 3
\\
\includegraphics[width=0.2\linewidth]{figures/faceBiometrics/interpFacePAD/pa_oa_ua_HighLow_s005_atk_05.png}
& 
\includegraphics[width=0.2\linewidth]{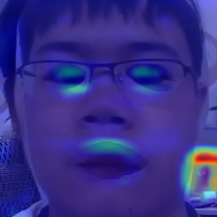}
&
\includegraphics[width=0.2\linewidth]{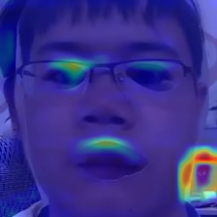}
&
\includegraphics[width=0.2\linewidth]{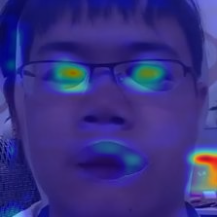}
\\
\textbf{One-Attack} 5 & Unseen-Attack 4 & Unseen-Attack 6 & Unseen-Attack 7 
\\
\includegraphics[width=0.2\linewidth]{figures/faceBiometrics/interpFacePAD/pa_oa_ua_HighLow_s005_atk_05_one_attack_5.png}
&
\includegraphics[width=0.2\linewidth]{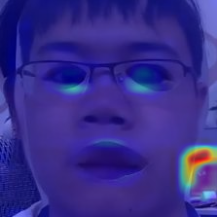}
&
\includegraphics[width=0.2\linewidth]{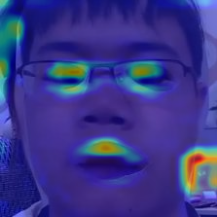}
&
\includegraphics[width=0.2\linewidth]{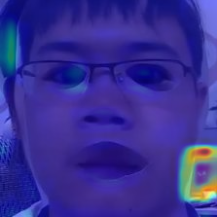}
\end{tabular}
\caption[Comparison of explanations in intraclass unseen-attack for an example PA sample of type $5$ presenting low $I\mu$ value.]{Comparison of explanations in intraclass unseen-attack for an example PA sample of type $5$ presenting low $I\mu$ value (obtained when the One-Attack$\#5$ is compared against the Unseen-Attacks$\{\#1,...,\#7\}\setminus{\#5}$).}
\label{fig:interpfacebiom_2pa_oa_ua_HL}
\end{figure}

Fig.~\ref{fig:interpfacebiom_3pa_oa_ua_HL} and Fig.~\ref{fig:interpfacebiom_2pa_oa_ua_HL} show an example of a PA sample presenting a higher $\mu({I\mu})$ in the one-attack scenario when compared to unseen-attack. These results further support the idea that a model trained with more than one PAI species is able to learn better patterns and become a more robust model with better generalisation capabilities.

\subsubsection{Attack average (${A\mu}$)}

\begin{figure}[!p]
\centering
\includegraphics[width=\linewidth]{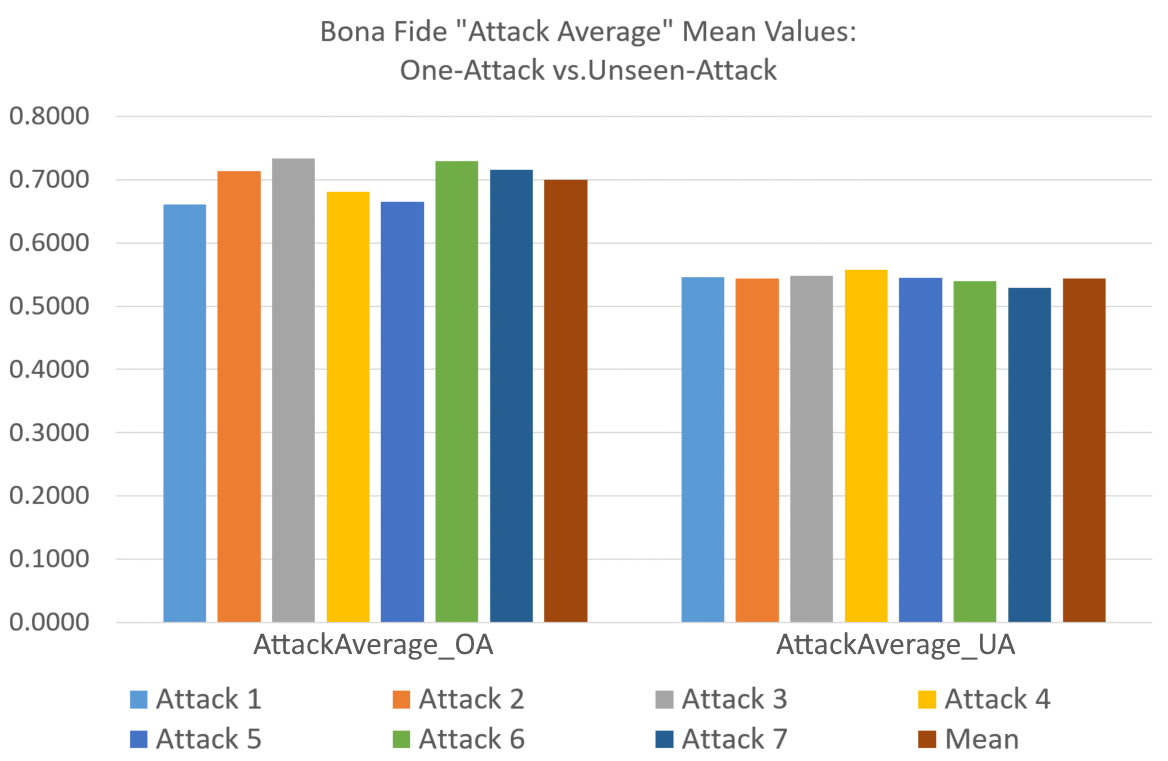}
\includegraphics[width=\linewidth]{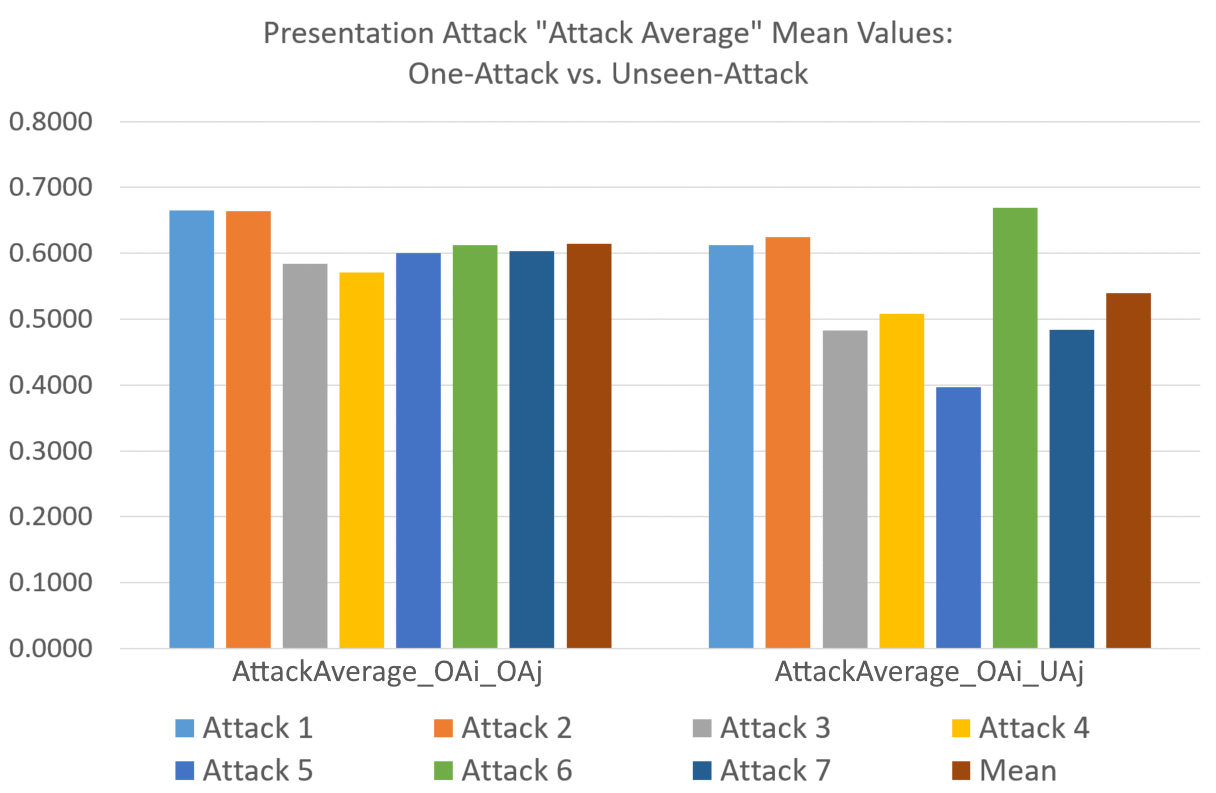} 
\caption{Mean $A\mu$ results for \emph{bona fide} and presentation attack samples in the one-attack (OA) and unseen-attack (UA, $i=1,...,7$) scenarios and respective mean value.}
\label{fig:interpfacebiom_atav_oa_ua}
\end{figure}

The mean and standard deviation of the attack average results (respectively, $\mu(A\mu)$ and $\sigma(A\mu)$) across the one-attack and unseen-attack scenarios for \emph{bona fide} and presentation attack samples are presented in Fig.~\ref{fig:interpfacebiom_atav_oa_ua}. 

Similar to the $I\mu$ results presented before, the $A\mu$ results show higher variability in the one-attack scenario. This suggests once again that a training setup that integrates more than one PAI species may be more successful at promoting the learning of more coherent features for the \emph{bona fide} class. The same is also verified for presentation attack samples, as the mean distance in the unseen-attack scenario is inferior to that in one-attack. 

Since each PAI species contains intrinsic specificities, one could hypothesise that grouping multiple of them into a single class ``presentation attack'' would confuse the model and result in lesser robustness. Yet, the results indicate that the models used to detect them seem to benefit from the integration of more attacks during the training phase.

It is interesting to observe the variability for presentation attack samples across the different attacks in the unseen-attack scenario: although, in general, the values are inferior to those in the one-attack scenario, some specific attacks present much lower results than others. This may denote that differences (and similarities) between PAI species seen in training and evaluation lead to models that are much more sensitive to unseen PAI species in the testing phase.

In particular, in the presentation attack graphs in Fig.~\ref{fig:interpfacebiom_atav_oa_ua}, Unseen-Attack$\#5$, consisting of paper masks with eyes and mouth cut off, presents the lowest $\mu(A\mu)$. This may result from the fact that, in this scenario, the model has seen multiple print-based PAI species (like complete photos, complete paper masks, and half paper masks) which help the model prepare for PAI species $\#5$. The Unseen-Attack$\#7$ (consisting of upper-half face paper masks) presents a higher result since these samples combine skin and paper in the facial area and may be more difficult to learn from the PAI species seen during training.

\subsection{Interclass comparison in the unseen-attack scenario}

\begin{figure}[!t]
\centering
\begin{tabular}{cccc}
\emph{Bona Fide} & Unseen-Attack 1 & Unseen-Attack 2 & Unseen-Attack 3
\\
\includegraphics[width=0.2\linewidth]{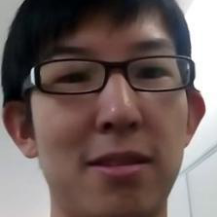}
& 
\includegraphics[width=0.2\linewidth]{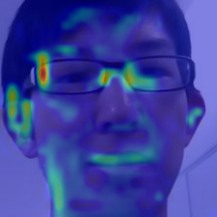}
&
\includegraphics[width=0.2\linewidth]{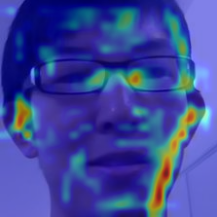}
&
\includegraphics[width=0.2\linewidth]{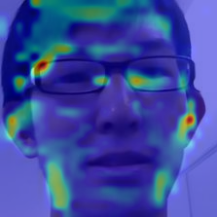}
\\
Unseen-Attack 4 & Unseen-Attack 5 & Unseen-Attack 6 & Unseen-Attack 7 
\\
\includegraphics[width=0.2\linewidth]{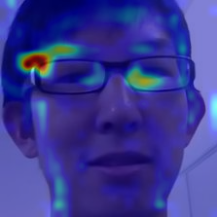}
&
\includegraphics[width=0.2\linewidth]{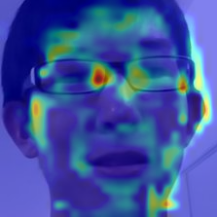}
&
\includegraphics[width=0.2\linewidth]{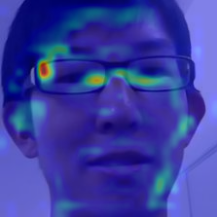}
&
\includegraphics[width=0.2\linewidth]{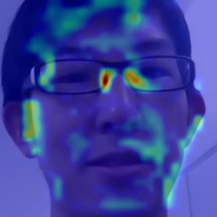}
\end{tabular}
\caption{Explanations for an example \emph{bona fide} samples with pairwise distance close to the obtained average ($\bar{d}_{BF}=0.54$).}
\label{fig:interpfacebiom_pw_bf_median}
\end{figure}

\begin{figure}[!t]
\centering
\begin{tabular}{cccc}
\emph{Attack $\#2$} & Unseen-Attack 1 & Unseen-Attack 3 & Unseen-Attack 4
\\
\includegraphics[width=0.2\linewidth]{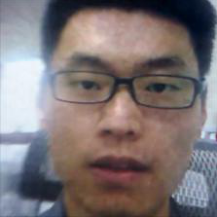}
& 
\includegraphics[width=0.2\linewidth]{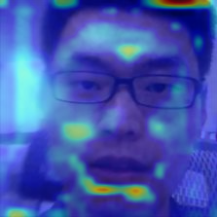}
&
\includegraphics[width=0.2\linewidth]{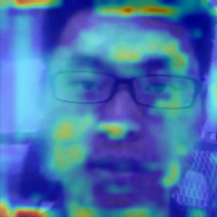}
&
\includegraphics[width=0.2\linewidth]{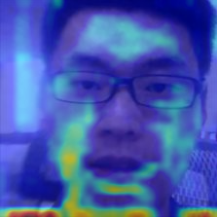}
\\
 & Unseen-Attack 5 & Unseen-Attack 6 & Unseen-Attack 7 
\\
&
\includegraphics[width=0.2\linewidth]{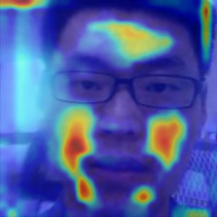}
&
\includegraphics[width=0.2\linewidth]{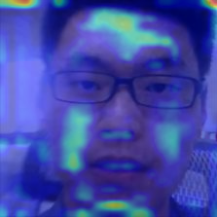}
&
\includegraphics[width=0.2\linewidth]{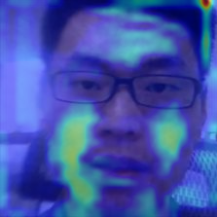}
\end{tabular}
\caption{Explanations for an example presentation attack sample of type $\#2$ with pairwise distance close to the obtained average ($\bar{d}_{PA}=0.52$).}
\label{fig:interpfacebiom_pw_pa_median}
\end{figure}

This experiment aimed to quantify the variability of the explanations between \emph{bona fide} and presentation attack samples. For the \emph{bona fide} samples, a result of $\bar{d}_{BF}=0.54$ was obtained, by averaging the value $\bar{d}_k$ of all \emph{bona fide} images ($I_k$, for $k=1,...,n$) obtained from a pairwise comparison of the explanations of all models in the unseen attack scenario. The associated standard deviation is $0.13$. As for the presentation attack samples, a result of $\bar{d}_{PA}=0.52$ was obtained by averaging the value $\bar{d}_k$ of all attack images ($I_k$, for $k=1,...,m$) obtained from a pairwise comparison of the explanations of Unseen-Attack$\#j$ for $j\in \{1,...,7\}\setminus{i}$ models. The associated standard deviation is $0.14$.

The obtained results are very similar and the standard deviation values are not only relatively high but similar for both sample classes. The similarity of the values does allude to any definitive comparative conclusion, but it does motivate further investigation. Fig.~\ref{fig:interpfacebiom_pw_bf_median} and Fig.~\ref{fig:interpfacebiom_pw_pa_median} depict examples of images whose pairwise distance results $\bar{d}_{k}$ are close to the respective mean values $\bar{d}_{BF}$ and $\bar{d}_{PA}$. A visual inspection allows us to conclude there is, in fact, considerable variability between explanations. Good examples of this are the Unseen-Attacks$\#4,\#6$ in comparison to $\#2,\#3,\#5$ in Fig.~\ref{fig:interpfacebiom_pw_bf_median}, and the Unseen-Attacks$\#3,\#4$ in comparison to $\#1,\#6$) in Fig.~\ref{fig:interpfacebiom_pw_pa_median}.

\begin{figure}[!t]
\centering
\begin{tabular}{cccc}
\emph{Bona Fide} & Unseen-Attack 1 & Unseen-Attack 2 & Unseen-Attack 3
\\
\includegraphics[width=0.2\linewidth]{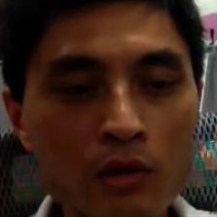}
& 
\includegraphics[width=0.2\linewidth]{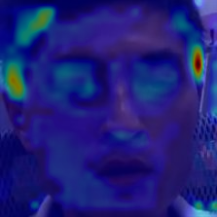}
&
\includegraphics[width=0.2\linewidth]{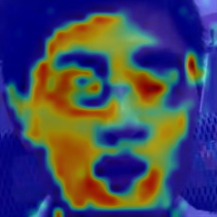}
&
\includegraphics[width=0.2\linewidth]{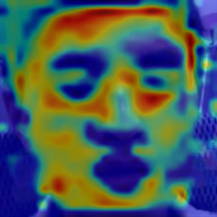}
\\
Unseen-Attack 4 & Unseen-Attack 5 & Unseen-Attack 6 & Unseen-Attack 7 
\\
\includegraphics[width=0.2\linewidth]{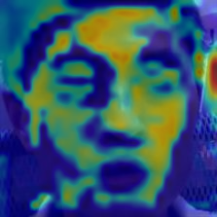}
&
\includegraphics[width=0.2\linewidth]{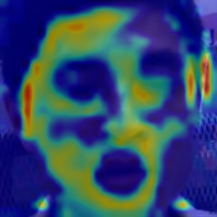}
&
\includegraphics[width=0.2\linewidth]{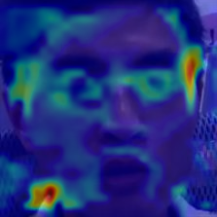}
&
\includegraphics[width=0.2\linewidth]{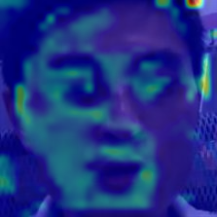}
\end{tabular}
\caption{Explanations for an example \emph{bona fide} sample with pairwise distance above the obtained average ($\bar{d}_{BF}=0.54$).}
\label{fig:interpfacebiom_pw_bf_abovemedian}
\end{figure}

\begin{figure}[!t]
\centering
\begin{tabular}{cccc}
\emph{Attack $\#7$} & Unseen-Attack 1 & Unseen-Attack 2 & Unseen-Attack 3
\\
\includegraphics[width=0.2\linewidth]{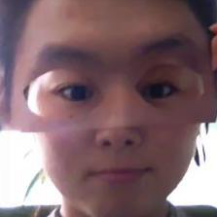}
& 
\includegraphics[width=0.2\linewidth]{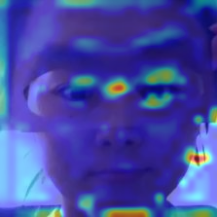}
&
\includegraphics[width=0.2\linewidth]{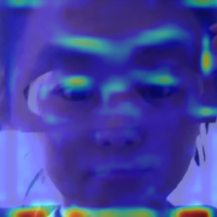}
&
\includegraphics[width=0.2\linewidth]{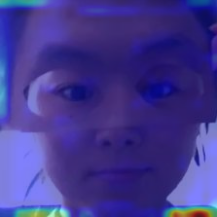}
\\
 & Unseen-Attack 4 & Unseen-Attack 5 & Unseen-Attack 6 
\\
&
\includegraphics[width=0.2\linewidth]{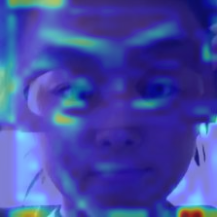}
&
\includegraphics[width=0.2\linewidth]{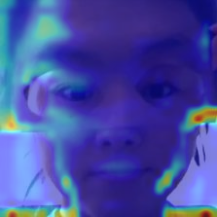}
&
\includegraphics[width=0.2\linewidth]{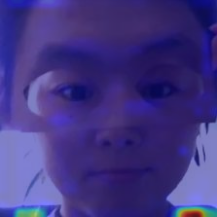}
\end{tabular}
\caption{Explanations for an example presentation attack sample of type $\#7$ with pairwise distance above the obtained average ($\bar{d}_{PA}=0.52$).}
\label{fig:interpfacebiom_pw_pa_abovemedian}
\end{figure}

Nevertheless, despite the observed variability, in both types of samples there are certain regions of the images that are consistently used by all models to make their decisions. This may denote that, despite the variety of training conditions and the resulting noise found in the explanations, the model is generally able to pinpoint some regions of the face that correspond to the real underlying label information. This idea is verified when the pairwise distance is above the median values (see Fig.~\ref{fig:interpfacebiom_pw_bf_abovemedian} and Fig.~\ref{fig:interpfacebiom_pw_pa_abovemedian}).

Naturally, this rationale based on subjective visual evaluations is limited, but this is a result of the current unavailability of a ground-truth for what is a good or meaningful explanation. These are still muddy unexplored grounds that require further research into the combination of interpretability methods and human expert knowledge in the field of biometrics.

\subsection{Intraclass comparison across different samples}

\begin{figure}[!t]
\centering
\includegraphics[width=0.75\linewidth]{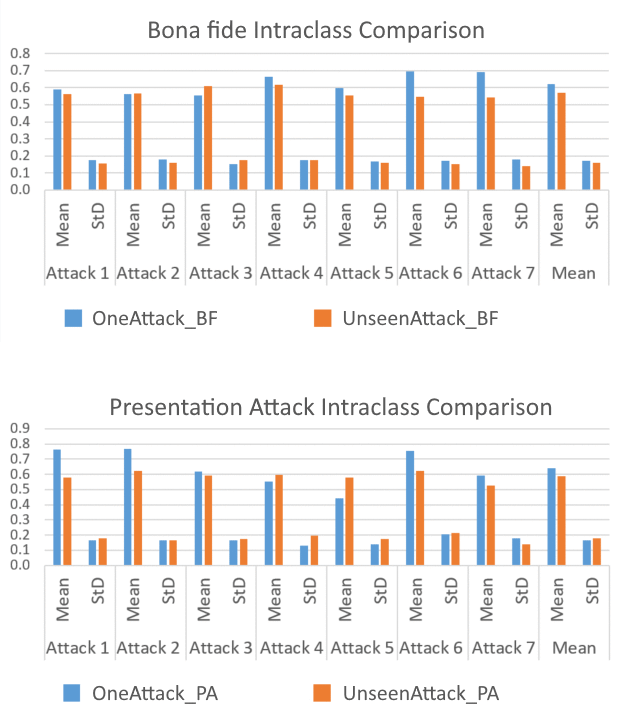} 
\caption{\emph{Bona fide} and presentation attack intraclass comparison mean and standard deviation results for the one-attack and unseen-attack ($i=1,...,7$) scenarios and respective overall results.}
\label{fig:interpfacebiom_ic_oa_ua}
\end{figure}

The intraclass comparison experiment aimed to analyse an expected property of a robust and sound PAD solution: explanations for different samples of the same class should be similar. If true, this means the model learned to behave in a coherent manner, looking for the same regions in the images to reach the same decisions. The results of this experiment are presented in Fig.~\ref{fig:interpfacebiom_ic_oa_ua}.

One can observe that the one-attack scenario (both for \emph{bona fide} and presentation attack samples) leads to overall higher variability in explanations. This confirms the idea, brought up previously in this section, that a greater variety of PAI species in the training phase will result in models that are more robust and consistent in the information they use for their decisions.

\section{Summary and Conclusions}

This study consisted of an analysis of the explanations produced for a face PAD model explored in different scenarios and settings. Both the standard one-attack and the more challenging unseen-attack scenarios were considered in this exploratory study. Additionally, intraclass and interclass experiments were also conducted to evaluate certain desirable properties of a robust and effective PAD solution. 

Focusing on the intraclass comparison of explanations' variability, it was possible to verify that the one-attack framework led to an increased mean distance value for both \emph{bona fide} and presentation attack samples. This denotes that the presence of more attacks during training has a positive effect on the generalisation capabilities of the models, despite one-attack performance metrics being generally better than those in unseen-attack.

As for the interclass comparison of explanations, it was found that both classes exhibit similar levels of variability. Further analysis of \emph{bona fide} and presentation attack samples has shown that, despite the wide variability in explanations, the models were generally able to pinpoint important regions of the face images that correspond to the true underlying labels.  

Overall, this exploratory study illustrates the deeper level of insight that can be obtained when biometric studies are combined with interpretability. If one is to overcome the elusive nature of sophisticated deep learning models and move towards more transparent approaches, it is important that evaluation frameworks allow the assessment of the quality and comparison of explanations.

It is clear that there is still much to do. Regardless of the biometric task at hand, it may be very difficult to find a consensual definition of a ``good'' explanation, or what behaviours an ideal model should exhibit. According to \citet{phillips2020four}, an explanation must describe how the system came to its conclusion and an ``accurate'' explanation (whatever that may mean) does not imply that a system provided the correct answer.

As such, the major open challenge is for the biometrics community to evolve from established evaluation metrics based on decision accuracy to novel explanation-based performance metrics. This work took one step forward in this direction. However, improving objectivity by combining subjective but knowledgeable opinions from several experts is essential to consolidate interpretability and thus enable more meaningful performance analysis on biometrics. Additionally, further efforts should be devoted to investigating the integration of the explanations on the training itself as a regularisation method to guide models through the learning of more meaningful features.

\part{Wellbeing Monitoring}\label{part:wellbeingMonitoring}
\chapter[Emotion Valence Classification in the Wild]{Emotion Valence\\Classification in the Wild}\label{ch:emotion}

\begin{tcolorbox}\footnotesize
{\large\bf Foreword on Author Contributions}

The research work described in this chapter was conducted within the Easy Ride project in collaboration with Tiago Gonçalves, Carolina Pinto, and Luís Sanhudo, under the supervision of Jaime S. Cardoso, Pedro Carvalho, Joaquim Fonseca, and Filipe Gonçalves. The author of this thesis contributed to this work on the conceptualisation and implementation of the video module, the evaluation of the unimodal and multimodal algorithms, the preparation of submissions to the EmotiW 2020 AV Group sub-challenge, and the preparation of the scientific publication.

The results of this work were disseminated in the form of an article in international conference proceedings:
\begin{itemize}[noitemsep, leftmargin=1em, nosep]
    \item \underline{J. R. Pinto}, T. Gonçalves, C. Pinto, L. Sanhudo, J. Fonseca, F. Gonçalves, P. Carvalho, and J. S. Cardoso, ``Audiovisual Classification of Group Emotion Valence Using Activity Recognition Networks,'' in \emph{Fourth IEEE International Conference on Image Processing, Applications and Systems (IPAS 2020)}, Dec.~2020.~\cite{Pinto2020Audiovisual}
\end{itemize}
This work was awarded the \emph{Best Session Paper Award} at the aforementioned international conference.

\end{tcolorbox}

\section{Context and Motivation}

Emotion recognition is a fast-growing research topic, due to its potential for enhanced human-computer interfaces and automatic services that immediately respond to the emotions of the user or client~\cite{Ferreira2018}. Horror videogames that adapt the gameplay and sound effects based on the player's fear, as well as autonomous vehicles that adapt the travel experience based on the occupants' emotions, are only two of the endless innovations attainable through emotion recognition~\cite{Mehta2018}.

State-of-the-art methods for emotion recognition are mainly based on facial expressions, and important hurdles have been overcome in this field~\cite{Mehta2018, Ferreira2018}. Group-level emotion is a fairly uncharted research topic that extends the analysis to the emotional state displayed by a group of people as a whole~\cite{Tan2017Group}. While there are several challenges in individual emotion recognition, approaches for group emotion recognition also need to deal with the variety of emotions, their valence, and arousal levels, that can differ among members of the same group. This topic was the focus of the EmotiW 2020~\cite{Dhall2020} sub-challenge that motivated this work. The scarce data and the difficulty in obtaining annotations is the reason why few have addressed this topic~\cite{Sun2016, Wei2017, Gupta2018}, and why current approaches still offer low accuracy levels.

The task of group emotion recognition shares some similarities with the recognition of human activity based on the video. Unlike the former, the latter boasts several large and thoroughly labelled datasets, such as the Kinetics~\cite{Carreira2017} or the ActivityNet~\cite{Heilbron2015}, even when restricting to data focused on groups rather than on individuals. These larger sets of available data have allowed for the development of very robust and high-performing algorithms, such as the I3D~\cite{Carreira2017}, the SlowFast networks~\cite{Feichtenhofer2019}, or the stagNet~\cite{Qi2020}. 

While methods based on visual information compose most of the literature, some works discuss the advantages of including additional sources of information, especially audio~\cite{Kazakos2019,Cosbey2019,Liang2019,Wang2019}. Specifically, it has been shown that using audio complements some of the flaws of video-based recognition~\cite{Kazakos2019}, despite offering subpar accuracy results when in a unimodal recognition system. These results have confirmed the advantages of combining audio information with a strong method for video-based recognition.

This work explores the novel application of inflated convolutional neural networks (CNN) to classify emotion valence at the group level in videos. The network uses weights pretrained for activity recognition, to take advantage of the greater availability of data to boost performance on our target task. We also study the use of audio for improved performance, through score-level fusion, with a Bi-LSTM network receiving spectral features. Throughout the experiments, we assess the performance of the proposed method for multimodal and unimodal classification, analyse its behaviour in different scenarios, and compare it directly with the EmotiW 2020 sub-challenge official baseline.

\section{Methodology}

\subsection{General overview}
\begin{figure}[t]
  \centering
  \includegraphics[width=\textwidth]{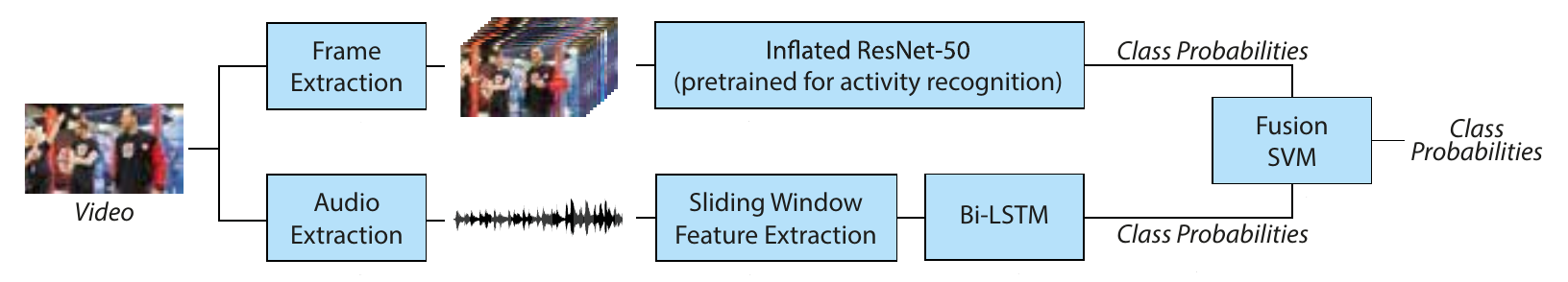}
  \caption[Illustration of the structure of the proposed method for audiovisual group emotion recognition.]{Illustration of the structure of the proposed method for audiovisual group emotion recognition (the proposed methodology for group emotion recognition processes a video in two streams: one processes concatenated video frames using an inflated ResNet-50 pretrained on a large activity recognition dataset, and the other extracts sliding window features from the audio and processes them using a bi-directional long short-term memory (Bi-LSTM) network; a support vector machine (SVM) classifier receives class probabilities from each stream and returns a final class prediction).}
  \label{fig:emotionval_teaser}
\end{figure}

The proposed algorithm is composed of three modules: a video-based emotion recognition model, an audio-based emotion recognition model, and a multimodal fusion module (see Fig.~\ref{fig:emotionval_teaser}). Video and audio-based emotion recognition modules are trained independently, while the fusion module, based on a multiclass SVM receives the softmax scores provided by the other two. Thus, the proposed method consists of a pipeline that relies on late audio-video fusion, at the score level, using a multi-class SVM emotion recognition classifier.

\subsection{Video-based emotion recognition}
\begin{figure}[!t]
  \centering
  \includegraphics[width=0.65\linewidth]{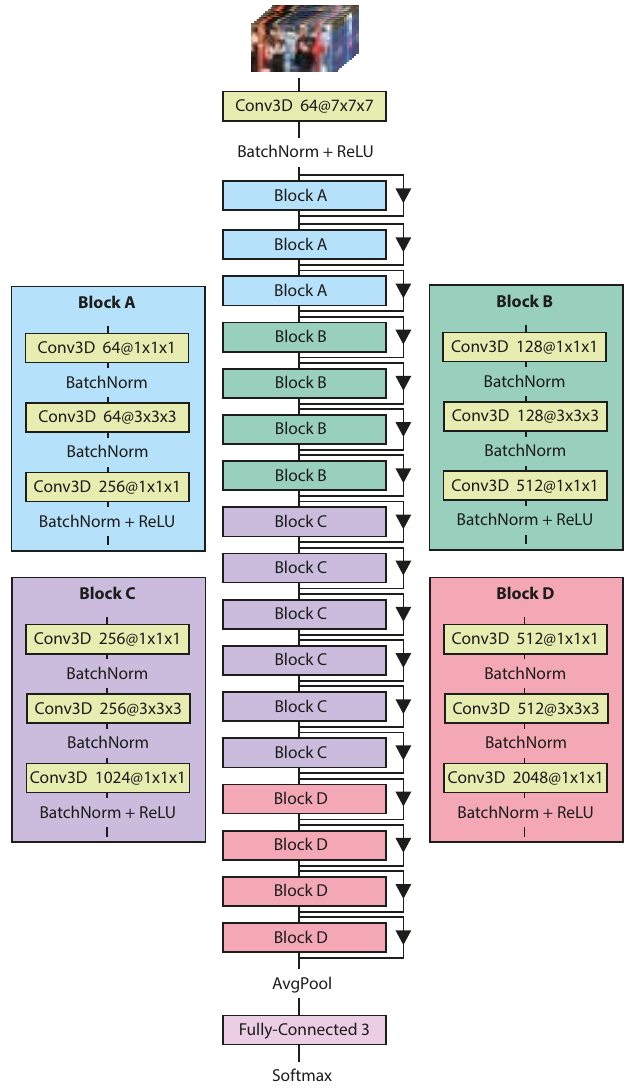}
  \caption{Structure of the video-based group emotion recognition module, based on an inflated ResNet-50.}\label{fig:emotionval_structure_video}
\end{figure}

The video-based emotion recognition module is based on an inflated bidimensional (2D) convolutional neural network (CNN), similar to I3D~\cite{Carreira2017}, the state-of-the-art in activity recognition. The model is an end-to-end network: it receives frames extracted from a video, ordered and concatenated over a time dimension, and returns class probabilities for that video. The architecture of the network follows the structure of a ResNet-50 (see Fig.~\ref{fig:emotionval_structure_video}), proposed by He~\etal~\cite{He2016}, whose name stands for residual networks. The shortcut connections that perform identity mapping on each residual learning block enable the stable training of models with more convolutional layers, resulting in deeper representations of the input data.

The inflated ResNet-50 consists of a bidimensional ResNet-50 model where the convolutional filters and layers have been converted into 3D. This allows them to process several frames simultaneously as a single input. Downsampling operation before the first block of each type enables learning multi-resolution features. This model has been pretrained\footnote{Multi-Moments in Time models. Available at: \url{https://github.com/zhoubolei/moments_models}.} to discriminate between $339$ activity classes on the Multi-Moments In Time database~\cite{Monfort2019}. To offer probability outputs for each of the three group-level emotion valence classes, the last fully-connected layer of this network is replaced by a three-neuron fully-connected layer, followed by softmax activation, trained on the EmotiW 2020 sub-challenge train dataset.

\subsection{Audio-based emotion recognition}

The audio-based recognition module (see Fig.~\ref{fig:emotionval_teaser}) is composed of two main processes: feature extraction on sliding windows, and a Bi-LSTM recognition model. Audio features were extracted using pyAudioAnalysis\footnote{pyAudioAnalysis. Available at: \url{https://github.com/tyiannak/pyAudioAnalysis}.} which contains an off-the-shelf feature set with 34 available features, including signal zero-crossing rate, signal energy, entropy of energy, spectral centroid, spectral spread, spectral entropy, spectral flux, spectral roll-off, mel-frequency cepstral coefficients (MFCC), chroma vector, and chroma deviation. All these features are extracted over sliding windows of $25$ milliseconds with a time step of $10$ milliseconds.

\begin{figure}[!t]
  \centering
  \includegraphics[width=0.65\linewidth]{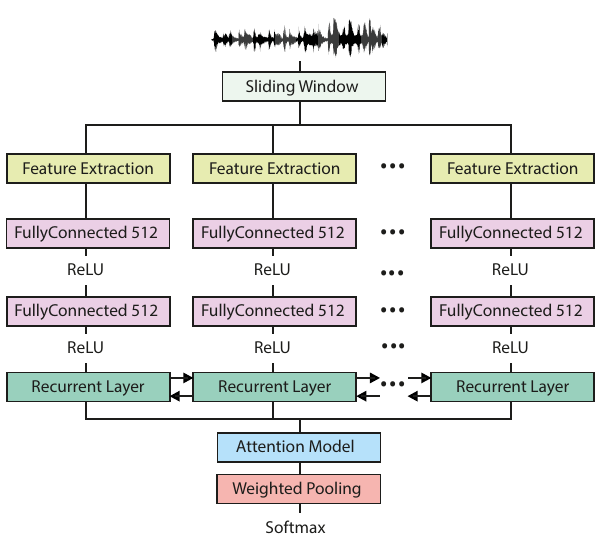}
  \caption[Structure of the audio-based group emotion recognition module, based on a Bi-LSTM network.]{Structure of the audio-based group emotion recognition module, based on a Bi-LSTM network~\cite{mirsamadi2017automatic}.}\label{fig:emotionval_structure_audio}
\end{figure}

The features are received by a Bi-LSTM model with local attention that returns the class probabilities for the respective audio (see Fig.~\ref{fig:emotionval_structure_audio}), adapted from~\cite{mirsamadi2017automatic}. Its weighted-pooling strategy enables the focus on the specific sound parts which contain strong emotional characteristics, controlled by an attention function trained simultaneously with the Bi-LSTM model.

\subsection{Score-level ensemble}
The softmax scores obtained by both video and audio-based emotion recognition models are then concatenated in a feature vector, composed of six class probability values. This vector is then given to a multi-class SVM classifier, which combines the separate audio and video predictions into a single decision.

\section{Experimental Setup}

\subsection{Data}
All the experiments were conducted on an adapted version of the ``Video-level Group AFfect'' (VGAF) dataset~\cite{sharma2019automatic} for the EmotiW 2020 AV Group-level sub-challenge~\cite{Dhall2020}. The VGAF is a video-based database that contains labels for emotion and cohesion. The data was collected from the YouTube platform and consists of videos under the creative commons license (CC0) and present keywords that correspond to the range of emotions and cohesion.

Since the number of individuals per video is variable, and the groups on each video can also present a varying number of persons over time, the videos have been divided so that each video clip has always the same number of persons per frame. Each VGAF clip was manually labelled by different annotators for emotion and cohesion and every annotator was informed of the basic concepts of emotion and cohesion. Only videos with mutual consensus were kept in the final database. The labels for group emotion are related to emotion valence (\textit{i.e.}, positive, neutral, and negative) whereas the group cohesion labels are in the range $[0-3]$, being $0$ the state of very low cohesion (dominance over the group members) and $3$ the state of very high cohesion.

For the EmotiW 2020 AV Group-Level Emotion sub-challenge, the task was the classification of group emotion. The VGAF dataset videos were divided into five-second videos: $2661$ for the train, $766$ for the validation, and $756$ for the test. Each video (except those in the test set) is accompanied by a discrete group emotion valence ground-truth label. In the training dataset, there are $802$ positive videos, $923$ neutral videos, and $936$ negative videos. In the validation set, there are $302$ positive videos, $280$ neutral videos, and $184$ negative videos.

\subsection{Baseline algorithm}

The baseline algorithm is the audiovisual group-level emotion recognition sub-challenge baseline~\cite{sharma2019automatic} of the EmotiW 2020 Grand Challenge. This method is composed of two streams, for audio and video data processing, fused at the feature level. The video stream is a pretrained Inception V3 network that separately processes frames extracted from a video. The extracted features are combined using a long short-term memory (LSTM) network. The audio stream is composed of a fully-connected network that receives OpenSMILE~\cite{eyben10real} features extracted from the audio. The outputs from the video and audio streams are concatenated and used by a fully-connected layer to offer two outputs: the probabilities for the three emotion valence classes and an emotion cohesion value on the $[0-3]$ range.

\subsection{Preprocessing}
The videos on the EmotiW 2020 AV Group-level Emotion sub-challenge were subject to preprocessing before being used by the proposed method. For each five-second video, ten frames were extracted, thus resulting in 2 frames per second. This is an adaptation from the original model pretrained on the Multi-Moments In Time database (MMIT), which worked at 5 frames per second. We found that reducing the frame rate did not harm performance and sped up the recognition process. Before being used on the audio-based recognition module, the audio was extracted from each file by converting them (originally in the MP4 format) to audio files (in the WAV format). Regarding audio, we noticed that after feature extraction some of the generated features could assume non-number values. For training purposes, we removed these samples and trained only with valid ones. For inference during validation/test, we replaced non-number values with zero.

\subsection{Training}
The audio-based recognition module was trained from scratch\footnote{Audio submodule code adapted from: \url{https://github.com/RayanWang/Speech_emotion_recognition_BLSTM}.}. The weights were randomly initialised, and the model was trained over a maximum of $200$ epochs, with a batch size of $128$, categorical-cross-entropy as the loss function, and using the Adam optimiser with an initial learning rate of $10^{-2}$. To prevent overfitting, we used dropout and early-stopping with a patience value of 15 epochs.

The video-based recognition module, pretrained on the MMIT database, was adapted to output probabilities for each of the three valence classes in group emotion recognition. This was achieved by replacing the last fully-connected layer with a new one, with three neurons.

Since this layer needs to be trained, all weights of the network have been frozen (except those of this layer). The network was briefly fine-tuned until convergence over a maximum of $250$ epochs, with batch size $32$, using the Adam optimiser with an initial learning rate of $10^{-5}$.

When training the audio-based module and the video-based module the hyperparameters have been selected empirically, to maximise performance in the validation set. The hyperparameters for the fusion module (\textit{e.g.}, the regularisation parameter ``C'', the kernel, or the polynomial degree of the kernel) were found through a grid search. Once the optimal hyperparameters were found, the train and validation set were combined to make a ``full train'' set, and thus take full advantage of all available labelled data for better performance in the test set.

\subsection{Experiments}
In this work, the aim is not only to assess the proposed method's performance for group-level emotion recognition but also to examine its behaviour in several conditions.

We use the accuracy metric and confusion matrices to examine the overall performance of the method and also analyse its class-wise accuracy. The performances of the multimodal method and its audio and video-based modules, separately, are evaluated in both the validation set (with available ground-truth labels) and the test set (accuracy values delivered by the EmotiW 2020 sub-challenge organisation upon request). The performance is compared with the official sub-challenge baseline, following the results reported in~\cite{sharma2019automatic}.

The performance is also evaluated according to the number of people in the video. Since the number of people in each video is not included, we use the MTCNN method~\cite{Zhang2016} on each frame of each video, and infer the group size based on the average number of detected faces: a group with less than five detected faces are considered small (total of $502$ videos on the validation set), otherwise, it is considered a large group ($264$ videos on the validation set). With this, we aim to evaluate the difficulties associated with recognising emotion in large groups, where cohesion is likely to be generally lower.

\section{Results and Discussion}

The performance results of the proposed method, and the comparison with the official sub-challenge baseline, on the validation and test sets, is presented, respectively, in Table~\ref{tab:emotionval_accuracy_val} and Table~\ref{tab:emotionval_accuracy_test}.

On the validation set, the performance offered by the proposed multimodal method is superior to the baseline. The accuracy attained by the video-only approach is close to that offered by the multimodal method, over $62\%$. This is evidence of the advantages of using pretrained networks (in this case, transferred from the task of human activity recognition). The audio-only approach offers considerably lower performance ($47\%$) than the audio-only baseline ($50\%$), which indicates the use of OpenSMILE features and fully-connected networks may be better fitted for group emotion recognition based on audio.

\begin{table}[!t]
  \centering
  \caption[Accuracy (\%) of the proposed method on the validation set.]{Accuracy (\%) of the proposed method on the validation set (A - only audio; V - only video; A+V - multimodal).}
  \label{tab:emotionval_accuracy_val}
  \begin{tabular}{lcccc}\hline
    \multirow{2}{*}{\textbf{Method}} & \multicolumn{4}{c}{\textbf{Accuracy}} \\
    & \textbf{\emph{Overall}} & \textbf{\emph{Positive}} & \textbf{\emph{Neutral}} & \textbf{\emph{Negative}} \\\hline
    Proposed (A) & 47.19 & 19.93 & \textbf{69.64} & 57.61\\
    Proposed (V) & \textbf{62.40} & \textbf{69.20} & 52.50 & 66.30\\
    Proposed (A+V) & 61.83 & 58.13 & 61.78 & \textbf{67.93}\\
    Baseline (A) & 50.23 & - & - & - \\
    Baseline (V) & 52.09 & - & - & -\\
    Baseline (A+V) & 50.23 & - & - & -\\\hline
\end{tabular}
\end{table}

\begin{table}[!t]
  \centering
  \caption[Accuracy (\%) of the proposed method on the test set.]{Accuracy (\%) of the proposed method on the test set (A - only audio; V - only video; A+V - multimodal).}
  \label{tab:emotionval_accuracy_test}
  \begin{tabular}{lcccc}\hline
    \multirow{2}{*}{\textbf{Method}} & \multicolumn{4}{c}{\textbf{Accuracy}} \\
    & \textbf{\emph{Overall}} & \textbf{\emph{Positive}} & \textbf{\emph{Neutral}} & \textbf{\emph{Negative}} \\\hline
    Proposed (V) & 58.86 & \textbf{55.76} & 57.93 & 63.04\\
    Proposed (A+V) & \textbf{65.74} & 54.38 & \textbf{77.99} & 60.00\\
    Baseline (V) & 42.00 & - & - & -\\
    Baseline (A+V) & 47.88 & 45.00 & 10.00 & \textbf{70.00} \\\hline
\end{tabular}
\end{table}

From the validation to the test set (Table~\ref{tab:emotionval_accuracy_test}), the official video-only baseline suffers a sharp performance decay (from $52\%$ to $42\%$ accuracy), which is also felt with the proposed video-only approach (albeit not as dramatic, from $62\%$ to $59\%$ accuracy). Fusing with audio on a multimodal approach reduced that decrease in the case of the official baseline (from $50\%$ to $48\%$ accuracy), and even reversed it in the case of the proposed method (from $62\%$ to $66\%$ accuracy). This confirms the idea present in the literature that, while audio alone is not suitable for recognition, it offers additional information that is essential for the robustness and accuracy of the method.

\begin{table}[!t]
  \centering
  \caption{Confusion matrix of audio-based recognition on the validation set.}
  \label{tab:emotionval_cm_audio}
  \begin{tabular}{lccc}\hline
    & \multicolumn{3}{c}{\textbf{Predicted Class}}\\
    \textbf{True Class} & \textbf{\emph{Positive}} & \textbf{\emph{Neutral}} & \textbf{\emph{Negative}}\\\hline
    \textbf{\emph{Positive}} & 60 & 139 & 102\\
    \textbf{\emph{Neutral}} & 35 & 195 & 50\\
    \textbf{\emph{Negative}} & 24 & 54 & 106\\\hline
\end{tabular}
\end{table}
 
\begin{table}[!t]
  \centering
  \caption{Confusion matrix of video-based recognition on the validation set.}
  \label{tab:emotionval_cm_video}
  \begin{tabular}{lccc}\hline
    & \multicolumn{3}{c}{\textbf{Predicted Class}}\\
    \textbf{True Class} & \textbf{\emph{Positive}} & \textbf{\emph{Neutral}} & \textbf{\emph{Negative}}\\\hline
    \textbf{\emph{Positive}} & 209 & 59 & 34\\
    \textbf{\emph{Neutral}}  & 98 & 147 & 35\\
    \textbf{\emph{Negative}} & 44 & 18 & 122\\\hline
\end{tabular}
\end{table}
 
\begin{table}[!t]
  \centering
  \caption{Confusion matrix of multimodal recognition on the validation set.}
  \label{tab:emotionval_cm_multi}
  \begin{tabular}{lccc}\hline
    & \multicolumn{3}{c}{\textbf{Predicted Class}}\\
    \textbf{True Class} & \textbf{\emph{Positive}} & \textbf{\emph{Neutral}} & \textbf{\emph{Negative}}\\\hline
    \textbf{\emph{Positive}} & 175 & 99 & 27\\
    \textbf{\emph{Neutral}} & 78 & 173 & 29 \\
    \textbf{\emph{Negative}} & 27 & 32 & 125 \\\hline
\end{tabular}
\end{table}

Analysing the class-wise accuracies and the confusion matrices (Table~\ref{tab:emotionval_cm_audio}, Table~\ref{tab:emotionval_cm_video}, and Table~\ref{tab:emotionval_cm_multi}), one can notice that video is, overall, the best modality to recognise emotions. The advantages of using video rely mainly on the ``extreme'' classes, positive and negative, which denote visual information is more advantageous to recognise strong group emotions. The proposed audio-only approach attains very poor accuracy in the positive class.

Since the positive class is the minority class in the training dataset, the results of the audio-only approach may partially be explained by this slight class imbalance. However, as mentioned before, the video-only approach does not verify this, which is fortunate when combining both approaches into the multimodal proposed method. Using both modalities slightly decreases the accuracy of positive and negative videos, when compared with the video-only approach, but takes advantage of the audio information to considerably improve accuracy on neutral videos and achieve overall better performance.

At last, the results of the group size study are presented in Table~\ref{tab:emotionval_accuracy_groups}. In both audio and video-only approaches, as well as the multimodal method, the recognition performance is higher in smaller groups. The performance values should serve as a rough reference, since the process of face detection may present errors, and the number of faces may not accurately describe the number of people in the video's group (which may include occluded faces or people facing the opposite direction of the camera).

Nevertheless, the performance differences are considerable and show expected behaviour: it should be harder for larger groups to consistently show the same emotion than smaller groups. Hence, emotion cohesion should be higher, on average, for smaller groups, and thus the certainty of the algorithms when recognising the emotion valence. This could perhaps be addressed using hierarchical methodologies (from individual-level to group-level) as used in current group activity recognition approaches (discussed in the related work section).

\begin{table}[t]
  \centering
  \caption[Accuracy (\%) on the validation set for videos of small groups \emph{vs.} large groups.]{Accuracy (\%) on the validation set for videos of small groups \emph{vs.} large groups (A - only audio; V - only video; A+V - multimodal).}
  \label{tab:emotionval_accuracy_groups}
  \begin{tabular}{ccc}\hline
    \multirow{2}{*}{\textbf{Method}} & \multicolumn{2}{c}{\textbf{Group Size}}\\
    & \textbf{$N<5$} & \textbf{$N\geq5$} \\\hline
    Proposed (A)   & 48.61 & 44.49 \\
    Proposed (V)   & 66.33 & 54.92 \\
    Proposed (A+V) & 65.74 & 54.37 \\\hline
\end{tabular}
\end{table}

\begin{figure}[t]
  \centering
  \includegraphics[width=0.48\linewidth]{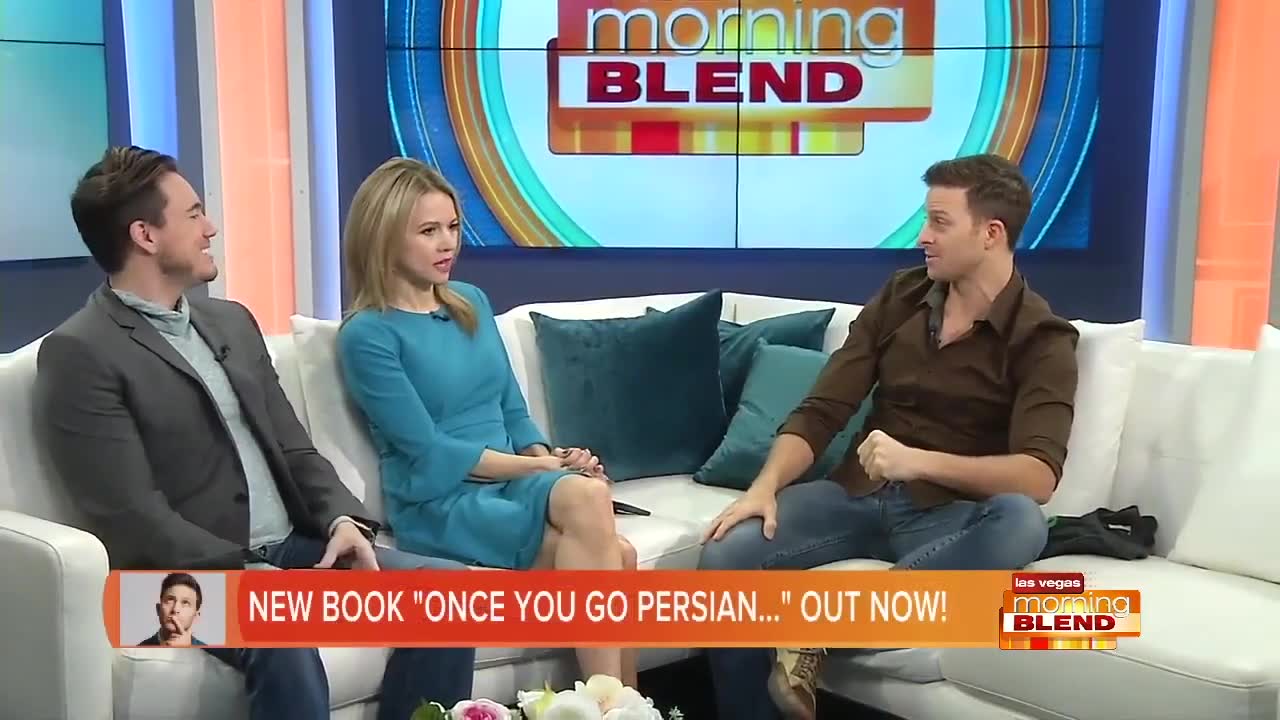}
  \includegraphics[width=0.48\linewidth]{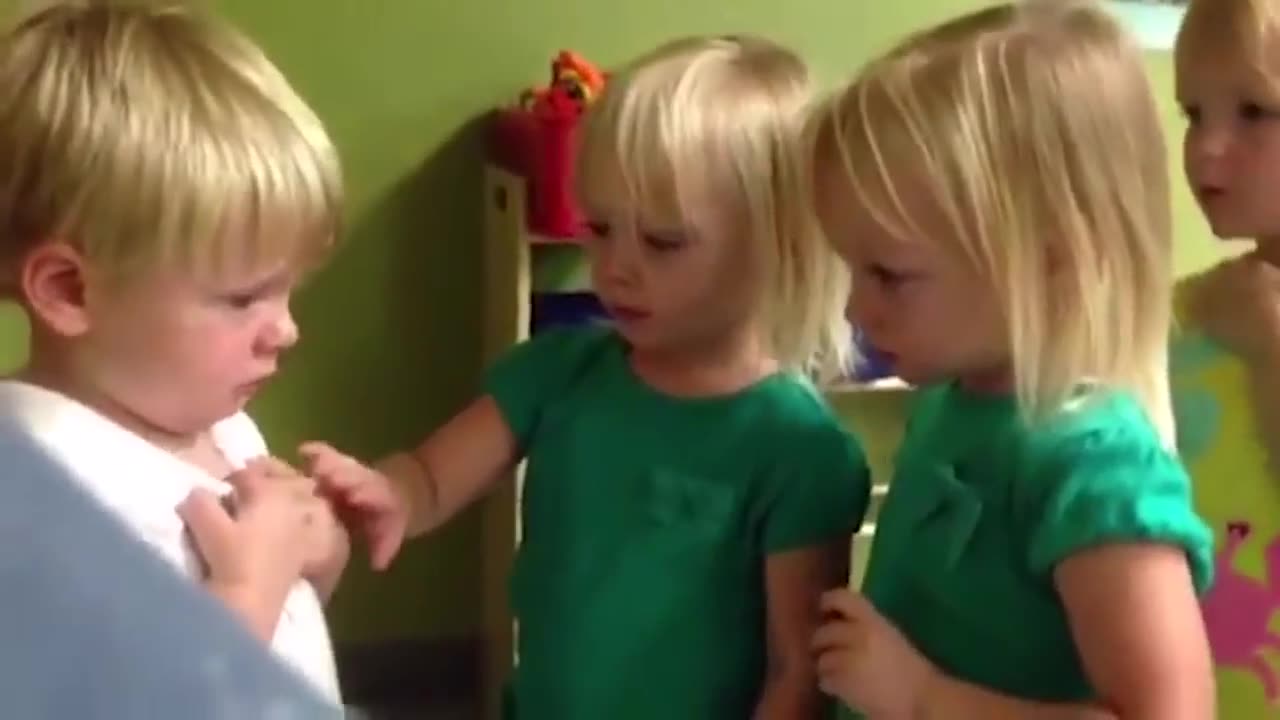}\\
  \includegraphics[width=0.48\linewidth]{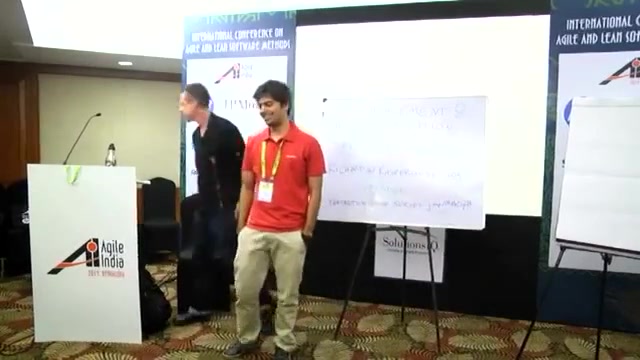}
  \includegraphics[width=0.48\linewidth]{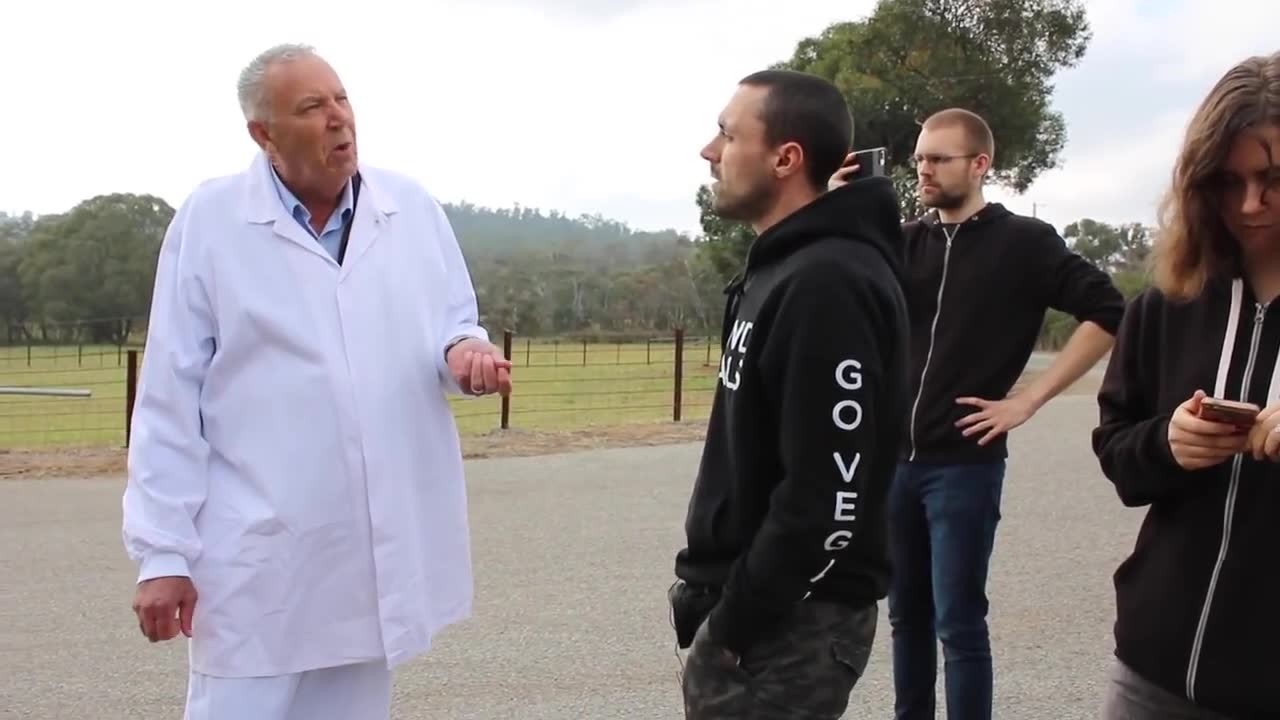}
  \caption[Some examples of validation set videos where the model offered unsuccessful predictions.]{Some examples of validation set videos where the model offered unsuccessful predictions (top left: label positive, predicted neutral; top right: label negative, predicted positive; bottom left: label positive, predicted neutral; bottom right: label negative, predicted neutral).}\label{fig:emotionval_examples}
\end{figure}

Through an analysis of some videos where the proposed model failed, a pattern emerged. While there are certain videos where the error was evident, there are several examples where it is very difficult to notice that an apparently neutral scene displays, in fact, a positive or negative group emotion.

Some examples are shown in Fig.~\ref{fig:emotionval_examples}. On the top left, is a short video of a calm conversation on a TV show that is labelled as positive. On the top right, is a negative emotion video that the proposed method classified as positive, since the video only covers the moment before the boy being bullied started crying. On the bottom left, a conference presentation that the method classified as neutral since the positive ground-truth emotion could only be verified by facial expressions. On the bottom right, a conversation is deemed neutral by the proposed method, where only a closer inspection of the audio shows that it is, in fact, part of a protest or a similar confrontation.

Although the information about the underlying ground-truth labels is indeed present in these example videos, it is hidden in contextual clues, expressions in small faces, or the content of conversations. Exploring ways to integrate these aspects into the recognition of group-level emotion could be the way to avoid the most common mistakes of the proposed method, and ultimately achieve better overall performance and robustness.

\section{Summary and Conclusions}

In this work we proposed a novel method for the automatic recognition of group emotion that uses a late-fusion multimodal approach, combining scores from both video and audio-based emotion recognition models that are used to feed a multiclass SVM that returns a final class probability. This method showed significant improvement against the baseline, confirming that the use of acquired knowledge from activity recognition is useful for group-emotion recognition and that the joint utilisation of audio and video benefits the learning of the model.

On the other hand, taking into account the maximum accuracy value, we believe that there is still room for further improvements. Further efforts should be devoted to the study of the links between the tasks of activity recognition and emotion recognition, especially at the group level. Approaches for abnormal behaviour recognition through video anomaly detection, such as~\cite{Augusto2020}, could offer meaningful improvements in the automatic distinction between positive and negative emotions. Also, this method could benefit from OpenSMILE features in the audio-based emotion recognition module, the development of multimodal approaches that are based on early-fusion (\textit{e.g.}, input or intermediate-layer levels), and the design of a ``fully'' end-to-end network that receives both video and audio as input and learns the relevant features for the classification task (\textit{e.g.}, through regularisation methods such as loss functions with different terms and weights).
\chapter[Activity and Violence Recognition in Shared Vehicles]{Activity and Violence\\Recognition in Shared Vehicles}\label{ch:activity}

\begin{tcolorbox}\footnotesize
{\large\bf Foreword on Author Contributions}

The research work described in this chapter was conducted in collaboration with Carolina Pinto, Afonso Sousa, and Leonardo Capozzi, under the supervision of Jaime S. Cardoso and Pedro Carvalho. The author of this thesis contributed to this work on the reformulation and implementation of the audio module, the conceptualisation and implementation of the cascade strategy, the preparation of data and the experimental setup, the evaluation of the parallel and cascade algorithms, the discussion of the results, and the preparation of the scientific publication.

The results of this work were disseminated in the form of an article in international conference proceedings:
\begin{itemize}[noitemsep, leftmargin=1em, nosep]
    \item \underline{J. R. Pinto}, P. Carvalho, C. Pinto, A. Sousa, L. G. Capozzi, and J. S. Cardoso, ``Streamlining Action Recognition in Autonomous Shared Vehicles with an Audiovisual Cascade Strategy,'' in \emph{17th International Conference on Computer Vision Theory and Applications (VISAPP)}, Feb.~2022.~\cite{Pinto2022Streamlining}
\end{itemize}

\end{tcolorbox}

\section{Context and Motivation}\label{sec:activityrec_introduction}

Human action or activity recognition is a vibrant and challenging research topic. Being able to recognise actions automatically is game-changing and often crucial for several industries, including the scenario of shared autonomous vehicles. Without a driver responsible for the vehicle's and occupant's security and integrity, it falls upon automatic recognition systems to monitor passenger well-being and actions, and eventually recognise harmful behaviours or even violence~\cite{Augusto2020}. However, the wide range of possible actions that can be portrayed, the variability in the way different individuals portray the same actions, the heterogeneity of sensors and the type of information captured and the influence of external factors still pose significant hurdles to this task.

Despite all the above-mentioned challenges, the topic of action recognition has thrived by following a very recognisable recipe for success. As in plenty of other pattern recognition tasks, the state-of-the-art gradually evolved towards larger and more sophisticated models based on deep learning methodologies~\cite{Carreira2017,Feichtenhofer2019,Qi2020}. These have achieved increasingly higher accuracy thanks to a growing number of massive databases typically using public video data gathered through online sourcing, such as Kinetics~\cite{Carreira2017}, Multi-Moments in Time (MMIT)~\cite{Monfort2019}, or ActivityNet~\cite{Heilbron2015}. This also means most research in action recognition is based on visual information (images or video). This is the case of the I3D~\cite{Carreira2017}, the methodology currently deemed the state-of-the-art in this topic. In fact, I3D goes further beyond simple visual spatial information by adopting a two-stream approach, including optical flow for temporal action encoding. Other approaches have explored recurrent networks for the same purpose~\cite{Kong2018,Pang2019,Hu2018a}, but have seldom managed to reach the accuracy level offered by the I3D method.

Despite the meaningful strides brought by such sophisticated methods and large databases, some limitations can be observed. On the one hand, the general nature of the data sourced to train and evaluate the state-of-the-art models lead to overly general results that may not be verified in more specific scenarios, such as in-vehicle passenger monitoring. On the other hand, hefty models based on visual information and optical flow (such as I3D) may offer very high accuracy, but their complexity does not allow for real-time applications in inexpensive limited hardware, such as embedded devices.

This work proposes a set of changes to the state-of-the-art I3D method to bring it closer to real applicability in edge computing scenarios: in this case, we focus on action recognition and violence detection in shared autonomous vehicles. First, inspired by~\cite{Pinto2020Audiovisual}, the current work discards the time-consuming optical flow component of I3D and introduces a lightweight model for action recognition with audio. Despite being less frequently used than video, audio is considered one of the most promising options for a multimodal system for action recognition~\cite{Kazakos2019,Cosbey2019,Liang2019}. This way, we obtain a simpler methodology that can use both video and audio modalities for a greater variety of information. Then, as each modality is likely to contribute differently to the recognition of each action, we propose a cascade strategy based on confidence score thresholding. This strategy allows a simplification of the multimodal pipeline by using only one (primary) modality as often as possible; the two modalities are used together only when the primary one is not enough for sufficiently confident predictions. Hence, it is possible to attain significant time and computing energy savings without overlooking classification accuracy.

This chapter is organised as follows: beyond this introduction, a description of the proposed multimodal methodology and cascade strategy is presented in section~\ref{sec:activityrec_parallel_vs_cascade}; the experimental setup is detailed in section~\ref{sec:activityrec_experiments}; section~\ref{sec:activityrec_results} presents and discusses the obtained results; and the conclusions drawn from this work are presented in section~\ref{sec:activityrec_conclusion}.

\section{Methodology}\label{sec:activityrec_parallel_vs_cascade}

\subsection{Multimodal pipeline}

\begin{figure}[!t]
\centering
\includegraphics[width=0.5\linewidth]{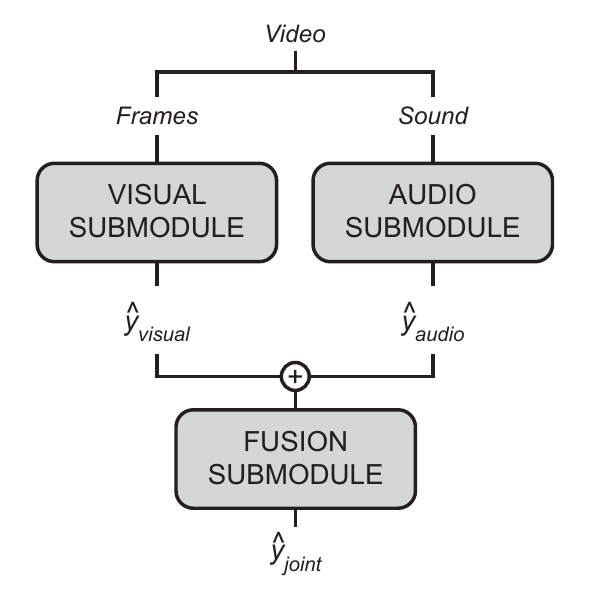}
\caption{Diagram of the full multimodal pipeline for activity recognition.}
\label{fig:activityrec_block_diagram_pipeline}
\end{figure}

The baseline consists of a multimodal pipeline for activity recognition based on an audio-visual module previously proposed for group emotion recognition~\cite{Pinto2020Audiovisual}. The pipeline is composed of three submodules (as illustrated in Fig.~\ref{fig:activityrec_block_diagram_pipeline}): the visual submodule, which processes visual data; the audio submodule, which processes sound data; and the fusion submodule, which combines individual decisions from the previous two submodules into joint multimodal classifications. The specific structures of each of these submodules are described below.

\subsubsection{Visual submodule}
As in~\cite{Pinto2020Audiovisual}, the visual submodule is based on an inflated ResNet-50~\cite{He2016} using pretrained weights for the Multi-Moments in Time (MMIT) activity recognition database~\cite{Monfort2019}. Using an inflated ResNet-50 ensures optimal performance by following the successful example of the state-of-the-art I3D method~\cite{Carreira2017}. Using model weights pretrained on the large MMIT database allows us to transfer deeper and more general knowledge to our narrower task of activity recognition inside vehicles. 

\begin{figure}[!t]
\centering
\includegraphics[width=0.5\linewidth]{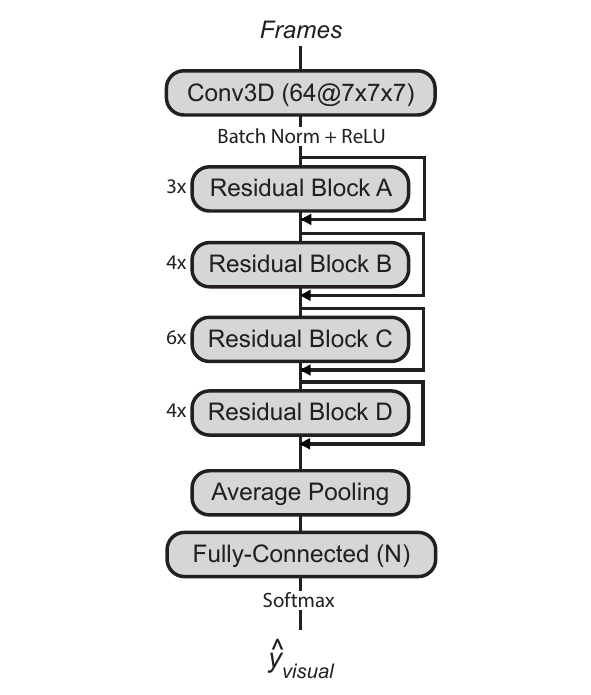}
\caption[Diagram of the visual submodule.]{Diagram of the visual submodule (more details on the ResNet-50 and the residual blocks in~\cite{He2016}).}
\label{fig:activityrec_block_diagram_video}
\end{figure}

The inflated ResNet-50 model (see Fig.~\ref{fig:activityrec_block_diagram_video}) is composed of seventeen residual blocks, each including three 3D convolutional layers with $64$ to $2048$ filters, batch normalisation and ReLU activation. Downsampling at each block allows the model to capture important information at different levels of resolution. After an average pooling layer, the last fully-connected layer, followed by a softmax activation function, offers probability scores for each of the $N$ considered activity labels.

\subsubsection{Audio submodule}
The audio submodule consists of a simple network based on a bi-directional long short-term memory (LSTM) model. These are known for their ability to encode temporal information, important for audio-related topics, and have been previously successful for tasks such as group emotion valence recognition~\cite{Pinto2020Audiovisual} or speech-based sentiment analysis~\cite{mirsamadi2017automatic}.

Unlike the visual submodule, which largely follows the method proposed in~\cite{Pinto2020Audiovisual} to approach the state-of-the-art performance of I3D, the audio submodule was reformulated. In the aforementioned work, the audio Bi-LSTM model received a set of cepstral, frequency, and energy handcrafted features extracted from each signal window. Moreover, it included multiple convolutional layers with $512$ filters each and an attention mechanism after the LSTM layer. In this work, we design a streamlined and faster audio submodule.

\begin{figure}[!t]
\centering
\includegraphics[width=0.5\linewidth]{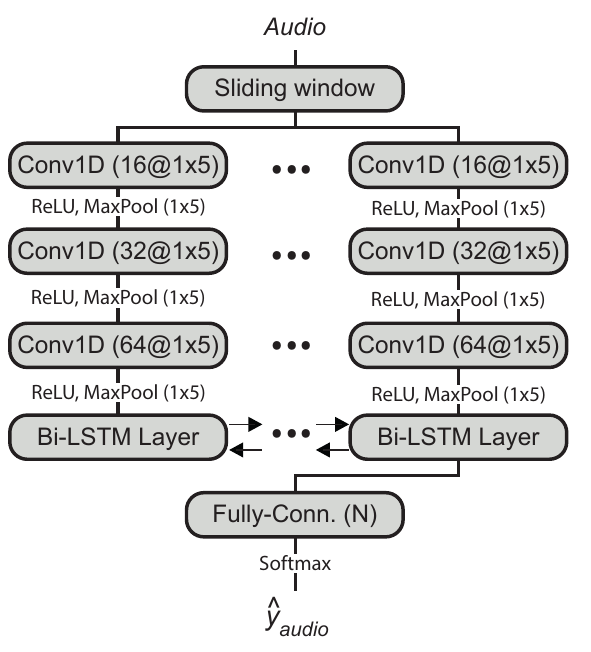}
\caption{Diagram of the audio submodule.}
\label{fig:activityrec_block_diagram_audio}
\end{figure}

The simplified and lighter Bi-LSTM model (see Fig.~\ref{fig:activityrec_block_diagram_audio}), with less trainable parameters, receives a raw audio signal divided into $100$ ms windows with $50$ ms overlap, without any preceding process of feature extraction. Each window is processed by three convolutional layers (with $16$, $32$, and $64$ $1\times5$ filters, respectively, stride $1$, and padding $2$), each followed by ReLU activation and max-pooling (with pooling size $5$). A Bi-LSTM layer receives features from the convolutional part for each window, and its output for the last window is sent to a fully-connected layer for classification (with $N$ neurons, one for each activity class, followed by softmax activation). In section~\ref{sec:activityrec_results}, we analyse the advantages of using the proposed audio submodule \emph{vs.} the one in~\cite{Pinto2020Audiovisual}.

\subsubsection{Fusion submodule}
The aforedescribed visual and audio submodules output their respective sets of class probability predictions for a given video. To combine the two separate sets of predictions for each task into a single audio-visual multimodal classification, the fusion submodule is used.

The fusion submodule is composed of a simple support vector machine (SVM) classifier. This classifier receives the probability score sets from the two previous submodules concatenated as a single unidimensional feature vector. The SVM model is trained to use these probability sets to output a joint class prediction for the respective video.

\subsection{Cascade strategy}

In the multimodal pipeline described above, all submodules are used for each instance (video) that needs to be classified. This means that regardless of the difficulty of a given video or the activity portrayed, both visual and sound data are always processed, resulting in two sets of unimodal class predictions which are then combined into a set of multimodal class probabilities.

Given the considerable complexity of both the residual-network-based visual submodule and the Bi-LSTM-based audio submodule, this multimodal pipeline is arguably too heavy for the target application. This is especially true considering, as observed in~\cite{Pinto2020Audiovisual}, that different classes may benefit much more from one of the modalities and thus not need the other one. Hence, we design a cascade strategy to explore the possibility of using just one of the modalities and ``turning off'' the remaining two submodules as often as possible. This aimed to achieve improved processing times and energy usage, offering an alternative or complement to model compression strategies.

In the proposed cascade strategy, one of the data modalities (visual or audio) is selected as the ``primary'' modality and, as such, the corresponding submodule is always used to offer a starting prediction. The probability score offered for the predicted class is considered a ``confidence score'': a measure of how confident the primary submodule is in the prediction it provided. If the confidence score is above a specific confidence threshold $T \in [0,1]$, the remaining modules remain unused, and the primary submodule predictions are considered final. However, if the aforementioned condition is not verified, the secondary submodule is called to offer additional information for more confident predictions, which are then combined into a single multimodal prediction (just like the original multimodal pipeline). 

The performance benefits of such a strategy are intimately related to the defined confidence threshold. If $T$ is too high, most of the instances will use both data sources, thus retaining the accuracy offered by the parallel pipeline but reaping very few benefits related to complexity or processing time. Conversely, if $T$ is too low, most instances will be classified using only the primary submodule, which may result in heavily impacted accuracy, despite the complexity benefits of the simplified pipeline. Section~\ref{sec:activityrec_results} includes thorough experimental results on the impact of the confidence score on the accuracy and processing requirements of the pipeline for activity classification.

\section{Experimental Setup}\label{sec:activityrec_experiments}

\subsection{Databases}

For generic scenarios, this work used the Multi-Moments in Time (MMIT) database~\cite{Monfort2019}, made available by the creators upon request. The MMIT database includes a total of $1~035~862$ videos, split between a training set ($1~025~862$ videos) and a validation set ($10~000$ videos). These correspond to a total of 339 classes, describing the main activity verified in each video. From those classes, only those related to the target scenario of in-vehicle passenger monitoring were included. This resulted in a subset of twenty-one classes: fighting/attacking, punching, pushing, sitting, sleeping, coughing, singing, speaking, discussing, pulling, slapping, hugging, kissing, reading, telephoning, studying, socialising, resting, celebrating, laughing, and eating. Train and test divisions use the official predefined MMIT dataset splits.

\begin{figure}[!t]
\centering
\includegraphics[width=0.5\linewidth]{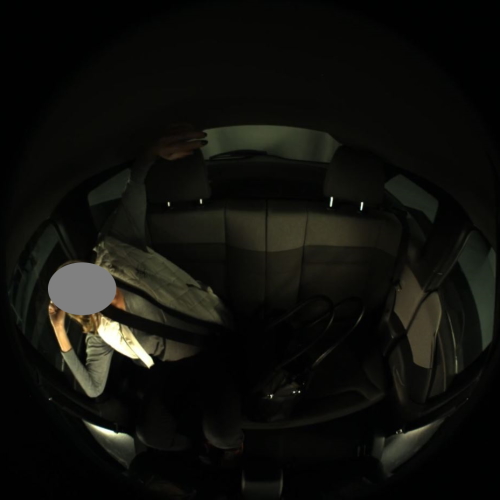}%
\includegraphics[width=0.5\linewidth]{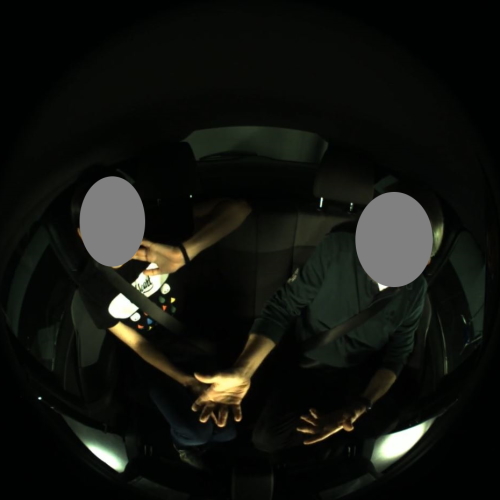}\\
\includegraphics[width=0.5\linewidth]{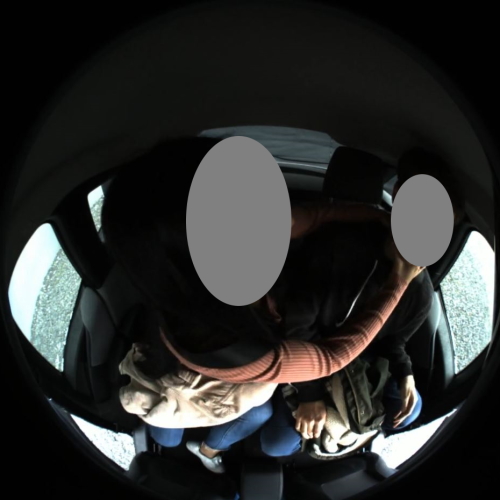}%
\includegraphics[width=0.5\linewidth]{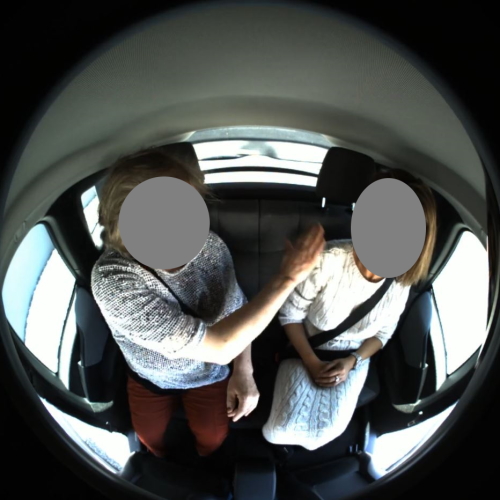}
\caption[Example frames from the in-vehicle dataset, depicting normal activities (top row) and violence between passengers (bottom row).]{Example frames from the in-vehicle dataset, depicting normal activities (top row) and violence between passengers (bottom row) (grey areas were used to protect the subjects' identities).}
\label{fig:activityrec_examples_hanau}
\end{figure}

For the in-vehicle scenario, a private dataset was used. The dataset includes a total of $490$ videos of the back seat of a car occupied by one or two passengers (see example frames in Fig.~\ref{fig:activityrec_examples_hanau}). Videos are acquired using a fish-eye camera to capture most of the interior of the car and microphones to acquire sound data. Each video includes annotations for forty-two action classes: ``entering'', ``leaving'', ``buckle on/off'', ``turning head'', ``lay down'', ``sleeping'', ``stretching'', ``changing seats'', ``changing clothes'', ``reading'', ``use mobile phone'', ``making a call'', ``posing'', ``waving hand'', ``drinking'', ``eating'', ``singing'', ``pick up item'', ``come closer'', ``handshaking'', ``talking'', ``dancing'', ``finger-pointing'', ``leaning forward'', ``tickling'', ``hugging'', ``kissing'', ``elbowing'', ``provocate'', ``pushing'', ``protecting oneself'', ``stealing'', ``screaming'', ``pulling'', ``arguing'', ``grabbing'', ``touching (sexual harassment)'', ``slapping'', ``punching'', ``strangling'', ``fighting'', and ``threatening with weapon''. Videos are randomly drawn into the train dataset ($70\%$) or the test dataset ($30\%$).

\subsection{Data preprocessing}

A total of $10$ frames, evenly spaced, was extracted from each video in the MMIT selected data subset (each with about 5 sec). These frames were concatenated over a third dimension following their temporal order to serve as input to the visual submodule. Two seconds of audio were extracted from each video and normalised to $16$ kHz sampling frequency to serve as input to the audio submodule.

For the in-vehicle dataset, each video can have multiple labels (the passengers portray different actions over the course of each acquisition). As such, each video period labelled as one of the 42 classes, is divided into two-second long individual samples. From each of these, 8 frames are extracted, resized and cropped into $224\times224$ squares, and concatenated over a third dimension to be used by the visual submodule, and the corresponding audio is resampled to $16$ kHz to be used by the audio submodule.

In the specific scenario of in-vehicle violence recognition, the aforementioned forty-two classes of the in-vehicle dataset have been clustered into three classes: normal car usage (from `entering' to `pick up item', in order of appearance), normal interactions (from `come closer' to `kissing'), and violence (from `elbowing' to `threatening with weapon').

\subsection{Model training}

The inflated ResNet-50 model used on the visual submodule uses the official pretrained weights from the MMIT database. Given that it was pretrained on the same database used in the laboratory experiments, this work took full advantage of this by setting most of the parameters of the network as non-trainable. The only parameters that were trained are those of the fully-connected layers which correspond to the classification of the selected twenty-one categories. This layer was optimised for a maximum of $250$ epochs according to categorical cross-entropy loss, with batch size 32, using the Adam optimiser with an initial learning rate of $10^{-4}$. For regularisation, dropout with a probability of 0.5 is used before the fully-connected layer. For the in-vehicle scenario, the training process is identical to the one described above. However, since the nature of the in-vehicle video data is substantially different from MMIT, the pretrained weights are also trained (not frozen) alongside the fully-connected layer for classification.

For both the generic and in-vehicle scenarios, the audio submodule was trained for a maximum of $200$ epochs with batch size $64$ and early-stopping patience of $25$ epochs. The optimisation was performed using Adam with an initial learning rate of $10^{-4}$ and cross-entropy loss.

\section{Results}\label{sec:activityrec_results}

After training the methodologies previously described, including the proposed cascade strategy, an overview of the obtained accuracy results is presented in Table~\ref{tab:activityrec_accuracy_summary}. It is clear the overall better performance of the proposed cascade pipeline, particularly with the audio submodule as the first block. Subsections~\ref{subsec:GenericScenario} and~\ref{subsec:vehicleScenario} offer a deeper discussion of the results for different configurations of the cascade pipeline for the more generic scenarios and the specific case of in-vehicle monitoring respectively.

\begin{table}[!t]
\caption{Summary of the accuracy (\%) results obtained in the various experimental scenarios.}\label{tab:activityrec_accuracy_summary} \centering
\begin{tabular}{lcccccccc}
  \hline
  \multirow{2}{*}{\textbf{Scenario}} && \multicolumn{2}{c}{\textbf{Unimodal}} && \textbf{Multimodal} && \multicolumn{2}{c}{\textbf{Cascade}}\\
  && \textbf{\textit{Audio}} & \textbf{\textit{Visual}} && \textbf{\textit{Parallel}} && \textbf{\textit{Audio-First}} & \textbf{\textit{Video-First}} \\\hline
  Generic (21 classes)      && 41.65 & 52.96 && 55.12 && \textbf{55.30} & 55.12 \\
  In-Vehicle (42 classes)   && 44.42 & 35.96 && 43.26 && \textbf{46.05} & 43.47 \\
  In-Vehicle (3 classes)    && 64.10 & 61.87 && 66.61 && \textbf{68.88} & 66.77 \\
  \hline
\end{tabular}
\end{table}

\subsection{Generic scenarios}
\label{subsec:GenericScenario}

\begin{figure}[!t]
    \centering
    \includegraphics[width=0.5\linewidth, trim={0.5cm 0 0.5cm 0}, clip]{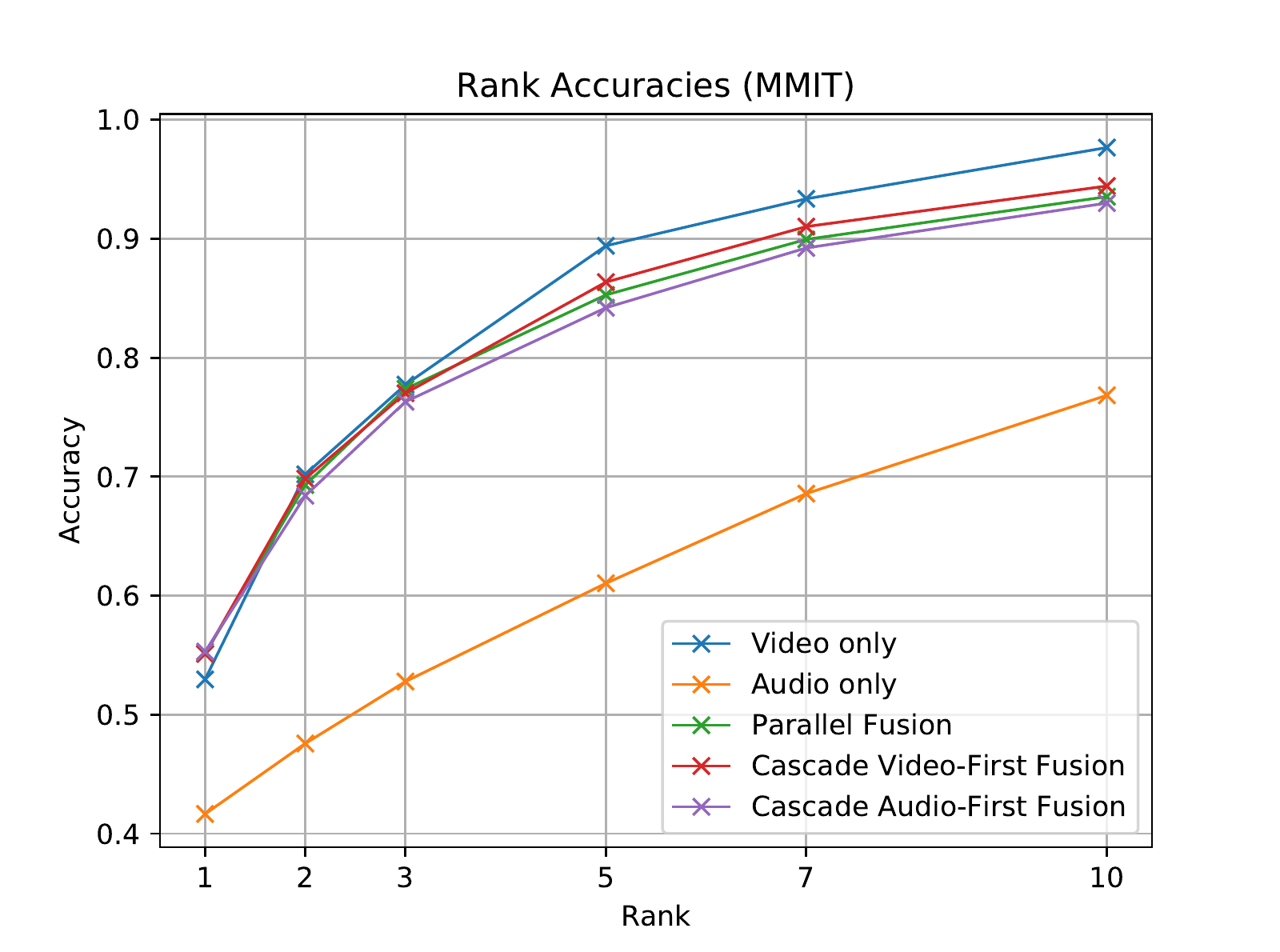}
    \caption{Rank accuracy results for the 21 selected classes from the MMIT database.}
    \label{fig:activityrec_mmit_rankacc}
\end{figure}

\begin{figure}[!t]
    \centering
    \includegraphics[width=0.5\linewidth, trim={0.5cm 0 0.5cm 0}, clip]{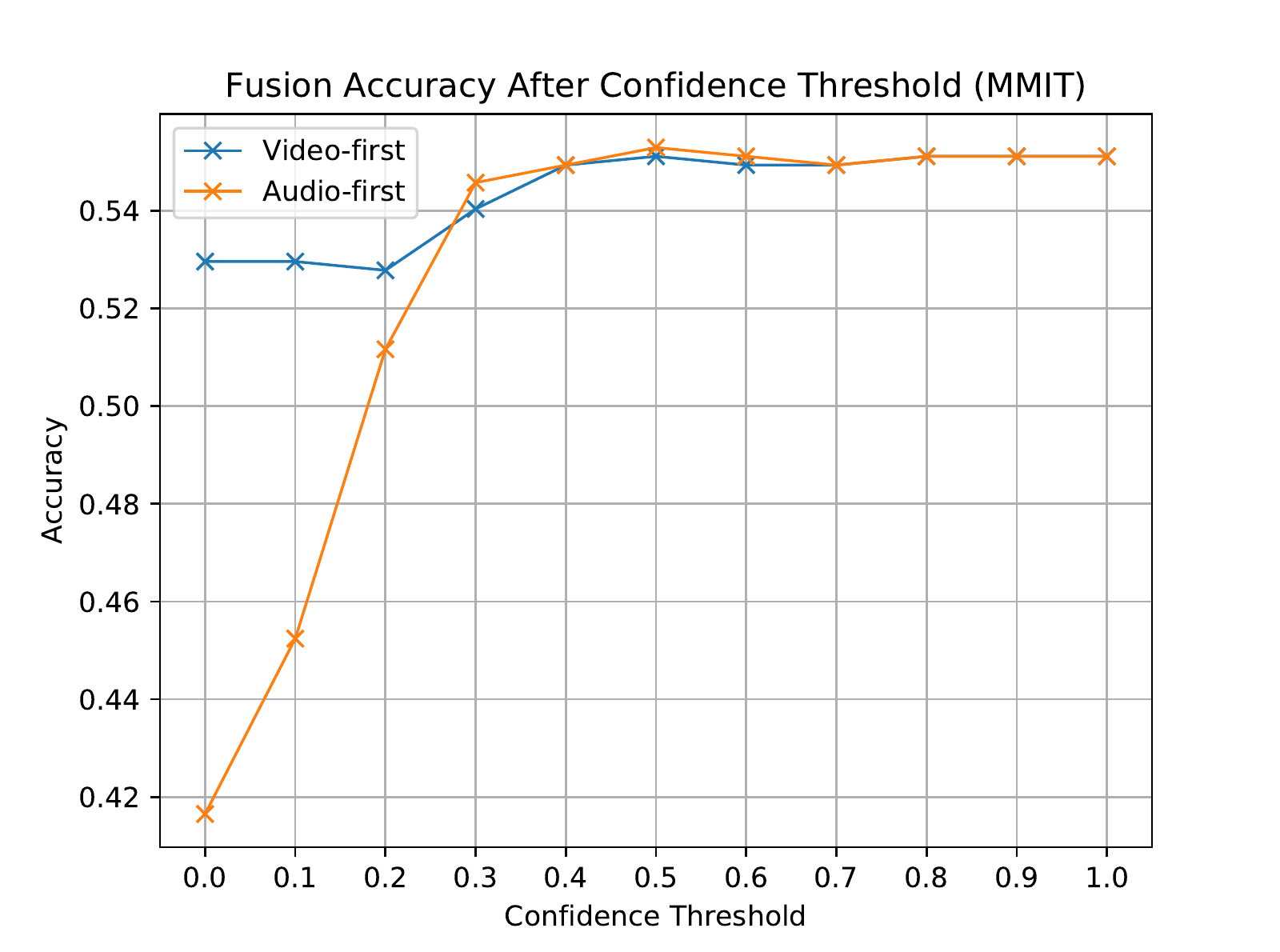}%
    \includegraphics[width=0.5\linewidth, trim={0.5cm 0 0.5cm 0}, clip]{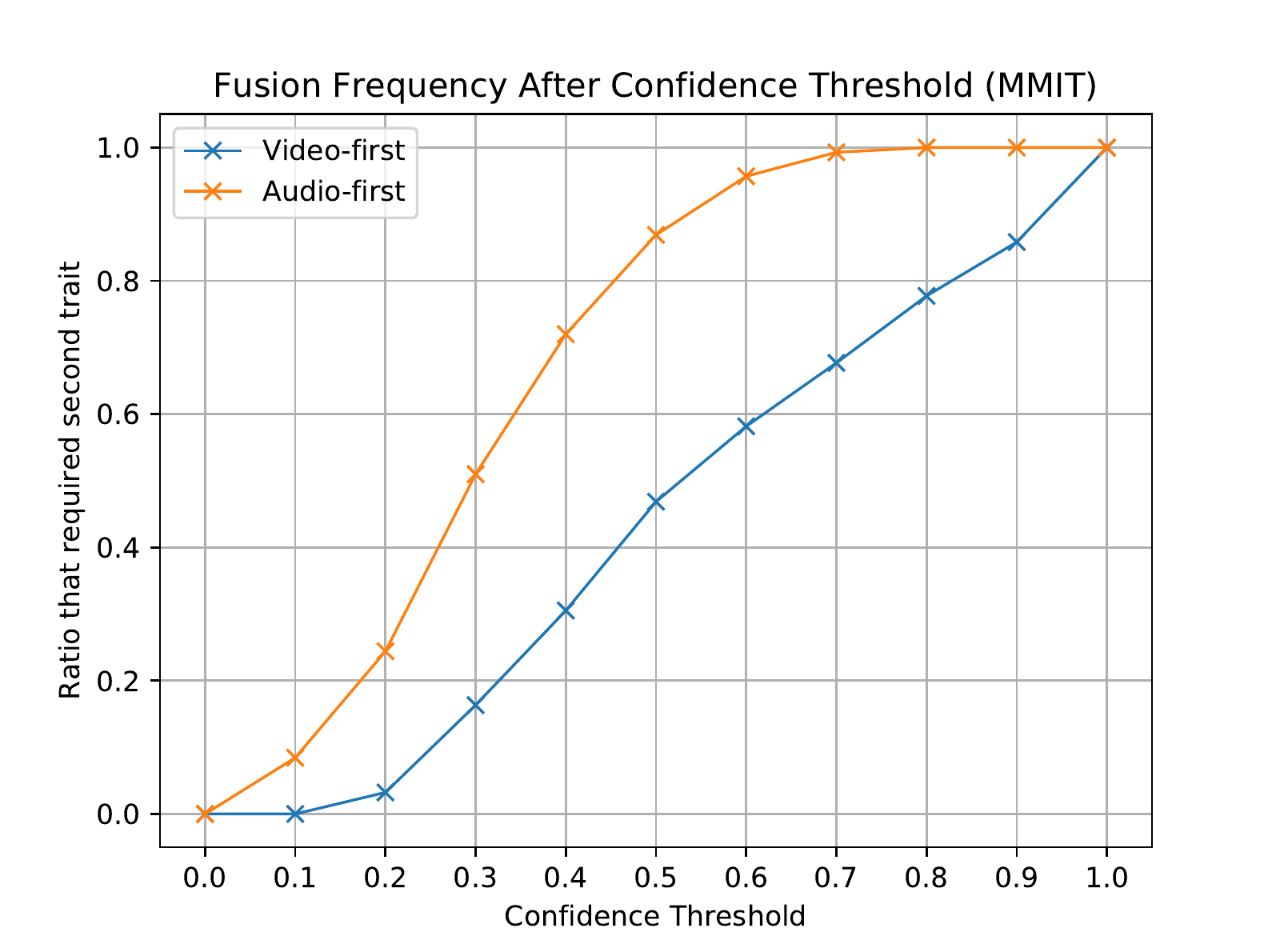}\\
    \caption[Cascade results for the 21 selected classes from the MMIT database.]{Cascade results for the 21 selected classes from the MMIT database (overall classification accuracy, on the left, and fraction of instances that need the secondary modality, on the right, for different confidence thresholds).}
    \label{fig:activityrec_mmit_cascade}
\end{figure}

For the laboratory experiments using the selected data from the MMIT database, the full parallel multimodal pipeline explored in this work offered $55.12\%$ accuracy. However, when considering the proposed cascade strategy based on confidence score thresholds, it was possible to achieve an improved accuracy score of $55.30\%$ (see Fig.~\ref{fig:activityrec_mmit_rankacc}). Beyond this relatively small accuracy improvement, the largest benefit of the proposed cascade algorithm is related to processing time. As presented in Fig.~\ref{fig:activityrec_mmit_cascade}, the best accuracy of $55.30\%$ is achieved with an audio-first cascade with a confidence score threshold $T=0.5$. This means it is possible to avoid the visual and fusion submodules for approximately $51\%$ of all instances without performance losses.

An analysis of size, number of parameters, and average run time per instance for each submodule (see Table~\ref{tab:activityrec_performance_summary}) shows that the visual model is the heftiest among the three submodules. Hence, being able to bypass it on more than half of the instances translates into significant time savings: while the full multimodal pipeline takes, on average, $85.9$ ms to predict an instance's activity label, the proposed cascade can do it in only $46.3$ ms, on average, without accuracy losses. This brings us closer to real applications using inexpensive hardware in the target in-vehicle scenario.

As visible in Table~\ref{tab:activityrec_performance_summary}, the proposed audio submodule has a total size of $1.70$ MB, approximately $230$ thousand parameters, and an average GPU run time of $8.30$ ms per instance. Conversely, the audio submodule used in \cite{Pinto2020Audiovisual}, on this task of activity recognition, has a total size of $3.94$ MB, approximately $1.026$ million parameters, and an average GPU run time of $1666$ ms per instance (due to the CPU-based handcrafted feature extraction process). Despite the significant reduction in run time, size, and complexity, the proposed audio submodule performed similarly \emph{vs}. the alternative ($41.65\%$ and $42.01\%$ accuracy, respectively).

\begin{table}[!t]
\caption[Summary of the size, total number of parameters, and average run times per instance of the three pipeline submodules for the in-lab scenario.]{Summary of the size, total number of parameters, and average run times per instance of the three pipeline submodules for the in-lab scenario (run times were computed using a NVidia GeForce GTX 1080 GPU, with the exception of the fusion submodule, computed on an Intel i7-8565U CPU).}\label{tab:activityrec_performance_summary} \centering
\setlength{\tabcolsep}{5pt} 
\begin{tabular}{lccc}
  \hline
  \textbf{Submodule} & \textbf{Size (MB)} & \textbf{Params.} & \textbf{Run Time (ms)}\\
  \hline
  Visual     & 176   & 46.2 M    & 77.18 \\
  Audio     & 1.70  & 220 K     & 8.30 \\
  Fusion    & 1.94  & -         & 0.38 \\
  \hline
\end{tabular}
\end{table}

\subsection{In-vehicle scenario}
\label{subsec:vehicleScenario}

\begin{figure}[!t]
    \centering
    \includegraphics[width=0.5\linewidth, trim={0.5cm 0 0.5cm 0}, clip]{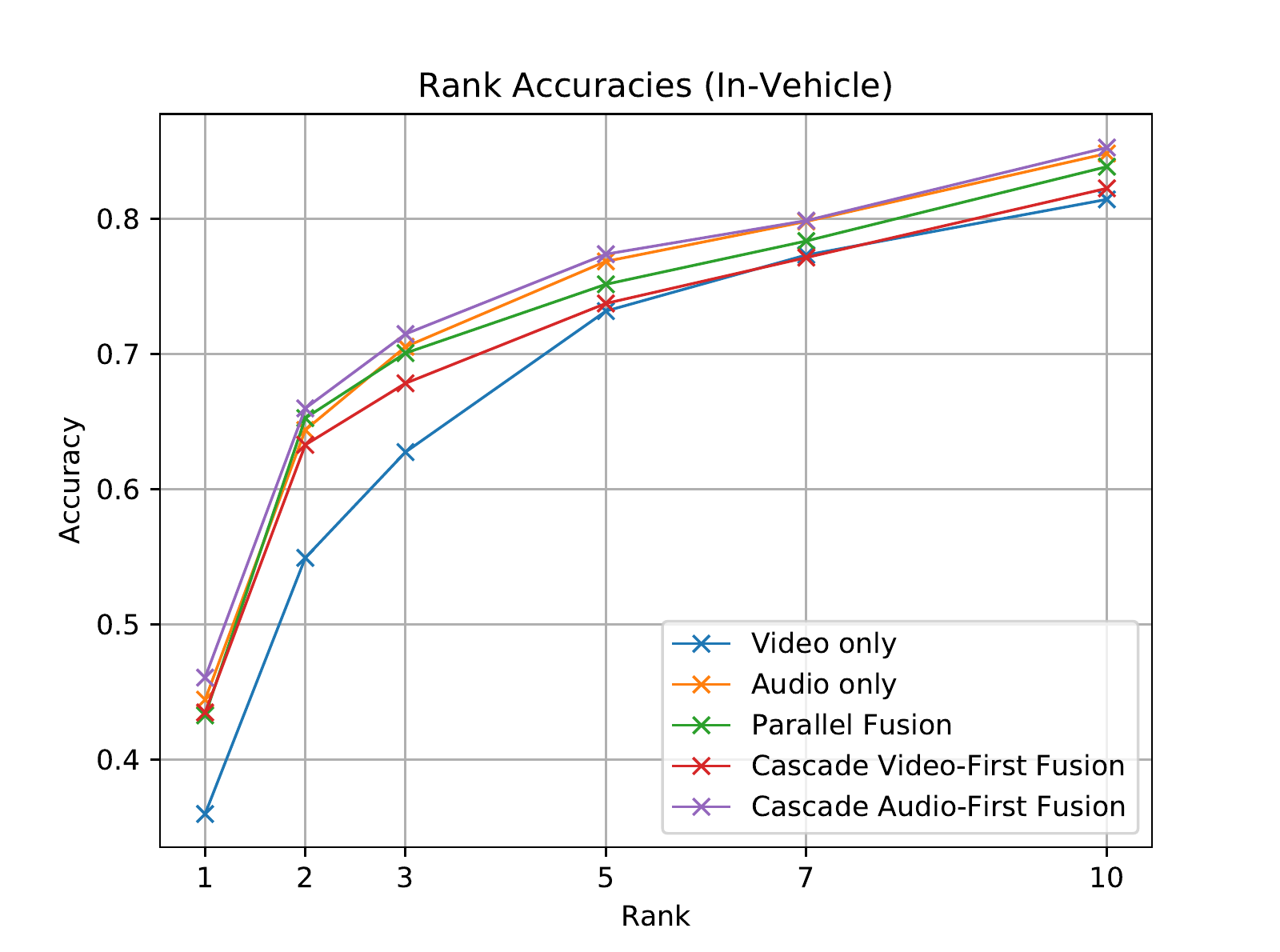}
    \caption{Rank accuracy results for the in-vehicle scenario with 42 classes.}
    \label{fig:activityrec_iv42_rankacc}
\end{figure}

\begin{figure}[!t]
    \centering
    \includegraphics[width=0.5\linewidth, trim={0.5cm 0 0.5cm 0}, clip]{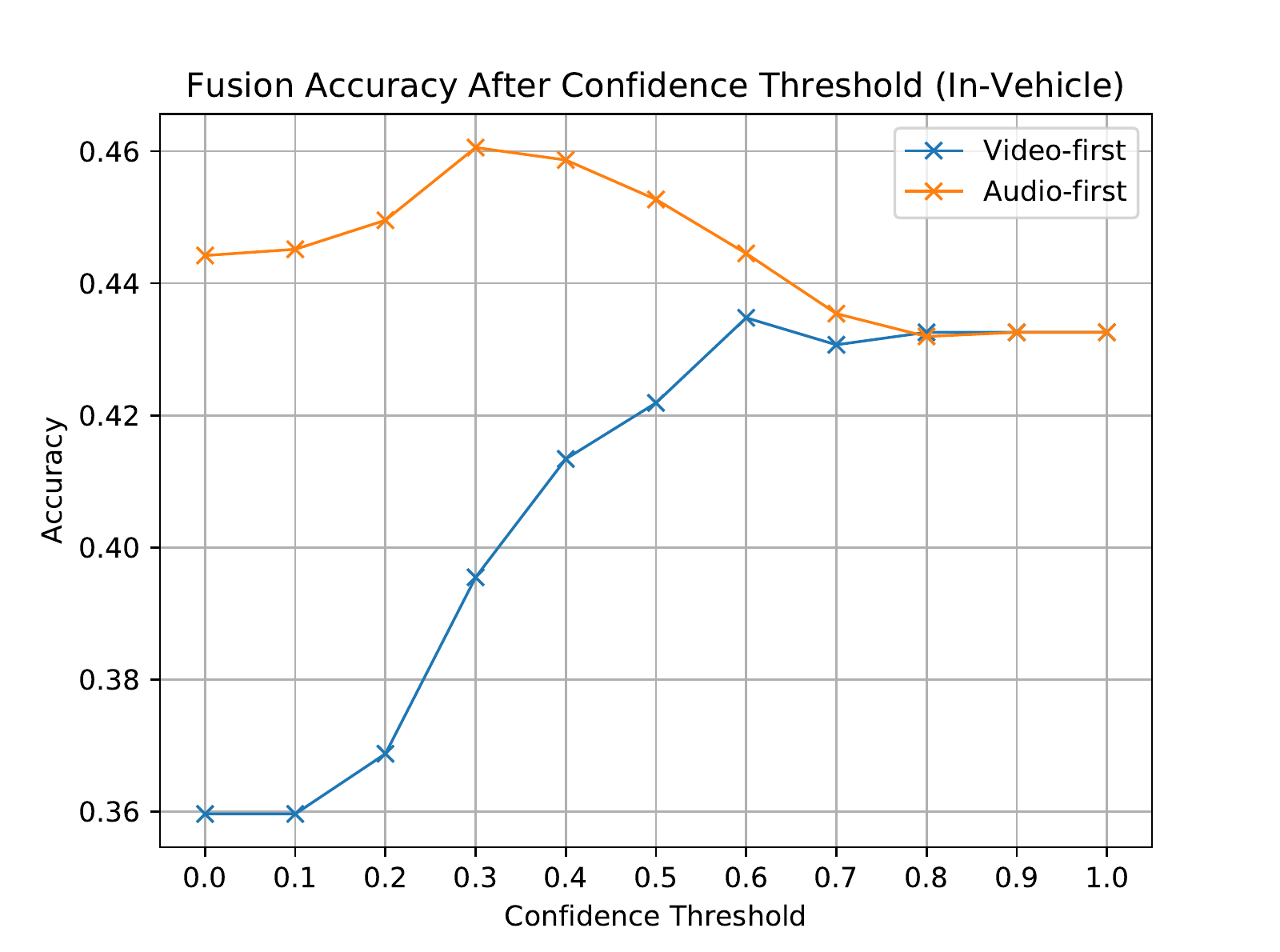}%
    \includegraphics[width=0.5\linewidth, trim={0.5cm 0 0.5cm 0}, clip]{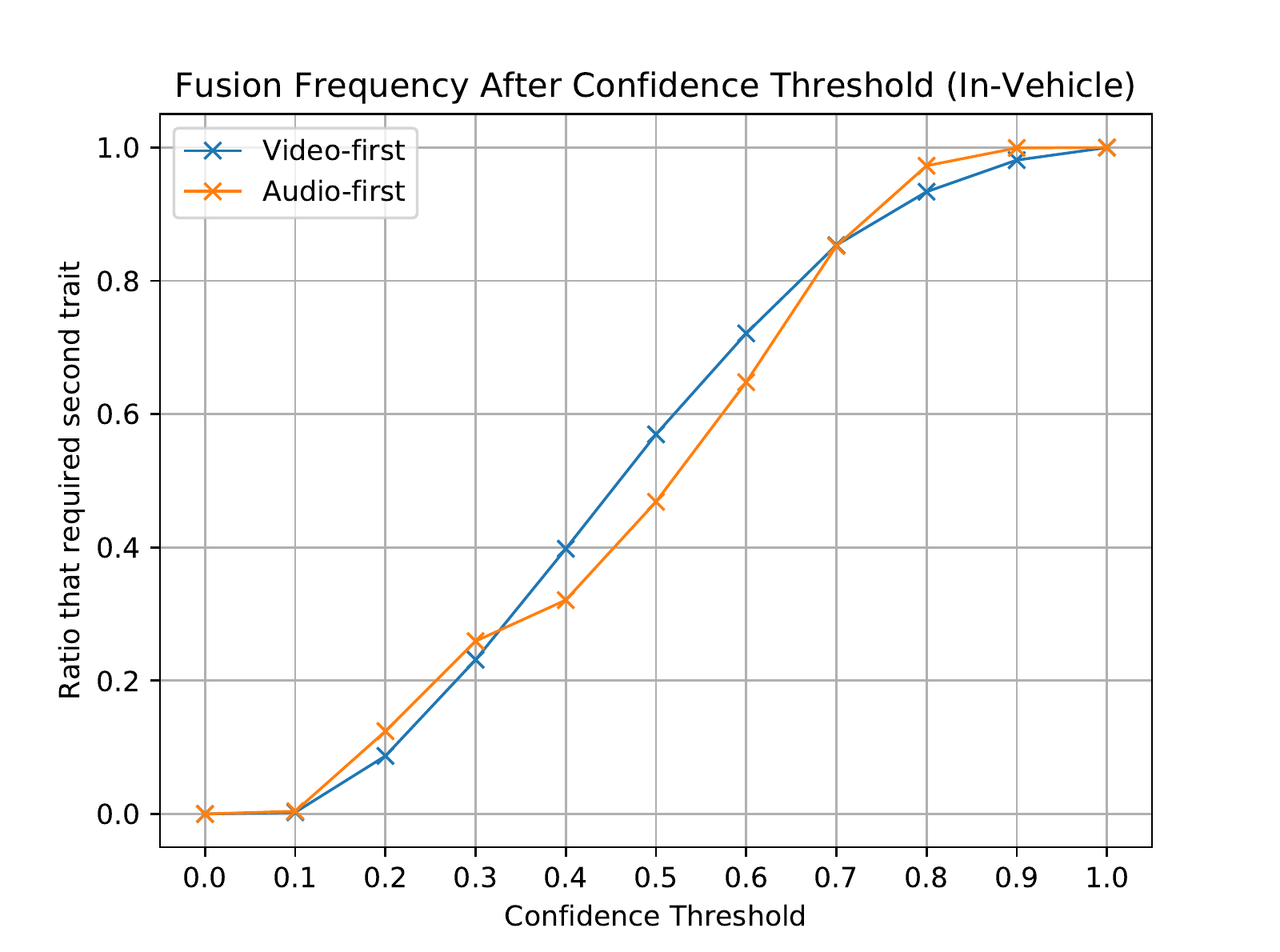}\\
    \caption[Cascade results in the in-vehicle scenario with 42 classes.]{Cascade results in the in-vehicle scenario with 42 classes (overall classification accuracy, on the left, and fraction of instances that need the secondary modality, on the right, for different confidence thresholds).}
    \label{fig:activityrec_iv42_cascade}
\end{figure}

The results on the data from the target in-vehicle scenario largely follow those discussed above for the laboratory experiments. On the 42-class activity recognition task, an audio-first cascade strategy achieved the best performance ($46.05\%$ accuracy) versus the full multimodal pipeline ($43.26\%$) and the best unimodal submodule ($44.42\%$). Similar accuracy improvements are verified up to rank 10 (see Fig.~\ref{fig:activityrec_iv42_rankacc}). With a confidence score threshold of $T=0.3$, this cascade strategy is able to avoid the visual submodule for approximately $74.1\%$ of the instances (see Fig.~\ref{fig:activityrec_iv42_cascade}). Considering the average run times presented in the previous subsection, this means the cascade is able to offer activity predictions in $28.4$ ms, on average, while offering considerably higher accuracy than the full multimodal pipeline (which would take $85.9$ ms).

\begin{figure}
    \centering
    \includegraphics[width=0.5\linewidth, trim={0.5cm 0 0.5cm 0}, clip]{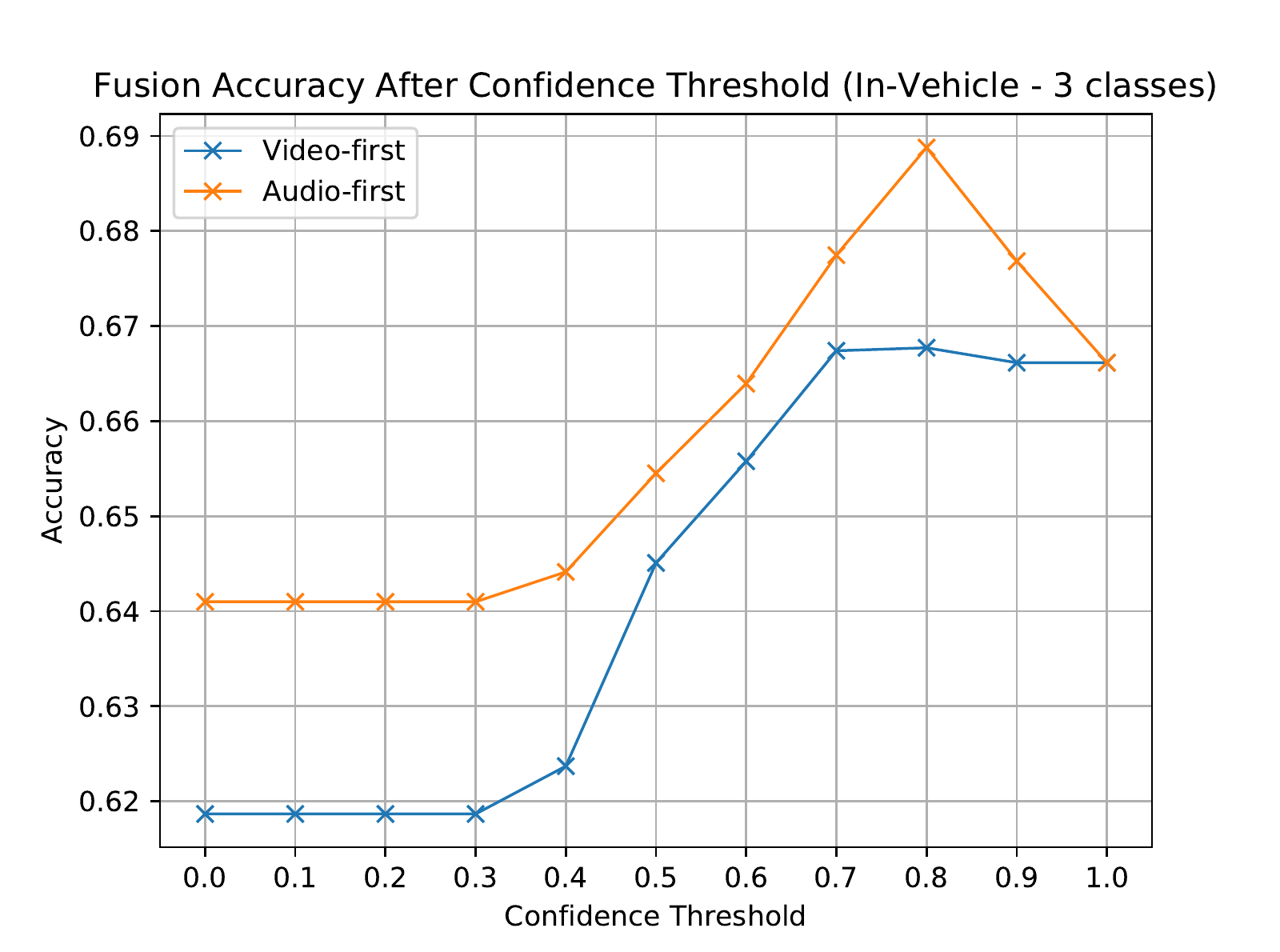}%
    \includegraphics[width=0.5\linewidth, trim={0.5cm 0 0.5cm 0}, clip]{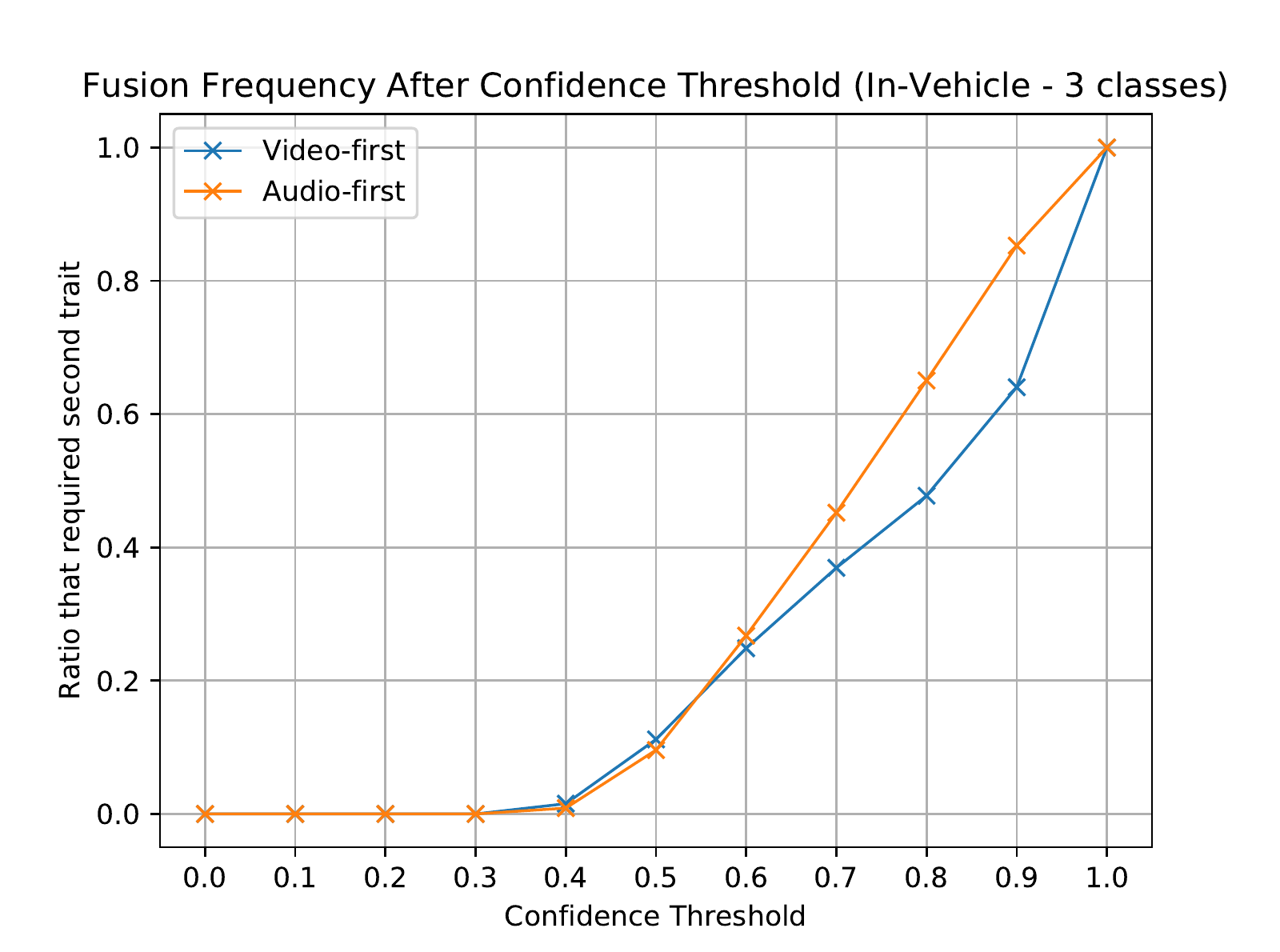}\\
    \caption[Cascade results in the in-vehicle scenario with 3 classes.]{Cascade results in the in-vehicle scenario with 3 classes (overall classification accuracy, on the left, and fraction of instances that need the secondary modality, on the right, for different confidence thresholds).}
    \label{fig:activityrec_iv3_cascade}
\end{figure}

For the three-class violence recognition task, the results follow the same trend, albeit with higher accuracy scores for all submodules and fusion strategies. The proposed cascade strategy with audio as the primary modality was able to attain $68.88\%$ accuracy, considerably better than the $66.61\%$ offered by the full multimodal pipeline. This accuracy corresponds to $T=0.8$, which enabled avoiding the visual submodule for $35\%$ of all instances (see Fig.~\ref{fig:activityrec_iv3_cascade}). While this value is lower than those reported for the previous experiments, it still translates into average time savings of $27.2$ ms per instance ($58.7$ ms for the cascade \emph{vs}. $85.9$ ms for the full pipeline), accompanied by a considerable improvement in accuracy.

\section{Summary and Conclusions}\label{sec:activityrec_conclusion}

This work explored a different strategy for the recognition of human activities, focusing on the scenario of autonomous shared vehicles. In addition to the inherent difficulties of automatically recognising human actions using audio-visual data, this specific scenario poses specific constraints regarding available hardware and energy consumption.

Inspired by state-of-the-art multimodal approaches, the main contributions are two-fold: a lighter-weight deep-learning based audio processing submodule; and a cascade processing pipeline. The proposed audio processing module demonstrated state-of-the-art performance while presenting lesser memory requirements and computational demands. With the submodules implemented, different configurations were tested for the cascade strategy to assess which one provides the best performance, taking into account two critical axes: accuracy and computational performance. Results show that by using audio as the first processing block, it was possible to obtain an accuracy score higher than the state-of-the-art, along with a significant reduction in processing/inference time.

The obtained results are interesting and reveal a high potential for further improvement. Modifications to the individual processing submodules could contribute to even higher accuracies while further reducing computational weight. The latter may benefit from a combination with model compression and acceleration techniques, such as quantisation, avoiding likely losses in accuracy due to compression. 

The proposed strategy demonstrated benefits from cascading the processing modules. Other early modules could bring other benefits by filtering out incoming audio-visual data, without relevant content (\emph{e.~g.}, without people present or without movement/sound).

\part{Broader Topics on Biometrics and Pattern Recognition}\label{part:broaderTopics}
\chapter[Learning Template Security on End-to-End Biometric Models]{Learning Template Security \\on End-to-End Biometric Models}\label{ch:secure}

\begin{tcolorbox}\footnotesize
{\large\bf Foreword on Author Contributions}

The research work described in this chapter was conducted entirely by the author of this thesis, under the supervision of Jaime S. Cardoso and Miguel V. Correia. The results of this work have been disseminated in the form of a journal article, an article in international conference proceedings, and an abstract in national conference proceedings:
\begin{itemize}[noitemsep, leftmargin=1em, nosep]
    \item \underline{J. R. Pinto}, M. V. Correia, and J. S. Cardoso, ``Secure Triplet Loss: Achieving Cancelability and Non-Linkability in End-to-End Deep Biometrics,'' \emph{IEEE Transactions on Biometrics, Behavior, and Identity Science}, 3 (2): 180--189, 2021.~\cite{Pinto2021Secure}
    \item \underline{J. R. Pinto}, J. S. Cardoso, and M. V. Correia, ``Secure Triplet Loss for End-to-End Deep Biometrics,'' in \emph{8th International Workshop on Biometrics and Forensics (IWBF 2020)}, Apr.~2020.~\cite{Pinto2020iwbf}
    \item \underline{J. R. Pinto}, M. V. Correia, and J. S. Cardoso, ``Achieving Cancellability in End-to-End Deep Biometrics with the Secure Triplet Loss,'' in \emph{26th Portuguese Conference on Pattern Recognition (RECPAD 2020)}, Oct.~2020.
\end{itemize}
This work was awarded the \emph{Computers Journal Best Paper Award} at the \emph{2020 International Workshop on Biometrics and Forensics} and the \emph{Best PhD Student Live Presentation}, awarded by the jury, at the \emph{2021 NIS Workshop}.

\end{tcolorbox}

\section{Context and Motivation}

It is easy to change our keys or passwords when a traditional authentication system is compromised, but it is very hard to change our compromised biometric characteristics. This is the reason why it is paramount that biometric templates are kept secure~\cite{Ignatenko2010, Nandakumar2015}. This is not easily achievable since, unlike password-based systems, biometric comparison is not binary and must also account for the natural intrasubject biometric variability~\cite{Jain2008, Nandakumar2015}.  

While several methods have been proposed to protect biometric templates, most require specific feature extraction or additional processes based on salting, biohashing, or cryptographic protection~\cite{Jain2008, Nandakumar2015}. Even those proposed for deep learning biometric methods~\cite{Pandey2016, Talreja2017} are integrated into end-to-end models, thus creating hurdles that often limit the achievable performance.

Hence, this chapter presents the Secure Triplet Loss, a reformulation of the well-known triplet loss that enables training end-to-end deep learning models to obtain secure biometric templates. Binding keys with the input within the model and considering key divergences in the objective function enables learning template cancelability. A component based on Kullback-Leibler divergence or distance statistics measures and actively promotes unlinkability. Thus, the proposed method aims to allow taking full advantage of the capabilities of end-to-end deep networks while still ensuring the security of the stored biometric data.

A detailed presentation of the Secure Triplet Loss can be found throughout the next sections. This training methodology was thoroughly evaluated for the task of identity verification, considering the template security properties of cancelability, unlinkability, non-invertibility, and secrecy leakage. It was thoroughly evaluated for both ECG and face, to confirm its solidity and flexibility, using the off-the-person University of Toronto ECG database (UofTDB)~\cite{Wahabi2014} and the unconstrained YouTube Faces~\cite{Wolf2011} database.

Moreover, it was studied in two scenarios: (a) training a model ``from scratch'' (initialised with random parameters), or (b) adapting an existing end-to-end biometric model to make it secure (taking advantage of pretrained weights and fine-tuning with the proposed method). The Secure Triplet Loss was also compared with the original triplet loss and competitive state-of-the-art approaches based on Bloom Filters~\cite{GomezBarrero2016} and Homomorphic Encryption~\cite{Drozdowski2019}. The code used in this work is available online\footnote{Secure Triplet Loss Github repository. Available on: \url{https://github.com/jtrpinto/SecureTL}.}.

\section{Related Work}

Beyond accounting for natural biometric characteristic variability, biometric data protection methods need to verify template cancelability, non-invertibility, and unlinkability. Cancelability (or revokability) means the templates can be easily and effectively rendered useless if they become compromised, generally through the change of a personal key that is bound with the template~\cite{punithavathi2017, tarek2016}.

Non-invertibility requires the transformation from biometric samples to templates to be as close to irreversible as possible. Thus, if the template is compromised, the original biometric sample cannot be reliably recovered or approximated~\cite{Nandakumar2015, Jain2008}. Finally, template unlinkability means it is difficult to assess if compromised templates from different biometric systems belong to the same identity~\cite{GomezBarrero2016}. 

One of the first template protection methods was the fuzzy commitment scheme proposed by \citet{Juels1999}, using cryptography and error-correcting codes for template cancelability. Later, \citet{Teoh2004} proposed BioHashing, an adaptation of the hashing process commonly applied to passwords to deal with fingerprint variability. A similar approach has been proposed by \citet{Sutcu2005}.

More recently, \citet{Rathgeb2013, Rathgeb2015} proposed the Bloom filter approach for alignment-free template cancelability and irreversibility. This approach was later adapted by \citet{GomezBarrero2016, gomezbarrero2018} to ensure template unlinkability, and by \citet{Drozdowski2018} for higher computational efficiency. \citet{Raja2019} proposed a highly efficient method using neighbourhood-preserving manifolds and hashing for biometric template protection in smartphones.

Among cryptography-based methods, homomorphic encryption (HE) approaches are particularly promising as HE allows arithmetic operations on the encrypted domain~\cite{Boddeti2018}. This allows the biometric comparison to be fully conducted on the encrypted domain, ensuring data security~\cite{Drozdowski2019}. Fully HE approaches, that allow for unlimited operations in the encrypted domain, most notably include Gentry's~\cite{Gentry2009}, Brakerski's~\cite{Brakerski2011}, and Fan-Vercauteren's~\cite{Fan2012} schemes.

HE has been successfully applied for biometric template protection in face~\cite{Boddeti2018, Drozdowski2019, Kolberg2020}, signature~\cite{GomezBarrero2016he}, and even multibiometric recognition~\cite{GomezBarrero2017}. However, with HE the protection of templates remains the responsibility of a separate process that should, ideally, be harmoniously integrated within the feature extraction algorithm.

Using deep learning, \citet{Pandey2016} proposed a template protection scheme that receives features from a convolutional neural network (CNN), quantises them, and applies hashing to obtain exact comparison despite the variability. Later, \citet{Talreja2017} used forward error control (FEC) decoding and hashing to protect biometric features extracted by deep neural networks. While these are applied to deep learning, they still require separate protection and comparison schemes. Hence, they are inadequate for recent state-of-the-art biometric recognition methods, which largely rely on end-to-end deep learning models for significantly improved performance.

Considering this, this work proposes the Secure Triplet Loss, a reformulation of the triplet loss~\cite{Chechik2010} that promotes cancelability and unlinkability in end-to-end biometric models. More importantly, it aims to achieve this while avoiding additional protection processes and decreases in performance relative to the original triplet loss.

\section{The Secure Triplet Loss}

\subsection{Original triplet loss}

The triplet loss~\cite{Chechik2010} has been widely used in deep learning to train networks to accurately determine whether or not two samples belong to the same class~\cite{Pinto2019b,Chen2017,Cheng2016}. During training, such networks receive three inputs (a triplet), in parallel: one is the anchor ($x_A$, the reference with identity $i_A$), the second is the positive sample ($x_P$, with identity $i_P = i_A$), and the third is the negative sample ($x_N$, with identity $i_N \neq i_A$). In biometrics, triplets are groups of three biometric samples (images or signals): the anchor and positive inputs correspond to the same individual, unlike the negative input.

For each input, the network will output a representation: \eg, for the anchor, $y_A = f(x_A)$. The three representations are then compared using a measure of distance or dissimilarity $d(y_1,y_2)$, and the network is optimised through the minimisation of the triplet loss function:
\begin{equation}
    l_{TL} = \max\left[ 0, \alpha + d(y_A, y_P) - d(y_A, y_N)\right ],
\end{equation}
which leads representations of the same class to be more similar than those of different classes, minimising $d(y_A, y_P)$ and maximising $d(y_A, y_N)$. The loss also aims to enforce a minimum margin $\alpha > 0$ between the two distances.

This is a generally successful strategy when training neural networks for biometric verification (assessing if the identities of a biometric template and a biometric query match). However, it does not address the important issue of security in biometrics, especially the topics of cancelability and unlinkability.

\begin{figure}
    \centering
    \includegraphics[width=0.7\linewidth]{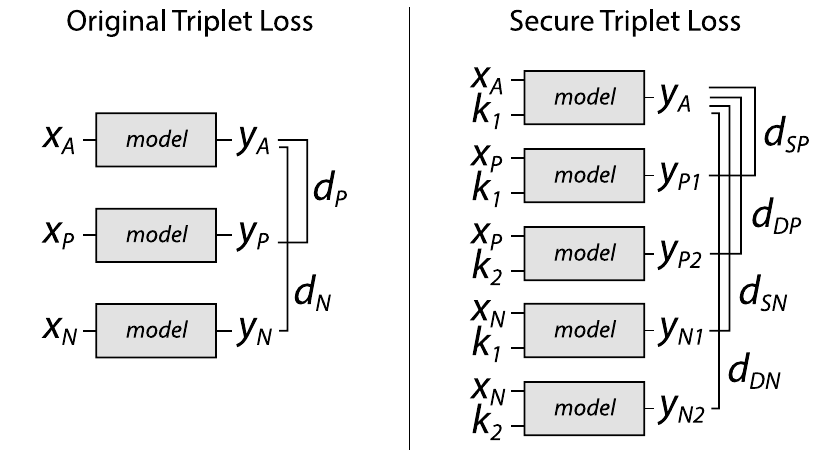}
    \caption[Comparison between the model training schemes of the original triplet loss and the proposed Secure Triplet Loss method.]{Comparison between the model training schemes of the original triplet loss and the proposed Secure Triplet Loss method~\cite{Pinto2020iwbf}.}
    \label{fig:templatesec_schema}
\end{figure}

\subsection{Learning cancelability}

In its initial formulation, proposed in~\cite{Pinto2020iwbf}, the Secure Triplet Loss modifies the original triplet loss to make the final sample representations cancelable (as illustrated in Fig.~\ref{fig:templatesec_schema}). Besides the triplet inputs ($x_A$, $x_P$, and $x_N$), the network also receives two different keys ($k_1$, $k_2$) that are bound with the inputs by the network itself.

Unlike the original triplet loss, $x_P$ and $x_N$ are processed by the network twice. First, they are combined with $k_1$ and then with $k_2$. The anchor $x_A$ is only bound with $k_1$. 
Thus, five representations are obtained: $y_{A} = f(x_A, k_1)$, $y_{P1} = f(x_P, k_1)$, $y_{P2} = f(x_P, k_2)$, $y_{N1} = f(x_N, k_1)$, $y_{N2} = f(x_N, k_2)$. From these, four distances are computed: $d_{SP} = d(y_{A}, y_{P1})$ (with matching identities and keys), $d_{DP} = d(y_{A}, y_{P2})$ (with matching identities but different keys), $d_{SN} = d(y_{A}, y_{N1})$ (with different identities but matching keys), and $d_{DN} = d(y_{A}, y_{N2})$ (with non-matching identities and keys).

\begin{figure}
    \centering
    \includegraphics[width=0.6\linewidth]{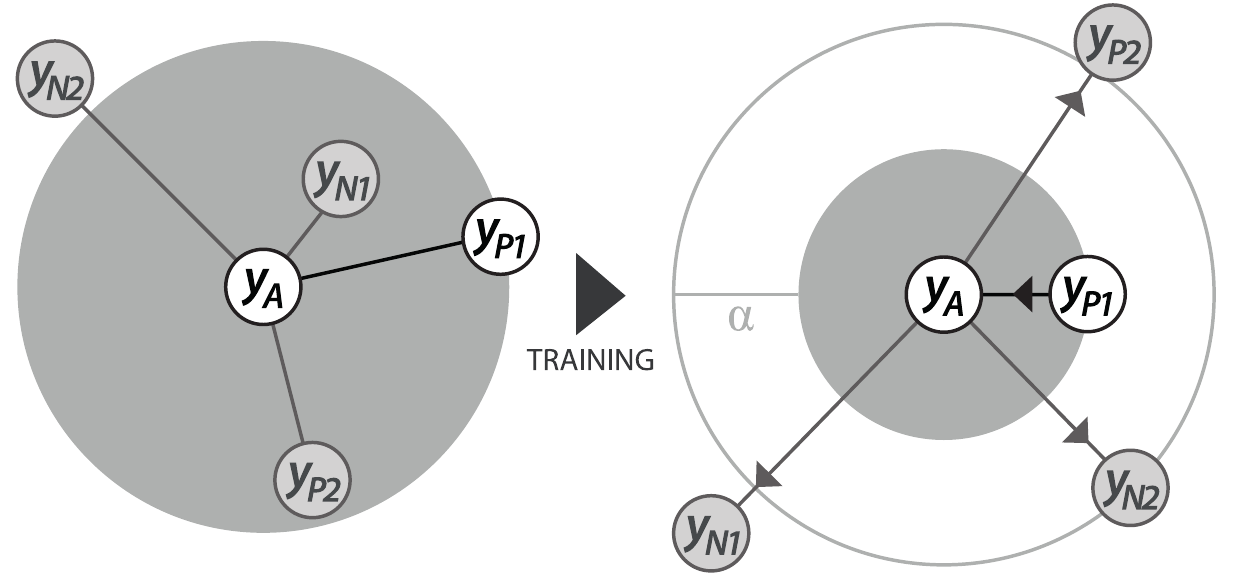}
    \caption[Illustration of the expected results when training with the proposed Secure Triplet Loss.]{Illustration of the expected results when training with the proposed Secure Triplet Loss (during training, the Secure Triplet Loss promotes the proximity between $y_A$ and $y_{P1}$, which match in identity and key, and larger distance to the three negative samples with a margin $\alpha$).}
    \label{fig:templatesec_secure_illustration}
\end{figure}

The objective is to minimise $d_{SP}$, when both the identities and the keys match, and maximise the remaining three distances (see Fig.~\ref{fig:templatesec_secure_illustration}). Hence, the loss is computed through:
\begin{equation}
    l_{STL} = \max \left( 0, \alpha + d_{SP} - d_{n} \right),
\end{equation}
where $d_n$ results from the combination of all three distances to be maximised. Here, we consider $d_n = \min(\{ d_{SN}, d_{DP}, d_{DN} \})$, with the three distances to be maximised being considered equally relevant. This results in:
\begin{equation}
    l_{STL} = \max \left[ 0, \alpha + d_{SP} - \min(\{ d_{SN}, d_{DP}, d_{DN} \}) \right].
\label{eq:templatesec_loss}
\end{equation}

As with triplet loss, $\alpha$ enforces a margin between positive and negative distances. In this case, the loss involves four distances, since it also takes into account whether or not the keys match. By minimising the loss in Eq.~\eqref{eq:templatesec_loss}, the network learns to deal with the intrasubject and intersubject variability of the biometric characteristic. More importantly, it learns to recognise when the keys do not match, even if the identity is the same. Hence, if the stored templates become compromised, they can easily be invalidated through a key change. However, as reported in~\cite{Pinto2020iwbf}, $l_{STL}$ fails to promote unlinkability.

\subsection{Promoting unlinkability}

Unlinkability can be achieved by combining the original formulation of the Secure Triplet Loss, $l_{STL}$, with a component that quantifies linkability in the representations output by the network during training, $l_L$. Thus, the proposed reformulation of the Secure Triplet Loss, as presented in~\cite{Pinto2021Secure}, follows the equation:
\begin{equation}
    l_{STL2} = \gamma l_{STL} + (1-\gamma)l_{L}.
\end{equation}

The $l_{STL}$ component is the original Secure Triplet Loss in Eq.~\eqref{eq:templatesec_loss}, focused on biometric performance and template cancelability, following the formulation in Eq.~\eqref{eq:templatesec_loss}. The parameter $\gamma\in[0,1]$ balances the $l_{STL}$ and $l_{L}$ loss components. The $l_{L}$ component is focused on measuring template linkability. To achieve unlinkability, one has to ensure similar distance values are obtained when keys don't match (regardless of whether or not the templates are from the same identity). Hence, $d_{DP}$ and $d_{DN}$ should assume similar values. This can be achieved using the Kullback-Leibler divergence (KLD), computed over each batch. This agrees with the reference linkability metric, which is also inspired by the KLD. In this case, this part of the loss becomes:
\begin{equation}
    l_{L} = D_{KL}(P_{d_{DP}}||P_{d_{DN}}) = \sum P_{d_{DP}}\log\left(\frac{P_{d_{DP}}}{P_{d_{DN}}}\right),
\label{eq:templatesec_loss_linkkld}
\end{equation}
where $P_{d_{DP}}$ and $P_{d_{DN}}$ are the probability density functions for the distributions of $d_{DP}$ and $d_{DN}$, respectively. To obtain these distributions and their respective probability density functions, this part of the loss cannot be computed over each triplet, instead being computed over each batch of triplets. For brevity, this formulation of the Secure Triplet Loss with Kullback-Leibler divergence-based linkability is from now on designated as \emph{SecureTL w/KLD}.

Alternatively, one can avoid estimating these distributions and the computation of the Kullback-Leibler divergence using simple statistics to promote linkability. If we consider $\mu(d_{DP})$ and $\sigma(d_{DP})$ as the mean and standard deviation, respectively, of the distances $d_{DP}$ on a given batch, and likewise $\mu(d_{DN})$ and $\sigma(d_{DN})$ for the distances $d_{DN}$ on the same batch, then we can reformulate
\begin{equation}
    l_{L} = |\mu(d_{DP}) - \mu(d_{DN})| + |\sigma(d_{DP}) - \sigma(d_{DN})|.
\end{equation}
This should lead the model to offer embeddings that result in similar distance scores when the keys do not match, regardless of whether or not the identities match, thus avoiding template linkability. Throughout the remainder of this chapter, for brevity, the formulation of the Secure Triplet Loss with this statistics-based linkability module is designated as \emph{SecureTL w/SL}.

\section{Experimental Setup}

The proposed methodology for learning secure biometric models was explored for the two biometric characteristics addressed within this doctoral work: the ECG and the face. This section presents the details of the models, the data, and the conducted experiments.

For either characteristic, keys have been randomly generated for each triplet, consisting of unidimensional arrays with $100$ binary values. Each key is processed after generation to verify unit $l2$ norm. For SecureTL w/KLD and SecureTL w/SL, the parameter $\gamma$ that controls the balance between the Secure Triplet Loss and the linkability component was set to $0.9$: this value has overall been able to offer good template unlinkability without considerably harming the validation performance and cancelability.

\begin{figure}[!t]
    \centering
    \includegraphics[width=0.8\linewidth]{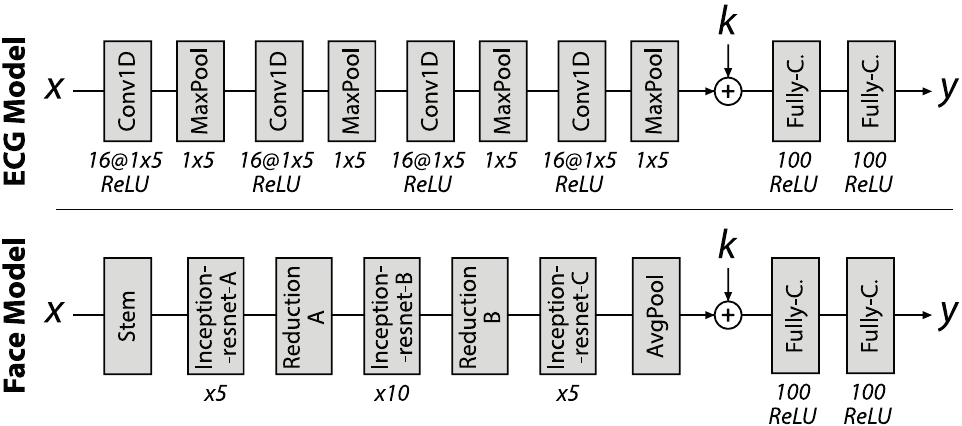}
    \caption[Architecture of the models used for ECG and face identity verification.]{Architecture of the models used for ECG and face identity verification ($x$ denotes an input biometric sample, $k$ a key, and $y$ a biometric template; the structure of the face model before concatenation with $k$ follows precisely the structure of the Inception-ResNet-v1, which is presented in higher detail in~\cite{Szegedy2016inception}).}
    \label{fig:templatesec_models}
\end{figure}

\subsection{ECG identity verification}

\subsubsection{Data}
The ECG data used comes from the University of Toronto ECG Database (UofTDB)~\cite{Wahabi2014}. This database includes recordings from $1019$ subjects over up to six sessions and five different positions. The signals are off-the-person (less obtrusive and more comfortable for realistic biometric applications) and have been acquired at $200$~Hz using dry electrodes on the pointer fingers. Each recording is generally $2$ to $5$ seconds long.

Data from the last $100$ identities were used for training, while the data from the remaining 919 subjects have been reserved for testing. From these $919$, one has been discarded for only having a total of $30$ seconds of data. Triplets have been generated by selecting an anchor from the first $30$~s of data from a subject and positive and negative samples from the remaining data of the same or another identity, respectively. From the $100$ training identities, $100~000$ triplets have been generated, with $20\%$ being used for validation. A total of $10~000$ triplets have been generated for testing. Each of the three samples in a triplet is a blindly-segmented five-second raw ECG sample, normalised to zero mean and unit variance. 

\subsubsection{Model}
The model for ECG identity verification (see Fig.~\ref{fig:templatesec_models}) is adapted from the end-to-end architectures proposed in~\cite{Pinto2019Deep, Pinto2019b} and described in Chapter~\ref{ch:ecgiden} and Chapter~\ref{ch:ecgauth}. The model is composed of four unidimensional convolutional layers (with $16$, $16$, $32$, and $32$ filters, respectively, with size $1\times5$, unit stride, and zero padding), each followed by ReLU activation and max-pooling (with $1\times3$ kernels and stride $3$). The model ends with two fully-connected layers, each with $100$ units and followed by ReLU activation.

Once trained, this model receives a $5$ second long raw ECG segment ($1000$ samples long at $200$ Hz sampling frequency) and outputs an embedding or template that can be compared to a reference through the Euclidean distance (during training) or through the normalised Euclidean distance~\cite{Wolfram} (with the trained model, to obtain dissimilarity scores in the $[0,1]$ range). In the case of Secure Triplet Loss, the feature vector $s(x)$ (the flattened feature maps from the last max-pooling layer) is concatenated with the key array $k$, and both are bound together by the fully-connected layers to make the final secure template $f(x,k)$.

The model was trained using the Adam~\cite{Kingma2015} optimiser, with an initial learning rate of $0.0001$ and $l2$ weight regularisation with $\lambda=0.001$. The training lasted a maximum of $250$ epochs, with batch size $32$, with early stopping based on validation loss with patience of $25$ epochs.

\subsection{Face identity verification}

\subsubsection{Data}
To fine-tune and evaluate the model, images from the YouTube Faces database~\cite{Wolf2011} were used. This database is composed of frames from $3425$ YouTube videos, depicting a total of $1595$ subjects (up to six videos of each subject). Each video corresponds to between $48$ and $6070$ frames. This work used the aligned images provided on the database, which resulted from face detection, cropping, and alignment. 

Each face image has been reduced to $70\%$ height and width and resized to $160\times160$ to match the input dimensions of the model. Ten random triplets have been generated for each of the first $500$ subjects on the database for a total of $5000$ training triplets, of which $1000$ have been used for validation. Ten random triplets have also been generated from each of the remaining identities, reserved for testing, resulting in a total of $10~950$ test triplets. Whenever possible, the anchor and positive samples corresponded to different videos of the same identity.

\subsubsection{Model}
The model for face identity verification (see Fig.~\ref{fig:templatesec_models}) is based on the Inception-ResNet~\cite{Szegedy2016inception}. This network has been pretrained\footnote{FaceNet Pytorch Package. Available on: \url{https://github.com/timesler/facenet-pytorch}.} for identification on the VGGFace2 dataset~\cite{Cao2018} and offered an accuracy of $99.63\%$ on the Labelled Faces in the Wild (LFW) dataset and $95.12\%$ on the YouTube Faces database~\cite{Schroff2015}. The original fully-connected layer has been replaced with two new fully-connected layers, each with $100$ units and followed by ReLU activation. For the Secure Triplet Loss, the first of these layers receives the feature vector $s(x)$ from the first part of the model, concatenated with the key $k$. The second outputs the template $y(x,k)$.

All layers on the model have been frozen, to take advantage of the pretrained parameters. The exceptions are the last convolutional block and the fully-connected layers that come, respectively, before and after the average pooling operation. The last convolutional block is fine-tuned to allow for small adjustments during training, while the fully-connected layers are newly created and thus require training. The model was trained for a maximum of $250$ epochs at batch size $32$, with early stopping based on validation $EER$ with a patience of $25$ epochs. As with the ECG model, the Adam optimiser was used with an initial learning rate of$0.0001$ and $l2$ regularisation with $\lambda=0.001$.

\subsection{Evaluation frameworks and metrics}

The experiments have been designed to quantify the performance of the models trained with the original and Secure Triplet Loss formulations, not only considering verification accuracy but also biometric security.

\subsubsection{Verification performance}
The verification performance is quantified through the measurement of false match rates ($FMR$) and false non-match rates ($FNMR$) over the range of possible decision thresholds (for these models, $t\in[0,1]$). These values are presented in $FMR$ \emph{vs.} $FNMR$ plots and detection error trade-off~(DET) curves and used to compute the equal error rate ($EER$), corresponding to the error where $FMR_V=FNMR$, and the $FNMR@FMR=0.01\%$.

\subsubsection{Cancelability}
Avoiding additional processes such as biohashing or template encryption, the proposed Secure Triplet Loss integrates cancelability into the single output of the system, the template $y(x,k)$, and is reflected in the distance measure $d$ between two templates. Although the proposed loss is designed to promote cancelability, this property may not necessarily be achieved.

Hence, the experiments with the Secure Triplet Loss include the measurement of cancelability error. The plots of false match \emph{vs.} false non-match rates over the dissimilarity/distance scores include both the false match rate based on identity (when identities don't match, denoted as $FMR_V$) and the false match rate based on cancelability (when keys don't match, denoted as $FMR_C$). The false non-match rate ($FNMR$) values are the same for identity and cancelability since they refer to situations when both identity and keys match. The value of cancelability false accept rate at the operation point that corresponds to the verification $EER$, $FMR_C@EER$, is also computed.

\subsubsection{Unlinkability}
The template unlinkability analysis followed the method described by Gomez-Barrero~\etal~\cite{GomezBarrero2016}. The test samples were paired into mated (different biometric samples from the same identity with different keys) and non-mated instances (different identities and keys). These have been used to compute $p(d|H_m)$ and $p(d|H_{nm})$: the probability density functions of the distance score $d$ given the instances are, respectively, mated (hypothesis $H_m$) or non-mated (hypothesis $H_{nm}$). From the likelihood ratio $LR(d) = p(d|H_{m})/p(d|H_{nm})$, $D_{\leftrightarrow}(d)$ is computed through
\begin{equation}\small
D_{\leftrightarrow}(d) = \begin{cases}
    0,  & \text{if } LR(d) \leq 1 \\
    2 \left( \left(1 + \me^{-\left(LR(d) - 1\right)}\right)^{-1}  -\frac{1}{2} \right), & \text{if } LR(d) > 1
  \end{cases}
\label{eq:templatesec_d_s}
\end{equation}
which allows to compute the $D_{\leftrightarrow}^{sys}$ linkability metric with
\begin{equation}\small
D_{\leftrightarrow}^{sys} = \int_{d_{min}}^{d_{max}} D_{\leftrightarrow}(d) \cdot p(d|H{m}) \diff d .
\label{eq:templatesec_d_sys}
\end{equation}

The $D_{\leftrightarrow}^{sys}$ is considered the main metric to quantify template linkability. A biometric system verifying perfect template unlinkability, which is highly desirable, will assume $D_{\leftrightarrow}^{sys} = 0$. A biometric system creating entirely linkable templates will verify $D_{\leftrightarrow}^{sys} = 1$.

\subsubsection{Non-invertibility and secrecy leakage}

Other aspects of template security offered by the proposed method were evaluated, namely non-invertibility and secrecy leakage. Non-invertibility is measured through the privacy leakage rate, which can be computed through the expression: 
\begin{equation}
    \frac{H(X|Y)}{H(X)} = 1 - \frac{I(X;Y)}{H(X)},
\end{equation}
where $X$ is the input biometric,  $Y$ is the output of the model, $H(X)$ denotes the entropy of $X$, $H(X|Y)$ denotes the conditional entropy of $X$ given $Y$, and $I(X;Y)$ denotes the mutual information between $X$ and $Y$. The privacy leakage rate, in the range $[0,1]$, should be as close to $1$ as possible: obtaining information on $X$ should be impossible even when one has all knowledge of $Y$. The secrecy leakage measures the mutual information between the stored template $Y$ and the key $K$, through the expression $I(Y;K)$. The keys are public, unlike the templates, so they should reveal as little information as possible on the templates. Hence, the secrecy leakage should be close to zero. 

These require the computation of some information theoretical measures, such as entropy and mutual information. This is very difficult in biometrics, due to the high dimensionality of the inputs and the feature sets, as well as their variability. In this work, entropy and mutual information were estimated using a Python implementation\footnote{Paul Brodersen's Entropy Estimators. Available on: \url{https://github.com/paulbrodersen/entropy_estimators}.} of the methods proposed in~\cite{Kozachenko1987} and in~\cite{Kraskov2004}, respectively, for continuous multivariate data. These methods, based on nearest neighbour statistics, were shown to be more accurate than the alternatives~\cite{doquire2012}. Since the processing cost of such estimations grows exponentially with the size of the dataset, a subset of $1000$ test anchors has been used for this test.

\section{Results and Discussion}

A general overview of the results obtained is presented in Table~\ref{tab:templatesec_ecg_results_summary} and Table~\ref{tab:templatesec_face_results_summary}, respectively for ECG and face identity verification. The following subsections discuss the results on verification performance, cancelability, and unlinkability, and the comparison with state-of-the-art alternatives.

\begin{table}[!t]
\centering
\caption{Summary of the test results for ECG identity verification.}\label{tab:templatesec_ecg_results_summary}
\begin{tabular}{lcccc}\hline
 & \multicolumn{2}{c}{\textbf{Performance}} & \textbf{Cancelability} & \textbf{Linkability}\\
\textbf{Method} & \textbf{$EER$ ($\%$)} & \textbf{$FNMR~@FMR_V=0.1\%$} & \textbf{$FMR_C@~EER$} & \textbf{$D_{\leftrightarrow}^{sys}$} \\\hline
Triplet Loss    & 12.56 & 0.9033 & -    & - \\
BF~\cite{GomezBarrero2016} & 15.76 & 0.9242 & 0.0075 & 0.234 \\
HE~\cite{Drozdowski2019}   & 12.49 & 0.9573 & 0.0806 & 0.002 \\
SecureTL             & 11.36 & 0.8362 & 0.0035 & 0.288 \\
SecureTL w/KLD       & 13.58 & 0.8700 & 0.0 & 0.005 \\
SecureTL w/SL        & 13.33 & 0.9458 & 0.0  & 0.004 \\
\hline 
\end{tabular}
\end{table}

\begin{table}[!t]
\centering
\caption{Summary of the test results for face identity verification.}\label{tab:templatesec_face_results_summary}
\begin{tabular}{lcccc}\hline
 & \multicolumn{2}{c}{\textbf{Performance}} & \textbf{Cancelability} & \textbf{Linkability}\\
\textbf{Method} & $EER$ ($\%$) & $FNMR~@FMR_V=0.1\%$ & $FMR_C@~EER$ & $D_{\leftrightarrow}^{sys}$ \\\hline
Triplet Loss  & 13.99 & 0.8496 & -    & - \\
BF~\cite{GomezBarrero2016}     & 17.07 & 0.9103 & 0.0396 & 0.245 \\ 
HE~\cite{Drozdowski2019}       & 15.06 & 0.8312 & 0.0371 & 0.001 \\
SecureTL           & 13.61 & 0.8314 & 0.0966 & 0.399 \\
SecureTL w/KLD     & 15.93 & 0.8586 & 0.0089 & 0.132 \\
SecureTL w/SL      & 15.15 & 0.8771 & 0.0182 & 0.070 \\
\hline 
\end{tabular}
\end{table}

\subsection{Verification performance}

\begin{figure}
    \centering
    \includegraphics[width=0.9\linewidth]{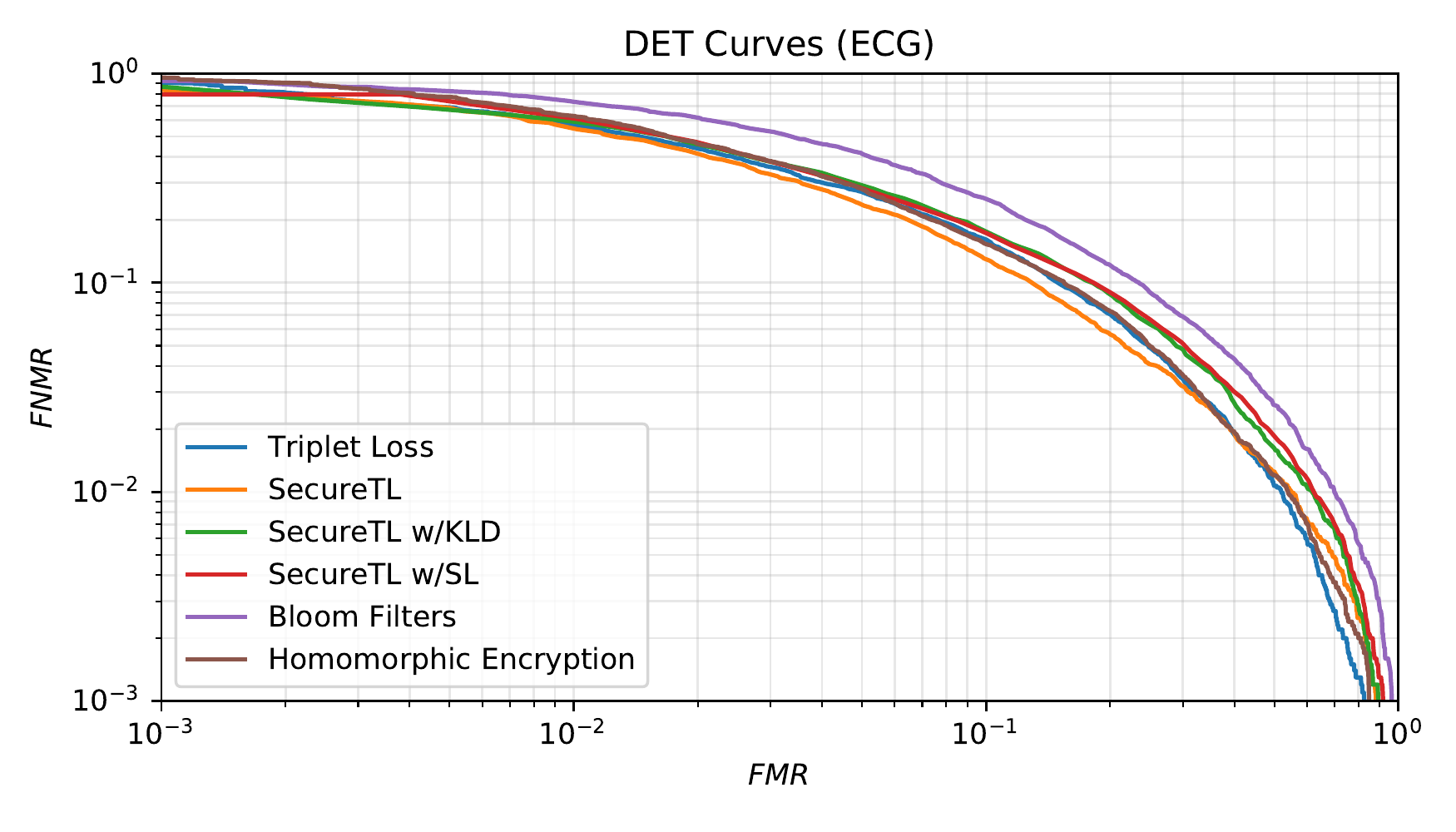}
    \caption{Detection Error Tradeoff (DET) curves for the ECG identity verification model when trained with the original triplet loss \emph{vs.} the proposed formulations of the Secure Triplet Loss.}
    \label{fig:templatesec_ecg_rocs}
\end{figure}

On ECG identity verification, the baseline method trained with triplet loss offered $12.56\%$ EER. This is similar to the results presented in the work that first proposed this end-to-end model~\cite{Pinto2019Deep, Pinto2019b}. As presented in Table~\ref{tab:templatesec_ecg_results_summary} and in the receiver-operating characteristic (ROC) curves in Fig.~\ref{fig:templatesec_ecg_rocs}, the first formulation of the Secure Triplet Loss (SecureTL), without considering linkability, attained $11.36\%$ EER, which is an improvement in performance over the original triplet loss despite the inclusion of a cancelability module.

The linkability-focused reformulations of the Secure Triplet Loss, which use the Kullback-Leibler divergence (SecureTL w/KLD) or distance statistics (SecureTL w/SL), led the model to attain, respectively, $13.58\%$ and $13.33\%$ EER. These results show that a small performance gap should be expected when considering both cancelability and linkability in the triplet loss. Recalling the performance improvements with the SecureTL formulation, it can be hypothesised that the performance decrease in SecureTL w/KLD and SecureTL w/SL is caused by measuring linkability in a separate loss module computed batch-by-batch. It is likely that, if linkability was better integrated into the Secure Triplet Loss, as was cancelability, then the performance gap would remain closed.  

\begin{figure}
    \centering
    \includegraphics[width=0.9\linewidth]{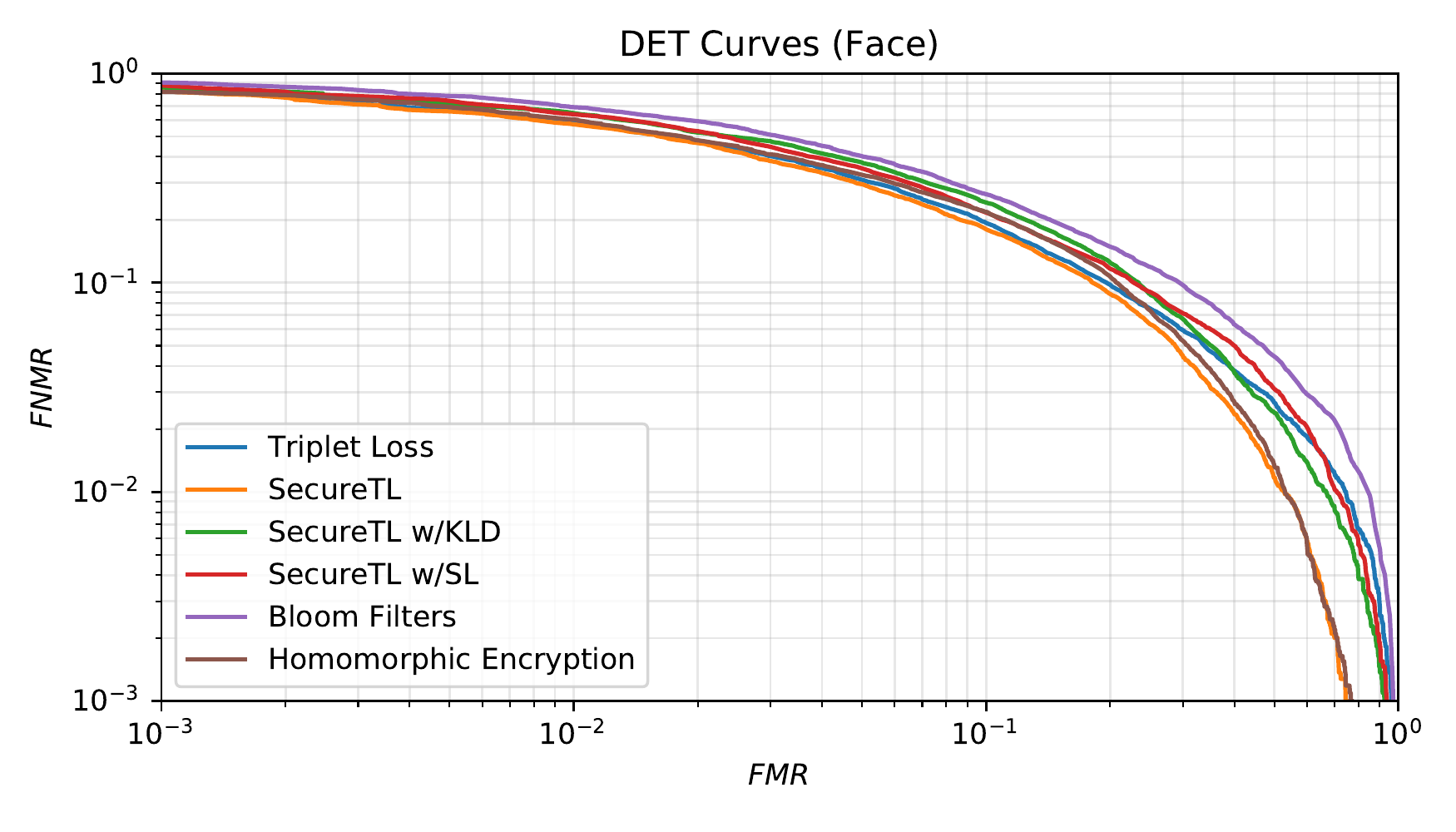}
    \caption{Detection Error Tradeoff (DET) curves for the face identity verification model when trained with the original triplet loss \emph{vs.} the proposed formulations of the Secure Triplet Loss.}
    \label{fig:templatesec_face_rocs}
\end{figure}

Nevertheless, the model trained with any of the proposed loss formulations still offers considerably better performance than the state-of-the-art methods. The best state-of-the-art method evaluated in the same conditions (in~\cite{Pinto2019b}) offered $21.82\%$ EER \emph{vs.} $13.58\%$ attained by SecureTL w/KLD and $13.33\%$ achieved by SecureTL w/SL. This denotes that the proposed method, while presenting a small performance gap with the linkability loss module, still retains most of the performance advantages associated with deep end-to-end models.

\begin{figure*}[p]
    \centering
    \includegraphics[width=0.78\linewidth]{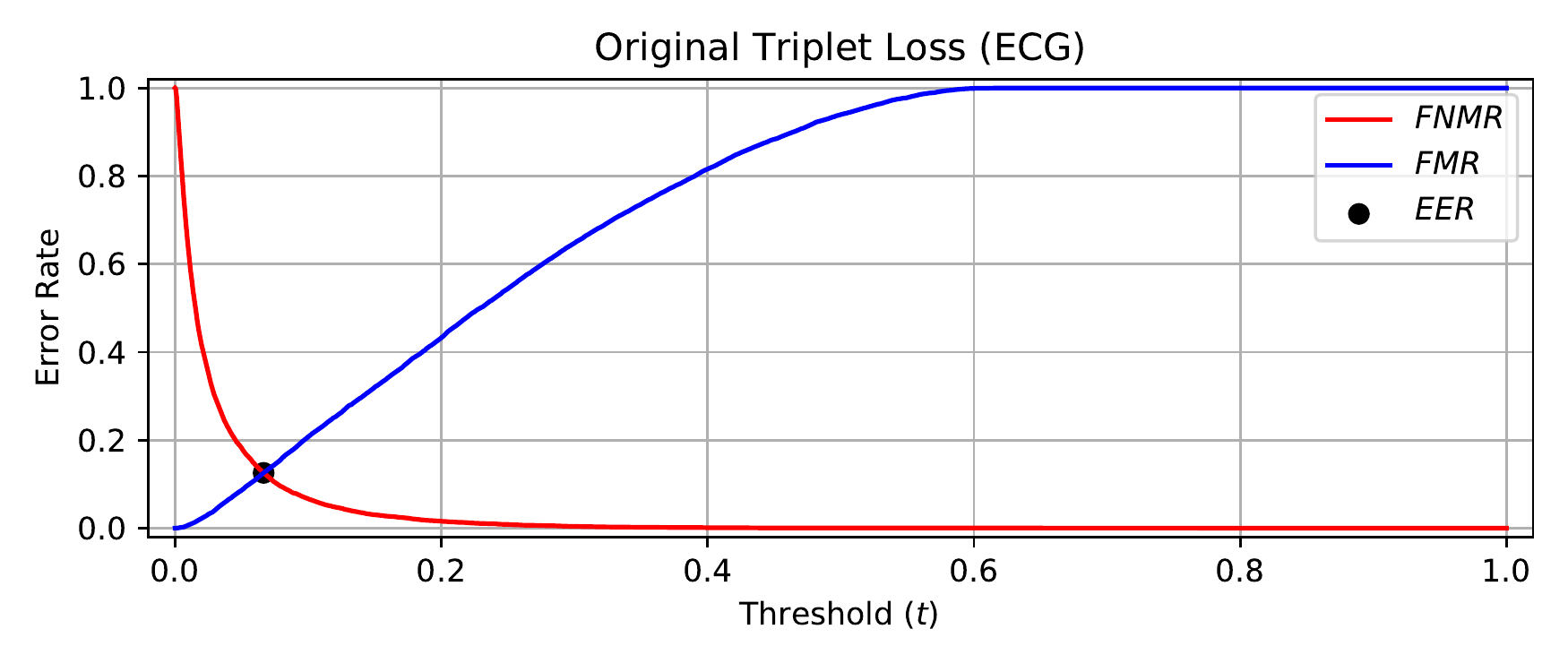}\\
    \includegraphics[width=0.78\linewidth]{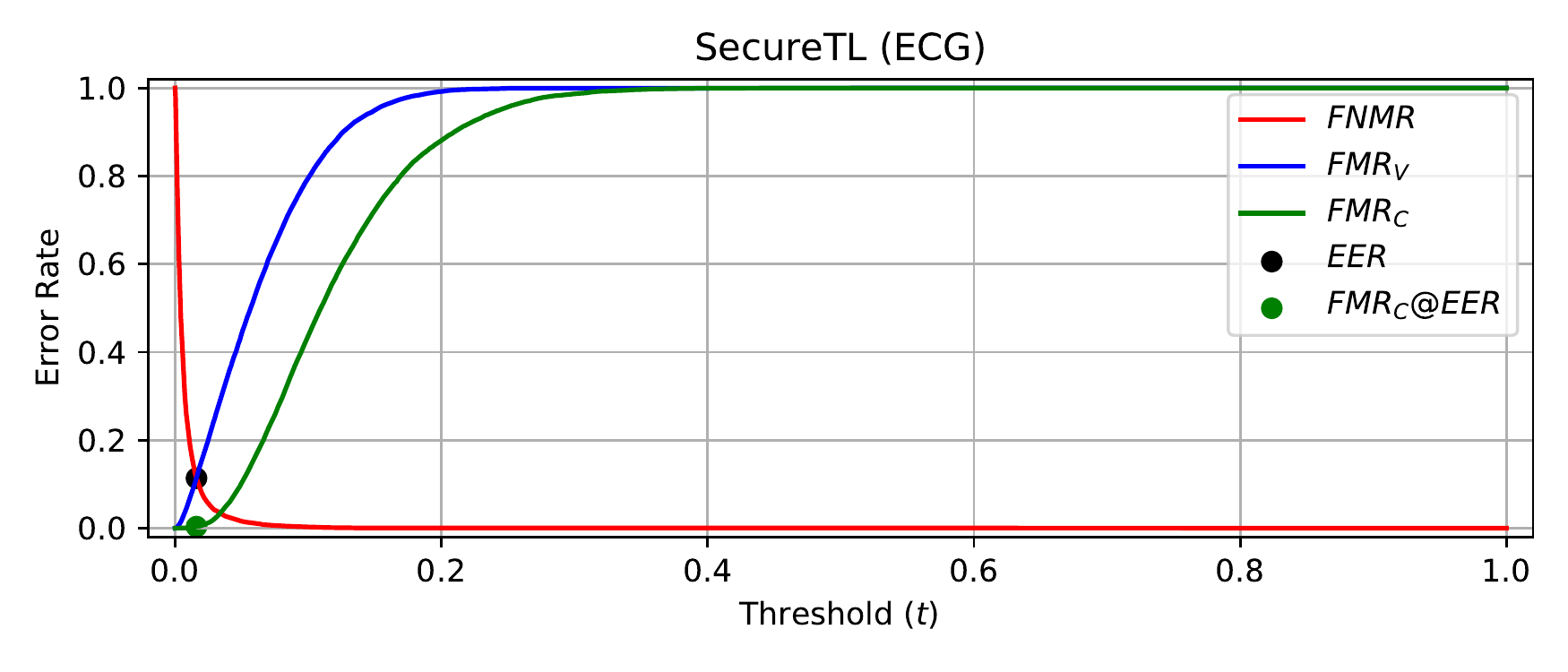}\\
    \includegraphics[width=0.78\linewidth]{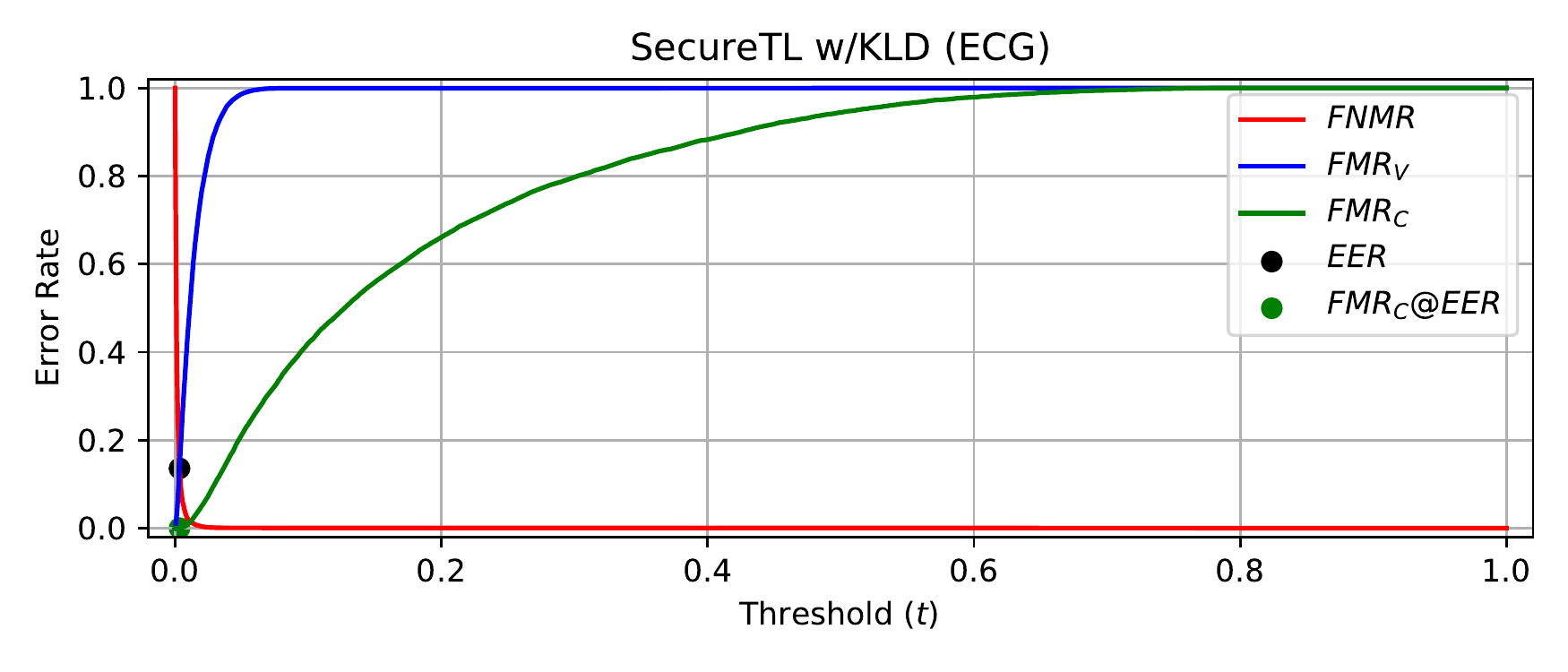}\\
    \includegraphics[width=0.78\linewidth]{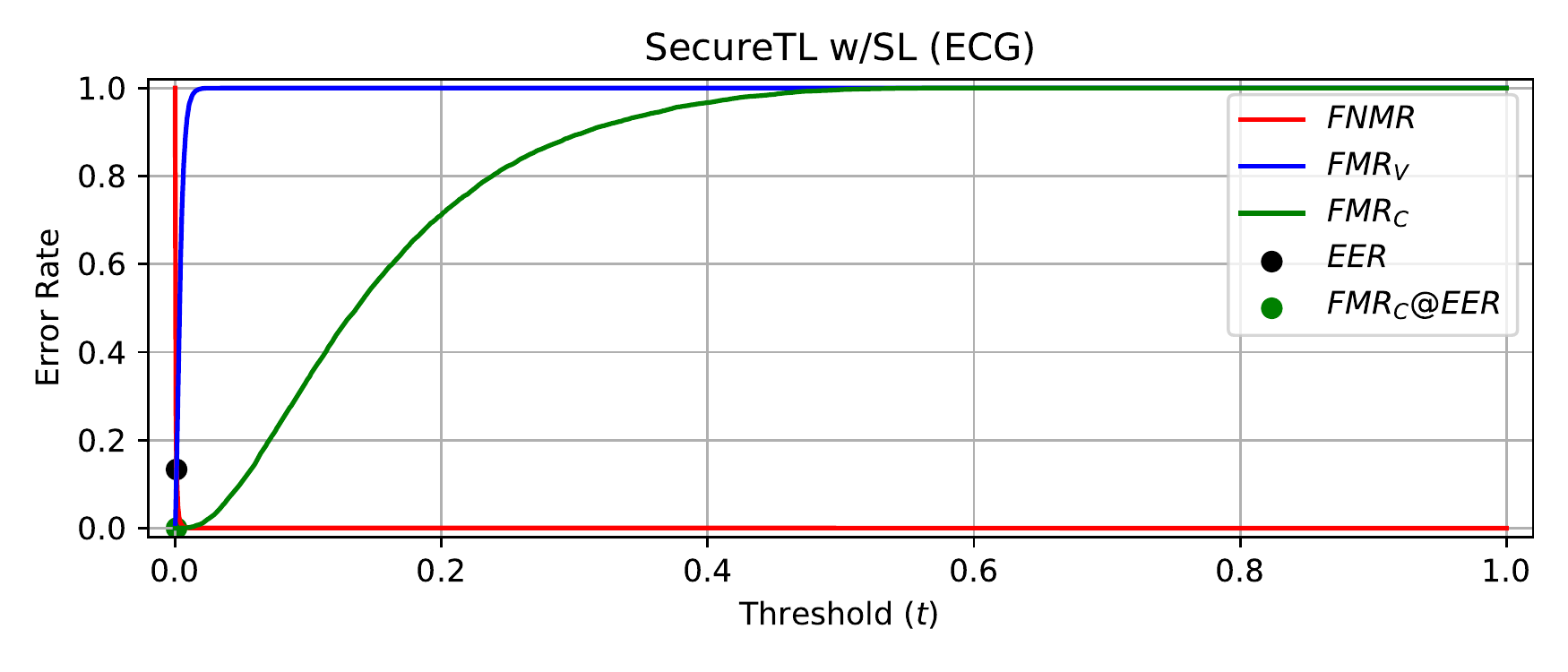}%
    \caption[False match rate ($FMR$) and false non-match rate ($FNMR$) curves w.r.t. the distance comparison threshold $t$, for ECG identity verification with triplet loss and the proposed Secure Triplet Loss formulations.]{False match rate ($FMR$) and false non-match rate ($FNMR$) curves w.r.t. the distance comparison threshold $t$, for ECG identity verification with triplet loss and the proposed Secure Triplet Loss formulations (the latter include both $FMR_P$, relative to verification error, and $FMR_C$, relative to cancelability error, as well as the $FMR_C$ that corresponds to the $EER$ point, $FMR_C@EER$).}
    \label{fig:templatesec_canc_eval_ecg}
\end{figure*}

\begin{figure*}[p]
    \centering
    \includegraphics[width=0.78\linewidth]{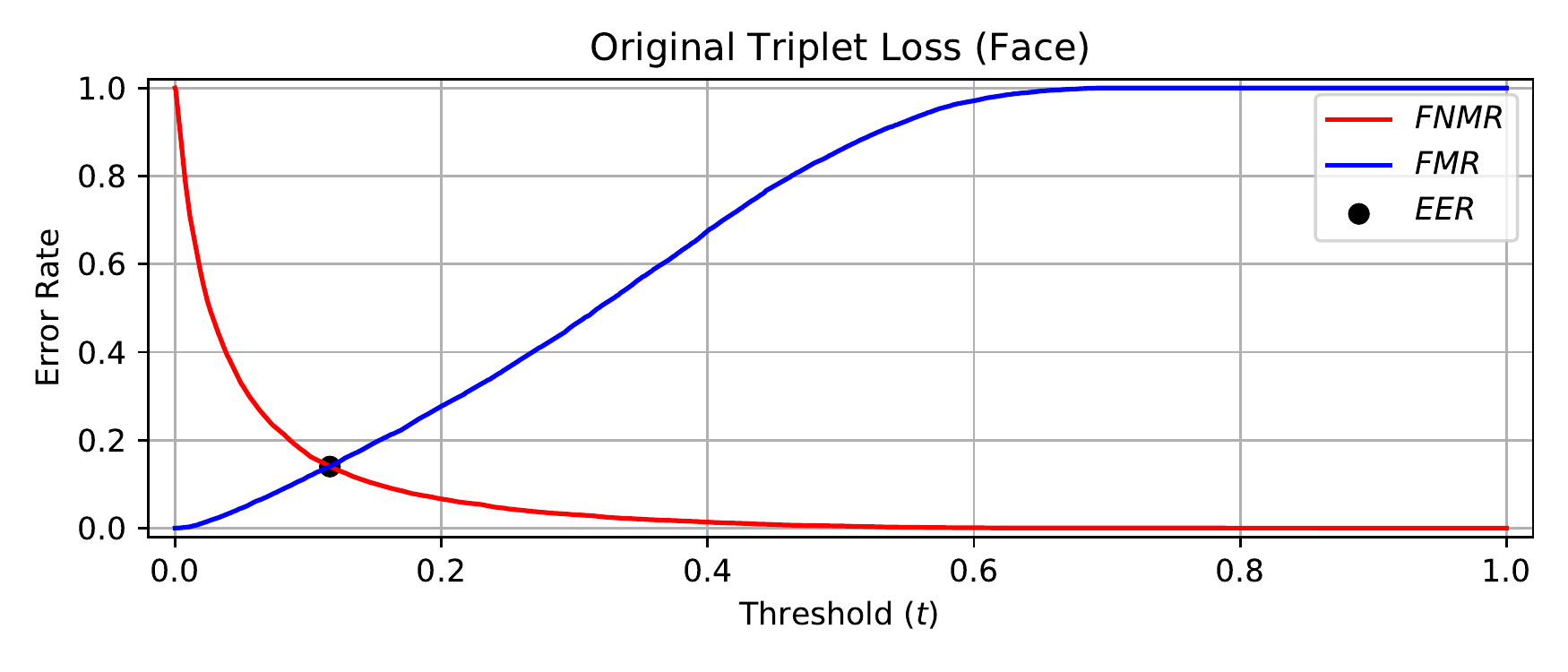}\\
    \includegraphics[width=0.78\linewidth]{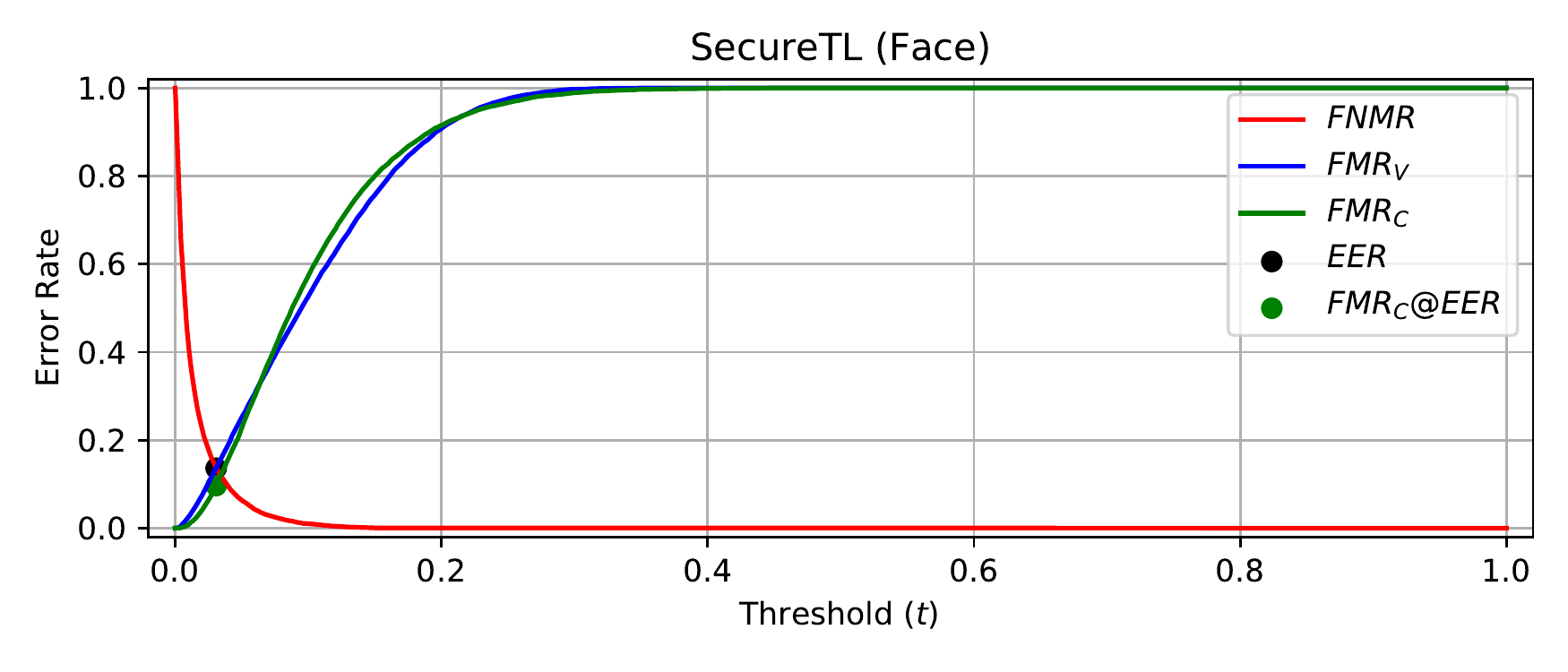}\\
    \includegraphics[width=0.78\linewidth]{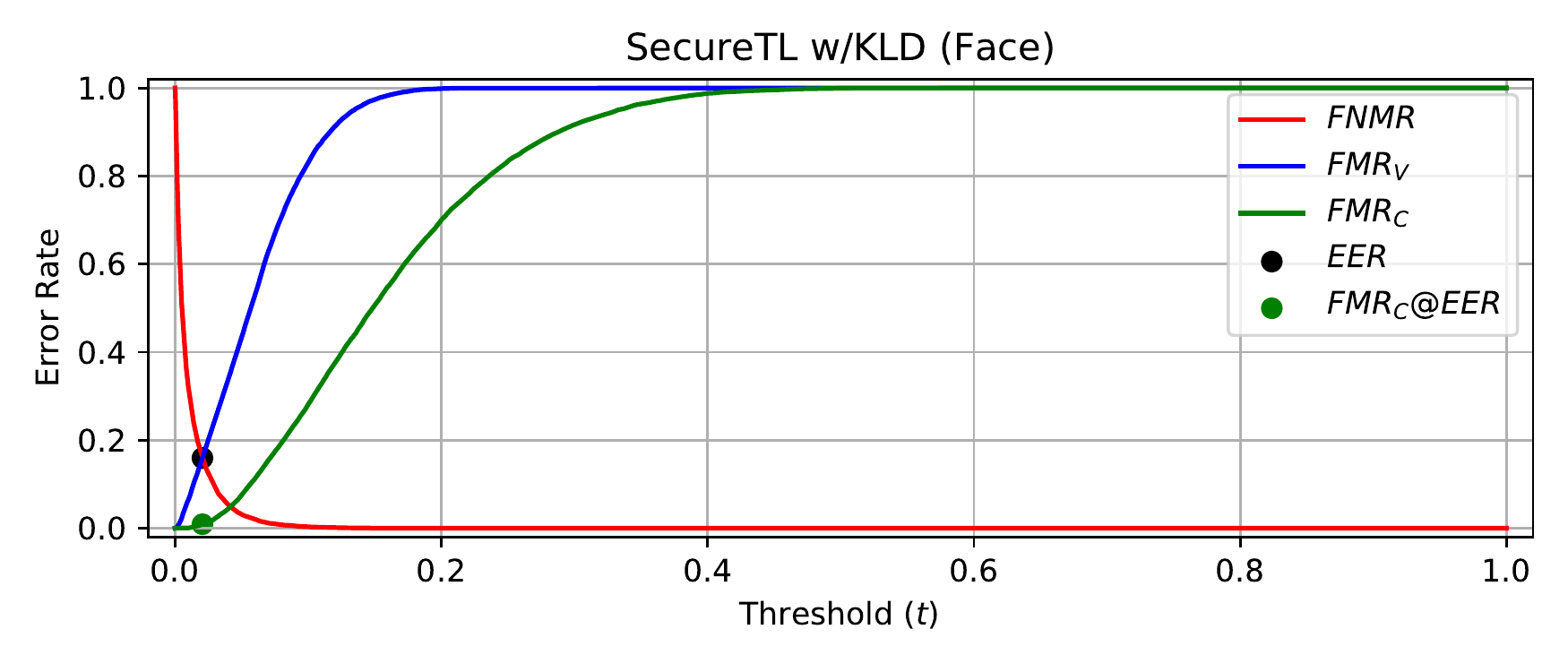}\\
    \includegraphics[width=0.78\linewidth]{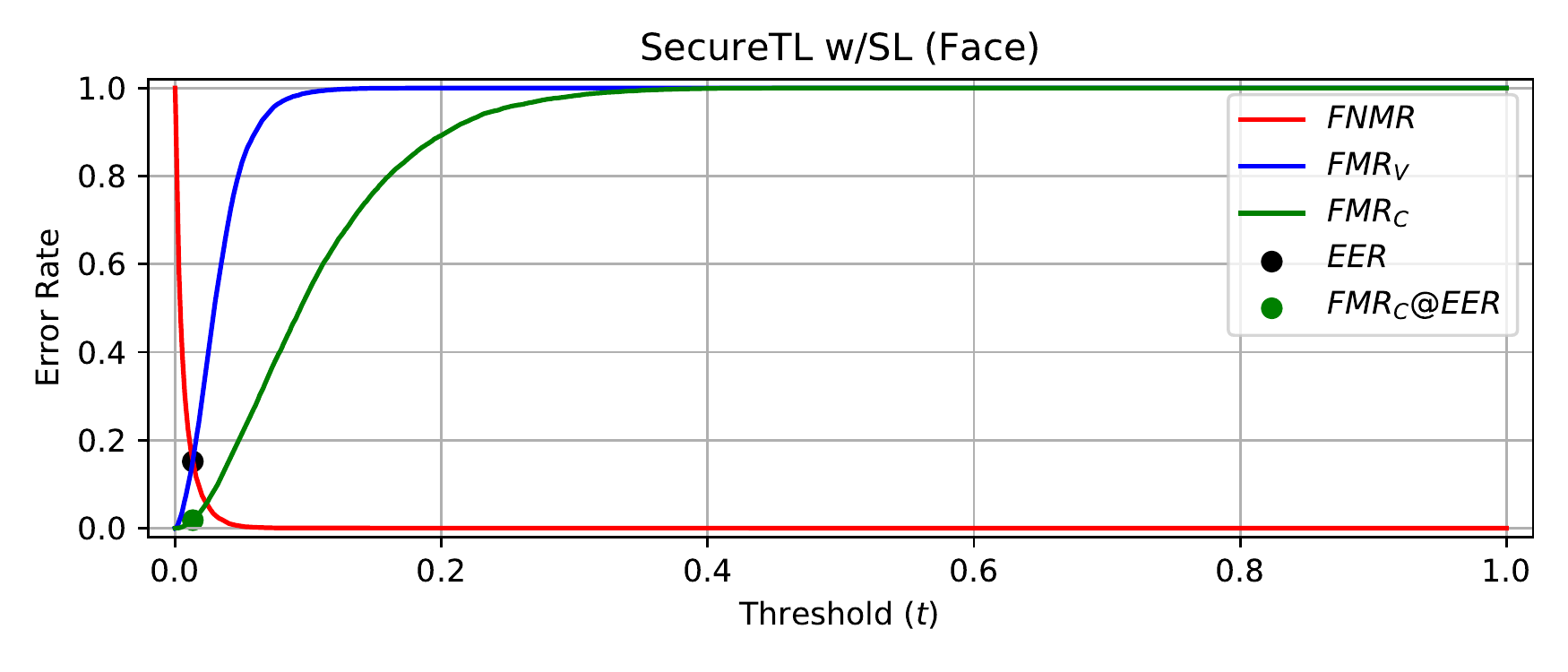}%
    \caption[False match rate ($FMR$) and false non-match rate ($FNMR$) curves w.r.t. the distance comparison threshold $t$, for face identity verification with triplet loss and the proposed Secure Triplet Loss formulations.]{False match rate ($FMR$) and false non-match rate ($FNMR$) curves w.r.t. the distance comparison threshold $t$, for face identity verification with triplet loss and the proposed Secure Triplet Loss formulations (the latter include both $FMR_P$, relative to verification error, and $FMR_C$, relative to cancelability error, as well as the $FMR_C$ that corresponds to the $EER$ point, $FMR_C@EER$).}
    \label{fig:templatesec_canc_eval_face}
\end{figure*}

For face identity verification, the performance results are presented in Table~\ref{tab:templatesec_face_results_summary} and in the ROC curves in Fig.~\ref{fig:templatesec_face_rocs}. The model trained with the triplet loss attained $13.99\%$ EER, which seems adequate given the difficulty of the evaluation settings: YouTube Faces provides a challenging framework for evaluation (noted by the $95.12\%$ accuracy achieved by the Inception-ResNet model on this database \emph{vs.} $99.63\%$ on the LFW database), disjoint subsets of identities are used for training/validation and testing, and each identity is only represented by a single template for each comparison (the gallery size is $1$). 

In harmony with the results on ECG, the model trained with the Secure Triplet Loss without a linkability component offered a small improvement in verification performance ($13.61\%$ EER). Likewise, the addition of a linkability-measuring term to the loss leads to a $2\%$ increase in EER. This confirms the aforementioned belief that the separate linkability loss term is affecting performance and improvements could be achieved by integrating it into the Secure Triplet Loss in a more cohesive way.

Overall, the verification performance results denote that it is possible to adequately train or fine-tune an end-to-end model with the proposed loss formulations. With either biometric characteristic, the performance difference between using KLD and distance statistics is not appreciable, which denotes these formulations may each be fitted for specific settings or used interchangeably.

\subsection{Cancelability evaluation}

As aforementioned, by integrating identity verification and template cancelability into a single comparison score, template cancelability is not necessarily ensured. Hence, the results of false match rates based on cancelability ($FMR_C$) are presented, in Fig.~\ref{fig:templatesec_canc_eval_ecg} and Fig.~\ref{fig:templatesec_canc_eval_face}, alongside the false match rates based on verification ($FMR_V$) and the common false non-match rates ($FNMR$).

In all cases, the $FMR_C$ is lower than $FMR_V$ at and around the $EER$ operation point. In most cases, $FMR_C$ at this point is very small and is lower than or equal to $FMR_V$ for all operation points, which is highly desirable. As presented in Table~\ref{tab:templatesec_ecg_results_summary} and Table~\ref{tab:templatesec_face_results_summary}, cancelability error is significantly lower in the ECG models. As shown by the results, SecureTL w/KLD and w/SL appear to be better at promoting cancelability than the original secure loss formulation, denoting that the linkability loss term could have a positive effect on cancelability. 

Considering these results and the increased difficulty experienced while fine-tuning the face models, one can conclude that the proposed Secure Triplet Loss is likely better fitted for training models from scratch than to adapting previously trained models to become secure. Nevertheless, the cancelability results, especially with the SecureTL w/KLD and SecureTL w/SL, are encouraging in either case.

\subsection{Unlinkability evaluation}

\begin{figure*}[t!]
    \centering
    \includegraphics[width=0.5\linewidth]{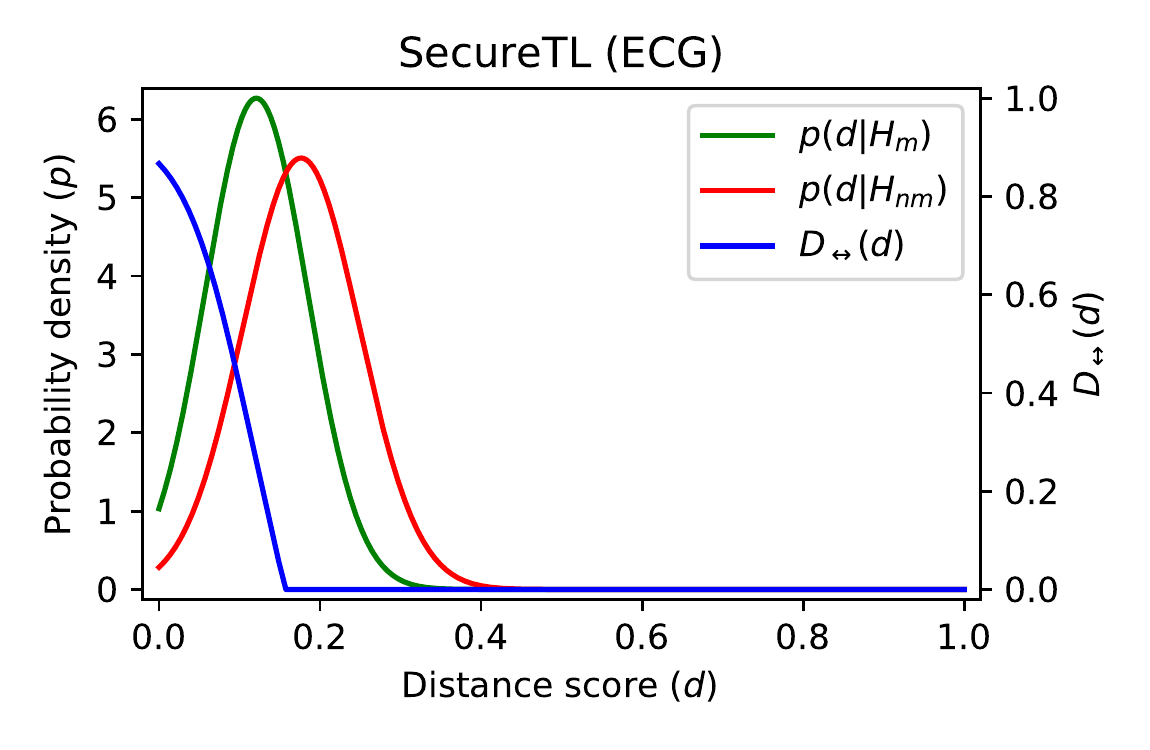}%
    \includegraphics[width=0.5\linewidth]{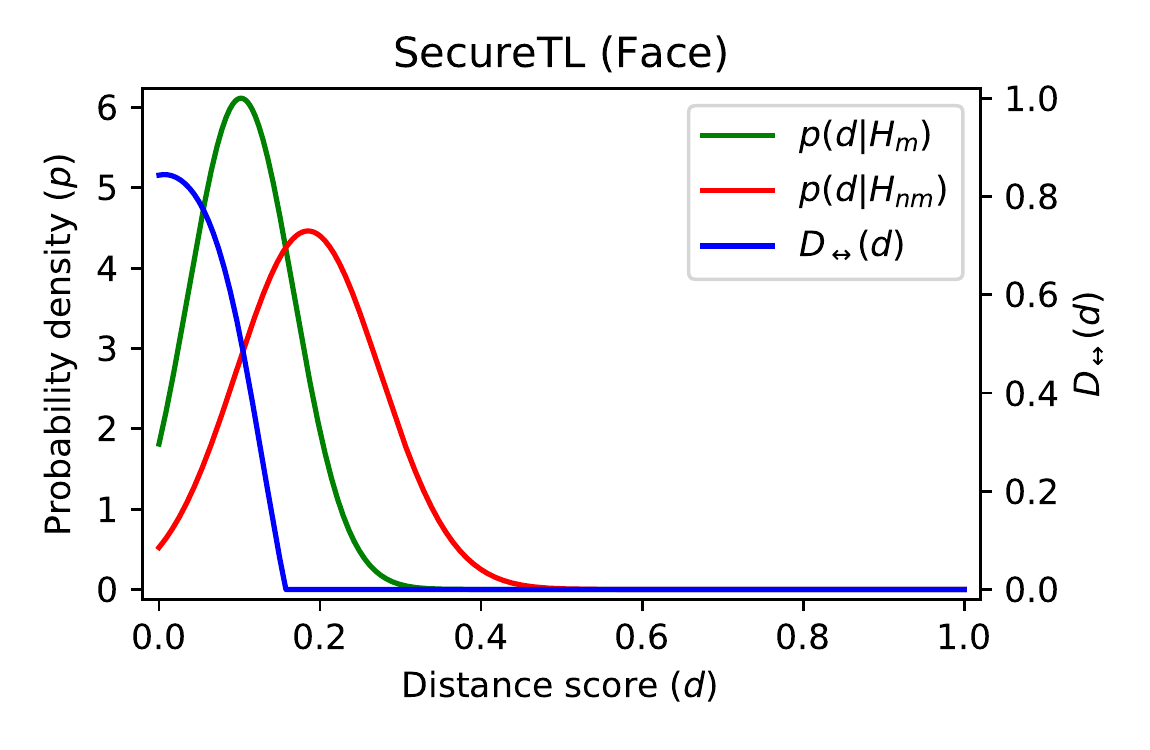}\\
    \includegraphics[width=0.5\linewidth]{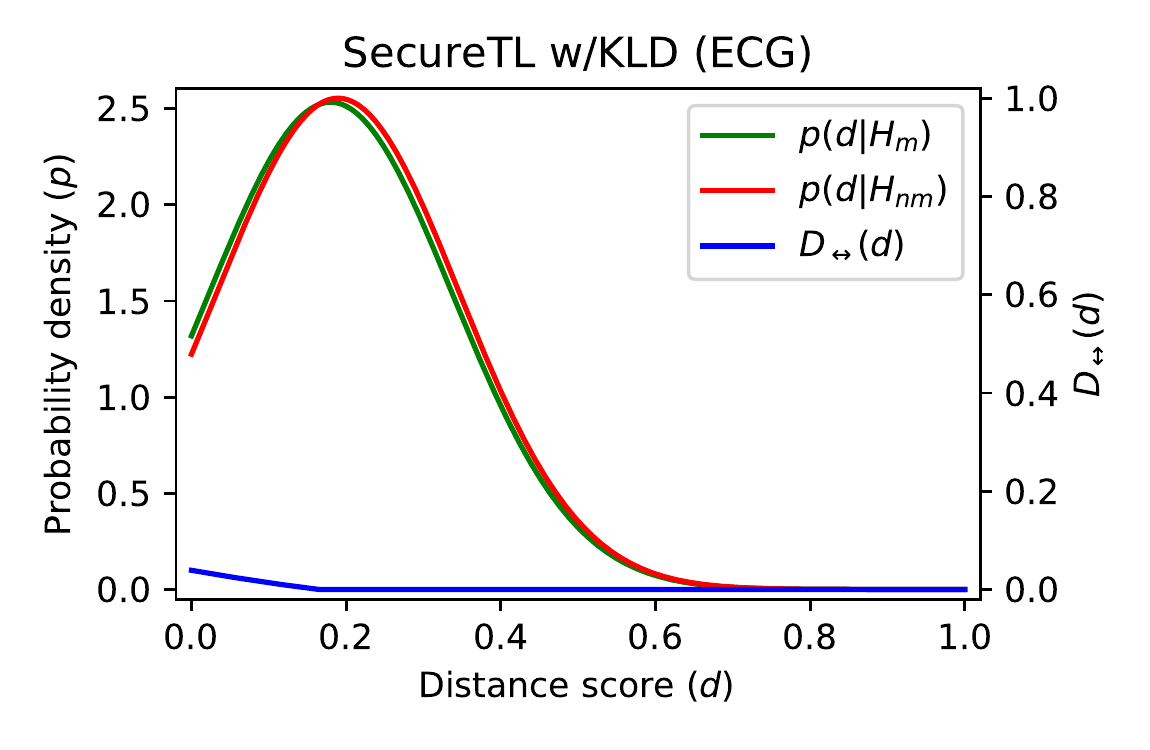}%
    \includegraphics[width=0.5\linewidth]{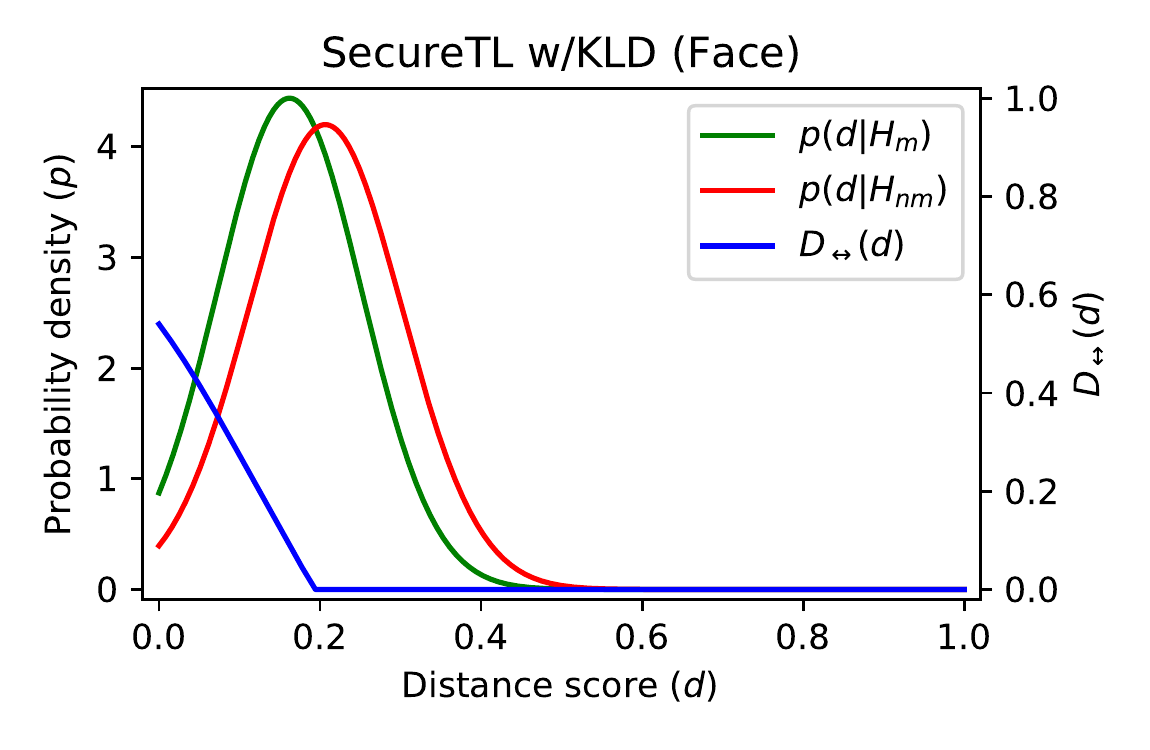}\\
    \includegraphics[width=0.5\linewidth]{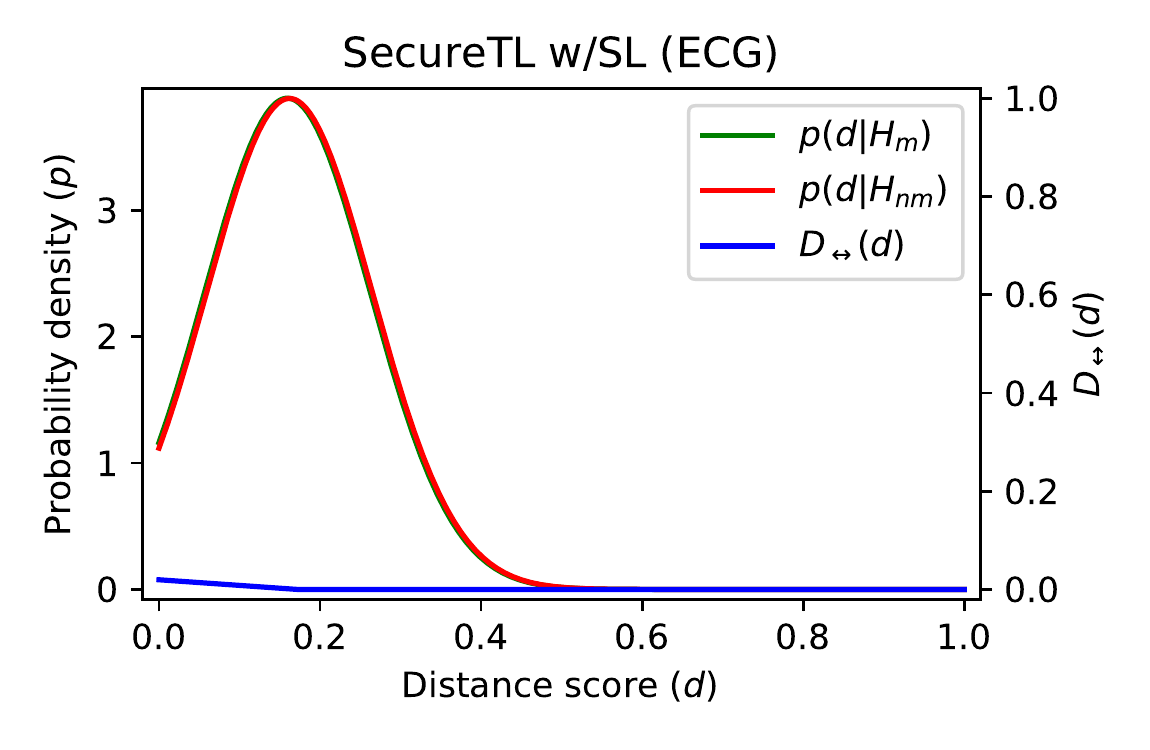}%
    \includegraphics[width=0.5\linewidth]{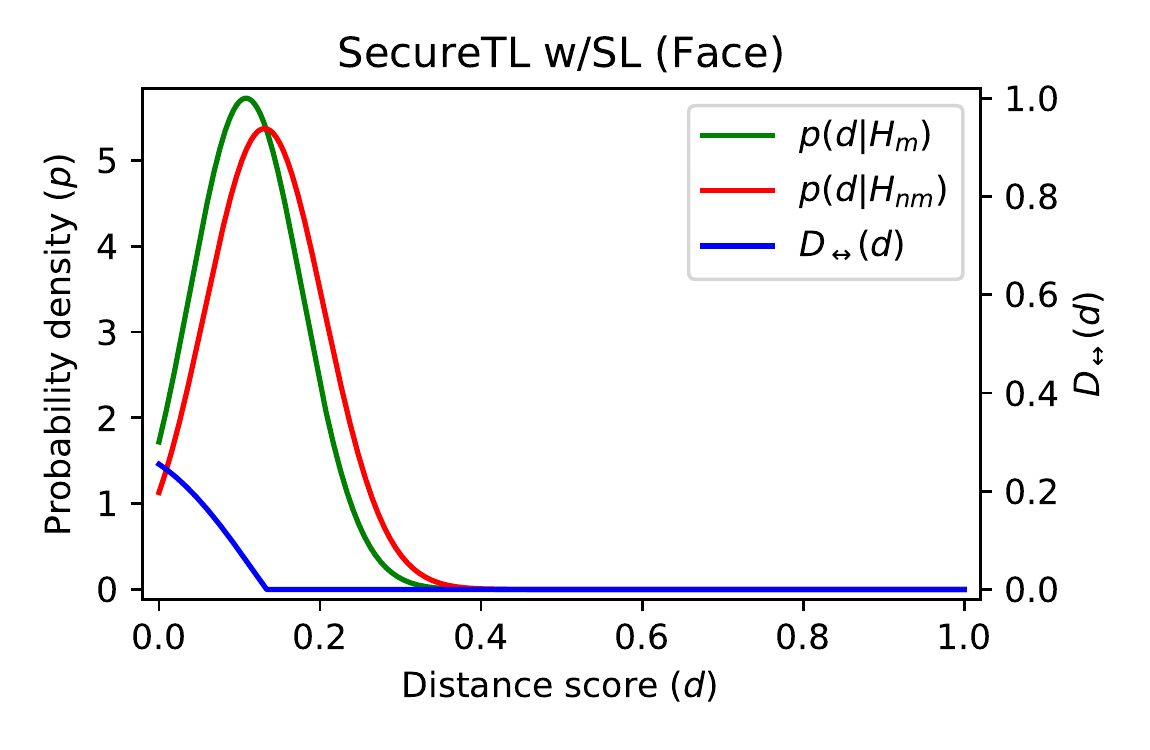}%
    \caption[Template linkability analysis for the ECG and face identity verification models.]{Template linkability analysis for the ECG and face identity verification models (following the procedure proposed by~\citet{GomezBarrero2016}).}
    \label{fig:templatesec_link_eval}
\end{figure*}

The results of the linkability analysis following the framework established in~\cite{GomezBarrero2016} are presented in Fig.~\ref{fig:templatesec_link_eval}. In both cases, the original formulation of the Secure Triplet Loss presents relatively high $D_{\leftrightarrow}^{sys}$ ($0.288$ for ECG and $0.399$ for face). However, the result with ECG is better than the equivalent reported earlier in~\cite{Pinto2020iwbf} ($0.67$). This results from the fact that linkability was not promoted by this loss during model training: hence, the model may achieve adequate unlinkability, but that would be accidental.

In the case of the proposed SecureTL w/KLD and SecureTL w/SL, linkability is actively promoted during training through the loss. The effects of this loss reformulation are clear: the probability density functions of mated and non-mated are more superposed, which indicates that it would be more difficult, as desired, to distinguish identities in pairs of templates where the keys do not match.

With ECG, $D_{\leftrightarrow}^{sys}$ assumes the values $0.005$ for SecureTL w/KLD and $0.004$ for SecureTL w/SL. With face, it assumes $0.132$ for SecureTL w/KLD and $0.070$ for SecureTL w/SL. All of these can be considered semi to fully-unlinkable. Just as with cancelability, the proposed method seems more adequate for training models from scratch than for fine-tuning existing biometric models. Additionally, using KLD appears to offer some advantages in linkability for ECG verification, but that should be weighted with the increased instability this alternative has shown during training, relative to SecureTL w/SL, especially in face verification.

\subsection{Non-invertibility and secrecy leakage}

Regarding other security metrics, the privacy leakage rate was estimated as $1$ for the model trained with any of the losses. This indicates that it is highly difficult for an attacker to recover the original biometric measurements $x$ based on compromised templates $y$ output by the model. This could be a result of using end-to-end deep learning models: recent research indicates that optimised deep models compress the inputs retaining only the information needed for the task~\cite{Tishby2015}. This means perfect non-invertibility can be achieved without carefully handcrafted feature extraction algorithms.

Similarly, all losses led the model to offer a perfect secrecy leakage rate of $0$, which denotes that the public keys used to make the templates cancelable reveal no information on them. These results on non-invertibility and secrecy leakage do not show a superiority of the proposed loss formulations over the original triplet loss but emphasise the meaningful advantages of using end-to-end deep learning models for secure biometrics. 

\subsection{Comparison with state-of-the-art approaches}

The proposed method was compared with two state-of-the-art approaches: Bloom Filters (BF) and Homomorphic Encryption (HE), as described in~\cite{GomezBarrero2016} and~\cite{Drozdowski2019}, respectively. To provide a fair and direct comparison between the template protection schemes, the features given to those methods were those output by the triplet loss baseline model.

The results are presented in Table~\ref{tab:templatesec_ecg_results_summary}, Table~\ref{tab:templatesec_face_results_summary}, Fig.~\ref{fig:templatesec_ecg_rocs}, and Fig.~\ref{fig:templatesec_face_rocs}. Both with face and ECG, the proposed method outperformed BF in $EER$, cancelability, and linkability. HE offered the best linkability results, at the cost of poor cancelability. Additionally, HE took significantly longer for biometric comparison than any of the alternatives, which may grant it limited real applicability. 

Although the error results are relatively high, the Secure Triplet Loss is competitive \emph{vs.} the state-of-the-art alternatives, especially on cancelability and linkability. Moreover, improved results are expected when the Secure Triplet Loss is used on more accurate biometric models.

\subsection{Effects of varying $\gamma$}

\begin{figure}[!t]
    \centering
    \includegraphics[width=0.9\linewidth]{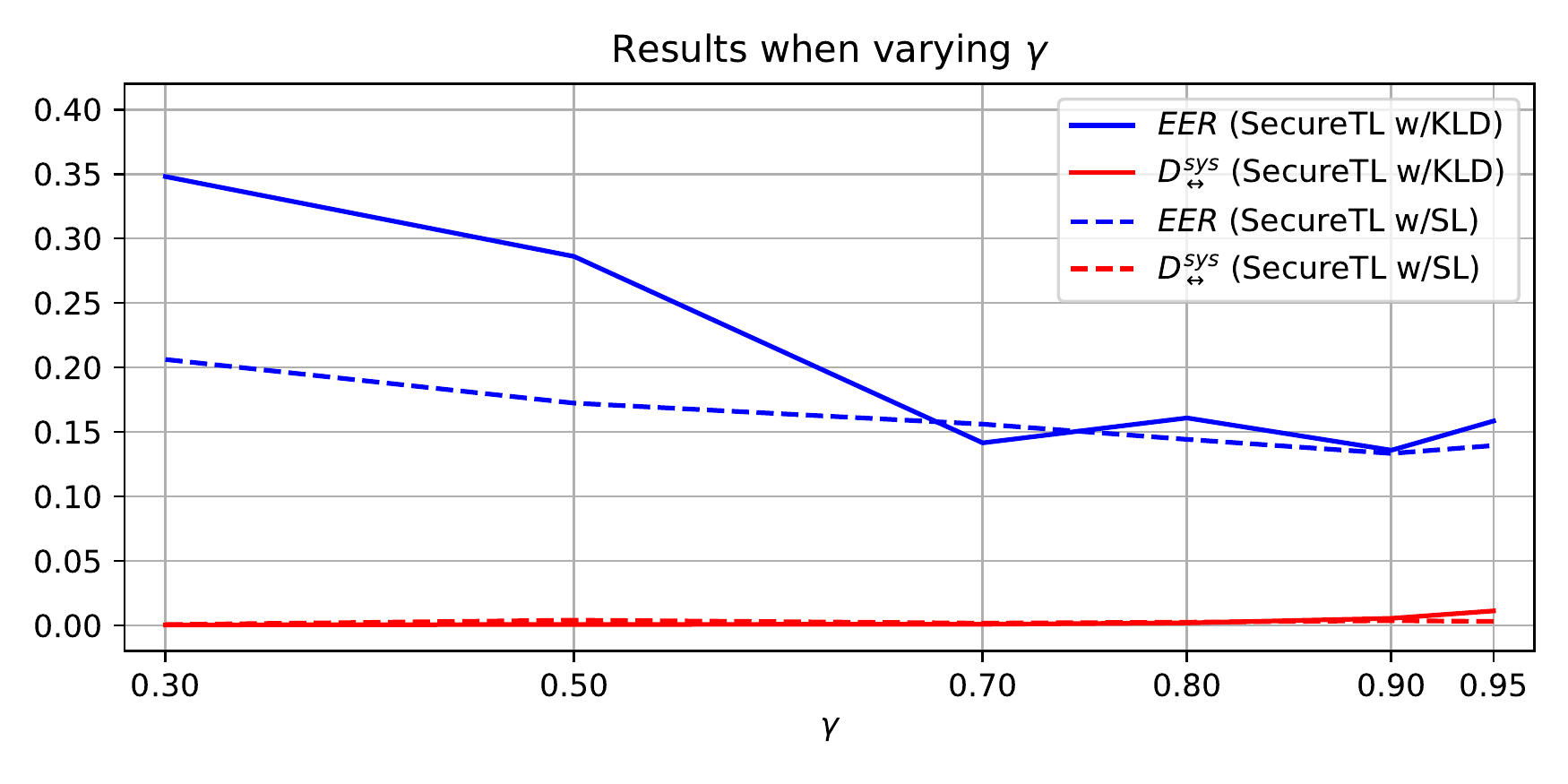}
    \caption{Results with the proposed loss when varying the $\gamma$ parameter.}
    \label{fig:templatesec_var_gamma}
\end{figure}

Fig.~\ref{fig:templatesec_var_gamma} presents the $EER$ and $D_{\leftrightarrow}^{sys}$ results obtained when varying the $\gamma$ parameter which balances the original secure triplet loss formulation and the template linkability component. As shown, lower $\gamma$ values ($\gamma<0.7$) lead to higher $EER$ with either SecureTL w/KLD or SecureTL w/SL, since template unlinkability takes precedence over verification accuracy on the loss that guides model training. For $\gamma\geq0.7$, lower $EER$ results are obtained, albeit with a slight increase in template linkability ($D_{\leftrightarrow}^{sys}$), especially for $\gamma>0.9$. Results may vary in other application scenarios depending on their specificities, but $0.7<\gamma<0.95$ should offer the highest likelihood of success.

\section{Summary and Conclusions}

This work presented the Secure Triplet Loss, a methodology focused on training end-to-end deep biometric models, without any additional processes, to verify template cancelability, unlinkability, and non-invertibility. The results on ECG and face identity verification show that the proposed method is not only able to fulfil this purpose, but also to adapt pretrained biometric models to offer secure templates, with competitive performance results.

However, there is still room for improvement. Further efforts should be devoted to designing ways to better integrate linkability in the Secure Triplet Loss, in order to avoid performance decreases. A scheme where linkability would be measured triplet-by-triplet (instead of batch-by-batch), similarly to cancelability, should lead to improved performance using the Secure Triplet Loss. This would also enable the formulation of triplet mining approaches for the proposed method. Nevertheless, the Secure Triplet Loss is, overall, a suitable and flexible general scheme for template protection in end-to-end deep biometrics.
\chapter[Self-Supervised Learning with Sequential Data]{Self-Supervised Learning\\with Sequential Data}\label{ch:selfsupervised}

\begin{tcolorbox}\footnotesize
{\large\bf Foreword on Author Contributions}

The research work described in this chapter was conducted entirely by the author of this thesis, under the supervision of Jaime S. Cardoso. The results of this work have been disseminated in the form of an article in international conference proceedings:
\begin{itemize}[noitemsep, leftmargin=1em, nosep]
    \item \underline{J. R. Pinto} and J. S. Cardoso, ``Self-Learning with Stochastic Triplet Loss,'' in \emph{International Joint Conference on Neural Networks (IJCNN 2020)}, Jul.~2020.~\cite{Pinto2020SelfLearning}
\end{itemize}

\end{tcolorbox}

\section{Context and Motivation}

In recent years, deep learning algorithms have offered improved performance over handcrafted methodologies in several pattern recognition tasks. These commonly take advantage of convolutional layers, which enable the autonomous learning of the most relevant features for the task at hand, and use fully-connected layers for more intricate decision boundaries~\cite{LeCun2015}. However, these improvements come with an important drawback: the need for labelled data.

Most tasks where such performance breakthroughs have been achieved are those where researchers have plenty of labelled data at their disposal. The ImageNet dataset enabled the training of deeper models for better performance in the detection and recognition of objects. Similarly, datasets such as VGGFace~\cite{Parkhi2015} and VGGFace2~\cite{Cao2018} helped in the development of improved models for biometric recognition based on face images.

However, for some pattern recognition problems, supervised data is scarce. In most of these cases, even though available data is plenty, the annotation process is cumbersome and/or expensive. This is very frequent in automatic medical image diagnosis tasks, where several imaging exams are usually available but lack specific annotations which typically would need to be offered by experts.

Another exemplary application is video surveillance. Given the current ubiquity of surveillance cameras, the availability of data is not a problem. However, the annotation of individuals on the recordings is a long and expensive endeavour. This limits the performance one can attain in these tasks since deeper models will be harder to train.

Yet another key application is continuous biometrics with electrocardiogram (ECG) signals~\cite{Pinto2018}. Deep learning has offered improved performance through increased robustness to signal noise and variability~\cite{Pinto2019Deep}. However, in scenarios with off-the-person signals (acquired during normal activity using few dry electrodes on the fingers and palms), the performance still fails to match the use of cleaner on-the-person signals (acquired on medical-grade settings from subjects at rest, using several wet electrodes on the chest and limbs)~\cite{Luz2018, Pinto2019b, Zhang2017}. This is verified because current off-the-person signal datasets are too few and too small to train larger and deeper models. 

Self-learning (SL) has emerged as a promising approach to learn from unlabelled data. Contrarily to unsupervised learning, SL uses contextual or prior information to automatically define labels and tasks, which then support the learning. Generally, self-learning methods are focused on specific applications, including visual representation learning~\cite{Feng2019, Gomez2017}, action classification~\cite{Fernando2017}, or human motion capture~\cite{Tung2017}, thus including details that restrict their use to those specific tasks.

On the other hand, several self-learning methods use simple low-level tasks for the training, such as ranking samples~\cite{Li2019} or recovering masked parts of an image~\cite{Trinh2019}, that do not necessarily guarantee that the learned parameters can be useful for high-level tasks. For example, features that prove adequate for the approximate reconstruction of biometric samples may not be good enough to discriminate their identities. Hence, there is currently a need for a more general and capable self-learning methodology.

In this work, we propose a novel formulation of the triplet loss~\cite{Chechik2010} for self-supervised learning with unlabelled data, unrestricted to specific problems. The triplet loss is particularly suited for this task since it does not require absolute labels for the samples. As samples are combined into triplets, only their relative label information is required. The proposed formulation finds its key application in sequential data scenarios, where mild assumptions about the data acquisition process enable the adoption of a stochastic adaptation of the triplet loss to trigger and sustain the learning.

In the experimental work, the proposed methodology is successfully applied to off-the-person ECG-based biometric identity verification tasks, using signals from the University of Toronto ECG Database (UofTDB)~\cite{Wahabi2014}, and to unconstrained face identity verification, using the YouTube Faces dataset~\cite{Wolf2011}. Specific stress experiments were conducted on the ECG-based identity verification task to evaluate the behaviour of the proposed methodology in strain conditions.

\section{The Stochastic Triplet Loss}\label{sec:selflearn_methodology}

The triplet loss~\cite{Chechik2010} uses triplets of data samples to train a network to accurately assess if two samples belong to the same class. Each triplet is composed by an anchor $x_a$ with identity $i_a$ and two other samples, one positive ($x_p$) and one negative ($x_n$), where $i_p = i_a \neq i_n$. The three samples are processed, in parallel, by the same network, which returns a learned representation of each of them ($y_a$, $y_p$, and $y_n$). A measure of distance $d$ is then used to compare the anchor-positive ($d_+ = d(y_a, y_p)$) and anchor-negative ($d_- = d(y_a, y_n)$) pairs of representations, which are used in the computation of the triplet loss. The triplet loss for a single triplet can be defined as:
\begin{equation}
    {\cal L}(x_a, x_p, x_n) = \max(0, \alpha + d_+ - d_-).
\end{equation}

During training, the goal to decrease the triplet loss will lead the model to adjust its weights to obtain a final representation which brings samples of the same class closer together (reducing $d_+$), and samples of different classes further apart (increasing $d_-$). Here, the margin parameter $\alpha$ will contribute to enforce a minimum distance margin between the samples of different classes.

In the standard triplet loss, it is certain that $x_p$ is sampled from the same class as $x_a$ and $x_n$ is sampled from a class different from the class of $x_a$. This can be rewritten as
$P(\mathbbm{I}_{i_a}(i_p)=1)=1$ and $P(\mathbbm{I}_{i_a}(i_n)=0)=1$, where $\mathbbm{I}_A(x)$ is the indicator function.

With unlabelled samples, there is an uncertainty associated with the generation of the triplets, arising from the possibility of errors during the selection of the positive and negative samples.
We generalise the previous assumption by modelling $\mathbbm{I}_{i_a}(i_p)$ as a random variable following a Bernoulli distribution with parameter $\beta$, \ie, $P(\mathbbm{I}_{i_a}(i_p)=1)=\beta$. Similarly, $\mathbbm{I}_{i_a}(i_n)$ is assumed to follow a Bernoulli distribution with parameter $\gamma$, \ie, $P(\mathbbm{I}_{i_a}(i_n)=0)=\gamma$.

Assuming the independence of $\mathbbm{I}_{i_a}(i_p)$ and $\mathbbm{I}_{i_a}(i_n)$, and conditioned on the true identity of the observations in the triplet, the triplet loss follows a multinoulli distribution, with:
\begin{equation}\small
\label{noisyLoss}
   {\cal L}(x_a, x_p, x_n) =
   \begin{dcases}
      \max(0, \alpha + d_+ -  d_-), & \mbox{ with probability } \beta\gamma \\
      \max(0, \alpha + d_- -  d_-), & \mbox{ with probability } (1-\beta)\gamma \\
      \max(0, \alpha + d_+ -  d_+), & \mbox{ with probability } \beta(1-\gamma) \\
      \max(0, \alpha + d_- -  d_+), & \mbox{ with probability } (1-\beta)(1-\gamma) \\
   \end{dcases}
\end{equation}

On average, the middle terms in Eq.~\eqref{noisyLoss} do not contribute to the learning. The last term in the equation negatively impacts the learning process. In practice, one would need more data/time to learn under the noisy sampling of the triplets. The parameters $\beta$ and $\gamma$ guide the training of the model through the triplet loss, and their values depend on the specificities of the task and the data. In ideal conditions, these should be as close as possible to $1$ to approximate the original triplet loss in supervised settings. Lower values would work against the purpose of the triplet loss and diminish its training effectiveness (or, equivalently, increase the difficulty of the training).

The proposed self-learning methodology can be used to train models with unsupervised data. During triplet generation, after the selection of an anchor sample, one can randomly draw one sample from the dataset to serve as the negative sample. Assuming a balanced dataset with $C$ classes will give $\gamma = 1 - 1/C$. If $C$ is large, the probability of errors in negative sample selection $p(i_a=i_n)$ will be very low (\eg, $0.1$ for $C=10$ or $0.01$ for $C=100$), and so will be their impact on the training process. More importantly, in practice, prior knowledge allows us to adopt a sampling strategy with a much higher probability of success.

The positive sample can be obtained through the transformation of the anchor according to $x_p = f(x_a)$. The transformation $f$ should be carefully defined in order to change the anchor according to an expected range of intraclass variability but without degrading the underlying label information carried by the sample. The probability $\beta$ will depend on the degree to which $f$ complies with this need. Similarly to the negative sample selection, prior knowledge of the data may be useful to maximise the probability of success. For example, when dealing with sequential data, choosing a positive sample closer in time to the anchor will increase the probability of both samples sharing the same label. However, the anchor and the positive sample will likely be more similar, which will restrict the model's robustness to intraclass variability. Hence, one should find a trade-off between ensuring intraclass variability and maximising the probability of success in positive sample generation.

An approximation of the expected value of the loss in Eq.~\eqref{noisyLoss} can be computed under some simplified conditions. Assuming a setting with two classes $C_1$ and $C_2$, with a probability density functions $p_1(x)$ and $p_2(x)$, respectively. If $x_a$ is sampled from either of the distributions, it results in $p_a (x_a) = \pi p_1(x_a) + (1-\pi)p_2(x_a)$, with $0\leq \pi\leq 1$.
Setting $x=[x'_a \  x'_p \ x'_n]'$ with a probability density function $p(x)=p(x_a,x_p,x_n)$ assumed to be equal to
\begin{align}
    \begin{split}
        p(x) = & \text{ } p(x_a,x_p,x_n) \\
        = & \text{ } p_a(x_a)p_p(x_p|x_a)p_n(x_n|x_a) \\
        = & \text{ } \pi p_1(x_a)p_1(x_p)p_2(x_n) + (1-\pi)p_2(x_a)p_2(x_p)p_1(x_n)
    \end{split},
\end{align}
the triplet loss between $x_a$, $x_p$, $x_n$ can be described using the Euclidean distance function as ${\mathbb E}_{x \sim p} \mathcal{L}(x_a,x_p,x_n)$ with
\begin{equation}
        \mathcal{L}(x_a,x_p,x_n) = \max\left(0, \alpha + ||r(x_a)-r(x_p)||^2 - ||r(x_a)-r(x_n)||^2\right),
\end{equation}
where $r(x)$ is the learned representation of $x$.

In the presence of the assumed noise model in the sampling process of the triplets $(x_a,x_b,x_c)$, the probability density function becomes 
\begin{align}
    \begin{split}
        g(x) = & \text{ } \beta\gamma p_a(x_a)p_p(x_p)p_n(x_n) \\
        & + (1-\beta)\gamma p_a(x_a)p_n(x_p)p_n(x_n) \\
        & + \beta(1-\gamma) p_a(x_a)p_p(x_p)p_p(x_n) \\
        & + (1-\beta)(1-\gamma)p_a(x_a)p_n(x_p)p_p(x_n)
    \end{split}.
\end{align}

With this, the triplet loss becomes:
\begin{align}\small
    \begin{split}
        {\mathbb E}_{x \sim g} {\cal L}(x_a,x_p,x_n) =  & \text{ } \beta\gamma {\mathbb E}_{x \sim p} {\cal L}(x_a,x_p,x_n) \\
        & + (1-\beta)\gamma {\mathbb E}_{x \sim h_1} {\cal L}(x_a,x_p,x_n) \\
        & + \beta(1-\gamma){\mathbb E}_{x \sim h_2} {\cal L}(x_a,x_p,x_n) \\
        & + (1-\beta)(1-\gamma){\mathbb E}_{x \sim p} {\cal L}(x_a,x_n,x_p)
    \end{split},
\end{align}
with:
\begin{align}
    \begin{split}
        h_1(x_a,x_p,x_n) = \pi p_1(x_a)p_2(x_p)p_2(x_n) + (1-\pi)p_2(x_a)p_1(x_p)p_1(x_n)
    \end{split}
\end{align}
and:
\begin{align}
    \begin{split}
        h_2(x_a,x_p,x_n) = \pi p_1(x_a)p_1(x_p)p_1(x_n) + (1-\pi)p_2(x_a)p_2(x_p)p_2(x_n)
    \end{split}.
\end{align}

Noting that the expected value of the {\em gradient} of the loss ${\cal L}(x_a,x_p,x_n)$ is zero under $h_1$ and $h_2$ (since $x_p$ and $x_n$ are sampled from the same distribution and the loss is symmetric), the impact of those two cases in a gradient-based learning scheme is small.
The total loss is then: 
\begin{align}\small
    \begin{split}
        & \beta\gamma \max(0, \alpha + ||y_a-y_p||^2 - ||y_a-y_n||^2 ) + \\
        & (1-\beta)(1-\gamma) \max(0, \alpha + ||y_a-y_n||^2 - ||y_a-y_p||^2 )
    \end{split},
\end{align}
under the $p(x)$ probability density function, where $y = r(x)$.

Finally, this loss can be compacted to:
\begin{align}\small
    \begin{split}
 & \max\left(0, \beta\gamma(\alpha +  ||y_a-y_p||^2 - ||y_a-y_n||^2 )\right) + \\
 & \max\left(0, (1-\beta)(1-\gamma)( \alpha +  ||y_a-y_n||^2 - ||y_a-y_p||^2 )\right)
    \end{split}.
\end{align}

In section~\ref{sec:selflearn_application}, example applications of this methodology are presented for the tasks of electrocardiogram-based biometric identity verification and face identity verification. 

\section{Application Scenarios}\label{sec:selflearn_application}

The proposed method can be used to train models relying solely on unsupervised data. On classification tasks, the negative sample can be generated through the random selection of a sample in the dataset. Assuming a large number of balanced classes, errors in negative sample selection should be rare. For the positive samples, the function $f(x)$ that generates them based on an anchor can be a data augmentation procedure. This should be carefully adjusted to cover the expected intraclass noise and variability while retaining the information pertaining to the underlying image label, which can be difficult. 

Alternatively, when training with sequential data, the triplet generation can forgo the data augmentation procedures. In these situations, depending on the acquisition context or protocol, the temporal distance or proximity between data can be used to infer the identity of the subjects. A sample that is very close in time to the anchor can safely be used as $x_p$. Similarly, a sample that is sufficiently distant in time to the anchor can be assumed to belong to a different user, and thus used as $x_n$. Some knowledge of the domain and the acquisition settings can be used to adjust the distance between $x_a$, $x_p$, and $x_n$ to maximise $\beta$ and $\gamma$.

Both aforementioned alternatives (entirely unsupervised or using sequential data) were explored for the applications described below, through the experiments described in section~\ref{sec:selflearn_settings}.

\subsection{ECG identity verification}

Deep learning models have previously shown improved robustness to off-the-person noise and variability in electrocardiogram-based biometrics~\cite{Pinto2019b, Pinto2019Deep}. However, to train such models and match the performances reported for cleaner on-the-person signals, one would need large databases of off-the-person acquisitions, which are currently unavailable~\cite{Pinto2018}. In such circumstances, a pretrained network would often be the natural option in computer vision tasks. However, these too are currently nonexistent for unidimensional physiological signals such as the electrocardiogram.

The integration of ECG sensors in everyday objects, \eg~using the CardioWheel steering wheel cover~\cite{Lourenco2015} for shared vehicles or similar solutions for shared bicycles or scooters, enables the continuous acquisition of data from several subjects over long periods. This large amount of collected data could be used to train deeper and more sophisticated models. However, this data is commonly unlabelled, as the identity of the users at the moment of acquisition cannot be easily verified.

The proposed methodology for self-learning can be applied to train models for ECG-based identity verification using such data. As aforementioned, perturbations based on data augmentation procedures can be applied to the anchor to generate a positive sample. Thus, the four most successful data augmentation procedures proposed by Pinto~\etal~\cite{Pinto2019Deep} were implemented. For each triplet, one of these was randomly selected to generate a positive sample from the anchor:
\begin{itemize}
    \item \emph{Cropping}: a smaller contiguous segment is taken from the anchor sample and resampled to match the anchor's length, to simulate slower heart rates;
    \item \emph{Baseline Wander}: a periodic undulation, with a frequency near $1$ Hz, is added to the anchor segment to simulate breathing movement artefacts; 
    \item \emph{Gaussian Noise}: Gaussian noise is added to the anchor signal, simulating high-frequency distortions similar to the electromyogram (EMG) and powerline interference; 
    \item \emph{Random Permutation}: the anchor is divided into $N$ subsegments, which are shuffled to generate a different sample that simulates discontinuities or sensor faults.
\end{itemize}

\begin{figure*}
    \centering
    \includegraphics[width=0.8\linewidth]{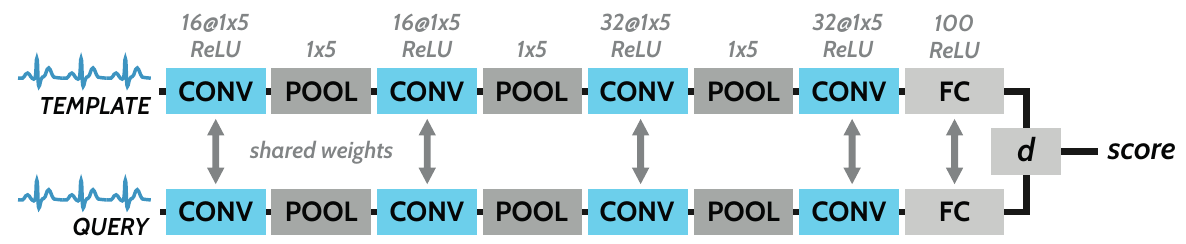}
    \caption{Architecture of the ECG identity verification model that was trained with the proposed methodology.}
    \label{fig:selflearn_model}
\end{figure*}

\begin{figure*}
    \centering
    \includegraphics[width=\linewidth]{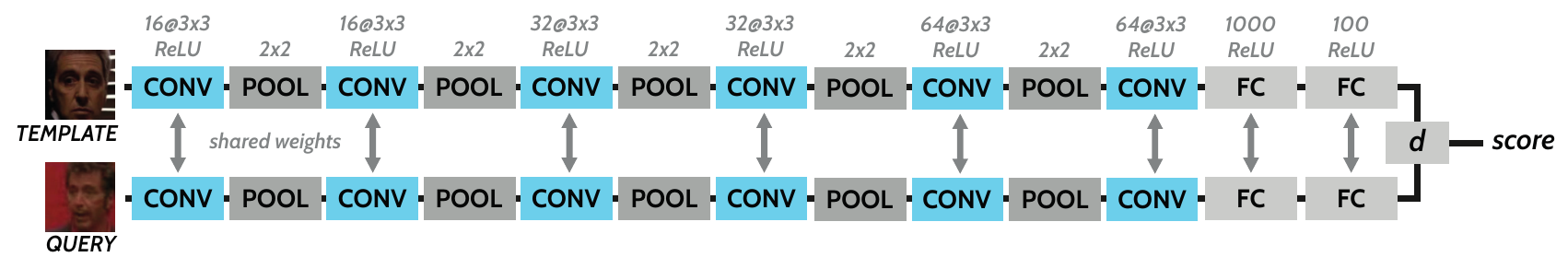}
    \caption{Architecture of the face identity verification model that was trained with the proposed methodology.}
    \label{fig:selflearn_model_face}
\end{figure*}

With continuous ECG recordings, it is possible to avoid errors in positive and negative sample selection. Having separate recordings for each person, positive samples are obtained through the selection of a segment of the anchor's recording. The negative sample is obtained from a different recording. In this case, there should be no errors in positive sample selection. Although there can be several recordings for the same person, errors in negative sample selection should be rare considering the large number of identities in the dataset and the balanced number of recordings per identity.

\subsection{Face identity verification} More face data are available now than ever before, especially from surveillance feeds or public videos shared on online social media platforms. However, as with ECG-based biometrics, the labelling of faces in acquired datasets is a tedious and lengthy task. Some researchers have taken advantage of online videos to build large datasets for face recognition, such as the YouTube Faces dataset from Wolf~\etal~\cite{Wolf2011}. However, these datasets are limited by the number of annotations available.

The proposed self-learning method can be used to train models for face verification without labelled data. In this case, common image data augmentation based on rotations, width and height shifts, and horizontal flips were used as the transformation function $f(x)$ that generates a positive sample $x_p$ based on an anchor $x_a$.

Having short videos, a random detected face from the same recording as the anchor can serve as a positive sample, while a negative sample can be drawn from a different recording. With some knowledge of the recordings, we minimise the probability of errors in positive and negative sample selection. Specifically, we know the YouTube Faces data consists of frames from short video recordings, with several people, with no more than one person per frame. Hence, although the short recordings lack much intrasubject trait variability, selecting triplets in the aforementioned way avoids errors in positive and negative sample selection.

\section{Experimental Setup}\label{sec:selflearn_settings}

\subsection{Data}
\subsubsection{ECG data}
The data used to train and evaluate the model is from the University of Toronto ECG Database (UofTDB)~\cite{Wahabi2014}. This database includes data from $1019$ subjects, acquired at $200$ Hz using dry metallic button electrodes, held by the subjects in contact with one finger of each hand. Each recording is $2-5$ minutes long, and each subject has recordings for up to five different postures (supine, tripod, exercise, standing, and sitting) on up to six sessions over a period of six months.

The data was divided for model training and evaluation as done by Pinto~\etal~\cite{Pinto2019b}. The last $100$ subjects (from subject $921$ to subject $1020$) were reserved for model training. The data from the remaining $918$ subjects were used for evaluation. One subject ($8$) was discarded for having too few data. From the $918$ subjects reserved for evaluation, the first $30$ seconds of the first recording were used for enrollment, while the remaining data were used for testing. This aimed to mimic a realistic context with scarce supervised data as expected in real ECG-based biometric applications.

\subsubsection{Face data} For face identity verification, data from the YouTube Faces database~\cite{Wolf2011} were used. This database contains frames from $3425$ videos of $1595$ subjects, sourced from YouTube. Each video is $48$ to $6070$ frames long, and there are up to six videos of each subject. This work used the aligned images provided on the database, which resulted from face detection, cropping, and alignment.

The first $150$ subjects (in alphabetical order) were used to build the dataset used in this work: the first $100$ subjects were reserved for training and validation, while the data from the remaining subjects were used for testing. Triplets were generated using this data subset, after resizing the images to $224\times224$, as detailed below in the experiments' description.

\subsection{Models}
The self-supervised training method proposed in this work was explored for ECG-based identity verification using an adapted version of the end-to-end network proposed by Pinto~\etal~\cite{Pinto2019b} (see Fig.~\ref{fig:selflearn_model}). The model receives two z-score normalised five-second raw ECG segments (a stored template and a query sample) and returns a measure of dissimilarity related to their identity.

The network is composed of four convolutional layers followed by a dense layer. A max-pooling layer (pooling size $1\times5$) follows each of the first three convolutional layers. The convolutional layers have $16$, $16$, $32$, and $32$ filters, respectively, with unit stride, without padding. The dense layer is composed of $100$ units. All convolutional and dense layers are followed by ReLU activation.

For face identity verification, the model is a simple convolutional neural network (see Fig.~\ref{fig:selflearn_model_face}), which receives two $224\times224$ RGB face images, normalised to $[0,1]$ intensities, and outputs a measure of their dissimilarity. It is composed of six convolutional layers interposed with five max-pooling layers (pooling size $2\times2$) and followed by two dense layers. The convolutional layers have $16$, $16$, $32$, $32$, $64$, and $64$ filters, respectively, with size $3\times3$, unit stride, without padding. The dense layers are composed of $1000$ and $100$ units, respectively. All convolutional and dense layers are followed by ReLU activation. 

Both models were trained using the Adam optimiser, with an initial learning rate of $0.0001$. As in \cite{Pinto2019b}, the Euclidean distance was used as distance measure $d$ during training, while for identity verification this was replaced by the normalised Euclidean distance for scores in $[0,1]$. The triplet loss margin was set as $\alpha=1.0$. A maximum of $200$ epochs was given, with batches of $12$ triplets, along with early stopping with a patience of $10$ epochs. Dropout was used before each dense layer, with rates of $0.5$ and $0.2$ for the ECG and the face models, respectively. L2 regularisation ($\lambda=0.01$) was used for the convolutional layers in both models.

\subsection{Evaluation metrics}
The evaluation metrics used are the False Acceptance Rate (FAR), the False Rejection Rate (FRR), the Equal Error Rate (EER), and the Receiver-Operating Characteristic (ROC) curve~\cite{Pinto2018}. The FAR measures the rate at which impostors meet a given acceptance threshold and are falsely granted access. The FRR measures the rate at which genuine users are incorrectly denied access due to their scores not meeting the given threshold. The EER corresponds to the error at the operation threshold where FAR and FRR have equal values. The ROC curve plots the values of $1-$FRR \emph{versus} FAR for the possible range of threshold values.

\subsection{Experiments' description}\label{subsec:selflearn_experiments}

\subsubsection{Without supervision}\label{subsubsec:selflearn_withoutsup}
In this experiment, the models were trained with triplets whose negative samples are drawn randomly from the entire respective dataset. The positive samples are created through the application of data augmentation procedure to the respective anchor samples. To train the ECG identity verification model, $100~000$ triplets were generated for the training, of which $10\%$ were used for validation during training, and $10~000$ triplets were generated for evaluation. For the face model, $10~000$ triplets were generated for the training, of which $20\%$ were used for validation, and $5000$ triplets were generated for testing.

Naturally, depending on the dataset used for training, the probability of error in the random selection of a negative sample will vary. In datasets with fewer classes, the probability of randomly selecting a negative sample whose class matches that of the anchor is greater than in datasets with more classes. Hence, the aforedescribed experiment with the ECG identity verification model was repeated, but giving the selection of a negative sample a probability $p_e = 1-\gamma$ of returning a sample from the anchor's identity. This probability of error was linked to a simulated number of subjects $N_s$, with $p_e=1/N_s$ and $N_s=\{2,5,10,20,50,100,200,500,1000\}$. This enabled the assessment of how a balanced dataset with fewer classes could impact the training process and the effect on the final identity verification performance.

\subsubsection{Using recordings}

This experiment used the recordings of the UofTDB database and the video recordings of YouTube Faces as a way to infer the identity of the samples through the temporal proximity between them, using prior knowledge to minimise triplet generation errors. Here, the positive sample is drawn from the same recording as the anchor, while the negative sample is drawn from a different recording. As each subject can have several recordings, there is an error associated with the selection of the negative sample, which can accidentally be selected from a different recording of the same subject. The number of generated ECG and face triplets used for training, validation, and testing, was the same as mentioned in~\ref{subsubsec:selflearn_withoutsup}.

When training the network with longer recordings spanning several users, as described in section~\ref{sec:selflearn_methodology}, the possible errors are different. Although the positive sample is selected in the temporal vicinity of the anchor, it can belong to a different identity. The negative sample, despite the distance from the anchor, can accidentally belong to the same user as the anchor. Hence, additional experiments were conducted where the positive and negative sample selection processes failed purposely with probability $p=\{0.05,0.1,0.2,0.3,0.5,0.7\}$, to assess the effect of such errors in the final model performance.

\section{Results and Discussion}

\subsection{ECG identity verification}
The baseline results correspond to the identity verification model trained with supervised data. The equal error rate of $12.56\%$ represents a small improvement over the corresponding result reported in~\cite{Pinto2019b}. Considering the evaluation data and conditions were the same, this method also offered significantly better results than the state-of-the-art methods implemented and tested in~\cite{Pinto2019b}: the Autoencoder-based solution proposed by Eduardo~\etal~\cite{Eduardo2017}, the AC/LDA method proposed by Agrafioti~\etal~\cite{Agrafioti2012}, and the DCT approach proposed by Pinto~\etal~\cite{Pinto2019Deep, Pinto2017}.

\begin{figure}[t!]
    \centering
    \includegraphics[width=0.65\linewidth]{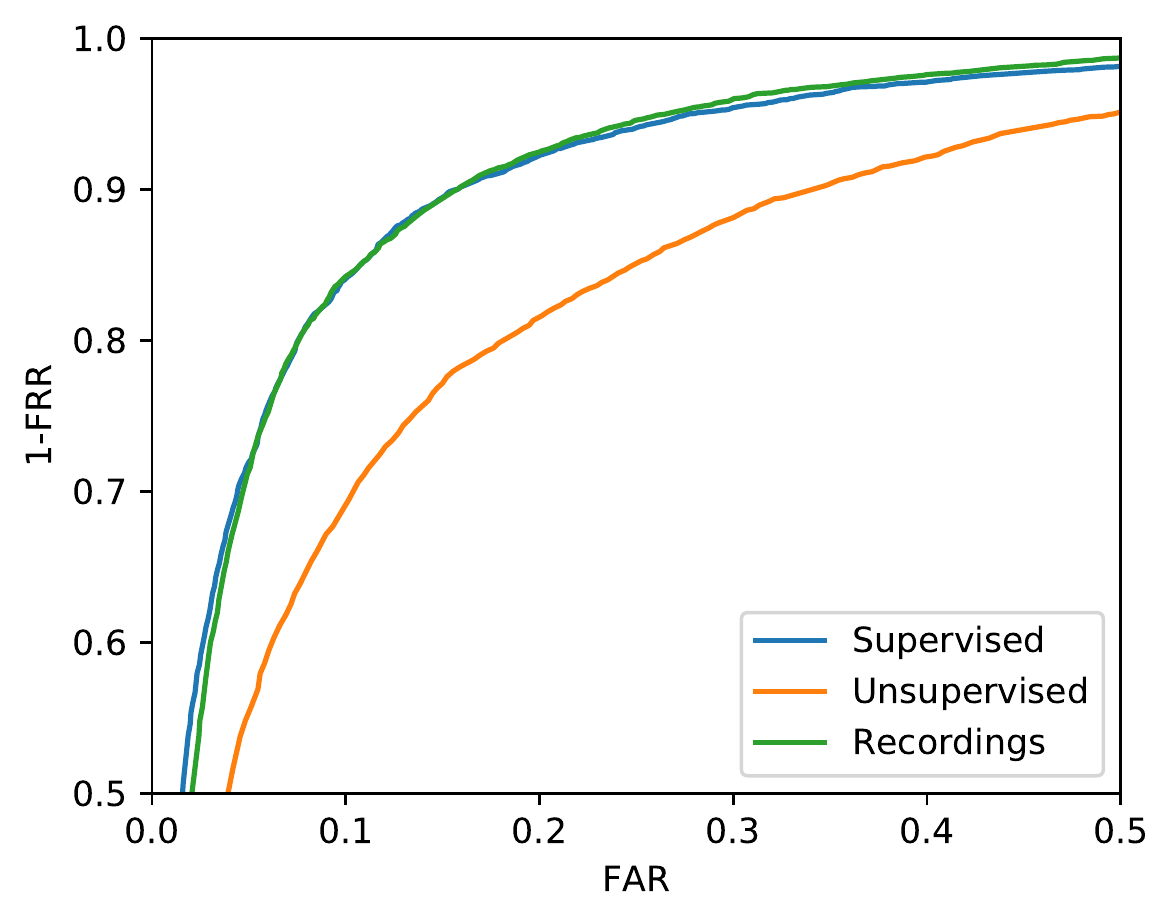}
    \caption{Comparison of the Receiver-Operating Characteristic curves on ECG identity verification for supervised training, unsupervised training, and recording-based supervision.}
    \label{fig:selflearn_comparison_roc}
\end{figure}

The performance results of the model trained with the two unsupervised training approaches are presented in Fig.~\ref{fig:selflearn_comparison_roc}, in comparison with the baseline results. The Equal Error Rate values were $12.56\%$, $19.19\%$, and $12.70\%$, for supervised, entirely unsupervised, and recording-based training, respectively. The difference between the performance with entirely unsupervised training and the performance with recording-based training denotes the data augmentation procedures have not been able to completely mimic the variability of the signals, and could perhaps be improved using optimised data augmentation~\cite{Cubuk2019, Lim2019}. Despite the worse performance attained with the entirely unsupervised training approach, all of these methods offered better performance than the handcrafted methods evaluated in the same settings in~\cite{Pinto2019b}, among which the best result was $21.82\%$ EER.

\subsection{Face identity verification}

\begin{figure}[t!]
    \centering
    \includegraphics[width=0.65\linewidth]{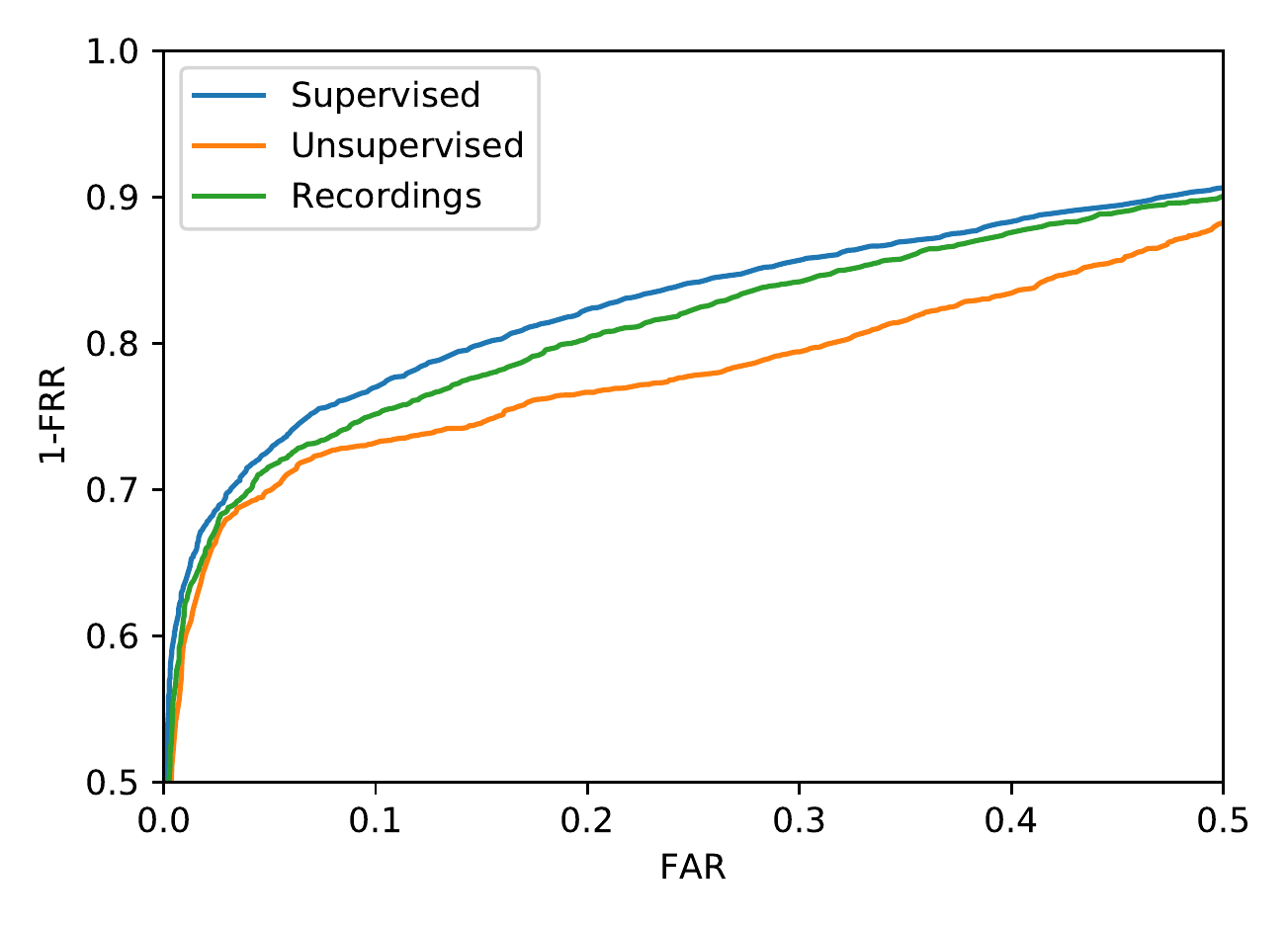}
    \caption{Comparison of the Receiver-Operating Characteristic curves on face verification for supervised training, unsupervised training, and recording-based supervision.}
    \label{fig:selflearn_comparison_roc_faces}
\end{figure}

As with the ECG-based identity verification task, the model trained with supervised data was used as a baseline for comparison of results in face identity verification. The performance offered by the baseline was $18.45\%$ EER. This is considerably higher than the state-of-the-art, which is explained by the relative simplicity of the implemented model and the relatively small dataset used. Nevertheless, the goal of this work was not to overcome or match the state-of-the-art in face verification but to illustrate how the proposed self-learning methodology can be applied to face biometrics with small performance losses relative to a supervised baseline in similar conditions.

The results with the proposed methodology are illustrated in Fig.~\ref{fig:selflearn_comparison_roc_faces}. When using entirely unsupervised data, the proposed method offered $22.81\%$ EER, a $4.36\%$ increase relative to the use of supervised data. With recording-based triplet generation, the model offered $19.77\%$ EER, a $1.32\%$ increase. These small performance losses when forgoing labels during training show that the model can learn without supervision using the stochastic triplet loss, as verified above for ECG biometrics. Besides these two applications, one should expect the proposed self-learning methodology to be successfully applied to similar problems.

\subsection{Stress experiments}

\begin{figure}[t!]
    \centering
    \includegraphics[width=0.65\linewidth]{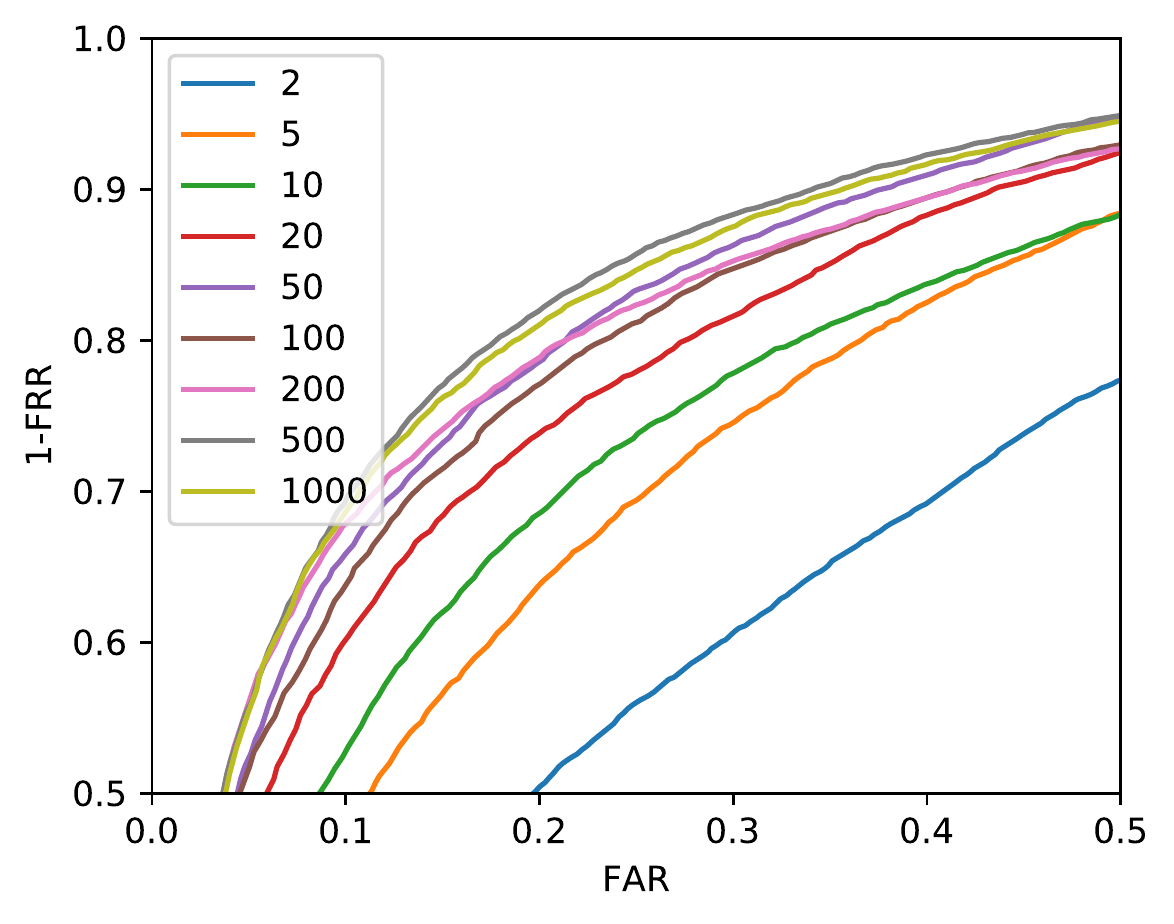}
    \caption{Receiver-Operating Characteristic curve for negative selection error based on the number of database subjects.}
    \label{fig:selflearn_subjects}
\end{figure}

As discussed in subsection~\ref{subsec:selflearn_experiments}, the success of the unsupervised triplet generation technique depends on the number of identities (classes) in the database. Hence, an experiment was performed on ECG identity verification to simulate the variation of the number of identities on the dataset, inducing errors in the negative sample selection with the respective probability. The results (see Fig.~\ref{fig:selflearn_subjects}) show that, although the performance worsens with fewer subjects, the errors have a very small effect for datasets with more than $20$ subjects. In fact, with $50$ subjects or more, the performance results stabilised around $20\%$ EER. Hence, a dataset with $50$ classes should be enough to adequately apply this method with better performance than handcrafted state-of-the-art approaches.

\begin{figure}[t!]
    \centering
    \includegraphics[width=0.65\linewidth]{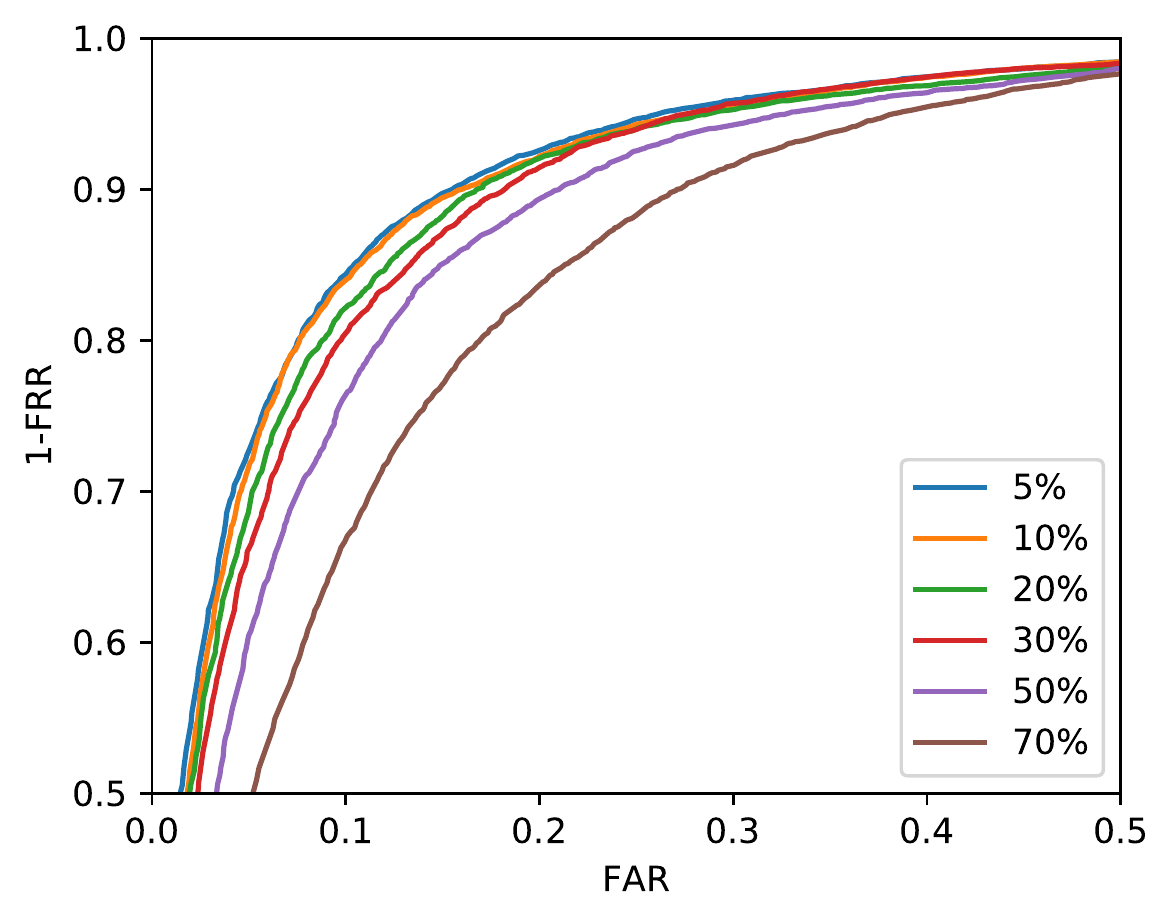}
    \caption{Receiver-Operating Characteristic curves for varying positive sample selection error probability.}
    \label{fig:selflearn_prob_pos}
\end{figure}

\begin{figure}[t!]
    \centering
    \includegraphics[width=0.65\linewidth]{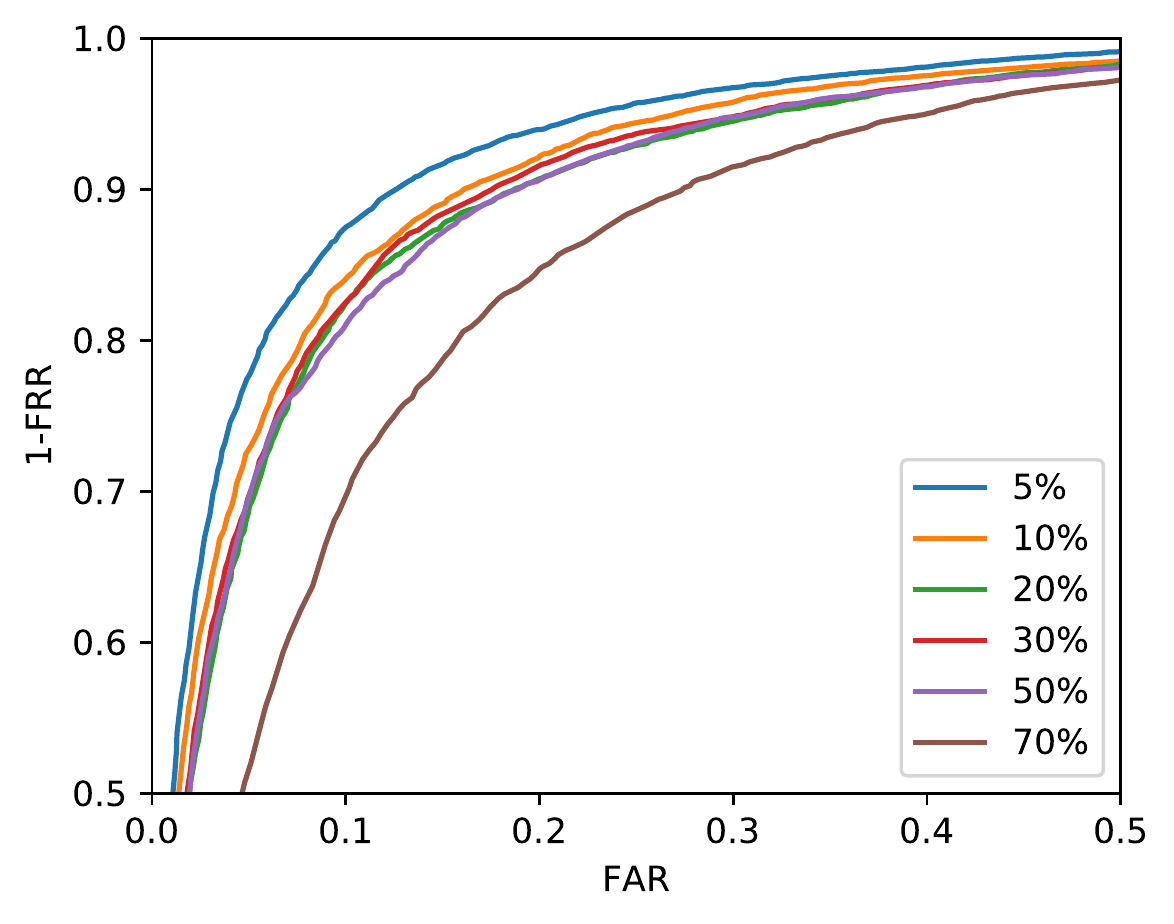}
    \caption{Receiver-Operating Characteristic curves for varying negative sample selection error probability.}
    \label{fig:selflearn_prob_neg}
\end{figure}

\begin{figure}[!t]
    \centering
    \includegraphics[width=0.65\linewidth]{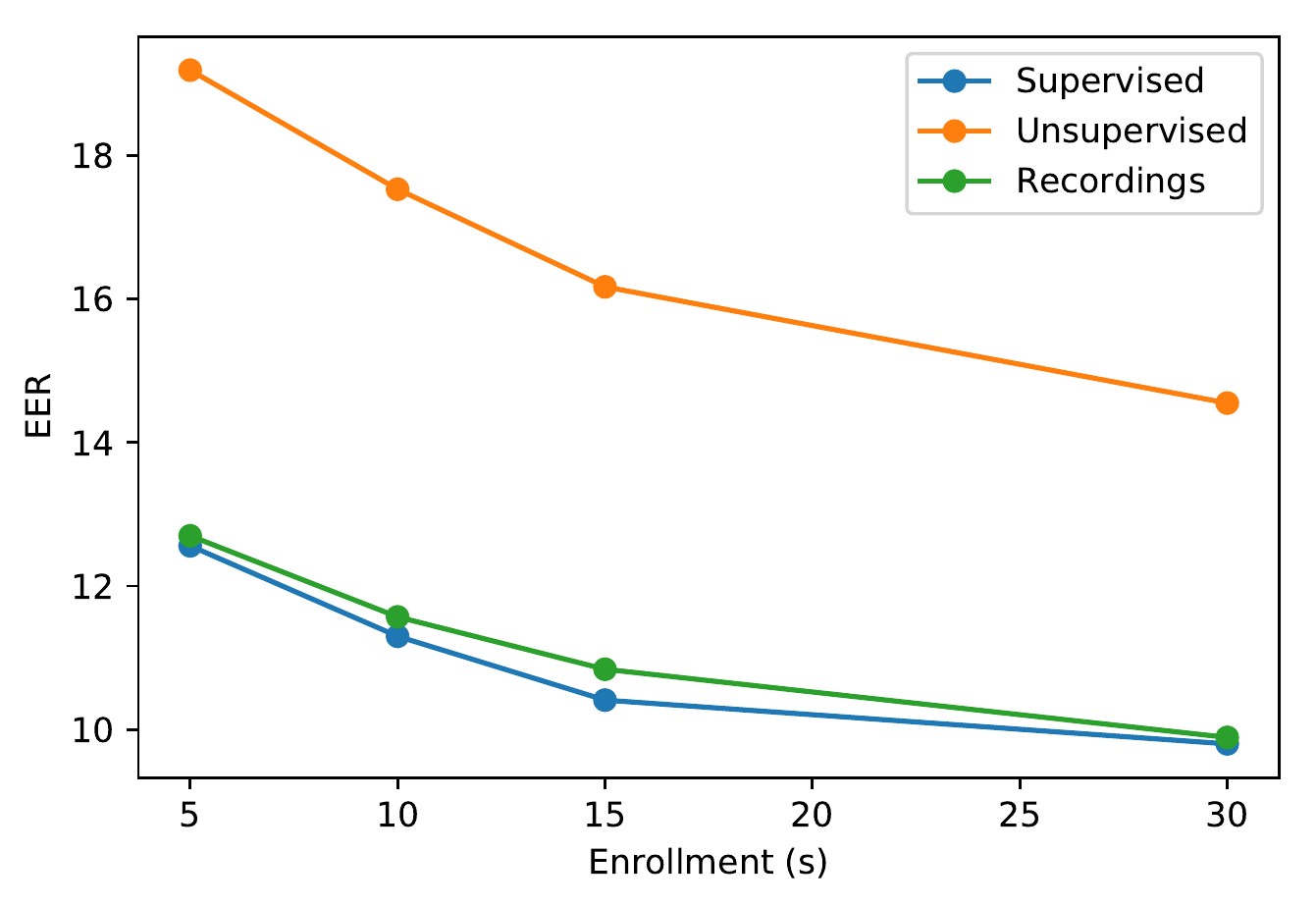}
    \caption{Equal Error Rate (EER) results when using more enrollment data from each subject.}
    \label{fig:selflearn_eer_enroll}
\end{figure}

For the training based on temporal proximity between samples, both the selection of positive samples and the selection of negative samples may fail. Hence, enforcing a probability of each error in the recording-based training experiments allows the study of the impact of such errors on the model's performance. The increase of either positive or negative sample selection error probabilities leads to a decrease in performance (see Fig.~\ref{fig:selflearn_prob_pos} and Fig.~\ref{fig:selflearn_prob_neg}). However, that decrease is relatively small unless the probabilities of error are over $50\%$. This means that some knowledge of the typical usage times and patterns during data acquisition would be enough to adjust the process of positive and negative sample selection and ensure the best results.

Results could be further improved using more enrollment data (see Fig.~\ref{fig:selflearn_eer_enroll}). As studied by Pinto~\etal~\cite{Pinto2019b}, instead of the simple one-vs-one comparisons performed in the aforedescribed experiments, which correspond to five-second enrollments, the query template can be compared with each of several enrollment templates from each person, and only the minimum score is considered. ECG identity verification performance with the proposed unsupervised and recording-based training approaches reached $14.55\%$ and $9.89\%$, respectively, when using thirty-second enrollments.

\section{Summary and Conclusions}

This work proposed a novel formulation of the triplet loss for self-supervised learning with unlabelled data. This method considers the uncertainty associated with the triplet generation in unsupervised settings and maximises the probability of success using prior knowledge.

In harmony with the goals of this thesis, the proposed methodology was applied to the task of ECG-based biometric identity verification, using transformations based on data augmentation or the temporal proximity between samples to generate valid triplets. The method offered better performance than handcrafted state-of-the-art methods, especially when using temporal proximity between samples, with performance results similar to supervised training.

This pattern was also confirmed on the task of unconstrained face identity verification. Training with entirely unsupervised data using the proposed triplet loss formulation resulted in just a small performance loss when compared with the use of supervised data. When generating triplets based on video streams, this loss was considerably smaller. 

It should be noted that the proposed method can be influenced by imbalanced classes and errors in the unsupervised triplet generation. However, the results of the stress experiments show its robustness is sufficient to avoid considerable impacts on performance in most cases. Thus, this method would, according to the presented results, be a viable training option in multiclass classification problems where only unlabelled data are available, especially with sequential data.

\part{Epilogue}\label{part:epilogue}
\chapter{Summary and Conclusions}\label{ch:summary}

The doctoral work presented throughout this thesis had the main goal of advancing the integration of biometric recognition and wellbeing monitoring solutions, especially focused on the case of intelligent vehicles. This goal was carefully targeted through two specific scenarios. The first was personalised wellbeing monitoring systems using biometrics, mainly focused on advancing ECG and face biometric solutions for improved driver assistance systems. The second was occupant monitoring for autonomous shared vehicles, through the development of solutions for efficient and robust group-based monitoring of passengers' wellbeing inside fully autonomous vehicles.

Always with this major goal in mind, several novel and meaningful contributions were achieved, as introduced in Chapter~\ref{ch:intro} and described in the various chapters of this document. The main contributions, resulting completely or mostly from the work of the author of this thesis, were:
\begin{itemize}
    \item The most comprehensive state-of-the-art survey on ECG biometrics to date. This work encompassed one hundred and twenty-five literature methods, a thorough description of the fundamentals behind the ECG as a biometric characteristic, and a critical overview of the evolution of this topic, including extensive considerations on the current research challenges and future opportunities. At the time of writing this thesis, the resulting survey article had been cited in the literature over one hundred and fifty times\footnote{João R. Pinto's Google Scholar. Available on:~\url{https://scholar.google.com/citations?user=hhF9Q8kAAAAJ}.};

    \item The first truly end-to-end approach for ECG-based biometrics, proposed for both identification and identity verification tasks, including the development of tailored data augmentation strategies and studies on fine-tuning and transfer learning. Meaningful and significant performance improvements were achieved in challenging and realistic evaluation scenarios using large \emph{off-the-person} databases. At the time of writing this thesis, the resulting book chapter and international conference article had been cited in the literature over forty times;

    \item A pioneering study on the relevance of ECG waveforms for the biometric identification task through explainability. End-to-end deep networks trained for biometric identification and analysed using multiple explainability tools offered new insights on the importance of the different parts of the ECG signal, especially the QRS complex, when distinguishing identities in diverse scenarios;

    \item A multimodal approach for audiovisual recognition of emotion valence in groups of individuals. The proposed approach, based on adapted state-of-the-art sound and video recognition modules, presented promising results in regards to both accuracy and efficiency in the recognition of emotional states of groups of people;

    \item A cascade strategy to streamline multimodal audiovisual activity recognition in time-sensitive scenarios. When evaluated for multiclass activity recognition and violence detection, including inside real vehicles, the proposed approach was able to offer improved accuracy with up to $66\%$ reduction in runtime; 

    \item The first approach to achieve template security in end-to-end biometric models. Aiming to combine the benefits of end-to-end deep learning with the security of template cancelability and unlinkability, the proposed Secure Triplet Loss was able to outperform the state-of-the-art alternatives and offer minimal performance gaps without requiring separate encryption processes or protection schemes. At the time of writing this thesis, the resulting journal article and international conference article had been cited in the literature fifteen times;

    \item An adaptation of the triplet loss to allow for fully unsupervised learning. Taking advantage of multiclass balance and the natural structure of sequential data, the proposed approach was able to considerably close the gap to supervised performance levels, even in the more challenging evaluation scenarios.
\end{itemize}

The secondary contributions of this doctoral work, resulting from collaborations within the scope of this thesis which benefitted partially from the work of the author, were:
\begin{itemize}
    \item The first study on long-term performance evolution of multiple state-of-the-art ECG biometric methodologies. This study highlighted the problem of intrasubject variability in ECG biometrics, even over relatively short time periods, and proposed multiple template/model update strategies to mitigate its negative effects on identification performance;

    \item An approach for recovering the full set of twelve standard ECG leads relying only on short single-lead blindly-segmented recordings. The proposed approach has offered promising results in a considerably more challenging evaluation scenario than those found in the literature, and paves the way towards robust and efficient methods for retrieving missing leads in more comfortable acquisition setups;

    \item Two novel strategies to reduce the performance gap in masked face recognition. Among the proposed approaches, the one based on multi-task contrastive learning outperformed the alternative methods and illustrated the benefits of promoting the similarity between latent masked and unmasked face image representations. At the time of writing this thesis, the resulting three international conference articles had been cited in the literature over fifty times;

    \item A pioneering study on interpretability for face biometrics, through the presentation attack detection task. This exploratory experiment illustrated how interpretability tools could be used to achieve a deeper understanding of the behaviour of biometric models and motivated their use in the next generation of biometric evaluation strategies.
\end{itemize}

Through these contributions, this doctoral project touched on several important topics related to each individual research area and the thesis theme as a whole. In ECG biometrics, the work addressed the topics of end-to-end models, transfer learning, data augmentation, long-term performance, template update, interpretability, and interlead conversion. In face biometrics, the topics of masked face recognition, presentation attack detection, and interpretability were covered. In wellbeing monitoring, this work focused on multimodal fusion, group emotion recognition, violence detection, in-vehicle scenarios, and optimisation/efficiency. Additionally, the broader topics of biometric template security and self-supervised learning were also addressed. 

Considering the achievements of this work, one can conclude that, although the ideal of truly personalised wellbeing monitoring is yet to be achieved, meaningful and valuable strides have been successfully taken to reach it. As such, a strong framework is now built to support future work towards the central goal of tightly integrating biometric recognition and wellbeing monitoring in a multimodal, seamless, continuous, and realistic way.

This conclusion is supported by the reception of this research within the scientific community. The work described in this thesis has directly resulted in twenty-four scientific publications, including five articles in peer-reviewed journals and eleven articles in the proceedings of international conferences. This increases to thirty-eight total publications if one also considers other minor contributions to other topics in biometrics, computer vision, and pattern recognition, which have not been addressed in this document. These had been welcomed by the scientific community with over three hundred citations by the time this thesis was written.

This doctoral work has been published in multiple major journals and conferences in the international biometrics community. These include the \emph{IEEE Transactions in Biometrics, Behavior and Identity Science}, \emph{IET Biometrics}, the \emph{International Joint Conference on Biometrics (IJCB)}, the \emph{International Conference on Biometrics: Theory, Applications and Systems (BTAS)}, and the \emph{International Conference of the Biometrics Special Interest Group (BIOSIG)}. Additionally, part of the work has also been published in reputed venues related to general machine learning, computer vision, or pattern recognition research, such as \emph{IEEE Access} or the \emph{International Joint Conference on Neural Networks (IJCNN)}. Part of the results of this work also contributed to the AUTOMOTIVE, Easy Ride, and Aurora research projects.

In recognition of the value of the achieved contributions, this doctoral research and its author have also been the recipients of multiple awards. These are the \emph{EAB Max Snijder Award} at the \emph{2022 European Biometrics Awards} organised by the European Association for Biometrics (EAB), the \emph{Computers Journal Best Paper Award} at the \emph{2020 International Workshop on Biometrics and Forensics}, the \emph{Best Session Paper Award} at the \emph{2020 IEEE International Conference on Image Processing, Applications and Systems (IPAS)}, and the \emph{Jury's Best Presentation Award} at the \emph{2021 NIS Workshop} organised by INESC TEC.

\chapter{Future Work Considerations}\label{ch:future}

Despite the results achieved in the doctoral work described throughout this thesis, plenty is yet to be done to achieve the full symbiotic integration of biometric recognition and wellbeing monitoring. The successful integration of these two tasks in real scenarios is a challenging endeavour in itself. Nevertheless, plenty of opportunities are yet to be explored in the topics of ECG biometrics, face biometrics, and wellbeing monitoring focused on this doctoral work, which would be essential to achieve our major objective.

When considering the current state of ECG biometrics, it is hard to disagree with the notion that data is the main problem to be tackled. Many would argue that face biometrics is more evolved than ECG biometrics because the ECG is more deeply affected by noise and variability. But this is misleading, since the face suffers (and heavily so in truly unconstrained scenarios) from most of the same factors that affect the ECG, including emotions, exercise, drowsiness, and medical conditions. Even the loss of information due to heavy noise or sensor contact losses in off-the-person ECG is analogous to common occlusions that can deeply limit the available information in unconstrained face images. Face biometrics is more developed thanks to the unprecedented magnitude of data available to train increasingly sophisticated and robust models. The number of subjects, the unconstrained nature of the data, the diversity of scenarios, the variety of acquisition sessions, and the comprehensiveness of current benchmark data in ECG biometrics pale in comparison to what can be found for face biometrics. 

As such, researchers should dedicate special efforts to creating larger (and better) datasets for ECG biometrics. Increased number of identities, longer recordings, more sessions across wider periods of time, and more realistic off-the-person scenarios are only some of the aspects that need to be verified by new datasets. Since the ECG could likely serve society better as part of multimodal solutions, new datasets could focus on the simultaneous (and continuous) acquisition of other traits alongside the ECG, especially face video. While such datasets remain unavailable, it would be interesting to find new ways to mitigate the effects of data scarcity. This includes the development of sophisticated pretrained models that could be fine-tuned to multiple tasks, the study of learnable 1D to 2D transformations to take advantage of image pretrained models, and the development of tailored solutions for unsupervised and self-supervised learning. At last, it is also important to standardise the way performance is evaluated in ECG biometrics, through the definition of complete and realistic benchmark datasets to assess and compare the performance of state-of-the-art methods.

Conversely, in face biometrics, data is plenty and models are considerably more accurate. However, as we could witness from the results of this doctoral research, there is still much to do regarding the robustness of face recognition in realistic scenarios. The advent of face masks revealed the feeble nature of current state-of-the-art approaches to unexpectedly drastic scenarios and reignited the old topic of occlusions in face biometrics. In fact, after over two years of heavy and dedicated research in masked face recognition, there is still a considerable performance gap \emph{vs}. unmasked scenarios. Masks may soon leave our society, but the possibility for such severe occlusions to be witnessed once again (in the shape of masks or any other object) is enough to warrant further research. As such, the successful example of multi-task contrastive learning should be followed with the development of more sophisticated multi-objective schemes for learning to ignore masked face regions. Moreover, the creation of larger datasets of real masked face images and videos is also a key step towards closing the performance gap in masked face recognition.

Additionally, given that face biometrics is a much more developed topic than ECG biometrics, interpretability is also a much greater opportunity in this topic. Face recognition solutions have permeated countless aspects of our society, including deeply sensitive applications such as border control or urban surveillance. This situation starts to raise doubts about the trustworthiness of such algorithms. Interpretability is essential to offer the deeper level of understanding needed to avoid biases, lead models away from undesirable behaviours, and ultimately make the general public more comfortable with the use of such technologies. As such, it is important that more attention is devoted to the topic of interpretability in biometrics (especially face biometrics), through the redefinition of evaluation standards and the development of approaches with interpretability and explainability as one of the main goals (and not just an afterthought).

On the topic of wellbeing monitoring, two major problems are evident from the results of the work conducted during this doctoral project. The first is performance in specific scenarios such as in-vehicle monitoring. Current models present promising results in general purpose scenarios, for which data has become plenty through online sourcing, but are still relatively weak for recognition in specific scenarios. These, naturally, should not be overlooked and researchers should dedicate efforts to better take advantage of generalistic models for specific scenarios, \emph{e.g.}, through more sophisticated methods for transfer learning and domain adaptation. The second problem is subject dependency. This is a well-documented problem in the literature and, while not as evident in the group-level emotion recognition work presented in this thesis, has been encountered and discussed frequently in emotion and drowsiness recognition works linked to this doctoral work and the AUTOMOTIVE project. It is troubling how performance can be affected when wellbeing monitoring models are applied to data from individuals they have never seen before. If wellbeing monitoring is to be applied in real scenarios, this problem should be tackled head-on through dedicated robustness studies and the reformulation of evaluation setups to ensure realistic subject-independent results.

After the aforementioned individual topics are addressed, the integration of biometrics and wellbeing, combined with the integration of multiple data sources, raises new problems in terms of efficiency and data security. Since these solutions would be deployed to edge scenarios, it is important to consider performance costs and develop models with efficiency as one of the main priorities. Multi-task and multimodal integrated solutions, minimising separate processing of information, should be chosen as frequently as possible. At last, since such solutions would deal with highly intimate data carrying identity and health information, it is paramount to continue the efforts on learnable template security to ensure all data is stored in a thoroughly protected way.

\part{Appendices}\label{part:appendices}
\appendix
\chapter{ECG Biometrics Literature Methods}\label{ch:appendix_ecg}

This appendix presents an overview of the methods proposed for ECG-based biometrics in the literature, from the beginning of the topic in 1999 till today, in Table~\ref{tab:ecg_unimodal_works}. This table includes a straightforward description of the most important aspects of ECG biometric methodologies: the denoising methods, the signal preparation pipeline, the feature extraction and dimensionality reduction schemes, and the decision models/strategies. Moreover, it includes the performance reported in the respective publications, alongside information on the evaluation scenario corresponding to such results: the dataset, the number of subjects, and whether the data has been acquired in off-the-person scenarios.

This survey of state-of-the-art approaches began with the research work in~\cite{Pinto2017b}, initially covering sixty-five literature methods. It was later expanded during this doctoral work to cover ninety-three publications in~\cite{Pinto2018}, which has since been cited nearly one hundred and fifty times by the scientific community. In this thesis, it presently covers a total of one hundred and twenty-five state-of-the-art methods for ECG biometrics.

This unprecedentedly comprehensive collection of information on ECG biometric methodologies aimed mainly to empower the easy pinpointing of open challenges and research opportunities. Hence, this work, as published before in~\cite{Pinto2018}, is thought to have helped accelerate the development of novel solutions in ECG biometrics and is expected to continue doing so in the near future. The information gathered in this appendix has also been used to thoroughly describe the evolution and current landscape of ECG biometrics in Chapter~\ref{ch:ecgprior}.

\afterpage{
\begin{landscape}
{\begin{spacing}{0.8}
\renewcommand{\arraystretch}{1.5}
\begin{longtable}{>{\raggedright}p{1.8cm} >{\centering}p{0.6cm} >{\raggedright}p{2.5cm} >{\raggedright}p{3cm} >{\raggedright}p{3.5cm} >{\raggedright}p{2.5cm} >{\raggedright}p{2.0cm} >{\centering}p{0.6cm} >{\centering}p{0.5cm} >{\raggedright}p{0.7cm} >{\raggedright}p{0.8cm}}
\caption[Summary of the surveyed state-of-the-art unimodal methods proposed for ECG biometrics.]{Summary of the surveyed state-of-the-art unimodal methods proposed for ECG biometrics (ordered by year of publication and first author name, DR -- Dimensionality Reduction, NS -- Number of Subjects, OP -- Off-the-Person).}
\label{tab:ecg_unimodal_works}\\
\hline
\textbf{Author} & \textbf{Year} & \textbf{Denoising} & \textbf{Preparation} & \textbf{Features/DR} & \textbf{Decision} & \textbf{Dataset} & \textbf{NS} & \textbf{OP} & \multicolumn{2}{l}{\textbf{Results}} \tabularnewline\hline
\endfirsthead
\hline
\textbf{Author} & \textbf{Year} & \textbf{Denoising} & \textbf{Preparation} & \textbf{Features/DR} & \textbf{Decision} & \textbf{Dataset} & \textbf{NS} & \textbf{OP} & \multicolumn{2}{l}{\textbf{Results}} \tabularnewline\hline
\endhead

Biel{ }\etal{ }\cite{Biel1999,Biel2001} & 1999 2001 & - & Siemens Megacart ECG Processing & 10 Lead I fiducial features / PCA & SIMCA & Private & 20 & No & IDR & 100\% \tabularnewline\hline

Kyoso{ }\etal{ }\cite{Kyoso2000} & 2000 & HPF 0.06 Hz + LPF 60 Hz + NF 50 Hz & Beat segment. and fiducial detection & PQ and QT times & Mahalanobis distance + LDA & Private & 3 & No & IDR & 99.5\%\tabularnewline\hline

Kyoso{ }\etal{ }\cite{Kyoso2001, Kyoso2001b} & 2001 & HPF 0.06 Hz + LPF 60 Hz + NF 50 Hz & Second-order derivative fiducial detect. & QRS duration and QT time & Mahalanobis distance + LDA & Private & 9 & No & IDR & 94.2\% \tabularnewline\hline

Shen{ }\etal{ }\cite{Shen2002} & 2002 & - & - & RQ, RS, ST, QS time, QT time, RS slope, QRS area & Correlation + DBNN & MIT NSR & 20 & No & IDR & 100\% \tabularnewline\hline

Palaniappan{ }\etal{ }\cite{Palaniappan2004} & 2004 & LPF 30 Hz & Adapted Pan-Tompkins fiducial detect. & R, QR, RS, QRS width, R-R, beat form factor & MLP; SFA & MIT NSR & 10 & No & IDR:\\MLP\\SFA & \hspace{0.001cm} \\96.2\%\\83.6\% \tabularnewline\hline

Israel{ }\etal{ }\cite{Israel2005} & 2005 & BPF 2--40 Hz & R detection, heartbeat segmentation and alignment by R-peaks & RQ, RS, RP, RL, RP', RT, RS', RT’, P and T widths, ST, PQ, PT, LQ, ST' / LDA & Contingency matrix majority voting & Private & 49 & No & IDR:\\Anxty.\\Norm. & \hspace{0.001cm}\\97\%\\98\% \tabularnewline\hline

Saechia{ }\etal{ }\cite{Saechia2005} & 2005 & - & PQRST heart rate normaliz., P, QRS, and T segmentation & Fourier transform of PQRST (whole), P, QRS, and T & Neural Networks & - & - & No & FRR:\\Whole\\Apart & \hspace{0.001cm}\\17.1\%\\2.85\% \tabularnewline\hline

\citet{Plataniotis2006} & 2006 & BPF 0.5--40 Hz & Fixed-length window blind segmentation & Autocorrelation coefficients / DCT & Norm. Euclidean dist. + Gaussian LLR & PTB & 14 & No & IDR\\FAR & 100\%\\0.02\% \tabularnewline\hline

Zhang{ }\etal{ }\cite{Zhang2006} & 2006 & - & R-peak and QRS detection, heartbeat segmentation & Amplitudes, durations, intervals, levels, \& areas / PCA & Bayes-minimum-error-rate & Private (leads I, II, V1, and V2) & 502 & No & IDR\\L.I\\L.II\\L.V1\\L.V2 &  \hspace{0.001cm}\\85.3\%\\92.0\%\\95.2\%\\97.4\% \tabularnewline\hline

Molina{ }\etal{ }\cite{Molina2007} & 2007 & Savitzky-Golay & Trahanias R detection, R-R segmentation & R-R segments & DTW path + kNN & Private & 10 & Yes & EER & 2\%\tabularnewline\hline

\citet{Wuebbeler2007} & 2007 & Moving Median + LPF 75 Hz & R-peak detection by thresholding & 2D QRS (combination of leads I, II, and III) & Temporal derivatives dist. + kNN & PTB & 74 & No & IDR\\EER & 98.1\%\\2.8\% \tabularnewline\hline

Agrafioti{ }\etal{ }\cite{Agrafioti2008} & 2008 & BPF 1--40 Hz & Fixed-length window blind segmentation & Normalized autocorrelation / DCT or LDA & Correlation + kNN & PTB + MIT NSR & 27 & No & IDR:\\DCT\\LDA & \hspace{0.001cm}\\96.3\%\\100\% \tabularnewline\hline

Chan{ }\etal{ }\cite{Chan2008} & 2008 & NF 60 Hz & Fid. detection with backward diff., alignment and outlier rej. by cross-corr. & Signal-averaged ECG & PRD, CC, WDIST + kNN & Private & 50 & Yes & IDR:\\PRD\\CC\\WD & \hspace{0.001cm}\\70\%\\80\%\\89\% \tabularnewline\hline

Irvine{ }\etal{ }\cite{Irvine2008} & 2008 & BPF 0.05--60 Hz & Beat segm. and alignment by R-peaks using AC, amplitude normalization & Covariance matrix eigenvectors / PCA & kNN & Private & 39 & No & IDR & 100\% \tabularnewline\hline

Boumbarov{ }\etal{ }\cite{Boumbarov2009} & 2009 & HPF 0.05 Hz + DWT soft thresholding & HMM-GMM PQRST segmentation & Cardiac cycle vector matrix / PCA and LDA & RBF NN & Private & 9 & No & IDR & 83.3\% \tabularnewline\hline

Fang{ }\etal{ }\cite{Fang2009} & 2009 & BPF 2--50 Hz & R detection, 5-beat average, amplitude normalization & Avg. beat phase space portrait & Correlation; Mutual nearest pt. dist. + kNN & Private (one or three leads) & 100 & No & IDR:\\ 1 l.\\ 3 l. &  \hspace{0.001cm}\\93\%\\99\% \tabularnewline\hline

Fatemian{ }\etal{ }\cite{Fatemian2009} & 2009 & DWT 3rd scale + Mov. Average & Discarding outlier beats, DWT QRS, T, \& P delineation  & Heart-rate normalized heartbeat construction & Correlation + kNN & PTB + MIT NSR & 27 & No & IDR & 99.6\%\tabularnewline\hline

Guennoun{ }\etal{ }\cite{Guennoun2009} & 2009 & LPF 30 Hz & - & Fiducial amplitude and time feat. / Physiological-state-indepen. feature select. & Mahalanobis dist. + Thresh. and Voting & Private & 16 & No & FRR\\FAR & 0.01\%\\0\% \tabularnewline\hline

Coutinho{ }\etal{ }\cite{Coutinho2010} & 2010 & BPF 2--30 Hz & Beat segment. and alignment, 10-beat avg. & Uniformly quantized avg. beats & Ziv-Merhav relative entropy + kNN & Private & 19 & Yes & IDR & 99.5\% \tabularnewline\hline

Fatemian{ }\etal{ }\cite{Fatemian2010} & 2010 & DWT 3rd scale & DWT beat delineation, P-wave time norm. & Avg. ensemble heartbeat & Correlation + kNN & Private & 21 & No & IDR\\EER &  95.4\%\\3.3\% \tabularnewline\hline

Ghofrani{ }\etal{ }\cite{Ghofrani2010} & 2010 & BPF 0.5--150 Hz + NF 50 Hz & - & AR; PSD; Lyapunov; Approximation Entropy; Higuchi Fractal Dim.; Shannon Entropy & kNN; MLP; PNN & PTB & 12 & No & IDR:\\AR\\ApEn\\Hig.\\Lya.\\Sha. & \hspace{0.001cm}\\98.6\%\\94.3\%\\87.4\%\\96.7\%\\92.8\% \tabularnewline\hline

Li{ }\etal{ }\cite{Li2010} & 2010 & - & Beat segment., amplitude and time norm. & Hermite poly. expansion; Cepstral features / HLDA & SVM + GMM-UBM fusion & MIT NSR & 18 & No & IDR\\EER & 98.3\%\\0.5\% \tabularnewline\hline

Murthy{ }\etal{ }\cite{Venkatesh2010} & 2010 & BPF & Pan-Tompkins & P, T, ST, PR, QRS and QT intervals / FLDA & DTW + kNN & MIT NSR & 15 & No & IDR & 96\% \tabularnewline\hline

Odinaka{ }\etal{ }\cite{Odinaka2010} & 2010 & HPF 0.5 Hz + LPF 500 Hz + NF 60 Hz & R detect., beat normaliz. and Hamming seg. & Log-STFT spectrogram / Bin selection & Gaussian models LLR & Private & 269 & No & IDR\\EER & 99\%\\0.37\% \tabularnewline\hline

Sasikala{ }\etal{ }\cite{Sasikala2010} & 2010 & Median Filters + DWT & QRS detection w/ Daubechies DWT, P and T detection & Fiducial amplitudes and differences & Correlation & MIT Arrh. & 10 & No & IDR & 62.7\% \tabularnewline\hline

Tawfiq{ }\etal{ }\cite{Tawfiq2010} & 2010 & HPF 1 Hz + LPF 40 Hz & R-alignment, amplit. norm., QRS segment. & QRS DCT coefficients & Neural Networks & Private & 22 & No & IDR & 99.1\% \tabularnewline\hline

Ye{ }\etal{ }\cite{Ye2010} & 2010 & BPF & Beat segment. with R location annotations & Daubechies DWT / ICA & RBF SVM & MIT Arrh.\\MIT NSR1\\MIT LT\\MIT NSR2 & 47\\18\\65\\18 & No\\No\\No\\No & IDR\\IDR\\IDR\\IDR & 99.6\%\\99.3\%\\98.1\%\\97.5\%\tabularnewline\hline

Coutinho{ }\etal{ }\cite{Coutinho2011} & 2011 & BPF 2--30 Hz & Beat segment., alignment, 10-beat avg. & User-tuned Lloyd-Max quantised avg. beat & Ziv-Merhav cross parsing similarity + kNN & Private & 19 & Yes & EER & 0.36\% \tabularnewline\hline

Lourenço{ }\etal{ }\cite{Lourenco2011} & 2011 & BPF 0.5--30 Hz & Engelse-Zeelenberg, beat segment., amplitude and time normaliz. & Avg. normalized beat & Euclidean dist. + kNN & Private & 16 & Yes & IDR\\EER & 94.3\%\\13\% \tabularnewline\hline

Matta{ }\etal{ }\cite{Matta2011} & 2011 & BPF & Fixed-length window blind segmentation & Autocorrelation coeff. / LDA & Euclidean dist. + kNN & Private & 10 & No & IDR\\TPIR3 & 75\%\\99\% \tabularnewline\hline

Safie{ }\etal{ }\cite{Safie2011} & 2011 & BPF 2--40 Hz & ECGPUWAVE fiducial detect., P-R \& P-T quality check, 5-beat avg. & Pulse Active Ratio & Euclidean dist. + kNN & PTB (healthy or w/ arrhythmias) & 112 & No & EER:\\Heal.\\Arrh. &  \hspace{0.001cm}\\9.98\%\\19.2\% \tabularnewline\hline

Shen{ }\etal{ }\cite{Shen2011} & 2011 & BPF 1--50 Hz & Pan-Tompkins, heartbeat segmentation & Amplitudes, durations, slopes, angles, and QRS area / LDA & Correlation + kNN & Private & 168 & Yes & IDR & 98\% \tabularnewline\hline

Sufi{ }\etal{ }\cite{Sufi2011} & 2011 & - & P, QRS, and T segmentation, cardioid 2D loop generation & Cardioid graph centroid, extremas, area, and perimeter & Straight line and percentage dist. + kNN & MIT Arrh. & - & No & MIDR\\FAR\\FRR & 1\%\\0.5\%\\0.5\% \tabularnewline\hline

Agrafioti{ }\etal{ }\cite{Agrafioti2012} & 2012 & BPF 1--40 Hz & Fixed-length window blind segmentation & Autocorrelation coeff. / LDA & Euclidean dist. + kNN & Private & 42 & No & EER & 3.96\%\tabularnewline\hline

Belgacem{ }\etal{ }\cite{Belgacem2012} & 2012 & BPF 1--40 Hz & Amplitude normal., beat segmentation and R-peak alignment & Avg. beat Daubechies DWT & Random Forest & MIT Arrh. + ST-T + MIT NSR + PTB + Private & 80 & No & FAR\\FRR & 0.60\%\\0.58\% \tabularnewline\hline

Lourenço{ }\etal{ }\cite{Lourenco2012} & 2012 & BPF 1--30 Hz & Steep-slope R detection, beat segment. & Segmented heartbeats & kNN & Private & 32 & Yes & EER & 9.39\% \tabularnewline\hline

Singh{ }\etal{ }\cite{Singh2012} & 2012 & - & QRS, P, and T delineation & Interval, angle, and amplitude fid. feat. & Euclidean dist. + kNN & MIT Arrh. + ST-T + MIT NSR + QT & 73 & No & EER & 10.8\% \tabularnewline\hline

Belgacem{ }\etal{ }\cite{Belgacem2013} & 2013 & BPF 1--40 Hz & Beat segment., amplitude normalization, 100-beat avg. & Avg. beat Daubechies DWT & Random Forest & MIT Arrh. + ST-T + MIT NSR + PTB + Private & 80 & No & IDR\\FAR\\FRR & 100\%\\0.63\%\\0.66\% \tabularnewline\hline

Coutinho{ }\etal{ }\cite{Coutinho2013} & 2013 & BPF 2--30 Hz & Beat segm. and alignment, 10-beat mean wave & Fid. latency and amplitude from mean waveform subsampling & Euclidean dist. + kNN & PTB\\\hspace{0.001cm}\\Private & 51\\\hspace{0.001cm}\\26 & No & IDR\\EER\\IDR\\EER & 99.9\%\\0.01\%\\99.6\%\\0.70\% \tabularnewline\cline{3-11}

&  & BPF 2--30 Hz & Beat segm. and alignment, 10-beat mean wave & User-tuned Lloyd-Max quantized heartbeats & Ziv-Merhav cross-pars. similarity + kNN & PTB\\\hspace{0.001cm}\\Private & 51\\\hspace{0.001cm}\\26 & No & IDR\\EER\\IDR\\EER & 99.4\%\\0.13\%\\99.9\%\\0.29\% \tabularnewline\hline

Labati{ }\etal{ }\cite{Labati2013} & 2013 & HPF 0.5 Hz + NF 50 Hz & R detection and QRS segment., rejection of low-correl. seg. & QRS segment set templates & Cross-corr. similarity mat. + kNN & E-HOL 24h & 185 & No & EER & 5.36\% \tabularnewline\hline

Matos{ }\etal{ }\cite{Matos2013} & 2013 & HPF 0.5 Hz + NF 50 and 150 Hz & R detection, beat segment., Hamming segmentation & STFT spectrogram + Spectral zoom / Bin selection & Gaussian models LLR & Private & 27 & No & EER & 10\% \tabularnewline\hline

Silva{ }\etal{ }\cite{Silva2013} & 2013 & BPF 5--20 Hz & R detection and beat segmentation & Mean and median ensemble beats & Euclidean and cosine dist. + kNN and SVM & Private & 63 & Yes & EER:\\kNN\\SVM & \hspace{0.001cm}\\0.99\%\\9.10\% \tabularnewline\hline

Wang{ }\etal{ }\cite{Wang2013} & 2013 & - & Sliding-window segmentation & Max-pooling representation elements & kNN & PTB & 100 & No & IDR & 99.5\% \tabularnewline\hline

Zhao{ }\etal{ }\cite{Zhao2013} & 2013 & HPF and DWT soft-thresh. & Beat segment. and normaliz., quality check & EEMD main IMFs and their PSD / PCA & kNN & ST-Change\\LTST\\PTB & 15\\18\\12 & No\\No\\No & IDR\\IDR\\IDR & 98.0\% \\ 95.8\% \\ 96.0\% \tabularnewline\hline

Ergin{ }\etal{ }\cite{Ergin2014} & 2014 & - & Segm. 2 s sliding windows & QRS fid., time domain, wavelet trans. and PSD & C4.5 and Bayesian Network  & MIT NSR & 18 & No & F-s.:\\C4.5\\Bay. & \hspace{0.001cm}\\0.97\%\\0.96\% \tabularnewline\hline

Iqbal{ }\etal{ }\cite{Iqbal2014} & 2014 & - & QRS detection and segment. & QRS cardioid graph coord. & MLP & Private & 30 & No & IDR & 96.4\% \tabularnewline\hline

Labati{ }\etal{ }\cite{Labati2014} & 2014 & HPF 0.5 Hz + NF 50 Hz & R detection, QRS segment. & QRS segments & Cross-corr. simil. kNN & E-HOL 24h & 185 & No & EER & 5.36\% \tabularnewline\hline

Lin{ }\etal{ }\cite{Lin2014} & 2014 & - & Time-delay space reconst., Chaos theory feature extract. & Corr. dimension Lyapunov exp. & SVM & Private & 26 & Yes & IDR & 81.7\% \tabularnewline\hline

Lourenço{ }\etal{ }\cite{Lourenco2014} & 2014 & - & QRS detection, DMEAN & Mean ensemble beats & SVM & Private & 63 & Yes & EER & 2.5\% \tabularnewline\hline

Matos{ }\etal{ }\cite{Matos2014} & 2014 & LPF 50 Hz &  Slope sum + thresh. for R det., beat segm. & STFT window features / Kullback-Leibler & LLR + kNN & Private & 10 & Yes & IDR\\EER & 100\%\\14\% \tabularnewline\hline

\citet{Pathoumvanh2014} & 2014 & BPF 0.4--40 Hz & R detection and hearbeat segmentation & CWT / FLDA & Euclidean dist. + kNN & Private (normal + increased HRV) & 10 & No & IDR:\\Norm.\\HRV & \hspace{0.001cm}\\97\%\\80\% \tabularnewline\hline

Zhou{ }\etal{ }\cite{Zhou2014} & 2014 & BPF 0.5--40 Hz &  R detection, interval vs. amplitude plot & Signal between 3 consec. R peaks & DTW path + kNN & Private & 20 & No & HTER & 1.45\%\tabularnewline\hline

Brás{ }\etal{ }\cite{Bras2015} & 2015 & NF 50 Hz + Moving Avg. + LPF 40 Hz & Amplitude norm., SAX conversion & Kolmogorov-based normalised rel. compression & kNN & PTB & 52 & No & IDR & 99.9\% \tabularnewline\hline

Choudhary{ }\etal{ }\cite{Choudhary2015} & 2015 & DCT & FOGD peak det., peak correction, heartbeat segmentation & Avg. ensemble beat & RMSE, PRD, NCC, WWPRD, WDIST + kNN & MIT Arrh. + STC + QT + MIT NSR + SLP & 127 & No & FAR\\FRR & 5.8\%\\11.6\% \tabularnewline\hline

Dar{ }\etal{ }\cite{Dar2015} & 2015 & Poly. line fitting & Local-maxima R det., QRS segmentation & Haar Transform / GBFS & kNN & MIT Arrh.\\MIT NSR\\ECG-ID & 47\\18\\90 & No\\No\\No & IDR\\IDR\\IDR & 93.1\%\\99.4\%\\83.2\% \tabularnewline\hline

Dar{ }\etal{ }\cite{Dar2015b} & 2015 & Poly. line fitting & Local-maxima R det., QRS segmentation & Haar Transform and HRV / GBFS & Random Forest & MIT Arrh.\\\hspace{0.001cm}\\MIT NSR\\\hspace{0.001cm}\\ECG-ID & 47\\\hspace{0.001cm}\\18\\\hspace{0.001cm}\\90 & No \\\hspace{0.001cm}\\No\\\hspace{0.001cm}\\No & IDR\\FAR\\IDR\\EER\\IDR\\FAR & 95.9\%\\4.1\%\\100\%\\0\%\\83.9\%\\16.1\% \tabularnewline\hline

\citet{Jahiruzzaman2015} & 2015 & BPF 0.5--45 Hz & None & CWT and Chaotic Encryption & Identification of unique CE sequen. & MIT Arrh. & 11 & No & IDR & 96.9\% \tabularnewline\hline

Carreiras{ }\etal{ }\cite{Carreiras2016} & 2016 & BPF 5--20 Hz & QRS det., beat seg. and align., DMEAN & Segmented heartbeats & kNN & Private & 618 & No & EER\\MIDR & 9.01\%\\15.6\% \tabularnewline\hline

Chun{ }\etal{ }\cite{Chun2016} & 2016 & DWT 3rd scale & Pan-Tompkins, heartbeat segmentation & Guided filtering avg. beat / PCA & DTW or Euclidean dist. + kNN & ECG-ID & 89 & No & EER:\\DTW\\Eucl. & \hspace{0.001cm}\\5.2\%\\2.4\% \tabularnewline\hline

Hejazi{ }\etal{ }\cite{Hejazi2016} & 2016 & DWT 3rd scale & Fixed-length window blind segmentation & Autocorrelation coeff. / KPCA & SVM & Private & 52 & Yes & IDR\\FAR\\FRR & 76.3\%\\3.5\%\\4.83\% \tabularnewline\hline

Louis{ }\etal{ }\cite{Louis2016} & 2016 & BPF 1--40 Hz & Pan-Tompkins, beat segment. \& alignment & 1D multi-res. LBP & Bagging & UofTDB & 1012 & Yes & EER & 7.89\% \tabularnewline\hline

Porée{ }\etal{ }\cite{Poree2016} & 2016 & LPF 45 Hz & Pan-Tompkins, beat segment. & 10 beat avg. ensemble & Discrimination coeff./kNN & Private & 14 & No & IDR & 100\% \tabularnewline\hline

Rezgui{ }\etal{ }\cite{Rezgui2016} & 2016 & BPF 2--40 Hz & ECGPUWAVE QRS detection, segmentation & Amplitudes, areas, intervals and fid. slopes & SVM & MIT NSR + Arrh. & No & 48 & IDR & 98.8\% \tabularnewline\hline

Waili{ }\etal{ }\cite{Tuerxunwaili2016} & 2016 & HPF 0.05 Hz + LPF 40 Hz & Pan-Tompkins, signal mean subtract. norm. & 12 QRS fid. amplitudes & MLP & PTB & 14 & No & IDR & 96\% \tabularnewline\hline

Camara{ }\etal{ }\cite{Camara2017} & 2017 & BPF 0.67--45 Hz & None & Walsh-Hadamard features, outliers rejected & kNN & MIT NSR & 10 & No & IDR & 94.8\% \tabularnewline\hline

\citet{Eduardo2017} & 2017 & BPF 5--20 Hz & Beat cropping, DMEAN outlier detection & Fully-connected autoencoder representations & kNN & Private & 709 & No & MIDR & 0.91\%  

\tabularnewline\hline

Islam{ }\etal{ }\cite{Islam2017} & 2017 & BPF 0.25--40 Hz & Curvature QRS detect., beat segm., time norm., AC outlier reject. & Avg. ensemble heartbeats / PCA & Euclidean dist. & Private & 112 & Yes & EER & 10.5\% \tabularnewline\hline

Karimian{ }\etal{ }\cite{Karimian2017} & 2017 & BPF 1--40 Hz & Pan-Tompkins, heartbeat segm., heart rate QT time normalization & DWT, Maximal Overlap DWT, DCT, Normalize Convolute Norm. encrypted by IOMBA & Key matching & PTB \\ BioSec & 290 \\ 13 & No\\Yes & Rel.\\Rel. & 97.4\%\\94.7\% \tabularnewline\hline

Komeili{ }\etal{ }\cite{Komeili2017} & 2017 & BPF 0.5--40 Hz & Heartbeat segm. and outlier removal, z-score norm. & CWT, STFT, AC, max., st. dev., kurtosis and skewness / MSFS & SVM & UofTDB (different session or posture) & 82\\ & Yes & EER:\\Sess.\\Post. & \hspace{0.001cm}\\6.9\%\\3.7\% \tabularnewline\hline 

Paiva{ }\etal{ }\cite{Paiva2017} & 2017 & - & Pan-Tompkins, LPF for Q, S, T detection & Fiducial distances ST, RT, and QT & SVM & PTB & 10 & No & IDR\\FAR\\FRR & 97.5\% \\ 5.71\% \\ 3.44\% \tabularnewline\hline

Pinto{ }\etal{ }\cite{Pinto2017} & 2017 & Savitzky-Golay + moving avg. & Trahanias, heartbeat segmentation, z-score, NCCC & DCT coefficients & SVM & Private & 6 & Yes & IDR\\EER & 94.9\%\\2.66\% \tabularnewline\hline

\citet{Salloum2017} & 2017 & - & Pan-Tompkins, beat segm., z-score norm. & Sequences of appended heartbeats & RNN with LSTM/GRU & ECG-ID\\MIT Arrh. & 90\\47 & No & IDR\\EER & 100\%\\0\%  

\tabularnewline\hline

Tan{ }\etal{ }\cite{Tan2017} & 2017 & BPF 2--50 Hz & Pan-Tompkins, P--QRS--T segmentation & Temporal, amplitude, and angle fid. + DWT coefficients & Random Forests + WDIST kNN & Private\\ECG-ID\\MIT Arrh.\\MIT NSR\\Combined & 30\\89\\47\\18\\184 & Yes\\No\\No\\No\\No & IDR\\IDR\\IDR\\IDR\\IDR & 99.4\%\\100\%\\100\%\\98.8\%\\99.5\% \tabularnewline\hline

Wieclaw{ }\etal{ }\cite{Wieclaw2017} & 2017 & BPF 4--35 Hz & Hamilton R detect., outlier rejection & Individual heartbeats & MLP & Private & 18 & Yes & IDR & 89\% \tabularnewline\hline

Zaghouani{ }\etal{ }\cite{Zaghouani2017} & 2017 & Median filter & Fixed-length window segm. & AC / DCT, feature security locking & Norm. Euclidean dist. & ECG-ID & 90 & No & EER & 15\% \tabularnewline\hline

\citet{Zhang2017} & 2017 & BPF 2--50 Hz & Normalisation, 2 s blind segmentation, DWT & Autocorrelation, component selection & Multiresolution 1D CNN & CEBSDB\\WECG\\FANTASIA\\MIT NSR\\STDB\\MIT Arrh.\\AFDB\\VFDB & 20\\22\\40\\18\\28\\47\\23\\22 & No & IDR & 99.0\%\\94.5\%\\97.2\%\\95.1\%\\90.3\%\\91.1\%\\93.9\%\\86.6\%
\tabularnewline\hline

\citet{Zhang2017b} & 2017 & None & Heartbeat detection and segmentation & Cardioid-like 2D representations & 2D CNN & Private & 10 & Yes & IDR & 98.4\% 

\tabularnewline\hline

Dong{ }\etal{ }\cite{Dong2018} & 2018 & - & Construction of 3D VCG with 12-lead ECG & Banks of state and errors from 2D VCG & Minimum L1 norm of bank of errors & PTB (healthy and ill subjects) & \hspace{0.001cm}\\14\\99\\113 & No & IDR:\\Healt.\\Ill\\All & \hspace{0.001cm}\\98.3\%\\93.3\%\\92.8\%
\tabularnewline\hline

\citet{Guven2018} & 2018 & HPF 0.5 Hz + LPF 150 Hz + mov. avg. & Z-score norm., 5 s window segmentation & AC, DCT, cepstral, and QRS features & Euclidean dist. + kNN & Private & 30\\45\\60 & Yes & IDR\\IDR\\IDR & 100\%\\100\%\\98.3\% \tabularnewline\hline

\citet{Kim2018} & 2018 & - & Min-max norm. Pan-Tompkins, beat ensemble & Haar Wavelet Transform & Fuzzy membership ANN & - & 73 & No & FRR\\FAR & 1.68\%\\5.84\%
\tabularnewline\hline

Lee{ }\etal{ }\cite{Lee2018} & 2018 & BPF 0.3--35 Hz, 6th order polynomial line fitting & R~\&~T~detection, R-R segmentation, resampling & R-R segments, including two or three heartbeats (hb.) & Cosine, euclidean, manhattan dists., \& CC & Private & 55 & No & IDR:\\2~hb.\\3~hb. & \hspace{0.001cm}\\89.9\%\\93.3\%
\tabularnewline\hline

\citet{Luz2018} & 2018 & BPF 0.5--40 Hz & Pan-Tompkins, beat segmentation, z-score normalisation, outlier removal & Raw 1D heartbeats and 2D spectrogram representations & 1D CNN + 2D CNN & CYBHi\\\hspace{0.001cm}\\\hspace{0.001cm}\\\hspace{0.001cm}\\UofTDB & 61\\\hspace{0.001cm}\\\hspace{0.001cm}\\\hspace{0.001cm}\\1019 & Yes\\\hspace{0.001cm}\\\hspace{0.001cm}\\\hspace{0.001cm}\\Yes & EER:\\Raw\\Spect.\\Fusion\\Raw\\Spect.\\Fusion &  \hspace{0.001cm}\\14.1\%\\26.4\%\\12.8\%\\16.9\%\\19.4\%\\14.3\%
\tabularnewline\hline

Pal{ }\etal{ }\cite{Pal2018} & 2018 & HPF 1 Hz, NF 50 Hz, LPF 40 Hz & DWT fiducial det., P-QRS-T segmentation & Interval,~amplitude, angle, and area fiducial features / KPCA & Euclidean distance & PTB & 100 & No & IDR & 97.1\% 
\tabularnewline\hline

\citet{Wu2018} & 2018 & BPF 1--40 Hz & Pan-Tompkins, beat segmentation & 1D CNN for feat. extraction and outlier removal & Attention-based LSTM & ECG-ID\\\hspace{0.001cm}\\MIT Arrh. & 90\\\hspace{0.001cm}\\47 & No\\\hspace{0.001cm}\\No & IDR\\EER\\IDR\\EER & 97.5\%\\0.52\%\\99.7\%\\0.02\% 
\tabularnewline\hline

\citet{Carvalho2019b} & 2019 & LPF 30 Hz, 1st order derivative & Lloyd-Max quantisation, 10 s segmentation & Extended-alphabet Finite-Context Model (xaFCM) & Normalised Relative Compression (NRC) & Private & 25 & No & IDR & 89.3\% 
\tabularnewline\hline

\citet{Chu2019} & 2019 & RNN-based noise removal & Threshold beat detection, segmentation, and concatenation & Parallel Multiscale 1D ResNet & Template similarity or fully-connected layer & ECG-ID\\\hspace{0.001cm}\\PTB\\\hspace{0.001cm}\\MIT Arrh. & 90\\\hspace{0.001cm}\\290\\\hspace{0.001cm}\\47 & No\\\hspace{0.001cm}\\No\\\hspace{0.001cm}\\No & IDR\\EER\\IDR\\EER\\IDR\\EER & 97.8\%\\2.00\%\\99.3\%\\0.59\%\\94.9\%\\4.74\%
\tabularnewline\hline

\citet{Ciocoiu2019, Ciocoiu2020} & 2019 2020 & BPF 1--40 Hz & DTW R-peak detection, beat segmentation & S-Transform, Gramian Angular Fields, Phase-Space Trajectories, or Recurrence Plots & 2D CNN & UofTDB\\\hspace{0.001cm}\\CYBHi & 52\\\hspace{0.001cm}\\65 & Yes\\\hspace{0.001cm}\\Yes & IDR\\EER\\IDR\\EER & 95.6\%\\5.48\%\\95\%\\8.6\%
\tabularnewline\hline

\citet{Hammad2019} & 2019 & - & 2 s blind segmentation, z-score norm. & None & 1D Res-Net with Attention & PTB\\CYBHi & 290\\65 & No\\Yes & IDR & 98.9\%\\99.3\%  
\tabularnewline\hline

\citet{Labati2018} & 2019 & HPF 0.5 Hz, NF & R detection, QRS cropping and concatenation, normalisation & 1D CNN & Softmax layer or template similarity & PTB\\E-HOL 24h & 52\\92 & No\\No & IDR\\EER & 100\%\\2.15\% 
\tabularnewline\hline

\citet{Pinto2019Deep} & 2019 & None & 5 s blind segmentation, z-score norm. & None & 1D CNN & UofTDB & 1019 & Yes & IDR & 96.1\%
\tabularnewline\hline

\citet{Pinto2019b} & 2019 & None & 5 s blind segmentation, z-score norm. & 1D CNN & Template similarity & UofTDB\\PTB\\CYBHi & 1018\\290\\128 & Yes\\No\\Yes & EER & 7.86\%\\11.0\%\\16.3\%
\tabularnewline\hline

\citet{Ranjan2019} & 2019 & BPF 1--20 Hz & Pan-Tompkins, QRS cropping, normalisation & Nine QRS concatenated in 2D & 2D CNN & ECG-ID & 90 & No & EER & 2\% 
\tabularnewline\hline

\citet{Zhang2019c} & 2019 & BPF 1--40 Hz & Blind segm., z-score norm. & Residual 1D CNN & kNN + majority voting & PTB\\CEBSDB\\MIT NSR & 234\\20\\18 & No\\No\\No & IDR & 98.7\%\\99.9\%\\92.9\% 
\tabularnewline\cline{3-11}

& & BPF 1--40 Hz & Blind segm., z-score norm. & Residual 1D CNN & SVM + majority voting & PTB\\CEBSDB\\MIT NSR & 234\\20\\18 & No\\No\\No & IDR & 99.5\%\\100\%\\95.3\%
\tabularnewline\hline

\citet{Zhang2019b} & 2019 & DWT hard thresholding & 3 s blind segmentation & Recurrence plots & GoogLeNet CNN & ECG-ID & 90 & No & IDR\\EER & 96.3\%\\5.82\% 
\tabularnewline\hline

\citet{Alduwaile2020} & 2020 & BPF & QRS detection, beat segm. & 1D CNN & Softmax layer & PTB & 100 & No & IDR & 99.9\%  
\tabularnewline\hline

\citet{Belo2020} & 2020 & Moving average & Blind segmentation, max. normalisation & Blind segment & RNN & FANTASIA\\\hspace{0.001cm}\\MIT NSR + Arrh. + LT\\CYBHi & 20\\\hspace{0.001cm}\\72\\\hspace{0.001cm}\\63 & No\\\hspace{0.001cm}\\No\\\hspace{0.001cm}\\Yes & IDR\\EER\\IDR\\EER\\IDR\\EER & 100\%\\0.02\%\\92.7\%\\1.25\%\\63.5\%\\4.3\%
\tabularnewline\cline{3-11}

& & Moving average & Blind segmentation, max. normalisation & Blind segment + extracted heartbeat & Temporal Dual-stream CNN & FANTASIA\\\hspace{0.001cm}\\MIT NSR + Arrh. + LT\\CYBHi & 20\\\hspace{0.001cm}\\72\\\hspace{0.001cm}\\63 & No\\\hspace{0.001cm}\\No\\\hspace{0.001cm}\\Yes & IDR\\EER\\IDR\\EER\\IDR\\EER & 99.1\%\\0.02\%\\96.4\%\\0.08\%\\100\%\\0.0\%
\tabularnewline\hline

\citet{Bento2020} & 2020 & Hann window filter, mov. avg. & Normalisation & Spectrogram & 2D CNN & FANTASIA\\ECG-ID & 40\\90 & No\\No & IDR & 99.4\%\\94.2\%
\tabularnewline\cline{3-11}

 & & Hann window filter, mov. avg. & Normalisation & Spectrogram & DenseNet & FANTASIA\\ECG-ID & 40\\90 & No\\No & IDR & 99.8\%\\96.9\% 
\tabularnewline\hline

\citet{Byeon2020} & 2020 & Convolution-based denoising & R-peak detection, beat segmentation & Log-spectrogram, melspectrogram, spectrogram, MFCC, and scalogram & Xception, ResNet, DenseNet ensembles  & PTB & 290 & No & IDR:\\Xcep.\\ResN.\\Dens. & \hspace{0.001cm}\\99.1\%\\99.0\%\\99.0\%
\tabularnewline\hline

\citet{Ingale2020} & 2020 & BPF 1--40 Hz & R-peak detection, beat segmentation & Thirty amplitude and time fiducial features or DWT & Euclidean distance or DTW & PTB\\\hspace{0.001cm}\\MIT Arrh.\\\hspace{0.001cm}\\CEBSDB\\\hspace{0.001cm}\\CYBHi\\\hspace{0.001cm}\\ECG-ID\\\hspace{0.001cm}\\Private & 290\\\hspace{0.001cm}\\47\\\hspace{0.001cm}\\20\\\hspace{0.001cm}\\125\\\hspace{0.001cm}\\90\\\hspace{0.001cm}\\1119 & No\\\hspace{0.001cm}\\No\\\hspace{0.001cm}\\No\\\hspace{0.001cm}\\Yes\\\hspace{0.001cm}\\No\\\hspace{0.001cm}\\Yes & IDR\\EER\\IDR\\EER\\IDR\\EER\\IDR\\EER\\IDR\\EER\\IDR\\EER & 100\%\\2\%\\100\%\\4\%\\100\%\\0\%\\100\%\\0.5\%\\96.7\%\\2.3\%\\100\%\\1.2\%
\tabularnewline\hline

\citet{Jyotishi2020} & 2020 & HPF 0.5 Hz + NF 50 Hz + NF 100 Hz & Min-max normalisation & 100 ms sliding windows & LSTM & PTB\\MIT Arrh.\\ECG-ID\\CYBHi & 290\\47\\90\\63 & No\\No\\No\\Yes & IDR & 97.3\%\\96.8\%\\93.1\%\\79.4\% 
\tabularnewline\hline

\citet{Kim2020} & 2020 & Derivative filter, moving avg. & Min-max normalisation, beat segmentation & Sequence of heartbeats & Bi-LSTM & MIT NSR\\\hspace{0.001cm}\\MIT Arrh. & 18\\\hspace{0.001cm}\\47 & No\\\hspace{0.001cm}\\No & IDR\\F-s.\\IDR\\F-s. & 100\%\\1.0\\99.8\%\\0.99 
\tabularnewline\hline

\citet{Lehmann2020} & 2020 & BPF 3--45 Hz & Engelse-Zeelenberg, beat segmentation, outlier removal & QRS fiducial statistics, feat. selection with correlation matrix & RF\\SVM\\MLP & Private & 20 & No & EER & 16.7\%\\25.7\%\\17.7\% 
\tabularnewline\hline

\citet{Li2020} & 2020 & BPF 1--40 Hz & Pan-Tompkins, beat segmentation and avg. ensemble & GNMF & Sparse representation-based matching & ECG-ID\\MIT Arrh. & 90\\47 & No\\No & IDR & 98.0\%\\100\% 
\tabularnewline\hline

\citet{Li2020b} & 2020 & BPF 2--50 Hz & Normalisation, R-peak detection, beat segmentation, outlier detection & Segmented heartbeats & Cascaded 1D CNN & FANTASIA\\CEBSDB\\MIT NSR\\STDB\\AFDB & 40\\20\\18\\28\\23 & No\\No\\No\\No\\No & IDR & 99.3\%\\93.1\%\\91.4\%\\92.7\%\\89.7\%
\tabularnewline\hline

\citet{Pinto2020iwbf} & 2020 & None & 5 s blind segmentation, z-score norm. & 1D CNN & Template similarity & UofTDB & 1018 & Yes & EER & 12.6\% 
\tabularnewline\hline

\citet{Pinto2020Explaining} & 2020 & None & 5 s blind segmentation, z-score norm. & None & 1D CNN & PTB\\UofTDB & 290\\1018 & No\\Yes & IDR & 97.7\%\\91.5\%  
\tabularnewline\hline

\citet{Randazzo2020} & 2020 & - & Heartbeat segmentation & AC/DCT & MLP & Private & 5 & Yes & IDR & 99.1\% 
\tabularnewline\hline

\citet{TiradoMartin2020} & 2020 & BPF 1--35 Hz & R detection, QRS segmentation & Differentiated QRS & MLP & Private & 55 & No & EER & 2.69\%
\tabularnewline\hline

\citet{Wang2020} & 2020 & BPF 1--40 Hz & Pan-Tompkins, beat segmentation & Multi-Scale Differential Features fusion of 1D Multi-Resolution LBPs (MSDF-1DMRLBP) & Euclidean distance & MIT Arrh.\\\hspace{0.001cm}\\ECG-ID\\\hspace{0.001cm}\\PTB\\\hspace{0.001cm}\\UofTDB & 47\\\hspace{0.001cm}\\90\\\hspace{0.001cm}\\248\\\hspace{0.001cm}\\46 & No\\\hspace{0.001cm}\\No\\\hspace{0.001cm}\\No\\\hspace{0.001cm}\\Yes & IDR\\EER\\IDR\\EER\\IDR\\EER\\IDR\\EER & 94.7\%\\2.73\%\\100\%\\3.3\%\\98.2\%\\2.55\%\\100\%\\2.17\% 
\tabularnewline\hline

\citet{Benouis2021} & 2021 & Savitzky-Golay filtering & Pan-Tompkins, beat segmentation & 1D Local Difference Patterns (1D LDP) & kNN, SVM, PNN & ECG-ID & 90 & No & IDR\\EER & 93.3\%\\3.05\%
\tabularnewline\hline

\citet{Ibtehaz2021} & 2021 & None & U-Net based R-peak detection, beat segmentation & 1D CNN & Softmax layer (identification) or siamese architecture (verification) & MIT Arrh.\\\hspace{0.001cm}\\ECG-ID\\\hspace{0.001cm}\\PTB\\\hspace{0.001cm}\\MIT NSR\\\hspace{0.001cm}\\CYBHi & 47\\\hspace{0.001cm}\\90\\\hspace{0.001cm}\\290\\\hspace{0.001cm}\\18\\\hspace{0.001cm}\\63 & No\\\hspace{0.001cm}\\No\\\hspace{0.001cm}\\No\\\hspace{0.001cm}\\No\\\hspace{0.001cm}\\Yes & IDR\\EER\\IDR\\EER\\IDR\\EER\\IDR\\EER\\IDR & 98.2\%\\6.36\%\\96.2\%\\1.29\%\\99.7\%\\5.66\%\\99.5\%\\5.17\%\\73.9\% 
\tabularnewline\hline

\citet{Ivanciu2021} & 2021 & - & Christov R-peak detection, beat segmentation & Heartbeat plot images & Siamese 2D CNN & ECG-ID & 90 & No & IDR\\FAR\\FRR\\Sens. & 86.5\%\\13.7\%\\12.7\%\\87.3\% 
\tabularnewline\hline

\citet{Pinto2021Secure} & 2021 & None & 5 s blind segmentation, z-score norm. & 1D CNN & Template similarity & UofTDB & 1018 & Yes & EER & 12.6\% 
\tabularnewline\hline

\citet{Srivastva2021} & 2021 & BPF 0.5--30 Hz & Pan-Tompkins, beat segmentation, z-score norm. & 2D heartbeat image & ImageNet pretrained DenseNet and ResNet models ensemble & PTB\\CYBHi & 290\\63 & No\\Yes & IDR & 99.7\%\\99.7\% 
\tabularnewline\hline

\citet{Tai2021} & 2021 & BPF 1--40 Hz & Normalisation, R-peak detection, fixed-window segmentation & CNN trained with N-Pair Loss & Euclidean distance & ECG-ID + E-HOL 24h + YSYW + MIT NSR & 2293 & No & AUC & 97.0\%  
\tabularnewline\hline 

\citet{Thentu2021} & 2021 & - & Pan-Tompkins, beat segmentation &  Multi-scale CWT representation & Various ImageNet pre-trained 2D networks & CEBSDB\\PTB & 20\\290 & No\\No & IDR & 99.9\%\\99.5\%  
\tabularnewline\hline

\citet{TiradoMartin2021} & 2021 & BPF 1--35 Hz & R-peak detection, QRS segmentation & CNN & LSTM + softmax fully-connected layer & Private & 105 & No & IDR\\EER & 94.6\%\\0.04\% 
\tabularnewline\hline

\citet{Wang2021} & 2021 & BPF 1--40 Hz & Pan-Tompkins, heartbeat segmentation & Segmented heartbeats and their STFT & Locality-preserving semantic space learning & MIT Arrh.\\PTB\\CYBHi & 47\\290\\63 &  No\\No\\Yes & IDR & 97.8\%\\98.9\%\\96.8\%
\tabularnewline\hline

\citet{Zhang2021} & 2021 & BPF 1--40 Hz & Blind segmentation, z-score normalisation & CNN & SVM & PTB\\CEBSDB\\MIT NSR & 234\\20\\18 & No\\No\\No & IDR & 98.8\%\\99.7\%\\93.6\% 
\tabularnewline\hline

\citet{Huang2022} & 2022 & DWT hard-thresholding denoising & Pan-Tompkins, heartbeat segmentation, normalisation & AC/DCT and Wavelet LBP histogram matrix factorisation & Euclidean distance matching & MIT Arrh.\\\hspace{0.001cm}\\PTB\\\hspace{0.001cm}\\CYBHi\\\hspace{0.001cm}\\UofTDB & 47\\\hspace{0.001cm}\\248\\\hspace{0.001cm}\\63\\\hspace{0.001cm}\\46 & No\\\hspace{0.001cm}\\No\\\hspace{0.001cm}\\Yes\\\hspace{0.001cm}\\Yes & IDR\\EER\\IDR\\EER\\IDR\\EER\\IDR\\EER & 95.7\%\\0.36\%\\86.3\%\\2.26\%\\88.5\%\\1.48\%\\87.4\%\\2.36\%  
\tabularnewline\hline

\citet{Kim2022} & 2022 & - & Pan-Tompkins, beat segmentation & Segmented heartbeats & AlexNet CNN with heart-rate based data augmentation & ECG-ID & 83 & No & IDR & 95.7\%
\tabularnewline\hline

\citet{Lee2022} & 2022 & - & - & ICA convolutional network with 2D scalogram representations & SVM & Private\\MIT Arrh. & 95\\47 & No\\No & IDR & 98.5\%\\96.0\% 
\tabularnewline\hline

\citet{Li2022} & 2022 & DWT adaptive soft-thresholding & Second-order diff. R-peak detection, beat segmentation & Uniform Manifold Approximation and Projection (UMAP), Stacked Extremely Randomised Trees (ETs) & XGBoost & ECG-ID\\PTB & 89\\71 & No\\No & IDR & 98.9\%\\95.8\% 
\tabularnewline\hline

\end{longtable}
\end{spacing}}
\end{landscape}
}



%
    \renewcommand{\bibname}{References}%
    \cleardoublepage%
    \phantomsection%
    \addcontentsline{toc}{part}{References}%
    \begin{singlespace}
\end{singlespace}


\end{document}